\pdfoutput=1
\PassOptionsToPackage{prologue,dvipsnames}{xcolor}

\documentclass[10pt]{article}
\usepackage{bedstyle}
\usepackage{latexsym}
\usepackage{CJKutf8}  % loaded in main.tex outside the shared preamble

\usepackage{multirow}
\usepackage{multicol} % balanced two-column prose above appendix samples

\usepackage{listings}
\usepackage{tabularx}
\usepackage{longtable}
\usepackage{algorithm}
\usepackage{algpseudocode}
\usepackage{graphicx}
\usepackage{colortbl}
\usepackage[dvipsnames]{xcolor}
\usepackage[table]{xcolor} 
\usepackage{fontawesome5}
\usepackage{pythonhighlight}
\usepackage{soul} % \hl{} line-wrapping text highlight, used for manual trace annotations
\usepackage{tikz}

\usepackage{supertabular}
\usepackage{xltabular}
\usepackage{amsmath}
\usepackage{amssymb}
\usepackage{xspace}
\usepackage{mathrsfs}
\usepackage{bm,dutchcal}
\usepackage{float}
\usepackage{cuted}   % full-width non-floating blocks in two-column mode (Appendix D)
\usepackage{capt-of} % \captionof outside float environments
\usepackage{placeins}
\usepackage{gensymb}
\usepackage{tcolorbox}
\tcbuselibrary{breakable, skins, listings, raster}

\usepackage{marvosym}

\usepackage{adjustbox}
\usepackage{lineno}
\usepackage{array}
\usepackage{booktabs} % DAShlines for tables
\usepackage{enumitem}
\usepackage{makecell}
\usepackage[normalem]{ulem}

\usepackage{hyperref}
\usepackage[acronym]{glossaries}
\definecolor{custom_purple}{RGB}{180,114,200}
\definecolor{custom_pink}{RGB}{230,150,190}
\definecolor{custom_darkgreen}{rgb}{0.0, 0.5, 0.0}

\definecolor{gtred}{RGB}{220,20,60} 
\definecolor{m2mblue}{RGB}{0,102,204}

\usepackage[table]{xcolor}
\usepackage{svg}
\usepackage{worldflags}
\definecolor{softgrey}{gray}{0.85}

\definecolor{bedYellow}{HTML}{F2C94C}
\definecolor{bedCyan}{HTML}{2EC4D6}
\definecolor{bedGreen}{HTML}{2EAD4B}
\definecolor{bedBlue}{HTML}{3772FF}
\definecolor{bedPurple}{HTML}{7B2CBF}
\definecolor{bedOrange}{HTML}{F77F00}
\definecolor{bedOrchid}{HTML}{DA70D6}
\definecolor{bedTeal}{HTML}{008080}
\definecolor{bedSlategray}{HTML}{708090}
\definecolor{bedGold}{HTML}{D4A017}
\definecolor{bedMagenta}{HTML}{D336B8}
\definecolor{bedRed}{HTML}{D62828}

\definecolor{DescriptReform_Color}{RGB}{230, 102, 140}

\definecolor{bg}{RGB}{248,248,255}
\definecolor{frame}{RGB}{80,80,180}
\definecolor{titlebg}{RGB}{60,60,150}
\definecolor{string}{RGB}{196,26,22}
\definecolor{keyword}{RGB}{0,102,204}
\definecolor{comment}{RGB}{0,140,0}
\definecolor{number}{RGB}{160,32,240}

\lstdefinestyle{mypython}{
    basicstyle=\ttfamily\footnotesize,
    keywordstyle=\color{keyword}\bfseries,
    stringstyle=\color{string},
    commentstyle=\color{comment}\itshape,
    numberstyle=\tiny\color{gray},
    numbers=left,
    stepnumber=1,
    numbersep=8pt,
    showstringspaces=false,
    breaklines=true,
    frame=none,
    columns=fullflexible
}

\newcommand{\tickok}{\,{\color{green!55!black}\ensuremath{\checkmark}}}
\newcommand{\tickno}{\,{\color{red!75!black}\ensuremath{\boldsymbol{\times}}}}
\newcommand{\tickpart}{\,{\textcolor{orange}{$\triangle$}}}

\newcommand{\tone}{\hyperref[sec:app_dataset_t1]{T1}}
\newcommand{\ttwo}{\hyperref[sec:app_dataset_t2]{T2}}
\newcommand{\tthree}{\hyperref[sec:app_dataset_t3]{T3}}
\newcommand{\tfour}{\hyperref[sec:app_dataset_t4]{T4}}
\newcommand{\tfive}{\hyperref[sec:app_dataset_t5]{T5}}
\newcommand{\tsix}{\hyperref[sec:app_dataset_t6]{T6}}
\newcommand{\tseven}{\hyperref[sec:app_dataset_t7]{T7}}
\newcommand{\teight}{\hyperref[sec:app_dataset_t8]{T8}}
\newcommand{\tnine}{\hyperref[sec:app_dataset_t9]{T9}}
\newcommand{\tten}{\hyperref[sec:app_dataset_t10]{T10}}
\newcommand{\televen}{\hyperref[sec:app_dataset_t11]{T11}}
\newcommand{\ttwelve}{\hyperref[sec:app_dataset_t12]{T12}}
\newcommand{\tthirteen}{\hyperref[sec:app_dataset_t13]{T13}}

\newcommand{\tonelong}{\hyperref[sec:app_dataset_t1]{T1: Volume}}
\newcommand{\ttwolong}{\hyperref[sec:app_dataset_t2]{T2: Packaging}}
\newcommand{\tthreelong}{\hyperref[sec:app_dataset_t3]{T3: Difference}}
\newcommand{\tfourlong}{\hyperref[sec:app_dataset_t4]{T4: Color Count}}
\newcommand{\tfivelong}{\hyperref[sec:app_dataset_t5]{T5: Color Distribution}}
\newcommand{\tsixlong}{\hyperref[sec:app_dataset_t6]{T6: Local Structure}}
\newcommand{\tsevenlong}{\hyperref[sec:app_dataset_t7]{T7: Color Dominance}}
\newcommand{\teightlong}{\hyperref[sec:app_dataset_t8]{T8: Silhouette Symmetry}}
\newcommand{\tninelong}{\hyperref[sec:app_dataset_t9]{T9: Visible Floors}}
\newcommand{\ttenlong}{\hyperref[sec:app_dataset_t10]{T10: Color Connected}}
\newcommand{\televenlong}{\hyperref[sec:app_dataset_t11]{T11: Projection}}
\newcommand{\ttwelvelong}{\hyperref[sec:app_dataset_t12]{T12: Reconstruct}}
\newcommand{\tthirteenlong}{\hyperref[sec:app_dataset_t13]{T13: Mixture-of-Shapes}}

\providecommand{\aclfloat}[1]{\input{#1}}
\providecommand{\arxivfloat}[1]{}

\definecolor{task_overview_blue}{HTML}{DAE8FC}
\definecolor{task_overview_orange}{HTML}{FFE6CC}
\definecolor{task_overview_yellow}{HTML}{FFF2CC}
\definecolor{task_overview_purple}{HTML}{E1D5E7}
\definecolor{task_overview_red}{HTML}{F8CECC}

\usepackage{wrapfig}
\makeatletter
\AtBeginDocument{\let\bed@hyperref\hyperref \def\hyperref{\leavevmode\bed@hyperref}}
\makeatother

\renewcommand{\aclfloat}[1]{}
\renewcommand{\arxivfloat}[1]{\input{#1}}

\makeatletter
\newcommand{\arxivhere}[1]{\par\begingroup
  \@namedef{figure*}{\@ifnextchar[\bed@here@{\bed@here@[]}}%
  \def\bed@here@[##1]{\par\vspace{\intextsep}\noindent\begin{minipage}{\textwidth}%
    \captionsetup{type=figure}}%
  \@namedef{endfigure*}{\end{minipage}\par\vspace{\intextsep}}%
  \input{#1}\endgroup\par}
\makeatother

\newcommand\khai[1]{}
\newcommand\hai[1]{}

\renewcommand{\twocolumn}[1][]{\clearpage#1}

\newcommand{\bedcofirst}{\text{\textcolor{custom_purple}{\faStar}}}

\DeclareRobustCommand{\BedBenchTitle}{%
  \begingroup\normalfont
  \raisebox{-0.5em}{%
    \includegraphics[height=1.9em]{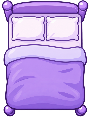}%
  }%
  \kern 0.0em % reduced spacing for tighter look
  \,%
  \endgroup
}

\DeclareRobustCommand{\BedBenchLogo}{%
  \begingroup\normalfont
  \raisebox{-0.2em}{%
    \includegraphics[height=1.2em]{images/BedBench_icon.png}%
  }%
  \kern 0.0em % reduced spacing for tighter look
  \textbf{\textcolor{custom_purple}{BedBench}}%
  \endgroup
}

\begin{document}

\bedtitleblock
  {\BedBenchTitle Where Do Vision Language Models Fail at 3D Reasoning?}
  {\textbf{Khai Le-Duc}$^{1,2,3\,\bedcofirst}$,
   \textbf{Nguyen Kim Hai Bui}$^{1,4\,\bedcofirst}$,
   \textbf{An Ngo}$^{1,5}$,
   \textbf{Nam D. Hoang}$^{\dagger}$,\\
   \textbf{Yueqi Song}$^{6}$,
   \textbf{Jiarui Liu}$^{6}$,
   \textbf{Thai Quoc Hoang}$^{7}$,
   \textbf{Fang Wu}$^{8}$,
   \textbf{Chris Ngo}$^{1}$,
   \textbf{Terry Jingchen Zhang}$^{3,9}$,\\
   \textbf{Mona T. Diab}$^{6}$,
   \textbf{Ran Xu}$^{7}$,
   \textbf{Silvio Savarese}$^{7,8}$,
   \textbf{Xiao-Ming Wu}$^{4}$}
  {
    % Affiliations as one logo strip, built by scripts/make_institution_marks.py (as in arXiv:2609.06396).
    \includegraphics[width=0.9\linewidth]{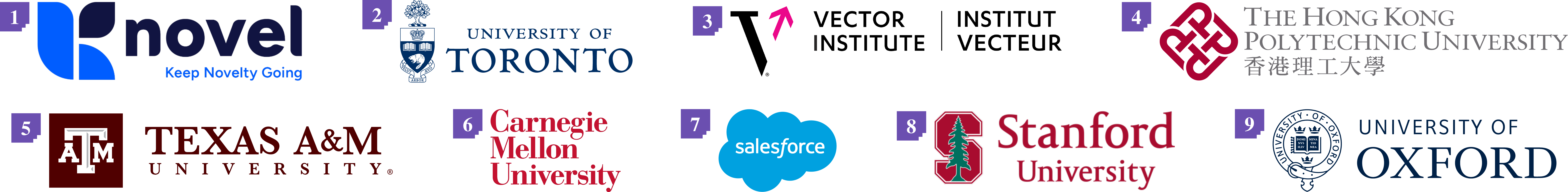}\\[0.6em]
   $^{\bedcofirst}$Co-first authors\quad
   $^{\dagger}$Independent Researcher\\[0.2em]
   % \quad
   % $^{\ddagger}$Work done partly during an internship at Knovel\\[0.2em]
   \textcolor{custom_purple}{\faEnvelope}\,\texttt{duckhai.le@mail.utoronto.ca}\quad
   \textcolor{custom_purple}{\faEnvelope}\,\texttt{nguyen-kim-hai.bui@connect.polyu.hk}}
  {% Vision-language models can get individual geometric judgments right while struggling to recover and consistently use the exact properties of a 3D object.
% To characterize this gap, we introduce BedBench, a controlled benchmark of thirteen procedurally generated tasks spanning geometric judgments and exact structural recovery.
Vision-language models can get individual judgments right while struggling to recover and consistently use the exact properties of a 3D object.
To characterize this gap, we introduce \BedBenchLogo, a controlled benchmark of thirteen procedurally generated tasks spanning individual judgments and composite answers.
Across twelve open-weight and four frontier models, per-color counting exposes a pronounced weakness: humans achieve 93.3\% accuracy, compared with 7.2\% for the best model.
Targeted fine-tuning raises one model's accuracy from 3.0\% to 68.6\%, but reveals a further difficulty: its answers about the most common colors remain inconsistent with its own per-color counts.
Supplying those counts as text improves its judgments, showing that recovering a property does not mean using it consistently.
% To examine why these difficulties persist, we test reasoning, scaling, and visual interventions, finding selective gains but unreliable exact answers.
% Probing Qwen3.5-27B shows that exact per-color counts remain recoverable at the language model's input, yet the model's answers rarely recover them (65.0\% for the probe vs. 3.0\% for the model).
Reasoning, scaling, and added visual cues each improve individual judgments, but none makes exact answers reliable.
To locate the difficulty, we probe Qwen3.5-27B: exact per-color counts remain recoverable at the language model's input, yet the model's answers rarely recover them (65.0\% for the probe vs. 3.0\% for the model).
Image-removal controls further show that some apparent successes require no visual evidence.
These findings motivate evaluating 3D reasoning by exact correctness, consistent use, and image dependence, rather than accuracy alone.
% These findings motivate evaluating 3D reasoning by exact recovery, consistent use, and image dependence, rather than accuracy alone.
% Across open-weight and frontier models, we {systematically characterize an approximate-to-exact failure boundary}: models often recover useful geometric quantities and cues while remaining unreliable on 
% jointly exact answers about task-relevant object properties.
% Across open-weight and frontier models, we identify a two-level reliability gap: fine-grained constituent estimates are often inaccurate, and improvements in these estimates do not necessarily yield jointly exact answers about object properties.
% reliable 3D structural recovery.
% Code and data are available at \url{https://anonymous.4open.science/r/BED} and \url{https://huggingface.co/datasets/BEDBench/BED}.

}
  % TODO: fill in the public URLs (empty URL = pill without a link).
  {\bedlink{\faGithub}{leduckhai/BedBench}{https://github.com/leduckhai/BedBench}\quad
   \bedlink{\faGlobe}{Project page}{https://leduckhai.github.io/BedBench}\quad
   \bedlink{\faDatabase}{Dataset}{https://huggingface.co/datasets/leduckhai/BedBench}}

\addtocontents{toc}{\protect\setcounter{tocdepth}{-1}}

% Figure 1 (task overview) right after the abstract; ACL: inside the introduction.
\begin{figure*}
    \centering
    \scriptsize
\begin{tikzpicture}
  \node[anchor=south west, inner sep=0] (img) at (0,0) {%
    \includegraphics[width=\linewidth]{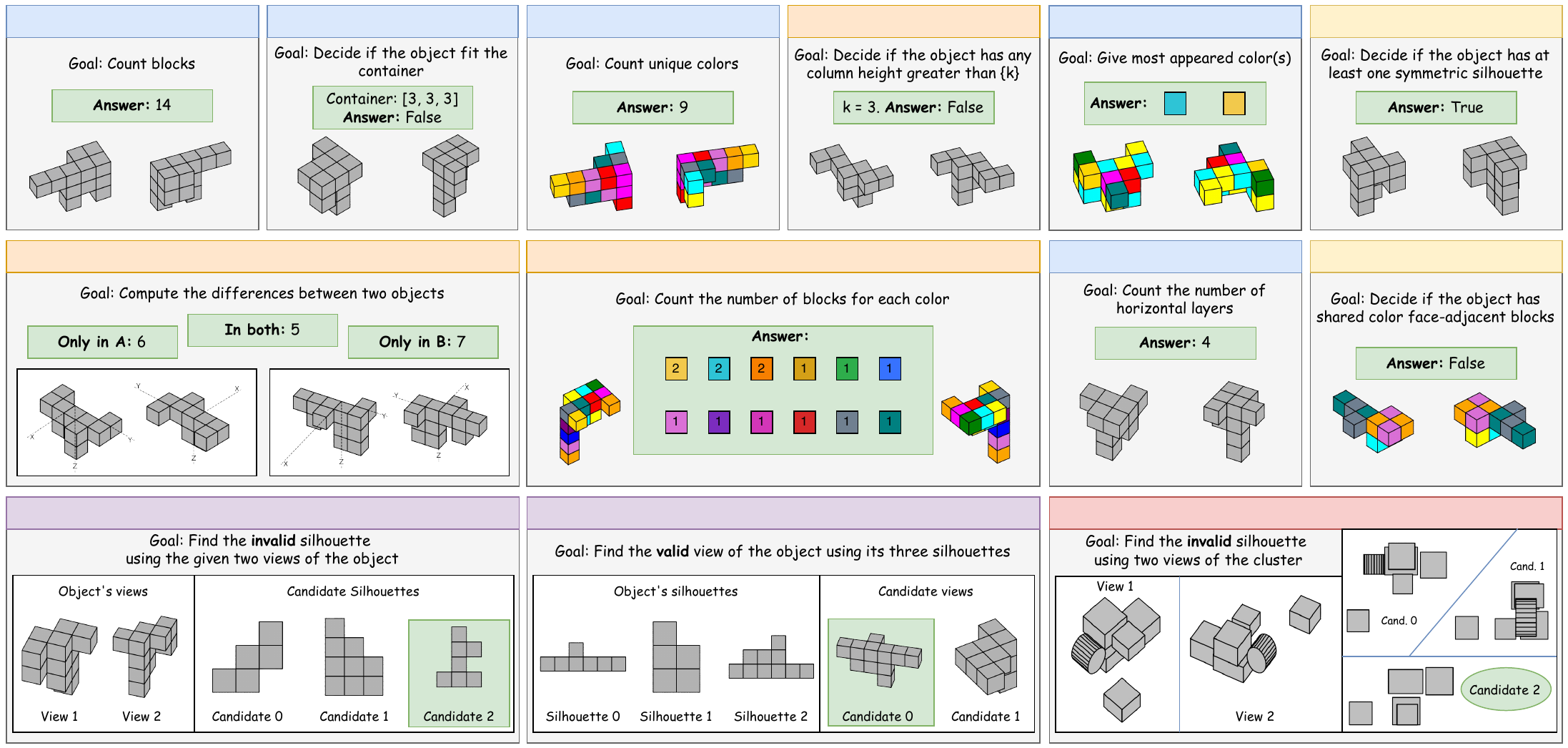}%
  };

  % Normalized coordinates: (0,0)=bottom-left, (1,1)=top-right
  \begin{scope}[x={(img.south east)}, y={(img.north west)}]
    \node[anchor=center] at (0.0842,0.9710) {\tonelong{}};
    \node[anchor=center] at (0.2508,0.9650) {\ttwolong{}};
    \node[anchor=center] at (0.4154,0.9710) {\tfourlong{}};
    \node[anchor=center] at (0.5819,0.9710) {\tsixlong{}};
    \node[anchor=center] at (0.7501,0.9710) {\tsevenlong{}};
    \node[anchor=center] at (0.9150,0.9650) {\teightlong{}};
    \node[anchor=center] at (0.1665,0.6570) {\tthreelong{}};
    \node[anchor=center] at (0.5003,0.6570) {\tfivelong{}};
    \node[anchor=center] at (0.7518,0.6570) {\tninelong{}};
    \node[anchor=center] at (0.9165,0.6570) {\ttenlong{}};
    \node[anchor=center] at (0.1662,0.3100) {\televenlong{}};
    \node[anchor=center] at (0.5013,0.3150) {\ttwelvelong{}};
    \node[anchor=center] at (0.8328,0.3100) {\tthirteenlong{}};
  \end{scope}
\end{tikzpicture}
    \caption{\BedBenchLogo \textbf{ tasks}, grouped by primary spatial demand: \sethlcolor{task_overview_blue}\hl{Aggregate quantity and support}, \sethlcolor{task_overview_orange}\hl{Compositional decomposition}, \sethlcolor{task_overview_yellow}\hl{Relational structure}, \sethlcolor{task_overview_purple}\hl{Cross-view geometry}, \sethlcolor{task_overview_red}\hl{Multi-object scene reasoning}. More in Appendix~\ref{sec:app_dataset_examples}. \textbf{Takeaway:} The tasks span counting, size checks, per-color counts, part relations, and cross-view matching.}
    \label{fig:task_overview}
\end{figure*}

\section{Introduction}
\label{sec:intro}
\begin{figure*}
    \centering
    \scriptsize
\begin{tikzpicture}
  \node[anchor=south west, inner sep=0] (img) at (0,0) {%
    \includegraphics[width=\linewidth]{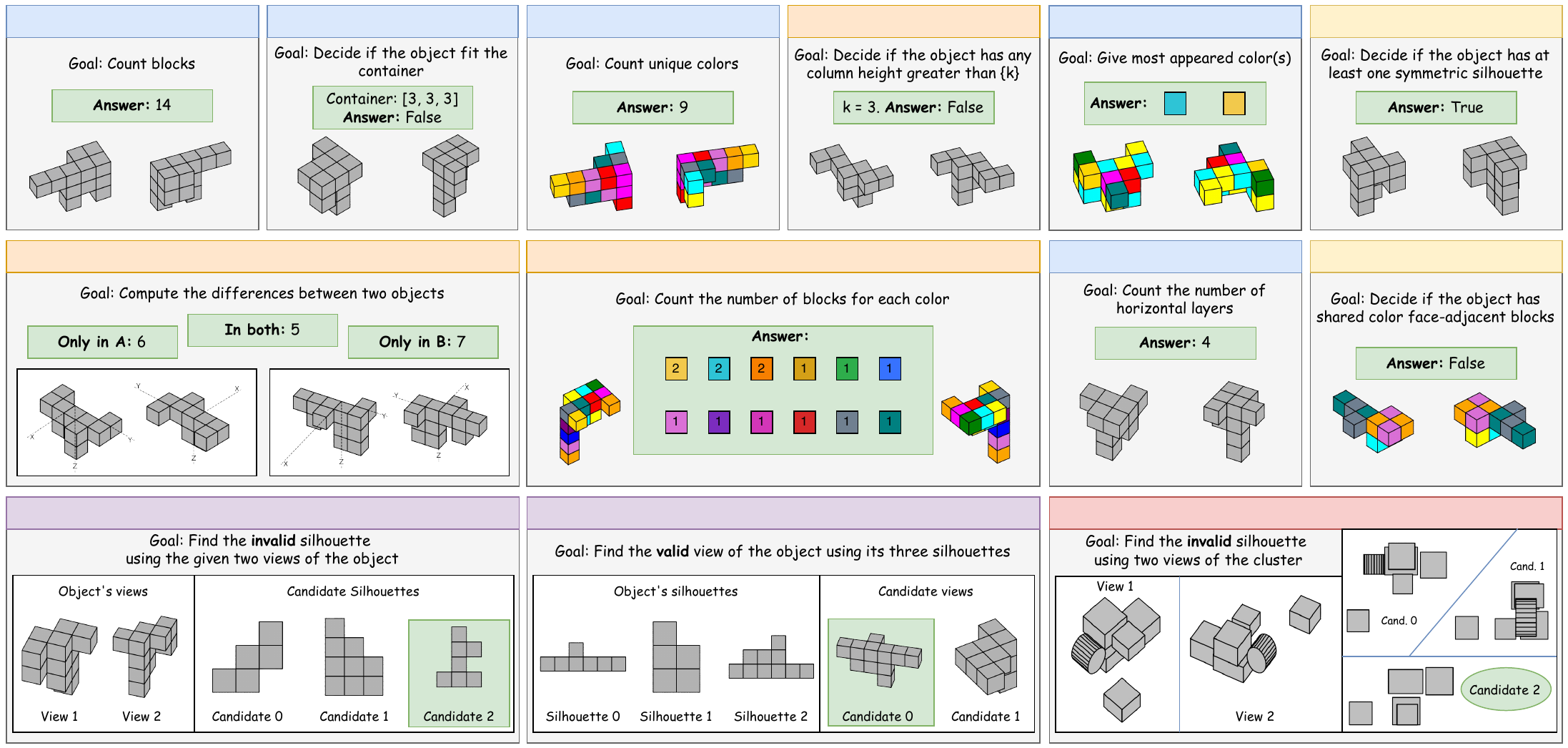}%
  };

  % Normalized coordinates: (0,0)=bottom-left, (1,1)=top-right
  \begin{scope}[x={(img.south east)}, y={(img.north west)}]
    \node[anchor=center] at (0.0842,0.9710) {\tonelong{}};
    \node[anchor=center] at (0.2508,0.9650) {\ttwolong{}};
    \node[anchor=center] at (0.4154,0.9710) {\tfourlong{}};
    \node[anchor=center] at (0.5819,0.9710) {\tsixlong{}};
    \node[anchor=center] at (0.7501,0.9710) {\tsevenlong{}};
    \node[anchor=center] at (0.9150,0.9650) {\teightlong{}};
    \node[anchor=center] at (0.1665,0.6570) {\tthreelong{}};
    \node[anchor=center] at (0.5003,0.6570) {\tfivelong{}};
    \node[anchor=center] at (0.7518,0.6570) {\tninelong{}};
    \node[anchor=center] at (0.9165,0.6570) {\ttenlong{}};
    \node[anchor=center] at (0.1662,0.3100) {\televenlong{}};
    \node[anchor=center] at (0.5013,0.3150) {\ttwelvelong{}};
    \node[anchor=center] at (0.8328,0.3100) {\tthirteenlong{}};
  \end{scope}
\end{tikzpicture}
    \caption{\BedBenchLogo \textbf{ tasks}, grouped by primary spatial demand: \sethlcolor{task_overview_blue}\hl{Aggregate quantity and support}, \sethlcolor{task_overview_orange}\hl{Compositional decomposition}, \sethlcolor{task_overview_yellow}\hl{Relational structure}, \sethlcolor{task_overview_purple}\hl{Cross-view geometry}, \sethlcolor{task_overview_red}\hl{Multi-object scene reasoning}. More in Appendix~\ref{sec:app_dataset_examples}. \textbf{Takeaway:} The tasks span counting, size checks, per-color counts, part relations, and cross-view matching.}
    \label{fig:task_overview}
\end{figure*}

Reasoning about a 3D object from images requires relating its visible parts across views: a model may need to count blocks, identify which parts share a color, or predict the object's appearance from another direction.
Vision-language models (VLMs) are increasingly evaluated on such spatial and 3D reasoning tasks \cite{liu2023visualspatialreasoning,ma20253dsrbenchcomprehensive3dspatial,xu2026spatialbenchbenchmarkingmultimodallarge,vsibench}, yet their scores do not always reveal how they reached an answer.
A correct answer may depend on visual evidence, but it may also follow from language priors or cues in multiple-choice options \cite{goyal2017making,agrawal2018dont,zheng2024large,balepur2024artifacts}.
Conversely, a low score may reflect a model limitation, ambiguity in the available views, or a task that is difficult for humans as well.

% Existing benchmarks cover diverse spatial skills, and several include human baselines, procedural generation, or comparisons between answer formats (Table~\ref{tab:benchmark_comparison}).
% Building on these works, we ask a more specific question: \textbf{does success on individual geometric judgments extend to exact recovery and consistent use of object properties?}
Building on these works, we ask a more specific question: \textbf{does success on individual judgments extend to exact answers and consistent use of object properties?}
Consider a model that can count the distinct colors (\tfourlong{}) and estimate the total number of blocks (\tonelong{}) in an object.
% These individual judgments do not establish that it can report the correct number of blocks of every color (\tfivelong{}).
% Recovering this full distribution requires associating every color with its correct count, illustrating what we mean by \emph{exact structural recovery}: the task-relevant quantities, attributes, and their associations must all be correct together.
Each is an \textbf{\emph{individual judgment}}: a single count, comparison, or decision about one property of the object.
Such judgments do not establish that the model can report the correct number of blocks of every color (\tfivelong{}).
That requires an \textbf{\emph{exact answer}}: a structured answer that is correct only when all its quantities, attributes, and their associations are correct together, here every color paired with its correct count.
Consistent use adds a further requirement: when asked which colors are most common, the model's answer should agree with those counts.

We introduce \BedBenchLogo\,(short for \textcolor{custom_purple}{\textbf{B}}lind \textcolor{custom_purple}{\textbf{3D}} \textcolor{custom_purple}{\textbf{Bench}}mark), a benchmark of thirteen procedurally generated tasks using voxel objects (i.e., 3D shapes built from unit cubes) and scenes of geometric primitives.
The tasks range from counting blocks (\tone{}) and checking whether an object fits inside a container (\ttwolong{}) to reporting per-color counts (\tfive{}) and matching an object to its 2D projections (\televenlong{}; Figure~\ref{fig:task_overview}).
% The generator provides exact ground truth and filters out samples detected as ambiguous at generation time, so partial geometric competence and exact structural recovery can be compared within one controlled task family.
The generator provides exact ground truth and filters out samples detected as ambiguous at generation time, so individual judgments and exact answers can be compared within one controlled task family.
Existing benchmarks include some of these properties (Table~\ref{tab:benchmark_comparison}); BedBench combines exact-answer scoring, consistency checks, image-dependence controls, and frozen-feature probes on the same objects, so it can also locate where a failure arises.
% We release the generator, data, model outputs, and code.

We evaluate twelve open-weight (OW) and four frontier VLMs, with 30 human participants as a reference for task difficulty.
% We first examine where exact answers fail and whether simpler queries, a 2D presentation, or task-specific training improve them.
We first compare models with humans across tasks to find which exact answers fail most clearly, and test whether simpler queries, a 2D presentation, or task-specific training improve them.
Since more reasoning, larger models, and added visual cues are the usual remedies for spatial failures, we then test each of them and probe the frozen visual features to locate where the difficulty arises.
Finally, because a high score is meaningful only if it depends on the image, we check whether successful answers require the images.
This yields four findings:
\begin{enumerate}[leftmargin=1.4em,itemsep=1pt,topsep=2pt]
  \item \textbf{Exact counting is unreliable, and learning it does not ensure consistent use.}
  Humans reach \textcolor{custom_darkgreen}{93.3\%} on \tfive{}, the best VLM \textcolor{red!70!black}{7.2\%}. Fine-tuning raises \tfive{} to 68.6\%, yet the model identifies the most common colors (\tseven{}) only 35.0\% of the time from images, versus 74.2\% when given its own counts as text (\S\ref{sec:3d_blind}).
  \item \textbf{Reasoning helps selected tasks but not exact answers.}
    Thinking raises \tfour{} by \textcolor{custom_darkgreen}{$+$13.9\,pp} on average, yet \tfive{} stays at \textcolor{red!70!black}{0.7--1.0\%}, larger budgets lower \tone{} and \televen{}, and visual misreadings often persist through the reasoning (\S\ref{sec:thinking_paradox}).

  % Larger reasoning budgets improve \tfour{} but lower \tone{} and \televen{}, and visual misreadings often persist through the reasoning (\S\ref{sec:thinking_paradox}).
  \item \textbf{Scaling and visual support leave exact answers unreliable, although visual features hold the counts.}
  Gains from scale and visual cues are selective. A probe at Qwen3.5-27B's language-model input recovers exact counts on \textcolor{custom_darkgreen}{65.0\%} of objects, against \textcolor{red!70!black}{3.0\%} for the VLM, pointing to difficulty in using available visual information (\S\ref{sec:scaling}).
  \item \textbf{Accuracy alone does not establish visual reasoning.}
  Without images, models keep some scores through constant answers, and in a fixed-layout MCQ one model reaches 100\% by always choosing the third option (\S\ref{sec:mcq}).
\end{enumerate}
Overall, models get individual judgments right while the exact answer stays wrong, and recovering a property does not mean using it consistently.

\section{Related Work}
\label{sec:related}
% arXiv version: Table 1 sits beside the start of Related Work (ACL: end of the section).
% arXiv version of tables/benchmark_comparison.tex (same content), wrapfig, as in arXiv:2608.07645 (simulation.tex); spans paragraphs.
\begin{wraptable}[39]{r}{7.7cm}
\centering
\scriptsize
\setlength{\tabcolsep}{1.6pt}
\renewcommand{\arraystretch}{1.05}
\renewcommand{\rothead}[1]{\rotatebox{90}{#1\,}}
\begin{tabular}{@{}>{\raggedright\arraybackslash}p{3.95cm} ccccc @{\hspace{4pt}} cccc@{}}
\toprule
\textbf{Dataset}
& \rothead{3D}
& \rothead{Pure}
& \rothead{Procedural}
& \rothead{Infinite}
& \rothead{OE}
& \rothead{Linked}
& \rothead{No-image}
& \rothead{Format}
& \rothead{Probe} \\
\midrule
\multicolumn{10}{@{}l}{\textit{Breadth-first aggregators}} \\
\hspace{3pt}SITE \citep{site_spatial}                 & \tickpart & \tickno & \tickno & \tickno & \tickno & \tickpart & \tickno & \tickno & \tickno \\
\hspace{3pt}EASI \citep{easi_holistic}                & \tickpart & \tickno & \tickno & \tickno & \tickpart & \tickno & \tickno & \tickno & \tickno \\
\hspace{3pt}OmniSpatial \citep{omnispatial}           & \tickpart & \tickno & \tickno & \tickno & \tickno & \tickno & \tickno & \tickno & \tickno \\
\hspace{3pt}Spatial-DISE \citep{spatialdise}          & \tickok & \tickpart & \tickpart & \tickno & \tickno & \tickno & \tickno & \tickno & \tickno \\
\hspace{3pt}SRBench \citep{mindthegap}                & \tickpart & \tickpart & \tickpart & \tickno & \tickno & \tickno & \tickno & \tickno & \tickno \\
\hspace{3pt}SpatiaLab \citep{wasi2026spatialab}       & \tickpart & \tickno & \tickno & \tickno & \tickok & \tickno & \tickno & \tickok & \tickno \\
\hspace{3pt}MindEdit-Bench \citep{yu2026mindedit}     & \tickpart & \tickno & \tickno & \tickno & \tickno & \tickno & \tickno & \tickno & \tickno \\
\addlinespace[1pt]
\multicolumn{10}{@{}l}{\textit{Cognitively grounded suites}} \\
\hspace{3pt}SPACE \citep{space_cognition}             & \tickok & \tickno & \tickok & \tickno & \tickno & \tickpart & \tickpart & \tickno & \tickno \\
\hspace{3pt}11Plus-Bench \citep{elevenplusbench}      & \tickpart & \tickok & \tickno & \tickno & \tickno & \tickno & \tickno & \tickno & \tickno \\
\hspace{3pt}Psychometric BSA \citep{psychometric_bsa} & \tickok & \tickok & \tickno & \tickno & \tickno & \tickno & \tickno & \tickno & \tickno \\
\hspace{3pt}SpatialViz-Bench \citep{wang2026spatialvizbenchcognitivelygroundedbenchmarkdiagnosing} & \tickok & \tickok & \tickok & \tickpart & \tickpart & \tickno & \tickpart & \tickno & \tickno \\
\hspace{3pt}STARE \citep{stare_unfolding}             & \tickpart & \tickno & \tickok & \tickno & \tickno & \tickpart & \tickpart & \tickno & \tickno \\
\addlinespace[1pt]
\multicolumn{10}{@{}l}{\textit{Geometry-native suites}} \\
\hspace{3pt}SPARE3D \citep{spare3d}                   & \tickok & \tickok & \tickok & \tickpart & \tickpart & \tickpart & \tickno & \tickno & \tickno \\
\hspace{3pt}LEGO-Puzzles \citep{legopuzzles}          & \tickok & \tickok & \tickpart & \tickno & \tickno & \tickpart & \tickno & \tickno & \tickno \\
\hspace{3pt}Inf-Bench \citep{infiniteladder}          & \tickpart & \tickok & \tickok & \tickok & \tickno & \tickpart & \tickpart & \tickno & \tickno \\
\addlinespace[1pt]
\multicolumn{10}{@{}l}{\textit{Procedural scene generators}} \\
\hspace{3pt}InfiniBench \citep{wang2025infinibench}   & \tickok & \tickno & \tickok & \tickok & \tickpart & \tickno & \tickno & \tickno & \tickno \\
\hspace{3pt}DynaSolidGeo \citep{wu2025dynasolidgeo}   & \tickpart & \tickok & \tickok & \tickok & \tickok & \tickno & \tickno & \tickno & \tickno \\
\hspace{3pt}IDEAL-Bench \citep{cai2026idealbench}     & \tickok & \tickno & \tickok & \tickpart & \tickpart & \tickno & \tickno & \tickno & \tickno \\
\hspace{3pt}SpatialEval \citep{spatialeval}         & \tickno & \tickpart & \tickpart & \tickpart & \tickno & \tickno & \tickok & \tickno & \tickno \\
\hspace{3pt}SpaMEM \citep{liao2026spamem}             & \tickok & \tickno & \tickok & \tickpart & \tickpart & \tickpart & \tickpart & \tickno & \tickno \\
\midrule
\textbf{\BedBenchLogo\,(ours)} & \tickok & \tickok & \tickok & \tickok & \tickok & \tickok & \tickok & \tickpart & \tickok \\
\bottomrule
\end{tabular}
\caption{\textbf{Comparison with representative spatial and 3D reasoning benchmarks.} \tickok{}~yes, \tickpart{}~partial, \tickno{}~no. \emph{3D}: volumetric stimuli throughout; \emph{Pure}: abstract geometric stimuli; \emph{Procedural}: programmatically generated items; \emph{Infinite}: generation not limited to a fixed source pool; \emph{OE}: open-ended answers; \emph{Linked}: questions on the same item whose answers entail one another, with their agreement evaluated (partial: linked questions without such an analysis); \emph{No-image}: image-removal controls (partial: image replaced by a text description); \emph{Format}: matched answer-format comparisons; \emph{Probe}: visual-feature probes against ground-truth targets. \textbf{Takeaway:} Several benchmarks share individual properties; \BedBenchLogo\,uniquely combines open-ended 3D questions, linked questions checked for agreement, image-removal controls, and probes of frozen visual features, which together locate where a failure arises (Section~\ref{sec:related}).
% \khai{The relationship to existing procedural 3D benchmarks should be sharper!
% The paper includes an extensive benchmark comparison table. However, several dimensions concern the presence of evaluation controls rather than fundamentally new reasoning tasks.
% The authors should articulate more precisely what scientific questions BedBench enables that \textbf{cannot} be answered using prior procedural benchmarks with additional diagnostics.}
}
\label{tab:benchmark_comparison}
\end{wraptable}

\paragraph{Benchmarks for spatial and 3D reasoning.}
Spatial benchmarks evaluate a broad range of skills, including scene understanding \cite{site_spatial,omnispatial,spatialdise}, mental rotation \cite{psychometric_bsa,wang2026spatialvizbenchcognitivelygroundedbenchmarkdiagnosing,stare_unfolding}, and geometric reasoning from drawings or assembled objects \cite{spare3d,legopuzzles}.
% Procedural benchmarks generate fresh instances to reduce reliance on memorized examples \cite{infiniteladder,wang2025infinibench,wu2025dynasolidgeo,cai2026idealbench,liao2026spamem}.
% BED builds on this approach by using the same underlying geometry for visual-cue comparisons and probe targets, alongside human evaluation, image removal, and an answer-format comparison.
% Table~\ref{tab:benchmark_comparison} summarizes the differences; Appendix~\ref{sec:extended_related_works} reviews individual benchmarks and their findings.
Procedural generation reduces memorization \cite{infiniteladder,wang2025infinibench,wu2025dynasolidgeo,cai2026idealbench,liao2026spamem}, but accuracy alone cannot show why a model fails. 
BedBench derives its tasks from the same generated objects, so their answers entail one another: the per-color counts of \tfive{} determine the number of distinct colors (\tfour{}) and the most common colors (\tseven{}).
Prior consistency checks for visual question answering measure whether answers to entailed questions agree \cite{hudson2019gqa,ribeiro2019redroses,ray2019sunny}; BedBench asks where a failure arises: whether a model fails to recover a property, to use it consistently, or to see it in 3D.
These questions require tasks whose answers entail each other on the same object and block-level ground truth for re-rendering, which adding diagnostics to an existing benchmark does not create.
Probing frozen features on the same targets, giving a model its own answer as text, and re-rendering the same blocks as 2D tiles answer them in turn.
Table~\ref{tab:benchmark_comparison} summarizes the design differences; Appendix~\ref{sec:extended_related_works} reviews individual benchmarks.

% \input{tables/task_overview}

% \paragraph{Related failure patterns.}
% Prior work has reported related failures in compositional reasoning, exact counting, exhaustive visual readout, and multi-view spatial reasoning
% \cite{li2025compositionalgap,counting_bottleneck,grid2matrix,threeviewsense}.
% Building on these findings, we examine what task accuracy establishes: whether constituent counting errors compound into failures of jointly exact recovery, whether task-specific training transfers to logically entailed questions, and whether apparent success depends on visual evidence.

\paragraph{Related failure patterns.}
Prior work reports failures in compositional reasoning, exact counting, reporting every element of an image, and multi-view spatial reasoning
\cite{li2025compositionalgap,counting_bottleneck,grid2matrix,threeviewsense}.
We extend these findings with diagnostics of exact answers, consistent use, and visual dependence.

% \paragraph{Shortcuts and evaluation controls.}
% Visual question answering models can exploit language priors, option positions, and other shortcuts \cite{goyal2017making,agrawal2018dont,zheng2024large,balepur2024artifacts,geirhos2020shortcut,lapuschkin2019unmasking}.
% These concerns also apply to spatial evaluation: some models score higher without images \cite{spatialeval}, and SpatiaLab reports lower model accuracy when matched questions require open-ended answers instead of MCQ \cite{wasi2026spatialab}.
% Image-removal controls have also been used to assess whether decoded spatial information contributes to model predictions \cite{liao2026decodable}.
% BED combines image removal with an open-ended/MCQ comparison on \tone{} to test whether an apparent gain requires visual input.

\paragraph{Visual representations and probes.}
Probing studies show that frozen visual features can encode information that a VLM fails to report, including counts and spatial properties \cite{rahmanzadehgervi2025visionlanguagemodelsblind,el2024probing,threeviewsense,grid2matrix,counting_bottleneck}.
% BED compares recovery of global quantities, per-part counts, and relations using targets from the same generator.
% Our pooled probes recover global count trends but rarely produce complete, correct per-part answers. This does not establish that the required information is absent from the patch features: prior work recovers binding information using quadratic probes or spatially selective readouts \cite{li2025objectbinding,michalkiewicz2026foveated}.
The choice of probe also matters: prior work recovers whether two patches belong to the same object, or which color goes with which shape, using quadratic probes or probes that attend to selected patches instead of averaging them \cite{li2025objectbinding,michalkiewicz2026foveated}.
Building on these studies, we test whether the behavioral gap between aggregate quantities and exact answers also appears in probes on averaged patch features, and whether a readout that attends to selected patches, trained on more objects, recovers exact per-color counts.
% We test whether the behavioral gap between approximate estimates and exact compositional answers also appears under the tested pooled readouts.

\paragraph{Reasoning, scaling, and training.}
Chain-of-Thought (CoT) prompting can improve reasoning, although its benefits depend on the task \cite{wei2022chain,kojima2022large,sprague2025cot,kancheti2026cot}.
Other work questions whether scaling alone is sufficient for spatial reasoning \cite{zhang2025scalingbeyond}, while targeted training, for example on geometry problems or with a staged spatial curriculum, has produced substantial gains \cite{euclids_gift,li2026spatialladder}.
% We use BED to compare the effects of reasoning budgets, model scale and architecture, visual cues, and task-specific fine-tuning, identifying which errors persist under the tested settings.
We test whether the gap between individual judgments and exact answers persists across reasoning budgets, model scale and architecture, visual cues, and task-specific fine-tuning.

\begin{table}[t]
\centering
\scriptsize
\setlength{\tabcolsep}{1.6pt}
\renewcommand{\arraystretch}{1.05}
\renewcommand{\rothead}[1]{\rotatebox{90}{#1\,}}
\begin{tabular}{@{}>{\raggedright\arraybackslash}p{3.95cm} ccccc @{\hspace{4pt}} cccc@{}}
\toprule
\textbf{Dataset}
& \rothead{3D}
& \rothead{Pure}
& \rothead{Procedural}
& \rothead{Infinite}
& \rothead{OE}
& \rothead{Linked}
& \rothead{No-image}
& \rothead{Format}
& \rothead{Probe} \\
\midrule
\multicolumn{10}{@{}l}{\textit{Breadth-first aggregators}} \\
\hspace{3pt}SITE \citep{site_spatial}                 & \tickpart & \tickno & \tickno & \tickno & \tickno & \tickpart & \tickno & \tickno & \tickno \\
\hspace{3pt}EASI \citep{easi_holistic}                & \tickpart & \tickno & \tickno & \tickno & \tickpart & \tickno & \tickno & \tickno & \tickno \\
\hspace{3pt}OmniSpatial \citep{omnispatial}           & \tickpart & \tickno & \tickno & \tickno & \tickno & \tickno & \tickno & \tickno & \tickno \\
\hspace{3pt}Spatial-DISE \citep{spatialdise}          & \tickok & \tickpart & \tickpart & \tickno & \tickno & \tickno & \tickno & \tickno & \tickno \\
\hspace{3pt}SRBench \citep{mindthegap}                & \tickpart & \tickpart & \tickpart & \tickno & \tickno & \tickno & \tickno & \tickno & \tickno \\
\hspace{3pt}SpatiaLab \citep{wasi2026spatialab}       & \tickpart & \tickno & \tickno & \tickno & \tickok & \tickno & \tickno & \tickok & \tickno \\
\hspace{3pt}MindEdit-Bench \citep{yu2026mindedit}     & \tickpart & \tickno & \tickno & \tickno & \tickno & \tickno & \tickno & \tickno & \tickno \\
\addlinespace[1pt]
\multicolumn{10}{@{}l}{\textit{Cognitively grounded suites}} \\
\hspace{3pt}SPACE \citep{space_cognition}             & \tickok & \tickno & \tickok & \tickno & \tickno & \tickpart & \tickpart & \tickno & \tickno \\
\hspace{3pt}11Plus-Bench \citep{elevenplusbench}      & \tickpart & \tickok & \tickno & \tickno & \tickno & \tickno & \tickno & \tickno & \tickno \\
\hspace{3pt}Psychometric BSA \citep{psychometric_bsa} & \tickok & \tickok & \tickno & \tickno & \tickno & \tickno & \tickno & \tickno & \tickno \\
\hspace{3pt}SpatialViz-Bench \citep{wang2026spatialvizbenchcognitivelygroundedbenchmarkdiagnosing} & \tickok & \tickok & \tickok & \tickpart & \tickpart & \tickno & \tickpart & \tickno & \tickno \\
\hspace{3pt}STARE \citep{stare_unfolding}             & \tickpart & \tickno & \tickok & \tickno & \tickno & \tickpart & \tickpart & \tickno & \tickno \\
\addlinespace[1pt]
\multicolumn{10}{@{}l}{\textit{Geometry-native suites}} \\
\hspace{3pt}SPARE3D \citep{spare3d}                   & \tickok & \tickok & \tickok & \tickpart & \tickpart & \tickpart & \tickno & \tickno & \tickno \\
\hspace{3pt}LEGO-Puzzles \citep{legopuzzles}          & \tickok & \tickok & \tickpart & \tickno & \tickno & \tickpart & \tickno & \tickno & \tickno \\
\hspace{3pt}Inf-Bench \citep{infiniteladder}          & \tickpart & \tickok & \tickok & \tickok & \tickno & \tickpart & \tickpart & \tickno & \tickno \\
\addlinespace[1pt]
\multicolumn{10}{@{}l}{\textit{Procedural scene generators}} \\
\hspace{3pt}InfiniBench \citep{wang2025infinibench}   & \tickok & \tickno & \tickok & \tickok & \tickpart & \tickno & \tickno & \tickno & \tickno \\
\hspace{3pt}DynaSolidGeo \citep{wu2025dynasolidgeo}   & \tickpart & \tickok & \tickok & \tickok & \tickok & \tickno & \tickno & \tickno & \tickno \\
\hspace{3pt}IDEAL-Bench \citep{cai2026idealbench}     & \tickok & \tickno & \tickok & \tickpart & \tickpart & \tickno & \tickno & \tickno & \tickno \\
\hspace{3pt}SpatialEval \citep{spatialeval}         & \tickno & \tickpart & \tickpart & \tickpart & \tickno & \tickno & \tickok & \tickno & \tickno \\
\hspace{3pt}SpaMEM \citep{liao2026spamem}             & \tickok & \tickno & \tickok & \tickpart & \tickpart & \tickpart & \tickpart & \tickno & \tickno \\
\midrule
\textbf{BedBench (ours)} & \tickok & \tickok & \tickok & \tickok & \tickok & \tickok & \tickok & \tickpart & \tickok \\
\bottomrule
\end{tabular}
\caption{\textbf{Comparison with representative spatial and 3D reasoning benchmarks.} \tickok{}~yes, \tickpart{}~partial, \tickno{}~no. \emph{3D}: volumetric stimuli throughout; \emph{Pure}: abstract geometric stimuli; \emph{Procedural}: programmatically generated items; \emph{Infinite}: generation not limited to a fixed source pool; \emph{OE}: open-ended answers; \emph{Linked}: questions on the same item whose answers entail one another, with their agreement evaluated (partial: linked questions without such an analysis); \emph{No-image}: image-removal controls (partial: image replaced by a text description); \emph{Format}: matched answer-format comparisons; \emph{Probe}: visual-feature probes against ground-truth targets. \textbf{Takeaway:} Several benchmarks share individual properties; BedBench uniquely combines open-ended 3D questions, linked questions checked for agreement, image-removal controls, and probes of frozen visual features, which together locate where a failure arises (Section~\ref{sec:related}).
% \khai{The relationship to existing procedural 3D benchmarks should be sharper!
% The paper includes an extensive benchmark comparison table. However, several dimensions concern the presence of evaluation controls rather than fundamentally new reasoning tasks.
% The authors should articulate more precisely what scientific questions BedBench enables that \textbf{cannot} be answered using prior procedural benchmarks with additional diagnostics.}
}
\label{tab:benchmark_comparison}
\end{table}

\section{The BedBench Benchmark}
\label{sec:benchmark}

% BED consists of thirteen tasks in which a model answers a question about a 3D object or scene from rendered images.
% For example, \tonelong{} presents two views of a voxel object and asks for its total block count, while \tfivelong{} asks for the count of each color.
% The underlying geometry supplies the ground-truth answers, allowing us to compare different questions and visual conditions without changing the object.

\paragraph{Task coverage.}
% Table~\ref{tab:task_overview} groups the tasks by the spatial information they require: aggregate quantity and support, compositional decomposition, relational structure, cross-view geometry, and multi-object scene reasoning.
% BedBench compares different questions about procedurally generated 3D objects and scenes, allowing us to examine both individual judgments and complete structured answers.
BedBench compares different questions about procedurally generated 3D objects and scenes, allowing us to examine both individual judgments and exact answers on the same objects.
Figure~\ref{fig:task_overview} groups the thirteen tasks by spatial demand and shows example inputs and ground-truth answers (more details in Appendix~\ref{sec:app_dataset_examples}).
% The outputs include counts, structured lists, binary decisions, and candidate selections.
% Cross-view tasks test the ability to relate different depictions of the same object, a difficulty also studied in prior work \cite{mindcube}.
% In \televenlong{} and \tthirteenlong{}, models select a valid or invalid orthographic silhouette from three candidates.
% \ttwelvelong{} reverses the direction: three silhouettes define an object, and the model selects its matching view from two candidates.

\paragraph{Task selection.}
We chose tasks whose answers are meant to be easier to check but harder to produce.
Producing answers requires imagining the object(s) from the given inputs and then mentally rotating, decomposing, comparing, or querying.

\paragraph{Generation and ambiguity checks.}
Each object has 10--15 connected unit blocks, rendered from two opposite views.
We reject objects whose two silhouettes match a sampled alternative with different orthographic projections, or that contain a block whose removal barely changes both silhouettes (IoU $\geq 0.9998$).
An independent audit after all experiments shows that excluding the one object with a block occluded in both views shifts model accuracies by at most 1.0\,pp, changing no findings.
\tthirteenlong{} scenes contain 6--8 collision-free boxes, cubes, and cylinders, each checked for visibility.
Details are in Appendix~\ref{sec:app_dataset}.
% Appendix~\ref{sec:app_dataset} gives the generation procedures, filtering thresholds, and distractor construction.

\paragraph{Visual conditions.}
We render matched variants with or without color and coordinate-axis overlays.
Colored objects use 4--12 colors from a 12-color palette, organized into six pairs of distinct hues with similar luminance, which are placed on neighboring blocks: \textcolor{bedRed}{red}/\textcolor{bedMagenta}{magenta}, \textcolor{bedBlue}{blue}/\textcolor{bedPurple}{purple}, \textcolor{bedGreen}{green}/\textcolor{bedGold}{gold}, \textcolor{bedCyan}{cyan}/\textcolor{bedYellow}{yellow}, \textcolor{bedOrange}{orange}/\textcolor{bedOrchid}{orchid}, and \textcolor{bedTeal}{teal}/\textcolor{bedSlategray}{slategray}.
% The adversarial coloring mode pairs distinct hues with similar luminance to reduce brightness-based boundary cues.
Only task-compatible variants are used.
Table~\ref{tab:visual_cue_effects} (Appendix~\ref{sec:app_visual_cues}) lists the default setting and variants for each task.

\section{Experimental Setup}
\label{sec:experiments}

\paragraph{Models.}
We evaluate twelve OW models from four families: Qwen3.5 \cite{qwen3.5}, Gemma4 \cite{gemma4}, GLM-4.6V-Flash \cite{vteam2025glm45vglm41vthinkingversatilemultimodal}, and Kimi-VL-A3B \cite{kimiteam2025kimivltechnicalreport}.
The roster includes dense models and mixture-of-experts (MoE) models, which activate only part of their parameters for each token.
We also evaluate four frontier models: GPT-5 \cite{singh2025openaigpt5card}, Grok-4.20 \cite{grok4.20}, Gemini-3-Flash/3.1-Pro \cite{gemini3flash,gemini3.1pro}.
Appendix~\ref{sec:app_model_details} gives more details about the roster.

\paragraph{Evaluation.}
Our evaluation uses 100 instances per task, except \televenlong{} and \tthirteenlong{}, which each contain 300 questions covering three projection planes for 100 objects/scenes.
Their default setting asks for the invalid candidate; the variant that asks for the valid candidate is reported in Appendix~\ref{sec:app_metrics_t11} and Appendix~\ref{sec:app_metrics_t13}.

\paragraph{Baselines.}
Our human baseline consists of 30 volunteers, each answering one randomly drawn instance per task under the default visual setting.
We report per-task human accuracy with 95\% Wilson CIs (Appendix~\ref{sec:human_baseline_details}).
For binary and MCQ tasks, the random reference is uniform selection among the available answers.
We also report a best-constant baseline: the accuracy of always giving the most frequent ground-truth answer for a task.

\paragraph{Metrics.}
Models are prompted to return answers using task-specific structured-output templates.
We parse the answer fields and report exact-match (EM) accuracy against ground truth.
% For \tfivelong{}, every color and its count must match; a numerically close answer receives no partial credit in accuracy.
We also report secondary metrics: mean absolute error (MAE) and root mean squared error (RMSE) give the size of numerical mistakes, and precision, recall, and F1 give the composition of the errors on binary tasks. 
% The latter separate a change in accuracy from a change in which way it leans when it answers yes or no (Appendix~\ref{sec:app_secondary_metrics}).

% \paragraph{Inference settings.}
% We evaluate OW models in Non-Thinking and Thinking modes using greedy decoding, with a default requested thinking budget of 1024 tokens.
% For Thinking, we use task-specific decomposition instructions that explicitly guide intermediate reasoning steps. 
% Non-Thinking uses concise task prompts without decomposition instructions.
% Prompt templates include unpaired, illustrative response-format examples specifying the expected output schema (Appendix~\ref{sec:app_prompts}).
% Each OW condition is evaluated in five repeated inference runs, with results summarized by their mean and standard deviation; this deviation reflects nondeterministic batched inference, and uncertainty over benchmark items is reported separately with paired bootstrap intervals.
% Frontier models are evaluated once in Thinking mode through provider APIs; their reasoning budgets are not guaranteed by providers (see Appendix~\ref{sec:app_exp_details}).

\paragraph{Inference settings.}
We evaluate OW models using greedy decoding in Non-Thinking and Thinking modes, with a default thinking budget of 1024 tokens.
Motivated by CoT prompting \cite{wei2022chain,kojima2022large}, Thinking combines reasoning mode with task-specific decomposition prompts to support multi-step reasoning, whereas Non-Thinking uses concise prompts without decomposition.
Both prompts also include a task-specific example response format (Appendix~\ref{sec:app_prompts}).
Each OW condition is repeated five times; we report mean and standard deviation across runs and paired bootstrap intervals over benchmark items.
Frontier models are evaluated once in Thinking mode via provider APIs with provider-dependent reasoning budgets (Appendix~\ref{sec:app_exp_details}).

% \paragraph{Diagnostic comparisons.}
% Matched color and axis variants test whether additional visual cues help, while image removal tests whether predictions depend on the images.
% On \tonelong{}, we also compare open-ended answers with five-option MCQ on the same items.
% We examine model scaling and dense versus MoE architectures, and conduct smaller studies of reasoning budgets and task-specific fine-tuning.
% The budget sweep covers two models and three tasks; fine-tuning uses Qwen3.5-27B on 800 \tfivelong{} examples with its vision encoder frozen.
% Image removal covers three models, so these diagnostic comparisons do not all use the full model roster.

\paragraph{Diagnostic comparisons.}
For \tfive{}, we add three controls: twelve single-color queries merged into one count vector, a 2D rendering of the same blocks as separate tiles, and LoRA fine-tuning of Qwen3.5-27B on 800 in-distribution examples with its vision encoder frozen.
We evaluate the fine-tuned model on \tfive{} and the related color tasks \tfour{}, \tseven{}, and \tten{}, and also ask it the \tseven{} question with its own \tfive{} counts given as text.
Across tasks, we vary the reasoning budget, model scale and architecture, and color and axis overlays, remove the images, and pose \tonelong{} as a five-option MCQ (Section~\ref{sec:results}).

% \paragraph{Diagnostic comparisons.}
% We use the following comparisons to connect performance differences to the diagnostic questions in Section~\ref{sec:results}:
% \begin{itemize}[leftmargin=1.4em,nosep]
%     \item \textbf{Counting controls, fine-tuning, and consistency.}
%     For \tfive{}, we decompose the 12-color distribution into twelve independent single-color count queries and merge the predictions into a complete count vector.
%     A matched 2D control presents the same blocks as individually visible tiles.
%     Task-specific fine-tuning uses Qwen3.5-27B on 800 in-distribution \tfivelong{} examples with its vision encoder frozen; we test the adapter on \tfive{} and on the related color tasks \tfour{}, \tseven{}, and \tten{}.
%     For the fine-tuned model, we compare direct \tseven{} answers with answers obtained by providing its \tfive{} counts as text.
%     \item \textbf{Reasoning budget.}
%     Our reasoning-budget sweep covers two models and three tasks.
%     \item \textbf{Model scale and architecture.}
%     We compare performance across model sizes and between dense and MoE architectures.
%     \item \textbf{Visual cues.}
%     Matched color and axis variants test whether additional visual cues help.
%     \item \textbf{Image dependence and answer format.}
%     Image removal tests whether predictions depend on the images; the image-removal comparison covers three models.
%     On \tonelong{}, we compare open-ended answers with a five-option MCQ control whose correct option is always third.
% \end{itemize}

\paragraph{Visual-feature probes.}
To examine which task targets can be recovered from visual representations, we train probes, small models that predict a target from the features of a frozen vision encoder.
We average the patch features of each view and probe 14 encoders.
The probes cover \tone{}--\tten{} and use five-fold cross-validation grouped by object, preventing the same object from appearing in both training and evaluation folds.
Probes are compared with shuffled-label controls; \televen{}--\tthirteen{} use training-free candidate comparisons (Appendix~\ref{sec:app_probing}).
For \tfive{}, we also train probes on the 800 generated objects used for fine-tuning and test them on the benchmark objects, using either averaged features or an attention readout over all patch tokens.

\section{Results and Diagnostic Analysis}
\label{sec:results}
% arXiv version of figures/fig_human_vlm_gap.tex (same content), wrapfig, as in arXiv:2608.07645 (simulation.tex).
\begin{wrapfigure}{r}{0.5\textwidth}
\vspace{-1.8em}
  \centering
  \includegraphics[width=\linewidth]{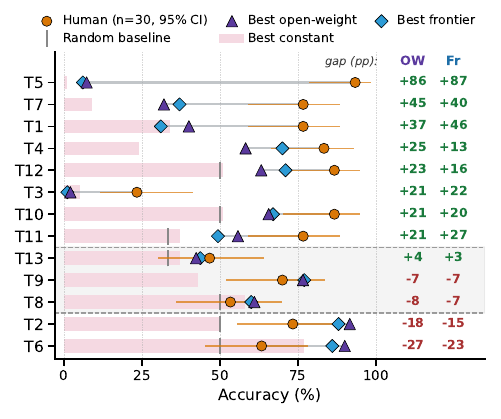}
  % \caption{\textbf{Human vs.\ best-model accuracy per task.} Right columns give human$-$model gaps (pp); dashed lines group tasks by whether the best OW result falls below, within, or above the human 95\% CI. \textbf{Takeaway:} \tfivelong{} shows the largest gap; the best models exceed the human interval only on \ttwo{} and \tsix{}.}
    \caption{\textbf{Human vs.\ best-model accuracy per task.} Shaded bars mark the best-constant baseline (always the most frequent answer); right columns give human$-$model gaps (pp); dashed lines group tasks by whether the best OW result falls below, within, or above the human 95\% CI. \textbf{Takeaway:} \tfivelong{} shows the largest gap; the best models exceed the human interval only on \ttwo{} and \tsix{}.}
  \label{fig:human_vlm_gap}
\vspace{-1.8em}
\end{wrapfigure}

Our results repeatedly show that models can get individual judgments right while the exact answer stays wrong.
We first examine why exact counting fails and whether training supports consistent related judgments (\S\ref{sec:3d_blind}).
We then test the usual remedies for spatial failures, reasoning (\S\ref{sec:thinking_paradox}) and scaling and visual support (\S\ref{sec:scaling}), to see whether any of them makes exact answers reliable and where the difficulty arises.
Finally, since a high score is meaningful only if it depends on the image, we test this dependence (\S\ref{sec:mcq}).

% Finally, since a high score is meaningful only if it depends on the image, we test this dependence (\S\ref{sec:mcq}).

% ------------------------------------------------------------------
% \subsection{Approximate Quantities Do Not Ensure Exact Compositional Answers}
\subsection{Exact Counting and Consistent Use Remain Distinct Challenges}
\label{sec:3d_blind}
\label{sec:human_gap}

\begin{figure}[t]
  \centering
  \includegraphics[width=\linewidth]{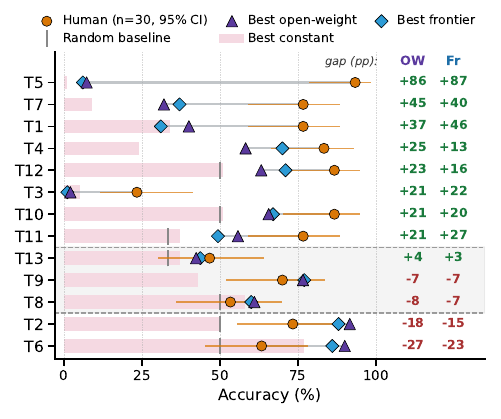}
  % \caption{\textbf{Human vs.\ best-model accuracy per task.} Right columns give human$-$model gaps (pp); dashed lines group tasks by whether the best OW result falls below, within, or above the human 95\% CI. \textbf{Takeaway:} \tfivelong{} shows the largest gap; the best models exceed the human interval only on \ttwo{} and \tsix{}.}
    \caption{\textbf{Human vs.\ best-model accuracy per task.} Shaded bars mark the best-constant baseline (always the most frequent answer); right columns give human$-$model gaps (pp); dashed lines group tasks by whether the best OW result falls below, within, or above the human 95\% CI. \textbf{Takeaway:} \tfivelong{} shows the largest gap; the best models exceed the human interval only on \ttwo{} and \tsix{}.}

  \label{fig:human_vlm_gap}
\end{figure}

\paragraph{Per-color counting (\tfivelong{}) exposes a selective model weakness.}
Averaged over tasks, humans reach 70.0\%, the strongest model (Gemini-3-Flash) 51.5\%, and the best-constant baseline 36.7\% (Table~\ref{tab:full_main_results}, Appendix~\ref{sec:app_extended_findings}).
VLMs perform strongly on some geometric decisions: the best models exceed the confidence interval upper bounds of humans on deciding whether the object fits in a given container (\ttwolong{}) and whether it has a vertical column taller than a given threshold (\tsixlong{}; there, the best model's 90.0\% also exceeds the 77.0\% obtained by always giving the most frequent answer).
Yet on \tfivelong{}, where every color count must be correct, the best VLM reaches only \textcolor{red!70!black}{7.2\%} accuracy, compared with \textcolor{custom_darkgreen}{93.3\%} for humans (Figure~\ref{fig:human_vlm_gap}).
The individual counts are already unreliable: across the 12 OW models in Non-Thinking mode, per-color MAE is 0.80 blocks over the full palette, worse than the 0.69 obtained by always predicting each color's mean count, although five models, including the two strongest, outperform it (Appendix~\ref{sec:app_floor_regimes}).
Thus, low exact-match accuracy cannot be attributed solely to a strict metric applied to otherwise reliable counts.

% \paragraph{Exact match requires every count to be correct.}
% A complete \tfivelong{} answer contains a count for every color, and a mistake in any one of them makes the entire answer incorrect.
% Averaged over the 12 open-weight models in Non-Thinking mode, \tfive{} has a MAE of 0.80 blocks per palette color, absent colors included, while exact-match accuracy is only 0.7\% (Table~\ref{tab:sec_floor}, Appendix~\ref{sec:app_floor_regimes}).
% These results motivate us to examine whether improving the numerical estimates also makes complete answers reliably correct.

% \paragraph{A single wrong count makes the whole answer wrong.} (merged into the paragraph above)
% We therefore ask whether making the counts more accurate would make complete answers correct.

\paragraph{Factorized querying does not consistently improve fine-grained color counting.}
We first test whether simplifying the requested output improves these estimates.
Instead of requesting all color counts in one response, we ask twelve independent single-color questions and merge their predictions.
For colors present in the object, average exact-count accuracy remains essentially unchanged, from 32.5\% under joint prediction to 33.2\% under factorized querying.
Expected exact match over subsets of $k$ present-color counts declines rapidly as $k$ grows, with closely aligned curves for joint and factorized predictions (Figure~\ref{fig:t5_k_color_present}).
The merged predictions achieve \textcolor{red!70!black}{0.0\%} full-distribution EM across all 12 OW models (Appendix~\ref{sec:app_t5_factorized}).
Thus, simplifying the output does not resolve the per-color counting errors that undermine exact answers.

% Thus, simplifying the output does not resolve the individual counting errors that undermine complete answers.

\begin{figure}[t]
  \centering
  \includegraphics[width=\linewidth]{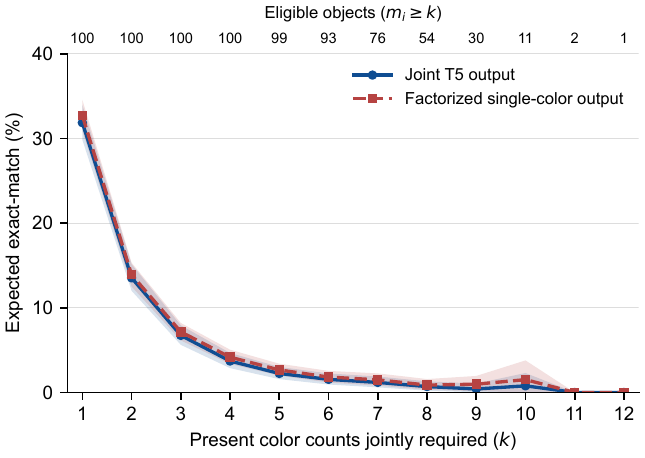}
  \caption{\textbf{Present-color subset exact match on \tfivelong{}.} Expected EM over $k$ present-color counts from saved joint and factorized predictions. Shading: pointwise 95\% object-bootstrap intervals. \textbf{Takeaway:} Exact recovery declines rapidly with $k$, with closely aligned joint and factorized mean curves.}
  \label{fig:t5_k_color_present}
\end{figure}

% arXiv version of figures/fig_t5_k_color_present.tex (same content), wrapfig, as in arXiv:2608.07645 (simulation.tex).
\begin{wrapfigure}{l}{0.5\textwidth}
  \centering
  \includegraphics[width=\linewidth]{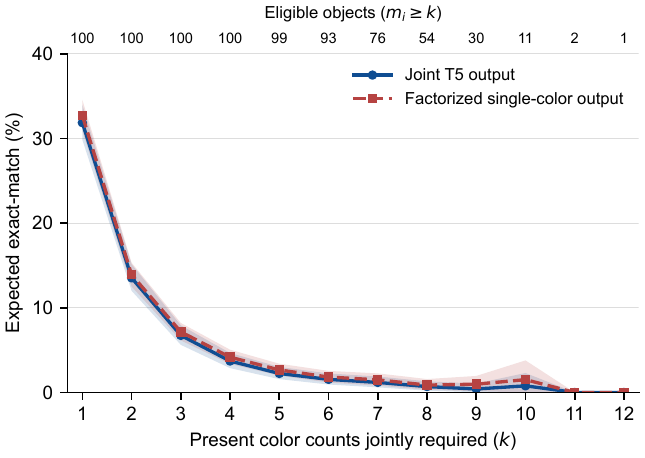}
  \caption{\textbf{Present-color subset exact match on \tfivelong{}.} Expected EM over $k$ present-color counts from saved joint and factorized predictions. Shading: pointwise 95\% object-bootstrap intervals. \textbf{Takeaway:} Exact recovery declines rapidly with $k$, with closely aligned joint and factorized mean curves.}
  \label{fig:t5_k_color_present}
\end{wrapfigure}

\paragraph{3D presentation amplifies constituent counting errors.}
Since simplifying the output offers little benefit, we next ask whether the 3D presentation makes counting difficult.
Our matched 2D control displays the same colored blocks as individually visible tiles while retaining the per-color counting task.
For Qwen3.5-27B, exact-match accuracy rises from 3.0\% in 3D to 52.4\% in 2D (\textcolor{custom_darkgreen}{$+$49.4\,pp}; 95\% CI $[+39.8, +59.0]$), while per-color MAE drops from 0.50 to 0.11.
After merging the frequently confused \textcolor{bedGold}{gold} and \textcolor{bedYellow}{yellow} categories in both predictions and ground truth, EM reaches \textcolor{custom_darkgreen}{89.8\%} in 2D versus \textcolor{red!70!black}{6.0\%} in 3D.
These results suggest that 3D presentation introduces substantial difficulty beyond counting individually visible colored blocks, although the control does not isolate occlusion, view integration, and layout effects (Table~\ref{tab:t5_2d_control}, Appendix~\ref{sec:app_2d_control}).

% \paragraph{Fine-tuning improves exact counting but leaves cross-task consistency unresolved.}
% We LoRA-fine-tune Qwen3.5-27B on 800 in-distribution \tfivelong{} examples while freezing its vision encoder.
% EM accuracy rises \textcolor{custom_darkgreen}{from 3.0\% to 68.6\%}, still \textcolor{red!70!black}{24.7\,pp} below the human reference (Table~\ref{tab:t5_finetune}, Appendix~\ref{sec:app_finetune}).
% The gain is task-specific: on the same objects, the dominant colors implied by the model's own \tfive{} answers are correct \textcolor{custom_darkgreen}{74.6\%} of the time, yet the model answers \tsevenlong{} directly only \textcolor{red!70!black}{35.0\%} of the time, and 40.8\% even when its \tfive{} answer is exactly correct (Table~\ref{tab:t5_finetune_transfer}, Appendix~\ref{sec:app_finetune}).
% Fine-tuning also improves related color tasks, but cross-task consistency remains limited: on the same objects, the dominant colors implied by the model's own \tfive{} answers are correct 74.6\% of the time, yet the model answers \tsevenlong{} directly only 35.0\% of the time, and 40.8\% even when its T5 answer is exactly correct (Table~\ref{tab:t5_finetune_transfer}, Appendix~\ref{sec:app_finetune}).
% The fine-tuned model's direct answers about the dominant colors are thus inconsistent with its own color counts: supervision improves the trained readout without a matching gain on a question those counts settle.
\paragraph{Fine-tuning improves counting but does not ensure consistent use.}
Keeping the original 3D views, we next test whether training improves both exact counting and related judgments.
We fine-tune Qwen3.5-27B on 800 in-distribution \tfivelong{} examples with LoRA on the language model only; exact match rises from 3.0\% to 68.6\% (Table~\ref{tab:t5_finetune}, Appendix~\ref{sec:app_finetune}).
To test consistent use, we ask which colors are most common (\tsevenlong{}), a question a correct set of counts already settles.
Accuracy is only 35.0\% when answering from images, but rises to 74.2\% when given the model's own \tfive{} counts as text (\textcolor{custom_darkgreen}{$+$39.2\,pp}; 95\% CI $[+26.8, +51.4]$).
When those counts are exactly correct, text-conditioned accuracy reaches 99.4\%, versus 40.8\% for direct visual questioning on the same objects.
Recovering a property therefore does not mean using it consistently: the model derives the judgment from explicit counts but not from images (Table~\ref{tab:t5_t7_from_counts}, Appendix~\ref{sec:app_finetune}).
This inconsistency is not specific to fine-tuning: per object, the direct \tseven{} answers of the 12 OW models match the most frequent colors in their own counts on only 21.6\% of objects (chance 2.4\%), and their direct \tone{} answers match the sum of their own counts no more often than chance (Table~\ref{tab:cross_task_consistency}, Appendix~\ref{sec:app_finetune}).
We next ask whether reasoning from the images can produce useful intermediate steps.
\subsection{Test-time Reasoning Leaves Exact Answers Unreliable}
\label{sec:thinking_paradox}

% \paragraph{Reasoning-oriented prompting improves selected tasks but leaves exact answers unreliable.}
% We use \emph{verbalizable steps} to describe intermediate operations organized explicitly in text (e.g., listing visible colors).
% Test-time reasoning does not reliably close the identified gap.
% Its largest gains occur on \ttwolong{} (\textcolor{custom_darkgreen}{$+$13.8\,pp}) and \tfourlong{} (\textcolor{custom_darkgreen}{$+$13.9\,pp}).
% Our qualitative examples illustrate these steps: on \ttwo{}, models compare estimated object dimensions with container boundaries; on \tfour{}, models accumulate a list of visible colors (Table~\ref{tab:qual_flip_t2},~\ref{tab:qual_flip_t4}, Appendix~\ref{sec:app_qual_thinking_flip}).
% Yet these benefits do not translate into reliable recovery of exact compositional structure.
% Complete per-color counting (\tfivelong{}) remains near floor, with mean accuracy increasing only \textcolor{custom_darkgreen}{from 0.7 to 1.0\%}, and structural difference (\tthreelong{}) remains similarly low.
% Reasoning also reduces accuracy on \tonelong{} and \tninelong{}; the decrease on \teightlong{} is smaller and its 95\% interval includes zero (Table~\ref{tab:thinking_delta_full}, Appendix~\ref{sec:app_thinking}).
% Together, these results suggest that reasoning can support verbalizable steps without reliably recovering exact compositional structure.
% These comparisons use the default 1024-token budget, so we next test whether a larger budget recovers the tasks where thinking initially fails.

\paragraph{Reasoning-oriented prompting improves selected tasks but leaves exact answers unreliable.}
We first examine whether reasoning helps models recover the exact information needed for subsequent judgments.
At the default 1024-token budget, the largest gains are on \ttwolong{} (\textcolor{custom_darkgreen}{$+$13.8\,pp}) and \tfourlong{} (\textcolor{custom_darkgreen}{$+$13.9\,pp}).
The qualitative examples show models comparing dimensions or accumulating lists of visible colors, intermediate operations that can be written out explicitly in text, which we call \emph{verbalizable steps} (Tables~\ref{tab:qual_flip_t2} and~\ref{tab:qual_flip_t4}, Appendix~\ref{sec:app_qual_thinking_flip}).
Yet \tfivelong{} remains near floor ({0.7\% to 1.0\%}), as does \tthreelong{}, while accuracy declines on \tonelong{} and \tninelong{}; the smaller decline on \teightlong{} has a 95\% CI including zero (Table~\ref{tab:thinking_delta_full}, Appendix~\ref{sec:app_thinking}).
Thus, the tested Thinking configuration helps selected individual judgments while exact answers remain unreliable.
We next test larger reasoning budgets.

\paragraph{Increasing the reasoning budget does not consistently improve 3D accuracy.}
We vary the reasoning budget for Qwen3.5-27B and Gemma4-31B on \tfourlong{} and \tonelong{}, the largest gain and loss at the default budget, and on the cross-view task \televenlong{} (Table~\ref{tab:thinking_scaling}, Appendix~\ref{sec:app_thinking_scaling}).
On \tfour{}, mean accuracy rises \textcolor{custom_darkgreen}{from 49.0\%} without thinking \textcolor{custom_darkgreen}{to 59.2\%} at 2048 tokens, then levels off. 
In contrast, every larger budget leaves mean accuracy on \tone{} below the budget-0 level (\textcolor{red!70!black}{37.4\%} to \textcolor{red!70!black}{19.8\%} at 4096 tokens), and budgets of 2048 tokens or more do the same on \televen{} (\textcolor{red!70!black}{51.5\%} to \textcolor{red!70!black}{45.8\%}).
For both tasks, raising the budget from 2048 to 4096 tokens changes accuracy by less than 1\,pp.
Our qualitative examples illustrate this contrast: a long \tfour{} trace organizes colors within an explicit coordinate lattice, whereas a long \tone{} trace continues to reconsider the alignment of the two views (Tables~\ref{tab:qual_budget_t1} and~\ref{tab:qual_budget_t4}, Appendix~\ref{sec:app_qual_thinking_budget}). 
Thus, more reasoning can help organize color enumeration but does not reliably correct the geometric estimates needed to recover 3D structure.
% To understand this contrast, we next examine whether the reasoning traces show models correcting their visual interpretations or continuing to reason from inaccurate ones.

\paragraph{Incorrect visual interpretations often persist through subsequent reasoning.}
To understand why additional reasoning helps some tasks but fails to recover others, we examine whether models revise their visual interpretations as reasoning proceeds.
In our exploratory manual audit, models often misinterpret the scene and continue reasoning from that initial misinterpretation without correcting it 
% In our manual exploratory audit, model usually read the scene incorrectly and then reiterating on that wrong perception 
% \textit{Wrong Perception} (i.e., the model's stated read of the scene is incorrect relative to the actual image) appears in 42 of 44 incorrect traces, and OW models frequently \textit{Reiterate} (i.e., repeat previously computed steps) 
% without correcting their initial interpretation 
(Figure~\ref{fig:thinking_trace_fsm_main}; full taxonomy in Table~\ref{tab:annot_taxonomy}, Appendix~\ref{sec:app_thinking}).
Together with the task-level results, this suggests that reasoning helps some individual judgments but can extend a mistaken scene interpretation, so the exact answer still stays wrong.
We next examine whether scaling and visual cues help, and what visual features make available.

\subsection{Scaling and Visual Support Leave Exact Answers Unreliable, Although Visual Features Hold the Counts}
\label{sec:scaling}

% arXiv version: state-machine figure (cited at the end of Section 5.2, same page) beside the
% first paragraphs of Section 5.3, below its heading (ACL: end of Section 5).
% arXiv version of figures/automaton_reasoning_traces_main.tex (same content), wrapfig, as in arXiv:2608.07645 (simulation.tex); spans paragraphs.
\begin{wrapfigure}{l}{0.42\textwidth}
  \centering
  \includegraphics[width=\linewidth]{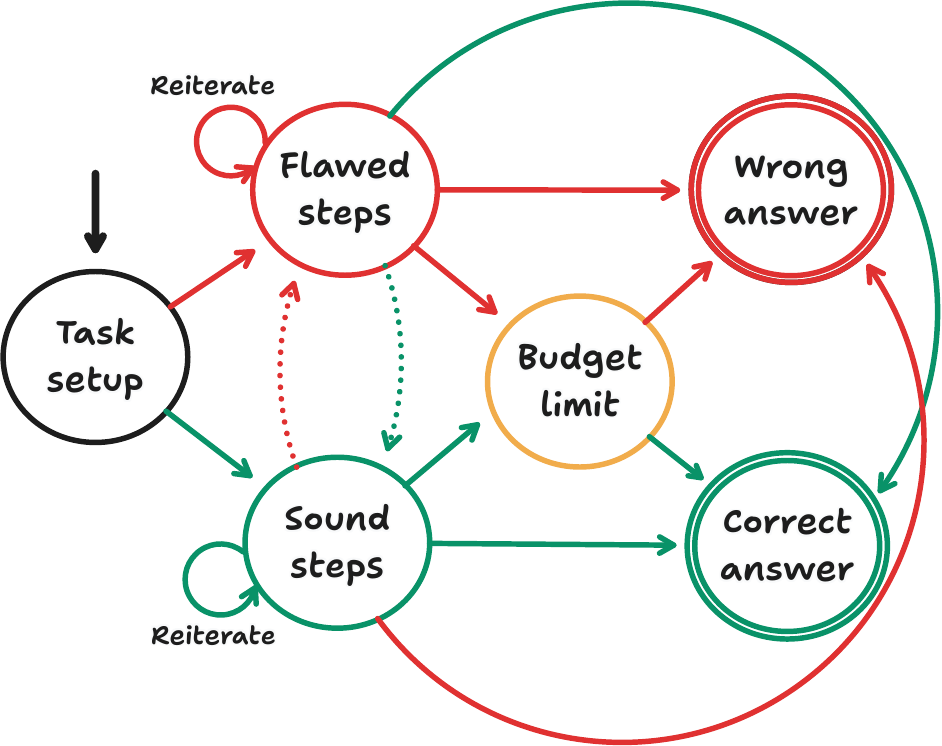}
  \caption{\textbf{State-machine summary of annotated reasoning traces.} Flawed steps include visual misreads; sound steps include correct perception and valid tactics. Dotted links class switches, and outer arcs trace--answer mismatches (taxonomy: Table~\ref{tab:annot_taxonomy}, Appendix~\ref{sec:app_thinking}). 
\textbf{Takeaway:} Visual misinterpretations occur in 42 of the 44 annotated traces that end in incorrect answers.}
  \label{fig:thinking_trace_fsm_main}
\end{wrapfigure}

\paragraph{Larger models improve individual judgments but leave exact answers unreliable.}
We first examine whether increasing model size closes the gap between individual judgments and exact answers.
Within Qwen3.5 in Non-Thinking mode, increasing model size from 0.8B to 27B improves \ttwolong{}, \tfourlong{}, \tsixlong{}, \tsevenlong{}, \tninelong{}, and \televenlong{} by \textcolor{custom_darkgreen}{18--42\,pp}, with paired bootstrap intervals that all exclude zero, whereas the remaining seven tasks change by at most \textcolor{custom_darkgreen}{6.4\,pp} in point estimate (Figure~\ref{fig:scale_analysis}; intervals in Appendix~\ref{sec:app_scale_full}).
This uneven improvement also appears in Gemma4, with a few task-specific exceptions; notably, both families remain near floor on \tthreelong{} and \tfivelong{} (more details in Appendix~\ref{sec:app_scaling_robustness}).
The mixture-of-experts variants follow the same pattern: Qwen3.5-35B-A3B and Gemma4-26B-A4B average 43.1\% and 38.5\% in Non-Thinking mode, below their dense counterparts of similar total size (46.8\% and 46.5\%), and both remain at 0.0\% on \tthree{} and \tfive{} (Table~\ref{tab:full_main_results}, Appendix~\ref{sec:app_extended_findings}).
Within the tested families, larger models therefore get more individual judgments right while the exact answer stays wrong.
We next ask whether visual cues help.

\begin{figure}[h]
  \centering
  % \includegraphics[width=\linewidth]{images/fig3_scale_analysis_v2.pdf}
  % \caption{\textbf{Accuracy across Qwen3.5 model sizes (Non-Thinking).} Language-model size grows from 0.8B to 35B-A3B; the vision encoder grows from 101M to about 450M parameters (447--461M from 9B onward). \textbf{Takeaway:} From 0.8B to 27B, six tasks gain 18--42\,pp, while the other seven change by at most 6.4\,pp.}
  \includegraphics[width=\linewidth]{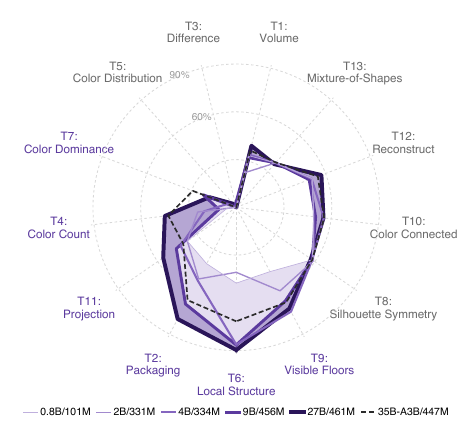}
    \caption{\textbf{Accuracy across Qwen3.5 model sizes (Non-Thinking).} Legend: LLM/ViT size; line width grows with size. Light band: gain from 0.8B to 9B (LLM and ViT grow); dark band: 9B to 27B (ViT nearly fixed). Dashed: 35B-A3B MoE. \textbf{Takeaway:} Six tasks (purple labels) gain 18--42\,pp; the rest change by at most 6.4\,pp.}

  \label{fig:scale_analysis}
\end{figure}

\paragraph{Additional visual cues leave exact answers unreliable.}
We add color or coordinate-axis overlays to the input images while keeping the geometry fixed (Appendix~\ref{sec:app_qual_visual_cues}).
Color helps most on \ttwelvelong{} (\textcolor{custom_darkgreen}{$+$6.8\,pp}), where it can mark corresponding parts across candidate views. On \tthreelong{}, adding color raises accuracy \textcolor{custom_darkgreen}{from 0.2\% to 3.7\%}, but complete structural-difference answers remain overwhelmingly incorrect. On \tfivelong{}, adding coordinate axes provides no improvement, with accuracy decreasing \textcolor{red!70!black}{from 0.7\% to 0.5\%} (Table~\ref{tab:visual_cue_effects}, Appendix~\ref{sec:app_visual_cues}).
Thus, visual cues help selected individual judgments while the exact answer stays wrong.
Since neither scale nor visual cues make exact answers reliable, and Qwen3.5 and Gemma4 both use SigLIP-2-family encoders, we next ask whether the frozen visual features hold the information exact answers require.

% arXiv version of figures/fig_scale_radar.tex (same content), wrapfig, as in arXiv:2608.07645 (simulation.tex); spans paragraphs.
\begin{wrapfigure}{r}{0.55\textwidth}
  \centering
  % \includegraphics[width=\linewidth]{images/fig3_scale_analysis_v2.pdf}
  % \caption{\textbf{Accuracy across Qwen3.5 model sizes (Non-Thinking).} Language-model size grows from 0.8B to 35B-A3B; the vision encoder grows from 101M to about 450M parameters (447--461M from 9B onward). \textbf{Takeaway:} From 0.8B to 27B, six tasks gain 18--42\,pp, while the other seven change by at most 6.4\,pp.}
  \includegraphics[width=\linewidth]{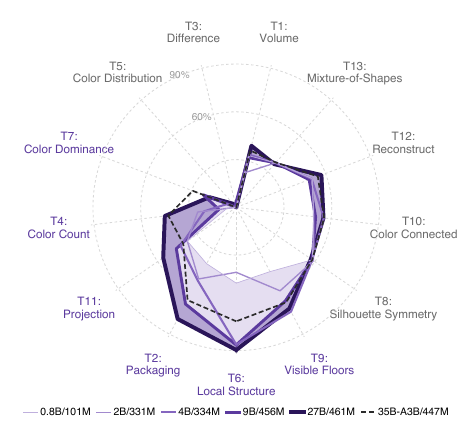}
    \caption{\textbf{Accuracy across Qwen3.5 model sizes (Non-Thinking).} Legend: LLM/ViT size; line width grows with size. Light band: gain from 0.8B to 9B (LLM and ViT grow); dark band: 9B to 27B (ViT nearly fixed). Dashed: 35B-A3B MoE. \textbf{Takeaway:} Six tasks (purple labels) gain 18--42\,pp; the rest change by at most 6.4\,pp.}

  \label{fig:scale_analysis}
\end{wrapfigure}

\paragraph{Pooled probes recover aggregate quantities but rarely exact counts.}
Probes on averaged frozen features recover quantitative information about the objects (e.g., volume for \tonelong{}, the number of distinct colors for \tfourlong{}, and the number of visible floors for \tninelong{}) even from the smallest deployed encoders, and this ability is maintained across the encoders used by VLMs of different sizes (Figure~\ref{fig:tower_scaling}, Appendix~\ref{sec:app_probing_scaling}).
Complete per-color counts, however, are rarely recovered by these probes, which are trained on the 100 benchmark objects: EM is at most \textcolor{red!70!black}{3.5\%} across encoders, including a depth-supervised one (Table~\ref{tab:probe_decodability}, Appendix~\ref{sec:app_probing_availability}).
This mirrors the VLM results, so we test whether it reflects the features or the small training set of the probes.

% arXiv version: radar chart beside this paragraph, where the page has room (ACL: after the scaling paragraph).
\paragraph{With more training data, frozen features yield exact counts.}
Using the 800 generated training objects and 100 benchmark test objects, attention probes recover exact counts from frozen Qwen3.5-27B and Gemma4-31B vision-encoder tokens on \textcolor{custom_darkgreen}{61.7\%} and \textcolor{custom_darkgreen}{44.7\%} of objects, versus \textcolor{red!70!black}{3.0\%} and \textcolor{red!70!black}{4.0\%} for the VLMs.
The same probe design reaches \textcolor{custom_darkgreen}{65.0\%} on Qwen3.5-27B's connector output, the image embeddings supplied to its language model (Table~\ref{tab:token_probe}, Appendix~\ref{sec:app_token_probe}).
Scores continue rising from 200 to 800 training objects.

% \paragraph{The evidence suggests an encoder bottleneck.} (previous version, 2026-10-10)
% We suggest that the vision encoder contributes to the difficulty of recovering exact structure: families with different language models but related vision encoders share persistent failures, and frozen-feature probes reproduce the gap between approximate quantities and exact answers.
% Together, these observations suggest that the visual features may represent exact attribute--quantity associations less effectively than aggregate geometry.
% Failures may also arise during feature pooling or downstream use: averaging can obscure spatial distinctions, while the language model may fail to use available information consistently.
% On \televenlong{}, Gemma4 features place valid projections closer to the object than invalid ones, while Qwen3.5 features show the opposite ordering (Figure~\ref{fig:tower_scaling}, Appendix~\ref{sec:app_probing_scaling}).
% Both distinguish the candidates systematically, yet the full VLMs remain unreliable.
% The substantial \tfive{} gain with a frozen encoder likewise shows that downstream training can improve access to its features.
% An encoder bottleneck is therefore a plausible contributor to unreliable exact recovery, while pooling and downstream processing may limit access to and use of available information.

\paragraph{The difficulty lies in using the visual features.}
% \paragraph{The evidence points to how the VLMs use their visual features.} \khai{should be findings, not a summary of the paragraph}
For Qwen3.5-27B, count information thus reaches the language-model input, pointing to difficulty in how the language model uses it.
This agrees with prior observation in simpler 2D tasks 
% where probes succeed but decoding fails 
\citep{rahmanzadehgervi2025visionlanguagemodelsblind}.
Three observations reinforce this interpretation.
First, training only the language model raises \tfive{} accuracy to 68.6\%, with the encoder and connector frozen.
Second, Qwen3.5-27B counts far more accurately when each block is a separate 2D tile (\S\ref{sec:3d_blind}).
Third, aggregate-quantity probe scores stay flat across encoder sizes while VLM accuracy rises (Figure~\ref{fig:tower_scaling}, Appendix~\ref{sec:app_probing_scaling}).
Recovering a property in the visual features thus does not mean using it consistently.

\subsection{Shortcut Controls Show Why Score Alone Is Insufficient}
\label{sec:mcq}
\label{sec:no_image}

\paragraph{Some scores can be reproduced without visual evidence.}
Interpreting recovery or use of object properties as visual competence also requires checking that successful answers depend on the images, so we remove the images and repeat the evaluation on three models.
Removing images substantially reduces accuracy on several tasks, while leaving others nearly unchanged (Table~\ref{tab:no_image_full_nt}, Appendix~\ref{sec:app_no_image_full}).
Inspecting the predictions explains how substantial scores can survive: without images, all three models answer \texttt{0} on every \ttenlong{} instance and \texttt{3} on every \tninelong{} instance, reaching 51\% and 43\% accuracy, respectively (Table~\ref{tab:prediction_collapse}, Appendix~\ref{sec:app_no_image_full}).
This raises a further question: \textit{can changing the answer format also produce an apparent improvement without requiring visual information?}

\paragraph{MCQ accuracy can come from the answer layout alone.}
The candidate-selection tasks in our benchmark (i.e., \televenlong{}, \ttwelvelong{}, \tthirteenlong{}) shuffle the position of the correct candidate, so always choosing one position yields at most 37.3\% on \televen{} and \tthirteen{} and 51.0\% on \ttwelve{} (Appendix~\ref{sec:app_mcq}).
As a control, we also pose \tonelong{} as a five-option MCQ with a fixed layout, in which the correct volume is always the third, middle option.
Under the default visual setting, accuracy rises by \textcolor{custom_darkgreen}{$+$39.0\,pp} on average across the OW models over open-ended prediction, with gains of up to \textcolor{custom_darkgreen}{$+$57.8\,pp} per model.
Without images, Gemma4-31B reaches 100\% by choosing the third option on every instance (Tables~\ref{tab:no_image_full_nt} and~\ref{tab:prediction_collapse}, Appendix~\ref{sec:app_no_image_full}).
Together, these controls show why scores must be checked for dependence on the image before they are interpreted as recovery of 3D quantities or structure.

\begin{figure}[t]
  \centering
  \includegraphics[width=\linewidth]{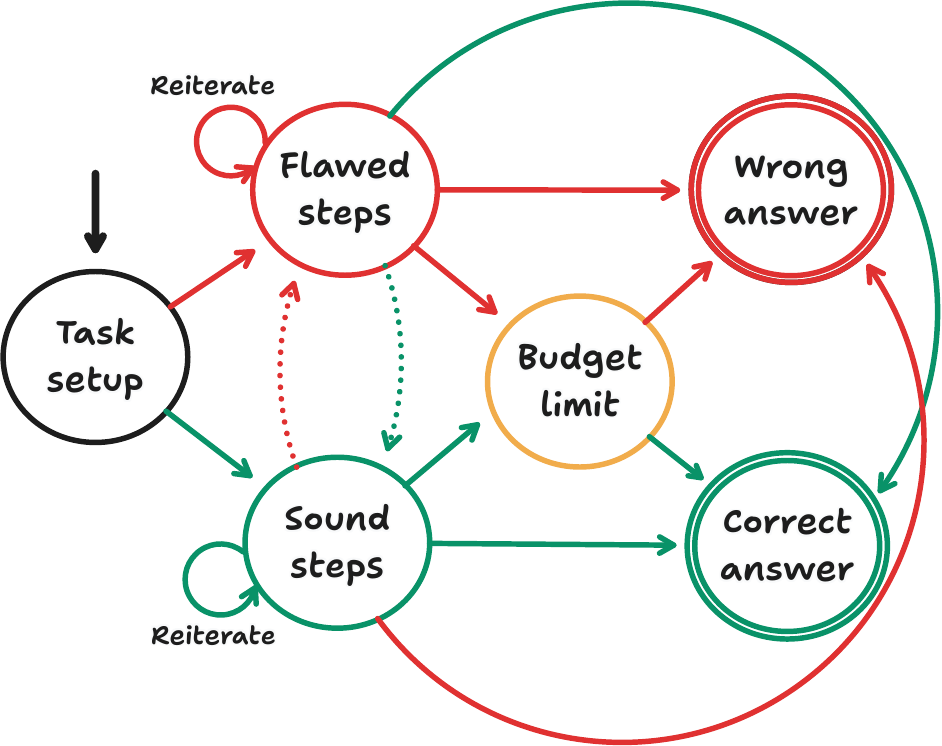}
  \caption{\textbf{State-machine summary of annotated reasoning traces.} Flawed steps include visual misreads; sound steps include correct perception and valid tactics. Dotted links class switches, and outer arcs trace--answer mismatches (taxonomy: Table~\ref{tab:annot_taxonomy}, Appendix~\ref{sec:app_thinking}). 
\textbf{Takeaway:} Visual misinterpretations occur in 42 of the 44 annotated traces that end in incorrect answers.}
  \label{fig:thinking_trace_fsm_main}
\end{figure}

\section{Conclusion}
% Across BedBench, models get individual judgments right while the exact answer stays wrong, and recovering a property does not mean using it consistently.
% For evaluation, 3D benchmarks should report exact answers with error sizes, test whether related answers agree, and check that scores depend on the image, since accuracy on individual judgments can overstate 3D understanding.
% For modeling, exact counts reach Qwen3.5-27B's language-model input, motivating training objectives that improve use of visual information and consistency.
% Future work could probe language-model layers to explain why the model does not use this information reliably, and extend these diagnostics to natural and embodied scenes.
% \khai{future work should say about some possible approaches to solve the problem we have found: e.g. " models get individual judgments right while the exact answer stays wrong, and recovering a property does not mean using it consistently"}

Across \BedBenchLogo, we find that models get individual judgments right while the exact answer stays wrong, and recovering a property does not mean using it consistently.
For evaluation, 3D benchmarks should report exact answers with error sizes, test whether related answers agree, and check that scores depend on the image, since accuracy on individual judgments can overstate 3D understanding.
For modeling, researchers should design training objectives that improve use of visual information and consistency.
For instance, models could be trained on exact labels across many tasks, stating intermediate structure before answering, with objectives that penalize disagreement between related answers.
We suggest future works to test these approaches, probe language-model layers to explain why the available information is not used reliably, and extend these diagnostics to natural and embodied scenes.

\section*{Limitations}

\paragraph{Scene realism and human baseline.}
BedBench uses procedurally generated voxel objects and geometric primitives under controlled rendering environments, which provides exact ground truth and lets us filter out ambiguous samples. While this controlled setup is essential for isolating fine-grained 3D reasoning failures, extending our findings to complex real-world or embodied environments requires assessing noisy textures, lighting variations, and complex scene geometries. Additionally, our human baseline covers 30 participants; broader and more diverse cohorts could further refine task-difficulty estimates and human variance.

\paragraph{Scope of probes and interventions.}
To pinpoint failure modes, we conducted extensive probing and behavioral interventions (e.g., targeted fine-tuning, counts-as-text prompting, and reasoning sweeps). However, due to computational and evaluation constraints, certain deeper analyses--such as token-level probing, internal LM hidden-state probing, and extensive fine-tuning--were executed on representative model families (e.g., Qwen3.5-27B) and selected tasks rather than the entire suite. Future work could generalize these internal state diagnostics across broader model families, training regimes, and embodied settings.

\section*{Ethical Considerations}

\paragraph{Human participants.}
The human baseline involved 30 volunteers recruited through the authors' networks (Appendix~\ref{sec:human_baseline_details}).
Participants were informed of the study purpose, approximate duration (20 minutes), voluntary nature, and right to withdraw, and provided informed consent.
For participants under 18, we additionally obtained parental or guardian consent and participant assent.
Participation was uncompensated.
The study used synthetic geometric renderings and involved no sensitive questions.

\paragraph{Participant data.}
We recorded per-task correctness, session time, and coarse self-reported demographics (age, sex, education), but no names, contact details, IP addresses, or free-text responses.
Participants consented to the release of anonymized responses; the released data contain only anonymous task verdicts and coarse demographic fields summarized in Appendix~\ref{sec:human_baseline_details}.
The sample, aged 16--35, serves as a reference for task solvability rather than a population estimate.

\paragraph{Models and intended use.}
Open-weight models were evaluated under their published licenses (Table~\ref{tab:model_details}, Appendix~\ref{sec:app_model_details}), and frontier models under applicable API terms.
Our proposed benchmark is intended for diagnosing 3D reasoning behavior, not certifying deployment suitability.
Image-removal controls (\S\ref{sec:mcq}) demonstrate why accuracy should be considered alongside image dependence.
The benchmark uses only synthetic scenes, with no personal imagery or scraped content, and presents no identified dual-use concerns.

\bibliography{main}

\clearpage
\appendix
\setcounter{secnumdepth}{3}
\raggedbottom

\begin{center}
    {\BedBenchTitle \Large\bfseries\color{bedaccent} Supplementary Material}
\end{center}
\tableofcontents
\addtocontents{toc}{\protect\setcounter{tocdepth}{3}}
\newpage

\section{Extended Related Work}
\label{sec:extended_related_works}

Prior work motivates four questions that guide BedBench: which spatial tasks remain difficult for VLMs, whether answers to related questions agree, whether correct answers depend on visual evidence, and which task-relevant quantities can be recovered from visual features.
This appendix expands Section~\ref{sec:related} by connecting benchmark design to these questions, then reviewing the reasoning and training interventions examined in our experiments.
The central connection is between task performance and its interpretation: human comparisons establish the size of a gap, consistency checks test whether a recovered property is used, shortcut controls test image dependence, and probes test recovery under a specified readout.

\subsection{Benchmarks and Human References}

\paragraph{Controlled geometry makes task demands explicit.}
The benchmarks closest to BedBench use geometric objects to test specific spatial operations.
SPARE3D studies relations between orthographic and isometric drawings \cite{spare3d}; LEGO-Puzzles tests understanding and planning over brick assemblies \cite{legopuzzles}; and Inf-Bench varies the number of deformation steps applied to generated shapes \cite{infiniteladder}.
These settings reveal different difficulties: SPARE3D's convolutional baselines perform close to random selection on three-view-to-isometric matching, while LEGO-Puzzles reports declining model accuracy as assembly plans grow longer.
Our benchmark builds on this use of explicit geometry, with tasks that distinguish total block count, per-color counts, relations between parts, and cross-view matching (Section~\ref{sec:benchmark}).
This distinction matters for our results: a model can succeed on a height-threshold decision while failing to report every per-color count correctly (Section~\ref{sec:3d_blind}).

\paragraph{Broad suites establish the range of spatial difficulties.}
SITE and EASI bring together existing spatial evaluations, while OmniSpatial, Spatial-DISE, and SRBench cover diverse skills such as orientation, mental rotation, and multi-view reasoning \cite{site_spatial,easi_holistic,omnispatial,spatialdise,mindthegap}.
Their findings provide context for our human--VLM comparisons.
For example, OmniSpatial reports 56.3\% accuracy for its best evaluated reasoning model, compared with 92.6\% for humans; Spatial-DISE reports a mean of 28.4\% across 32 models, compared with 76.8\% for humans \cite{omnispatial,spatialdise}.
Video and multi-image benchmarks extend this evaluation to scene understanding across observations \cite{vsibench,mmsibench}, while spatial-relation benchmarks examine how models interpret positions and viewpoints \cite{liu2023visualspatialreasoning,kamath2023whatsupvisionlanguagemodels,ma20253dsrbenchcomprehensive3dspatial,xu2026spatialbenchbenchmarkingmultimodallarge,fu2024blink}.
These studies establish broad weaknesses; our benchmark complements them by comparing task demands and diagnostic conditions within a controlled family of rendered objects and scenes.

\paragraph{Cognitive tasks help distinguish different spatial operations.}
SPACE, 11Plus-Bench, and psychometric batteries adapt tests of human spatial ability, while SpatialViz-Bench and STARE examine visualization and spatial transformations \cite{space_cognition,elevenplusbench,psychometric_bsa,wang2026spatialvizbenchcognitivelygroundedbenchmarkdiagnosing,stare_unfolding}.
Their task-level analyses show why a single average can be misleading.
On STARE, for example, o1 achieves 81.8\% on 2D transformations but 51.3\% on cube-net folding, where humans achieve 99.0\% \cite{stare_unfolding}.
SpatialViz-Bench's error analysis attributes roughly 60\% of failures to perception and spatial transformation \cite{wang2026spatialvizbenchcognitivelygroundedbenchmarkdiagnosing}.
These observations motivate examining the structure of errors on BedBench, including the difference between counts with a small average error and a completely correct per-color answer (Section~\ref{sec:3d_blind} and Appendix~\ref{sec:app_secondary_metrics}).

\paragraph{Procedural generation supports fresh instances and controlled comparisons.}
Inf-Bench generates shape-deformation problems, InfiniBench constructs configurable 3D scenes, and DynaSolidGeo regenerates solid-geometry questions from expert-curated seeds \cite{infiniteladder,wang2025infinibench,wu2025dynasolidgeo}.
DynaSolidGeo reports lower performance on regenerated questions than on its static source questions, motivating evaluation beyond a fixed collection of examples \cite{wu2025dynasolidgeo}.
Other generators support different diagnostic targets: IDEAL-Bench evaluates structured indoor 3D layouts, while SpaMEM compares reasoning with oracle textual histories against maintaining spatial information from visual streams \cite{cai2026idealbench,liao2026spamem}.
MindEdit-Bench instead uses private indoor scenes and counterfactual object edits to test spatial predictions whose answers are not directly visible in the input \cite{yu2026mindedit}.
Our benchmark uses procedural geometry to provide both matched visual-cue variants and exact targets for behavioral evaluation and probing.
Generation enables these comparisons, but does not by itself eliminate answer priors or establish that a model uses the image; those questions require the controls below.

\paragraph{Human references make low model scores interpretable.}
Human comparisons are already an important part of spatial evaluation, although their protocols differ.
For example, SPARE3D reports results for untrained and trained participants \cite{spare3d}, whereas the psychometric study of \citet{psychometric_bsa} uses published human norms.
Scores across these studies are therefore contextual evidence, rather than measurements on a common scale or protocol.
We collect responses from 30 participants on the BedBench tasks and report per-task confidence intervals (Section~\ref{sec:experiments}).
This comparison identifies per-color counting as a particularly large gap, while also showing tasks on which models perform well (Section~\ref{sec:3d_blind}).
Human accuracy helps distinguish a model weakness from a task that is difficult for both groups; additional controls are needed to explain how the model reaches its answers.

\subsection{Consistency Across Related Questions}
\label{sec:app_rw_consistency}
\paragraph{Prior consistency checks detect disagreement between related answers.}
GQA generates questions from scene graphs of real images and uses their functional programs to identify entailed questions, measuring whether a model's answers to them agree \cite{hudson2019gqa}.
\citet{ribeiro2019redroses} automatically derive implied question--answer pairs for VQA and SQuAD and use them to expose inconsistencies in existing models, and ConVQA pairs entailed questions about visual facts to measure and improve consistency \cite{ray2019sunny}.
Among spatial benchmarks, LEGO-Puzzles checks whether successively generated assembly steps form a valid sequence \cite{legopuzzles}, which tests coherence across generated steps rather than agreement between answers to entailed questions, while SITE, SPACE, STARE, SPARE3D, Inf-Bench, and SpaMEM \cite{site_spatial,space_cognition,stare_unfolding,spare3d,infiniteladder,liao2026spamem} pose related tasks on the same items without analyzing whether the answers agree (Table~\ref{tab:benchmark_comparison}).
These measures show that related answers disagree, but not which answer is at fault or why.
\paragraph{BedBench locates where a failure arises.}
Because all tasks share block-level ground truth, the per-color counts of \tfive{} entail the answers to \tfour{} and \tseven{}.
Probing frozen features on the same targets tests whether the visual representation encodes a property (Appendix~\ref{sec:app_probing}); given its own \tfive{} counts as text, the fine-tuned model derives the \tseven{} answer almost perfectly when those counts are correct, while its direct visual answers often disagree (Appendix~\ref{sec:app_finetune}); and re-rendering the same blocks as 2D tiles tests whether the 3D presentation is the obstacle (Appendix~\ref{sec:app_2d_control}).

\subsection{Shortcut Controls and Answer Formats}

\paragraph{Correct answers do not necessarily require visual evidence.}
Visual question answering studies show that models can exploit language priors, while analyses of multiple-choice evaluation identify sensitivity to option positions and other answer artifacts \cite{goyal2017making,agrawal2018dont,zheng2024large,balepur2024artifacts}.
These are examples of shortcut learning, in which a predictive cue supports success without the intended reasoning process \cite{geirhos2020shortcut,lapuschkin2019unmasking}.
The concern extends to spatial tasks: SpatialEval reports settings in which models score higher without images \cite{spatialeval}.
Several spatial benchmarks instead replace the image with a text description \cite{space_cognition,stare_unfolding,infiniteladder,wang2026spatialvizbenchcognitivelygroundedbenchmarkdiagnosing}, which tests reasoning from text rather than reliance on answer priors.
We therefore measure accuracy after image removal and inspect the resulting predictions.
The constant answers documented in Section~\ref{sec:mcq} show how label frequencies can sustain a score even when no visual evidence is available.

\paragraph{Matched answer formats reveal what candidate options contribute.}
SpatiaLab evaluates spatial questions in both multiple-choice and open-ended formats and reports lower model accuracy in the open-ended setting \cite{wasi2026spatialab}.
Its human accuracy also decreases under this change, and its open-ended responses are evaluated by an LLM judge.
The comparison therefore demonstrates format sensitivity, while leaving several possible reasons for the score difference.
The BedBench volume task (\tone{}) permits an exact-match comparison between a generated integer answer and selection from five candidate answers on the same items.
We pair this format comparison with image removal: the no-image MCQ result in Section~\ref{sec:mcq} demonstrates that a large apparent improvement can occur without counting the depicted blocks.

\paragraph{Image ablation also tests interpretations of probe results.}
\citet{liao2026decodable} show that decodable spatial information and improved predictions need not reflect visual grounding: an intervention can retain its gain when the image is replaced with a gray blank.
Their analysis connects interpretability claims to the same image-dependence question raised by behavioral evaluation.
BedBench brings these established diagnostics together with generator-derived targets and matched cue comparisons (Table~\ref{tab:benchmark_comparison}).
The experiments have different coverage: image removal uses three models, and the open-ended/MCQ comparison is restricted to \tone{} (Section~\ref{sec:experiments}).
The contribution is the combined diagnostic setting, with each comparison supporting a specific interpretation of the reported scores.

\subsection{Visual Representations and the Choice of Readout}

\paragraph{A VLM can fail to report information that a probe recovers.}
Probing studies train small predictors on frozen features to test which targets those features make accessible.
Prior work finds recoverable information about elementary visual structure and spatial properties despite errors in end-to-end VLM answers \cite{rahmanzadehgervi2025visionlanguagemodelsblind,el2024probing}.
Related results cover block counts from orthographic views, stone counts on synthetic boards, and color-grid contents \cite{threeviewsense,counting_bottleneck,grid2matrix}.
These findings motivate our comparison between probe predictions and model answers: poor task accuracy alone does not establish that the visual representation lacks useful information.

\paragraph{Different analyses examine different possible sources of error.}
Work on embedding norms, attention to image regions, and arbitration between visual and linguistic signals investigates how models use visual information \cite{beyond_semantics,attention_spatial_focus,arbitration_failure}.
Other work examines the contribution of positional encoding within the visual encoder \cite{mulset}.
\citet{alam2026spatialblindspot,alam2026freelunch} trace spatial failures to CLIP-style encoder objectives and to flattening images into 1D token sequences; in controlled LLaVA variants, alternative encoder objectives and 2D positional encodings improve spatial reasoning but do not resolve it.
SIT-Bench approaches the reasoning component separately by supplying coordinate-aware textual scene descriptions to text-only LLMs, which still exhibit errors in global spatial consistency \cite{guo2026sitbench}.
These studies leave several plausible explanations for a spatial error, including feature encoding, access to the relevant features, and subsequent reasoning.
Our behavioral comparisons and frozen-feature probes examine complementary parts of this problem.

% \paragraph{BedBench compares aggregate quantities with exact per-part targets.}
% Our generator supplies both global quantities, such as total block count, and structured targets, such as the count assigned to each color.
% Across 14 frozen encoders, probes on mean-pooled patch features recover global count trends, but exact per-part answers remain poor; several relational targets are also difficult under these readouts (Section~\ref{sec:scaling}).
% The successful count readouts are consistent with prior counting studies, while the structured targets ask whether recovery extends to all components of an answer.
% The comparison does not establish a contradiction with earlier work, because both the prediction targets and the readout protocols differ.
% It shows why approximate numerical recovery and exact structural recovery should be evaluated separately.

\paragraph{Pooling limits what a failed probe can establish.}
\citet{li2025objectbinding} recover whether two patches belong to the same object using a quadratic probe on frozen patch embeddings.
\citet{michalkiewicz2026foveated} hold the encoder fixed and recover localized color--shape binding with an attention-based readout that selects relevant patches before pooling, whereas global average pooling performs poorly under clutter.
These results demonstrate that binding information can be accessible through one readout and difficult to recover through another.
They directly qualify the interpretation of our probes: poor performance from mean-pooled features does not prove that the original patch tokens lack the target information.
On \tfive{}, attention and token-sum readouts over patch tokens, trained on more objects, recover exact counts that pooled probes miss (Appendix~\ref{sec:app_token_probe}); for the other tasks, such readouts remain untested.

\subsection{Reasoning, Scaling, and Training Interventions}

\paragraph{Additional reasoning has task-dependent benefits.}
Chain-of-thought prompting can improve reasoning, but its benefits vary across tasks, including spatial tasks \cite{wei2022chain,kojima2022large,sprague2025cot,kancheti2026cot}.
Moreover, a written reasoning trace need not faithfully describe the computation that produced an answer \cite{turpin2023language,dziri2023faith}.
BedBench therefore separates enabling Thinking mode from increasing its token budget and reports the effects by task (Section~\ref{sec:thinking_paradox}).
The selective gains and persistent errors motivate this distinction, while the annotated traces provide descriptive evidence about recurring visual misreadings rather than a causal account of the model's computation.

\paragraph{Visual cues test whether additional spatial information helps.}
Human mental-rotation studies examine how color, reference axes, and viewpoint affect performance \cite{khooshabeh2008visual,branoff1998coordinate,parsons1995inability,wraga1999influence}.
These studies motivate our matched renderings: the geometry stays fixed while color or coordinate-axis overlays change.
The model results show modest color benefits and predominantly negative axis effects (Section~\ref{sec:scaling}).

\paragraph{Training successes elsewhere define the scope of our results.}
Prior work questions whether scaling current training recipes is sufficient for spatial reasoning \cite{zhang2025scalingbeyond}, while geometric surrogate tasks and the staged SpatialLadder curriculum demonstrate benefits from targeted training \cite{euclids_gift,li2026spatialladder}.
Comparisons of reinforcement learning and supervised fine-tuning also report differences in compositional generalization \cite{li2025compositionalgap}.
The scaling comparisons and the per-color counting fine-tuning experiment on BedBench examine which errors persist under the configurations we test (Sections~\ref{sec:3d_blind} and~\ref{sec:scaling}).
Fine-tuning on 800 in-distribution examples makes most complete per-color answers correct, but the model's direct answers about the dominant colors often disagree with those counts, although it derives them correctly when given its own counts as text (Appendix~\ref{sec:app_finetune}); this result applies to one model trained with its vision encoder frozen.
Together, the intervention results identify unresolved targets for future methods, while the human references, shortcut controls, and probes provide complementary ways to interpret any improvement.

\clearpage
\section{Extended Findings and Detailed Analyses}
\label{sec:app_extended_findings}

Throughout, the same pattern recurs: models get individual judgments right while the exact answer stays wrong, and recovering a property does not mean using it consistently.
This section extends the main findings in Section~\ref{sec:results} with the detailed ablations, figures, and tables that support the paper's central claims.
Table~\ref{tab:full_main_results} reports the full per-model accuracy matrix summarized in Section~\ref{sec:3d_blind}.

\begin{table*}[ht]
\centering
\resizebox{\textwidth}{!}{%
\begin{tabular}{llccccccccccccc|c}
\toprule
Model & Mode & \tone{} & \ttwo{} & \tthree{} & \tfour{} & \tfive{} & \tsix{} & \tseven{} & \teight{} & \tnine{} & \tten{} & \televen{} & \ttwelve{} & \tthirteen{} & Avg \\
\midrule
\rowcolor{gray!15} Random baseline & -- & -- & 50.0 & -- & -- & -- & 50.0 & -- & 50.0 & -- & 50.0 & 33.3 & 50 & 33.3 & 24.4 \\
\rowcolor{purple!15} Best constant & -- & 34.0 & 50.0 & 5.0 & 24.0 & 1.0 & 77.0 & 9.0 & 58.0 & 43.0 & 51.0 & 37.3 & 51.0 & 37.3 & 36.7 \\
\midrule
\rowcolor{orange!15} Human baseline & -- & 76.7 & 73.3 & 23.3 & 83.3 & 93.3 & 63.3 & 76.7 & 53.3 & 70.0 & 86.7 & 76.7 & 86.7 & 46.7 & 70.0 \\
\midrule
Qwen3.5-0.8B & NT & 33.4 & 40.0 & 0.0 & 6.0 & 0.0 & 47.6 & 0.0 & 58.0 & 45.6 & 49.0 & 37.3 & 51.0 & 37.3 & 31.2 \\
 & TH & 20.8 & 50.4 & 0.2 & 19.8 & 0.4 & 66.4 & 0.0 & 52.2 & 27.0 & 46.4 & 37.3 & 50.6 & 37.0 & 31.4 \\
Qwen3.5-2B & NT & 22.4 & 50.8 & 0.0 & 24.4 & 0.0 & 40.8 & 0.0 & 59.0 & 59.2 & 49.0 & 37.3 & 51.0 & 37.3 & 33.2 \\
 & TH & 14.6 & 57.0 & 0.0 & 25.8 & 0.0 & 62.2 & 0.0 & 54.2 & 43.4 & 49.0 & 39.6 & 51.0 & 36.7 & 33.3 \\
Qwen3.5-4B & NT & 34.0 & 52.0 & 0.0 & 20.2 & 0.0 & 86.2 & 6.0 & 58.0 & 74.0 & 55.4 & 42.5 & 50.6 & 37.3 & 39.7 \\
 & TH & 28.6 & 62.4 & 0.0 & 43.8 & 1.0 & 78.8 & 8.4 & 57.8 & 65.8 & 53.4 & 37.1 & 50.4 & 35.3 & 40.2 \\
Qwen3.5-9B & NT & 32.2 & 68.6 & \textbf{2.0} & 11.0 & 0.0 & 86.4 & 22.2 & 58.0 & 66.4 & 50.2 & 45.7 & 49.0 & 37.3 & 40.7 \\
 & TH & 33.2 & 65.4 & 0.2 & 45.2 & 1.0 & 86.8 & 21.4 & 58.8 & 68.0 & 56.8 & 41.7 & 49.6 & 36.2 & 43.4 \\
Qwen3.5-27B & NT & 39.8 & 79.2 & 0.0 & 45.2 & 3.0 & 89.6 & 18.0 & 57.0 & 71.8 & 55.2 & \textbf{55.7} & 57.2 & 36.2 & 46.8 \\
 & TH & 24.2 & 88.2 & 0.0 & 57.2 & 1.4 & 87.0 & 24.8 & 59.8 & 76.6 & 65.6 & 50.5 & 57.6 & 38.5 & 48.6 \\
Qwen3.5-35B-A3B & NT & 36.8 & 65.8 & 0.0 & 43.0 & 0.0 & 71.6 & 29.2 & 58.0 & 67.4 & 55.6 & 40.4 & 54.8 & 37.6 & 43.1 \\
 & TH & 22.6 & 77.0 & 0.0 & 58.2 & 0.6 & 82.4 & 24.4 & 58.2 & 69.0 & 65.6 & 41.5 & 53.6 & 37.9 & 45.5 \\
\midrule
Gemma4-E2B & NT & 8.2 & 59.0 & 0.0 & 11.8 & 0.0 & 77.6 & 6.4 & 57.0 & 50.6 & 53.0 & 36.3 & 52.0 & 36.9 & 34.5 \\
 & TH & 10.8 & 74.8 & 0.2 & 41.6 & 0.0 & 74.6 & 8.0 & 55.4 & 22.2 & 59.0 & 33.2 & 49.0 & 33.5 & 35.6 \\
Gemma4-E4B & NT & \textbf{40.0} & 55.0 & 0.0 & 24.8 & 1.0 & 77.0 & 14.0 & 58.0 & 51.0 & 51.0 & 41.5 & 48.0 & 35.3 & 38.2 \\
 & TH & 16.8 & 78.6 & 0.0 & 36.2 & 0.0 & 75.6 & 9.2 & 56.6 & 46.4 & 53.2 & 45.5 & 47.0 & 34.9 & 38.5 \\
Gemma4-26B-A4B & NT & 14.0 & 58.0 & 0.0 & 27.4 & 0.0 & 81.0 & 22.2 & \textbf{61.0} & 44.8 & 55.0 & 46.3 & 54.6 & 36.3 & 38.5 \\
 & TH & 15.2 & 85.8 & 0.0 & 55.0 & 0.0 & 73.4 & 19.8 & 50.4 & 55.4 & 56.8 & 45.2 & 48.2 & 34.3 & 41.5 \\
Gemma4-31B & NT & 35.0 & 76.0 & 0.0 & 52.8 & 4.0 & 83.0 & 32.0 & 56.0 & 61.0 & 61.4 & 47.3 & 54.0 & 42.3 & 46.5 \\
 & TH & 17.2 & \textbf{91.6} & 0.0 & 56.2 & \textbf{7.2} & 74.8 & 27.0 & 58.2 & 42.6 & 58.2 & 48.6 & 50.8 & 40.3 & 44.1 \\
\midrule
GLM-4.6V-Flash & NT & 22.2 & 51.0 & 0.0 & 38.2 & 0.0 & 78.0 & 17.6 & 58.0 & 61.2 & 53.0 & 45.4 & 48.0 & 37.3 & 39.2 \\
 & TH & 38.0 & 73.8 & 0.0 & 47.0 & 0.0 & \textbf{90.0} & 17.0 & 59.0 & 62.2 & 59.8 & 44.4 & 63.2 & 37.7 & 45.5 \\
\midrule
Kimi-VL-A3B & NT & 31.8 & 50.0 & 0.0 & 20.2 & 0.0 & 23.0 & 0.6 & 60.8 & 17.0 & 49.0 & 41.3 & 60.4 & 37.1 & 30.1 \\
 & TH & 0.6 & 66.4 & 0.0 & 6.0 & 0.0 & 23.0 & 3.6 & 43.6 & 2.0 & 49.0 & 37.3 & 53.4 & 37.3 & 24.8 \\
\midrule
\rowcolor{blue!07} GPT-5 (High) & TH & 14.0 & 88.0 & 0.0 & 64.0 & 3.0 & 80.0 & 24.0 & 60.0 & 51.0 & 65.0 & 33.0 & 61.0 & 40.3 & 44.9 \\
\rowcolor{blue!07} Grok-4.20 (High) & TH & 13.0 & 82.0 & 0.0 & 52.0 & 1.0 & 67.0 & 29.0 & 57.0 & 27.0 & 62.0 & 44.0 & 69.0 & 41.0 & 41.8 \\
\rowcolor{blue!07} Gemini-3.1-Pro (High) & TH & 18.0 & 82.0 & 1.0 & 66.0 & 6.0 & 79.0 & 35.0 & 51.0 & \textbf{77.0} & 57.0 & 39.3 & 57.0 & 42.0 & 46.9 \\
\rowcolor{blue!07} Gemini-3-Flash (High) & TH & 31.0 & 88.0 & 0.0 & \textbf{70.0} & 4.0 & 86.0 & \textbf{37.0} & 60.0 & 63.0 & \textbf{67.0} & 49.3 & \textbf{71.0} & \textbf{43.7} & \textbf{51.5} \\
\bottomrule
\end{tabular}

}
\caption{\textbf{Main results on BedBench.} Accuracy (\%) across all 13 tasks in their default settings. Each open-weight cell is the 5-run mean for one task$\times$model$\times$inference mode condition; \sethlcolor{blue!07}\hl{the frontier rows} are single-run TH results; \sethlcolor{orange!15}\hl{the human row} aggregates 30 participants (one question per task per participant; see Appendix~\ref{sec:human_baseline_details}). \sethlcolor{purple!15}\hl{The best-constant row} gives the accuracy of always outputting the most frequent ground-truth answer (or candidate position) for each task. \textbf{Bold} = best model accuracy per task and highest model average. NT = Non-Thinking, TH = Thinking mode. \textbf{Takeaway:} Humans average 70.0\%, while the strongest model reaches 51.5\%, 14.8\,pp above the best-constant average of 36.7\%; the gap is largest on \tfive{}, where humans reach 93.3\% and no model exceeds 7.2\%.}
\label{tab:full_main_results}
\end{table*}

% arXiv version of tables/table1_main_results.tex (same content) with a top gap, since \intextsep is 0pt here. Edit both copies when the table changes.
\begin{table*}[ht]
\vspace{10pt}
\centering
\resizebox{\textwidth}{!}{%
\begin{tabular}{llccccccccccccc|c}
\toprule
Model & Mode & \tone{} & \ttwo{} & \tthree{} & \tfour{} & \tfive{} & \tsix{} & \tseven{} & \teight{} & \tnine{} & \tten{} & \televen{} & \ttwelve{} & \tthirteen{} & Avg \\
\midrule
\rowcolor{gray!15} Random baseline & -- & -- & 50.0 & -- & -- & -- & 50.0 & -- & 50.0 & -- & 50.0 & 33.3 & 50 & 33.3 & 24.4 \\
\rowcolor{purple!15} Best constant & -- & 34.0 & 50.0 & 5.0 & 24.0 & 1.0 & 77.0 & 9.0 & 58.0 & 43.0 & 51.0 & 37.3 & 51.0 & 37.3 & 36.7 \\
\midrule
\rowcolor{orange!15} Human baseline & -- & 76.7 & 73.3 & 23.3 & 83.3 & 93.3 & 63.3 & 76.7 & 53.3 & 70.0 & 86.7 & 76.7 & 86.7 & 46.7 & 70.0 \\
\midrule
Qwen3.5-0.8B & NT & 33.4 & 40.0 & 0.0 & 6.0 & 0.0 & 47.6 & 0.0 & 58.0 & 45.6 & 49.0 & 37.3 & 51.0 & 37.3 & 31.2 \\
 & TH & 20.8 & 50.4 & 0.2 & 19.8 & 0.4 & 66.4 & 0.0 & 52.2 & 27.0 & 46.4 & 37.3 & 50.6 & 37.0 & 31.4 \\
Qwen3.5-2B & NT & 22.4 & 50.8 & 0.0 & 24.4 & 0.0 & 40.8 & 0.0 & 59.0 & 59.2 & 49.0 & 37.3 & 51.0 & 37.3 & 33.2 \\
 & TH & 14.6 & 57.0 & 0.0 & 25.8 & 0.0 & 62.2 & 0.0 & 54.2 & 43.4 & 49.0 & 39.6 & 51.0 & 36.7 & 33.3 \\
Qwen3.5-4B & NT & 34.0 & 52.0 & 0.0 & 20.2 & 0.0 & 86.2 & 6.0 & 58.0 & 74.0 & 55.4 & 42.5 & 50.6 & 37.3 & 39.7 \\
 & TH & 28.6 & 62.4 & 0.0 & 43.8 & 1.0 & 78.8 & 8.4 & 57.8 & 65.8 & 53.4 & 37.1 & 50.4 & 35.3 & 40.2 \\
Qwen3.5-9B & NT & 32.2 & 68.6 & \textbf{2.0} & 11.0 & 0.0 & 86.4 & 22.2 & 58.0 & 66.4 & 50.2 & 45.7 & 49.0 & 37.3 & 40.7 \\
 & TH & 33.2 & 65.4 & 0.2 & 45.2 & 1.0 & 86.8 & 21.4 & 58.8 & 68.0 & 56.8 & 41.7 & 49.6 & 36.2 & 43.4 \\
Qwen3.5-27B & NT & 39.8 & 79.2 & 0.0 & 45.2 & 3.0 & 89.6 & 18.0 & 57.0 & 71.8 & 55.2 & \textbf{55.7} & 57.2 & 36.2 & 46.8 \\
 & TH & 24.2 & 88.2 & 0.0 & 57.2 & 1.4 & 87.0 & 24.8 & 59.8 & 76.6 & 65.6 & 50.5 & 57.6 & 38.5 & 48.6 \\
Qwen3.5-35B-A3B & NT & 36.8 & 65.8 & 0.0 & 43.0 & 0.0 & 71.6 & 29.2 & 58.0 & 67.4 & 55.6 & 40.4 & 54.8 & 37.6 & 43.1 \\
 & TH & 22.6 & 77.0 & 0.0 & 58.2 & 0.6 & 82.4 & 24.4 & 58.2 & 69.0 & 65.6 & 41.5 & 53.6 & 37.9 & 45.5 \\
\midrule
Gemma4-E2B & NT & 8.2 & 59.0 & 0.0 & 11.8 & 0.0 & 77.6 & 6.4 & 57.0 & 50.6 & 53.0 & 36.3 & 52.0 & 36.9 & 34.5 \\
 & TH & 10.8 & 74.8 & 0.2 & 41.6 & 0.0 & 74.6 & 8.0 & 55.4 & 22.2 & 59.0 & 33.2 & 49.0 & 33.5 & 35.6 \\
Gemma4-E4B & NT & \textbf{40.0} & 55.0 & 0.0 & 24.8 & 1.0 & 77.0 & 14.0 & 58.0 & 51.0 & 51.0 & 41.5 & 48.0 & 35.3 & 38.2 \\
 & TH & 16.8 & 78.6 & 0.0 & 36.2 & 0.0 & 75.6 & 9.2 & 56.6 & 46.4 & 53.2 & 45.5 & 47.0 & 34.9 & 38.5 \\
Gemma4-26B-A4B & NT & 14.0 & 58.0 & 0.0 & 27.4 & 0.0 & 81.0 & 22.2 & \textbf{61.0} & 44.8 & 55.0 & 46.3 & 54.6 & 36.3 & 38.5 \\
 & TH & 15.2 & 85.8 & 0.0 & 55.0 & 0.0 & 73.4 & 19.8 & 50.4 & 55.4 & 56.8 & 45.2 & 48.2 & 34.3 & 41.5 \\
Gemma4-31B & NT & 35.0 & 76.0 & 0.0 & 52.8 & 4.0 & 83.0 & 32.0 & 56.0 & 61.0 & 61.4 & 47.3 & 54.0 & 42.3 & 46.5 \\
 & TH & 17.2 & \textbf{91.6} & 0.0 & 56.2 & \textbf{7.2} & 74.8 & 27.0 & 58.2 & 42.6 & 58.2 & 48.6 & 50.8 & 40.3 & 44.1 \\
\midrule
GLM-4.6V-Flash & NT & 22.2 & 51.0 & 0.0 & 38.2 & 0.0 & 78.0 & 17.6 & 58.0 & 61.2 & 53.0 & 45.4 & 48.0 & 37.3 & 39.2 \\
 & TH & 38.0 & 73.8 & 0.0 & 47.0 & 0.0 & \textbf{90.0} & 17.0 & 59.0 & 62.2 & 59.8 & 44.4 & 63.2 & 37.7 & 45.5 \\
\midrule
Kimi-VL-A3B & NT & 31.8 & 50.0 & 0.0 & 20.2 & 0.0 & 23.0 & 0.6 & 60.8 & 17.0 & 49.0 & 41.3 & 60.4 & 37.1 & 30.1 \\
 & TH & 0.6 & 66.4 & 0.0 & 6.0 & 0.0 & 23.0 & 3.6 & 43.6 & 2.0 & 49.0 & 37.3 & 53.4 & 37.3 & 24.8 \\
\midrule
\rowcolor{blue!07} GPT-5 (High) & TH & 14.0 & 88.0 & 0.0 & 64.0 & 3.0 & 80.0 & 24.0 & 60.0 & 51.0 & 65.0 & 33.0 & 61.0 & 40.3 & 44.9 \\
\rowcolor{blue!07} Grok-4.20 (High) & TH & 13.0 & 82.0 & 0.0 & 52.0 & 1.0 & 67.0 & 29.0 & 57.0 & 27.0 & 62.0 & 44.0 & 69.0 & 41.0 & 41.8 \\
\rowcolor{blue!07} Gemini-3.1-Pro (High) & TH & 18.0 & 82.0 & 1.0 & 66.0 & 6.0 & 79.0 & 35.0 & 51.0 & \textbf{77.0} & 57.0 & 39.3 & 57.0 & 42.0 & 46.9 \\
\rowcolor{blue!07} Gemini-3-Flash (High) & TH & 31.0 & 88.0 & 0.0 & \textbf{70.0} & 4.0 & 86.0 & \textbf{37.0} & 60.0 & 63.0 & \textbf{67.0} & 49.3 & \textbf{71.0} & \textbf{43.7} & \textbf{51.5} \\
\bottomrule
\end{tabular}

}
\caption{\textbf{Main results on \BedBenchLogo.} Accuracy (\%) across all 13 tasks in their default settings. Each open-weight cell is the 5-run mean for one task$\times$model$\times$inference mode condition; \sethlcolor{blue!07}\hl{the frontier rows} are single-run TH results; \sethlcolor{orange!15}\hl{the human row} aggregates 30 participants (one question per task per participant; see Appendix~\ref{sec:human_baseline_details}). \sethlcolor{purple!15}\hl{The best-constant row} gives the accuracy of always outputting the most frequent ground-truth answer (or candidate position) for each task. \textbf{Bold} = best model accuracy per task and highest model average. NT = Non-Thinking, TH = Thinking mode. \textbf{Takeaway:} Humans average 70.0\%, while the strongest model reaches 51.5\%, 14.8\,pp above the best-constant average of 36.7\%; the gap is largest on \tfive{}, where humans reach 93.3\% and no model exceeds 7.2\%.}
\label{tab:full_main_results}
\end{table*}

\subsection{Human gap and per-color counting}
\label{sec:app_model_human_training}

\paragraph{Overview.} This section expands Section~\ref{sec:3d_blind}. It asks whether the exact-recovery failures reflect a limitation specific to current VLMs or a difficulty intrinsic to the tasks, characterizes the per-color counting errors on \tfivelong{}, and reports three controls: factorized querying, a 2D visibility control, and task-specific fine-tuning with tests of consistent use, including a per-object check of whether each model's direct answers agree with its own counts.

\paragraph{Experimental setup.} We combine a human reference, error-magnitude metrics, and three controls on \tfivelong{}.
\begin{itemize}[leftmargin=1.4em,nosep]
    \item \textbf{Human baseline.} Thirty participants each answer one default-setting question per task (Appendix~\ref{sec:human_baseline_details}); we report per-task human accuracy with Wilson 95\% CIs alongside the best OW and best frontier cell from Table~\ref{tab:full_main_results} (Table~\ref{tab:human_gap}), and compute the Spearman rank correlation between human and best-VLM accuracy across the 13 tasks.
    \item \textbf{Error metrics.} Error-magnitude metrics (MAE, RMSE) are unweighted averages over the 12 OW models under the default visual setting, computed separately for Non-Thinking (NT) and Thinking (TH) modes; each model $\times$ mode cell is a 5-run mean.
    \item \textbf{Controls.} Factorized querying (Appendix~\ref{sec:app_t5_factorized}), a 2D visibility control (Appendix~\ref{sec:app_2d_control}), and task-specific fine-tuning (Appendix~\ref{sec:app_finetune}); each subsection gives its own setup.
\end{itemize}

\subsubsection{VLM-human gap concentrates on exact structure and cross-view tasks}
\label{sec:app_human_gap}

Table~\ref{tab:human_gap} reports per-task human accuracy and its gap to the best OW and best frontier cell.

% arXiv version of tables/human_gap.tex (same content), wrapfig as in arXiv:2608.07645.
\begin{wraptable}{L}{0.5\textwidth}
\centering
\setlength{\tabcolsep}{4pt}
\scriptsize
\begin{adjustbox}{max width=\linewidth}
\begin{tabular}{lcccc}
\toprule
Task & Random & \makecell[c]{Best\\OW} & \makecell[c]{Best\\Frontier} & \makecell[c]{Human\\(95\% CI)} \\
\midrule
\midrule
 \tfivelong{} & 0.0 & 7.2 & 6.0 & 93.3 [78.7, 98.2] \\
 \tsevenlong{}    & 0.0 & 32.0 & 37.0 & 76.7 [59.1, 88.2] \\
 \tonelong{}             & 0.0 & 40.0 & 31.0 & 76.7 [59.1, 88.2] \\
 \tfourlong{}        & 0.0 & 58.2 & 70.0 & 83.3 [66.4, 92.7] \\
 \ttwelvelong{}       & 50.0 & 63.2 & 71.0 & 86.7 [70.3, 94.7] \\
 \tthreelong{}         & 0.0 & 2.0 & 1.0 & 23.3 [11.8, 40.9] \\
 \ttenlong{}   & 50.0 & 65.6 & 67.0 & 86.7 [70.3, 94.7] \\
 \televenlong{}        & 33.3 & 55.7 & 49.3 & 76.7 [59.1, 88.2] \\
\midrule
 \tthirteenlong{} & 33.3 & 42.3 & 43.7 & 46.7 [30.2, 63.9] \\
 \tninelong{}     & 0.0 & 76.6 & 77.0 & 70.0 [52.1, 83.3] \\
 \teightlong{} & 50.0 & 61.0 & 60.0 & 53.3 [36.1, 69.8] \\
\midrule
 \ttwolong{}          & 50.0 & 91.6 & 88.0 & 73.3 [55.6, 85.8] \\
 \tsixlong{}    & 50.0 & 90.0 & 86.0 & 63.3 [45.5, 78.1] \\
\midrule
\midrule
\textit{Average}       & 24.4 & 52.7 & 52.8 & 70.0 \\
\bottomrule
\end{tabular}
\end{adjustbox}
% \caption{Per-task accuracy (\%). \textbf{Best OW} selects the highest five-run mean across 12 models and both inference modes; \textbf{Best Frontier} selects the highest single-run score across four systems (full results in Table~\ref{tab:full_main_results}, Appendix~\ref{sec:app_extended_findings}). \textbf{Human} averages 30 responses per task, with 95\% Wilson confidence intervals. Horizontal rules group tasks by whether Best OW falls below, within, or above the human interval. The Average row summarizes the displayed task scores; the Best OW and Best Frontier averages combine per-task best results, which may come from different models. Figure~\ref{fig:human_vlm_gap} plots the gaps.}
\caption{\textbf{Human vs.\ best-model accuracy per task (\%).} \textbf{Best OW} selects the highest five-run mean across 12 models and both inference modes; \textbf{Best Frontier} selects the highest single-run score across four systems (full results in Table~\ref{tab:full_main_results}, Appendix~\ref{sec:app_extended_findings}). \textbf{Human} averages 30 responses per task, with 95\% Wilson confidence intervals. Horizontal rules group tasks by whether Best OW falls below, within, or above the human interval. Best OW and Best Frontier averages combine per-task bests, which may come from different models. Figure~\ref{fig:human_vlm_gap} plots the gaps. \textbf{Takeaway:} The best OW model falls below the human interval on eight tasks, most on \tfive{} (7.2\% vs.\ 93.3\%), and exceeds it only on \ttwo{} and \tsix{}.}
\label{tab:human_gap}
\vspace{-1em}
\end{wraptable}

\paragraph{\textcolor{yellow}{\faLightbulb} The gap concentrates on tasks that require reading out a full 3D structure.}
The largest margins appear where the answer is a full structure rather than a single number. On \tfivelong{}, humans exceed the best OW cell by \textcolor{custom_darkgreen}{$+$86.1\,pp} and the best frontier cell by \textcolor{custom_darkgreen}{$+$87.3\,pp}, the largest separation. Across the two VLM references, humans also lead on \tonelong{} and \tsevenlong{} by \textcolor{custom_darkgreen}{$+$37} to \textcolor{custom_darkgreen}{$+$46\,pp}, on the cross-view tasks \televenlong{} and \ttwelvelong{} by \textcolor{custom_darkgreen}{$+$16} to \textcolor{custom_darkgreen}{$+$27\,pp}, and on \tthreelong{}, although human accuracy there is itself low (23.3\%). Both evaluated model families fail on these same tasks (Appendix~\ref{sec:app_cross_family}).

\begin{table}[t]
\centering
\setlength{\tabcolsep}{4pt}
\scriptsize
\begin{adjustbox}{max width=\columnwidth}
\begin{tabular}{lcccc}
\toprule
Task & Random & \makecell[c]{Best\\OW} & \makecell[c]{Best\\Frontier} & \makecell[c]{Human\\(95\% CI)} \\
\midrule
\midrule
 \tfivelong{} & 0.0 & 7.2 & 6.0 & 93.3 [78.7, 98.2] \\
 \tsevenlong{}    & 0.0 & 32.0 & 37.0 & 76.7 [59.1, 88.2] \\
 \tonelong{}             & 0.0 & 40.0 & 31.0 & 76.7 [59.1, 88.2] \\
 \tfourlong{}        & 0.0 & 58.2 & 70.0 & 83.3 [66.4, 92.7] \\
 \ttwelvelong{}       & 50.0 & 63.2 & 71.0 & 86.7 [70.3, 94.7] \\
 \tthreelong{}         & 0.0 & 2.0 & 1.0 & 23.3 [11.8, 40.9] \\
 \ttenlong{}   & 50.0 & 65.6 & 67.0 & 86.7 [70.3, 94.7] \\
 \televenlong{}        & 33.3 & 55.7 & 49.3 & 76.7 [59.1, 88.2] \\
\midrule
 \tthirteenlong{} & 33.3 & 42.3 & 43.7 & 46.7 [30.2, 63.9] \\
 \tninelong{}     & 0.0 & 76.6 & 77.0 & 70.0 [52.1, 83.3] \\
 \teightlong{} & 50.0 & 61.0 & 60.0 & 53.3 [36.1, 69.8] \\
\midrule
 \ttwolong{}          & 50.0 & 91.6 & 88.0 & 73.3 [55.6, 85.8] \\
 \tsixlong{}    & 50.0 & 90.0 & 86.0 & 63.3 [45.5, 78.1] \\
\midrule
\midrule
\textit{Average}       & 24.4 & 52.7 & 52.8 & 70.0 \\
\bottomrule
\end{tabular}
\end{adjustbox}
% \caption{Per-task accuracy (\%). \textbf{Best OW} selects the highest five-run mean across 12 models and both inference modes; \textbf{Best Frontier} selects the highest single-run score across four systems (full results in Table~\ref{tab:full_main_results}, Appendix~\ref{sec:app_extended_findings}). \textbf{Human} averages 30 responses per task, with 95\% Wilson confidence intervals. Horizontal rules group tasks by whether Best OW falls below, within, or above the human interval. The Average row summarizes the displayed task scores; the Best OW and Best Frontier averages combine per-task best results, which may come from different models. Figure~\ref{fig:human_vlm_gap} plots the gaps.}
\caption{\textbf{Human vs.\ best-model accuracy per task (\%).} \textbf{Best OW} selects the highest five-run mean across 12 models and both inference modes; \textbf{Best Frontier} selects the highest single-run score across four systems (full results in Table~\ref{tab:full_main_results}, Appendix~\ref{sec:app_extended_findings}). \textbf{Human} averages 30 responses per task, with 95\% Wilson confidence intervals. Horizontal rules group tasks by whether Best OW falls below, within, or above the human interval. Best OW and Best Frontier averages combine per-task bests, which may come from different models. Figure~\ref{fig:human_vlm_gap} plots the gaps. \textbf{Takeaway:} The best OW model falls below the human interval on eight tasks, most on \tfive{} (7.2\% vs.\ 93.3\%), and exceeds it only on \ttwo{} and \tsix{}.}
\label{tab:human_gap}
\end{table}

\paragraph{\textcolor{yellow}{\faLightbulb} Where VLMs exceed the human interval, the task is a binary decision.}
The best OW cell exceeds the human mean by \textcolor{custom_darkgreen}{$+$26.7\,pp} on \tsixlong{} and \textcolor{custom_darkgreen}{$+$18.3\,pp} on \ttwolong{}. \tsix{} belongs to the same \emph{Compositional decomposition} group as \tthree{} and \tfive{}, but it only asks whether one column exceeds a threshold and requires no per-part decomposition.

\paragraph{\textcolor{yellow}{\faLightbulb} The human$-$VLM gap is concentrated, not spread across a shared difficulty axis.}
Per-task human and VLM accuracies show no detectable association (Spearman $\rho = -0.01$ against the best OW cell and $\rho = 0.02$ against the best frontier cell), but with 13 tasks a bootstrap 95\% CI spans roughly $[-0.7, 0.7]$, so we rely on the distribution of the gap instead. Humans lead on 9 of 13 tasks, yet \tfive{}, \tseven{}, and \tone{} account for 59\% of the total positive gap, while the best OW cell exceeds the human mean on \tninelong{}, \teightlong{}, \ttwo{}, and \tsix{}. A uniformly hard benchmark would give a gap of similar sign and size across tasks, not 93.3\% human accuracy on \tfive{} alongside a best-OW score 26.7\,pp above humans on \tsix{}. The small-gap rows (\tnine{}, \teight{}, \tthirteenlong{}) lie within CI overlap, whereas the \tfive{} lower bound of 78.7\% remains far above every VLM cell (best overall 7.2\%, Gemma4-31B Thinking), so the selective gap survives the uncertainty of single-trial human estimates.

\subsubsection{Floor accuracy hides two failure depths}
\label{sec:app_floor_regimes}
\label{sec:app_secondary_metrics}

\paragraph{\textcolor{yellow}{\faLightbulb} \tfivelong{} counts are often wrong; \tthreelong{} errors are larger still.}
\tthree{} and \tfive{} are the two lowest-accuracy tasks (Table~\ref{tab:full_main_results}), yet their errors differ in size (Table~\ref{tab:sec_floor}). The \tfive{} MAE averages over all 12 palette colors, including absent colors that models mostly predict correctly as zero. On the colors present in the object, MAE is 1.02 blocks (Non-Thinking) against a mean true count of 1.48, and only 32.5\% are counted exactly. Over-counts are about 3.6 times as frequent as under-counts (52.7\%~vs.\ 14.8\% of present colors); one possible explanation is that blocks visible in both views are counted twice, which we do not test directly.
Exact match is low because errors occur in many color fields. These errors are positively correlated, so exact match is in fact higher than independence would predict: the observed per-color error rates would give an expected palette-vector EM of only 0.07\%, compared with the observed 0.68\%. Allowing every color to be off by one block raises exact match only to 20.9\%, and the palette MAE (0.80) exceeds that of always predicting each color's mean count (0.69); only five models do better, led by Gemma4-31B (0.41) and Qwen3.5-27B (0.50). On the aggregate count tasks, by contrast, the best OW cells beat a constant median prediction (MAE 0.82 vs.\ 1.04 on \tone{}, 0.44 vs.\ 1.22 on \tfour{}, and 0.25 vs.\ 0.65 on \tnine{}). On \tthree{}, errors are larger still (MAE 3.70 over its three components): when two objects must be compared across views, the answers are far from the right structure.

\begin{table}[t]
\centering
\small
\setlength{\tabcolsep}{4pt}
\begin{tabular}{llccc}
\toprule
Task & Mode & Acc (\%) & RMSE & MAE \\
\midrule
\multirow{2}{*}{\tthreelong{}} & NT & 0.2 & 4.14 & 3.70 \\
 & TH & 0.1 & 4.19 & 3.58 \\
\midrule
\multirow{2}{*}{\tfivelong{}} & NT & 0.7 & 1.31 & 0.80 \\
 & TH & 1.0 & 1.33 & 0.74 \\
\bottomrule
\end{tabular}
\caption{\textbf{Exact match vs.\ error magnitude on the numeric floor tasks.} Averaged over 12 OW models under the default setting; each cell is a 5-run mean. \tthree{} MAE and RMSE average its three components; \tfive{} MAE and RMSE average all 12 palette colors, including colors absent from the object. \textbf{Takeaway:} Both tasks are near zero under exact match (both $\le$1\%), while \tthree{} errors are larger (RMSE $3.2\times$, MAE $4.6\times$).}
\label{tab:sec_floor}
\end{table}

% \paragraph{\textcolor{yellow}{\faLightbulb} Fine-tuning separates error size from exact correctness on \tfivelong{}.} The same contrast appears in the fine-tuning experiment (Appendix~\ref{sec:app_finetune}). Task-specific LoRA shrinks the \tfive{} error magnitude (\textcolor{red!70!black}{$-$35\%} MAE) while exact match rises only from 3.0\% to 9.4\%. Supervision therefore makes the counts more accurate without making complete answers reliable, which argues against a pure data-coverage explanation and is consistent with a limitation in combining estimates into exact answers or in accessing the required visual information. Operationally, this identifies \tfive{} as the cleanest probe of exact compositional answers in BED, and \tthree{} as its probe of cross-view integration.

\subsubsection{Factorized querying does not consistently improve fine-grained color counting}
\label{sec:app_t5_factorized}

% arXiv version of tables/sec_floor.tex (same content), wrapfig as in arXiv:2608.07645.
\begin{wraptable}{r}{0.5\textwidth}
\centering
\small
\setlength{\tabcolsep}{4pt}
\begin{tabular}{llccc}
\toprule
Task & Mode & Acc (\%) & RMSE & MAE \\
\midrule
\multirow{2}{*}{\tthreelong{}} & NT & 0.2 & 4.14 & 3.70 \\
 & TH & 0.1 & 4.19 & 3.58 \\
\midrule
\multirow{2}{*}{\tfivelong{}} & NT & 0.7 & 1.31 & 0.80 \\
 & TH & 1.0 & 1.33 & 0.74 \\
\bottomrule
\end{tabular}
\caption{\textbf{Exact match vs.\ error magnitude on the numeric floor tasks.} Averaged over 12 OW models under the default setting; each cell is a 5-run mean. \tthree{} MAE and RMSE average its three components; \tfive{} MAE and RMSE average all 12 palette colors, including colors absent from the object. \textbf{Takeaway:} Both tasks are near zero under exact match (both $\le$1\%), while \tthree{} errors are larger (RMSE $3.2\times$, MAE $4.6\times$).}
\label{tab:sec_floor}
\end{wraptable}

\paragraph{Experimental setup.} We query each of the 12 OW models (Non-Thinking, default setting) with twelve independent single-color questions per \tfivelong{} instance over five runs (72,000 prompts) and merge the answers into a count vector. Present-color accuracy is the share of the 766 color--instance pairs with a positive count that are counted exactly, pooled over runs without majority voting; a color missing from a joint answer counts as zero, and no parse failures occurred.
To measure how exactness degrades as more counts must be correct together, we compute for each instance and run the expected exact match of a random subset of $k$ colors,
\[
    \operatorname{EM}_i(k)=\frac{\binom{c_i}{k}}{\binom{m_i}{k}},
    \qquad 1\leq k\leq m_i,
\]
where $m_i$ is the number of eligible colors (all 12, or only those present) and $c_i$ the number counted correctly (the numerator is zero when $c_i<k$), and average over instances, runs, and models. This is a descriptive analysis of saved predictions, not a new prompting experiment. Shaded bands in Figures~\ref{fig:t5_k_color_present} and~\ref{fig:t5_k_color_all} are pointwise 95\% intervals from 10,000 bootstrap resamples of objects, keeping all models and runs of an object together.

\paragraph{\textcolor{yellow}{\faLightbulb} On average, factorization leaves present-color accuracy nearly unchanged.}
Present-color accuracy changes from 32.52\% under joint prediction to 33.16\% under factorized querying (\textcolor{custom_darkgreen}{$+0.64$\,pp}), but the effect varies by model: five models improve and seven decline, from \textcolor{custom_darkgreen}{$+26.06$\,pp} (Qwen3.5-0.8B) to \textcolor{red!70!black}{$-32.58$\,pp} (Gemma4-E4B) (Table~\ref{tab:t5_factorized_models}).

\begin{table}[t]
\centering
\small
\setlength{\tabcolsep}{5pt}
\begin{tabular}{@{}lrrr@{}}
\toprule
Model & Joint (\%) & Factorized (\%) & $\Delta$ (pp) \\
\midrule
Qwen3.5-0.8B & 44.49 & 70.55 & \textcolor{custom_darkgreen}{$+26.06$} \\
Qwen3.5-2B & 26.63 & 26.61 & \textcolor{red!70!black}{$-0.03$} \\
Qwen3.5-4B & 35.04 & 35.74 & \textcolor{custom_darkgreen}{$+0.70$} \\
Qwen3.5-9B & 15.35 & 34.07 & \textcolor{custom_darkgreen}{$+18.72$} \\
Qwen3.5-27B & 48.22 & 47.49 & \textcolor{red!70!black}{$-0.73$} \\
Qwen3.5-35B-A3B & 30.13 & 36.92 & \textcolor{custom_darkgreen}{$+6.79$} \\
\midrule
Gemma4-E2B & 37.02 & 28.33 & \textcolor{red!70!black}{$-8.69$} \\
Gemma4-E4B & 41.85 & 9.27 & \textcolor{red!70!black}{$-32.58$} \\
Gemma4-26B-A4B & 17.65 & 15.74 & \textcolor{red!70!black}{$-1.91$} \\
Gemma4-31B & 53.55 & 52.53 & \textcolor{red!70!black}{$-1.02$} \\
\midrule
GLM-4.6V-Flash & 28.85 & 27.62 & \textcolor{red!70!black}{$-1.23$} \\
Kimi-VL-A3B & 11.44 & 13.05 & \textcolor{custom_darkgreen}{$+1.62$} \\
\midrule
\textbf{Mean} & \textbf{32.52} & \textbf{33.16} & \textcolor{custom_darkgreen}{$\mathbf{+0.64}$} \\
\bottomrule
\end{tabular}
\caption{\textbf{Joint vs.\ factorized querying on \tfivelong{}.} Present-color exact-count accuracy (\%) under Non-Thinking mode with color inputs, micro-averaged over color--instance pairs with ground-truth count $>0$ across five runs per model. $\Delta$ denotes factorized minus joint accuracy in percentage points. \textbf{Takeaway:} Factorization improves five models and worsens seven, yielding only a \textcolor{custom_darkgreen}{$+0.64$\,pp} average change across the 12 models.}
\label{tab:t5_factorized_models}
\end{table}
% arXiv only: tables/t5_factorized_models.tex (Table) and figures/fig_t5_k_color_all.tex (Figure) side by side,
% same content as the ACL files (edit both when a caption or table changes). The figure image height is computed
% so that both columns, captions included, end at the same height (any remainder becomes space above the caption).
\begin{figure}[t]
\setbox0=\hbox{\begin{minipage}[t]{0.45\textwidth}
\vspace{0pt}
\centering
\small
\setlength{\tabcolsep}{3pt}
\captionsetup{position=bottom}
\begin{tabular}{@{}lrrr@{}}
\toprule
Model & Joint (\%) & Factorized (\%) & $\Delta$ (pp) \\
\midrule
Qwen3.5-0.8B & 44.49 & 70.55 & \textcolor{custom_darkgreen}{$+26.06$} \\
Qwen3.5-2B & 26.63 & 26.61 & \textcolor{red!70!black}{$-0.03$} \\
Qwen3.5-4B & 35.04 & 35.74 & \textcolor{custom_darkgreen}{$+0.70$} \\
Qwen3.5-9B & 15.35 & 34.07 & \textcolor{custom_darkgreen}{$+18.72$} \\
Qwen3.5-27B & 48.22 & 47.49 & \textcolor{red!70!black}{$-0.73$} \\
Qwen3.5-35B-A3B & 30.13 & 36.92 & \textcolor{custom_darkgreen}{$+6.79$} \\
\midrule
Gemma4-E2B & 37.02 & 28.33 & \textcolor{red!70!black}{$-8.69$} \\
Gemma4-E4B & 41.85 & 9.27 & \textcolor{red!70!black}{$-32.58$} \\
Gemma4-26B-A4B & 17.65 & 15.74 & \textcolor{red!70!black}{$-1.91$} \\
Gemma4-31B & 53.55 & 52.53 & \textcolor{red!70!black}{$-1.02$} \\
\midrule
GLM-4.6V-Flash & 28.85 & 27.62 & \textcolor{red!70!black}{$-1.23$} \\
Kimi-VL-A3B & 11.44 & 13.05 & \textcolor{custom_darkgreen}{$+1.62$} \\
\midrule
\textbf{Mean} & \textbf{32.52} & \textbf{33.16} & \textcolor{custom_darkgreen}{$\mathbf{+0.64}$} \\
\bottomrule
\end{tabular}
\captionof{table}{\textbf{Joint vs.\ factorized querying on \tfivelong{}.} Present-color exact-count accuracy (\%) under Non-Thinking mode with color inputs, micro-averaged over color--instance pairs with ground-truth count $>0$ across five runs per model. $\Delta$ denotes factorized minus joint accuracy in percentage points. \textbf{Takeaway:} Factorization improves five models and worsens seven, yielding only a \textcolor{custom_darkgreen}{$+0.64$\,pp} average change across the 12 models.}
\label{tab:t5_factorized_models}
\end{minipage}}%
\setbox2=\hbox{\begin{minipage}[b]{0.53\textwidth}
\captionsetup{position=bottom}
\captionof{figure}{\textbf{All-color subset exact match on \tfivelong{}.} Expected EM over $k$ counts from the full 12-color palette, including zero counts for absent colors, using the same 100 objects at every $k$. Lines average 12 OW models and five runs under Non-Thinking mode; shading gives pointwise 95\% object-bootstrap intervals. Present-color curves appear in Figure~\ref{fig:t5_k_color_present}. \textbf{Takeaway:} Joint outputs outperform factorized outputs across the full palette, alongside their higher absent-color accuracy.}
\label{fig:t5_k_color_all}
\end{minipage}}%
\dimen0=\dimexpr\ht0+\dp0-\ht2-\dp2\relax
\setbox4=\hbox{\includegraphics[width=0.53\textwidth,height=\dimen0,keepaspectratio]{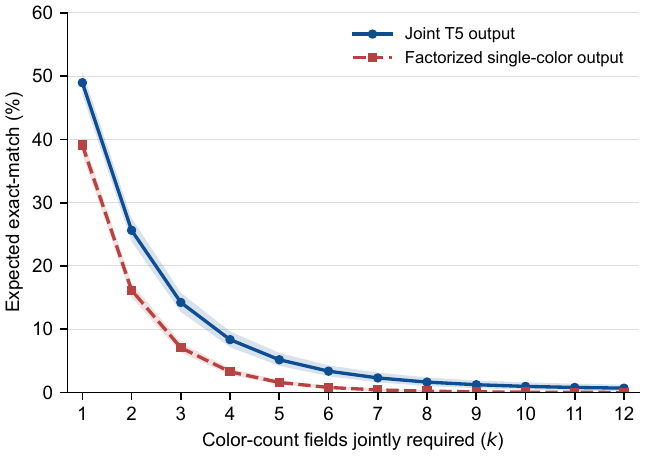}}%
\dimen0=\dimexpr\dimen0-\ht4\relax
\noindent\box0\hfill
\begin{minipage}[t]{0.53\textwidth}
\vspace{0pt}\offinterlineskip
\hbox to\linewidth{\hss\box4\hss}
\vskip\dimen0
\box2
\end{minipage}
\end{figure}

\paragraph{\textcolor{yellow}{\faLightbulb} Subset exact match falls quickly under either strategy.}
For present colors, subset EM falls from 31.88\% (joint) and 32.72\% (factorized) at $k=1$ to 3.71\% and 4.17\% at $k=4$ (Figure~\ref{fig:t5_k_color_present}). Beyond $k=4$, fewer objects have enough present colors (54 at $k=8$, one at $k=12$), so the tail also reflects object composition. Over the full palette, subset EM falls from 48.98\% to 8.35\% (joint) and from 39.13\% to 3.27\% (factorized) between $k=1$ and $k=4$; at $k=12$, palette-vector EM is 0.683\% versus 0.0\% (Figure~\ref{fig:t5_k_color_all}).

\begin{figure}[ht]
  \centering
  \includegraphics[width=\linewidth]{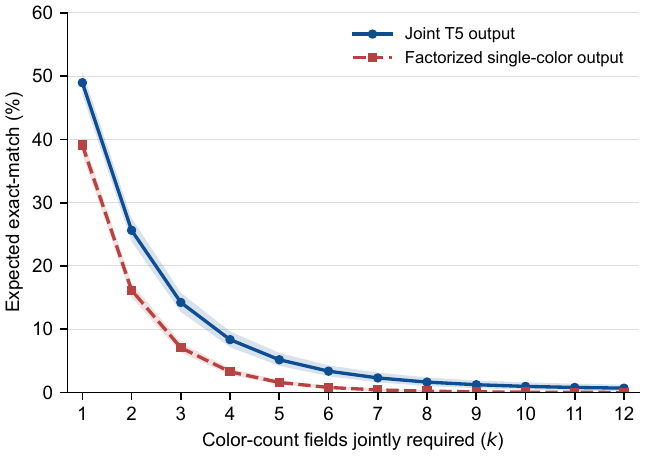}
  \caption{\textbf{All-color subset exact match on \tfivelong{}.} Expected EM over $k$ counts from the full 12-color palette, including zero counts for absent colors, using the same 100 objects at every $k$. Lines average 12 OW models and five runs under Non-Thinking mode; shading gives pointwise 95\% object-bootstrap intervals. Present-color curves appear in Figure~\ref{fig:t5_k_color_present}. \textbf{Takeaway:} Joint outputs outperform factorized outputs across the full palette, alongside their higher absent-color accuracy.}
  \label{fig:t5_k_color_all}
\end{figure}

\paragraph{\textcolor{yellow}{\faLightbulb} Factorization adds positive counts for absent colors.}
Exact zero-count accuracy on absent colors falls from \textcolor{red!70!black}{78.02\%} under joint prediction to \textcolor{red!70!black}{49.66\%} under factorized querying, so positive counts for absent colors rise from 21.98\% to 50.34\%. Merged factorized vectors therefore reach \textcolor{red!70!black}{0.0\%} full-distribution EM for every model.

\paragraph{\textcolor{yellow}{\faLightbulb} A benchmark-matched prompt example improves counts but not exact match.}
With Qwen3.5-27B, an illustrative example matched to the benchmark raises present-color accuracy by $+$10.6\,pp (95\% CI $[+7.6, +14.0]$) and lowers MAE by 0.09 (95\% CI $[0.07, 0.12]$), but full-distribution EM reaches only 5.0\% ($+$2.0\,pp, 95\% CI including zero; Table~\ref{tab:t5_prompt_ablation}).

\begin{table}[t]
\centering
\small
\setlength{\tabcolsep}{4.5pt}
\begin{tabular}{@{}lrrrr@{}}
\toprule
Prompt & EM $\uparrow$ & MAE $\downarrow$ & Present $\uparrow$ & Over $\downarrow$ \\
\midrule
Original      & 3.0 & 0.50 & 48.2 & 40.6 \\
No example    & 1.4 & 0.51 & 45.5 & 43.9 \\
Benchmark-matched & 5.0 & 0.41 & 58.9 & 29.8 \\
\bottomrule
\end{tabular}
\caption{\textbf{Prompt-example ablation on \tfivelong{}.} Qwen3.5-27B under Non-Thinking mode with color inputs, mean over five runs on the same 100 instances. The original prompt contains an illustrative 28-block answer; the benchmark-matched example has 11 blocks and 8 colors, matching the benchmark distribution. EM is full-distribution exact match, MAE is the per-color absolute error in blocks over the 12-color palette, Present is exact-count accuracy on colors present in the object, and Over is the share of present colors that are over-counted. \textbf{Takeaway:} Removing the example leaves all metrics within their paired 95\% CIs, and a benchmark-matched example improves present-color accuracy by \textcolor{custom_darkgreen}{$+10.6$\,pp}, while EM stays \textcolor{red!70!black}{$\le 5.0\%$}.}
\label{tab:t5_prompt_ablation}
\end{table}

\subsubsection{A 2D visibility control separates basic enumeration from 3D presentation}
\label{sec:app_2d_control}

% arXiv version of tables/t5_prompt_ablation.tex (same content), wrapfig as in arXiv:2608.07645.
\begin{wraptable}{L}{0.48\textwidth}
\centering
\small
\setlength{\tabcolsep}{4.5pt}
\begin{tabular}{@{}lrrrr@{}}
\toprule
Prompt & EM $\uparrow$ & MAE $\downarrow$ & Present $\uparrow$ & Over $\downarrow$ \\
\midrule
Original      & 3.0 & 0.50 & 48.2 & 40.6 \\
No example    & 1.4 & 0.51 & 45.5 & 43.9 \\
Benchmark-matched & 5.0 & 0.41 & 58.9 & 29.8 \\
\bottomrule
\end{tabular}
\caption{\textbf{Prompt-example ablation on \tfivelong{}.} Qwen3.5-27B under Non-Thinking mode with color inputs, mean over five runs on the same 100 instances. The original prompt contains an illustrative 28-block answer; the benchmark-matched example has 11 blocks and 8 colors, matching the benchmark distribution. EM is full-distribution exact match, MAE is the per-color absolute error in blocks over the 12-color palette, Present is exact-count accuracy on colors present in the object, and Over is the share of present colors that are over-counted. \textbf{Takeaway:} Removing the example leaves all metrics within their paired 95\% CIs, and a benchmark-matched example improves present-color accuracy by \textcolor{custom_darkgreen}{$+10.6$\,pp}, while EM stays \textcolor{red!70!black}{$\le 5.0\%$}.}
\label{tab:t5_prompt_ablation}
\end{wraptable}

\paragraph{Experimental setup.} For each of the 100 \tfivelong{} objects, we draw one image in which every block appears as a separate square tile with the same color, black edges, and image size as the 3D renders; tile positions are shuffled with a fixed seed, no tile touches or covers another, and the image contains no text. The control therefore keeps each object's color multiset and ground-truth count vector unchanged. The prompt is the \tfive{} prompt with only its first sentence changed to describe an image of separate square tiles; decoding, answer schema, and the illustrative example are unchanged. We evaluate zero-shot Qwen3.5-27B under Non-Thinking mode for five runs and compare it with its saved 3D runs on the default color setting (Table~\ref{tab:t5_2d_control}). Intervals are paired object-bootstrap 95\% intervals (10{,}000 resamples, runs averaged per object).

\begin{table*}[t]
\centering
\small
\setlength{\tabcolsep}{6.5pt}
\begin{tabular}{@{}lrrrrrr@{}}
\toprule
Input & EM $\uparrow$ & EM (gold\,=\,yellow) $\uparrow$ & MAE $\downarrow$ & Present $\uparrow$ & Over $\downarrow$ & Total \\
\midrule
3D (two views) & 3.0 & 6.0 & 0.50 & 48.2 & 40.6 & 14.9 \\
2D tiles & 52.4 & 89.8 & 0.11 & 91.0 & 3.6 & 11.5 \\
\midrule
2D $-$ 3D & $+$49.4 & $+$83.8 & $-$0.39 & $+$42.8 & $-$37.0 & $-$3.4 \\
95\% CI & $[+39.8, +59.0]$ & $[+75.6, +91.0]$ & $[-0.43, -0.34]$ & $[+37.6, +47.9]$ & $[-41.8, -32.2]$ & \\
\bottomrule
\end{tabular}
\caption{\textbf{2D visibility control on \tfivelong{}.} Qwen3.5-27B, Non-Thinking, color setting, mean over five runs on the same 100 objects. Metrics as in Table~\ref{tab:t5_prompt_ablation}; EM (gold\,=\,yellow) merges the two colors in predictions and ground truth as a post-hoc diagnostic, and Total is the mean predicted number of blocks (ground truth 11.4). \textbf{Takeaway:} With every block visible as a separate tile, EM rises by \textcolor{custom_darkgreen}{$+49.4$\,pp}, and the gap remains after merging gold and yellow.}
\label{tab:t5_2d_control}
\end{table*}

\paragraph{\textcolor{yellow}{\faLightbulb} Separate tiles are counted accurately apart from one color confusion.}
In 2D, the model counts present colors exactly 91.0\% of the time.
Of the 238 incorrect 2D answers (over five runs), 187 differ from the ground truth only by one gold block reported as yellow; on the 34 objects without gold blocks, 2D EM is 100\%.
Merging gold and yellow is a post-hoc diagnostic and does not replace the original exact-match metric; under it, EM is 89.8\% in 2D and 6.0\% in 3D, so the 3D gap remains.
The mean predicted total is 11.5 blocks in 2D and 14.9 in 3D, against a ground truth of 11.4, so the model over-counts mainly in 3D.
The control changes occlusion, the need to integrate two views, and the visual layout together, so it cannot separate their contributions, and it covers one model.

% arXiv version of tables/t5_2d_control.tex (same content). Edit both copies when the table changes.
\begin{table*}[h]
\vspace{10pt}
\centering
\small
\setlength{\tabcolsep}{6.5pt}
\begin{tabular}{@{}lrrrrrr@{}}
\toprule
Input & EM $\uparrow$ & EM (gold\,=\,yellow) $\uparrow$ & MAE $\downarrow$ & Present $\uparrow$ & Over $\downarrow$ & Total \\
\midrule
3D (two views) & 3.0 & 6.0 & 0.50 & 48.2 & 40.6 & 14.9 \\
2D tiles & 52.4 & 89.8 & 0.11 & 91.0 & 3.6 & 11.5 \\
\midrule
2D $-$ 3D & $+$49.4 & $+$83.8 & $-$0.39 & $+$42.8 & $-$37.0 & $-$3.4 \\
95\% CI & $[+39.8, +59.0]$ & $[+75.6, +91.0]$ & $[-0.43, -0.34]$ & $[+37.6, +47.9]$ & $[-41.8, -32.2]$ & \\
\bottomrule
\end{tabular}
\caption{\textbf{2D visibility control on \tfivelong{}.} Qwen3.5-27B, Non-Thinking, color setting, mean over five runs on the same 100 objects. Metrics as in Table~\ref{tab:t5_prompt_ablation}; EM (gold\,=\,yellow) merges the two colors in predictions and ground truth as a post-hoc diagnostic, and Total is the mean predicted number of blocks (ground truth 11.4). \textbf{Takeaway:} With every block visible as a separate tile, EM rises by \textcolor{custom_darkgreen}{$+49.4$\,pp}, and the gap remains after merging gold and yellow.}
\label{tab:t5_2d_control}
\end{table*}

\begin{table*}[h]
\centering
\setlength{\tabcolsep}{5pt}
\small
\begin{tabular}{l c c c c c c}
\toprule
& \multicolumn{3}{c}{Zero-shot} & \multicolumn{3}{c}{LoRA, benchmark palette} \\
\cmidrule(lr){2-4}\cmidrule(lr){5-7}
Variant & Acc & MAE & RMSE & Acc & MAE & RMSE \\
\midrule
color (default) & 3.0\,$\pm$0.0 & 0.502\,$\pm$0.002 & 0.937\,$\pm$0.003 & 68.6\,$\pm$0.5 & 0.033\,$\pm$0.001 & 0.182\,$\pm$0.003 \\
color\,+\,axes & 3.0\,$\pm$0.0 & 0.481\,$\pm$0.002 & 0.923\,$\pm$0.004 & 66.8\,$\pm$1.8 & 0.036\,$\pm$0.002 & 0.191\,$\pm$0.004 \\
\bottomrule
\end{tabular}
\caption{\textbf{LoRA fine-tuning of Qwen3.5-27B on \tfivelong{}.} The adapter is trained on 800 generated \tfive{} instances rendered with the benchmark palette, the 12-color, luminance-paired coloring of BedBench. Each cell is the 5-run mean $\pm$ std on the 100 benchmark instances. \textbf{Takeaway:} Training on the benchmark palette raises exact match from 3.0\% to 68.6\%.}
\label{tab:t5_finetune}
\end{table*}

% arXiv version of tables/t5_finetune.tex (same content) with a top gap, since \intextsep is 0pt here. Edit both copies when the table changes.
\begin{table*}[h]
\vspace{10pt}
\centering
\setlength{\tabcolsep}{5pt}
\small
\begin{tabular}{l c c c c c c}
\toprule
& \multicolumn{3}{c}{Zero-shot} & \multicolumn{3}{c}{LoRA, benchmark palette} \\
\cmidrule(lr){2-4}\cmidrule(lr){5-7}
Variant & Acc & MAE & RMSE & Acc & MAE & RMSE \\
\midrule
color (default) & 3.0\,$\pm$0.0 & 0.502\,$\pm$0.002 & 0.937\,$\pm$0.003 & 68.6\,$\pm$0.5 & 0.033\,$\pm$0.001 & 0.182\,$\pm$0.003 \\
color\,+\,axes & 3.0\,$\pm$0.0 & 0.481\,$\pm$0.002 & 0.923\,$\pm$0.004 & 66.8\,$\pm$1.8 & 0.036\,$\pm$0.002 & 0.191\,$\pm$0.004 \\
\bottomrule
\end{tabular}
\caption{\textbf{LoRA fine-tuning of Qwen3.5-27B on \tfivelong{}.} The adapter is trained on 800 generated \tfive{} instances rendered with the benchmark palette, the 12-color, luminance-paired coloring of BedBench. Each cell is the 5-run mean $\pm$ std on the 100 benchmark instances. \textbf{Takeaway:} Training on the benchmark palette raises exact match from 3.0\% to 68.6\%.}
\label{tab:t5_finetune}
\end{table*}

\begin{table}[t]
\centering
\setlength{\tabcolsep}{4pt}
\small
\begin{tabular}{l c c c}
\toprule
Task (variant) & Zero-shot & LoRA & $\Delta$ \\
\midrule
\tfour{} (color) & 45.2\,$\pm$1.3 & 69.6\,$\pm$0.5 & $+$24.4 \\
\tfour{} (color\,+\,axes) & 45.8\,$\pm$0.8 & 61.8\,$\pm$0.4 & $+$16.0 \\
\tseven{} (color) & 18.0\,$\pm$0.0 & 35.0\,$\pm$0.7 & $+$17.0 \\
\tseven{} (color\,+\,axes) & 18.4\,$\pm$0.5 & 36.4\,$\pm$0.5 & $+$18.0 \\
\tten{} (color) & 55.2\,$\pm$0.4 & 62.4\,$\pm$1.5 & $+$7.2 \\
\bottomrule
\end{tabular}
\caption{\textbf{Transfer of the \tfive{} LoRA adapter (benchmark palette) to related color tasks.} Accuracy (\%) in Non-Thinking mode, 5-run mean $\pm$ std. Table~\ref{tab:t5_t7_from_counts} compares direct \tseven{} answers with answers derived from the model's own \tfive{} counts. \textbf{Takeaway:} Fine-tuning on \tfive{} improves the related color tasks only partially, most on \tfour{} and least on \tten{}.}
\label{tab:t5_finetune_transfer}
\end{table}

\begin{table*}[h]
\centering
\setlength{\tabcolsep}{6pt}
\small
\begin{tabular}{l c c}
\toprule
\tseven{} evaluation condition & Color & Color\,+\,axes \\
\midrule
Direct visual questioning & 35.0\,$\pm$0.7 & 36.4\,$\pm$0.5 \\
Self-predicted \tfive{} counts as text & 74.2\,$\pm$0.8 & 73.2\,$\pm$1.3 \\
Ground-truth \tfive{} counts as text & 99.0\,$\pm$0.0 & -- \\
\midrule
Agreement with the argmax of the input counts & 99.6\,$\pm$0.5 & 99.0\,$\pm$0.0 \\
Dominant colors implied by self-predicted counts & 74.6 & 74.2 \\
Text-conditioned accuracy, \tfive{} answer exactly correct & 99.4 & 98.5 \\
Direct visual accuracy, \tfive{} answer exactly correct & 40.8 & 37.4 \\
\bottomrule
\end{tabular}
\caption{\textbf{\tseven{} answered from block counts given as text.} Fine-tuned Qwen3.5-27B, Non-Thinking, accuracy (\%) as 5-run mean $\pm$ std on the same 100 objects; the last three rows pool 5 runs $\times$ 100 objects, and the last two use only the 343 and 334 answers with an exactly correct \tfive{} answer. The implied row scores the argmax of the model's own \tfive{} counts without querying \tseven{}. Ground-truth counts do not depend on the visual setting and are evaluated once (``--''). Paired object-bootstrap 95\% CIs (10{,}000 resamples): self-predicted counts $-$ direct $+$39.2\,pp $[+26.8, +51.4]$ (color) and $+$36.8\,pp $[+23.6, +50.0]$ (color\,+\,axes); ground-truth $-$ self-predicted counts $+$24.8\,pp $[+16.4, +34.0]$ (color). \textbf{Takeaway:} Given its own counts as text, the model derives the dominant colors almost perfectly whenever those counts are correct, unlike when it answers from the images.}
\label{tab:t5_t7_from_counts}
\end{table*}

\subsubsection{Task-specific fine-tuning substantially improves counting but yields incomplete cross-task transfer}
\label{sec:app_finetune}

\paragraph{Experimental setup.} We use a TRL-plus-PEFT LoRA pipeline (rank $r{=}16$, $\alpha{=}32$, dropout 0.05, applied to the language tower projections \texttt{q,k,v,o,gate,up,down}); the vision encoder and connector are kept frozen and the base weights are loaded in 4-bit NF4 with bf16 compute. The training data are 800 \tfive{} instances, plus 100 for validation, produced by the benchmark generator with a separate random seed, so they follow the benchmark's object, color, and rendering distribution without sharing any object with it; the test set is the 100 benchmark instances. Each instance provides the same two perspective views and the same prompt template used at evaluation. We train for 3 epochs with AdamW (lr $2{\times}10^{-4}$, cosine schedule, warmup 0.1, effective batch size 8) on four NVIDIA RTX 3090 GPUs with data parallelism, keep the checkpoint with the lowest validation loss, and evaluate the resulting adapter under the same default-setting protocol as the zero-shot baseline (5-run mean $\pm$ std). Table~\ref{tab:t5_finetune} reports \tfive{}; Table~\ref{tab:t5_finetune_transfer} reports transfer to related color tasks. To test whether the inconsistency we find is specific to fine-tuning, we also measure per-object consistency for every model, before and after fine-tuning. On the same objects, colored renderings, and runs, a model's direct answer to \tfour{} or \tseven{} is compared with the answer implied by its own \tfive{} counts $\hat{\mathbf c}$, $|\{c:\hat c_c>0\}|$ and $\arg\max_c \hat c_c$, and its direct \tone{} answer, where a colored \tone{} run exists, with $\sum_c \hat c_c$. Agreement measures consistency with the model's own counts, not correctness, and leaves the input unchanged; as a chance level, we permute the direct answers across objects within each run (200 permutations).

% arXiv version of tables/t5_finetune_transfer.tex (same content), wrapfig as in arXiv:2608.07645.
\begin{wraptable}{r}{0.45\textwidth}
\centering
\setlength{\tabcolsep}{4pt}
\small
\begin{tabular}{l c c c}
\toprule
Task (variant) & Zero-shot & LoRA & $\Delta$ \\
\midrule
\tfour{} (color) & 45.2\,$\pm$1.3 & 69.6\,$\pm$0.5 & $+$24.4 \\
\tfour{} (color\,+\,axes) & 45.8\,$\pm$0.8 & 61.8\,$\pm$0.4 & $+$16.0 \\
\tseven{} (color) & 18.0\,$\pm$0.0 & 35.0\,$\pm$0.7 & $+$17.0 \\
\tseven{} (color\,+\,axes) & 18.4\,$\pm$0.5 & 36.4\,$\pm$0.5 & $+$18.0 \\
\tten{} (color) & 55.2\,$\pm$0.4 & 62.4\,$\pm$1.5 & $+$7.2 \\
\bottomrule
\end{tabular}
\caption{\textbf{Transfer of the \tfive{} LoRA adapter (benchmark palette) to related color tasks.} Accuracy (\%) in Non-Thinking mode, 5-run mean $\pm$ std. Table~\ref{tab:t5_t7_from_counts} compares direct \tseven{} answers with answers derived from the model's own \tfive{} counts. \textbf{Takeaway:} Fine-tuning on \tfive{} improves the related color tasks only partially, most on \tfour{} and least on \tten{}.}
\label{tab:t5_finetune_transfer}
\end{wraptable}

\paragraph{\textcolor{yellow}{\faLightbulb} In-distribution supervision teaches \tfive{}.}
Fine-tuning raises exact match on \tfivelong{} \textcolor{custom_darkgreen}{from 3.0\% to 68.6\%} on the trained color setting and to \textcolor{custom_darkgreen}{66.8\%} on color\,+\,axes, which was not used in training, while MAE drops \textcolor{red!70!black}{93.4\%} and RMSE \textcolor{red!70!black}{80.6\%}. The best fine-tuned cell remains 24.7\,pp below the human mean of 93.3\% and 10.1\,pp below its Wilson lower bound of 78.7\%. Training only the language model therefore recovers much of the information already supplied by the frozen visual pathway: a probe on the connector output reaches 65.0\% exact match (Appendix~\ref{sec:app_token_probe}). Reasoning, scale, and visual cues do not yield comparable recovery in our experiments.

\paragraph{\textcolor{yellow}{\faLightbulb} Direct answers about the dominant colors disagree with the model's own counts.}
\tsevenlong{} asks for the most frequent colors, which follow directly from a correct \tfive{} answer. The dominant colors implied by the fine-tuned model's own \tfive{} answers are correct \textcolor{custom_darkgreen}{74.6\%} of the time, but the model answers the \tseven{} prompt correctly only \textcolor{red!70!black}{35.0\%} of the time, and only 40.8\% on objects whose \tfive{} answer is exactly correct (Table~\ref{tab:t5_t7_from_counts}). Gains on the other color tasks are partial (\tfourlong{} $+$24.4\,pp, \ttenlong{} $+$7.2\,pp; Table~\ref{tab:t5_finetune_transfer}).

% arXiv version of tables/cross_task_consistency.tex (same content), wrapfig as in arXiv:2608.07645.
\begin{wraptable}{r}{0.5\textwidth}
\centering
\small
\setlength{\tabcolsep}{2pt}
\begin{tabular}{lccc}
\toprule
Models & \tone{} & \tfour{} & \tseven{} \\
\midrule
OW, Non-Thinking (12) & 4.0 (5.4) & 32.1 (16.8) & 21.6 (2.4) \\
OW, Thinking (12) & 9.6 (7.8) & 55.1 (17.7) & 21.9 (3.1) \\
Frontier (4) & -- & 65.8 (16.7) & 39.0 (5.0) \\
\midrule
Qwen3.5-27B, zero-shot & 8.8 (11.8) & 51.8 (17.8) & 23.8 (1.7) \\
Qwen3.5-27B, fine-tuned & -- & 69.6 (19.9) & 36.8 (2.2) \\
\bottomrule
\end{tabular}
\caption{\textbf{Per-object agreement between direct answers and the answer implied by the model's own \tfive{} counts} (\%, chance in parentheses): \tone{} vs.\ $\sum_c\hat c_c$, \tfour{} vs.\ $|\{c:\hat c_c>0\}|$, \tseven{} vs.\ $\arg\max_c\hat c_c$. Same objects, colored renderings, and runs; group rows average the models, and the Qwen3.5-27B rows are Non-Thinking. Chance permutes the direct answers across objects. ``--'': no colored \tone{} run. With the default uncolored \tone{} renderings, OW agreement is 5.2\% (chance 6.0\%) in Non-Thinking and 8.7\% (7.3\%) in Thinking mode. \textbf{Takeaway:} Direct answers rarely agree with the counts the same model reports, before and after fine-tuning, and for \tone{} no more often than chance.}
\label{tab:cross_task_consistency}
\end{wraptable}

\paragraph{\textcolor{yellow}{\faLightbulb} Given explicit counts, the model derives the dominant colors.}
We ask the fine-tuned model the \tseven{} question without images, listing per-color block counts in the prompt: either its own \tfive{} answer for the same object, run, and setting, or the ground truth (Table~\ref{tab:t5_t7_from_counts}). From its own counts, it answers correctly \textcolor{custom_darkgreen}{74.2\%} of the time on the color setting, $+$39.2\,pp over direct visual questioning (paired 95\% CI $[+26.8, +51.4]$), and from ground-truth counts 99.0\%; all misses come from one object whose ten colors are tied at one block, where the model misspells a color name. Its answer matches the highest stated count in 99.6\% of cases, and when its \tfive{} answer is exactly correct, text-conditioned accuracy is 99.4\% against \textcolor{red!70!black}{40.8\%} from the images. The model thus derives the judgment from explicit counts but does not consistently produce it from the images.

\paragraph{\textcolor{yellow}{\faLightbulb} VLMs also answer inconsistently with their own counts.}
The counts-as-text test changes the input; the per-object check does not (Table~\ref{tab:cross_task_consistency}). Before fine-tuning, the direct \tseven{} answers of Qwen3.5-27B match the most frequent colors in its own counts on 23.8\% of objects (chance 1.7\%), and fine-tuning raises this only to 36.8\%. Across the 12 OW models, agreement is 21.6\% for \tseven{} (chance 2.4\%) and 32.1\% for \tfour{} (chance 16.8\%) in Non-Thinking mode, and frontier models reach 39.0\% and 65.8\%. Direct \tone{} answers match the sum of the model's own counts on only 4.0\% of objects, below the 5.4\% chance level, yet they are more often correct than that sum (31.8\% vs.\ 3.8\%), so the models do not answer \tone{} by adding up the counts they report. The check also separates the two failure types: when the \tfive{} answer is wrong, \tseven{} agreement stays at 21.3\%, so models are rarely wrong but consistent, and even the 41 exactly correct OW \tfive{} answers are followed by a consistent \tseven{} answer only 61.0\% of the time.

\begin{table}[t]
\centering
\small
\setlength{\tabcolsep}{2pt}
\begin{tabular}{lccc}
\toprule
Models & \tone{} & \tfour{} & \tseven{} \\
\midrule
OW, Non-Thinking (12) & 4.0 (5.4) & 32.1 (16.8) & 21.6 (2.4) \\
OW, Thinking (12) & 9.6 (7.8) & 55.1 (17.7) & 21.9 (3.1) \\
Frontier (4) & -- & 65.8 (16.7) & 39.0 (5.0) \\
\midrule
Qwen3.5-27B, zero-shot & 8.8 (11.8) & 51.8 (17.8) & 23.8 (1.7) \\
Qwen3.5-27B, fine-tuned & -- & 69.6 (19.9) & 36.8 (2.2) \\
\bottomrule
\end{tabular}
\caption{\textbf{Per-object agreement between direct answers and the answer implied by the model's own \tfive{} counts} (\%, chance in parentheses): \tone{} vs.\ $\sum_c\hat c_c$, \tfour{} vs.\ $|\{c:\hat c_c>0\}|$, \tseven{} vs.\ $\arg\max_c\hat c_c$. Same objects, colored renderings, and runs; group rows average the models, and the Qwen3.5-27B rows are Non-Thinking. Chance permutes the direct answers across objects. ``--'': no colored \tone{} run. With the default uncolored \tone{} renderings, OW agreement is 5.2\% (chance 6.0\%) in Non-Thinking and 8.7\% (7.3\%) in Thinking mode. \textbf{Takeaway:} Direct answers rarely agree with the counts the same model reports, before and after fine-tuning, and for \tone{} no more often than chance.}
\label{tab:cross_task_consistency}
\end{table}

% \paragraph{Sensitivity to the training distribution.}
% An adapter trained with the same hyperparameters on a different
% 8-color distribution reaches only 9.4\% EM on \tfive{},
% compared with 68.6\% when trained on the benchmark palette
% (Table~\ref{tab:t5_finetune}).
% Because the two settings differ in both color coverage
% and assignment distribution, this comparison suggests sensitivity
% to the training distribution without isolating its cause.

\twocolumn
\subsection{Thinking helps selected tasks but leaves exact answers unreliable}

\paragraph{Overview.} This section expands Section~\ref{sec:thinking_paradox}: which tasks gain or lose from Thinking mode, what the exposed reasoning traces show, and whether a larger reasoning budget helps.

\paragraph{Experimental setup.} We combine a quantitative mode contrast, a controlled budget sweep, a qualitative trace audit, and class-balanced metrics for the binary tasks. \begin{itemize}[leftmargin=1.4em,nosep]
    \item For the mode contrast, we compare Non-Thinking and Thinking modes under the default visual setting across all 12 OW VLMs (Table~\ref{tab:thinking_delta_full}).
    \item For the budget sweep, we vary the reasoning budget over $\{0, 1024, 2048, 4096\}$ tokens on the two largest OW models (Qwen3.5-27B, Gemma4-31B) for three tasks selected to span the win/loss spectrum of the mode contrast: \tonelong{}, \tfourlong{}, and \televenlong{} (Table~\ref{tab:thinking_scaling}).
    \item For the trace audit, we read reasoning traces on a seeded random draw of instances (one per task, aligned across the best-performing OW model for that task and the four frontier models on identical inputs) and manually annotate the salient spans in the exposed traces using the span-level behavior taxonomy of Table~\ref{tab:annot_taxonomy}. Per-task trace panels with the highlighted, tagged spans are provided in Appendix~\ref{sec:app_qual_thinking_panels}. For OW models, we collect the raw thinking traces from model inference. For GPT-5 and the two Gemini models, the OpenRouter and Vertex AI endpoints used in our evaluation did not expose reasoning traces, so we manually collected the available meta-reasoning summaries through their respective chat interfaces for the qualitative audit. For Grok-4.20, we collected the meta-reasoning trace returned through OpenRouter. Because exposed reasoning traces need not faithfully reflect the computation underlying a model's answer, particularly for frontier models where only summarized reasoning is available, we use these annotations to characterize recurring trace patterns rather than as direct evidence of internal causal mechanisms.
    \item For the binary tasks, precision, recall, and F1 are macro-averaged over the two classes and then averaged over the 12 OW models under the default visual setting; each model $\times$ mode cell is a 5-run mean (Table~\ref{tab:sec_thinking}).
\end{itemize}

% arXiv version of tables/thinking_delta.tex (same content), placed here with [h]. Edit both copies when the table changes.
\begin{table*}[h]
\vspace{10pt}
\centering
\setlength{\tabcolsep}{5.5pt}
\small
\begin{tabular}{l|cccc|cccl}
\toprule
Task & Non-Thinking & Thinking & $\Delta_\text{thinking}$ & 95\% CI & $\uparrow$ & $\downarrow$ & $\approx$ & Verdict \\
\midrule
\tonelong{} & 29.2 & 20.2 & \bf \textcolor{red!70!black}{$-8.9$} & $[-11.9, -6.0]$ & 2 & 8 & 2 & Non-Thinking wins \\
\ttwolong{} & 58.8 & 72.6 & \textcolor{custom_darkgreen}{$+13.8$} & $[+9.3, +18.4]$ & 11 & 1 & 0 & Thinking wins \\
\tthreelong{} & 0.2 & 0.1 & \textcolor{red!70!black}{$-0.1$} & $[-0.4, +0.1]$ & 0 & 0 & 12 & Tie \\
\tfourlong{} & 27.1 & 41.0 & \bf \textcolor{custom_darkgreen}{$+13.9$} & $[+8.9, +19.1]$ & 10 & 1 & 1 & Thinking wins \\
\tfivelong{} & 0.7 & 1.0 & \textcolor{custom_darkgreen}{$+0.3$} & $[-0.3, +1.0]$ & 1 & 0 & 11 & Tie \\
\tsixlong{} & 70.2 & 72.9 & \textcolor{custom_darkgreen}{$+2.8$} & $[+0.4, +5.1]$ & 4 & 5 & 3 & Thinking wins \\
\tsevenlong{} & 14.0 & 13.6 & \textcolor{red!70!black}{$-0.4$} & $[-2.2, +1.4]$ & 3 & 4 & 5 & Tie \\
\teightlong{} & 58.2 & 55.4 & \textcolor{red!70!black}{$-2.9$} & $[-6.3, +0.5]$ & 2 & 4 & 6 & Non-Thinking wins$^\dagger$ \\
\tninelong{} & 55.8 & 48.4 & \textcolor{red!70!black}{$-7.4$} & $[-10.3, -4.6]$ & 2 & 7 & 3 & Non-Thinking wins \\
\ttenlong{} & 53.1 & 56.1  & \textcolor{custom_darkgreen}{$+3.0$} & $[-0.2, +6.2]$ & 6 & 3 & 3 & Thinking wins$^\dagger$ \\
\televenlong{} & 43.1 & 41.8 & \textcolor{red!70!black}{$-1.3$} & $[-2.4, -0.2]$ & 2 & 5 & 5 & Tie \\
\ttwelvelong{} & 52.6 & 52.0 &  \textcolor{red!70!black}{$-0.5$} & $[-3.4, +2.3]$ & 1 & 4 & 7 & Tie \\
\tthirteenlong{} & 37.4 & 36.6  & \textcolor{red!70!black}{$-0.7$} & $[-2.1, +0.6]$ & 1 & 3 & 8 & Tie \\
\midrule
\textbf{Total} & & & \textcolor{custom_darkgreen}{$+0.9$} & -- & 4 & 3 & 6 & -- \\
\bottomrule
\end{tabular}
\caption{\textbf{Per-task effect of Thinking.} Comparison at the default 1024-token budget under the default visual setting. Non-Thinking and Thinking report mean accuracy (\%) over the 12 OW models. $\uparrow$/$\downarrow$/$\approx$ count how many of the 12 models win, lose, or tie under Thinking by the same 2\,pp threshold (not per-model significance). Verdict labels each task by its mean $\Delta_\text{thinking}$: Thinking wins above $+$2\,pp, Non-Thinking wins below $-$2\,pp, and Tie otherwise. The 95\% CI is a paired bootstrap interval for the mean $\Delta_\text{thinking}$ (10{,}000 resamples of objects, scenes for \tthirteen{}, and pairs for \tthree{}, with models and runs fixed; no multiplicity correction). $^\dagger$The interval includes zero. \textbf{Takeaway:} Thinking helps most on \ttwo{} and \tfour{} (\textcolor{custom_darkgreen}{$+$13.8} and \textcolor{custom_darkgreen}{$+$13.9\,pp}), hurts most on \tone{} and \tnine{} (\textcolor{red!70!black}{$-$8.9} and \textcolor{red!70!black}{$-$7.4\,pp}), and moves the mean across tasks by only \textcolor{custom_darkgreen}{$+$0.9\,pp}.}
\label{tab:thinking_delta_full}
\end{table*}

\subsubsection{When thinking helps and what the traces reveal}
\label{sec:app_thinking}

Table~\ref{tab:thinking_delta_full} gives the per-task breakdown at the default 1024-token budget. By the 2\,pp threshold, 4 tasks show a Thinking win, 3 a Non-Thinking win, and 6 tie; the paired bootstrap intervals exclude zero for all wins and losses except \teightlong{} and \ttenlong{}. Excluding Kimi-VL-A3B, whose Thinking checkpoint collapses on several tasks, changes only \teight{}, which becomes a tie ($-1.6$\,pp), and shrinks the \tonelong{} drop from $-8.9$ to $-6.9$\,pp. The trace audit covers 457 tagged spans across 65 traces.

\begin{table*}
\centering
\setlength{\tabcolsep}{5.5pt}
\small
\begin{tabular}{l|cccc|cccl}
\toprule
Task & Non-Thinking & Thinking & $\Delta_\text{thinking}$ & 95\% CI & $\uparrow$ & $\downarrow$ & $\approx$ & Verdict \\
\midrule
\tonelong{} & 29.2 & 20.2 & \bf \textcolor{red!70!black}{$-8.9$} & $[-11.9, -6.0]$ & 2 & 8 & 2 & Non-Thinking wins \\
\ttwolong{} & 58.8 & 72.6 & \textcolor{custom_darkgreen}{$+13.8$} & $[+9.3, +18.4]$ & 11 & 1 & 0 & Thinking wins \\
\tthreelong{} & 0.2 & 0.1 & \textcolor{red!70!black}{$-0.1$} & $[-0.4, +0.1]$ & 0 & 0 & 12 & Tie \\
\tfourlong{} & 27.1 & 41.0 & \bf \textcolor{custom_darkgreen}{$+13.9$} & $[+8.9, +19.1]$ & 10 & 1 & 1 & Thinking wins \\
\tfivelong{} & 0.7 & 1.0 & \textcolor{custom_darkgreen}{$+0.3$} & $[-0.3, +1.0]$ & 1 & 0 & 11 & Tie \\
\tsixlong{} & 70.2 & 72.9 & \textcolor{custom_darkgreen}{$+2.8$} & $[+0.4, +5.1]$ & 4 & 5 & 3 & Thinking wins \\
\tsevenlong{} & 14.0 & 13.6 & \textcolor{red!70!black}{$-0.4$} & $[-2.2, +1.4]$ & 3 & 4 & 5 & Tie \\
\teightlong{} & 58.2 & 55.4 & \textcolor{red!70!black}{$-2.9$} & $[-6.3, +0.5]$ & 2 & 4 & 6 & Non-Thinking wins$^\dagger$ \\
\tninelong{} & 55.8 & 48.4 & \textcolor{red!70!black}{$-7.4$} & $[-10.3, -4.6]$ & 2 & 7 & 3 & Non-Thinking wins \\
\ttenlong{} & 53.1 & 56.1  & \textcolor{custom_darkgreen}{$+3.0$} & $[-0.2, +6.2]$ & 6 & 3 & 3 & Thinking wins$^\dagger$ \\
\televenlong{} & 43.1 & 41.8 & \textcolor{red!70!black}{$-1.3$} & $[-2.4, -0.2]$ & 2 & 5 & 5 & Tie \\
\ttwelvelong{} & 52.6 & 52.0 &  \textcolor{red!70!black}{$-0.5$} & $[-3.4, +2.3]$ & 1 & 4 & 7 & Tie \\
\tthirteenlong{} & 37.4 & 36.6  & \textcolor{red!70!black}{$-0.7$} & $[-2.1, +0.6]$ & 1 & 3 & 8 & Tie \\
\midrule
\textbf{Total} & & & \textcolor{custom_darkgreen}{$+0.9$} & -- & 4 & 3 & 6 & -- \\
\bottomrule
\end{tabular}
\caption{\textbf{Per-task effect of Thinking.} Comparison at the default 1024-token budget under the default visual setting. Non-Thinking and Thinking report mean accuracy (\%) over the 12 OW models. $\uparrow$/$\downarrow$/$\approx$ count how many of the 12 models win, lose, or tie under Thinking by the same 2\,pp threshold (not per-model significance). Verdict labels each task by its mean $\Delta_\text{thinking}$: Thinking wins above $+$2\,pp, Non-Thinking wins below $-$2\,pp, and Tie otherwise. The 95\% CI is a paired bootstrap interval for the mean $\Delta_\text{thinking}$ (10{,}000 resamples of objects, scenes for \tthirteen{}, and pairs for \tthree{}, with models and runs fixed; no multiplicity correction). $^\dagger$The interval includes zero. \textbf{Takeaway:} Thinking helps most on \ttwo{} and \tfour{} (\textcolor{custom_darkgreen}{$+$13.8} and \textcolor{custom_darkgreen}{$+$13.9\,pp}), hurts most on \tone{} and \tnine{} (\textcolor{red!70!black}{$-$8.9} and \textcolor{red!70!black}{$-$7.4\,pp}), and moves the mean across tasks by only \textcolor{custom_darkgreen}{$+$0.9\,pp}.}
\label{tab:thinking_delta_full}
\end{table*}

% arXiv version of tables/thinking_scaling.tex (same content), wrapfig as in arXiv:2608.07645. Edit both copies when the table changes.
\begin{wraptable}{r}{0.5\textwidth}
\centering
\scriptsize
\begin{tabular}{llcccc}
\toprule
Task & Model & 0 & 1024 & 2048 & 4096 \\
\midrule
\midrule
\multirow{3}{*}{\tonelong{}}
 & Qwen3.5-27B  & \textbf{39.8} & 24.2 & {27.6} & 26.2 \\
 & Gemma4-31B   & \textbf{35.0} & {17.2} & 13.0 & 13.4 \\
\cmidrule{2-6}
 & \textit{Average} & \textbf{37.4} & {20.7} & 20.3 & 19.8 \\
\midrule
\midrule
\multirow{3}{*}{\tfourlong{}}
 & Qwen3.5-27B  & 45.2 & {57.2} & \textbf{60.8} & {57.2} \\
 & Gemma4-31B   & 52.8 & 56.2 & {57.6} & \textbf{60.0} \\
 \cmidrule{2-6}
 & \textit{Average} & 49.0 & 56.7 & \textbf{59.2} & {58.6} \\
\midrule
\midrule
\multirow{3}{*}{\televenlong{}}
 & Qwen3.5-27B  & \textbf{55.7} & {50.5} & 45.4 & 45.2 \\
 & Gemma4-31B   & {47.3} & \textbf{48.6} & 46.3 & 46.5 \\
 \cmidrule{2-6}
 & \textit{Average} & \textbf{51.5} & {49.6} & 45.9 & 45.8 \\
\bottomrule
\end{tabular}
\caption{\textbf{Effect of the reasoning budget.} Default-setting accuracy (\%) of Qwen3.5-27B and Gemma4-31B at four reasoning-token budgets, on three tasks spanning the win/loss spectrum of Table~\ref{tab:thinking_delta_full} (Appendix~\ref{sec:app_thinking}): \tone{} (most hurt by Thinking), \tfour{} (most helped), and \televen{} (near the middle). Each cell is a 5-run mean; \textbf{boldface} marks the best budget per row. \textbf{Takeaway:} On average, every nonzero budget leaves \tone{} below budget~0, as do budgets of 2048 tokens or more for \televen{}, while \tfour{} improves at every nonzero budget and levels off after 2048 tokens (paired bootstrap intervals in Appendix~\ref{sec:app_thinking_scaling}).}
\label{tab:thinking_scaling}
\end{wraptable}
\paragraph{\textcolor{yellow}{\faLightbulb} Thinking gains concentrate on tasks with perceptually grounded or verbalizable steps.}
\emph{Good Perception} is the tag most associated with correct outcomes: it appears in 16 of the 21 correct traces but only 3 of the 44 incorrect ones, mostly in the two Gemini models (30 of its 43 spans). GPT-5 is the only model that uses tools (all 27 \emph{Tool Use} spans), probing crops and pixel values mostly on \tthreelong{} and the color tasks; four of its seven tool-using traces end correct. \emph{Good Tactic} and \emph{Good Intuition} are rare (3 and 6 spans), with two of the three \emph{Good Tactic} spans on \ttwolong{}. On \ttwo{} and \tfourlong{}, the clearest Thinking wins, the traces contain grounded intermediate steps such as dimension comparisons or explicit color enumeration, consistent with prior work linking chain-of-thought gains to explicit intermediate steps \cite{sprague2025cot,wei2022chain,kojima2022large}.

\paragraph{\textcolor{yellow}{\faLightbulb} Perceptual errors frequently persist through subsequent reasoning.}
\emph{Wrong Perception} is the most frequent tag (222 of 457 spans, applied to all seven models) and appears in 42 of 44 incorrect traces, especially on \tonelong{}, \tthreelong{}, \tfivelong{}, \tninelong{}, and \televenlong{}, where Thinking does not improve or reduces accuracy (Table~\ref{tab:annot_taxonomy}). The OW traces also often repeat earlier steps instead of re-inspecting the image (87 of the 100 \emph{Reiterating} spans) and end with a \emph{Budget Hit} on all 13 tasks, so many continue from an incorrect scene interpretation until the budget runs out. Other failures are model-specific: Grok reasons about its own summarization instruction (\emph{Harnessing}, 27 spans) and anchors on the prompt's illustrative example (\emph{Prior Bias}), and Qwen3.5-27B reads a green cube as yellow on \tfourlong{} (\emph{Color Confusion}). A second pass finds traces dominated by \emph{Wrong Perception} or truncated mid-derivation that are nonetheless scored correct (\emph{Correct by Luck}), and a Grok \ttwelvelong{} trace that states the ground-truth candidate but outputs another (\emph{Think--Answer Mismatch}; Table~\ref{tab:annot_taxonomy}, rows 12--13). Figure~\ref{fig:thinking_trace_fsm_main} (Section~\ref{sec:results}) summarizes these paths; the most common failure path runs from flawed steps to the budget limit and a wrong answer. These patterns resemble persistent visual failures in slow-thinking VLMs \cite{rahmanzadehgervi2025visionlanguagemodelsblind} and compounding errors in multi-step reasoning \cite{dziri2023faith}, but they remain descriptive, since exposed reasoning need not reflect the basis of an answer \cite{turpin2023language}.

\subsubsection{Larger reasoning budgets do not recover the affected tasks}
\label{sec:app_thinking_scaling}

% The first-order pattern in \ref{sec:app_thinking} motivates a controlled budget sweep: if the deficit on 3D-structural tasks were merely a matter of insufficient reasoning length, larger budgets should attenuate it. We select two large OW models (Qwen3.5-27B, Gemma4-31B) and choose three representative tasks that span the win/loss spectrum of Table~\ref{tab:thinking_delta_full}: \tonelong{} (most hurt by thinking), \tfourlong{} (most helped by thinking), and \televenlong{} (a cross-view geometry task that lies near the middle of the distribution). Results are reported in Table~\ref{tab:thinking_scaling}.

Table~\ref{tab:thinking_scaling} reports the budget sweep. Intervals are paired object-level bootstrap intervals over the two-model average (10{,}000 resamples), with models and runs held fixed.

\paragraph{\textcolor{yellow}{\faLightbulb} Linguistically verbalizable tasks absorb additional reasoning tokens.}
\tfourlong{} is the only task whose accuracy increases with budget. Averaged over the two models, it rises from \textcolor{custom_darkgreen}{$49.0\%$} at budget~0 to \textcolor{custom_darkgreen}{$56.7\%$} at 1024 and \textcolor{custom_darkgreen}{$59.2\%$} at 2048 ($+10.2$\,pp, 95\% CI $[+3.4, +17.3]$), then levels off ($58.6\%$ at 4096; $-0.6$\,pp vs.\ 2048, $[-4.6, +3.5]$), while Gemma4-31B keeps improving (\textcolor{custom_darkgreen}{$52.8 \rightarrow 60.0\%$}). Counting distinct colors decomposes naturally into identifying a color, checking whether it has been seen, and incrementing a counter.

\begin{table}[t]
\centering
\scriptsize
\begin{tabular}{llcccc}
\toprule
Task & Model & 0 & 1024 & 2048 & 4096 \\
\midrule
\midrule
\multirow{3}{*}{\tonelong{}}
 & Qwen3.5-27B  & \textbf{39.8} & 24.2 & {27.6} & 26.2 \\
 & Gemma4-31B   & \textbf{35.0} & {17.2} & 13.0 & 13.4 \\
\cmidrule{2-6}
 & \textit{Average} & \textbf{37.4} & {20.7} & 20.3 & 19.8 \\
\midrule
\midrule
\multirow{3}{*}{\tfourlong{}}
 & Qwen3.5-27B  & 45.2 & {57.2} & \textbf{60.8} & {57.2} \\
 & Gemma4-31B   & 52.8 & 56.2 & {57.6} & \textbf{60.0} \\
 \cmidrule{2-6}
 & \textit{Average} & 49.0 & 56.7 & \textbf{59.2} & {58.6} \\
\midrule
\midrule
\multirow{3}{*}{\televenlong{}}
 & Qwen3.5-27B  & \textbf{55.7} & {50.5} & 45.4 & 45.2 \\
 & Gemma4-31B   & {47.3} & \textbf{48.6} & 46.3 & 46.5 \\
 \cmidrule{2-6}
 & \textit{Average} & \textbf{51.5} & {49.6} & 45.9 & 45.8 \\
\bottomrule
\end{tabular}
\caption{\textbf{Effect of the reasoning budget.} Default-setting accuracy (\%) of Qwen3.5-27B and Gemma4-31B at four reasoning-token budgets, on three tasks spanning the win/loss spectrum of Table~\ref{tab:thinking_delta_full} (Appendix~\ref{sec:app_thinking}): \tone{} (most hurt by Thinking), \tfour{} (most helped), and \televen{} (near the middle). Each cell is a 5-run mean; \textbf{boldface} marks the best budget per row. \textbf{Takeaway:} On average, every nonzero budget leaves \tone{} below budget~0, as do budgets of 2048 tokens or more for \televen{}, while \tfour{} improves at every nonzero budget and levels off after 2048 tokens (paired bootstrap intervals in Appendix~\ref{sec:app_thinking_scaling}).}
\label{tab:thinking_scaling}
\end{table}

\paragraph{\textcolor{yellow}{\faLightbulb} 3D-structural tasks degrade rather than recover with more thinking.}
On \tonelong{}, average accuracy drops from \textcolor{red!70!black}{$37.4\%$} at budget~0 to \textcolor{red!70!black}{$20.7\%$} at 1024 and stays there ($20.3\%$ at 2048 and $19.8\%$ at 4096; \textcolor{red!70!black}{$-17.6$\,pp}, $[-25.7, -9.4]$; $-0.5$\,pp from 2048 to 4096, $[-4.9, +3.9]$), with Gemma4-31B falling \textcolor{red!70!black}{from $35.0\%$ to $13.4\%$}. On \televenlong{}, Qwen3.5-27B declines monotonically (\textcolor{red!70!black}{$55.7 \rightarrow 50.5 \rightarrow 45.4 \rightarrow 45.2\%$}), and the average drops \textcolor{red!70!black}{$-5.7$\,pp} from budget~0 to 4096 ($[-9.5, -2.0]$), although the change at 1024 ($-2.0$\,pp, $[-4.9, +0.9]$) is not distinguishable from zero. Unlike \tfour{}, these tasks require judgments about latent 3D structure, and the decline is a test-time-compute analogue of the inverse-scaling effects discussed by \citet{mckenzie2023inverse}. Even \tone{}, which can in principle be decomposed into per-layer counts, does not recover; together with the trace audit, this suggests that additional tokens do not reliably correct inaccurate scene estimates in the tested models, tasks, and budgets.

\subsubsection{Thinking shifts decision behavior on binary tasks}
\label{sec:app_thinking_policy}

\begin{table}[t]
\centering
\small
\setlength{\tabcolsep}{7pt}
\begin{tabular}{llcccc}
\toprule
Task & Mode & Acc (\%) & Prec & Recall & F1 \\
\midrule
\multirow{2}{*}{\teight{}} & NT & 58.2 & 0.383 & 0.510 & 0.398 \\
 & TH & 55.4 & 0.543 & 0.504 & 0.430 \\
\midrule
\multirow{2}{*}{\ttwo{}} & NT & 58.8 & 0.685 & 0.588 & 0.506 \\
 & TH & 72.6 & 0.783 & 0.726 & 0.697 \\
\midrule
\multirow{2}{*}{\tsix{}} & NT & 70.2 & 0.692 & 0.664 & 0.595 \\
 & TH & 72.9 & 0.670 & 0.733 & 0.675 \\
\midrule
\multirow{2}{*}{\tten{}} & NT & 53.1 & 0.546 & 0.531 & 0.411 \\
 & TH & 56.1 & 0.558 & 0.564 & 0.509 \\
\bottomrule
\end{tabular}
\caption{\textbf{Binary tasks under accuracy vs.\ macro-averaged metrics.} Non-Thinking (NT) vs.\ Thinking (TH), averaged over 12 OW models under the default setting; each cell is a 5-run mean. Thinking lowers accuracy on \teight{} and raises it on \ttwo{}, \tsix{}, and \tten{}; the \teight{} and \tten{} changes have 95\% intervals that include zero (Table~\ref{tab:thinking_delta_full}). \textbf{Takeaway:} On \teight{}, Thinking lowers accuracy but raises precision and F1; on \ttwo{}, \tsix{}, and \tten{}, recall rises more than precision.}
\label{tab:sec_thinking}
\end{table}

% arXiv version of tables/sec_thinking.tex (same content), wrapfig as in arXiv:2608.07645. Edit both copies when the table changes.
\begin{wraptable}{r}{0.5\textwidth}
\centering
\small
\setlength{\tabcolsep}{4pt}
\begin{tabular}{llcccc}
\toprule
Task & Mode & Acc (\%) & Prec & Recall & F1 \\
\midrule
\multirow{2}{*}{\teight{}} & NT & 58.2 & 0.383 & 0.510 & 0.398 \\
 & TH & 55.4 & 0.543 & 0.504 & 0.430 \\
\midrule
\multirow{2}{*}{\ttwo{}} & NT & 58.8 & 0.685 & 0.588 & 0.506 \\
 & TH & 72.6 & 0.783 & 0.726 & 0.697 \\
\midrule
\multirow{2}{*}{\tsix{}} & NT & 70.2 & 0.692 & 0.664 & 0.595 \\
 & TH & 72.9 & 0.670 & 0.733 & 0.675 \\
\midrule
\multirow{2}{*}{\tten{}} & NT & 53.1 & 0.546 & 0.531 & 0.411 \\
 & TH & 56.1 & 0.558 & 0.564 & 0.509 \\
\bottomrule
\end{tabular}
\caption{\textbf{Binary tasks under accuracy vs.\ macro-averaged metrics.} Non-Thinking (NT) vs.\ Thinking (TH), averaged over 12 OW models under the default setting; each cell is a 5-run mean. Thinking lowers accuracy on \teight{} and raises it on \ttwo{}, \tsix{}, and \tten{}; the \teight{} and \tten{} changes have 95\% intervals that include zero (Table~\ref{tab:thinking_delta_full}). \textbf{Takeaway:} On \teight{}, Thinking lowers accuracy but raises precision and F1; on \ttwo{}, \tsix{}, and \tten{}, recall rises more than precision.}
\label{tab:sec_thinking}
\end{wraptable}

\paragraph{\textcolor{yellow}{\faLightbulb} Where thinking lowers accuracy on \teightlong{}, the metric pattern is consistent with a shifted decision boundary.}
On \teight{}, Thinking lowers accuracy by \textcolor{red!70!black}{$2.9$\,pp} (95\% CI $[-6.3, +0.5]$) but raises precision \textcolor{custom_darkgreen}{from 0.383 to 0.543} and F1 \textcolor{custom_darkgreen}{from 0.398 to 0.430}, while recall stays essentially unchanged (\textcolor{red!70!black}{0.510 to 0.504}), consistent with a shift in the yes/no decision rather than better classification.

\paragraph{\textcolor{yellow}{\faLightbulb} Where thinking raises accuracy, recall rises more than precision.}
On \ttwolong{}, \tsixlong{}, and \ttenlong{}, recall rises more than precision. The largest change is on \ttwo{} (recall \textcolor{custom_darkgreen}{$+0.138$}, F1 \textcolor{custom_darkgreen}{$+0.191$}), and on \tsix{} precision even falls slightly (\textcolor{red!70!black}{$-0.022$}) while recall rises by \textcolor{custom_darkgreen}{$+0.070$}. Like the \teight{} pattern, these changes describe output behavior and do not by themselves show that Thinking recovers 3D structure.

\begingroup
\footnotesize
\renewcommand{\arraystretch}{1.25}
\begin{table*}[h]
\small
\centering
\begin{tabularx}{\textwidth}{@{}c >{\raggedright\arraybackslash}p{1.8cm} >{\raggedright\arraybackslash}p{4.4cm} >{\raggedright\arraybackslash}p{2cm} X@{}}
\toprule
\# & Name & Description & Models & Representative quote (qualitative sample) \\
\midrule
\multicolumn{5}{@{}l}{\textit{Good behavior}} \\
1 & Tool Use & Invokes external code execution to inspect pixel values or crop the input image. & GPT & GPT-5, \tfive{}: samples RGB pixel values via \texttt{numpy} to separate look-alike hues (Tab.~\ref{tab:qual_trace_t5}, Appendix~\ref{sec:app_qual_thinking_panels}) \\
2 & Good Perception & An isolated span where the model's stated read of the scene is correct and matches the ground-truth structure. & All & Gemini-3-Flash, \tseven{}: \textit{``we clearly see a diagonal row of 2 teal blocks forming part of the central structure''} (matches GT; Tab.~\ref{tab:qual_trace_t7}, Appendix~\ref{sec:app_qual_thinking_panels}) \\
3 & Good Tactic & A valid problem-solving strategy or check applied correctly, e.g.\ comparing sorted dimensions or framing a task as set difference. & Grok, Gemini & Gemini-3-Flash, \ttwo{}: \textit{``Comparing the sorted dimensions: $2\le2$ (Fits) \ldots the orientation is valid''} (Tab.~\ref{tab:qual_trace_t2}, Appendix~\ref{sec:app_qual_thinking_panels}) \\
4 & Good Intuition & A correct-magnitude estimate reached by heuristic guess rather than a completed derivation. & GPT-5, Grok & GPT-5, \tone{}: \textit{``I estimate a total of around 10 cubes''} (GT 10; Tab.~\ref{tab:qual_trace_t1}, Appendix~\ref{sec:app_qual_thinking_panels}) \\
\midrule
\multicolumn{5}{@{}l}{\textit{Bad behavior}} \\
5 & Reiterating & Restates or redoes a step already computed, with no new evidence. & All & Gemma4-31B, \tone{}: \textit{``Wait, let's re-examine.''} (repeated $\sim$6$\times$ in the same box; Tab.~\ref{tab:qual_trace_t1}, Appendix~\ref{sec:app_qual_thinking_panels}) \\
6 & Prior Bias & Anchors the final answer on the prompt's illustrative example or output-format template. & Grok & Grok, \televen{}: \textit{``But the example is \{ 'candidate\_idx': 2, \}, so mine is 0''} (Tab.~\ref{tab:qual_trace_t11}, Appendix~\ref{sec:app_qual_thinking_panels}) \\
7 & Wrong Perception & An isolated span where the model's stated read of the scene is incorrect relative to the actual image. & All & Gemma4-31B, \tone{}: \textit{``One block at the back-right. * One block in the middle. * One block at the front-left''} (Tab.~\ref{tab:qual_trace_t1}, Appendix~\ref{sec:app_qual_thinking_panels}) \\
8 & Color Confusion & Misidentifies or conflates two visually adjacent or same-family colors, e.g.\ \textcolor{bedYellow}{yellow} vs.\ \textcolor{bedGold}{gold}, \textcolor{bedMagenta}{magenta} vs.\ \textcolor{bedPurple}{purple}. & Open weights & Qwen3.5-27B, \tfour{}: ``Next to the \textcolor{bedGreen}{green} cube (right): \textcolor{bedYellow}{Yellow} cube'' (Tab.~\ref{tab:qual_trace_t4}, Appendix~\ref{sec:app_qual_thinking_panels}, in this situation, it should be \textcolor{bedGold}{gold}, not \textcolor{bedYellow}{yellow}) \\
9 & Harnessing & The thinking derails into reasoning about the system's own meta instructions instead of the task. & Grok & Grok, \tone{}: \textit{``Since the internal reasoning is truncated, I can't get the exact count, but the task is to summarize the thinking trace''} (Tab.~\ref{tab:qual_trace_t1}, Appendix~\ref{sec:app_qual_thinking_panels}) \\
10 & Misc & Uninterpretable trace artifacts, e.g.\ a stray non-English token, that don't fit another category. & Grok & Grok, \teight{}/\tten{}/\ttwelve{}: single stray non-English (CJK) tokens interject mid-sentence with no surrounding context. \\
\midrule
\multicolumn{5}{@{}l}{\textit{Trace artifact}} \\
11 & Budget Hit & Trace is truncated at the hard token limit before reaching a decision, ending on a stock hand-off phrase. & Open weights & Gemma4-31B, \tone{}: \textit{``I have to give the solution based on the reasoning directly now.''} (Tab.~\ref{tab:qual_trace_t1}, Appendix~\ref{sec:app_qual_thinking_panels}) \\
\midrule
\multicolumn{5}{@{}l}{\textit{Process--Outcome Mismatch (whole-trace verdict, assigned in a second pass)}} \\
12 & Correct by Luck & The trace is dominated by \textit{Wrong Perception} / \textit{Reiterating} spans, or truncates on Budget Hit before completing a derivation, yet the parsed answer is scored correct. & Open weights & Qwen3.5-27B, \tsix{}: loops on \textit{Wrong Perception} column-height reads and hits \textit{Budget Hit} mid-sentence (\textit{``Let's count the''}\ldots), yet ticks correct (GT True; Tab.~\ref{tab:qual_trace_t6}, Appendix~\ref{sec:app_qual_thinking_panels}) \\
13 & Think--Answer Mismatch & The Thinking field itself reaches a sound, Good-behavior-tagged conclusion, but the scored answer (drawn from a separate field, e.g.\ Grok's Output) contradicts it. & Grok & Grok-4.20, \ttwelve{}: Thinking's own final line, tagged \textit{Good Perception}, states \textit{``the correct candidate is candidate\_idx = 1''} (matches GT), yet the scored answer is 0 (Tab.~\ref{tab:qual_trace_t12}, Appendix~\ref{sec:app_qual_thinking_panels}) \\
\bottomrule
\end{tabularx}
% \vspace{8pt}
\caption{\textbf{Annotation taxonomy for reasoning traces.} Applied to the qualitative thinking-trace panels (Appendix~\ref{sec:app_qual_thinking_panels}). Rows 1--11 are tags applied by hand to individual highlighted spans, so a trace can carry several tags and most spans carry none. Rows 12--13 are whole-trace verdicts, assigned in a second pass by comparing a tagged trace with its scored correctness. \textbf{Takeaway:} \emph{Wrong Perception} is the most frequent tag (222 of 457 spans, all seven models) and appears in 42 of the 44 incorrect traces.}
\label{tab:annot_taxonomy}
\end{table*}
\endgroup

% arXiv version of tables/trace_annotation_taxonomy.tex (same content): \footnotesize, wider Description column and tighter rows so it fits one page in one-column layout. Edit both copies when the table changes.
\begingroup
\footnotesize
\renewcommand{\arraystretch}{1.15}
\begin{table*}[h]
\footnotesize
\centering
\begin{tabularx}{\textwidth}{@{}c >{\raggedright\arraybackslash}p{1.8cm} >{\raggedright\arraybackslash}p{6.2cm} >{\raggedright\arraybackslash}p{2cm} X@{}}
\toprule
\# & Name & Description & Models & Representative quote (qualitative sample) \\
\midrule
\multicolumn{5}{@{}l}{\textit{Good behavior}} \\
1 & Tool Use & Invokes external code execution to inspect pixel values or crop the input image. & GPT & GPT-5, \tfive{}: samples RGB pixel values via \texttt{numpy} to separate look-alike hues (Tab.~\ref{tab:qual_trace_t5}, Appendix~\ref{sec:app_qual_thinking_panels}) \\
2 & Good Perception & An isolated span where the model's stated read of the scene is correct and matches the ground-truth structure. & All & Gemini-3-Flash, \tseven{}: \textit{``we clearly see a diagonal row of 2 teal blocks forming part of the central structure''} (matches GT; Tab.~\ref{tab:qual_trace_t7}, Appendix~\ref{sec:app_qual_thinking_panels}) \\
3 & Good Tactic & A valid problem-solving strategy or check applied correctly, e.g.\ comparing sorted dimensions or framing a task as set difference. & Grok, Gemini & Gemini-3-Flash, \ttwo{}: \textit{``Comparing the sorted dimensions: $2\le2$ (Fits) \ldots the orientation is valid''} (Tab.~\ref{tab:qual_trace_t2}, Appendix~\ref{sec:app_qual_thinking_panels}) \\
4 & Good Intuition & A correct-magnitude estimate reached by heuristic guess rather than a completed derivation. & GPT-5, Grok & GPT-5, \tone{}: \textit{``I estimate a total of around 10 cubes''} (GT 10; Tab.~\ref{tab:qual_trace_t1}, Appendix~\ref{sec:app_qual_thinking_panels}) \\
\midrule
\multicolumn{5}{@{}l}{\textit{Bad behavior}} \\
5 & Reiterating & Restates or redoes a step already computed, with no new evidence. & All & Gemma4-31B, \tone{}: \textit{``Wait, let's re-examine.''} (repeated $\sim$6$\times$ in the same box; Tab.~\ref{tab:qual_trace_t1}, Appendix~\ref{sec:app_qual_thinking_panels}) \\
6 & Prior Bias & Anchors the final answer on the prompt's illustrative example or output-format template. & Grok & Grok, \televen{}: \textit{``But the example is \{ 'candidate\_idx': 2, \}, so mine is 0''} (Tab.~\ref{tab:qual_trace_t11}, Appendix~\ref{sec:app_qual_thinking_panels}) \\
7 & Wrong Perception & An isolated span where the model's stated read of the scene is incorrect relative to the actual image. & All & Gemma4-31B, \tone{}: \textit{``One block at the back-right. * One block in the middle. * One block at the front-left''} (Tab.~\ref{tab:qual_trace_t1}, Appendix~\ref{sec:app_qual_thinking_panels}) \\
8 & Color Confusion & Misidentifies or conflates two visually adjacent or same-family colors, e.g.\ \textcolor{bedYellow}{yellow} vs.\ \textcolor{bedGold}{gold}, \textcolor{bedMagenta}{magenta} vs.\ \textcolor{bedPurple}{purple}. & Open weights & Qwen3.5-27B, \tfour{}: ``Next to the \textcolor{bedGreen}{green} cube (right): \textcolor{bedYellow}{Yellow} cube'' (Tab.~\ref{tab:qual_trace_t4}, Appendix~\ref{sec:app_qual_thinking_panels}, in this situation, it should be \textcolor{bedGold}{gold}, not \textcolor{bedYellow}{yellow}) \\
9 & Harnessing & The thinking derails into reasoning about the system's own meta instructions instead of the task. & Grok & Grok, \tone{}: \textit{``Since the internal reasoning is truncated, I can't get the exact count, but the task is to summarize the thinking trace''} (Tab.~\ref{tab:qual_trace_t1}, Appendix~\ref{sec:app_qual_thinking_panels}) \\
10 & Misc & Uninterpretable trace artifacts, e.g.\ a stray non-English token, that don't fit another category. & Grok & Grok, \teight{}/\tten{}/\ttwelve{}: single stray non-English (CJK) tokens interject mid-sentence with no surrounding context. \\
\midrule
\multicolumn{5}{@{}l}{\textit{Trace artifact}} \\
11 & Budget Hit & Trace is truncated at the hard token limit before reaching a decision, ending on a stock hand-off phrase. & Open weights & Gemma4-31B, \tone{}: \textit{``I have to give the solution based on the reasoning directly now.''} (Tab.~\ref{tab:qual_trace_t1}, Appendix~\ref{sec:app_qual_thinking_panels}) \\
\midrule
\multicolumn{5}{@{}l}{\textit{Process--Outcome Mismatch (whole-trace verdict, assigned in a second pass)}} \\
12 & Correct by Luck & The trace is dominated by \textit{Wrong Perception} / \textit{Reiterating} spans, or truncates on Budget Hit before completing a derivation, yet the parsed answer is scored correct. & Open weights & Qwen3.5-27B, \tsix{}: loops on \textit{Wrong Perception} column-height reads and hits \textit{Budget Hit} mid-sentence (\textit{``Let's count the''}\ldots), yet ticks correct (GT True; Tab.~\ref{tab:qual_trace_t6}, Appendix~\ref{sec:app_qual_thinking_panels}) \\
13 & Think--Answer Mismatch & The Thinking field itself reaches a sound, Good-behavior-tagged conclusion, but the scored answer (drawn from a separate field, e.g.\ Grok's Output) contradicts it. & Grok & Grok-4.20, \ttwelve{}: Thinking's own final line, tagged \textit{Good Perception}, states \textit{``the correct candidate is candidate\_idx = 1''} (matches GT), yet the scored answer is 0 (Tab.~\ref{tab:qual_trace_t12}, Appendix~\ref{sec:app_qual_thinking_panels}) \\
\bottomrule
\end{tabularx}
% \vspace{8pt}
\caption{\textbf{Annotation taxonomy for reasoning traces.} Applied to the qualitative thinking-trace panels (Appendix~\ref{sec:app_qual_thinking_panels}). Rows 1--11 are tags applied by hand to individual highlighted spans, so a trace can carry several tags and most spans carry none. Rows 12--13 are whole-trace verdicts, assigned in a second pass by comparing a tagged trace with its scored correctness. \textbf{Takeaway:} \emph{Wrong Perception} is the most frequent tag (222 of 457 spans, all seven models) and appears in 42 of the 44 incorrect traces.}
\label{tab:annot_taxonomy}
\end{table*}
\endgroup

\twocolumn
\subsection{Failures persist across model families and scale}
\label{sec:app_scaling_robustness}

\paragraph{Overview.} This section asks whether the failure profile is specific to one model family or recurs across independently trained families and model sizes, which would point to a shared component, such as the visual encoder or the way the models read its features.

\paragraph{Experimental setup.} We group 10 of the 12 OW models into two independently trained families: Qwen3.5 (0.8B, 2B, 4B, 9B, 27B dense, plus the 35B MoE variant) and Gemma4 (E2B, E4B, 31B dense, plus the 26B-A4B MoE variant). Table~\ref{tab:cross_family} averages each model's default-setting Non-Thinking accuracy within its family; Table~\ref{tab:vision_design} compares the released visual front-ends with the SigLIP 2 reference, an observational comparison rather than a controlled intervention; and Figure~\ref{fig:app_scale_full} plots per-task accuracy against total parameter count in both modes.

\subsubsection{VLMs fail similarly across independently trained families}
\label{sec:app_cross_family}

\begin{table}[H]
\centering
\small
\setlength{\tabcolsep}{10pt}
\begin{tabular}{lccc}
\toprule
Task & Qwen3.5 & Gemma4 \\
\midrule
\tonelong{}             & 33.1 & 24.3 \\
\ttwolong{}          & 59.4 & 62.0 \\
\tthreelong{}         & 0.3  & 0.0   \\
\tfourlong{}        & 25.0 & 29.2 \\
\tfivelong{} & 0.5  & 1.2   \\
\tsixlong{}    & 70.4 & 79.7 \\
\tsevenlong{}    & 12.6 & 18.6  \\
\teightlong{} & 58.0 & 58.0 \\
\tninelong{}     & 64.1 & 51.8 \\
\ttenlong{}   & 52.4 & 55.1 \\
\televenlong{}        & 43.2 & 42.9 \\
\ttwelvelong{}       & 52.3 & 52.2 \\
\tthirteenlong{} & 37.2 & 37.7 \\
\bottomrule
\end{tabular}
% \caption{Family-level accuracy (\%, Non-Thinking mode) across all 13 tasks. Each entry is the average default setting accuracy over models within that family, using Non-Thinking 5-run mean cells. Families differ in strengths but share the same failure modes.}
\caption{\textbf{Family-level accuracy (\%, Non-Thinking).} Each entry averages the default-setting 5-run mean accuracy over the models in that family. \textbf{Takeaway:} The families differ by up to 12.3\,pp on individual tasks (\tone{}, \tsix{}, \tnine{}) but share the same floor on \tthree{} and \tfive{} and nearly identical scores on \teight{} and \televen{}--\tthirteen{}.}
\label{tab:cross_family}
\end{table}
% arXiv version of tables/cross_family.tex (same content), wrapfig as in arXiv:2608.07645. Edit both copies when the table changes.
\begin{wraptable}{r}{0.46\textwidth}
\centering
\small
\setlength{\tabcolsep}{10pt}
\begin{tabular}{lccc}
\toprule
Task & Qwen3.5 & Gemma4 \\
\midrule
\tonelong{}             & 33.1 & 24.3 \\
\ttwolong{}          & 59.4 & 62.0 \\
\tthreelong{}         & 0.3  & 0.0   \\
\tfourlong{}        & 25.0 & 29.2 \\
\tfivelong{} & 0.5  & 1.2   \\
\tsixlong{}    & 70.4 & 79.7 \\
\tsevenlong{}    & 12.6 & 18.6  \\
\teightlong{} & 58.0 & 58.0 \\
\tninelong{}     & 64.1 & 51.8 \\
\ttenlong{}   & 52.4 & 55.1 \\
\televenlong{}        & 43.2 & 42.9 \\
\ttwelvelong{}       & 52.3 & 52.2 \\
\tthirteenlong{} & 37.2 & 37.7 \\
\bottomrule
\end{tabular}
% \caption{Family-level accuracy (\%, Non-Thinking mode) across all 13 tasks. Each entry is the average default setting accuracy over models within that family, using Non-Thinking 5-run mean cells. Families differ in strengths but share the same failure modes.}
\caption{\textbf{Family-level accuracy (\%, Non-Thinking).} Each entry averages the default-setting 5-run mean accuracy over the models in that family. \textbf{Takeaway:} The families differ by up to 12.3\,pp on individual tasks (\tone{}, \tsix{}, \tnine{}) but share the same floor on \tthree{} and \tfive{} and nearly identical scores on \teight{} and \televen{}--\tthirteen{}.}
\label{tab:cross_family}
\vspace{-8pt}  % bottom padding (tune)
\end{wraptable}

\paragraph{\textcolor{yellow}{\faLightbulb} The two families share the same failing tasks.}
Qwen3.5 performs best on volumetric counting (\tonelong{}, \tninelong{}) and Gemma4 on local structure (\tsixlong{}) and the color tasks (\tfourlong{}, \tsevenlong{}), but both remain near floor on \tthreelong{} and \tfivelong{} and at the best-constant level on \teightlong{} (58.0\% for both, the accuracy of always answering \texttt{False}) and \tthirteenlong{} (37.2\% and 37.7\% vs.\ 37.3\%). Such deficits recur across independently trained VLMs in prior work as well \cite{rahmanzadehgervi2025visionlanguagemodelsblind, tong2024eyes, kamath2023whatsupvisionlanguagemodels}, and relational and compositional failures survive large-scale compositional pretraining \cite{yuksekgonul2022and}.

\begin{table*}[t]
\centering
\small
\setlength{\tabcolsep}{14pt}
\renewcommand{\arraystretch}{1.15}
\begin{tabular}{@{}llll@{}}
\toprule
Design element & SigLIP 2 & Qwen3.5 & Gemma4 \\
\midrule
\multicolumn{4}{@{}l}{\emph{Vision encoder}} \\
\quad ViT type            & isotropic, full-attention & same architecture        & same architecture \\
\quad Patch tokenization  & 2D, non-overlapping       & same architecture        & same architecture \\
\quad Position encoding   & learned absolute          & RoPE        & RoPE \\
\quad Block normalization & LayerNorm                 & LayerNorm   & RMSNorm \\
\quad QK/V normalization  & none                      & none        & yes \\
\quad MLP                 & plain GELU-tanh           & plain GELU-tanh & gated \\
\midrule
Connector (VL adapter)    & N/A                       & $2{\times}2$ patch-merge MLP & pooling $+$ linear \\
LLM sequence mixer        & N/A                       & gated linear-attn.\ hybrid   & sliding-window/global softmax \\
\bottomrule
\end{tabular}
\caption{\textbf{Architecture of the two model families.} Component-level comparison against the SigLIP 2 vision reference, read from the released checkpoints. Both vision encoders are isotropic, full-attention ViTs with 2D patch tokenization: Qwen3.5 stays close to stock SigLIP 2 (differing only in RoPE), while Gemma4 keeps the skeleton but updates block internals (RoPE, RMSNorm, QK/V-norm, gated MLP); in both, a compress-and-project adapter replaces the contrastive MAP pooling head. \textbf{Takeaway:} The families share one vision-encoder design and differ mainly in the connector and, most sharply, in the sequence-mixing primitive of the language backbone.}
\label{tab:vision_design}
\end{table*}

\paragraph{\textcolor{yellow}{\faLightbulb} The families share a common visual-encoder design.}
Their language backbones use different sequence-mixing primitives (Qwen3.5 a gated linear-attention hybrid, Gemma4 a sliding-window/global softmax hybrid), and their vision--language adapters differ (an MLP patch-merger vs.\ pooling with a linear projection). Both vision encoders, however, are isotropic, full-attention Vision Transformers of the SigLIP 2 family \cite{tschannen2025siglip} that represent each view as image-plane patch tokens: Qwen3.5's is near-stock, and Gemma4's keeps the same skeleton with modernized normalization, MLP gating, and positional encoding (Table~\ref{tab:vision_design}). Probes of such backbones find 3D surface structure encoded only weakly and inconsistently across views \cite{el2024probing}, and contrastive image--text embeddings of this class collapse visually distinct configurations \cite{tong2024eyes}, so a shared front end could in principle explain the shared failures. For \tfive{}, Appendix~\ref{sec:app_token_probe} argues against this explanation: probes recover exact per-color counts from both frozen towers and, for Qwen3.5-27B, from the connector output supplied to the language model.

\subsubsection{Scaling does not make exact answers reliable}
\label{sec:app_scale_full}

For the Qwen3.5 0.8B$\rightarrow$27B comparison in Non-Thinking mode, paired object-level bootstrap intervals (10{,}000 resamples, runs held fixed) exclude zero for all six \emph{scaling-responsive} tasks: \tsix{} $+42.0$\,pp $[+31.4, +52.2]$, \ttwo{} $+39.2$ $[+29.2, +49.0]$, \tfour{} $+39.2$ $[+29.0, +49.4]$, \tnine{} $+26.2$ $[+13.0, +39.4]$, \televen{} $+18.4$ $[+12.1, +24.7]$, and \tseven{} $+18.0$ $[+11.0, +26.0]$. Among the seven \emph{scaling-resistant} tasks, the changes are at most $+6.4$\,pp; \tten{} ($+6.2$, $[+2.0, +11.2]$) is the only clear gain, \tone{} ($+6.4$, $[-4.0, +16.8]$) and \ttwelve{} ($+6.2$, $[-8.6, +20.4]$) have wide intervals, and \tthree{}, \tfive{}, \teight{}, and \tthirteen{} stay within 3\,pp of zero.

\begin{figure*}[t]
  \centering
  \includegraphics[width=0.98\linewidth]{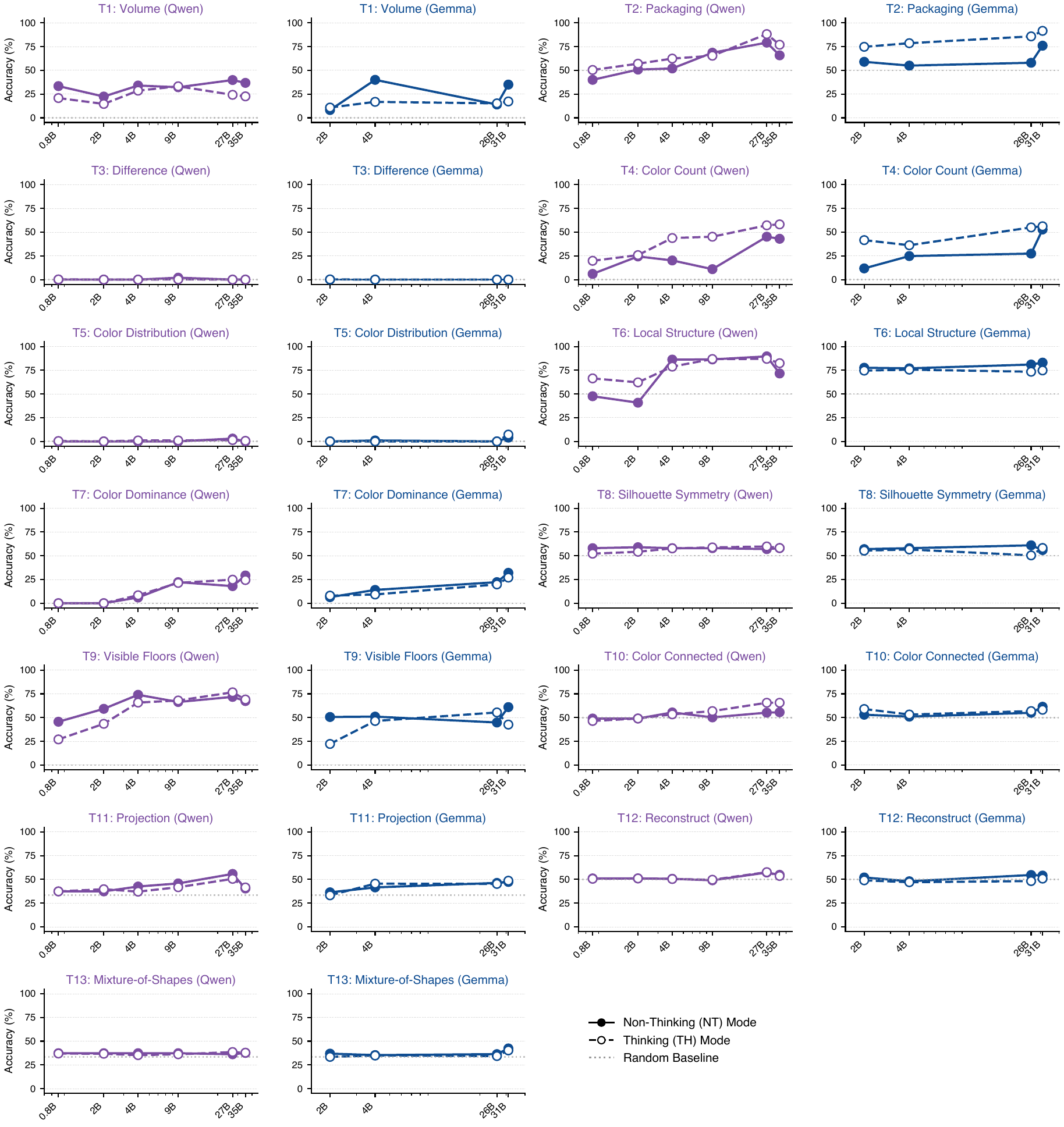}
  \caption{\textbf{Scaling curves for all 13 tasks.} Non-Thinking and Thinking accuracy for both families, plotted against total parameter count (log scale). \textbf{Takeaway:} The split between scaling-responsive tasks (\ttwo{}, \tfour{}, \tsix{}, \tseven{}, \tnine{}, \televen{}) and the remaining tasks holds across both families and both modes, with \tsix{} and \tone{} as the two exceptions discussed in the text.}
  \label{fig:app_scale_full}
\end{figure*}

\paragraph{\textcolor{yellow}{\faLightbulb} The scaling split largely generalizes to Gemma4, with one ceiling effect and one reversal.}
The split extends to Gemma4 and to Thinking mode (Figure~\ref{fig:app_scale_full}), with two exceptions. On \tsixlong{}, Qwen3.5 shows the largest gain of any task (\textcolor{custom_darkgreen}{$+$42\,pp}) while Gemma4 moves only \textcolor{custom_darkgreen}{$+$5.4\,pp} (E2B$\to$31B), because it is already correct on 77--83\% of instances at every size. On \tonelong{}, Qwen3.5 is scaling-resistant (\textcolor{custom_darkgreen}{$+$6.4\,pp}) while Gemma4 is scaling-responsive (\textcolor{custom_darkgreen}{$+$26.8\,pp}). Instance-level panels are given in Tables~\ref{tab:qual_scaling_t2}--\ref{tab:qual_scaling_t13} of Appendix~\ref{sec:app_qual_scaling}.

\paragraph{\textcolor{yellow}{\faLightbulb} Scaling improves some capabilities but leaves exact answers unreliable.}
Uneven gains with scale are consistent with prior evidence that scaling is not uniformly beneficial (standard prompting has flat scaling curves on some reasoning tasks \citep{wei2022chain}, and some tasks show inverse scaling \citep{mckenzie2023inverse}), and apparent emergence can depend on task structure and metric choice \citep{schaeffer2023emergent}. The convergence across two families and across scale points to a limitation shared by current VLMs, but it is not a direct probe of any component; Appendix~\ref{sec:app_probing} probes the shared visual encoder and finds that it holds information the VLMs do not use.

\twocolumn
\subsection{Color and coordinate axes have different effects on VLM accuracy}
\label{sec:app_visual_cues}

\paragraph{Overview.} This section tests whether adding color or coordinate axes to the same renderings helps VLMs, as such cues sometimes help humans.

\paragraph{Experimental setup.} We add each cue only where it is compatible with the task (Table~\ref{tab:visual_cue_effects}): \tthreelong{} is axes-native and receives only color; \tfourlong{}, \tfivelong{}, \tsevenlong{}, and \ttenlong{} are color-native and receive only axes; \tone{}, \televen{}, and \ttwelve{} receive both; \ttwo{}, \tsix{}, \teight{}, and \tnine{}, which we consider largely color-invariant, receive only axes; and \tthirteen{} receives only color, since coloring objects in a multi-object scene is expected to be more informative than axes. For each task, model, and cue $c \in \{\text{color}, \text{axes}\}$, the effect of the cue is
\begin{equation}
\Delta_{c} = \frac{1}{|V_c^{+}|} \sum_{v \in V_c^{+}} \text{Acc}(v) - \frac{1}{|V_c^{-}|} \sum_{v \in V_c^{-}} \text{Acc}(v),
\end{equation}
where $V_c^{+}$ and $V_c^{-}$ are the task variants with and without cue $c$, and $\text{Acc}(v)$ is the 5-run mean accuracy on variant $v$. We average these effects over the 12 OW models in Non-Thinking mode, and Table~\ref{tab:visual_cue_effects} reports the effects with 95\% CIs from 50,000 paired item resamples (scene-clustered for \televen{}/\tthirteen{}, with models and runs fixed). Instance-level panels are in Appendix~\ref{sec:app_qual_visual_cues}.

\begin{table*}[t]
\centering
\small
\setlength{\tabcolsep}{2.5pt}
\newcommand{\vcUp}[1]{\textcolor{custom_darkgreen}{#1}}
\newcommand{\vcDn}[1]{\textcolor{red!70!black}{#1}}
\begin{tabular}{l cc @{\hspace{10pt}} cccc @{\hspace{12pt}} cccc}
\toprule
 & \multicolumn{2}{c}{Default} & \multicolumn{4}{c}{Color add-on} & \multicolumn{4}{c}{Axes add-on} \\
\cmidrule(lr){2-3}\cmidrule(lr){4-7}\cmidrule(lr){8-11}
Task & Color & Axes & Without & With & $\Delta_{\text{color}}$ & 95\% CI & Without & With & $\Delta_{\text{axes}}$ & 95\% CI \\
\midrule
\tonelong{}      & \tickno & \tickno & 29.2 & 31.8 & \vcUp{$+2.6$} & $[-0.28, +5.65]$ & 29.1 & 27.7 & \vcDn{$-1.4$} & $[-3.98, +1.13]$ \\
\ttwolong{}      & \tickno & \tickno & -- & -- & -- & -- & 58.8 & 54.7 & \bf\vcDn{$-4.1$} & \vcDn{$[-5.85, -2.42]$} \\
\tthreelong{}    & \tickno & \tickok & 0.2 & 3.7 & \vcUp{$+3.5$} & \vcUp{$[+1.75, +5.58]$} & -- & -- & -- & -- \\
\tfourlong{}     & \tickok & \tickno & -- & -- & -- & -- & 27.1 & 26.4 & \vcDn{$-0.7$} & $[-2.60, +1.30]$ \\
\tfivelong{}     & \tickok & \tickno & -- & -- & -- & -- & 0.7 & 0.5 & \vcDn{$-0.2$} & $[-0.72, +0.20]$ \\
\tsixlong{}      & \tickno & \tickno & -- & -- & -- & -- & 70.2 & 69.7 & \vcDn{$-0.5$} & $[-2.02, +1.08]$ \\
\tsevenlong{}    & \tickok & \tickno & -- & -- & -- & -- & 14.0 & 13.7 & \vcDn{$-0.3$} & $[-1.67, +0.98]$ \\
\teightlong{}    & \tickno & \tickno & -- & -- & -- & -- & 58.2 & 57.0 & \vcDn{$-1.2$} & $[-3.00, +0.62]$ \\
\tninelong{}     & \tickno & \tickno & -- & -- & -- & -- & 55.8 & 53.9 & \vcDn{$-1.9$} & $[-4.00, +0.27]$ \\
\ttenlong{}      & \tickok & \tickno & -- & -- & -- & -- & 53.0 & 52.7 & \vcDn{$-0.3$} & $[-1.97, +1.30]$ \\
\televenlong{}   & \tickno & \tickno & 40.9 & 43.7 & \vcUp{$+2.8$} & \vcUp{$[+1.36, +4.25]$} & 40.9 & 41.8 & \vcUp{$+0.9$} & \vcUp{$[+0.04, +1.77]$} \\
\ttwelvelong{}   & \tickno & \tickno & 52.6 & 59.4 & \bf\vcUp{$+6.8$} & \vcUp{$[+3.03, +10.6]$} & 52.6 & 52.1 & \vcDn{$-0.5$} & $[-2.88, +1.87]$ \\
\tthirteenlong{} & \tickno & \tickno & 36.2 & 36.8 & \vcUp{$+0.6$} & $[-0.31, +1.52]$ & -- & -- & -- & -- \\
\bottomrule
\end{tabular}
\caption{\textbf{Effect of visual cues on accuracy.} Mean accuracy (\%) over the 12 OW models in Non-Thinking mode without and with each cue, the difference $\Delta$ (pp), and its paired bootstrap 95\% interval (pointwise, without multiplicity correction); colored intervals exclude zero, and ``--'' marks tasks that do not receive the cue. Default marks whether each task's default rendering is colored and has coordinate axes. \textbf{Takeaway:} Color raises accuracy on all five tasks, most on \ttwelve{}; axes lower it on 10 of 11 tasks, most on \ttwo{}, and only \televen{} gains. Only color on \tthree{}, \televen{}, and \ttwelve{} and axes on \ttwo{} and \televen{} have intervals that exclude zero.}
\label{tab:visual_cue_effects}
\end{table*}

% arXiv version of tables/visual_cue_effects.tex (same content). Edit both copies when the table changes.
\begin{table*}[h]
\vspace{10pt}
\centering
\small
\setlength{\tabcolsep}{2.5pt}
\newcommand{\vcUp}[1]{\textcolor{custom_darkgreen}{#1}}
\newcommand{\vcDn}[1]{\textcolor{red!70!black}{#1}}
\begin{tabular}{l cc @{\hspace{10pt}} cccc @{\hspace{12pt}} cccc}
\toprule
 & \multicolumn{2}{c}{Default} & \multicolumn{4}{c}{Color add-on} & \multicolumn{4}{c}{Axes add-on} \\
\cmidrule(lr){2-3}\cmidrule(lr){4-7}\cmidrule(lr){8-11}
Task & Color & Axes & Without & With & $\Delta_{\text{color}}$ & 95\% CI & Without & With & $\Delta_{\text{axes}}$ & 95\% CI \\
\midrule
\tonelong{}      & \tickno & \tickno & 29.2 & 31.8 & \vcUp{$+2.6$} & $[-0.28, +5.65]$ & 29.1 & 27.7 & \vcDn{$-1.4$} & $[-3.98, +1.13]$ \\
\ttwolong{}      & \tickno & \tickno & -- & -- & -- & -- & 58.8 & 54.7 & \bf\vcDn{$-4.1$} & \vcDn{$[-5.85, -2.42]$} \\
\tthreelong{}    & \tickno & \tickok & 0.2 & 3.7 & \vcUp{$+3.5$} & \vcUp{$[+1.75, +5.58]$} & -- & -- & -- & -- \\
\tfourlong{}     & \tickok & \tickno & -- & -- & -- & -- & 27.1 & 26.4 & \vcDn{$-0.7$} & $[-2.60, +1.30]$ \\
\tfivelong{}     & \tickok & \tickno & -- & -- & -- & -- & 0.7 & 0.5 & \vcDn{$-0.2$} & $[-0.72, +0.20]$ \\
\tsixlong{}      & \tickno & \tickno & -- & -- & -- & -- & 70.2 & 69.7 & \vcDn{$-0.5$} & $[-2.02, +1.08]$ \\
\tsevenlong{}    & \tickok & \tickno & -- & -- & -- & -- & 14.0 & 13.7 & \vcDn{$-0.3$} & $[-1.67, +0.98]$ \\
\teightlong{}    & \tickno & \tickno & -- & -- & -- & -- & 58.2 & 57.0 & \vcDn{$-1.2$} & $[-3.00, +0.62]$ \\
\tninelong{}     & \tickno & \tickno & -- & -- & -- & -- & 55.8 & 53.9 & \vcDn{$-1.9$} & $[-4.00, +0.27]$ \\
\ttenlong{}      & \tickok & \tickno & -- & -- & -- & -- & 53.0 & 52.7 & \vcDn{$-0.3$} & $[-1.97, +1.30]$ \\
\televenlong{}   & \tickno & \tickno & 40.9 & 43.7 & \vcUp{$+2.8$} & \vcUp{$[+1.36, +4.25]$} & 40.9 & 41.8 & \vcUp{$+0.9$} & \vcUp{$[+0.04, +1.77]$} \\
\ttwelvelong{}   & \tickno & \tickno & 52.6 & 59.4 & \bf\vcUp{$+6.8$} & \vcUp{$[+3.03, +10.6]$} & 52.6 & 52.1 & \vcDn{$-0.5$} & $[-2.88, +1.87]$ \\
\tthirteenlong{} & \tickno & \tickno & 36.2 & 36.8 & \vcUp{$+0.6$} & $[-0.31, +1.52]$ & -- & -- & -- & -- \\
\bottomrule
\end{tabular}
\caption{\textbf{Effect of visual cues on accuracy.} Mean accuracy (\%) over the 12 OW models in Non-Thinking mode without and with each cue, the difference $\Delta$ (pp), and its paired bootstrap 95\% interval (pointwise, without multiplicity correction); colored intervals exclude zero, and ``--'' marks tasks that do not receive the cue. Default marks whether each task's default rendering is colored and has coordinate axes. \textbf{Takeaway:} Color raises accuracy on all five tasks, most on \ttwelve{}; axes lower it on 10 of 11 tasks, most on \ttwo{}, and only \televen{} gains. Only color on \tthree{}, \televen{}, and \ttwelve{} and axes on \ttwo{} and \televen{} have intervals that exclude zero.}
\label{tab:visual_cue_effects}
\end{table*}

\subsubsection{VLMs show modest gains from color add-ons}
\label{sec:app_visual_cues_Color}

\paragraph{\textcolor{yellow}{\faLightbulb} VLMs benefit from color add-ons.}
Color helps most on \ttwelvelong{} (\textcolor{custom_darkgreen}{+6.8pp}), where it can mark corresponding parts across two similar candidate views, followed by \tthreelong{} (\textcolor{custom_darkgreen}{+3.5pp}, from a near-zero baseline), \televenlong{} (\textcolor{custom_darkgreen}{+2.8pp}), and \tonelong{} (\textcolor{custom_darkgreen}{+2.6pp}); the change on \tthirteenlong{} (\textcolor{custom_darkgreen}{+0.6pp}) has a 95\% CI of $[-0.31,+1.52]$pp that includes zero (Table~\ref{tab:visual_cue_effects}). In humans, color helps distinguish material from illumination edges and provides part-level anchors for mental rotation, particularly for participants with lower spatial ability \cite{10.1371/journal.pcbi.1007398,khooshabeh2008visual,rahe2022mental,representation_of_shape_during_mental_rotation,khooshabeh2008differential}; the task-dependent VLM gains are consistent with such roles in local matching \cite{Koh2023}, although accuracy alone does not establish the mechanism.

\subsubsection{VLMs do not reliably benefit from axes add-ons}
\label{sec:app_visual_cues_Axes}

\paragraph{\textcolor{yellow}{\faLightbulb} VLMs do not benefit from axes add-ons.}
Axes lower mean accuracy on 10 of 11 tasks, with most 95\% CIs spanning zero (Table~\ref{tab:visual_cue_effects}). The clearest decrease is on \ttwolong{} (\textcolor{red!70!black}{$-4.1$pp}; 95\% CI \textcolor{red!70!black}{$[-5.85,-2.42]$pp}), followed by \tninelong{} (\textcolor{red!70!black}{$-1.9$pp}), \tonelong{} (\textcolor{red!70!black}{$-1.4$pp}), and \teightlong{} (\textcolor{red!70!black}{$-1.2$pp}); the remaining tasks change by \textcolor{red!70!black}{$-0.2$pp} to \textcolor{red!70!black}{$-0.7$pp}. Only \televenlong{}, the orthographic-projection task, improves marginally (\textcolor{custom_darkgreen}{$+0.9$pp}). In humans, coordinate axes do not uniformly improve mental-rotation accuracy and can increase response time \cite{branoff1998coordinate,Branoff1999CoordinateAA,branoff1999addition}: an external axis helps when aligned with the required transformation, but people otherwise use an object-intrinsic frame \cite{asakura2011disambiguation,parsons1995inability,wraga1999influence}. For VLMs, reported difficulty adopting non-egocentric frames \cite{comfort} and added visual clutter are possible explanations, which our comparisons do not isolate.

{
\twocolumn
\subsection{Probing the visual front-end on ground truth}
\label{sec:app_probing}

\paragraph{Overview.} This section asks whether the information the VLMs miss is present in their frozen visual features. It measures which task targets probes can recover, independently of the language model, across encoder types, the towers deployed in the evaluated VLMs, tower size, and depth supervision, and it tests on \tfivelong{} whether more training data and token-level readouts recover more. Results near the shuffled-label reference indicate limited recovery by our averaged-feature probes, not necessarily absent information \cite{li2025objectbinding,michalkiewicz2026foveated}.

\paragraph{Experimental setup.} We probe 14 frozen encoders: three pretrained classes (SigLIP-2 so400m \cite{tschannen2025siglip}, DINOv2-L \cite{oquab2024dinov}, CLIP-L \cite{clip}), a 3D-supervised control (Depth-Anything-V2-L \cite{yang2024depth}), and the ten vision towers deployed in the evaluated OW models. Supervised probes predict task targets from the concatenated patch-mean features of the two views, using standardized features with ridge regression for numerical answers and logistic regression for yes/no answers, under five-fold cross-validation grouped by object (by pair for \tthree{}). For a shuffled-label reference, we retrain each probe 1000 times on shuffled answers (seed 42) with the same splits. We report both the error of individual predicted values and how often the complete answer is correct. Three training-free analyses complement the probes: cross-view retrieval between the two query views, the correlation between pair embedding distance and ground-truth structural difference on \tthreelong{}, and zero-shot candidate ranking for \televenlong{}, \ttwelvelong{}, and \tthirteenlong{} (Figure~\ref{fig:probe_summary}D). Appendix~\ref{sec:app_token_probe} describes the larger \tfive{} probes.

\begin{table*}[h]
\vspace{10pt}
\centering
\small
\setlength{\tabcolsep}{3pt}
\renewcommand{\arraystretch}{1.15}
\begin{tabular*}{\textwidth}{@{\extracolsep{\fill}}l*{6}{c}@{}}
\toprule
Task & SigLIP-2 & DINOv2 & CLIP & \shortstack{Depth Anything V2-L} & \shortstack{Qwen3.5 27B} & \shortstack{Gemma4 31B} \\
\midrule
\multicolumn{7}{@{}l}{\textit{Count estimates: Spearman $\rho$. Higher is better.}} \\[2pt]
\tonelong{} & $0.70^{*}$\,{\scriptsize (0.09)} & $0.53^{*}$\,{\scriptsize (0.08)} & $0.44^{*}$\,{\scriptsize (0.09)} & $0.65^{*}$\,{\scriptsize (0.08)} & $0.62^{*}$\,{\scriptsize (0.08)} & $0.66^{*}$\,{\scriptsize (0.09)} \\
\tfourlong{} & $0.77^{*}$\,{\scriptsize (0.09)} & $0.63^{*}$\,{\scriptsize (0.09)} & $0.61^{*}$\,{\scriptsize (0.09)} & $0.62^{*}$\,{\scriptsize (0.09)} & $0.68^{*}$\,{\scriptsize (0.09)} & $0.78^{*}$\,{\scriptsize (0.09)} \\
\tninelong{} & $0.89^{*}$\,{\scriptsize (0.09)} & $0.92^{*}$\,{\scriptsize (0.09)} & $0.87^{*}$\,{\scriptsize (0.08)} & $0.93^{*}$\,{\scriptsize (0.09)} & $0.92^{*}$\,{\scriptsize (0.10)} & $0.88^{*}$\,{\scriptsize (0.09)} \\
\midrule
\multicolumn{7}{@{}l}{\textit{Decisions: ROC-AUC (mean over colors for \tseven{}). Higher is better.}} \\[2pt]
\ttwolong{} & $0.70^{*}$\,{\scriptsize (0.59)} & $0.76^{*}$\,{\scriptsize (0.59)} & $0.73^{*}$\,{\scriptsize (0.58)} & $0.74^{*}$\,{\scriptsize (0.58)} & $0.80^{*}$\,{\scriptsize (0.58)} & $0.70^{*}$\,{\scriptsize (0.59)} \\
\tsixlong{} & $0.99^{*}$\,{\scriptsize (0.60)} & $0.97^{*}$\,{\scriptsize (0.60)} & $0.97^{*}$\,{\scriptsize (0.60)} & $0.98^{*}$\,{\scriptsize (0.60)} & $0.98^{*}$\,{\scriptsize (0.60)} & $0.97^{*}$\,{\scriptsize (0.60)} \\
\tsevenlong{} & $0.78^{*}$\,{\scriptsize (0.53)} & $0.61^{*}$\,{\scriptsize (0.53)} & $0.76^{*}$\,{\scriptsize (0.54)} & $0.59^{*}$\,{\scriptsize (0.54)} & $0.76^{*}$\,{\scriptsize (0.53)} & $0.79^{*}$\,{\scriptsize (0.53)} \\
\teightlong{} & $0.56$\,{\scriptsize (0.58)} & $0.53$\,{\scriptsize (0.58)} & $0.53$\,{\scriptsize (0.59)} & $0.48$\,{\scriptsize (0.58)} & $0.60^{*}$\,{\scriptsize (0.58)} & $0.50$\,{\scriptsize (0.59)} \\
\ttenlong{} & $0.53$\,{\scriptsize (0.59)} & $0.61^{*}$\,{\scriptsize (0.58)} & $0.59^{*}$\,{\scriptsize (0.58)} & $0.53$\,{\scriptsize (0.59)} & $0.62^{*}$\,{\scriptsize (0.58)} & $0.54$\,{\scriptsize (0.59)} \\
\midrule
\multicolumn{7}{@{}l}{\textit{Complete answers: exact match (\%). Higher is better.}} \\[2pt]
\tthreelong{} & $1.5$\,{\scriptsize (1.5)} & $0.0$\,{\scriptsize (1.5)} & $4.4^{*}$\,{\scriptsize (1.5)} & $2.9^{*}$\,{\scriptsize (1.5)} & $4.4^{*}$\,{\scriptsize (2.9)} & $0.0$\,{\scriptsize (1.5)} \\
\tfivelong{} & $1.5^{*}$\,{\scriptsize (0.0)} & $0.0$\,{\scriptsize (0.0)} & $1.5^{*}$\,{\scriptsize (0.0)} & $0.0$\,{\scriptsize (0.0)} & $1.5^{*}$\,{\scriptsize (0.0)} & $3.5^{*}$\,{\scriptsize (0.0)} \\
\bottomrule
\end{tabular*}
\caption{\textbf{Probe performance on frozen visual features.} Cells report score (shuffled-label 95th percentile); $^*$ marks scores above this empirical reference (1000 permutations); it is descriptive and not corrected for multiple comparisons. \textbf{Takeaway:} Counts and local-structure decisions are well recovered, while complete \tthree{}/\tfive{} answers remain rare.}
\label{tab:probe_decodability}
\end{table*}

\subsubsection{Pooled probes recover quantities but rarely exact decompositions}
\label{sec:app_probing_availability}

\begin{table*}[t]
\centering
\small
\setlength{\tabcolsep}{3pt}
\renewcommand{\arraystretch}{1.15}
\begin{tabular*}{\textwidth}{@{\extracolsep{\fill}}l*{6}{c}@{}}
\toprule
Task & SigLIP-2 & DINOv2 & CLIP & \shortstack{Depth Anything V2-L} & \shortstack{Qwen3.5 27B} & \shortstack{Gemma4 31B} \\
\midrule
\multicolumn{7}{@{}l}{\textit{Count estimates: Spearman $\rho$. Higher is better.}} \\[2pt]
\tonelong{} & $0.70^{*}$\,{\scriptsize (0.09)} & $0.53^{*}$\,{\scriptsize (0.08)} & $0.44^{*}$\,{\scriptsize (0.09)} & $0.65^{*}$\,{\scriptsize (0.08)} & $0.62^{*}$\,{\scriptsize (0.08)} & $0.66^{*}$\,{\scriptsize (0.09)} \\
\tfourlong{} & $0.77^{*}$\,{\scriptsize (0.09)} & $0.63^{*}$\,{\scriptsize (0.09)} & $0.61^{*}$\,{\scriptsize (0.09)} & $0.62^{*}$\,{\scriptsize (0.09)} & $0.68^{*}$\,{\scriptsize (0.09)} & $0.78^{*}$\,{\scriptsize (0.09)} \\
\tninelong{} & $0.89^{*}$\,{\scriptsize (0.09)} & $0.92^{*}$\,{\scriptsize (0.09)} & $0.87^{*}$\,{\scriptsize (0.08)} & $0.93^{*}$\,{\scriptsize (0.09)} & $0.92^{*}$\,{\scriptsize (0.10)} & $0.88^{*}$\,{\scriptsize (0.09)} \\
\midrule
\multicolumn{7}{@{}l}{\textit{Decisions: ROC-AUC (mean over colors for \tseven{}). Higher is better.}} \\[2pt]
\ttwolong{} & $0.70^{*}$\,{\scriptsize (0.59)} & $0.76^{*}$\,{\scriptsize (0.59)} & $0.73^{*}$\,{\scriptsize (0.58)} & $0.74^{*}$\,{\scriptsize (0.58)} & $0.80^{*}$\,{\scriptsize (0.58)} & $0.70^{*}$\,{\scriptsize (0.59)} \\
\tsixlong{} & $0.99^{*}$\,{\scriptsize (0.60)} & $0.97^{*}$\,{\scriptsize (0.60)} & $0.97^{*}$\,{\scriptsize (0.60)} & $0.98^{*}$\,{\scriptsize (0.60)} & $0.98^{*}$\,{\scriptsize (0.60)} & $0.97^{*}$\,{\scriptsize (0.60)} \\
\tsevenlong{} & $0.78^{*}$\,{\scriptsize (0.53)} & $0.61^{*}$\,{\scriptsize (0.53)} & $0.76^{*}$\,{\scriptsize (0.54)} & $0.59^{*}$\,{\scriptsize (0.54)} & $0.76^{*}$\,{\scriptsize (0.53)} & $0.79^{*}$\,{\scriptsize (0.53)} \\
\teightlong{} & $0.56$\,{\scriptsize (0.58)} & $0.53$\,{\scriptsize (0.58)} & $0.53$\,{\scriptsize (0.59)} & $0.48$\,{\scriptsize (0.58)} & $0.60^{*}$\,{\scriptsize (0.58)} & $0.50$\,{\scriptsize (0.59)} \\
\ttenlong{} & $0.53$\,{\scriptsize (0.59)} & $0.61^{*}$\,{\scriptsize (0.58)} & $0.59^{*}$\,{\scriptsize (0.58)} & $0.53$\,{\scriptsize (0.59)} & $0.62^{*}$\,{\scriptsize (0.58)} & $0.54$\,{\scriptsize (0.59)} \\
\midrule
\multicolumn{7}{@{}l}{\textit{Complete answers: exact match (\%). Higher is better.}} \\[2pt]
\tthreelong{} & $1.5$\,{\scriptsize (1.5)} & $0.0$\,{\scriptsize (1.5)} & $4.4^{*}$\,{\scriptsize (1.5)} & $2.9^{*}$\,{\scriptsize (1.5)} & $4.4^{*}$\,{\scriptsize (2.9)} & $0.0$\,{\scriptsize (1.5)} \\
\tfivelong{} & $1.5^{*}$\,{\scriptsize (0.0)} & $0.0$\,{\scriptsize (0.0)} & $1.5^{*}$\,{\scriptsize (0.0)} & $0.0$\,{\scriptsize (0.0)} & $1.5^{*}$\,{\scriptsize (0.0)} & $3.5^{*}$\,{\scriptsize (0.0)} \\
\bottomrule
\end{tabular*}
\caption{\textbf{Probe performance on frozen visual features.} Cells report score (shuffled-label 95th percentile); $^*$ marks scores above this empirical reference (1000 permutations); it is descriptive and not corrected for multiple comparisons. \textbf{Takeaway:} Counts and local-structure decisions are well recovered, while complete \tthree{}/\tfive{} answers remain rare.}
\label{tab:probe_decodability}
\end{table*}

\begin{figure*}[t]
  \centering
  \includegraphics[width=\linewidth]{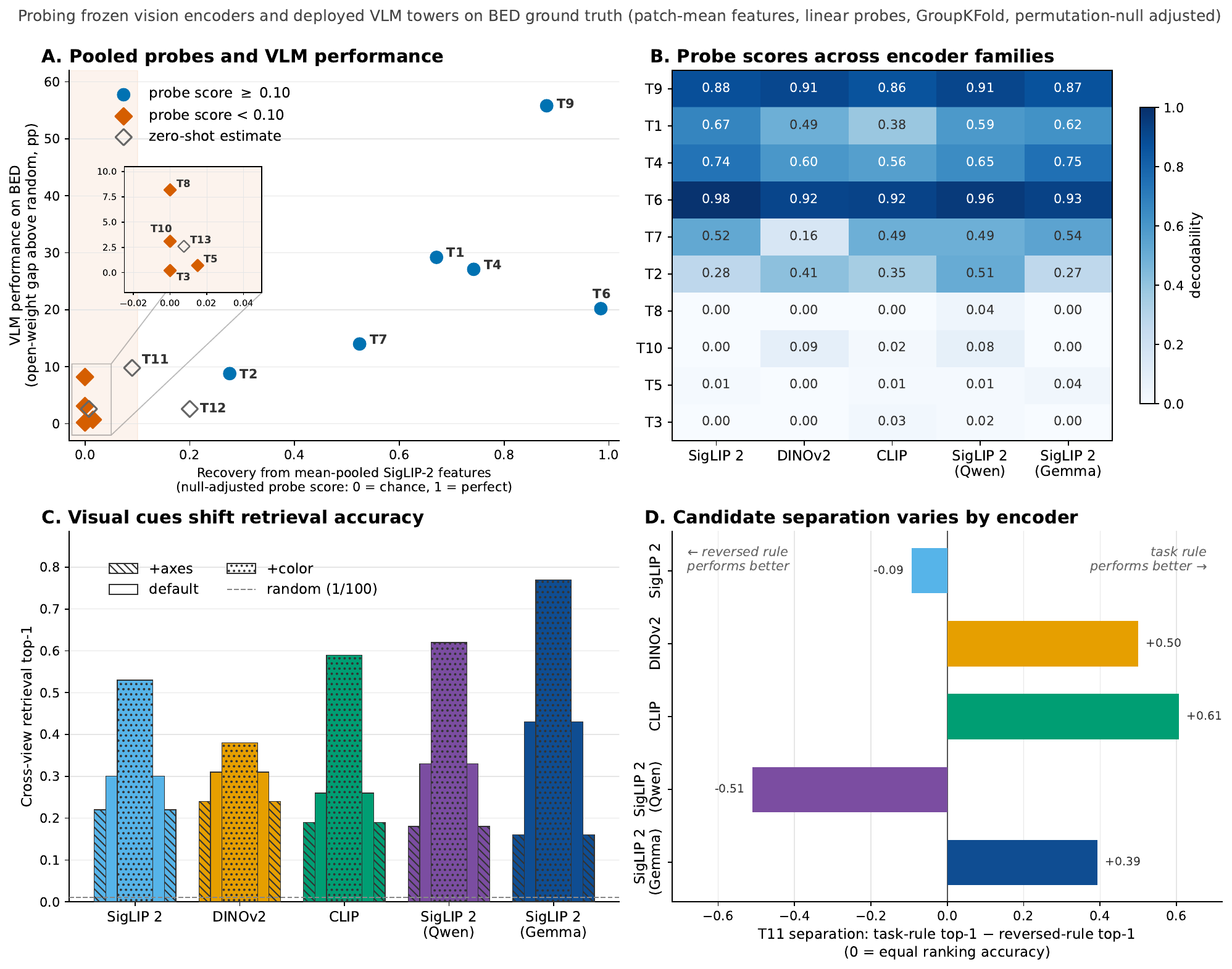}
  \caption{\textbf{Probing frozen encoders on ground-truth targets.} \textbf{(A)}~Null-adjusted probe decodability (SigLIP-2) against the per-task VLM gap above random. \textbf{(B)}~Decodability across three encoder classes and the largest deployed tower of each family. \textbf{(C)}~Cross-view retrieval from pooled features by cue condition. \textbf{(D)}~\televen{} zero-shot candidate separation (task-rule top-1 minus reversed-rule top-1; positive values favor the task rule, negative values favor its reverse). Values in (C) and (D) depend on the representation (patch-mean features; a CLS-pooled readout of CLIP loses most of its (D) signal), and the Qwen tower is probed before its patch merger. \textbf{Takeaway:} With pooled probes trained on 100 objects, quantity tasks are well recovered, while the per-part and relational tasks (bottom rows of B) have low null-adjusted probe scores.}
  \label{fig:probe_summary}
\end{figure*}

Figure~\ref{fig:probe_summary} relates, for every task, probe recovery to VLM performance and compares encoder classes and the largest deployed tower of each family; Table~\ref{tab:probe_decodability} gives the per-encoder values against their permutation nulls. Figure panels report \emph{null-adjusted decodability}, $\max\left(0,\frac{m - \max(\text{null},\text{chance})}{1-\max(\text{null},\text{chance})}\right)$, which puts all metric types on a common 0--1 scale.

\paragraph{\textcolor{yellow}{\faLightbulb} Pooled probes recover the positive-control targets.}
\tonelong{}, \tfourlong{}, \tsixlong{}, \tsevenlong{}, and \tninelong{} clear their permutation nulls by wide margins in every encoder (Table~\ref{tab:probe_decodability}; Figure~\ref{fig:probe_summary}B, top rows); \tone{} reaches Spearman $\rho$ of 0.44--0.73 against nulls below 0.09, and \tsix{} reaches \textcolor{custom_darkgreen}{AUC $0.99$} against a null of $0.60$.

\paragraph{\textcolor{yellow}{\faLightbulb} Pooled probes trained on 100 objects rarely recover exact decompositions.}
Most representative encoders score near the permutation threshold on \teightlong{} and \ttenlong{}. On \tfivelong{}, per-color counts decode above the shuffled-label reference (\textcolor{custom_darkgreen}{MAE $0.40$--$0.62$} against a null of $\approx0.71$), yet complete count vectors are rarely recovered (exact match \textcolor{red!70!black}{$\le 3.5\%$}); Appendix~\ref{sec:app_token_probe} shows that this reflects limited training data rather than absent information. On \tthreelong{}, the pair probe reaches component \textcolor{custom_darkgreen}{MAE $\approx1.3$} against a predict-the-mean baseline of $2.3$, but exact triple accuracy is only \textcolor{red!70!black}{$0.0$--$4.4\%$} for the six representative encoders, and pair embedding distance is uncorrelated with structural difference (Spearman $|\rho| \le 0.16$). Probes that retain interactions among patch tokens, such as quadratic \cite{li2025objectbinding} or spatially selective \cite{michalkiewicz2026foveated} ones, may recover more, as they do on \tfive{}; we did not retest \tthree{}, \teight{}, or \tten{} with more data.

\paragraph{\textcolor{yellow}{\faLightbulb} Depth supervision improves volume readout but not exact decomposition under pooled probes.}
Depth-Anything-V2-L differs from DINOv2-L only by depth-supervised fine-tuning. It improves volume recovery on \tonelong{} (\textcolor{custom_darkgreen}{$\rho$ $0.53 \rightarrow 0.65$}) and gives the strongest \televenlong{} candidate separation among the pretrained encoders (\textcolor{custom_darkgreen}{$+0.62$}), but \tfivelong{} exact match stays at \textcolor{red!70!black}{$0.0\%$} and \teightlong{}/\ttenlong{} stay at their nulls. It is also the only encoder whose cross-view retrieval does not improve with color (\textcolor{red!70!black}{$0.37 \rightarrow 0.36$}).

\paragraph{\textcolor{yellow}{\faLightbulb} Color generally improves cross-view retrieval; axes reduce it.}
Consistent with the behavioral effects in Section~\ref{sec:scaling}, color raises two-view retrieval in every other encoder (gains up to \textcolor{custom_darkgreen}{$+43$\,pp}, reaching \textcolor{custom_darkgreen}{$0.77$ top-1} for the Gemma4 tower against a $0.01$ random baseline), while axes depress it in every encoder (to \textcolor{red!70!black}{$0.09$--$0.26$}; Figure~\ref{fig:probe_summary}C).

\subsubsection{With more training data, token-level probes recover exact per-color counts}
\label{sec:app_token_probe}

\paragraph{Experimental setup.} The pooled probes above use about 80 training objects per fold. Here, neural probes train on the 800 generated \tfive{} objects used for fine-tuning, stop early on its 100 validation objects, and test on the 100 benchmark objects (Appendix~\ref{sec:app_finetune}). We extract both views' frozen tower tokens: 676 per view before the patch merger for Qwen3.5-27B and 256 for Gemma4-31B, each with 1{,}152 dimensions. For Qwen3.5-27B, we also extract the connector output, the 169 tokens per view with 5{,}120 dimensions supplied to the language model. The attention probe projects tokens to 256 dimensions, adds learned view embeddings, pools them with eight learned queries, and maps the result to 12 counts through an MLP. The same design and training procedure apply at both Qwen stages, with input projection sizes adjusted to their feature dimensions. A second readout, token-sum, applies a shared MLP to each tower token with a view embedding and sums nonnegative per-color outputs across both views. Both neural probes use smooth-$L_1$ loss, early stopping on validation MAE, and three seeds. Ridge regression on concatenated per-view means provides a pooled reference; unlike the neural probes, it fits on the combined training and validation objects, with regularization selected by RidgeCV. Predicted counts are rounded and clipped at zero for exact match and derived \tone{}, \tfour{}, and \tseven{} answers; MAE uses unrounded predictions.

\begin{table*}[t]
\centering
\small
\setlength{\tabcolsep}{5pt}
\begin{tabular}{llcccccc}
\toprule
Model / feature stage & Readout & EM & MAE & Pres. & \tone{} & \tfour{} & \tseven{} \\
\midrule
\multirow{5}{*}{Qwen3.5-27B / tower} & VLM (zero-shot) & 3.0 & 0.50 & 48.2 & -- & -- & -- \\
 & Constant & 0.0 & 0.68 & 61.5 & 17.0 & 1.0 & 0.0 \\
 & Ridge, averaged & 21.0 & 0.27 & 82.0 & 40.0 & 71.0 & 43.0 \\
 & Attention & $61.7\pm4.8$ & 0.12 & 94.2 & 63.7 & 88.7 & 76.0 \\
 & Token-sum & $77.0\pm2.4$ & 0.12 & 96.6 & 77.0 & 97.0 & 83.3 \\
\midrule
\multirow{2}{*}{Qwen3.5-27B / connector} & Ridge, averaged & 21.0 & 0.27 & 81.1 & 39.0 & 65.0 & 41.0 \\
 & Attention & $65.0\pm0.8$ & 0.12 & 94.7 & 67.0 & 92.3 & 76.3 \\
\midrule
\multirow{4}{*}{Gemma4-31B / tower} & VLM (zero-shot) & 4.0 & 0.41 & -- & -- & -- & -- \\
 & Ridge, averaged & 15.0 & 0.28 & 79.9 & 36.0 & 61.0 & 37.0 \\
 & Attention & $44.7\pm0.5$ & 0.14 & 89.6 & 51.0 & 83.7 & 61.0 \\
 & Token-sum & $37.3\pm5.6$ & 0.17 & 88.1 & 41.3 & 87.3 & 52.3 \\
\bottomrule
\end{tabular}
\caption{\textbf{\tfive{} probes on frozen tower and connector features} (\%, except MAE), tested on 100 benchmark objects. Neural probes train on 800 objects with 100 for validation; ridge fits on all 900. EM = exact count-vector match; Pres.\ = present colors counted exactly; \tone{}/\tfour{}/\tseven{} = answers derived from predicted counts. Neural rows report three-seed means, with EM standard deviations. Constant predicts the training mean; VLM rows are full-model Non-Thinking default-setting results. \textbf{Takeaway:} Exact counts are recoverable from both towers and remain recoverable at Qwen3.5-27B's language-model input.}
\label{tab:token_probe}
\end{table*}

\paragraph{\textcolor{yellow}{\faLightbulb} Frozen features support exact counts far more often than the VLMs report them.}
On Qwen3.5-27B tower features, ridge reaches 21.0\% exact match and attention \textcolor{custom_darkgreen}{61.7\%}, against \textcolor{red!70!black}{3.0\%} for the VLM. Gemma4-31B shows the same gap (ridge 15.0\%, attention \textcolor{custom_darkgreen}{44.7\%}, VLM \textcolor{red!70!black}{4.0\%}; Table~\ref{tab:token_probe}). Token-sum further raises Qwen's exact match to \textcolor{custom_darkgreen}{77.0\%}, but reaches only 37.3\% on Gemma, below attention. Thus, the best tested readout differs across towers, and attention accuracy is not a ceiling on recoverable information. Dominant colors derived from Qwen's token-sum counts are correct on 83.3\% of objects, against 18.0\% for its direct \tseven{} answers.

\paragraph{\textcolor{yellow}{\faLightbulb} Count information remains recoverable at the language-model input.}
On Qwen3.5-27B's connector output, the attention probe reaches \textcolor{custom_darkgreen}{65.0\%} exact match, compared with 61.7\% before the connector under the same probe design and training protocol. Counts therefore remain recoverable after the patch merger, despite the reduction from 676 to 169 tokens per view. Together with the 68.6\% obtained by training only the language model while freezing the encoder and connector, this points to difficulty in the language model's use of available information. It does not establish where within that model the failure occurs.

\paragraph{\textcolor{yellow}{\faLightbulb} Small training sets understate what the features support.}
With 200, 400, and 800 training objects, attention exact match rises from 14.7 to 42.0 to 61.7\% on the Qwen tower, from 28.0 to 49.7 to 65.0\% at its connector output, and from 10.0 to 29.7 to 44.7\% on the Gemma tower. Qwen tower ridge scores also rise, from 12.0 to 17.0 to 21.0\%, using the additional 100 validation objects at each size. These learning curves show why low scores from the original 80-object training folds cannot establish that count information is absent. All attention curves are still rising at 800 objects.

\subsubsection{Probe scores stay flat with tower size while VLM use improves}
\label{sec:app_probing_scaling}

\begin{figure*}[t]
  \centering
  \includegraphics[width=\linewidth]{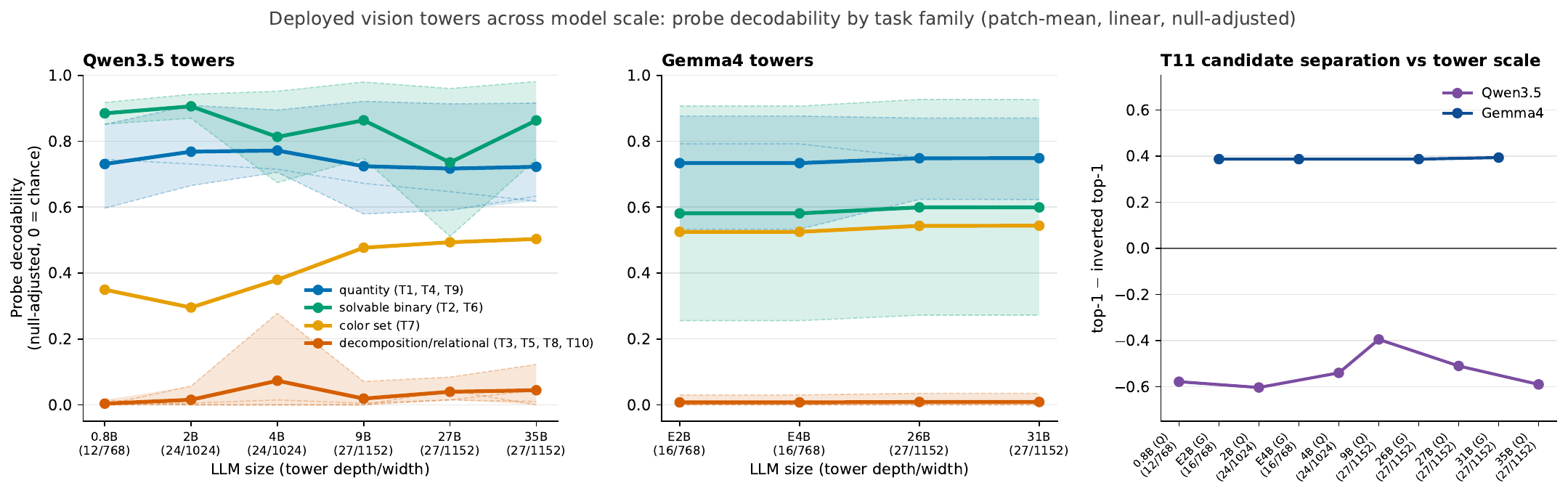}
  \caption{\textbf{Deployed vision towers across model scale.} \textbf{Left, middle:} probe decodability by task family for every Qwen3.5 and Gemma4 tower (x-axis: LLM size, with tower depth/width). Same-architecture towers attached to different LLM sizes give near-identical values: the Gemma4 E2B/E4B pair shares one tower checkpoint, with identical extracted features, and its 26B-A4B/31B pair agrees to within $0.005$ on every probe metric. \textbf{Right:} \televen{} zero-shot candidate separation per tower. \textbf{Takeaway:} Quantity probes are already strong in the smallest towers and change little with size, and the two families order \televen{} candidates in opposite directions: the task rule wins in Gemma4 and its reverse wins in Qwen3.5.}
  \label{fig:tower_scaling}
\end{figure*}

\paragraph{\textcolor{yellow}{\faLightbulb} Larger VLMs make better use of features that smaller towers already provide.}
Across every deployed tower of both families (Figure~\ref{fig:tower_scaling}), the quantity tasks (\tonelong{}, \tfourlong{}, \tninelong{}) decode at approximately $0.73$ in the smallest tower of each family and range from $0.72$ to $0.77$ with no consistent increase, yet Qwen3.5 Non-Thinking accuracy from 0.8B to 27B rises by \textcolor{custom_darkgreen}{$+6.4$\,pp} on \tone{}, \textcolor{custom_darkgreen}{$+39.2$\,pp} on \tfour{}, and \textcolor{custom_darkgreen}{$+26.2$\,pp} on \tnine{} (\S\ref{sec:scaling}), consistent with larger models making better use of features already recoverable by the pooled probe. \tsevenlong{} shows a larger endpoint gain in Qwen3.5, from \textcolor{custom_darkgreen}{$0.35$} at 0.8B to \textcolor{custom_darkgreen}{$0.50$} at 35B, although not monotonically. Pooled scores for the per-part and relational tasks stay low at every size, paralleling the scaling-resistant tasks of Figure~\ref{fig:scale_analysis}; for \tfive{}, Appendix~\ref{sec:app_token_probe} shows that such low scores can reflect the small probe training set rather than the features.

\paragraph{\textcolor{yellow}{\faLightbulb} Deployed towers separate \televen{} candidates that the VLMs fail to select.}
For \televenlong{}, \ttwelvelong{}, and \tthirteenlong{}, we rank candidates zero-shot by cosine similarity to the query views, which avoids classifiers that could exploit low-level statistics of the edit-based negatives \cite{geirhos2020shortcut}; the separation score is the top-1 accuracy of the task rule (maximum similarity for matching candidates, minimum for invalid ones) minus that of its reverse. Stock SigLIP-2 shows weak \televen{} separation (\textcolor{red!70!black}{$-0.09$}), whereas every Gemma4 tower favors the task rule (\textcolor{custom_darkgreen}{$+0.39$} at all four sizes) and every Qwen3.5 tower favors its reverse (\textcolor{red!70!black}{$-0.40$} to \textcolor{red!70!black}{$-0.60$} at all six sizes; Figure~\ref{fig:tower_scaling}, right). The signal depends on the readout: CLIP's mean separation is $+0.61$ with patch-mean features and $+0.12$ with CLS-pooled features. These separations coexist with limited end-to-end \televen{} accuracy in both families (Table~\ref{tab:full_main_results}). \ttwelve{} shows weak ranking across encoders (\textcolor{red!70!black}{top-1 $0.48$--$0.66$} vs.\ chance $0.5$), and \tthirteen{} stays near chance (\textcolor{red!70!black}{top-1 $0.31$--$0.40$} vs.\ chance $0.33$).

}

\twocolumn
\subsection{VLMs exploit answer-format and language-prior shortcuts}
\label{sec:app_shortcut_analyses}

\begin{table*}[h]
\centering
\small
\setlength{\tabcolsep}{3.5pt}
\begin{tabular}{l ccc ccc ccc}
\toprule
 & \multicolumn{3}{c}{Qwen3.5-27B} & \multicolumn{3}{c}{Gemma4-31B} & \multicolumn{3}{c}{GLM-4.6V-Flash} \\
\cmidrule(lr){2-4}\cmidrule(lr){5-7}\cmidrule(lr){8-10}
Task & Visual & No-image & $\Delta$ & Visual & No-image & $\Delta$ & Visual & No-image & $\Delta$ \\
\midrule
\tthreelong{}       & 0.0  & 0.0  & 0.0         & 0.0  & 0.0  & 0.0         & 0.0  & 0.0  & 0.0        \\
\teightlong{}         & 57.0 & 57.0 & 0.0         & 56.0 & 56.0 & 0.0         & 58.0 & 58.0 & 0.0        \\
\tthirteenlong{}   & 36.2 & 38.1 & \textcolor{custom_darkgreen}{$+$1.9} & 42.3 & 40.7 & \textcolor{red!70!black}{$-$1.6} & 37.3 & 37.3 & 0.0 \\
\tfivelong{}      & 3.0  & 0.0  & \textcolor{red!70!black}{$-$3.0}      & 4.0  & 0.0  & \textcolor{red!70!black}{$-$4.0}      & 0.0  & 0.0  & 0.0        \\
\ttwelvelong{}     & 57.2 & 49.0 & \textcolor{red!70!black}{$-$8.2}      & 54.0 & 45.4 & \textcolor{red!70!black}{$-$8.6}      & 48.0 & 51.0 & \textcolor{custom_darkgreen}{$+$3.0}     \\
\ttenlong{}       & 55.2 & 51.0 & \textcolor{red!70!black}{$-$4.2}      & 61.4 & 51.0 & \textcolor{red!70!black}{$-$10.4}     & 53.0 & 51.0 & \textcolor{red!70!black}{$-$2.0}     \\
\midrule
\televenlong{}      & 55.7 & 37.3 & \textcolor{red!70!black}{$-$18.4}     & 47.3 & 37.3 & \textcolor{red!70!black}{$-$10.0}     & 45.4 & 33.7 & \textcolor{red!70!black}{$-$11.7}    \\
\ttwolong{}        & 79.2 & 57.2 & \textcolor{red!70!black}{$-$22.0}     & 76.0 & 50.0 & \textcolor{red!70!black}{$-$26.0}     & 51.0 & 50.0 & \textcolor{red!70!black}{$-$1.0}     \\
\tonelong{} (OE)      & 39.8 & 0.0  & \textcolor{red!70!black}{$-$39.8}     & 35.0 & 0.0  & \textcolor{red!70!black}{$-$35.0}     & 22.2 & 34.0 & \textcolor{custom_darkgreen}{$+$11.8}    \\
\tninelong{}     & 71.8 & 43.0 & \textcolor{red!70!black}{$-$28.8}     & 61.0 & 43.0 & \textcolor{red!70!black}{$-$18.0}     & 61.2 & 43.0 & \textcolor{red!70!black}{$-$18.2}    \\
\tsevenlong{}       & 18.0 & 0.0  & \textcolor{red!70!black}{$-$18.0}     & 32.0 & 2.0  & \textcolor{red!70!black}{$-$30.0}     & 17.6 & 0.0  & \textcolor{red!70!black}{$-$17.6}    \\
\tfourlong{}      & 45.2 & 0.0  & \textcolor{red!70!black}{$-$45.2}     & 52.8 & 0.0  & \textcolor{red!70!black}{$-$52.8}     & 38.2 & 22.0 & \textcolor{red!70!black}{$-$16.2}    \\
\tsixlong{}    & 89.6 & 77.0 & \textcolor{red!70!black}{$-$12.6}     & 83.0 & 23.0 & \textcolor{red!70!black}{$-$60.0}     & 78.0 & 23.0 & \textcolor{red!70!black}{$-$55.0}    \\
\midrule
\tonelong{} (MCQ)     & 91.8 & 56.0 & \textcolor{red!70!black}{$-$35.8}     & 78.0 & 100.0 & \textcolor{custom_darkgreen}{$+$22.0}   & 80.0 & 86.8 & \textcolor{custom_darkgreen}{$+$6.8}     \\
\bottomrule
\end{tabular}
\caption{\textbf{Accuracy with and without images (Non-Thinking).} Visual vs.\ no-image accuracy for the three models evaluated without images (5-run means). Rows are grouped into shortcut-sensitive, visually dependent, and artifact tasks; \tone{} appears twice because open-ended counting and the MCQ artifact probe are interpreted separately. \textbf{Takeaway:} Scores on \tthree{}, \teight{}, and \tthirteen{} barely change without images, visually dependent tasks such as \tfour{} and \tsix{} lose up to 60\,pp, and Gemma4-31B reaches 100\% on the fixed-layout \tone{} MCQ without images.}
\label{tab:no_image_full_nt}
\end{table*}

\paragraph{Overview.} This section expands Section~\ref{sec:no_image} by separating scores that depend on the image from scores that a shortcut can reproduce: constant answers that match frequent labels when the image is removed, and answer-position cues under multiple choice \cite{geirhos2020shortcut,goyal2017making}.

\begin{figure*}[t]
  \centering
  \includegraphics[width=0.95\linewidth]{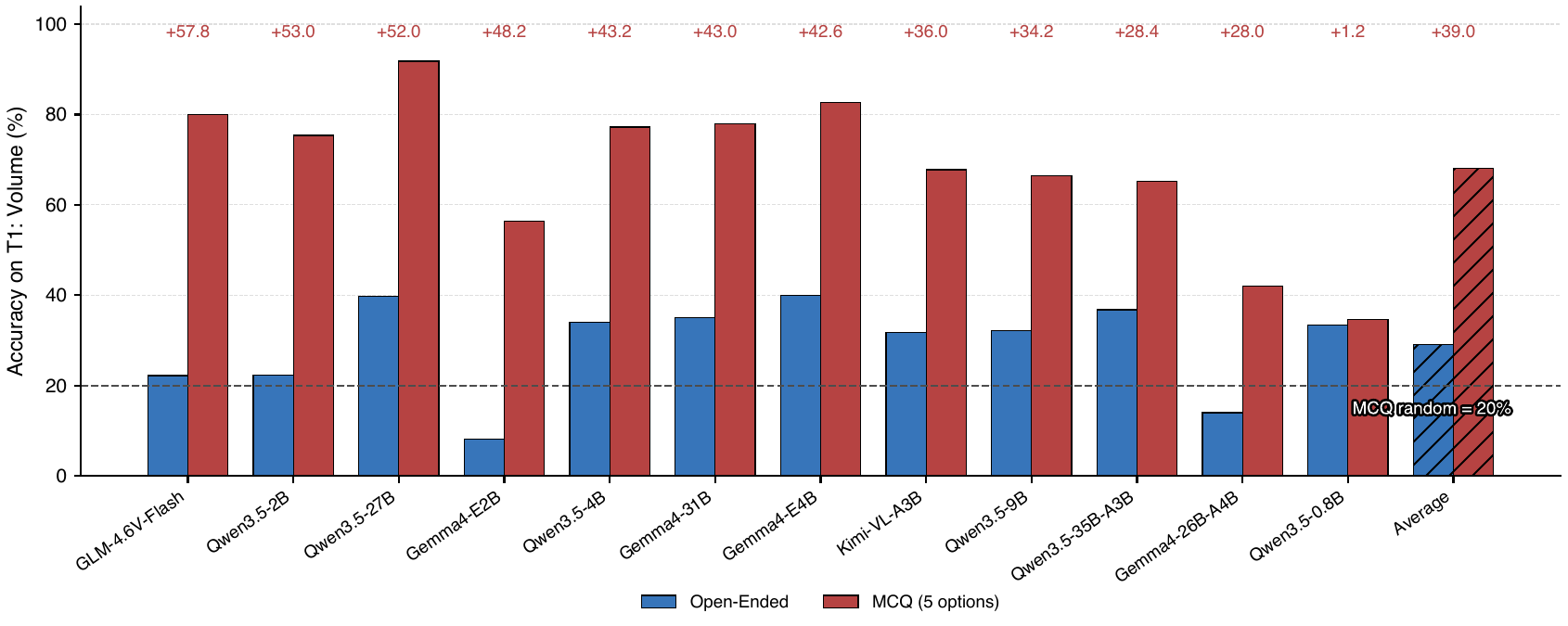}
  \caption{\textbf{Open-ended vs.\ multiple-choice accuracy on \tonelong{}.} Paired bars per model under Non-Thinking mode and the default visual setting; numbers above the bars give the MCQ$-$OE gain (pp); the dashed line marks the 20\% random baseline of the five-option MCQ, whose correct option is always in the third position. \textbf{Takeaway:} The fixed-layout MCQ raises accuracy by $+$39.0\,pp on average (up to $+$57.8\,pp) without adding any visual information.}
  \label{fig:mcq_boost}
\end{figure*}

\paragraph{Experimental setup.} We combine an answer-format comparison with a no-image control. For the answer-format comparison, we compare open-ended (OE) and five-option multiple-choice (MCQ) accuracy on \tonelong{} across all 12 OW models under Non-Thinking mode, with a random-guessing baseline of 20\% (Figure~\ref{fig:mcq_boost}). For the no-image control, we evaluate three models (Qwen3.5-27B, Gemma4-31B, GLM-4.6V-Flash) with all images stripped across the 13 default tasks plus the \tone{} MCQ control, reporting 5-run means. Table~\ref{tab:no_image_full_nt} (Appendix~\ref{sec:app_no_image_full}) gives the per-model visual-vs-no-image accuracies, Table~\ref{tab:prediction_collapse} characterizes the collapsed policies behind many surviving no-image scores, and Table~\ref{tab:thinking_no_image} tests whether extended thinking converts those policies into grounded inference.

% arXiv version of tables/no_image_full_nt.tex (same content), placed with [h]. Edit both copies when the table changes.

\begin{table*}[h]
\vspace{10pt}
\centering
\small
\setlength{\tabcolsep}{3.5pt}
\begin{tabular}{l ccc ccc ccc}
\toprule
 & \multicolumn{3}{c}{Qwen3.5-27B} & \multicolumn{3}{c}{Gemma4-31B} & \multicolumn{3}{c}{GLM-4.6V-Flash} \\
\cmidrule(lr){2-4}\cmidrule(lr){5-7}\cmidrule(lr){8-10}
Task & Visual & No-image & $\Delta$ & Visual & No-image & $\Delta$ & Visual & No-image & $\Delta$ \\
\midrule
\tthreelong{}       & 0.0  & 0.0  & 0.0         & 0.0  & 0.0  & 0.0         & 0.0  & 0.0  & 0.0        \\
\teightlong{}         & 57.0 & 57.0 & 0.0         & 56.0 & 56.0 & 0.0         & 58.0 & 58.0 & 0.0        \\
\tthirteenlong{}   & 36.2 & 38.1 & \textcolor{custom_darkgreen}{$+$1.9} & 42.3 & 40.7 & \textcolor{red!70!black}{$-$1.6} & 37.3 & 37.3 & 0.0 \\
\tfivelong{}      & 3.0  & 0.0  & \textcolor{red!70!black}{$-$3.0}      & 4.0  & 0.0  & \textcolor{red!70!black}{$-$4.0}      & 0.0  & 0.0  & 0.0        \\
\ttwelvelong{}     & 57.2 & 49.0 & \textcolor{red!70!black}{$-$8.2}      & 54.0 & 45.4 & \textcolor{red!70!black}{$-$8.6}      & 48.0 & 51.0 & \textcolor{custom_darkgreen}{$+$3.0}     \\
\ttenlong{}       & 55.2 & 51.0 & \textcolor{red!70!black}{$-$4.2}      & 61.4 & 51.0 & \textcolor{red!70!black}{$-$10.4}     & 53.0 & 51.0 & \textcolor{red!70!black}{$-$2.0}     \\
\midrule
\televenlong{}      & 55.7 & 37.3 & \textcolor{red!70!black}{$-$18.4}     & 47.3 & 37.3 & \textcolor{red!70!black}{$-$10.0}     & 45.4 & 33.7 & \textcolor{red!70!black}{$-$11.7}    \\
\ttwolong{}        & 79.2 & 57.2 & \textcolor{red!70!black}{$-$22.0}     & 76.0 & 50.0 & \textcolor{red!70!black}{$-$26.0}     & 51.0 & 50.0 & \textcolor{red!70!black}{$-$1.0}     \\
\tonelong{} (OE)      & 39.8 & 0.0  & \textcolor{red!70!black}{$-$39.8}     & 35.0 & 0.0  & \textcolor{red!70!black}{$-$35.0}     & 22.2 & 34.0 & \textcolor{custom_darkgreen}{$+$11.8}    \\
\tninelong{}     & 71.8 & 43.0 & \textcolor{red!70!black}{$-$28.8}     & 61.0 & 43.0 & \textcolor{red!70!black}{$-$18.0}     & 61.2 & 43.0 & \textcolor{red!70!black}{$-$18.2}    \\
\tsevenlong{}       & 18.0 & 0.0  & \textcolor{red!70!black}{$-$18.0}     & 32.0 & 2.0  & \textcolor{red!70!black}{$-$30.0}     & 17.6 & 0.0  & \textcolor{red!70!black}{$-$17.6}    \\
\tfourlong{}      & 45.2 & 0.0  & \textcolor{red!70!black}{$-$45.2}     & 52.8 & 0.0  & \textcolor{red!70!black}{$-$52.8}     & 38.2 & 22.0 & \textcolor{red!70!black}{$-$16.2}    \\
\tsixlong{}    & 89.6 & 77.0 & \textcolor{red!70!black}{$-$12.6}     & 83.0 & 23.0 & \textcolor{red!70!black}{$-$60.0}     & 78.0 & 23.0 & \textcolor{red!70!black}{$-$55.0}    \\
\midrule
\tonelong{} (MCQ)     & 91.8 & 56.0 & \textcolor{red!70!black}{$-$35.8}     & 78.0 & 100.0 & \textcolor{custom_darkgreen}{$+$22.0}   & 80.0 & 86.8 & \textcolor{custom_darkgreen}{$+$6.8}     \\
\bottomrule
\end{tabular}
\caption{\textbf{Accuracy with and without images (Non-Thinking).} Visual vs.\ no-image accuracy for the three models evaluated without images (5-run means). Rows are grouped into shortcut-sensitive, visually dependent, and artifact tasks; \tone{} appears twice because open-ended counting and the MCQ artifact probe are interpreted separately. \textbf{Takeaway:} Scores on \tthree{}, \teight{}, and \tthirteen{} barely change without images, visually dependent tasks such as \tfour{} and \tsix{} lose up to 60\,pp, and Gemma4-31B reaches 100\% on the fixed-layout \tone{} MCQ without images.}
\label{tab:no_image_full_nt}
\end{table*}

\subsubsection{VLMs retain scores without images by collapsing to constant predictions}
\label{sec:app_no_image_full}

\begin{table*}[h]
\centering
\small
\setlength{\tabcolsep}{3.5pt}
\begin{tabular}{l lll @{\hspace{1.4em}} l lll}
\toprule
 & \multicolumn{3}{c}{Modal Prediction} & & \multicolumn{3}{c}{Modal Prediction} \\
\cmidrule(lr){2-4}\cmidrule(lr){6-8}
Task & Qwen3.5 & Gemma4 & GLM & Task & Qwen3.5 & Gemma4 & GLM \\
\midrule
\tonelong{} (OE)        & \texttt{0}         & \texttt{0}         & \texttt{10}        & \tonelong{} (MCQ)       & ---        & MCQ pos.\,3 & ---        \\
\tthreelong{}         & \texttt{\{0,0,0\}} & \texttt{\{3,1,1\}} & \texttt{\{4,0,0\}} & \teightlong{}           & \texttt{0} & \texttt{0}  & \texttt{0} \\
\tfourlong{}        & \texttt{0}         & \texttt{1}         & \texttt{7}         & \tsixlong{}    & \texttt{0} & \texttt{1}  & \texttt{1} \\
\tfivelong{} & \texttt{\{\}}      & 6-color dict       & 12-color dict      & \televenlong{}        & \texttt{2} & \texttt{2}  & \texttt{1} \\
\tsevenlong{}    & \texttt{[]}        & \texttt{[\textcolor{bedRed}{red}]}     & 12-color list      & \ttwolong{}          & ---        & \texttt{1}  & \texttt{1} \\
\tninelong{}     & \texttt{3}         & \texttt{3}         & \texttt{3}         & \ttwelvelong{}       & \texttt{0} & ---         & \texttt{1} \\
\ttenlong{}         & \texttt{0}         & \texttt{0}         & \texttt{0}         & \tthirteenlong{} & ---        & ---         & \texttt{2} \\
\bottomrule
\end{tabular}
\caption{\textbf{Constant predictions without images (Non-Thinking).} Each cell shows the modal prediction; ``---'' means no collapse observed for that model. Qwen3.5\,=\,Qwen3.5-27B, Gemma4\,=\,Gemma4-31B, GLM\,=\,GLM-4.6V-Flash. \tthree{} predictions abbreviated as \texttt{\{i,a,b\}} (intersection, only-a, only-b). The color abbreviations denote fixed palettes emitted regardless of input: ``6-color dict'' $=$ \texttt{\{\textcolor{bedBlue}{blue}, \textcolor{bedCyan}{cyan}, \textcolor{bedGreen}{green}, \textcolor{bedMagenta}{magenta}, \textcolor{bedRed}{red}, \textcolor{bedYellow}{yellow}\}}, each color mapped to count~1; ``12-color dict'' $=$ the full 12-color palette, each count~1; ``12-color list'' $=$ the same palette as a list of names without counts. All modal shares are 100\% except Qwen3.5\,\teight{} (99.0\%), Gemma4\,\teight{} (98.0\%), GLM\,\tfive{} (89.8\%). \textbf{Takeaway:} Without images, all three models give one fixed answer on every \tnine{} and \tten{} instance (\texttt{3} and \texttt{0}), which accounts for their no-image scores of 43\% and 51\% (Table~\ref{tab:no_image_full_nt}, Appendix~\ref{sec:app_no_image_full}).}
\label{tab:prediction_collapse}
\end{table*}

\paragraph{\textcolor{yellow}{\faLightbulb} Removing the image leaves some scores nearly intact.}
In Table~\ref{tab:no_image_full_nt}, \ttwolong{}, \tfourlong{}, \tsixlong{}, \tsevenlong{}, \tninelong{}, and \televenlong{} drop by \textcolor{red!70!black}{$13$--$43$\,pp} on average without the image. \teightlong{} is unchanged for all three models, \tthreelong{} and \tfivelong{} stay at or near floor, and \tthirteenlong{} changes by \textcolor{custom_darkgreen}{$+1.9$}, \textcolor{red!70!black}{$-1.6$}, and $0.0$\,pp for Qwen3.5-27B, Gemma4-31B, and GLM-4.6V-Flash. Image dependence also varies by model: GLM-4.6V-Flash gains on \tonelong{} OE (\textcolor{custom_darkgreen}{$+$11.8\,pp}) and \ttwelvelong{} (\textcolor{custom_darkgreen}{$+$3.0\,pp}), where the other two models drop.

\paragraph{\textcolor{yellow}{\faLightbulb} Constant answers explain many surviving scores.}
Most Non-Thinking no-image cells collapse to a single prediction for every sample with zero run-to-run variance (Table~\ref{tab:prediction_collapse}), an answer-prior behavior also documented in visual question answering \cite{goyal2017making,agrawal2018dont}. All three models output \texttt{0} on every \ttenlong{} instance and \texttt{3} on every \tninelong{} instance; on \tthreelong{}, \tfivelong{}, and \tsevenlong{}, they repeat a fixed triple, color-count dictionary, or color list. When the constant answer is the majority label, the score equals the best-constant baseline of Table~\ref{tab:full_main_results}: 51\% on \tten{}, 43\% on \tnine{}, and 77\% for Qwen3.5-27B on \tsixlong{}, where Gemma4-31B and GLM-4.6V-Flash instead always predict the minority label and reach 23\%. Such scores measure label frequency rather than task competence \cite{lapuschkin2019unmasking}, and which constant a model falls back on differs across checkpoints.

\begin{table*}[t]
\centering
\small
\setlength{\tabcolsep}{12pt}
\begin{tabular}{lcccl}
\toprule
Task & Non-Thinking & Thinking & $\Delta$ & Effect \\
\midrule
\tonelong{} (OE)   & 11.3 & 15.0 & \textcolor{custom_darkgreen}{$+$3.7}  & small gain, still weak \\
\tonelong{} (MCQ)  & 80.9 & 55.3 & \textcolor{red!70!black}{$-$25.7} & disrupts answer-position prior \\
\tthreelong{}    & 0.0  & 0.0  & 0.0     & no recovery \\
\tfourlong{}   & 7.3  & 8.7  & \textcolor{custom_darkgreen}{$+$1.4}   & minimal gain \\
\tfivelong{}   & 0.0  & 0.0  & 0.0     & no recovery \\
\tsixlong{} & 41.0 & 51.4 & \textcolor{custom_darkgreen}{$+$10.4} & breaks collapse \\
\tsevenlong{}    & 0.7  & 2.5  & \textcolor{custom_darkgreen}{$+$1.8}  & still near floor \\
\teightlong{}      & 57.0 & 61.0 & \textcolor{custom_darkgreen}{$+$4.0}  & changes prior, not mechanism \\
\tninelong{}   & 43.0 & 32.0 & \textcolor{red!70!black}{$-$11.0} & hurts despite more reasoning \\
\ttenlong{}    & 51.0 & 49.7 & \textcolor{red!70!black}{$-$1.3}  & no recovery \\
\ttwolong{}     & 52.4 & 56.7 & \textcolor{custom_darkgreen}{$+$4.3}  & marginal gain \\
\televenlong{}    & 36.1 & 34.5 & \textcolor{red!70!black}{$-$1.6}  & no effect \\
\ttwelvelong{}   & 48.5 & 49.1 & \textcolor{custom_darkgreen}{$+$0.6}  & negligible gain \\
\tthirteenlong{} & 38.7 & 39.3 & \textcolor{custom_darkgreen}{$+$0.6}  & negligible gain \\
\bottomrule
\end{tabular}
\caption{\textbf{Effect of Thinking without images.} Non-Thinking and Thinking accuracy (\%) averaged over the three models evaluated without images, under matched task and ablation conditions; each model/mode value is a 5-run mean. \textbf{Takeaway:} Without images, Thinking changes most tasks by only a few points, breaks the \tsix{} collapse ($+$10.4\,pp), lowers \tnine{} by 11.0\,pp, and lowers the \tone{} MCQ score by 25.7\,pp by disrupting the answer-position prior.}
\label{tab:thinking_no_image}
\end{table*}

\paragraph{\textcolor{yellow}{\faLightbulb} Thinking perturbs the constant answers but does not restore visual grounding.}
Without images, Thinking breaks the \tsixlong{} collapse (\textcolor{custom_darkgreen}{$+$10.4\,pp}), but its accuracy (51.4\%) stays far below the visual Non-Thinking average of the same three models (83.5\%), and it lowers \tninelong{} by \textcolor{red!70!black}{11.0\,pp} by moving off the frequent constant answer (Table~\ref{tab:thinking_no_image}) \cite{sprague2025cot}. On \tonelong{} MCQ, it reduces the no-image score by \textcolor{red!70!black}{25.7\,pp}, weakening the answer-position shortcut \cite{zheng2024large}. It recovers nothing on the structured tasks: \tthreelong{} and \tfivelong{} stay at 0.0\%, and \tsevenlong{} rises only from 0.7\% to 2.5\%.

\subsubsection{VLMs may appear stronger under multiple choice}
\label{sec:app_mcq}

% \paragraph{\textcolor{yellow}{\faLightbulb} Reformatting \tonelong{} as multiple choice inflates accuracy without adding 3D competence.}
% Figure~\ref{fig:mcq_boost} compares open-ended (OE) and five-option multiple-choice (MCQ) accuracy on \tone{} under Non-Thinking mode, where the MCQ random-guessing baseline is 20\%. Changing only the answer format -- the underlying image and 3D question are identical -- induces a large artificial boost across almost all models. The inflation averages \textcolor{custom_darkgreen}{$+$39.1\,pp} and spans $1.1\times$ to $4.2\times$ across the suite. It is most extreme for GLM-4.6V-Flash (19.3\% $\rightarrow$ 81.5\%) and Qwen3.5-2B (23.0\% $\rightarrow$ 82.3\%), and even the largest OW models more than double (e.g., Gemma4-31B, 34.5\% $\rightarrow$ 83.2\%); only the smallest model, Qwen3.5-0.8B, fails to exploit the MCQ structure. Because no visual information is added, the boost measures the option-elimination and answer-position cues the format exposes rather than any gain in volumetric counting: LLMs carry a documented selection bias toward particular option positions \cite{zheng2024large}, and can recover correct choices from the options alone even when the question is withheld \cite{balepur2024artifacts}. This is the same artifact isolated by the no-image probe: under MCQ prompting without images, one model reaches 100\% on \tone{} by exploiting answer position alone (Table~\ref{tab:no_image_full_nt}, Appendix~\ref{sec:app_no_image_full}). MCQ-only spatial benchmarks can therefore overstate VLM 3D capability.

\paragraph{\textcolor{yellow}{\faLightbulb} Reformatting \tonelong{} as a fixed-layout multiple choice inflates accuracy without adding 3D competence.}
In the candidate tasks \televenlong{}, \ttwelvelong{}, and \tthirteenlong{}, the position of the correct candidate is shuffled (positions 0/1/2 in 87/101/112 of the 300 \televen{} and \tthirteen{} questions, and 0/1 in 49/51 of the 100 \ttwelve{} questions), so always choosing one position yields at most 37.3\% and 51.0\%, respectively. The \tone{} MCQ control instead orders the five options by value, with the correct volume always third. With the same images and question, this fixed layout raises accuracy over open-ended answers by \textcolor{custom_darkgreen}{$+$39.0\,pp} on average (range \textcolor{custom_darkgreen}{$+$1.2} to \textcolor{custom_darkgreen}{$+$57.8\,pp}; Figure~\ref{fig:mcq_boost}), most for GLM-4.6V-Flash (\textcolor{custom_darkgreen}{22.2\% $\rightarrow$ 80.0\%}) and Qwen3.5-2B (\textcolor{custom_darkgreen}{22.4\% $\rightarrow$ 75.4\%}); the largest dense OW models also gain more than 40\,pp (Qwen3.5-27B, \textcolor{custom_darkgreen}{39.8\% $\rightarrow$ 91.8\%}; Gemma4-31B, \textcolor{custom_darkgreen}{35.0\% $\rightarrow$ 78.0\%}), and only Qwen3.5-0.8B barely benefits ($+$1.2\,pp). The gain reflects option-position cues rather than counting: LLMs show selection biases toward particular option positions \cite{zheng2024large} and can pick correct answers from the options alone \cite{balepur2024artifacts}, and without images Gemma4-31B reaches 100\% on \tone{} MCQ by always choosing the third option (Table~\ref{tab:no_image_full_nt}).

\clearpage
\section{Implementation Details}
\label{sec:app_implementation}
% Scope float spacing to this appendix. Top-align float-only columns/pages
% and put unused space below the blocks, not above or between them.
\begingroup
\makeatletter
\setlength{\@fptop}{0pt}
\setlength{\@fpsep}{12pt}
\setlength{\@fpbot}{0pt plus 1fil}
\setlength{\@dblfptop}{0pt}
\setlength{\@dblfpsep}{12pt}
\setlength{\@dblfpbot}{0pt plus 1fil}
\makeatother
\raggedbottom
\setlength{\floatsep}{12pt plus 2pt minus 2pt}
\setlength{\textfloatsep}{12pt plus 2pt minus 2pt}
\setlength{\intextsep}{12pt plus 2pt minus 2pt}

This appendix describes how we construct inputs and ground truth, run model inference, and collect the human reference used in Section~\ref{sec:3d_blind}.
The generation pipeline first samples geometry, selects and checks views, and renders the supported visual conditions; task construction then turns these reusable samples into questions.
We give the generation procedures below, followed by inference settings, the model roster, and the human-evaluation protocol.

\subsection{Dataset Generation Details}
\label{sec:app_dataset}

\paragraph{Pseudocode notation.}
We write $O$ for a voxel object and $\Pi(O)=\{\pi_{xy},\pi_{yz},\pi_{zx}\}$ for its canonical projections.
For a query view $q=(a,e)$, $d_{\mathrm{axis}}(q)$ is its angular distance to the nearest canonical direction checked by the sampler.
Projection similarity $\mathrm{sim}(\pi,\pi')$ is the maximum IoU under the filter's symmetry transforms; $|\pi\mathbin{\triangle}\pi'|$ counts differing cells.
The set $\mathcal{V}$ contains supported render variants, and $\kappa_p$ assigns each cell the color of its frontmost voxel on plane $p$.
For multi-object scenes, $\mu(G)$ denotes the distribution of scene mutations.
The algorithms describe the main generation steps; finite search budgets and fallback outcomes are explicit where they affect the interpretation of an accepted sample.

\paragraph{Connected voxel objects.}
Each voxel object is built from 10--15 unit blocks on an integer grid.
Starting from a block at the origin, the generator proposes additions next to existing blocks and accepts them only if they are unoccupied, connected to the structure, and supported by the ground or a block immediately below.
Block coordinates satisfy $|x|,|y|\leq4$ and $0\leq z\leq6$.
This explicit geometry supplies the counts, extents, and spatial relations used by the voxel tasks and visual-feature probes (Sections~\ref{sec:benchmark} and~\ref{sec:scaling}).
Algorithm~\ref{alg:grow_voxel_object} summarizes object construction.

\begin{algorithm}[!htb]
\caption{Grow a Connected Voxel Object}
\label{alg:grow_voxel_object}
\footnotesize
\raggedright
\algrenewcommand{\algorithmicindent}{1em}
\begin{algorithmic}[1]
\Require block-count bounds $n_{\min}, n_{\max}$, XY radius $R$, height cap $H$, attempt budget $B$, horizontal-growth probability $p_{\mathrm{xy}}$
\Ensure connected voxel object $O$, or failure
    \State Sample $n \sim [n_{\min}, n_{\max}]$
    \State Initialize $O \gets \{(0,0,0)\}$; $t \gets 0$
    \While{$|O| < n \wedge t < B$}
        \State Set $t \gets t+1$
        \State Sample anchor $b=(x,y,z) \in O$
        \State Sample $\xi \sim \mathrm{Bernoulli}(p_{\mathrm{xy}})$
        \If{$\xi = 1$}
            \State Sample $(\Delta x,\Delta y) \in \{(\pm1,0),(0,\pm1)\}$
            \State Set $c \gets (x+\Delta x,\ y+\Delta y,\ z)$
        \Else
            \State Set $c \gets (x,\ y,\ z+1)$
        \EndIf
        \If{$|c_x| \le R \wedge |c_y| \le R \wedge 0 \le c_z \le H$}
            \If{$c \notin O \wedge (c_z=0 \;\vee\; (c_x,c_y,c_z-1)\in O)$}
                \If{$\exists\, o \in O$: $\|c-o\|_1=1$}
                    \State $O \gets O \cup \{c\}$
                \EndIf
            \EndIf
        \EndIf
    \EndWhile
\If{$|O| < n$}
    \State report failure and stop \Comment{caller may resample}
\EndIf
\State Sort $O$ by $(z,y,x)$
\State \Return $O$
\end{algorithmic}
\end{algorithm}

\paragraph{Query-view selection.}
Two oblique query views show each voxel object from opposite sides.
The first view $q_1=(a_1,e_1)$ is sampled with azimuth in $0$--$360\degree$ and elevation in $18$--$55\degree$, at least $45\degree$ from every canonical viewing direction, which places its elevation between $45\degree$ and $55\degree$.
The second view is placed at the opposite azimuth, $a_1+180\degree$, with elevation $\min(e_1+12\degree,\,55\degree)$.
Across the 100 evaluated objects, the two views therefore differ by $180\degree$ in azimuth and by $0.3$--$10.0\degree$ in elevation (median $5.5\degree$), and second views lie $35.0$--$55.0\degree$ from the nearest canonical direction.
Footprint checks require a projected span of at least 2.2 units in each image dimension, a bounding-box area of at least 6.0 square units, and an aspect ratio no greater than 4.2, so that the object does not appear too narrow or too small.
All evaluated view pairs satisfy these checks (Algorithm~\ref{alg:select_query_views}).

\begin{algorithm}[!htb]
\caption{Select Two Query Views}
\label{alg:select_query_views}
\footnotesize
\raggedright
\algrenewcommand{\algorithmicindent}{1em}
\begin{algorithmic}[1]
\Require voxel object $O$, azimuth range $\mathcal{A}$, elevation range $\mathcal{E}=[e_{\min},e_{\max}]$, axis-separation threshold $\theta_{\mathrm{axis}}$, elevation offset $\delta_e$, minimum footprint span $s_{\min}$, minimum footprint area $A_{\min}$, maximum aspect ratio $\rho_{\max}$, view-attempt budget $B_q$
\Ensure selected query views $(q_1,q_2)$
\For{$t=1,\ldots,B_q$}
    \State Sample $q_1=(a_1,e_1)$ with $a_1\in\mathcal{A}$, $e_1\in\mathcal{E}$, and $d_{\mathrm{axis}}(q_1)\ge\theta_{\mathrm{axis}}$
    \State Set $q_2 \gets (a_1+180\degree,\ \min\{e_1+\delta_e,\ e_{\max}\})$
    \State Project all cube vertices of $O$ under $q_1$ and $q_2$
    \State For each $q \in \{q_1,q_2\}$, compute $(s_x(q),s_y(q))$, $A(q)=s_x(q)s_y(q)$, and $\rho(q)=\max\{s_x/s_y,\ s_y/s_x\}$
    \If{$\min\{s_x(q),s_y(q)\}\ge s_{\min} \wedge A(q)\ge A_{\min}$ for both $q$}
        \If{$\rho(q)\le \rho_{\max}$ for both $q$}
            \State \Return $(q_1,q_2)$
        \EndIf
    \EndIf
\EndFor
\State \Return the last sampled pair \Comment{not reached for the evaluated objects}
\end{algorithmic}
\end{algorithm}

\paragraph{Sampled ambiguity checks.}
The generator tests whether alternative geometry can produce nearly identical silhouettes in both query views while changing the orthographic projections.
For each candidate, it samples 500 alternative objects and renders their masks using the candidate's view limits.
An alternative constitutes a detected collision if its mask intersection over union (IoU) is at least 0.995 in both views and at least one canonical projection differs.
The candidate is then rejected and resampled (Algorithm~\ref{alg:reject_two_view_collisions}).

\begin{algorithm}[!htb]
\caption{Reject Two-View Collisions}
\label{alg:reject_two_view_collisions}
\footnotesize
\raggedright
\algrenewcommand{\algorithmicindent}{1em}
\begin{algorithmic}[1]
\Require voxel object $O$, query views $(q_1,q_2)$, collision-trial budget $T$, mask-matching threshold $\tau_{\mathrm{col}}$
\Ensure object with no collision detected in $T$ trials, or rejection
\State Render base masks $(M_1,M_2)$
\State Compute $\Pi(O)=\{\pi_{xy}(O),\pi_{yz}(O),\pi_{zx}(O)\}$
\For{$t \gets 1$ to $T$}
    \State Sample an alternative voxel object $O'$
    \State Render $(M'_1,M'_2)$ for $O'$ using the fixed limits of $O$
    \If{$\mathrm{IoU}(M_1,M'_1)\ge\tau_{\mathrm{col}}$ and $\mathrm{IoU}(M_2,M'_2)\ge\tau_{\mathrm{col}}$}
        \State Compute $\Pi(O')$
        \If{$\Pi(O') \neq \Pi(O)$}
            \State reject $O$ and stop
        \EndIf
    \EndIf
\EndFor
\State \Return $O$
\end{algorithmic}
\end{algorithm}

\paragraph{Hidden-block filtering.}
A visibility check tests whether removing a block changes the rendered masks.
The filter rejects an object if any block's removal changes fewer than 8 pixels in each $256\times256$ view and both ablated masks keep an IoU of at least 0.9998 with the original masks (Algorithm~\ref{alg:filter_hidden_blocks}).

\begin{algorithm}[!htb]
\caption{Filter Hidden-Block Objects}
\label{alg:filter_hidden_blocks}
\footnotesize
\raggedright
\algrenewcommand{\algorithmicindent}{1em}
\begin{algorithmic}[1]
\Require voxel object $O$, query views $(q_1,q_2)$, minimum mask-change threshold $\delta_{\min}$, unchanged-mask IoU threshold $\tau_{\mathrm{vis}}$
\Ensure object passing the hidden-block check, or rejection
\State Render the base silhouette masks $M_1$ and $M_2$
\For{each block $b \in O$}
    \State Form the ablated object $O^- \gets O \setminus \{b\}$
    \State Render the ablated masks $M^-_1$ and $M^-_2$
    \State Compute $\delta_i \gets |M_i \oplus M^-_i|$ and $\iota_i \gets \mathrm{IoU}(M_i,M^-_i)$ for $i \in \{1,2\}$
    \If{$\delta_1 < \delta_{\min}$ and $\delta_2 < \delta_{\min}$ and $\iota_1 \ge \tau_{\mathrm{vis}}$ and $\iota_2 \ge \tau_{\mathrm{vis}}$}
            \State reject $O$ and stop
    \EndIf
\EndFor
\State \Return $O$
\end{algorithmic}
\end{algorithm}

\paragraph{Multi-object scenes.}
Each \hyperref[sec:app_dataset_t13]{T13: Mixture-of-Shapes} scene contains 6--8 boxes, cubes, or cylinders.
After the first primitive, each placement proposal attaches to an existing primitive with probability 0.45, either on top or beside it; otherwise, the proposal starts on the floor in $[-3,3]^2$.
Intersecting placements are rejected.
The two query views use azimuth/elevation pairs $(45\degree,30\degree)$ and $(150\degree,40\degree)$.
A depth-aware rasterizer checks visibility at $220\times220$ resolution: each primitive must have at least 220 visible pixels in each query view and 180 in each canonical view, with at least 8\% of its projected area visible in every view.
The configured maximum hidden-area ratios are 0.92 for query views and 0.95 for canonical views.
Algorithm~\ref{alg:sample_multi_object_scene} summarizes scene validation and distractor construction.

\paragraph{Matched renderings and color assignment.}
The visual-cue comparisons cross color (colored or uniform gray) with coordinate axes (present or absent), while preserving the underlying geometry.
Only task-compatible variants are exported: color questions retain color, and object comparison retains axes (Table~\ref{tab:visual_cue_effects}, Appendix~\ref{sec:app_visual_cues}).
An adversarial color-assignment mode pairs neighboring blocks using hand-designed color pairs, motivated by contextual effects in color perception \cite{westland2023universal}.
The pairs are \textcolor{bedRed}{red}/\textcolor{bedMagenta}{magenta}, \textcolor{bedBlue}{blue}/\textcolor{bedPurple}{purple}, \textcolor{bedGreen}{green}/\textcolor{bedGold}{gold}, \textcolor{bedCyan}{cyan}/\textcolor{bedYellow}{yellow}, \textcolor{bedOrange}{orange}/\textcolor{bedOrchid}{orchid}, and \textcolor{bedTeal}{teal}/\textcolor{bedSlategray}{slategray}.
They are chosen to combine distinct hues with relatively similar luminance, with the aim of reducing brightness-based boundary cues (Algorithm~\ref{alg:assign_adversarial_colors}).

\begin{algorithm}[!htb]
\caption{Assign Adversarial Colors}
\label{alg:assign_adversarial_colors}
\footnotesize
\raggedright
\algrenewcommand{\algorithmicindent}{1em}
\begin{algorithmic}[1]
\Require voxel object $O$, color-pair set $\mathcal{C}$
\Ensure colored object $\widetilde{O}$
% \If{mode is \textsc{Disegno-Failure}}
    \State Build the $6$-neighbor graph $G_O=(O,E)$
    \State Shuffle the block order
    \For{each uncolored voxel $b \in O$}
        \State Select an uncolored neighbor $b'$ if one exists
        \State Sample $(c_1,c_2) \sim \mathcal{C}$ \Comment{color-pair set}
        \State Randomly swap $(c_1,c_2)$
        \State Set $\widetilde{O}(b)\gets c_1$ and, if present, $\widetilde{O}(b')\gets c_2$
    \EndFor
% \Else
%     \State Rank voxels by a depth score $d(b)$
%     \State Assign warm colors to $\{b:\mathrm{rank}(b)\le |O|/2\}$
%     \State Assign cool colors to $\{b:\mathrm{rank}(b) > |O|/2\}$
% \EndIf
\State \Return $\widetilde{O}$
\end{algorithmic}
\end{algorithm}

\paragraph{Projection distractors.}
For \hyperref[sec:app_dataset_t11]{T11: Projection}, the generator edits occupied cells in a correct projection using add, remove, or shift operations.
A cross-plane filter rejects proposals whose maximum IoU under symmetry with another positive plane is at least 0.85, or whose symmetric difference from that plane is fewer than 2 cells.
The search runs for up to 600 proposals per plane (Algorithm~\ref{alg:generate_hard_negative_projections}).
Every distractor for the 100 evaluated objects passes this filter, with a maximum cross-plane similarity of 0.833 and at least 2 differing cells, and no evaluated object has a skipped plane.
\hyperref[sec:app_dataset_t12]{T12: Reconstruct} uses a different construction: the anchor object's three positive silhouettes are paired with candidate query views from the anchor and a different object.

\begin{algorithm}[!htb]
\caption{Generate Projection Distractors}
\label{alg:generate_hard_negative_projections}
\footnotesize
\raggedright
\algrenewcommand{\algorithmicindent}{1em}
\begin{algorithmic}[1]
\Require positive projections $\Pi$, edit settings, similarity threshold $\tau_{\mathrm{neg}}$, minimum cell difference $d_{\min}$, attempt budget $B_{\mathrm{neg}}$
\Ensure per-plane distractors or skipped-plane markers, with search-status flags
\For{each plane $p\in\{xy,yz,zx\}$}
    \If{$\pi_p$ meets the enabled thin-bar skip rule}
        \State mark plane $p$ as skipped; continue
    \EndIf
    \State $\mathrm{accepted}\gets\mathrm{false}$
    \For{$t=1,\ldots,B_{\mathrm{neg}}$}
        \State Propose $\pi^-_p$ using configured add, remove, or shift edits
        \State Apply the configured connectivity, hole, and cell-count constraints
        \If{for every $p'\ne p$, $\mathrm{sim}(\pi^-_p,\pi_{p'})<\tau_{\mathrm{neg}}$ and $|\pi^-_p\mathbin{\triangle}\pi_{p'}|\ge d_{\min}$}
            \State $\mathrm{accepted}\gets\mathrm{true}$; break
        \EndIf
    \EndFor
    \If{not $\mathrm{accepted}$}
        \State Retain the last proposal and flag possible cross-plane similarity \Comment{not reached for the evaluated objects}
    \EndIf
\EndFor
\State \Return distractors and status flags
\end{algorithmic}
\end{algorithm}

\paragraph{Scene distractors.}
For each \hyperref[sec:app_dataset_t13]{T13: Mixture-of-Shapes} projection plane, the generator proposes a modified scene by swapping two primitives, deleting one, resizing one, lifting one, or translating one horizontally.
Resizing multiplies the X extent by 1.2--1.6 and the Y and Z extents independently by 0.9--1.2; lifting adds 0.8--1.6 units to height, and translation uses $\pm1.2$ or $\pm1.8$ units along X or Y.
A proposal is accepted only if it is non-intersecting, both scenes pass visibility checks in shared view limits, and their silhouette IoU on the target plane lies strictly between 0.55 and 0.88.
The generator allows 120 attempts per plane and reports failure if none succeeds.

\begin{algorithm*}[!tp]
\caption{Sample and Validate a Multi-Object Scene}
\label{alg:sample_multi_object_scene}
\small
\raggedright
\algrenewcommand{\algorithmicindent}{1em}
\begin{algorithmic}
\Require primitive and placement settings, visibility thresholds, query views $\mathcal{Q}$, render variants $\mathcal{V}$, attempt budgets $B_{\mathrm{scene}},B_{\mathrm{obj}},B_{\mathrm{neg}}$
\Ensure reusable multi-object sample $m$, or failure
\end{algorithmic}
\medskip
\begin{minipage}[t]{0.48\textwidth}
\vspace{0pt}
\textit{Scene construction and visibility}
\begin{algorithmic}[1]
\State $\mathrm{scene\_valid}\gets\mathrm{false}$
\For{$t=1,\ldots,B_{\mathrm{scene}}$}
    \State Sample target size $n\in[n_{\min},n_{\max}]$; initialize $G\gets\varnothing$
    \For{$j=1,\ldots,B_{\mathrm{obj}}$}
        \State Sample a primitive's type, dimensions, and orientation
        \State Propose attachment with probability $p_{\mathrm{att}}$ if $G\ne\varnothing$; otherwise propose floor placement
        \State Add the proposal to $G$ if non-intersecting
        \If{$|G|=n$}
            \State break
        \EndIf
    \EndFor
    \If{$|G|=n$ and all query and canonical views pass visibility checks}
        \State $\mathrm{scene\_valid}\gets\mathrm{true}$; break
    \EndIf
\EndFor
\If{not $\mathrm{scene\_valid}$}
    \State report failure and stop
\EndIf
\State Assign colors consistently across views of $G$
\algstore{bedsample_multi_object_scene}
\end{algorithmic}
\end{minipage}\hfill
\begin{minipage}[t]{0.48\textwidth}
\vspace{0pt}
\textit{Projection distractors and export}
\begin{algorithmic}[1]
\algrestore{bedsample_multi_object_scene}
\For{each plane $p\in\{xy,yz,zx\}$}
    \State $\mathrm{accepted}\gets\mathrm{false}$
    \For{$j=1,\ldots,B_{\mathrm{neg}}$}
        \State Sample $G^-_p\sim\mu(G)$ \Comment{swap, delete, move, resize, or lift}
        \If{$G^-_p$ is intersecting}
            \State continue
        \EndIf
        \State Validate both scenes on plane $p$ using shared view limits
        \State Compute silhouette overlap $\iota_p$
        \If{both views pass and $\tau^-_{\min}<\iota_p<\tau^-_{\max}$}
            \State $\mathrm{accepted}\gets\mathrm{true}$; break
        \EndIf
    \EndFor
    \If{not $\mathrm{accepted}$}
        \State report failure and stop
    \EndIf
\EndFor
\State Render query views and positive/negative silhouettes under $\mathcal{V}$
\State Save geometry, renders, and validation reports as $m$
\State \Return $m$
\end{algorithmic}
\end{minipage}
\end{algorithm*}

\paragraph{Reusable samples and task construction.}
Each accepted voxel sample stores its geometry, query images, canonical projections, color assignments, and task metadata (Algorithm~\ref{alg:construct_single_object_sample}).
Single-object tasks derive their targets from this shared sample; \hyperref[sec:app_dataset_t3]{T3: Difference} compares two objects, \hyperref[sec:app_dataset_t11]{T11: Projection} selects among projection candidates, and \hyperref[sec:app_dataset_t12]{T12: Reconstruct} matches silhouettes to an object view.
\hyperref[sec:app_dataset_t13]{T13: Mixture-of-Shapes} applies projection selection to the separate scene pool.
Section~\ref{sec:experiments} specifies the evaluation subset, and Appendix~\ref{sec:app_dataset_examples} illustrates the resulting questions.

\begin{algorithm}[!htb]
\caption{Construct One Single-Object Sample}
\label{alg:construct_single_object_sample}
\footnotesize
\raggedright
\algrenewcommand{\algorithmicindent}{1em}
\begin{algorithmic}[1]
\Require voxel object $O$ passing the enabled filters, query views $(q_1,q_2)$, supported visual-variant set $\mathcal{V}$
\Ensure reusable single-object sample $s$
\State Assign standard or adversarial colors to $O$
\State Compute block count, object extents, container dimensions and fit label, and per-color counts
\State Compute positive projections $\Pi(O)$
\State Generate negative projections $\Pi^-(O)$
\State Derive plane-wise color maps $\kappa_p$ \Comment{frontmost voxel wins}
\State Render query views under $\mathcal{V}$
\State Render $\Pi(O)$ and $\Pi^-(O)$ in the matching variants from $\mathcal{V}$
\State Save geometry, renders, projection edits, and metadata as sample $s$
\State \Return $s$
\end{algorithmic}
\end{algorithm}

\paragraph{Final ambiguity audit.}
This post-hoc audit, implemented independently of the generator, checks whether any evaluated object has an answer that cannot be recovered from the query views, and measures how much such objects affect the reported results.
The hidden-block filter (Algorithm~\ref{alg:filter_hidden_blocks}) operates on silhouette masks, so a block passes it whenever its removal changes the outline, even if none of its faces are visible.
The audit was run after all experiments were completed.
It tests every single-block edit of the 100 evaluated objects: recoloring any block, removing any block, and adding a supported block to any empty face-adjacent cell.
Each edited object is rendered with the benchmark renderer at the original query cameras and compared with the original object in both views.
One object (ID~32) contains a block that is almost entirely occluded in both views: recoloring it leaves both images unchanged, and removing it only thins one outline by about one pixel in one view.
In the filter's silhouette masks, this removal changes 0 pixels in one view and 29 of 18{,}706 foreground pixels in the other (IoU 0.9985 $<$ 0.9998), so the filter treats the block as visible.
No other edit of any object leaves both views unchanged apart from such outline changes.
For this object, the per-color counts of \tfivelong{} cannot be recovered from the images, and the answers to \tfourlong{}, \tsevenlong{}, and \ttenlong{} change for 8, 7, and 2 of the 11 alternative colors of the hidden block, respectively.
Its volume (\tonelong{}) and the projections used in \televenlong{} are visible only through the outline change.
The other tasks are unaffected: no edit changes the \ttwolong{}, \tsixlong{}, \teightlong{}, or \tninelong{} answer, no evaluated \tthreelong{} pair contains this object, and \ttwelvelong{} does not use these query images.

We keep the object in all reported results.
Excluding it from every task built on the 100 shared objects changes each of the 768 open-weight accuracy cells (12 models, both inference modes, all tasks and visual-condition variants) by at most 0.97\,pp.
It changes no verdict or win/loss count in Table~\ref{tab:thinking_delta_full} (Appendix~\ref{sec:app_thinking}), no sign of a color or axes effect in Table~\ref{tab:visual_cue_effects} (Appendix~\ref{sec:app_visual_cues}), and no task's position in the ordering of tasks by mean accuracy.
Each frontier model answered every instance once, so its accuracy on any affected task changes by at most 1.0\,pp.
In the human study, no participant received this object on \tone{}, \tfour{}, \tfive{}, \tseven{}, or \televen{}; four of the 30 participants received it on \tten{}, and excluding their responses changes human \tten{} accuracy from 86.7\% to 88.5\%.

\subsection{General Experiment Setup}
\label{sec:app_exp_details}

\paragraph{Open-weight inference.}
We serve the 12 OW models through an OpenAI-compatible vLLM (v0.19.1) endpoint on NVIDIA H100, NVIDIA A40, and NVIDIA GeForce RTX 3090 GPUs, typically using one to four GPUs depending on the model and task.
The evaluation requests greedy decoding with \texttt{temperature}$=0$ and \texttt{top\_k}$=1$.
Thinking mode uses the model-specific configuration and a default requested reasoning budget of 1024 tokens, where supported; the budget sweep is described in Appendix~\ref{sec:app_thinking_scaling}.
Each OW condition in the main evaluation is run five times, and we report the mean and standard deviation across these runs.
Although decoding is greedy, the runs are not bit-identical: each run sends up to 100 concurrent requests, so vLLM's continuous batching (and, for some models, multi-GPU execution) groups requests differently across runs, and the resulting floating-point differences can change the selected token when two candidates are nearly tied.
No sampling seed is involved.
Under the default setting, an average of 90.8\% of Non-Thinking and 53.7\% of Thinking outputs per model and task are byte-identical across all five runs; longer reasoning traces leave more opportunities for such divergence.
The reported standard deviation therefore measures inference nondeterminism, not uncertainty over the benchmark items.
For the latter, we report paired item-level bootstrap intervals that resample objects (scenes for \tthirteen{}, object pairs for \tthree{}) jointly across models, modes, and runs, with models and runs held fixed: for visual cues (Table~\ref{tab:visual_cue_effects}, Appendix~\ref{sec:app_visual_cues}), thinking (Table~\ref{tab:thinking_delta_full}, Appendix~\ref{sec:app_thinking}), the reasoning budget (Appendix~\ref{sec:app_thinking_scaling}), and model scale (Appendix~\ref{sec:app_scale_full}).
Prompts and answer schemas appear in Appendix~\ref{sec:app_prompts}, and Section~\ref{sec:experiments} gives the evaluation sizes and coverage of the diagnostic subsets.

\paragraph{Frontier inference.}
The four frontier models use the same task prompts and default visual conditions, with one Thinking-mode evaluation per model.
GPT-5 and Grok-4.20 are queried through OpenRouter; Gemini-3-Flash and Gemini-3.1-Pro are queried through Vertex AI.
The API requests set \texttt{temperature}$=0$; the OpenAI-model requests omit \texttt{top\_k}.
Output and reasoning budgets are requested limits whose enforcement depends on the provider.

\paragraph{Answer parsing and scoring.}
Task-specific parsers extract the answer fields used for exact-match accuracy and the secondary metrics defined in Section~\ref{sec:experiments}.
Responses that cannot be parsed are scored as incorrect for accuracy; the secondary metrics are computed over the parsed responses.
For per-color counting, exact match requires the complete color-to-count mapping to be correct, while numerical-error metrics measure how far the counts deviate.
For binary tasks, precision, recall, and F1 are computed separately for each class and then averaged with equal class weights (macro-averaging); undefined class scores are set to zero.

\subsection{Model Details}
\label{sec:app_model_details}

Table~\ref{tab:model_details} identifies the 12 OW checkpoints and four provider-hosted models evaluated in Section~\ref{sec:experiments}.
The parameter column distinguishes total and active parameters for mixture-of-experts (MoE) models and retains provider-reported effective counts where applicable.
Context windows and modalities describe model capabilities; our experiments use the image-and-text inputs specified by each task.
Kimi-VL-A3B releases separate checkpoints for the two modes: Non-Thinking results use Kimi-VL-A3B-Instruct and Thinking results use Kimi-VL-A3B-Thinking-2506; all other OW models use one checkpoint for both modes.

\begin{table*}[t]
\centering
\scriptsize
\setlength{\tabcolsep}{3pt}
\renewcommand{\arraystretch}{1.15}
\begin{adjustbox}{width=\textwidth}
\begin{tabular}{lllllllllll}
\toprule
\makecell[l]{Model} & \makecell[l]{Family /\\Provider} & \makecell[l]{Open /\\Proprietary} & \makecell[l]{Architecture} & \makecell[l]{Parameters} & \makecell[l]{Release} & \makecell[l]{Knowledge\\cutoff} & \makecell[l]{Context\\window} & \makecell[l]{Modalities} & \makecell[l]{License /\\Access} & \makecell[l]{Official\\source} \\
\midrule
Qwen3.5-0.8B & Qwen & Open & \makecell[l]{Dense\\hybrid} & 0.87B & 2026-02-28 & Not public & \makecell[l]{262k native} & Text, image & Apache-2.0 & \makecell[l]{\href{https://hf.co/Qwen/Qwen3.5-0.8B}{HF}} \\
Qwen3.5-2B & Qwen & Open & \makecell[l]{Dense\\hybrid} & 2.27B & 2026-02-28 & Not public & 262k native & Text, image & Apache-2.0 & \makecell[l]{\href{https://hf.co/Qwen/Qwen3.5-2B}{HF}} \\
Qwen3.5-4B & Qwen & Open & \makecell[l]{Dense\\hybrid} & 4.66B & 2026-02-27 & Not public & \makecell[l]{262k native,\\$\sim$1.01M ext.} & Text, image & Apache-2.0 & \makecell[l]{\href{https://hf.co/Qwen/Qwen3.5-4B}{HF}} \\
Qwen3.5-9B & Qwen & Open & \makecell[l]{Dense\\hybrid} & 9.65B & 2026-02-27 & Not public & \makecell[l]{262k native,\\$\sim$1.01M ext.} & Text, image & Apache-2.0 & \makecell[l]{\href{https://hf.co/Qwen/Qwen3.5-9B}{HF}} \\
Qwen3.5-27B & Qwen & Open & \makecell[l]{Dense\\hybrid} & 27.8B & 2026-02-24 & Not public & \makecell[l]{262k native,\\$\sim$1.01M ext.} & Text, image & Apache-2.0 & \makecell[l]{\href{https://hf.co/Qwen/Qwen3.5-27B}{HF}} \\
Qwen3.5-35B-A3B & Qwen & Open & \makecell[l]{MoE\\hybrid} & \makecell[l]{35B total\\(3B act.)} & 2026-02-24 & Not public & \makecell[l]{262k native,\\$\sim$1.01M ext.} & Text, image & Apache-2.0 & \makecell[l]{\href{https://hf.co/Qwen/Qwen3.5-35B-A3B}{HF}} \\
\midrule
Gemma4-E2B & Google & Open & \makecell[l]{Dense\\hybrid attn.} & \makecell[l]{2.3B eff.\\(5.1B total)} & 2026-03-02 & Not public & 128k & \makecell[l]{Text, image,\\audio} & Apache-2.0 & \makecell[l]{\href{https://hf.co/google/gemma-4-E2B-it}{HF}} \\
Gemma4-E4B & Google & Open & \makecell[l]{Dense\\hybrid attn.} & \makecell[l]{4.5B eff.\\(8B total)} & 2026-03-02 & Not public & 128k & \makecell[l]{Text, image,\\audio} & Apache-2.0 & \makecell[l]{\href{https://hf.co/google/gemma-4-E4B-it}{HF}} \\
Gemma4-26B-A4B & Google & Open & \makecell[l]{MoE\\hybrid attn.} & \makecell[l]{25.2B total\\(3.8B act.)} & 2026-03-11 & Not public & 256k & Text, image & Apache-2.0 & \makecell[l]{\href{https://hf.co/google/gemma-4-26B-A4B-it}{HF}} \\
Gemma4-31B & Google & Open & \makecell[l]{Dense\\hybrid attn.} & 30.7B & 2026-03-11 & Not public & 256k & Text, image & Apache-2.0 & \makecell[l]{\href{https://hf.co/google/gemma-4-31B-it}{HF}} \\
\midrule
GLM-4.6V-Flash & Z.ai & Open & \makecell[l]{Dense VLM\\with tool use} & 9B & 2025-12-07 & Not public & 128k & Text, image & MIT & \makecell[l]{\href{https://hf.co/zai-org/GLM-4.6V-Flash}{HF}} \\
Kimi-VL-A3B & Moonshot AI & Open & \makecell[l]{MoE VLM\\(MoonViT)} & \makecell[l]{16B total\\(3B act.)} & 2025-04-09 & Not public & 128k & \makecell[l]{Text, image,\\video} & MIT & \makecell[l]{\href{https://hf.co/moonshotai/Kimi-VL-A3B-Instruct}{HF (NT)} +\\\href{https://hf.co/moonshotai/Kimi-VL-A3B-Thinking-2506}{HF (TH)}} \\
\midrule
GPT-5 & OpenAI & Proprietary & \makecell[l]{Unified\\system} & Not public & 2025-08-07 & 2024-09-30 & 400k & Text, image & Closed API & \makecell[l]{\href{https://developers.openai.com/api/docs/models/gpt-5}{Docs} +\\\href{https://openai.com/index/gpt-5-system-card/}{Card}} \\
Grok-4.20 & xAI & Proprietary & \makecell[l]{Reasoning model\\with tools} & Not public & 2026-03-09 & Not public & 1M & Text, vision & Closed API & \makecell[l]{\href{https://docs.x.ai/developers/models/grok-4.20}{Docs}} \\
Gemini-3-Flash & Google & Proprietary & \makecell[l]{Preview\\multimodal} & Not public & 2025-12-17 & 2025-01 & 1M / 64k & \makecell[l]{Text, image,\\video, audio, PDF} & \makecell[l]{Closed API\\(preview)} & \makecell[l]{\href{https://ai.google.dev/gemini-api/docs/gemini-3}{Docs} +\\\href{https://blog.google/products-and-platforms/products/gemini/gemini-3-flash/}{Blog}} \\
Gemini-3.1-Pro & Google & Proprietary & \makecell[l]{Preview\\multimodal} & Not public & 2026-02-19 & 2025-01 & 1M / 64k & \makecell[l]{Text, image,\\video, audio, PDF} & \makecell[l]{Closed API\\(preview)} & \makecell[l]{\href{https://ai.google.dev/gemini-api/docs/gemini-3}{Docs} +\\\href{https://blog.google/innovation-and-ai/models-and-research/gemini-models/gemini-3-1-pro/}{Blog}} \\
\bottomrule
\end{tabular}
\end{adjustbox}
\caption{Evaluated model roster and provider metadata. Parameter counts distinguish total, active (act.), and effective (eff.) counts where reported; context windows are measured in tokens. See the text for date conventions and undisclosed fields.}
\label{tab:model_details}
\end{table*}

\paragraph{Reading the model metadata.}
``Open'' denotes OW models served locally, and ``Proprietary'' denotes provider-hosted models.
The release column uses checkpoint-repository creation dates for OW models and public announcement dates for frontier models; these are different date conventions.
``Not public'' marks fields not disclosed in the cited sources.

\subsection{Human Baseline Details}
\label{sec:human_baseline_details}

\paragraph{Participants.}
We recruit 30 volunteers aged 16--35 years (mean $23.2$, std. $4.1$), comprising 19 male and 11 female participants.
Their highest completed education is high school ($n=8$), Bachelor's ($n=16$), Master's ($n=4$), or PhD ($n=2$).

\paragraph{Questions and response procedure.}
Each participant answers 13 randomly drawn questions, one per task, under the task's default visual condition.
Participants receive the same task instructions as the VLMs, with the output-format specification removed, and may revisit earlier questions before submitting.
This gives 30 responses per task and 390 responses overall.

\paragraph{Accuracy and uncertainty.}
Each response receives a binary verdict, and per-task accuracy is the fraction of correct responses among the 30 participants.
We report 95\% Wilson confidence intervals for these proportions; the mean accuracy across the 13 tasks is 70.0\%.
Table~\ref{tab:human_gap} gives the comparison with model performance, and Appendix~\ref{sec:app_human_gap} provides further analysis.

\paragraph{Completion time.}
We record total session time because participants can move between questions and revise their answers.
Mean completion time is 1234.1 seconds (approximately 20.6 minutes).
Dividing by 13 gives about 95 seconds per question.

% Flush appendix floats before the following section changes column layout.
\FloatBarrier
\endgroup

\clearpage
\section{Dataset Examples}
\label{sec:app_dataset_examples}
\newcommand{\cdchip}[1]{\textcolor{#1}{\rule{2.4ex}{2.4ex}}}
\newcommand{\cdnum}[3]{{\setlength{\fboxsep}{1pt}\colorbox{#1}{\makebox[2.2ex][c]{\textcolor{#2}{\scriptsize\bfseries\strut #3}}}}}
\newcommand{\threewrap}[1]{\begin{tabular}[t]{@{}c@{\hspace{1pt}}c@{\hspace{1pt}}c@{}}#1\end{tabular}}
\newcommand{\cdY}[1]{\cdnum{bedYellow}{black}{#1}}
\newcommand{\cdC}[1]{\cdnum{bedCyan}{black}{#1}}
\newcommand{\cdG}[1]{\cdnum{bedGreen}{white}{#1}}
\newcommand{\cdB}[1]{\cdnum{bedBlue}{white}{#1}}
\newcommand{\cdP}[1]{\cdnum{bedPurple}{white}{#1}}
\newcommand{\cdO}[1]{\cdnum{bedOrange}{black}{#1}}
\newcommand{\cdOr}[1]{\cdnum{bedOrchid}{black}{#1}}
\newcommand{\cdT}[1]{\cdnum{bedTeal}{white}{#1}}
\newcommand{\cdS}[1]{\cdnum{bedSlategray}{white}{#1}}
\newcommand{\cdGo}[1]{\cdnum{bedGold}{black}{#1}}
\newcommand{\cdM}[1]{\cdnum{bedMagenta}{white}{#1}}
\newcommand{\cdR}[1]{\cdnum{bedRed}{white}{#1}}
\newcommand{\qualimageheight}{1.05in}
\newcommand{\qualsamplegap}{10pt}
\newcommand{\qualinputrow}[1]{\multicolumn{9}{@{}c@{}}{\includegraphics[width=0.92\textwidth,height=\qualimageheight,keepaspectratio]{#1}}\\[-0.5mm]}
\newcommand{\micon}[1]{\makebox[\linewidth][c]{\includegraphics[height=4.5ex]{images/app_qual/#1}}}
\newcommand{\miconlbl}[2]{\makebox[\linewidth][c]{\includegraphics[height=4.5ex]{images/app_qual/#1}\,#2}}
\newcommand{\modelrow}{\makebox[\linewidth][c]{\textbf{GT}} & \micon{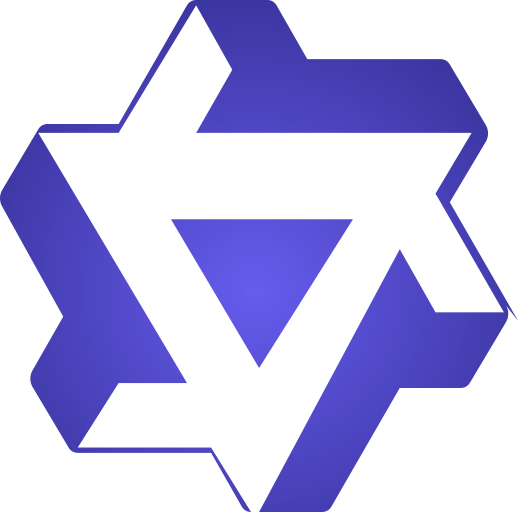} & \micon{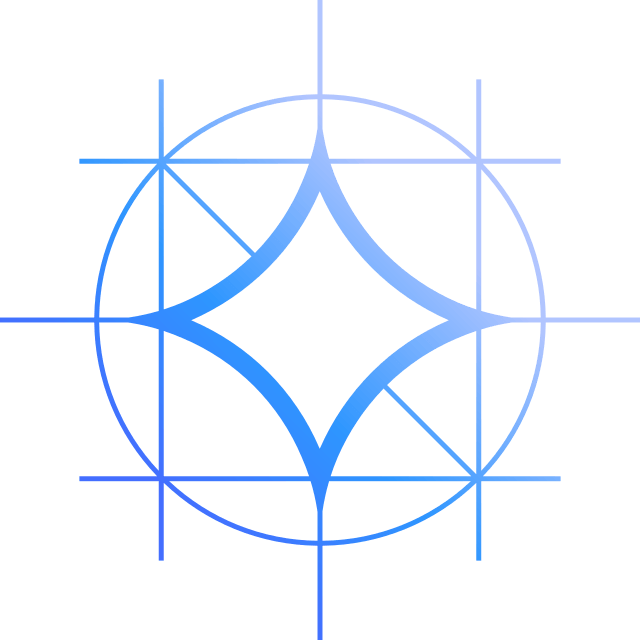} & \micon{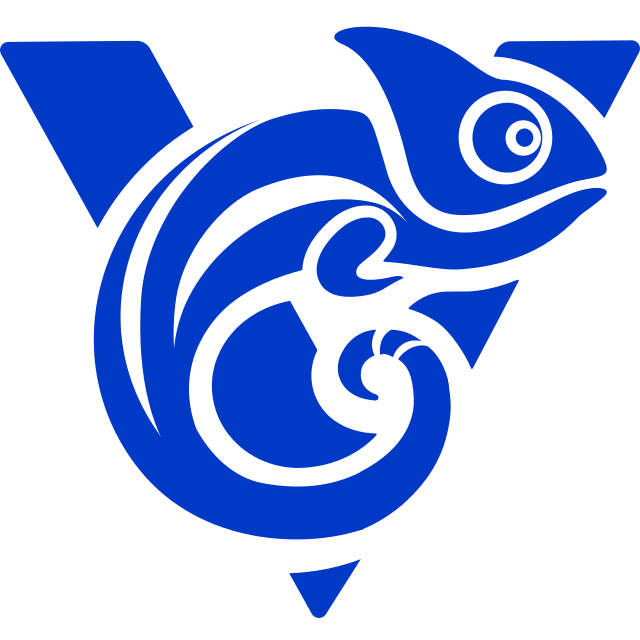} & \micon{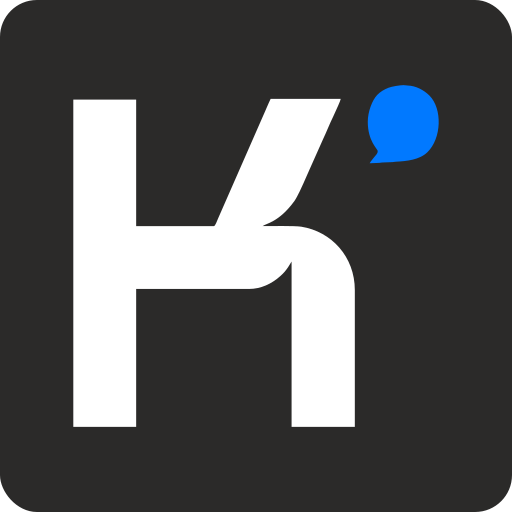} & \micon{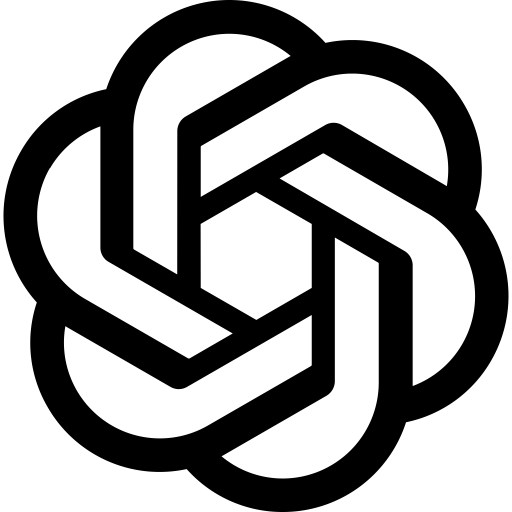} & \micon{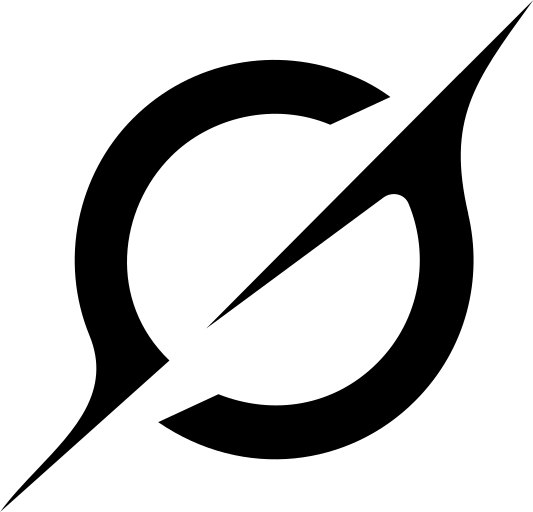} & \micon{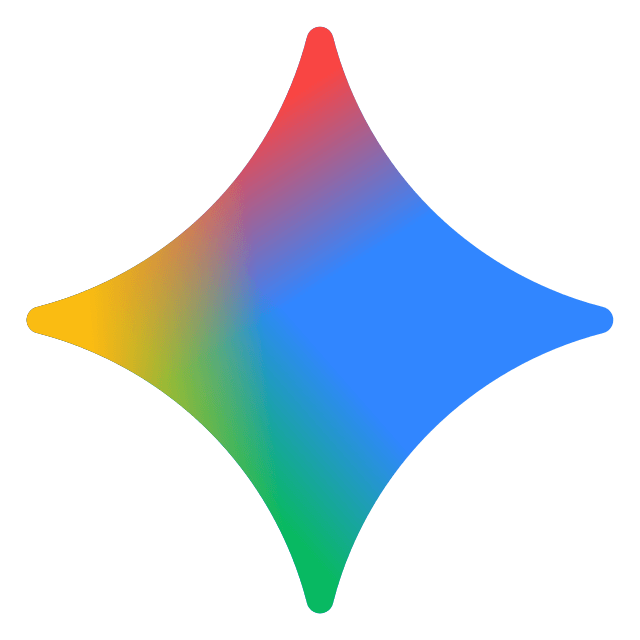} & \micon{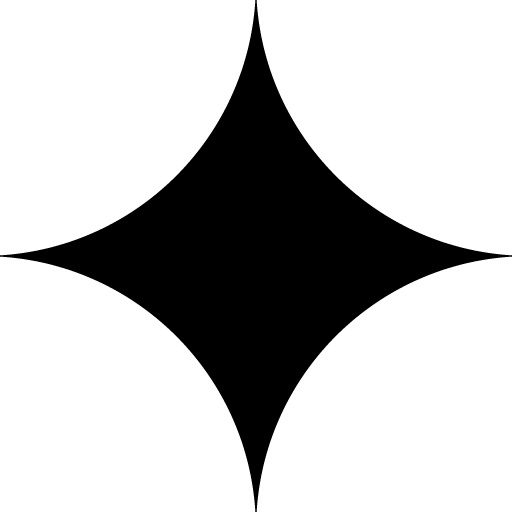} \\}
\newcommand{\promptline}[1]{\multicolumn{9}{@{}l@{}}{\parbox{0.92\textwidth}{\itshape ``\ldots{}Your task is to #1\ldots{}''}}\\[-0.2mm]}

For each of the thirteen tasks, this appendix shows three instances under the default visual setting: the prompt excerpt, the rendered image inputs, the ground-truth answer, and the parsed answer from each of eight primary models (Qwen3.5-27B, Gemma4-31B, GLM-4.6V-Flash, Kimi-VL-A3B, GPT-5, Grok-4.20, Gemini-3-Flash, Gemini-3.1-Pro), marked \tickok/\tickno for correct/incorrect.

\paragraph{Shared conventions.}
For voxel tasks, each block is a unit cube at an integer lattice position, with Z defining the vertical direction. Labels are computed from object coordinates and assigned colors. Each physical block contributes once to an object-level count, even when it appears in both query views. Color categories are the twelve names specified in the prompts, rather than shading variations. The definitions below state each task's target and label convention; the examples show the default condition.

\newcommand{\qualcanonicaltable}[3]{%
\par\noindent
\begin{minipage}{\textwidth}
\centering
\small
\renewcommand{\arraystretch}{1.16}
\setlength{\tabcolsep}{2pt}
\begin{tabular}{*{9}{>{\centering\arraybackslash}p{0.10\textwidth}}}
\toprule
#1
\bottomrule
\end{tabular}
\captionof{table}{#2}
\label{#3}
\end{minipage}
\par\vspace{\qualsamplegap}
}

% Keep each task on one page: heading, balanced two-column prose, then samples.
% The fixed page-top block prevents the samples from floating ahead of the prose.
\twocolumn[{%
\subsection{T1: Volume}
\label{sec:app_dataset_t1}

\setlength{\multicolsep}{3pt}
\begin{multicols}{2}
T1 belongs to the \sethlcolor{task_overview_blue}\hl{Aggregate quantity and support} group (Figure~\ref{fig:task_overview}). Given two views of one object, the target is the total number of occupied voxel positions, which also gives its volume in cubic block units. Each block counts once, regardless of how many faces or views show it. The hidden-block filter is designed to make each voxel visible in at least one view (Appendix~\ref{sec:app_dataset}); combining the views therefore requires avoiding duplicate counts while retaining blocks seen in only one. The default response is an integer; the multiple-choice control presents candidate totals for the same target.
\end{multicols}
\vspace{10pt}

\qualcanonicaltable{
\promptline{count the total number of blocks that make up the 3D object}
\qualinputrow{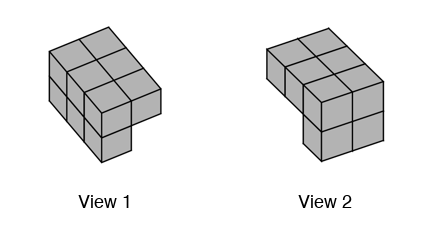}
\modelrow
10 & 10\tickok & 10\tickok & 10\tickok & 10\tickok & 10\tickok & 10\tickok & 10\tickok & 10\tickok \\
}{T1 Volume instance where all eight models count correctly.}{tab:qual_canonical_t1_74}

\qualcanonicaltable{
\promptline{count the total number of blocks that make up the 3D object}
\qualinputrow{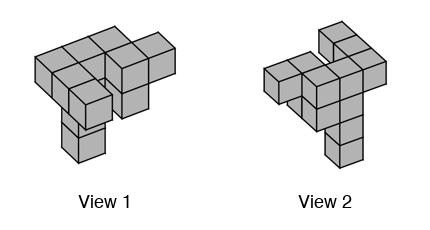}
\modelrow
12 & 12\tickok & 12\tickok & 12\tickok & 10\tickno & 12\tickok & 13\tickno & 16\tickno & 10\tickno \\
}{T1 Volume instance with a mixed split.}{tab:qual_canonical_t1_5}

\qualcanonicaltable{
\promptline{count the total number of blocks that make up the 3D object}
\qualinputrow{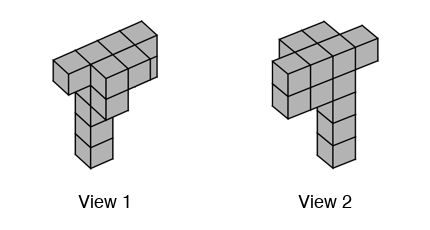}
\modelrow
13 & 12\tickno & 11\tickno & 12\tickno & 10\tickno & 11\tickno & 15\tickno & 10\tickno & 8\tickno \\
}{T1 Volume instance where all eight models miscount.}{tab:qual_canonical_t1_37}
\vspace{-\qualsamplegap}% cancel the gap after the last sample
}]

\twocolumn[{%
\subsection{T2: Packaging}
\label{sec:app_dataset_t2}

\setlength{\multicolsep}{3pt}
\begin{multicols}{2}
T2 tests spatial extents within the \sethlcolor{task_overview_blue}\hl{Aggregate quantity and support} group. Given two object views and three container dimensions, the target is whether the object fits under an axis-aligned reorientation. Each object extent is the maximum occupied coordinate minus the minimum plus one. The label is true if any permutation of the three extents is no larger than the corresponding container dimensions, allowing equality. Mental rotation thus changes which object axis aligns with each container axis; the label procedure does not test arbitrary oblique rotations. Unlike T1, this judgment depends on the object's extents rather than its total block count.
\end{multicols}
\vspace{10pt}

\qualcanonicaltable{
\promptline{determine whether the object can fit in the boundary in any possible orientation. Boundary: \text{[5,4,5]}}
\qualinputrow{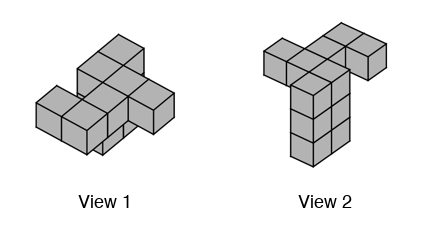}
\modelrow
True & True\tickok & True\tickok & True\tickok & True\tickok & True\tickok & True\tickok & True\tickok & True\tickok \\
}{T2 Packaging instance where all eight models answer correctly.}{tab:qual_canonical_t2_8}

\qualcanonicaltable{
\promptline{determine whether the object can fit in the boundary in any possible orientation. Boundary: \text{[2,3,2]}}
\qualinputrow{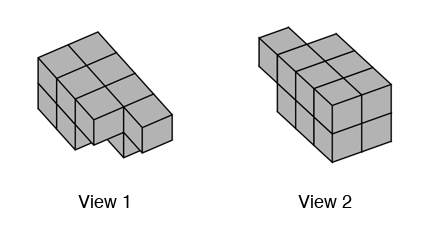}
\modelrow
False & False\tickok & False\tickok & True\tickno & True\tickno & False\tickok & False\tickok & False\tickok & True\tickno \\
}{T2 Packaging instance with a mixed split.}{tab:qual_canonical_t2_16}

\qualcanonicaltable{
\promptline{determine whether the object can fit in the boundary in any possible orientation. Boundary: \text{[3,2,3]}}
\qualinputrow{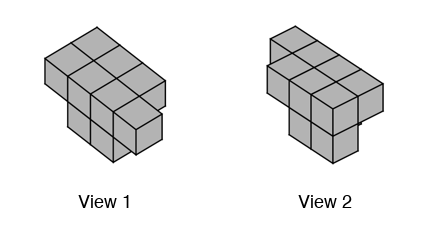}
\modelrow
False & True\tickno & True\tickno & True\tickno & True\tickno & True\tickno & False\tickok & True\tickno & True\tickno \\
}{T2 Packaging instance with a mixed split (1/8 correct).}{tab:qual_canonical_t2_98}
\vspace{-\qualsamplegap}% cancel the gap after the last sample
}]

\twocolumn[{%
\subsection{T3: Difference}
\label{sec:app_dataset_t3}

\setlength{\multicolsep}{3pt}
\begin{multicols}{2}
T3 opens the \sethlcolor{task_overview_orange}\hl{Compositional decomposition} group. Given two views each of objects A and B, it requests the numbers of shared blocks and blocks unique to each object. Correspondence uses identical integer positions in the generator coordinate frame, whose axes appear in the images; neither object is independently translated or rotated to maximize overlap. The three labels are the sizes of the intersection and the two set differences of the occupied-position sets. In the color-aware variant, shared blocks must also match in color. T3 therefore requires a structured partition across objects, rather than a single aggregate count.
\end{multicols}
\vspace{10pt}

\qualcanonicaltable{
\promptline{analyze both objects and determine the difference in their block compositions}
\qualinputrow{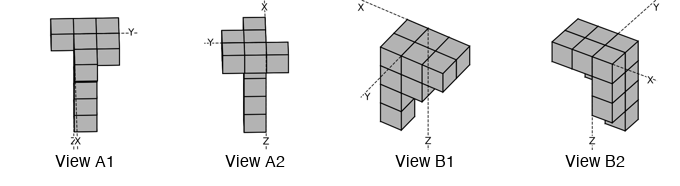}
\modelrow
$(6,4,6)$ & $(10,2,2)$\tickno & $(7,3,3)$\tickno & $(11,0,0)$\tickno & $(2,4,5)$\tickno & $(13,1,1)$\tickno & $(8,4,3)$\tickno & $(4,11,10)$\tickno & $(6,4,6)$\tickok \\
}{T3 Difference instance with a mixed split.}{tab:qual_canonical_t3_41_vs_89}

\qualcanonicaltable{
\promptline{analyze both objects and determine the difference in their block compositions}
\qualinputrow{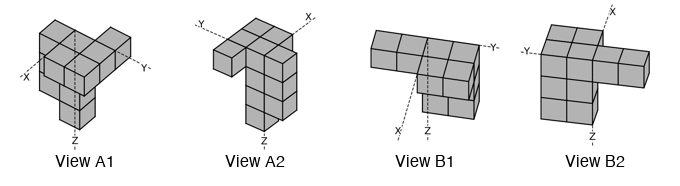}
\modelrow
$(7,4,3)$ & $(10,2,2)$\tickno & $(7,3,3)$\tickno & $(8,4,5)$\tickno & $(2,4,5)$\tickno & $(9,1,2)$\tickno & $(6,4,5)$\tickno & $(7,1,3)$\tickno & $(7,4,2)$\tickno \\
}{T3 Difference instance where all eight models return an incorrect triple with small per-component error.}{tab:qual_canonical_t3_19_vs_51}

\qualcanonicaltable{
\promptline{analyze both objects and determine the difference in their block compositions}
\qualinputrow{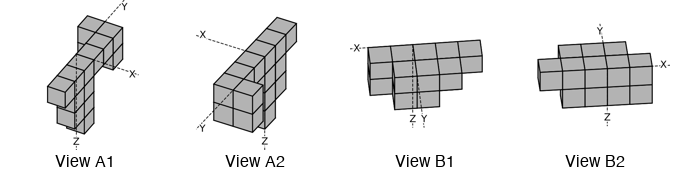}
\modelrow
$(2,12,12)$ & $(10,2,0)$\tickno & $(7,3,3)$\tickno & $(0,13,12)$\tickno & $(2,4,5)$\tickno & $(10,2,1)$\tickno & $(2,4,5)$\tickno & $(4,8,11)$\tickno & $(14,2,4)$\tickno \\
}{T3 Difference instance where all eight models return an incorrect triple with large per-component error.}{tab:qual_canonical_t3_2_vs_10}
\vspace{-\qualsamplegap}% cancel the gap after the last sample
}]

\twocolumn[{%
\subsection{T4: Color Count}
\label{sec:app_dataset_t4}

\setlength{\multicolsep}{3pt}
\begin{multicols}{2}
T4 returns to the \sethlcolor{task_overview_blue}\hl{Aggregate quantity and support} group, replacing T1's occupancy count with a categorical count. Given two colored views, the target is the number of distinct palette categories assigned to at least one block. Repeated occurrences of a color count as one category, including occurrences seen in both views. The label is the cardinality of the set of assigned colors. Unlike T5, T4 requires identifying which colors are present without determining how many blocks belong to each color.
\end{multicols}
\vspace{10pt}

\qualcanonicaltable{
\promptline{count the number of unique colors in the object}
\qualinputrow{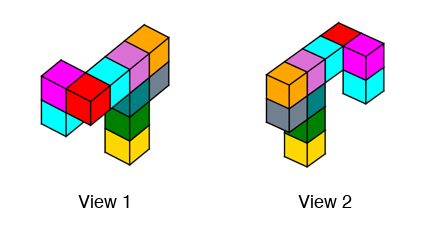}
\modelrow
9 & 9\tickok & 9\tickok & 9\tickok & 9\tickok & 9\tickok & 9\tickok & 9\tickok & 9\tickok \\
}{T4 Color Count instance where all eight models answer correctly.}{tab:qual_canonical_t4_40}

\qualcanonicaltable{
\promptline{count the number of unique colors in the object}
\qualinputrow{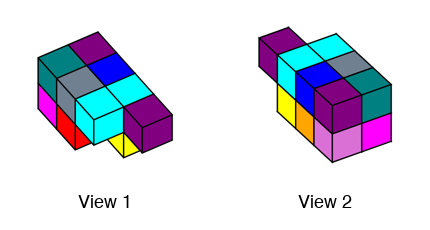}
\modelrow
10 & 11\tickno & 11\tickno & 11\tickno & 10\tickok & 10\tickok & 10\tickok & 10\tickok & 11\tickno \\
}{T4 Color Count instance with a mixed split.}{tab:qual_canonical_t4_16}

\qualcanonicaltable{
\promptline{count the number of unique colors in the object}
\qualinputrow{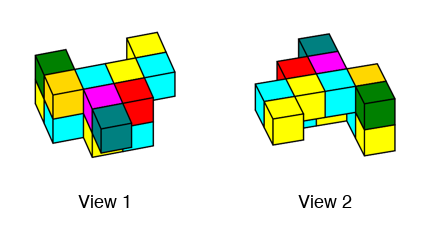}
\modelrow
7 & 6\tickno & 6\tickno & 6\tickno & 9\tickno & 6\tickno & 6\tickno & 6\tickno & 6\tickno \\
}{T4 Color Count instance where all eight models answer incorrectly.}{tab:qual_canonical_t4_4}
\vspace{-\qualsamplegap}% cancel the gap after the last sample
}]

\twocolumn[{%
\subsection{T5: Color Distribution}
\label{sec:app_dataset_t5}
% Leave room for the taller color-distribution answer rows.
\renewcommand{\qualimageheight}{0.95in}

\setlength{\multicolsep}{3pt}
\begin{multicols}{2}
T5 is the compositional-decomposition sibling of T4. Given two colored views, it requests a dictionary mapping every color present in the object to its block count. Each physical block contributes once to its assigned color, requiring both color-to-block association and aggregation within each category. Exact-match accuracy requires the same color keys and counts as the target dictionary; per-color error metrics instead use twelve-dimensional vectors with omitted palette categories assigned zero. T5 is the paper's clearest VLM-specific failure case (Section~\ref{sec:human_gap}), despite the comparatively higher accuracy on T4.
\end{multicols}
\vspace{10pt}

\qualcanonicaltable{
\promptline{give the color distribution in the object}
\qualinputrow{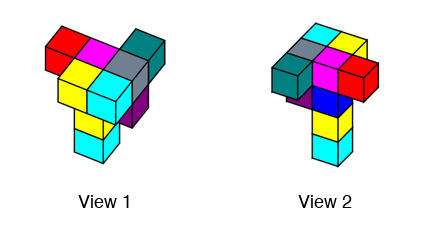}
\modelrow
\threewrap{\cdB{1}&\cdC{2}&\cdM{1}\\\cdP{1}&\cdR{1}&\cdS{1}\\\cdT{1}&\cdY{2}&} & \threewrap{\cdB{1}&\cdC{3}&\cdM{1}\\\cdP{1}&\cdR{1}&\cdS{1}\\\cdT{1}&\cdY{3}&}\tickno & \threewrap{\cdB{1}&\cdC{2}&\cdM{1}\\\cdP{1}&\cdR{1}&\cdS{1}\\\cdT{1}&\cdY{2}&}\tickok & \threewrap{\cdC{3}&\cdM{1}&\cdP{1}\\\cdR{2}&\cdS{1}&\cdT{2}\\\cdY{3}&&}\tickno & \threewrap{\cdB{1}&\cdC{5}&\cdM{3}\\\cdP{1}&\cdR{3}&\cdS{2}\\\cdT{3}&\cdY{5}&}\tickno & \threewrap{\cdB{1}&\cdC{2}&\cdM{1}\\\cdP{1}&\cdR{1}&\cdS{1}\\\cdT{1}&\cdY{2}&}\tickok & \threewrap{\cdB{1}&\cdC{3}&\cdM{1}\\\cdP{1}&\cdR{1}&\cdS{1}\\\cdT{1}&\cdY{2}&}\tickno & \threewrap{\cdB{1}&\cdC{2}&\cdM{1}\\\cdP{1}&\cdR{1}&\cdS{1}\\\cdT{1}&\cdY{2}&}\tickok & \threewrap{\cdB{1}&\cdC{3}&\cdM{1}\\\cdP{1}&\cdR{1}&\cdS{1}\\\cdT{1}&\cdY{3}&}\tickno \\
}{T5 Color Distribution instance with a mixed split.}{tab:qual_canonical_t5_70}

\qualcanonicaltable{
\promptline{give the color distribution in the object}
\qualinputrow{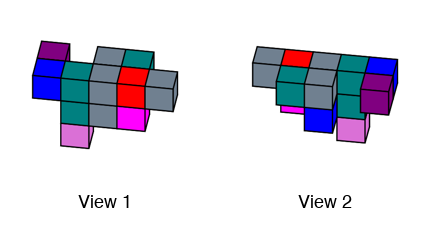}
\modelrow
\threewrap{\cdB{2}&\cdM{1}&\cdOr{1}\\\cdP{1}&\cdR{1}&\cdS{4}\\\cdT{3}&&} & \threewrap{\cdB{2}&\cdM{1}&\cdOr{1}\\\cdP{1}&\cdR{2}&\cdS{5}\\\cdT{4}&&}\tickno & \threewrap{\cdB{2}&\cdM{1}&\cdOr{1}\\\cdP{1}&\cdR{1}&\cdS{5}\\\cdT{4}&&}\tickno & \threewrap{\cdB{3}&\cdM{2}&\cdP{2}\\\cdR{1}&\cdS{5}&\cdT{4}}\tickno & \threewrap{\cdB{5}&\cdC{8}&\cdGo{2}\\\cdM{4}&\cdOr{10}&\cdP{5}\\\cdR{5}&\cdS{8}&\cdT{8}}\tickno & \threewrap{\cdB{1}&\cdM{1}&\cdOr{2}\\\cdP{2}&\cdR{1}&\cdS{5}\\\cdT{3}&&}\tickno & \threewrap{\cdB{2}&\cdM{2}&\cdP{2}\\\cdR{2}&\cdS{6}&\cdT{4}}\tickno & \threewrap{\cdB{2}&\cdM{1}&\cdOr{1}\\\cdP{1}&\cdR{1}&\cdS{4}\\\cdT{3}&&}\tickok & \threewrap{\cdB{2}&\cdM{1}&\cdOr{1}\\\cdP{1}&\cdR{1}&\cdS{3}\\\cdT{3}&&}\tickno \\
}{T5 Color Distribution instance with a mixed split.}{tab:qual_canonical_t5_38}

\qualcanonicaltable{
\promptline{give the color distribution in the object}
\qualinputrow{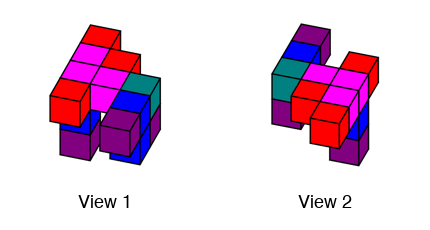}
\modelrow
\threewrap{\cdB{3}&\cdM{3}&\cdP{3}\\\cdR{3}&\cdT{1}&} & \threewrap{\cdB{4}&\cdM{4}&\cdP{4}\\\cdR{5}&\cdT{2}&}\tickno & \threewrap{\cdB{3}&\cdM{4}&\cdP{4}\\\cdR{4}&\cdT{2}&}\tickno & \threewrap{\cdB{3}&\cdM{4}&\cdP{4}\\\cdR{6}&\cdT{2}&}\tickno & \threewrap{\cdB{5}&\cdGo{1}&\cdM{6}\\\cdOr{3}&\cdP{5}&\cdR{9}\\\cdT{3}&&}\tickno & \threewrap{\cdB{4}&\cdM{4}&\cdP{5}\\\cdR{4}&\cdT{2}&}\tickno & \threewrap{\cdB{4}&\cdM{4}&\cdP{4}\\\cdR{4}&\cdT{1}&}\tickno & \threewrap{\text{--}&&}\tickno & \threewrap{\cdB{4}&\cdM{4}&\cdP{4}\\\cdR{5}&\cdT{2}&}\tickno \\
}{T5 Color Distribution instance where all eight models answer incorrectly.}{tab:qual_canonical_t5_94}
\vspace{-\qualsamplegap}% cancel the gap after the last sample
}]

\twocolumn[{%
\subsection{T6: Local Structure}
\label{sec:app_dataset_t6}

\setlength{\multicolsep}{3pt}
\begin{multicols}{2}
T6 completes the \sethlcolor{task_overview_orange}\hl{Compositional decomposition} group with a local binary judgment. Given two object views and a threshold $k$, the target is whether any vertical column contains more than $k$ blocks. A column contains blocks sharing the same X and Y coordinates, and its height is measured by its block count. The evaluated threshold is $k=3$, so exactly three blocks do not satisfy the condition. The label thresholds the largest column count. This tests identifying a qualifying local structure without requiring the full decomposition requested by T3 or T5.
\end{multicols}
\vspace{10pt}

\qualcanonicaltable{
\promptline{determine whether the object has at least one column (vertical stack of blocks) with a height greater than 3}
\qualinputrow{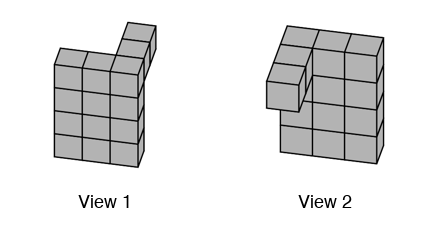}
\modelrow
True & True\tickok & True\tickok & True\tickok & True\tickok & True\tickok & True\tickok & True\tickok & True\tickok \\
}{T6 Local Structure instance where all eight models answer correctly.}{tab:qual_canonical_t6_34}

\qualcanonicaltable{
\promptline{determine whether the object has at least one column (vertical stack of blocks) with a height greater than 3}
\qualinputrow{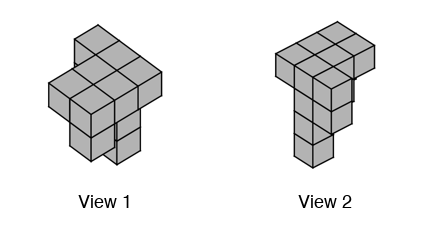}
\modelrow
True & True\tickok & False\tickno & False\tickno & True\tickok & False\tickno & True\tickok & True\tickok & True\tickok \\
}{T6 Local Structure instance with a mixed split.}{tab:qual_canonical_t6_12}

\qualcanonicaltable{
\promptline{determine whether the object has at least one column (vertical stack of blocks) with a height greater than 3}
\qualinputrow{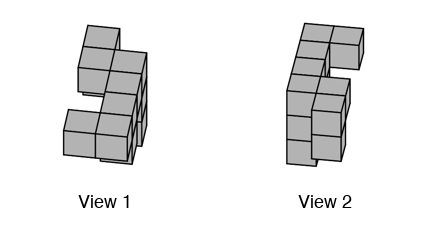}
\modelrow
False & True\tickno & False\tickok & True\tickno & True\tickno & True\tickno & True\tickno & True\tickno & True\tickno \\
}{T6 Local Structure instance with a mixed split.}{tab:qual_canonical_t6_93}
\vspace{-\qualsamplegap}% cancel the gap after the last sample
}]

\twocolumn[{%
\subsection{T7: Color Dominance}
\label{sec:app_dataset_t7}

\setlength{\multicolsep}{3pt}
\begin{multicols}{2}
T7 sits in the \sethlcolor{task_overview_blue}\hl{Aggregate quantity and support} group alongside T4, but asks for a comparison across color counts. Given two colored views, the target is the set of colors with the largest number of blocks, including all colors tied for the maximum. Dominance refers to block count, rather than projected image area. Accuracy requires the exact set, with no missing or additional colors; output order does not matter. The answer follows from the full T5 distribution, making T7 useful for checking whether separately elicited answers agree.
\end{multicols}
\vspace{10pt}

\qualcanonicaltable{
\promptline{determine the dominant color(s) in the object --- the color(s) with the highest block count}
\qualinputrow{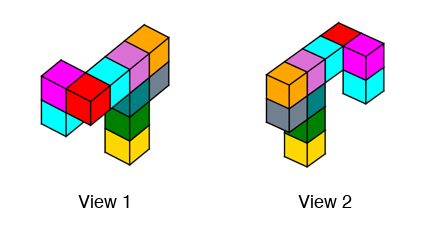}
\modelrow
\threewrap{\cdchip{bedCyan}&&} & \threewrap{\cdchip{bedCyan}&\cdchip{bedOrchid}&\cdchip{bedTeal}\\\cdchip{bedGreen}&\cdchip{bedYellow}&\cdchip{bedMagenta}\\\cdchip{bedRed}&\cdchip{bedOrange}&\cdchip{bedSlategray}}\tickno & \threewrap{\cdchip{bedCyan}&&}\tickok & \threewrap{\cdchip{bedCyan}&&}\tickok & \threewrap{\cdchip{bedCyan}&\cdchip{bedMagenta}&\cdchip{bedYellow}\\\cdchip{bedGreen}&\cdchip{bedOrange}&\cdchip{bedOrchid}\\\cdchip{bedTeal}&\cdchip{bedSlategray}&}\tickno & \threewrap{\cdchip{bedCyan}&&}\tickok & \threewrap{\cdchip{bedCyan}&&}\tickok & \threewrap{\cdchip{bedCyan}&&}\tickok & \threewrap{\cdchip{bedCyan}&&}\tickok \\
}{T7 Color Dominance instance with a mixed split.}{tab:qual_canonical_t7_40}

\qualcanonicaltable{
\promptline{determine the dominant color(s) in the object --- the color(s) with the highest block count}
\qualinputrow{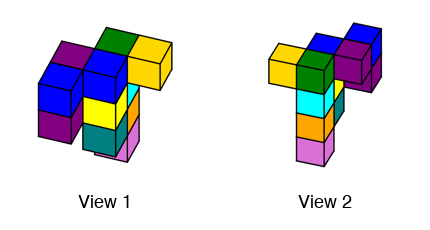}
\modelrow
\threewrap{\cdchip{bedBlue}&\cdchip{bedPurple}&} & \threewrap{\cdchip{bedBlue}&\cdchip{bedPurple}&}\tickok & \threewrap{\cdchip{bedBlue}&\cdchip{bedPurple}&}\tickok & \threewrap{\cdchip{bedBlue}&\cdchip{bedPurple}&}\tickok & \threewrap{\cdchip{bedBlue}&\cdchip{bedPurple}&\cdchip{bedYellow}}\tickno & \threewrap{\cdchip{bedBlue}&&}\tickno & \threewrap{\cdchip{bedBlue}&&}\tickno & \threewrap{\cdchip{bedYellow}&\cdchip{bedBlue}&\cdchip{bedPurple}}\tickno & \threewrap{\cdchip{bedBlue}&\cdchip{bedPurple}&\cdchip{bedYellow}}\tickno \\
}{T7 Color Dominance instance with a mixed split.}{tab:qual_canonical_t7_35}

\qualcanonicaltable{
\promptline{determine the dominant color(s) in the object --- the color(s) with the highest block count}
\qualinputrow{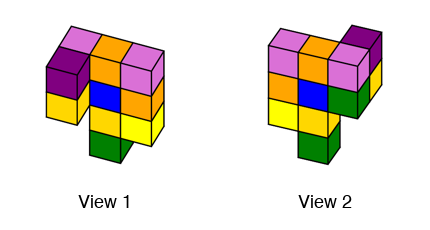}
\modelrow
\threewrap{\cdchip{bedOrchid}&\cdchip{bedOrange}&\cdchip{bedGold}\\\cdchip{bedGreen}&&} & \threewrap{\cdchip{bedYellow}&\cdchip{bedOrchid}&\cdchip{bedOrange}\\\cdchip{bedGreen}&\cdchip{bedBlue}&\cdchip{bedPurple}}\tickno & \threewrap{\cdchip{bedYellow}&\cdchip{bedOrange}&}\tickno & \threewrap{\cdchip{bedYellow}&\cdchip{bedOrange}&\cdchip{bedPurple}\\\cdchip{bedOrchid}&&}\tickno & \threewrap{\cdchip{bedYellow}&\cdchip{bedGreen}&\cdchip{bedBlue}\\\cdchip{bedOrange}&\cdchip{bedPurple}&\cdchip{bedMagenta}}\tickno & \threewrap{\cdchip{bedYellow}&&}\tickno & \threewrap{\cdchip{bedYellow}&&}\tickno & \threewrap{\cdchip{bedYellow}&&}\tickno & \threewrap{\cdchip{bedGold}&\cdchip{bedOrange}&}\tickno \\
}{T7 Color Dominance instance where all eight models answer incorrectly.}{tab:qual_canonical_t7_99}
\vspace{-\qualsamplegap}% cancel the gap after the last sample
}]

\twocolumn[{%
\subsection{T8: Silhouette Symmetry}
\label{sec:app_dataset_t8}

\setlength{\multicolsep}{3pt}
\begin{multicols}{2}
T8 opens the \sethlcolor{task_overview_yellow}\hl{Relational structure} group. The two input views provide evidence for inferring the object's shape and assessing its silhouette symmetry. The label is true if at least one canonical orthographic projection onto XY, YZ, or ZX is horizontally or vertically mirror-symmetric. Each projection is a set of occupied two-dimensional cells obtained by dropping the remaining coordinate; reflection about either corresponding bounding-box midline must leave that set unchanged. This tests a global geometric relation in the object's projections, independently of its assigned colors.
\end{multicols}
\vspace{10pt}

\qualcanonicaltable{
\promptline{determine whether the object's silhouette is symmetric (vertically or horizontally)}
\qualinputrow{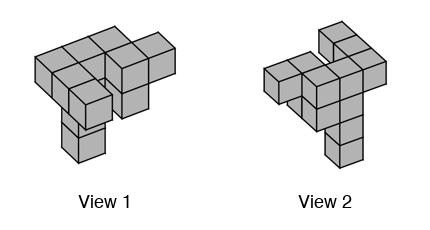}
\modelrow
False & False\tickok & False\tickok & False\tickok & False\tickok & False\tickok & False\tickok & False\tickok & False\tickok \\
}{T8 Silhouette Symmetry instance where all eight models answer correctly.}{tab:qual_canonical_t8_5}

\qualcanonicaltable{
\promptline{determine whether the object's silhouette is symmetric (vertically or horizontally)}
\qualinputrow{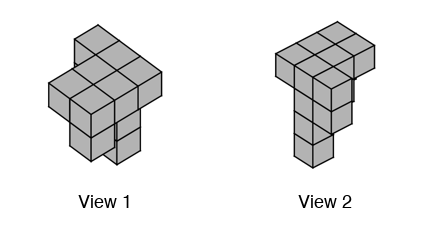}
\modelrow
True & False\tickno & False\tickno & False\tickno & True\tickok & False\tickno & False\tickno & True\tickok & True\tickok \\
}{T8 Silhouette Symmetry instance with a mixed split.}{tab:qual_canonical_t8_12}

\qualcanonicaltable{
\promptline{determine whether the object's silhouette is symmetric (vertically or horizontally)}
\qualinputrow{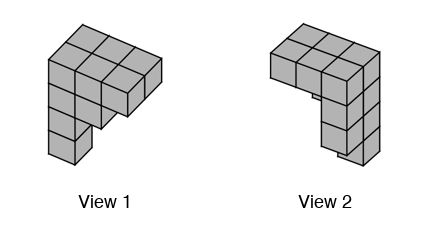}
\modelrow
True & False\tickno & False\tickno & False\tickno & False\tickno & False\tickno & False\tickno & False\tickno & False\tickno \\
}{T8 Silhouette Symmetry instance where all eight models answer incorrectly.}{tab:qual_canonical_t8_89}
\vspace{-\qualsamplegap}% cancel the gap after the last sample
}]

\twocolumn[{%
\subsection{T9: Visible Floors}
\label{sec:app_dataset_t9}

\setlength{\multicolsep}{3pt}
\begin{multicols}{2}
T9 returns to the \sethlcolor{task_overview_blue}\hl{Aggregate quantity and support} group, aggregating occupancy along the height axis. Given two object views, the target is the number of distinct occupied Z coordinates. A floor is a horizontal lattice level containing at least one block, regardless of its block count or how many views show it. Labels count these occupied levels from the object coordinates without a separate pixel-visibility threshold. T9 therefore tests aggregation by height, complementing the total-block and color-based judgments in this group.
\end{multicols}
\vspace{10pt}

\qualcanonicaltable{
\promptline{count the number of visible layers (floors/horizontal levels) in the object}
\qualinputrow{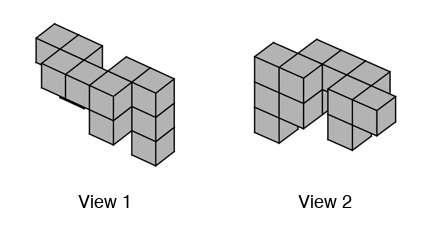}
\modelrow
3 & 3\tickok & 3\tickok & 3\tickok & 4\tickno & 3\tickok & 3\tickok & 3\tickok & 3\tickok \\
}{T9 Visible Floors instance with a mixed split.}{tab:qual_canonical_t9_3}

\qualcanonicaltable{
\promptline{count the number of visible layers (floors/horizontal levels) in the object}
\qualinputrow{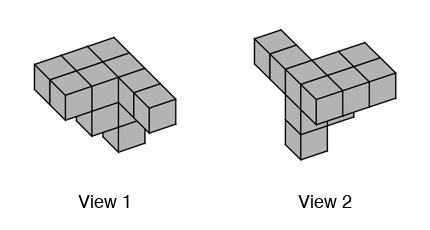}
\modelrow
3 & 3\tickok & 2\tickno & 3\tickok & 4\tickno & 3\tickok & 4\tickno & 3\tickok & 3\tickok \\
}{T9 Visible Floors instance with a mixed split.}{tab:qual_canonical_t9_71}

\qualcanonicaltable{
\promptline{count the number of visible layers (floors/horizontal levels) in the object}
\qualinputrow{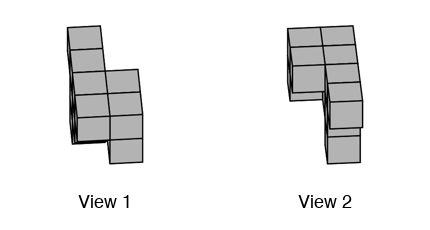}
\modelrow
3 & 5\tickno & 4\tickno & 4\tickno & 4\tickno & 5\tickno & 4\tickno & 6\tickno & 5\tickno \\
}{T9 Visible Floors instance where all eight models answer incorrectly.}{tab:qual_canonical_t9_59}
\vspace{-\qualsamplegap}% cancel the gap after the last sample
}]

\twocolumn[{%
\subsection{T10: Color Connected}
\label{sec:app_dataset_t10}

\setlength{\multicolsep}{3pt}
\begin{multicols}{2}
T10 is the second \sethlcolor{task_overview_yellow}\hl{Relational structure} task. Given two colored views, the target is whether any pair of face-adjacent blocks shares the same assigned color. Face adjacency requires integer coordinates to differ by one along exactly one axis and agree along the other two; edge or corner contact does not qualify. A single matching pair makes the label true, without requiring all blocks of that color to form a connected component. Together, T8 and T10 test relations at different scales: global reflective symmetry and local adjacency.
\end{multicols}
\vspace{10pt}

\qualcanonicaltable{
\promptline{determine whether there exists at least one pair of adjacent blocks (sharing a face) that share the same color}
\qualinputrow{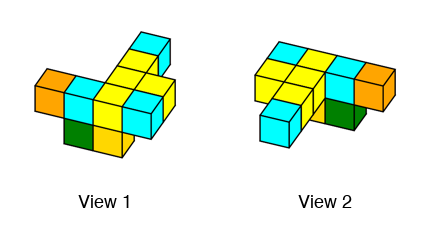}
\modelrow
True & True\tickok & True\tickok & True\tickok & True\tickok & True\tickok & True\tickok & True\tickok & True\tickok \\
}{T10 Color Connected instance where all eight models answer correctly.}{tab:qual_canonical_t10_7}

\qualcanonicaltable{
\promptline{determine whether there exists at least one pair of adjacent blocks (sharing a face) that share the same color}
\qualinputrow{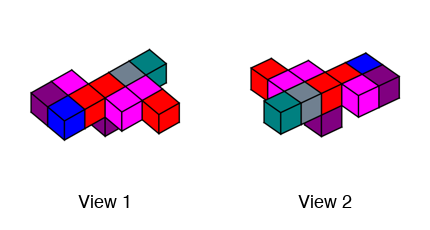}
\modelrow
True & True\tickok & False\tickno & False\tickno & True\tickok & False\tickno & True\tickok & True\tickok & True\tickok \\
}{T10 Color Connected instance with a mixed split.}{tab:qual_canonical_t10_58}

\qualcanonicaltable{
\promptline{determine whether there exists at least one pair of adjacent blocks (sharing a face) that share the same color}
\qualinputrow{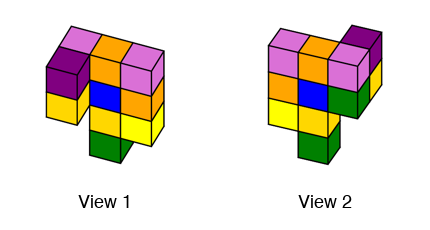}
\modelrow
False & True\tickno & True\tickno & False\tickok & True\tickno & True\tickno & True\tickno & True\tickno & True\tickno \\
}{T10 Color Connected instance with a mixed split.}{tab:qual_canonical_t10_99}
\vspace{-\qualsamplegap}% cancel the gap after the last sample
}]

\twocolumn[{%
\subsection{T11: Projection}
\label{sec:app_dataset_t11}

\setlength{\multicolsep}{3pt}
\begin{multicols}{2}
T11 opens the \sethlcolor{task_overview_purple}\hl{Cross-view geometry} group. Given two object views and three candidate orthographic projections, it asks for the valid or invalid candidate's index. Candidates cover the three principal planes in shuffled order. Invalid-selection trials contain two original projections and one perturbed projection; valid-selection trials contain one original and two perturbed projections. Distractors are generated by editing occupied two-dimensional cells, and labels record the designated candidate's index after shuffling, from 0 to 2. The task tests consistency between object views and projections; the examples below use invalid selection.
\end{multicols}
\vspace{10pt}

\qualcanonicaltable{
\promptline{identify the invalid projection}
\qualinputrow{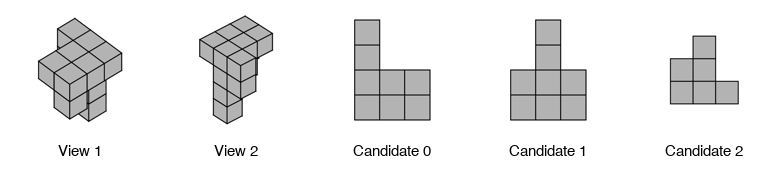}
\modelrow
2 & 2\tickok & 0\tickno & 2\tickok & 2\tickok & 2\tickok & 2\tickok & 2\tickok & 2\tickok \\
}{T11 Projection instance with a mixed split.}{tab:qual_canonical_t11_36}

\qualcanonicaltable{
\promptline{identify the invalid projection}
\qualinputrow{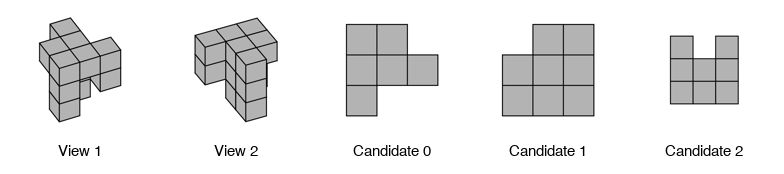}
\modelrow
2 & 2\tickok & 1\tickno & 2\tickok & 2\tickok & 1\tickno & 2\tickok & 0\tickno & 1\tickno \\
}{T11 Projection instance with a mixed split.}{tab:qual_canonical_t11_5}

\qualcanonicaltable{
\promptline{identify the invalid projection}
\qualinputrow{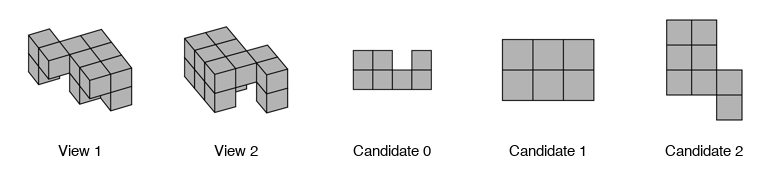}
\modelrow
0 & 2\tickno & 1\tickno & 1\tickno & 2\tickno & 2\tickno & 1\tickno & 2\tickno & 2\tickno \\
}{T11 Projection instance where all eight models answer incorrectly.}{tab:qual_canonical_t11_28}
\vspace{-\qualsamplegap}% cancel the gap after the last sample
}]

\twocolumn[{%
\subsection{T12: Reconstruct}
\label{sec:app_dataset_t12}

\setlength{\multicolsep}{3pt}
\begin{multicols}{2}
T12 tests \sethlcolor{task_overview_purple}\hl{Cross-view geometry} in the reverse direction. Given three canonical projections of one object and two candidate query views, it asks which candidate was rendered from that object. The other view depicts a different generated object whose block count differs by at most two. Projection order and candidate positions are randomized, and the label is the source object's candidate index, 0 or 1. T12 evaluates selection between supplied views rather than requiring an explicit voxel reconstruction, complementing T11's selection among projections.
\end{multicols}
\vspace{10pt}

\qualcanonicaltable{
\promptline{identify the positive view}
\qualinputrow{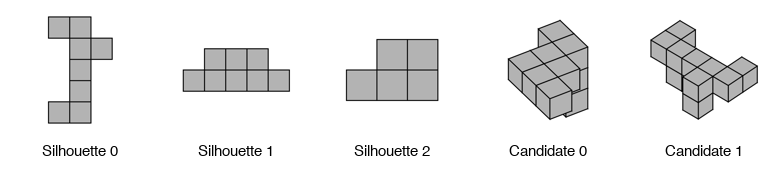}
\modelrow
1 & 1\tickok & 1\tickok & 1\tickok & 1\tickok & 1\tickok & 1\tickok & 1\tickok & 1\tickok \\
}{T12 Reconstruct instance where all eight models answer correctly.}{tab:qual_canonical_t12_22}

\qualcanonicaltable{
\promptline{identify the positive view}
\qualinputrow{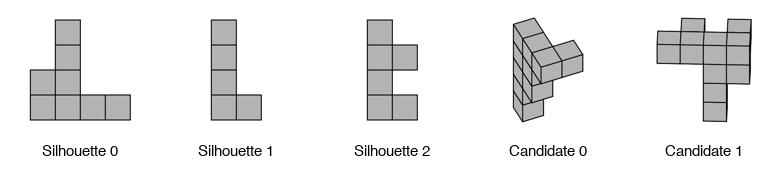}
\modelrow
1 & 1\tickok & 1\tickok & 1\tickok & 1\tickok & 0\tickno & 0\tickno & 0\tickno & 0\tickno \\
}{T12 Reconstruct instance with a mixed split.}{tab:qual_canonical_t12_0}

\qualcanonicaltable{
\promptline{identify the positive view}
\qualinputrow{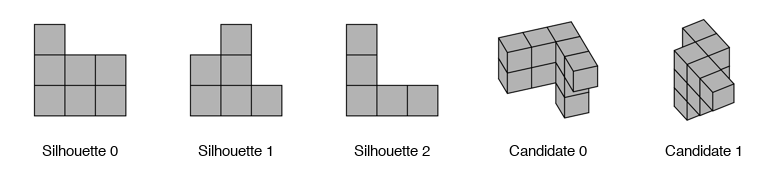}
\modelrow
0 & 1\tickno & 1\tickno & 1\tickno & 1\tickno & 1\tickno & 0\tickok & 1\tickno & 1\tickno \\
}{T12 Reconstruct instance with a mixed split.}{tab:qual_canonical_t12_6}
\vspace{-\qualsamplegap}% cancel the gap after the last sample
}]

\twocolumn[{%
\subsection{T13: Mixture-of-Shapes}
\label{sec:app_dataset_t13}

\setlength{\multicolsep}{3pt}
\begin{multicols}{2}
T13 forms the \sethlcolor{task_overview_red}\hl{Multi-object scene reasoning} group, extending projection selection to scenes with multiple geometric primitives and inter-object occlusion. Given two scene views and three candidate canonical projections in shuffled order, the task asks for the valid or invalid candidate. As in T11, trials contain one invalid candidate among two valid ones, or one valid among two invalid ones, with labels indexed from 0 to 2. Here, distractors are projections of modified three-dimensional scenes rather than edits to voxel projection cells. The examples below use invalid selection.
\end{multicols}
\vspace{10pt}

\qualcanonicaltable{
\promptline{identify the invalid silhouette}
\qualinputrow{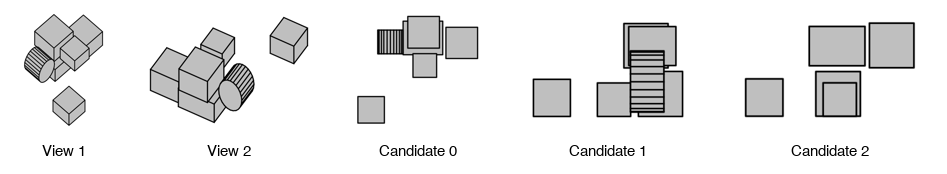}
\modelrow
2 & 2\tickok & 2\tickok & 2\tickok & 2\tickok & 2\tickok & 2\tickok & 2\tickok & 2\tickok \\
}{T13 Mixture-of-Shapes instance where all eight models answer correctly.}{tab:qual_canonical_t13_32}

\qualcanonicaltable{
\promptline{identify the invalid silhouette}
\qualinputrow{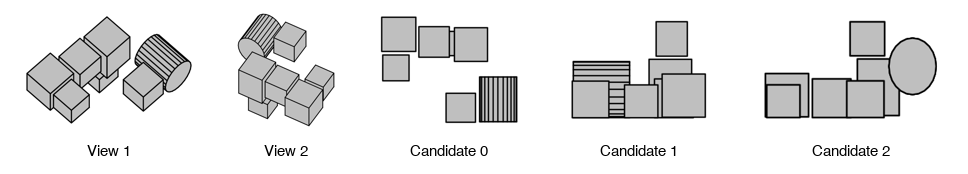}
\modelrow
2 & 2\tickok & 0\tickno & 2\tickok & 2\tickok & 0\tickno & 1\tickno & 0\tickno & 2\tickok \\
}{T13 Mixture-of-Shapes instance with a mixed split.}{tab:qual_canonical_t13_5}

\qualcanonicaltable{
\promptline{identify the invalid silhouette}
\qualinputrow{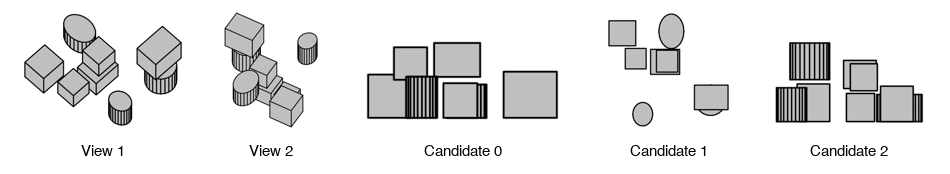}
\modelrow
0 & 2\tickno & 1\tickno & 2\tickno & 2\tickno & 2\tickno & 2\tickno & 2\tickno & 2\tickno \\
}{T13 Mixture-of-Shapes instance where all eight models answer incorrectly.}{tab:qual_canonical_t13_41}
\vspace{-\qualsamplegap}% cancel the gap after the last sample
}]

\clearpage
\section{Qualitative Results}
\label{sec:app_qualitative_results}
\subsection{Visual-Cue Add-Ons: Instance-Level Evidence}
\label{sec:app_qual_visual_cues}

These panels give the instance-level qualitative counterpart to the aggregate color-effect analysis in Appendix~\ref{sec:app_visual_cues_Color} and the aggregate axes-effect analysis in Appendix~\ref{sec:app_visual_cues_Axes}. Each panel below pairs the add-on absent and add-on present renderings of one task instance.

\newcommand{\qualinputrowVC}[1]{\multicolumn{5}{@{}c@{}}{\includegraphics[width=0.92\textwidth,height=1.05in,keepaspectratio]{#1}}\\[-0.5mm]}
\newcommand{\promptlineVC}[1]{\multicolumn{5}{@{}l@{}}{\parbox{0.92\textwidth}{\itshape ``\ldots{}#1\ldots{}''}}\\[-0.2mm]}
\newcommand{\modelrowVC}{\makebox[\linewidth][c]{\textbf{GT}} & \micon{qwen_icon.png} & \micon{gemma_icon.png} & \micon{glmv_icon.png} & \micon{kimi_icon.png} \\}
\newcommand{\fourwrap}[1]{\begin{tabular}[t]{@{}c@{\hspace{1pt}}c@{\hspace{1pt}}c@{\hspace{1pt}}c@{}}#1\end{tabular}}
\newcommand{\fivewrap}[1]{\begin{tabular}[t]{@{}c@{\hspace{1pt}}c@{\hspace{1pt}}c@{\hspace{1pt}}c@{\hspace{1pt}}c@{}}#1\end{tabular}}

\onecolumn

\begin{table}[H]
\centering
\small
\renewcommand{\arraystretch}{1.16}
\setlength{\tabcolsep}{2pt}
\begin{tabular}{*{5}{>{\centering\arraybackslash}p{\dimexpr0.18\textwidth+3.2pt\relax}}}
\toprule
\multicolumn{5}{l}{\textbf{T1 Volume}}\\
\promptlineVC{count the total number of blocks that make up the 3D object}
\multicolumn{5}{l}{\textit{(a) Color absent.}}\\
\qualinputrowVC{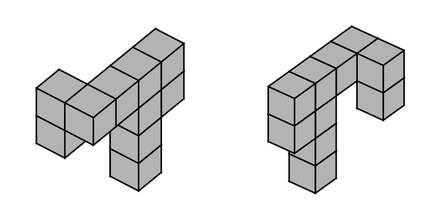}
\modelrowVC
$10$ & $12$\tickno & $11$\tickno & $12$\tickno & $10$\tickok \\
\addlinespace[2pt]
\multicolumn{5}{l}{\textit{(b) Color present.}}\\
\qualinputrowVC{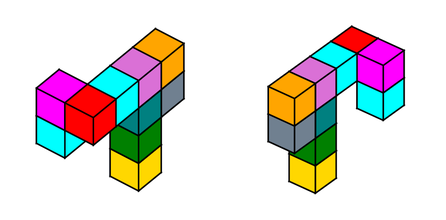}
\modelrowVC
$10$ & $10$\tickok & $10$\tickok & $12$\tickno & $10$\tickok \\
\bottomrule
\end{tabular}
\caption{T1 Volume, color add-on qualitative panel. Each block compares the same sample with and without color. Color present is likely to help.}
\label{tab:qual_color_cue_t1}
\end{table}

\begin{table}[H]
\centering
\small
\renewcommand{\arraystretch}{1.16}
\setlength{\tabcolsep}{2pt}
\begin{tabular}{*{5}{>{\centering\arraybackslash}p{\dimexpr0.18\textwidth+3.2pt\relax}}}
\toprule
\multicolumn{5}{l}{\textbf{T3 Difference}}\\
\promptlineVC{analyze both objects and determine the difference in their block compositions}
\multicolumn{5}{l}{\textit{(a) Color absent.}}\\
\qualinputrowVC{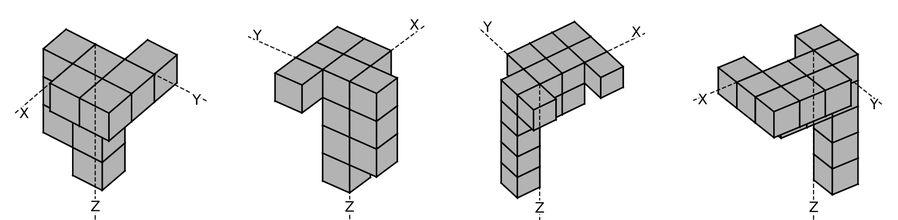}
\modelrowVC
$(5,6,10)$ & $(10,2,2)$\tickno & $(8,4,5)$\tickno & $(8,4,5)$\tickno & $(2,4,5)$\tickno \\
\addlinespace[2pt]
\multicolumn{5}{l}{\textit{(b) Color present (color-aware difference).}}\\
\qualinputrowVC{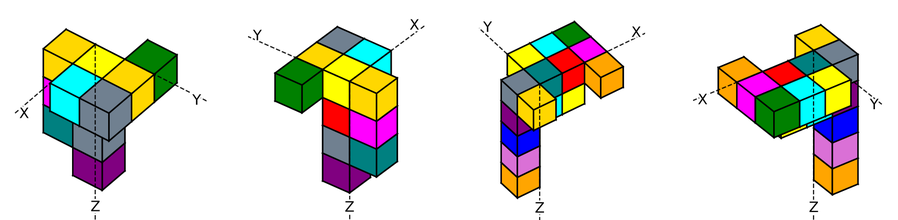}
\modelrowVC
$(0,11,15)$ & $(0,10,10)$\tickno & $(5,6,10)$\tickno & $(0,9,12)$\tickno & $(2,5,5)$\tickno \\
\bottomrule
\end{tabular}
\caption{T3 Difference, color add-on qualitative panel. Each block compares the same sample with and without color. Color present is likely to help.}
\label{tab:qual_color_cue_t3}
\end{table}

\begin{table}[H]
\centering
\small
\renewcommand{\arraystretch}{1.16}
\setlength{\tabcolsep}{2pt}
\begin{tabular}{*{5}{>{\centering\arraybackslash}p{\dimexpr0.18\textwidth+3.2pt\relax}}}
\toprule
\multicolumn{5}{l}{\textbf{T11 Projection}}\\
\promptlineVC{identify the invalid projection}
\multicolumn{5}{l}{\textit{(a) Color absent.}}\\
\qualinputrowVC{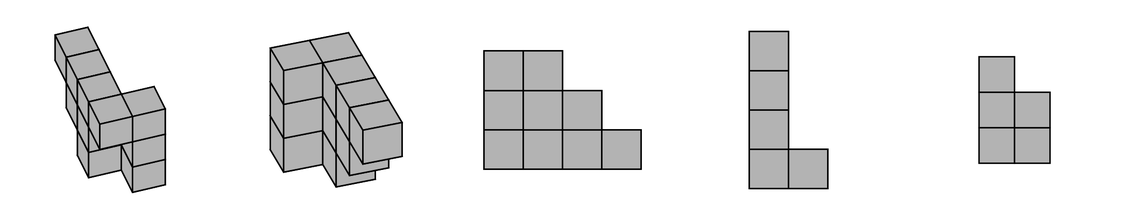}
\modelrowVC
$2$ & $0$\tickno & $0$\tickno & $2$\tickok & $0$\tickno \\
\addlinespace[2pt]
\multicolumn{5}{l}{\textit{(b) Color present.}}\\
\qualinputrowVC{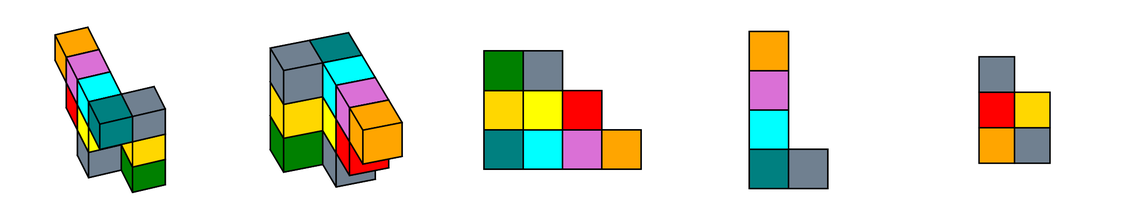}
\modelrowVC
$2$ & $2$\tickok & $2$\tickok & $2$\tickok & $2$\tickok \\
\bottomrule
\end{tabular}
\caption{T11 Projection, color add-on qualitative panel. Each block compares the same sample with and without color. Color present is likely to help.}
\label{tab:qual_color_cue_t11}
\end{table}

\begin{table}[H]
\centering
\small
\renewcommand{\arraystretch}{1.16}
\setlength{\tabcolsep}{2pt}
\begin{tabular}{*{5}{>{\centering\arraybackslash}p{\dimexpr0.18\textwidth+3.2pt\relax}}}
\toprule
\multicolumn{5}{l}{\textbf{T12 Reconstruct}}\\
\promptlineVC{identify the positive view}
\multicolumn{5}{l}{\textit{(a) Color absent.}}\\
\qualinputrowVC{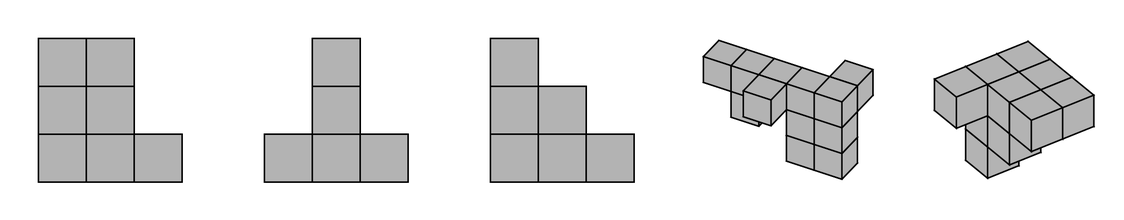}
\modelrowVC
$1$ & $0$\tickno & $0$\tickno & $0$\tickno & $0$\tickno \\
\addlinespace[2pt]
\multicolumn{5}{l}{\textit{(b) Color present.}}\\
\qualinputrowVC{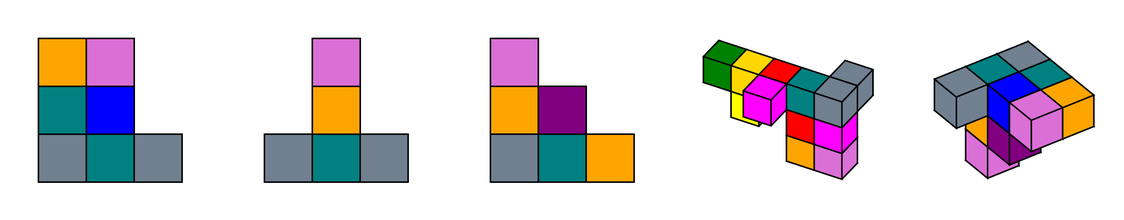}
\modelrowVC
$1$ & $1$\tickok & $1$\tickok & $1$\tickok & $1$\tickok \\
\bottomrule
\end{tabular}
\caption{T12 Reconstruct, color add-on qualitative panel. Each block compares the same sample with and without color. Color present is likely to help.}
\label{tab:qual_color_cue_t12}
\end{table}

\begin{table}[H]
\centering
\small
\renewcommand{\arraystretch}{1.16}
\setlength{\tabcolsep}{2pt}
\begin{tabular}{*{5}{>{\centering\arraybackslash}p{\dimexpr0.18\textwidth+3.2pt\relax}}}
\toprule
\multicolumn{5}{l}{\textbf{T13 Mixture-of-Shapes}}\\
\promptlineVC{identify the invalid silhouette}
\multicolumn{5}{l}{\textit{(a) Color absent.}}\\
\qualinputrowVC{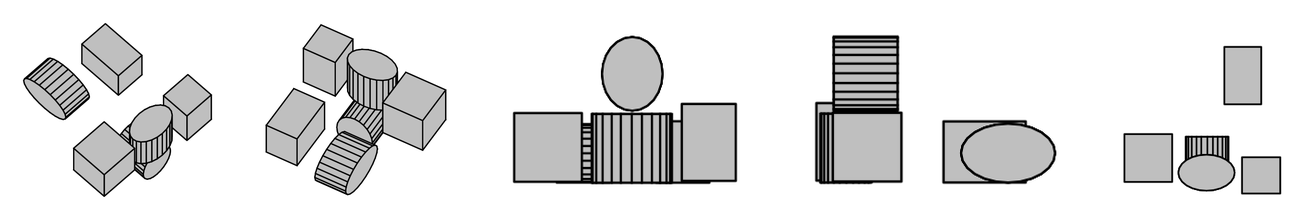}
\modelrowVC
$2$ & $1$\tickno & $2$\tickok & $2$\tickok & $2$\tickok \\
\addlinespace[2pt]
\multicolumn{5}{l}{\textit{(b) Color present.}}\\
\qualinputrowVC{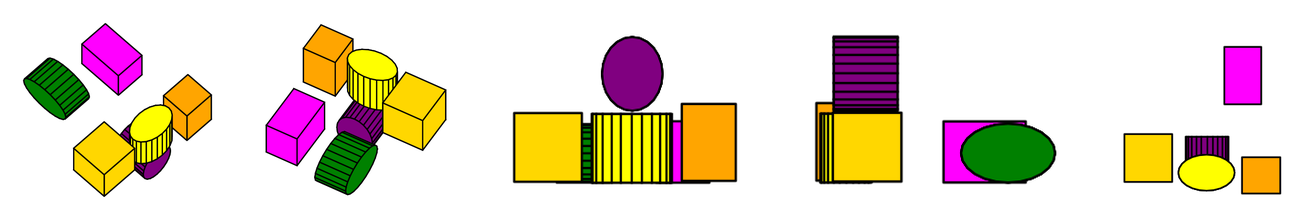}
\modelrowVC
$2$ & $2$\tickok & $1$\tickno & $2$\tickok & $2$\tickok \\
\bottomrule
\end{tabular}
\caption{T13 Mixture-of-Shapes, color add-on qualitative panel. Each block compares the same sample with and without color. Color present is likely to help.}
\label{tab:qual_color_cue_t13}
\end{table}

\begin{table}[H]
\centering
\small
\renewcommand{\arraystretch}{1.16}
\setlength{\tabcolsep}{2pt}
\begin{tabular}{*{5}{>{\centering\arraybackslash}p{\dimexpr0.18\textwidth+3.2pt\relax}}}
\toprule
\multicolumn{5}{l}{\textbf{T1 Volume}}\\
\promptlineVC{count the total number of blocks that make up the 3D object}
\multicolumn{5}{l}{\textit{(a) Axes absent.}}\\
\qualinputrowVC{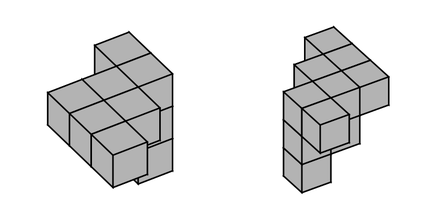}
\modelrowVC
$10$ & $10$\tickok & $10$\tickok & $10$\tickok & $10$\tickok \\
\addlinespace[2pt]
\multicolumn{5}{l}{\textit{(b) Axes present.}}\\
\qualinputrowVC{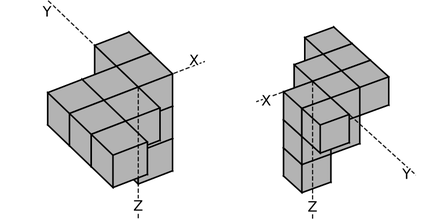}
\modelrowVC
$10$ & $12$\tickno & $11$\tickno & $12$\tickno & $10$\tickok \\
\bottomrule
\end{tabular}
\caption{T1 Volume, axes add-on qualitative panel. Each block compares the same sample with and without a coordinate-axis. Axes present is likely to hurt.}
\label{tab:qual_axes_cue_t1}
\end{table}

\begin{table}[H]
\centering
\small
\renewcommand{\arraystretch}{1.16}
\setlength{\tabcolsep}{2pt}
\begin{tabular}{*{5}{>{\centering\arraybackslash}p{\dimexpr0.18\textwidth+3.2pt\relax}}}
\toprule
\multicolumn{5}{l}{\textbf{T2 Packaging}}\\
\promptlineVC{determine whether the object can fit in the boundary in any possible orientation. Boundary: \text{[2,3,2]}}
\multicolumn{5}{l}{\textit{(a) Axes absent.}}\\
\qualinputrowVC{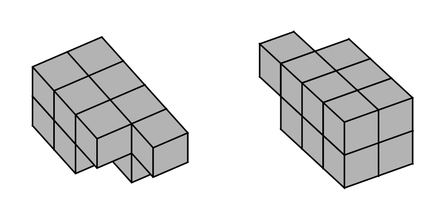}
\modelrowVC
False & False\tickok & False\tickok & True\tickno & True\tickno \\
\addlinespace[2pt]
\multicolumn{5}{l}{\textit{(b) Axes present.}}\\
\qualinputrowVC{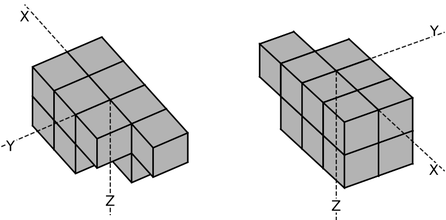}
\modelrowVC
False & True\tickno & True\tickno & True\tickno & True\tickno \\
\bottomrule
\end{tabular}
\caption{T2 Packaging, axes add-on qualitative panel. Each block compares the same sample with and without a coordinate-axis. Axes present is likely to hurt.}
\label{tab:qual_axes_cue_t2}
\end{table}

\begin{table}[H]
\centering
\small
\renewcommand{\arraystretch}{1.16}
\setlength{\tabcolsep}{2pt}
\begin{tabular}{*{5}{>{\centering\arraybackslash}p{\dimexpr0.18\textwidth+3.2pt\relax}}}
\toprule
\multicolumn{5}{l}{\textbf{T4 Color Count}}\\
\promptlineVC{count the number of unique colors in the object}
\multicolumn{5}{l}{\textit{(a) Axes absent.}}\\
\qualinputrowVC{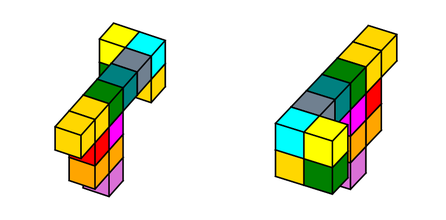}
\modelrowVC
$10$ & $10$\tickok & $10$\tickok & $10$\tickok & $10$\tickok \\
\addlinespace[2pt]
\multicolumn{5}{l}{\textit{(b) Axes present.}}\\
\qualinputrowVC{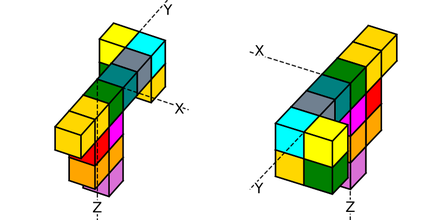}
\modelrowVC
$10$ & $9$\tickno & $9$\tickno & $10$\tickok & $10$\tickok \\
\bottomrule
\end{tabular}
\caption{T4 Color Count, axes add-on qualitative panel. Each block compares the same sample with and without a coordinate-axis. Axes present is likely to hurt.}
\label{tab:qual_axes_cue_t4}
\end{table}

\begin{table}[H]
\centering
\small
\renewcommand{\arraystretch}{1.16}
\setlength{\tabcolsep}{2pt}
\begin{tabular}{*{5}{>{\centering\arraybackslash}p{\dimexpr0.18\textwidth+3.2pt\relax}}}
\toprule
\multicolumn{5}{l}{\textbf{T5 Color Distribution}}\\
\promptlineVC{give the color distribution in the object}
\multicolumn{5}{l}{\textit{(a) Axes absent.}}\\
\qualinputrowVC{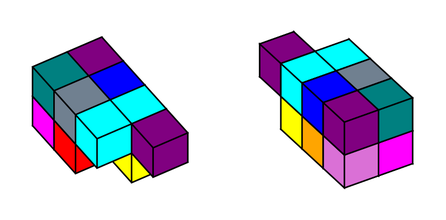}
\modelrowVC
\fivewrap{\cdY{1}&\cdC{2}&\cdB{1}&\cdP{2}&\cdO{1}\\\cdOr{1}&\cdT{1}&\cdS{1}&\cdM{1}&\cdR{1}} & \fivewrap{\cdY{2}&\cdC{4}&\cdB{2}&\cdP{3}&\cdO{1}\\\cdOr{1}&\cdT{2}&\cdS{2}&\cdM{2}&\cdR{1}}\tickno & \fivewrap{\cdY{1}&\cdC{2}&\cdB{1}&\cdP{2}&\cdO{1}\\\cdOr{1}&\cdT{1}&\cdS{1}&\cdM{1}&\cdR{1}}\tickok & \fivewrap{\cdY{1}&\cdC{3}&\cdG{1}&\cdB{1}&\cdP{3}\\\cdOr{2}&\cdT{2}&\cdS{1}&\cdM{2}&\cdR{1}} \tickno & \fivewrap{\cdY{3}&\cdC{6}&\cdB{3}&\cdP{6}&\cdT{3}\\\cdS{3}&\cdGo{1}&\cdM{3}&\cdR{3}&}\tickno \\
\addlinespace[2pt]
\multicolumn{5}{l}{\textit{(b) Axes present.}}\\
\qualinputrowVC{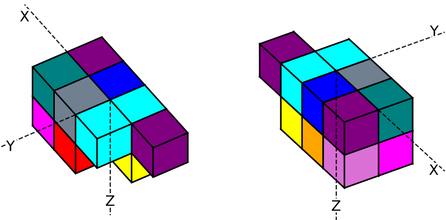}
\modelrowVC
\fivewrap{\cdY{1}&\cdC{2}&\cdB{1}&\cdP{2}&\cdO{1}\\\cdOr{1}&\cdT{1}&\cdS{1}&\cdM{1}&\cdR{1}} & \fivewrap{\cdY{2}&\cdC{4}&\cdB{1}&\cdP{3}&\cdO{1}\\\cdOr{1}&\cdT{1}&\cdS{1}&\cdM{2}&\cdR{1}}\tickno & \fivewrap{\cdY{1}&\cdC{3}&\cdB{1}&\cdP{3}&\cdT{1}\\\cdS{1}&\cdM{1}&\cdR{1}&&}\tickno & \fivewrap{\cdY{2}&\cdC{4}&\cdB{2}&\cdP{4}&\cdO{1}\\\cdT{2}&\cdS{1}&\cdM{2}&\cdR{2}&}\tickno & \fivewrap{\cdY{2}&\cdC{8}&\cdP{1}&\cdO{2}&\cdOr{10}\\\cdR{5}&&&&}\tickno \\
\bottomrule
\end{tabular}
\caption{T5 Color Distribution, axes add-on qualitative panel. Each block compares the same sample with and without a coordinate-axis. Axes present is likely to hurt.}
\label{tab:qual_axes_cue_t5}
\end{table}

\begin{table}[H]
\centering
\small
\renewcommand{\arraystretch}{1.16}
\setlength{\tabcolsep}{2pt}
\begin{tabular}{*{5}{>{\centering\arraybackslash}p{\dimexpr0.18\textwidth+3.2pt\relax}}}
\toprule
\multicolumn{5}{l}{\textbf{T6 Local Structure}}\\
\promptlineVC{determine whether the object has at least one column (vertical stack of blocks) with a height greater than 3}
\multicolumn{5}{l}{\textit{(a) Axes absent.}}\\
\qualinputrowVC{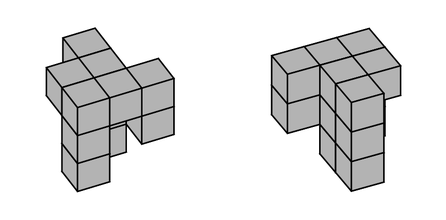}
\modelrowVC
False & False\tickok & False\tickok & False\tickok & True\tickno \\
\addlinespace[2pt]
\multicolumn{5}{l}{\textit{(b) Axes present.}}\\
\qualinputrowVC{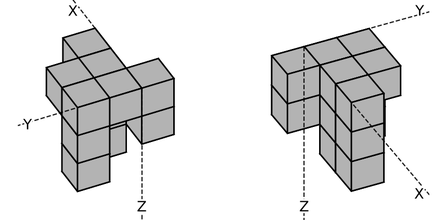}
\modelrowVC
False & False\tickok & False\tickok & True\tickno & True\tickno \\
\bottomrule
\end{tabular}
\caption{T6 Local Structure, axes add-on qualitative panel. Each block compares the same sample with and without a coordinate-axis. Axes present is likely to hurt.}
\label{tab:qual_axes_cue_t6}
\end{table}

\begin{table}[H]
\centering
\small
\renewcommand{\arraystretch}{1.16}
\setlength{\tabcolsep}{2pt}
\begin{tabular}{*{5}{>{\centering\arraybackslash}p{\dimexpr0.18\textwidth+3.2pt\relax}}}
\toprule
\multicolumn{5}{l}{\textbf{T7 Color Dominance}}\\
\promptlineVC{determine the dominant color(s) in the object --- the color(s) with the highest block count}
\multicolumn{5}{l}{\textit{(a) Axes absent.}}\\
\qualinputrowVC{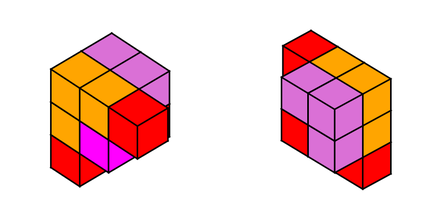}
\modelrowVC
\fivewrap{\cdchip{bedOrange}&\cdchip{bedOrchid}&\cdchip{bedRed}&&} & \fivewrap{\cdchip{bedOrange}&\cdchip{bedOrchid}&\cdchip{bedRed}&&}\tickok & \fivewrap{\cdchip{bedOrange}&\cdchip{bedOrchid}&\cdchip{bedRed}&&}\tickok & \fivewrap{\cdchip{bedYellow}&\cdchip{bedPurple}&\cdchip{bedRed}&&}\tickno & \fivewrap{\cdchip{bedPurple}&\cdchip{bedOrange}&\cdchip{bedRed}&&}\tickno \\
\addlinespace[2pt]
\multicolumn{5}{l}{\textit{(b) Axes present.}}\\
\qualinputrowVC{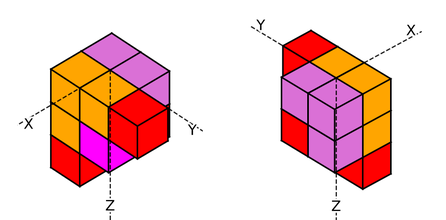}
\modelrowVC
\fivewrap{\cdchip{bedOrange}&\cdchip{bedOrchid}&\cdchip{bedRed}&&} & \fivewrap{\cdchip{bedOrange}&\cdchip{bedOrchid}&\cdchip{bedMagenta}&\cdchip{bedRed}&}\tickno & \fivewrap{\cdchip{bedOrange}&&&&}\tickno & \fivewrap{\cdchip{bedYellow}&\cdchip{bedPurple}&\cdchip{bedRed}&&}\tickno & \fivewrap{\cdchip{bedPurple}&\cdchip{bedOrange}&\cdchip{bedRed}&&}\tickno \\
\bottomrule
\end{tabular}
\caption{T7 Color Dominance, axes add-on qualitative panel. Each block compares the same sample with and without a coordinate-axis. Axes present is likely to hurt.}
\label{tab:qual_axes_cue_t7}
\end{table}

\begin{table}[H]
\centering
\small
\renewcommand{\arraystretch}{1.16}
\setlength{\tabcolsep}{2pt}
\begin{tabular}{*{5}{>{\centering\arraybackslash}p{\dimexpr0.18\textwidth+3.2pt\relax}}}
\toprule
\multicolumn{5}{l}{\textbf{T8 Silhouette Symmetry}}\\
\promptlineVC{determine whether the object's silhouette is symmetric (vertically or horizontally)}
\multicolumn{5}{l}{\textit{(a) Axes absent.}}\\
\qualinputrowVC{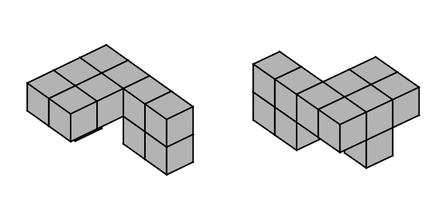}
\modelrowVC
False & False\tickok & False\tickok & False\tickok & False\tickok \\
\addlinespace[2pt]
\multicolumn{5}{l}{\textit{(b) Axes present.}}\\
\qualinputrowVC{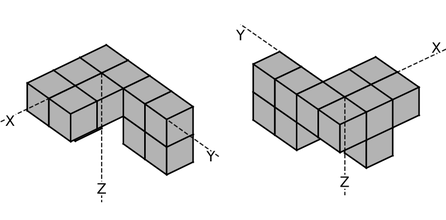}
\modelrowVC
False & True\tickno & True\tickno & False\tickok & False\tickok \\
\bottomrule
\end{tabular}
\caption{T8 Silhouette Symmetry, axes add-on qualitative panel. Each block compares the same sample with and without a coordinate-axis. Axes present is likely to hurt.}
\label{tab:qual_axes_cue_t8}
\end{table}

\begin{table}[H]
\centering
\small
\renewcommand{\arraystretch}{1.16}
\setlength{\tabcolsep}{2pt}
\begin{tabular}{*{5}{>{\centering\arraybackslash}p{\dimexpr0.18\textwidth+3.2pt\relax}}}
\toprule
\multicolumn{5}{l}{\textbf{T9 Visible Floors}}\\
\promptlineVC{count the number of visible layers (floors/horizontal levels) in the object}
\multicolumn{5}{l}{\textit{(a) Axes absent.}}\\
\qualinputrowVC{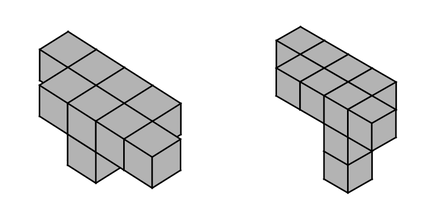}
\modelrowVC
$3$ & $3$\tickok & $3$\tickok & $3$\tickok & $4$\tickno \\
\addlinespace[2pt]
\multicolumn{5}{l}{\textit{(b) Axes present.}}\\
\qualinputrowVC{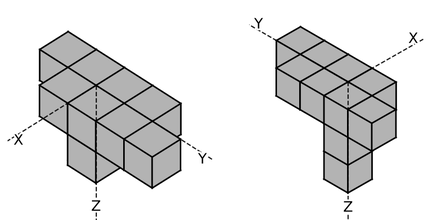}
\modelrowVC
$3$ & $3$\tickok & $2$\tickno & $2$\tickno & $4$\tickno \\
\bottomrule
\end{tabular}
\caption{T9 Visible Floors, axes add-on qualitative panel. Each block compares the same sample with and without a coordinate-axis. Axes present is likely to hurt.}
\label{tab:qual_axes_cue_t9}
\end{table}

\begin{table}[H]
\centering
\small
\renewcommand{\arraystretch}{1.16}
\setlength{\tabcolsep}{2pt}
\begin{tabular}{*{5}{>{\centering\arraybackslash}p{\dimexpr0.18\textwidth+3.2pt\relax}}}
\toprule
\multicolumn{5}{l}{\textbf{T10 Color Connected}}\\
\promptlineVC{determine whether there exists at least one pair of adjacent blocks (sharing a face) that share the same color}
\multicolumn{5}{l}{\textit{(a) Axes absent.}}\\
\qualinputrowVC{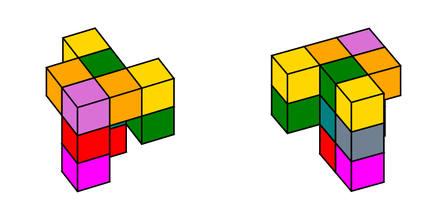}
\modelrowVC
False & True\tickno & False\tickok & False\tickok & True\tickno \\
\addlinespace[2pt]
\multicolumn{5}{l}{\textit{(b) Axes present.}}\\
\qualinputrowVC{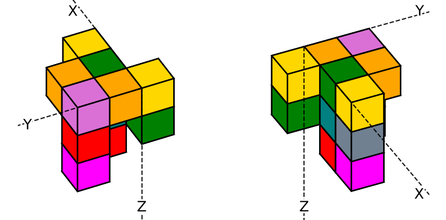}
\modelrowVC
False & True\tickno & True\tickno & True\tickno & True\tickno \\
\bottomrule
\end{tabular}
\caption{T10 Color Connected, axes add-on qualitative panel. Each block compares the same sample with and without a coordinate-axis. Axes present is likely to hurt.}
\label{tab:qual_axes_cue_t10}
\end{table}

\begin{table}[H]
\centering
\small
\renewcommand{\arraystretch}{1.16}
\setlength{\tabcolsep}{2pt}
\begin{tabular}{*{5}{>{\centering\arraybackslash}p{\dimexpr0.18\textwidth+3.2pt\relax}}}
\toprule
\multicolumn{5}{l}{\textbf{T11 Projection}}\\
\promptlineVC{identify the invalid projection}
\multicolumn{5}{l}{\textit{(a) Axes absent.}}\\
\qualinputrowVC{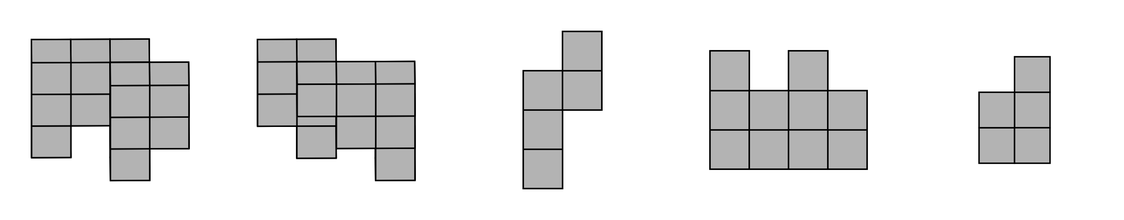}
\modelrowVC
$2$ & $2$\tickok & $0$\tickno & $2$\tickok & $0$\tickno \\
\addlinespace[2pt]
\multicolumn{5}{l}{\textit{(b) Axes present.}}\\
\qualinputrowVC{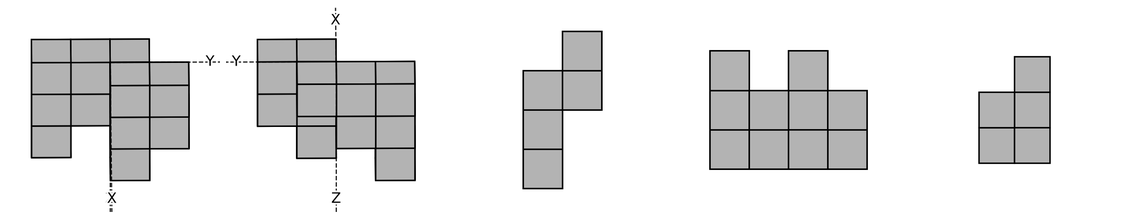}
\modelrowVC
$2$ & $1$\tickno & $0$\tickno & $1$\tickno & $2$\tickok \\
\bottomrule
\end{tabular}
\caption{T11 Projection, axes add-on qualitative panel. Each block compares the same sample with and without a coordinate-axis. Axes present is likely to hurt.}
\label{tab:qual_axes_cue_t11}
\end{table}

\begin{table}[H]
\centering
\small
\renewcommand{\arraystretch}{1.16}
\setlength{\tabcolsep}{2pt}
\begin{tabular}{*{5}{>{\centering\arraybackslash}p{\dimexpr0.18\textwidth+3.2pt\relax}}}
\toprule
\multicolumn{5}{l}{\textbf{T12 Reconstruct}}\\
\promptlineVC{identify the positive view}
\multicolumn{5}{l}{\textit{(a) Axes absent.}}\\
\qualinputrowVC{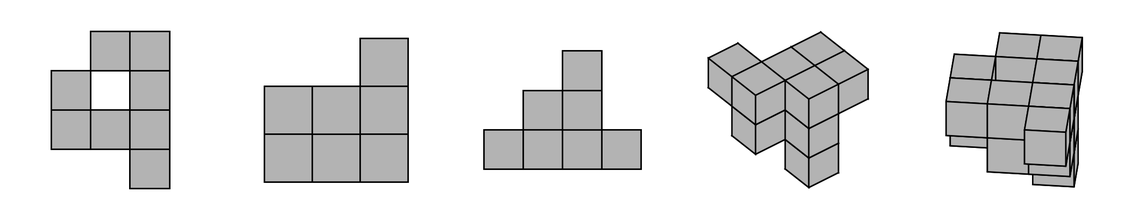}
\modelrowVC
$1$ & $1$\tickok & $1$\tickok & $1$\tickok & $1$\tickok \\
\addlinespace[2pt]
\multicolumn{5}{l}{\textit{(b) Axes present.}}\\
\qualinputrowVC{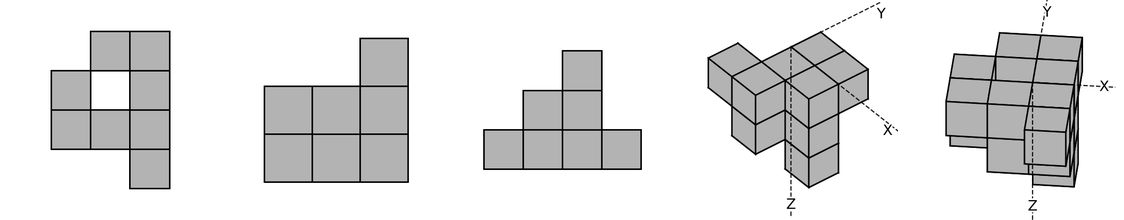}
\modelrowVC
$1$ & $0$\tickno & $0$\tickno & $0$\tickno & $0$\tickno \\
\bottomrule
\end{tabular}
\caption{T12 Reconstruct, axes add-on qualitative panel. Each block compares the same sample with and without a coordinate-axis. Axes present is likely to hurt.}
\label{tab:qual_axes_cue_t12}
\end{table}

\twocolumn
\subsection{Parameter Scaling: Instance-Level Evidence}
\label{sec:app_qual_scaling}

These panels give the instance-level qualitative counterpart to the scaling-curve analysis in Appendix~\ref{sec:app_scale_full} and to the Qwen3.5-only split in Figure~\ref{fig:scale_analysis}, covering all 13 BedBench tasks in two descriptive groups based on the observed Qwen3.5 scaling curves: scaling-responsive tasks (\ttwolong{}, \tfourlong{}, \tsixlong{}, \tsevenlong{}, \tninelong{}, \televenlong{}) and scaling-resistant tasks (\tonelong{}, \tthreelong{}, \tfivelong{}, \teightlong{}, \ttenlong{}, \ttwelvelong{}, \tthirteenlong{}) across the Qwen3.5 and Gemma4 model ladders.

\newcommand{\scalehdr}{\textbf{Model} & \textbf{GT} & \makecell{0.8B} & \makecell{2B\\\scriptsize(E2B)} & \makecell{4B\\\scriptsize(E4B)} & \makecell{9B} & \makecell{27B\\\scriptsize(26B-A4B)} & \makecell{35B-A3B\\\scriptsize(31B)} \\}
\newcommand{\qualinputrowSm}[1]{\multicolumn{8}{@{}c@{}}{\includegraphics[width=0.92\textwidth,height=1.05in,keepaspectratio]{#1}}\\[-0.5mm]}
\newcommand{\promptlineSm}[1]{\multicolumn{8}{@{}l@{}}{\parbox{0.92\textwidth}{\itshape ``\ldots{}#1\ldots{}''}}\\[-0.2mm]}

\onecolumn

\begingroup
\small
\renewcommand{\arraystretch}{1.16}
\setlength{\tabcolsep}{2pt}
\begin{longtable}{>{\raggedright\arraybackslash}p{0.08\textwidth}*{7}{>{\centering\arraybackslash}p{0.12\textwidth}}}
\toprule
\endfirsthead
\toprule
\multicolumn{8}{l}{\tablename~\thetable{} continued.}\\
\midrule
\endhead
\multicolumn{8}{l}{\textbf{T2 Packaging -- scaling-responsive}}\\
\promptlineSm{determine whether the object can fit in the boundary in any orientation. Boundary: [2,2,2]}
\qualinputrowSm{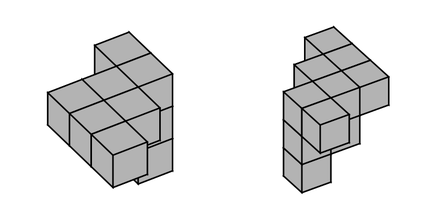}
\scalehdr
Qwen3.5 & \multirow{2}{*}{False} & True\tickno & True\tickno & True\tickno & True\tickno & False\tickok & False\tickok \\
Gemma4 & & -- & True\tickno & False\tickok & -- & False\tickok & False\tickok \\
\bottomrule
\caption{T2 Packaging, scaling-responsive instance, across the Qwen3.5 (0.8B--35B-A3B) and Gemma4 (E2B--31B, in parentheses) parameter ladders under Non-Thinking mode. Qwen3.5 flips from wrongly answering True at 0.8B--9B to correctly answering False (the object cannot fit the [2,2,2] boundary in any orientation) at 27B/35B-A3B; Gemma4 flips one step earlier, at E4B and up. Both families converge on the correct answer only at their larger checkpoints, mirroring the coarse-cue scaling gains quantified in Appendix~\ref{sec:app_scale_full}.}
\label{tab:qual_scaling_t2}\\
\end{longtable}
\endgroup

\begingroup
\small
\renewcommand{\arraystretch}{1.16}
\setlength{\tabcolsep}{2pt}
\begin{longtable}{>{\raggedright\arraybackslash}p{0.08\textwidth}*{7}{>{\centering\arraybackslash}p{0.12\textwidth}}}
\toprule
\endfirsthead
\toprule
\multicolumn{8}{l}{\tablename~\thetable{} continued.}\\
\midrule
\endhead
\multicolumn{8}{l}{\textbf{T4 Color Count -- scaling-responsive}}\\
\promptlineSm{count the number of unique colors in the object}
\qualinputrowSm{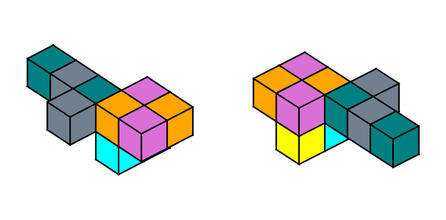}
\scalehdr
Qwen3.5 & \multirow{2}{*}{$6$} & $10$\tickno & $10$\tickno & $7$\tickno & $7$\tickno & $6$\tickok & $6$\tickok \\
Gemma4 & & -- & $11$\tickno & $8$\tickno & -- & $6$\tickok & $6$\tickok \\
\bottomrule
\caption{T4 Color Count, scaling-responsive instance, across the Qwen3.5 (0.8B--35B-A3B) and Gemma4 (E2B--31B, in parentheses) parameter ladders under Non-Thinking mode. Both families only reach the ground truth (6) at their two largest checkpoints, illustrating the scaling-responsive trend quantified in Appendix~\ref{sec:app_scale_full}.}
\label{tab:qual_scaling_t4}\\
\end{longtable}
\endgroup

\begingroup
\small
\renewcommand{\arraystretch}{1.16}
\setlength{\tabcolsep}{2pt}
\begin{longtable}{>{\raggedright\arraybackslash}p{0.08\textwidth}*{7}{>{\centering\arraybackslash}p{0.12\textwidth}}}
\toprule
\endfirsthead
\toprule
\multicolumn{8}{l}{\tablename~\thetable{} continued.}\\
\midrule
\endhead
\multicolumn{8}{l}{\textbf{T6 Local Structure -- scaling-responsive}}\\
\promptlineSm{determine whether any column (vertical stack of blocks) has a height greater than 3}
\qualinputrowSm{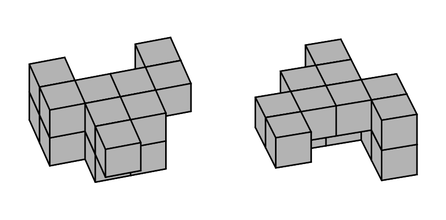}
\scalehdr
Qwen3.5 & \multirow{2}{*}{False} & True\tickno & True\tickno & False\tickok & False\tickok & False\tickok & False\tickok \\
Gemma4 & & -- & False\tickok & False\tickok & -- & False\tickok & False\tickok \\
\bottomrule
\caption{T6 Local Structure, scaling-responsive instance, across the Qwen3.5 (0.8B--35B-A3B) and Gemma4 (E2B--31B, in parentheses) parameter ladders under Non-Thinking mode,. The Qwen3.5 ladder flips from wrongly answering True at 0.8B/2B to correctly answering False from 4B upward, the trend quantified in Appendix~\ref{sec:app_scale_full}; Gemma4 is already correct at every checkpoint on this instance, so this row demonstrates the Qwen-side scaling gain rather than a matched two-family transition.}
\label{tab:qual_scaling_t6}\\
\end{longtable}
\endgroup

\begingroup
\small
\renewcommand{\arraystretch}{1.16}
\setlength{\tabcolsep}{2pt}
\begin{longtable}{>{\raggedright\arraybackslash}p{0.08\textwidth}*{7}{>{\centering\arraybackslash}p{0.12\textwidth}}}
\toprule
\endfirsthead
\toprule
\multicolumn{8}{l}{\tablename~\thetable{} continued.}\\
\midrule
\endhead
\multicolumn{8}{l}{\textbf{T7 Color Dominance -- scaling-responsive}}\\
\promptlineSm{determine the dominant color(s) in the object --- the color(s) with the highest block count}
\qualinputrowSm{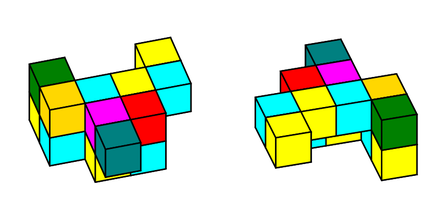}
\scalehdr
Qwen3.5 & \multirow{2}{*}{\fourwrap{\cdchip{bedCyan}&\cdchip{bedYellow}&&}} & \fourwrap{\cdchip{bedCyan}&\cdchip{bedGreen}&\cdchip{bedMagenta}&\cdchip{bedRed}\\\cdchip{bedTeal}&\cdchip{bedYellow}&&}\tickno & \fourwrap{\cdchip{bedCyan}&\cdchip{bedGreen}&\cdchip{bedMagenta}&\cdchip{bedRed}\\\cdchip{bedTeal}&\cdchip{bedYellow}&&}\tickno & \fourwrap{\cdchip{bedCyan}&\cdchip{bedYellow}&&}\tickok & \fourwrap{\cdchip{bedCyan}&\cdchip{bedYellow}&&}\tickok & \fourwrap{\cdchip{bedCyan}&\cdchip{bedYellow}&&}\tickok & \fourwrap{\cdchip{bedCyan}&\cdchip{bedYellow}&&}\tickok \\
Gemma4 & & -- & \fourwrap{\cdchip{bedCyan}&\cdchip{bedGreen}&\cdchip{bedRed}&\cdchip{bedYellow}}\tickno & \fourwrap{\cdchip{bedCyan}&\cdchip{bedYellow}&&}\tickok & -- & \fourwrap{\cdchip{bedCyan}&\cdchip{bedYellow}&&}\tickok & \fourwrap{\cdchip{bedCyan}&\cdchip{bedYellow}&&}\tickok \\
\bottomrule
\caption{T7 Color Dominance, scaling-responsive instance, across the Qwen3.5 (0.8B--35B-A3B) and Gemma4 (E2B--31B, in parentheses) parameter ladders under Non-Thinking mode. The two smallest checkpoints in each family over-list several non-dominant colors alongside the correct pair; every checkpoint at 4B/E4B and above converges exactly on the correct two-color set \{\textcolor{bedCyan}{cyan}, \textcolor{bedYellow}{yellow}\}, mirroring the coarse-cue scaling gains quantified in Appendix~\ref{sec:app_scale_full} (Figure~\ref{fig:scale_analysis}).}
\label{tab:qual_scaling_t7}\\
\end{longtable}
\endgroup

\begingroup
\small
\renewcommand{\arraystretch}{1.16}
\setlength{\tabcolsep}{2pt}
\begin{longtable}{>{\raggedright\arraybackslash}p{0.08\textwidth}*{7}{>{\centering\arraybackslash}p{0.12\textwidth}}}
\toprule
\endfirsthead
\toprule
\multicolumn{8}{l}{\tablename~\thetable{} continued.}\\
\midrule
\endhead
\multicolumn{8}{l}{\textbf{T9 Visible Floors -- scaling-responsive}}\\
\promptlineSm{count the number of visible layers (floors) in the object}
\qualinputrowSm{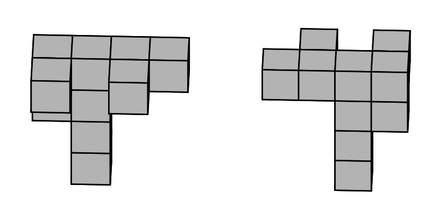}
\scalehdr
Qwen3.5 & \multirow{2}{*}{$4$} & $3$\tickno & $3$\tickno & $4$\tickok & $4$\tickok & $4$\tickok & $4$\tickok \\
Gemma4 & & -- & $3$\tickno & $4$\tickok & -- & $4$\tickok & $4$\tickok \\
\bottomrule
\caption{T9 Visible Floors, scaling-responsive instance, across the Qwen3.5 (0.8B--35B-A3B) and Gemma4 (E2B--31B, in parentheses) parameter ladders under Non-Thinking mode. Both families walk from wrongly answering 3 at their two smallest checkpoints to correctly answering 4 (the ground truth) from their third checkpoint upward, the same shape as the T4 instance above; the aggregate T9 curve is noisier in between than T2/T4/T6 (Appendix~\ref{sec:app_scale_full}), but this instance isolates the same net small-to-large improvement.}
\label{tab:qual_scaling_t9}\\
\end{longtable}
\endgroup

\begingroup
\small
\renewcommand{\arraystretch}{1.16}
\setlength{\tabcolsep}{2pt}
\begin{longtable}{>{\raggedright\arraybackslash}p{0.08\textwidth}*{7}{>{\centering\arraybackslash}p{0.12\textwidth}}}
\toprule
\endfirsthead
\toprule
\multicolumn{8}{l}{\tablename~\thetable{} continued.}\\
\midrule
\endhead
\multicolumn{8}{l}{\textbf{T11 Projection -- scaling-responsive}}\\
\promptlineSm{identify the invalid projection among the three candidates}
\qualinputrowSm{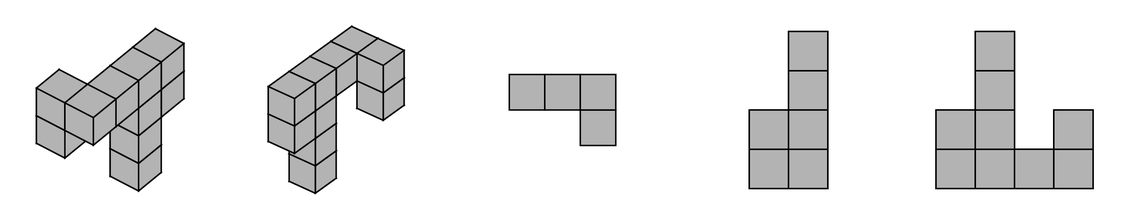}
\scalehdr
Qwen3.5 & \multirow{2}{*}{1} & 2\tickno & 2\tickno & 1\tickok & 1\tickok & 1\tickok & 1\tickok \\
Gemma4 & & -- & 2\tickno & 1\tickok & -- & 1\tickok & 1\tickok \\
\bottomrule
\caption{T11 Projection, scaling-responsive instance, across the Qwen3.5 (0.8B--35B-A3B) and Gemma4 (E2B--31B, in parentheses) parameter ladders under Non-Thinking mode. Both families flip from the wrong candidate (2) at their two smallest checkpoints to the correct 1 from their third checkpoint upward, the same shape as the T4/T9/T7 instances above, illustrating the responsive trend quantified in Appendix~\ref{sec:app_scale_full} (Figure~\ref{fig:scale_analysis}).}
\label{tab:qual_scaling_t11}\\
\end{longtable}
\endgroup

\begingroup
\small
\renewcommand{\arraystretch}{1.16}
\setlength{\tabcolsep}{2pt}
\begin{longtable}{>{\raggedright\arraybackslash}p{0.08\textwidth}*{7}{>{\centering\arraybackslash}p{0.12\textwidth}}}
\toprule
\endfirsthead
\toprule
\multicolumn{8}{l}{\tablename~\thetable{} continued.}\\
\midrule
\endhead
\multicolumn{8}{l}{\textbf{T1 Volume -- scaling-resistant for Qwen3.5, scaling-responsive for Gemma4}}\\
\promptlineSm{count the total number of blocks that make up the 3D object}
\qualinputrowSm{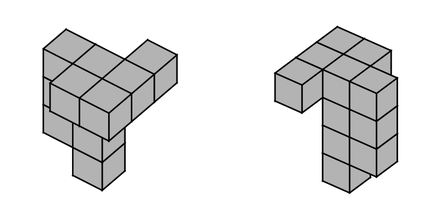}
\scalehdr
Qwen3.5 & \multirow{2}{*}{$11$} & $10$\tickno & $12$\tickno & $12$\tickno & $12$\tickno & $12$\tickno & $12$\tickno \\
Gemma4 & & -- & $14$\tickno & $11$\tickok & -- & $13$\tickno & $11$\tickok \\
\bottomrule
\caption{T1 Volume, an instance that isolates the family-level asymmetry underlying T1's aggregate curve in Appendix~\ref{sec:app_scale_full} (Figure~\ref{fig:scale_analysis}): across the Qwen3.5 (0.8B--35B-A3B) and Gemma4 (E2B--31B, in parentheses) parameter ladders under Non-Thinking mode. Qwen3.5 is genuinely scaling-resistant on this instance, landing on the same wrong value ($12$) at 5 of its 6 sizes regardless of scale; Gemma4 is scaling-responsive, flipping from wrong (E2B $=14$) to exactly correct (E4B $=11$, 31B $=11$), with a dip back to wrong at the intervening MoE checkpoint (26B-A4B $=13$) consistent with the paper's broader finding that MoE checkpoints do not reliably help. Aggregated over all 100 rows, Qwen3.5's 0.8B$\to$27B delta on T1 is only $+6.4$pp while Gemma4's E2B$\to$31B delta is $+26.8$pp, this instance is representative of the two families' opposite directions.}
\label{tab:qual_scaling_t1}\\
\end{longtable}
\endgroup

\begingroup
\small
\renewcommand{\arraystretch}{1.16}
\setlength{\tabcolsep}{2pt}
\begin{longtable}{>{\raggedright\arraybackslash}p{0.08\textwidth}*{7}{>{\centering\arraybackslash}p{0.12\textwidth}}}
\toprule
\endfirsthead
\toprule
\multicolumn{8}{l}{\tablename~\thetable{} continued.}\\
\midrule
\endhead
\multicolumn{8}{l}{\textbf{T3 Difference -- scaling-resistant}}\\
\promptlineSm{determine the difference in the two objects' block compositions}
\qualinputrowSm{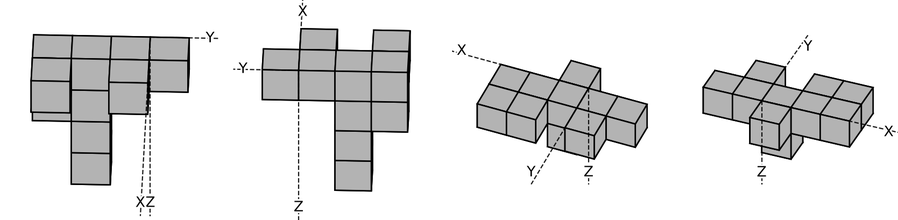}
\scalehdr
Qwen3.5 & \multirow{2}{*}{$(2,8,8)$} & $(1,1,1)$\tickno & $(10,10,10)$\tickno & $(10,2,2)$\tickno & $(6,4,4)$\tickno & $(10,4,2)$\tickno & $(10,2,2)$\tickno \\
Gemma4 & & -- & $(6,2,2)$\tickno & $(3,3,2)$\tickno & -- & $(10,2,0)$\tickno & $(7,3,2)$\tickno \\
\bottomrule
\caption{T3 Difference, scaling-resistant instance, across the Qwen3.5 (0.8B--35B-A3B) and Gemma4 (E2B--31B, in parentheses) parameter ladders under Non-Thinking mode. Every one of the 10 sizes misses the (intersection, only-A, only-B) triple, and there is no monotonic approach to GT $(2,8,8)$ as size grows -- e.g.\ Qwen3.5-27B $(10,4,2)$ is not closer than Qwen3.5-4B $(10,2,2)$ -- illustrating that scale does not recover the cross-object voxel-set alignment this task requires (Appendix~\ref{sec:app_scale_full}).}
\label{tab:qual_scaling_t3}\\
\end{longtable}
\endgroup

% T5 chip-grid macros (full-size chips, 3-col \threewrap, matching C1).
% Letter-only macro names (LaTeX disallows digits in \newcommand names).
% Qwen ladder A..F = 0.8B, 2B, 4B, 9B, 27B, 35B-A3B; Gemma G..J = E2B, E4B, 26B-A4B, 31B.
\newcommand{\tfiveGTscale}{\threewrap{\cdOr{2}&\cdO{2}&\cdP{1}\\\cdR{1}&\cdY{2}&\cdC{2}}}
\newcommand{\tfiveSA}{\threewrap{\cdM{1}&\cdR{1}&\cdO{1}\\\cdP{1}&\cdY{1}&\cdC{1}\\\cdT{1}&\cdGo{1}&\cdB{1}\\\cdS{1}&&}}
\newcommand{\tfiveSB}{\threewrap{\cdY{4}&\cdC{3}&\cdR{1}\\\cdO{4}&\cdP{2}&\cdM{3}\\\cdGo{2}&&}}
\newcommand{\tfiveSC}{\threewrap{\cdY{3}&\cdC{3}&\cdR{2}\\\cdO{4}&\cdOr{4}&\cdP{4}}}
\newcommand{\tfiveSD}{\threewrap{\cdY{3}&\cdC{3}&\cdM{4}\\\cdR{2}&\cdO{4}&\cdP{3}}}
\newcommand{\tfiveSE}{\threewrap{\cdY{3}&\cdC{3}&\cdM{4}\\\cdR{2}&\cdO{4}&\cdP{2}}}
\newcommand{\tfiveSF}{\threewrap{\cdY{3}&\cdC{3}&\cdO{4}\\\cdP{3}&\cdR{2}&\cdOr{3}}}
\newcommand{\tfiveSG}{\threewrap{\cdM{1}&\cdP{3}&\cdR{2}\\\cdO{3}&\cdC{3}&\cdY{2}}}
\newcommand{\tfiveSH}{\threewrap{\cdP{2}&\cdM{2}&\cdO{3}\\\cdR{2}&\cdY{3}&\cdC{3}\\\cdOr{1}&&}}
\newcommand{\tfiveSI}{\threewrap{\cdOr{3}&\cdP{3}&\cdO{3}\\\cdR{3}&\cdC{3}&\cdY{3}}}
\newcommand{\tfiveSJ}{\threewrap{\cdOr{3}&\cdP{3}&\cdO{4}\\\cdR{3}&\cdC{3}&\cdY{2}}}

\begingroup
\small
\renewcommand{\arraystretch}{1.16}
\setlength{\tabcolsep}{2pt}
\begin{longtable}{>{\raggedright\arraybackslash}p{0.08\textwidth}*{7}{>{\centering\arraybackslash}p{0.12\textwidth}}}
\toprule
\endfirsthead
\toprule
\multicolumn{8}{l}{\tablename~\thetable{} continued.}\\
\midrule
\endhead
\multicolumn{8}{l}{\textbf{T5 Color Distribution -- scaling-resistant}}\\
\promptlineSm{give the color distribution in the object}
\qualinputrowSm{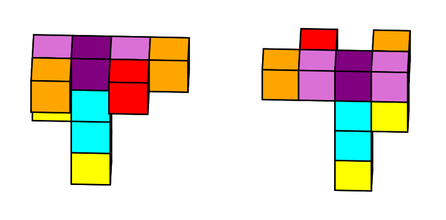}
\scalehdr
Qwen3.5 & \multirow{2}{*}{\tfiveGTscale} & \tfiveSA\tickno & \tfiveSB\tickno & \tfiveSC\tickno & \tfiveSD\tickno & \tfiveSE\tickno & \tfiveSF\tickno \\
Gemma4 & & -- & \tfiveSG\tickno & \tfiveSH\tickno & -- & \tfiveSI\tickno & \tfiveSJ\tickno \\
\bottomrule
\caption{T5 Color Distribution, scaling-resistant instance, across the Qwen3.5 (0.8B--35B-A3B) and Gemma4 (E2B--31B, in parentheses) parameter ladders under Non-Thinking mode. Every model size in both families misassigns the per-color block counts, including the two largest checkpoints per family that solve T4, illustrating that added scale does not recover exact per-color counts (Appendix~\ref{sec:app_scale_full}).}
\label{tab:qual_scaling_t5}\\
\end{longtable}
\endgroup

\begingroup
\small
\renewcommand{\arraystretch}{1.16}
\setlength{\tabcolsep}{2pt}
\begin{longtable}{>{\raggedright\arraybackslash}p{0.08\textwidth}*{7}{>{\centering\arraybackslash}p{0.12\textwidth}}}
\toprule
\endfirsthead
\toprule
\multicolumn{8}{l}{\tablename~\thetable{} continued.}\\
\midrule
\endhead
\multicolumn{8}{l}{\textbf{T8 Silhouette Symmetry -- scaling-resistant}}\\
\promptlineSm{determine whether the object's silhouette is symmetric (vertically or horizontally)}
\qualinputrowSm{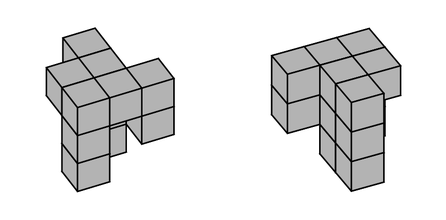}
\scalehdr
Qwen3.5 & \multirow{2}{*}{True} & False\tickno & False\tickno & False\tickno & False\tickno & False\tickno & False\tickno \\
Gemma4 & & -- & False\tickno & False\tickno & -- & False\tickno & False\tickno \\
\bottomrule
\caption{T8 Silhouette Symmetry, scaling-resistant instance, across the Qwen3.5 (0.8B--35B-A3B) and Gemma4 (E2B--31B, in parentheses) parameter ladders under Non-Thinking mode. All 10 sizes in both families answer False on a silhouette that is in fact symmetric (GT True), never once pairing the mirrored columns needed to detect it; the modal answer does not shift with scale, illustrating a scale-invariant default-to-False response under uncertainty at every point on the ladder (Appendix~\ref{sec:app_scale_full}).}
\label{tab:qual_scaling_t8}\\
\end{longtable}
\endgroup

\begingroup
\small
\renewcommand{\arraystretch}{1.16}
\setlength{\tabcolsep}{2pt}
\begin{longtable}{>{\raggedright\arraybackslash}p{0.08\textwidth}*{7}{>{\centering\arraybackslash}p{0.12\textwidth}}}
\toprule
\endfirsthead
\toprule
\multicolumn{8}{l}{\tablename~\thetable{} continued.}\\
\midrule
\endhead
\multicolumn{8}{l}{\textbf{T10 Color Connected -- scaling-resistant}}\\
\promptlineSm{determine whether there exists at least one pair of adjacent blocks (sharing a face) that share the same color}
\qualinputrowSm{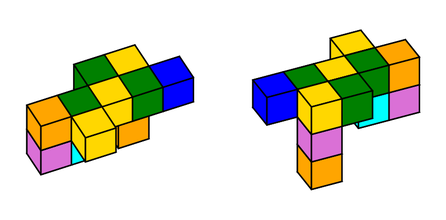}
\scalehdr
Qwen3.5 & \multirow{2}{*}{False} & True\tickno & True\tickno & True\tickno & True\tickno & True\tickno & True\tickno \\
Gemma4 & & -- & False\tickok & False\tickok & -- & False\tickok & True\tickno \\
\bottomrule
\caption{T10 Color Connected, scaling-resistant instance, across the Qwen3.5 (0.8B--35B-A3B) and Gemma4 (E2B--31B, in parentheses) parameter ladders under Non-Thinking mode. The entire Qwen3.5 ladder answers True (wrong) at every size from 0.8B to 35B-A3B, showing no scaling trend; Gemma4 is correct at E2B/E4B/26B-A4B but regresses back to the wrong answer at its largest checkpoint (31B). Neither family shows the monotonic small-to-large convergence seen for the responsive tasks above, illustrating T10's flat aggregate curve in Appendix~\ref{sec:app_scale_full} (Figure~\ref{fig:scale_analysis}).}
\label{tab:qual_scaling_t10}\\
\end{longtable}
\endgroup

\begingroup
\small
\renewcommand{\arraystretch}{1.16}
\setlength{\tabcolsep}{2pt}
\begin{longtable}{>{\raggedright\arraybackslash}p{0.08\textwidth}*{7}{>{\centering\arraybackslash}p{0.12\textwidth}}}
\toprule
\endfirsthead
\toprule
\multicolumn{8}{l}{\tablename~\thetable{} continued.}\\
\midrule
\endhead
\multicolumn{8}{l}{\textbf{T12 Reconstruct -- scaling-resistant}}\\
\promptlineSm{identify the positive view among the two candidates}
\qualinputrowSm{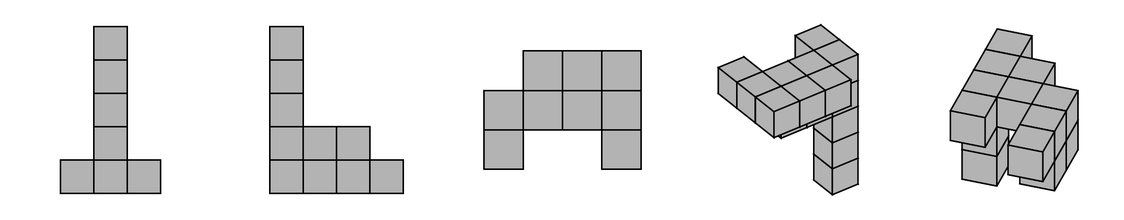}
\scalehdr
Qwen3.5 & \multirow{2}{*}{0} & 1\tickno & 1\tickno & 1\tickno & 1\tickno & 1\tickno & 1\tickno \\
Gemma4 & & -- & 1\tickno & 1\tickno & -- & 1\tickno & 1\tickno \\
\bottomrule
\caption{T12 Reconstruct, scaling-resistant instance, across the Qwen3.5 (0.8B--35B-A3B) and Gemma4 (E2B--31B, in parentheses) parameter ladders under Non-Thinking mode. All 10 sizes across both families collapse onto the same wrong candidate (1), never selecting the correct 0: added scale changes nothing about which cross-view silhouette is picked.}
\label{tab:qual_scaling_t12}\\
\end{longtable}
\endgroup

\begingroup
\small
\renewcommand{\arraystretch}{1.16}
\setlength{\tabcolsep}{2pt}
\begin{longtable}{>{\raggedright\arraybackslash}p{0.08\textwidth}*{7}{>{\centering\arraybackslash}p{0.12\textwidth}}}
\toprule
\endfirsthead
\toprule
\multicolumn{8}{l}{\tablename~\thetable{} continued.}\\
\midrule
\endhead
\multicolumn{8}{l}{\textbf{T13 Mixture-of-Shapes -- scaling-resistant}}\\
\promptlineSm{identify the invalid silhouette among the three candidates}
\qualinputrowSm{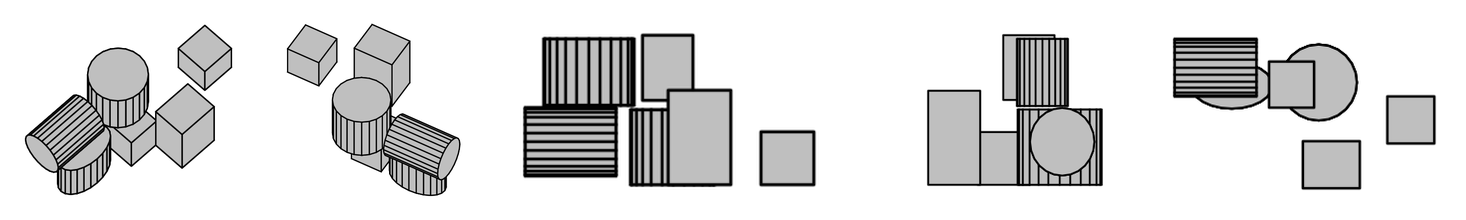}
\scalehdr
Qwen3.5 & \multirow{2}{*}{2} & 2\tickok & 2\tickok & 2\tickok & 2\tickok & 2\tickok & 2\tickok \\
Gemma4 & & -- & 2\tickok & 2\tickok & -- & 2\tickok & 2\tickok \\
\bottomrule
\caption{T13 Mixture-of-Shapes, scaling-resistant instance, across the Qwen3.5 (0.8B--35B-A3B) and Gemma4 (E2B--31B, in parentheses) parameter ladders under Non-Thinking mode. All 10 sizes across both families collapse onto the same correct candidate (2). This example shows agreement across scales; it is not an instance of persistent failure. The aggregate scaling trends are reported in Appendix~\ref{sec:app_scale_full}.}
\label{tab:qual_scaling_t13}\\
\end{longtable}
\endgroup

\twocolumn
\subsection{Thinking Budget: Instance-Level Evidence}
\label{sec:app_qual_thinking_budget}

These panels give the instance-level qualitative counterpart to the budget sweep in Table~\ref{tab:thinking_scaling} (Appendix~\ref{sec:app_thinking_scaling}), covering the three tasks that sweep spans: \tonelong{} and \televenlong{}, where added reasoning tokens reduce accuracy, and \tfourlong{}, where they raise it.

\onecolumn
% Auto-generated by scripts/gen_thinking_budget_section.py -- do not edit by hand.
\newcommand{\budgetfull}{\dimexpr\textwidth-4pt\relax}
\newcommand{\budgetpromptline}[1]{\multicolumn{6}{@{}l@{}}{\parbox{\budgetfull}{\itshape ``\ldots{}#1\ldots{}''}}\\[-0.2mm]}
\newcommand{\budgetinputrow}[1]{\multicolumn{6}{@{}c@{}}{\includegraphics[width=\budgetfull,height=1.25in,keepaspectratio]{#1}}\\[-0.5mm]}
\newcommand{\budgetmodelhdr}{\textbf{Model} & \textbf{GT} & \textbf{NT@$0$} & \textbf{TH@$1024$} & \textbf{TH@$2048$} & \textbf{TH@$4096$}\\ \addlinespace[1pt]}
% \\[x] would set the ROW DEPTH to x, so a tall paragraph would eat the gap and
% short quotes would look further apart. \addlinespace is a true \noalign skip
% between rows, so every gap is the same regardless of quote length.
\newcommand{\budgettrace}[3]{\multicolumn{6}{@{}p{\budgetfull}@{}}{\footnotesize\textbf{#1, TH@#2:} #3}\\ \addlinespace[4mm]}
\newcommand{\budgettracelast}[3]{\multicolumn{6}{@{}p{\budgetfull}@{}}{\footnotesize\textbf{#1, TH@#2:} #3}\\ \addlinespace[1mm]}

\begin{table}[H]
\centering
\begingroup
\small
\renewcommand{\arraystretch}{1.16}
\setlength{\tabcolsep}{2pt}
\begin{tabular}{>{\centering\arraybackslash}m{0.12\textwidth}>{\centering\arraybackslash}m{0.08\textwidth}*{4}{>{\centering\arraybackslash}m{\dimexpr0.2\textwidth-6pt\relax}}}
\toprule
\multicolumn{6}{@{}l@{}}{\textbf{T1 Volume} \hfill \textit{budget verdict: correct at budget $0$, drifts wrong as the budget grows}}\\
\budgetpromptline{count the total number of blocks that make up the 3D object}
\budgetinputrow{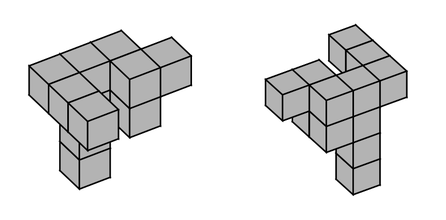}
\midrule
\budgetmodelhdr
\micon{qwen_icon.png} & \multirow{2}{*}{$12$} & $12$\tickok & $12$\tickok & $13$\tickno & $9$\tickno \\
\micon{gemma_icon.png} &  & $12$\tickok & $14$\tickno & $14$\tickno & $14$\tickno \\
\addlinespace[2pt]
\budgettrace{Qwen3.5-27B}{1024}{[\ldots] Top block. Let's re-examine Image 1. In Image 1, the front column has: Top block (part of the main cross). Middle block. Bottom block. So the column is 3 blocks high below the main cross? Or is the top block part of the cross? Let's look at the junction. In Image 1, the front-most block of the main horizontal layer is connected to a block below it. So, there is \textit{[trace cut off at the budget cap]}}
\budgettrace{Qwen3.5-27B}{2048}{[\ldots] Behind the central junction: A block. This is getting complicated. Let's try to map the blocks directly. Let's assume the structure is a ``plus'' sign with a tail. Let's look at Image 2 again. It looks like a standard ``plus'' sign on top, with a vertical stem. Top layer: Center block. Left arm: 2 blocks. Right arm: 2 blocks. Back arm: 1 block. Front arm: This is where the vertical stem is. So the ``front'' block of the \textit{[trace cut off at the budget cap]}}
\budgettrace{Qwen3.5-27B}{4096}{[\ldots] To its ``back'' (up-right in image), there is a block. This could be the center block? No, the center block is to the left. Let's re-orient. In Image 2, the column is on the right. To the left of the column top is a block. Let's call this C1. To the left of C1 is C2. To the left of C2 is C3. So we have a row of 4 blocks? Column top + 3 to the left. This would mean the ``left'' arm is very long. Let's look at Image 1 again. \textit{[trace cut off at the budget cap]}}
\budgettrace{Gemma4-31B}{1024}{[\ldots] Back row: 3 blocks. Middle row: 2 blocks. Wait, let's look at View 1 again. Top surface: Back row: 3 blocks. Middle row: 2 blocks. Wait, the rightmost block in the back row is (2,0,2). \textit{[trace cut off at the budget cap]}}
\budgettrace{Gemma4-31B}{2048}{[\ldots] Let me re-count the top faces in Image 1. 1 (left), 2 (mid-left), 3 (mid-right), 4 (right), 5 (back-mid-right), 6 (back-right). Yes, 6 blocks on the top layer. Middle layer (z=1): \textit{[trace cut off at the budget cap]}}
\budgettracelast{Gemma4-31B}{4096}{[\ldots] They are under (1,0,1) and (2,0,1). So (1,0,0) and (2,0,0). Let's check Image 2. In Image 2, we see the bottom blocks. They are under (1,0,1) and (2,0,1). Yes, (1,0,0) and (2,0,0). Wait, let me re-examine the middle layer in Image 2. In Image 2, we see: (0,0,1) - leftmost. (1,0,1) - next. (2,0,1) \textit{[trace cut off at the budget cap]}}
\bottomrule
\end{tabular}
\endgroup
\caption{\tonelong{}, thinking-budget qualitative panel. Both models answer correctly without thinking and neither recovers the count once the budget grows: Qwen3.5-27B moves $12 \rightarrow 12 \rightarrow 13 \rightarrow 9$ and Gemma4-31B settles on $14$ from $1024$ tokens onward. The quoted traces show the budget spent re-orienting the two views and relabeling arms rather than fixing a block lattice. Budgets run left to right: $0$ (non-thinking), $1024$, $2048$, and $4096$ reasoning tokens; each cell is the 5-run majority vote of that model at that budget on the single default-setting instance shown.}
\label{tab:qual_budget_t1}
\end{table}

\begin{table}[H]
\centering
\begingroup
\small
\renewcommand{\arraystretch}{1.16}
\setlength{\tabcolsep}{2pt}
\begin{tabular}{>{\centering\arraybackslash}m{0.12\textwidth}>{\centering\arraybackslash}m{0.08\textwidth}*{4}{>{\centering\arraybackslash}m{\dimexpr0.2\textwidth-6pt\relax}}}
\toprule
\multicolumn{6}{@{}l@{}}{\textbf{T4 Color Count} \hfill \textit{budget verdict: wrong at budget $0$, recovers the exact count at $4096$}}\\
\budgetpromptline{count the number of unique colors in the object}
\budgetinputrow{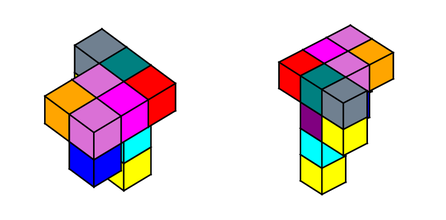}
\midrule
\budgetmodelhdr
\micon{qwen_icon.png} & \multirow{2}{*}{$10$} & $12$\tickno & $11$\tickno & $11$\tickno & $10$\tickok \\
\micon{gemma_icon.png} &  & $11$\tickno & $10$\tickok & $10$\tickok & $10$\tickok \\
\addlinespace[2pt]
\budgettrace{Qwen3.5-27B}{1024}{[\ldots] In front of Slategray: Orchid. In front of Teal: Magenta. To the right of Teal: Red. To the left of Orchid: Orange. This implies a 2x3 grid on top? Or a cross? Let's assume a central block. Let's look at the ``cross'' shape in View 1. Center: Magenta? No. Let's look at the connections. The block with the top face ``Orchid'' (light purple) is connected to ``Orange'' (left), ``Magenta \textit{[trace cut off at the budget cap]}}
\budgettrace{Qwen3.5-27B}{2048}{[\ldots] Wait, let's look at the colors again. View 1 Top Layer: Back Left: Slategray. Back Right: Teal. Middle Left: Orchid. Middle Right: Magenta. Front Left: Orange. Front Middle Left: Orchid. Front Middle Right: Magenta. Front Right: Red. This seems like too many blocks. Let's look at the connections. The block labeled ``Orchid'' in the middle (below Slategray) is connected to the block labeled ``Magenta'' (below Teal). \textit{[trace cut off at the budget cap]}}
\budgettrace{Qwen3.5-27B}{4096}{[\ldots] The block at (x=1, z=0) is Magenta. Below it is Cyan. So Cyan is at y=1. Below Cyan is Yellow. So Yellow is at y=0. So we have a stack at (x=1, z=0): Yellow (y=0), Cyan (y=1), Magenta (y=2). And a stack at (x=0, z=0): Blue (y=1), Orchid (y=2). Wait, in View 1, the front-most Orchid is at the same level as the Magenta block? Yes, they look adjacent. So at y=2, we have Orchid (front left) and Magenta \textit{[trace cut off at the budget cap]}}
\budgettrace{Gemma4-31B}{1024}{[\ldots] Combined list: 1. Slategray 2. Teal 3. Red 4. Orange 5. Orchid 6. Magenta 7. Cyan 8. Yellow 9. Blue 10. Purple Let's double-check the colors again. Image 1: Top: slategray, teal, red Middle: orange, orchid, magenta, red (side) Bottom: orchid (side), magenta (side), \textit{[trace cut off at the budget cap]}}
\budgettrace{Gemma4-31B}{2048}{[\ldots] 2. teal 3. red 4. orange 5. magenta 6. orchid 7. blue 8. cyan 9. yellow 10. purple Let me double-check the colors in Image 1 again. Top row: slategray, teal, red. Middle row: orange, magenta, red. Bottom row: orchid, magenta, red. Below orchid: blue. Below magenta: cyan. Below cyan: yellow. Colors: sl \textit{[trace cut off at the budget cap]}}
\budgettracelast{Gemma4-31B}{4096}{[\ldots] cyan (below magenta) yellow (below cyan) Image 2: orchid (top left) orange (top right) magenta (middle left) orchid (middle middle) orange (middle right) teal (below magenta) slategray (below orchid) purple (below teal) yellow (below slategray) cyan (below purple) \textit{[trace cut off at the budget cap]}}
\bottomrule
\end{tabular}
\endgroup
\caption{\tfourlong{}, thinking-budget qualitative panel. Both models start wrong and reach the exact count as the budget grows. The budget-$4096$ trace has committed to an explicit coordinate lattice with one color assigned per cell, which is the grounded intermediate the count is read off; at $1024$ the same model is still debating whether the top face is a $2\times3$ grid or a cross. Budgets run left to right: $0$ (non-thinking), $1024$, $2048$, and $4096$ reasoning tokens; each cell is the 5-run majority vote of that model at that budget on the single default-setting instance shown.}
\label{tab:qual_budget_t4}
\end{table}

\begin{table}[H]
\centering
\begingroup
\small
\renewcommand{\arraystretch}{1.16}
\setlength{\tabcolsep}{2pt}
\begin{tabular}{>{\centering\arraybackslash}m{0.12\textwidth}>{\centering\arraybackslash}m{0.08\textwidth}*{4}{>{\centering\arraybackslash}m{\dimexpr0.2\textwidth-6pt\relax}}}
\toprule
\multicolumn{6}{@{}l@{}}{\textbf{T11 Projection} \hfill \textit{budget verdict: Qwen drifts wrong at $4096$; Gemma recovers at $2048$}}\\
\budgetpromptline{identify the invalid projection}
\budgetinputrow{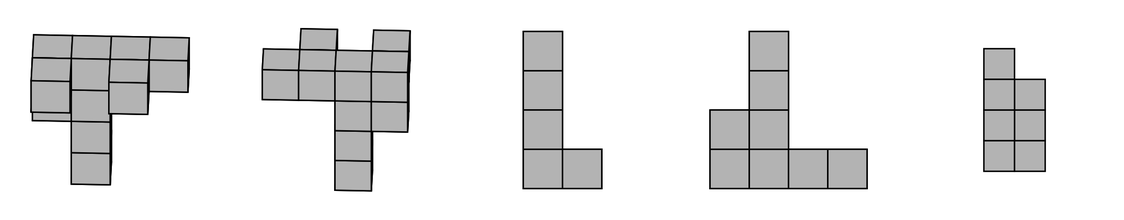}
\midrule
\budgetmodelhdr
\micon{qwen_icon.png} & \multirow{2}{*}{$2$} & $2$\tickok & $2$\tickok & $2$\tickok & $1$\tickno \\
\micon{gemma_icon.png} &  & $0$\tickno & $0$\tickno & $2$\tickok & $2$\tickok \\
\addlinespace[2pt]
\budgettrace{Qwen3.5-27B}{1024}{[\ldots] Fifth layer down: 1 block (col 2). Wait, the stem is clearly visible. Let's count the vertical stack in the stem. In View 1, the stem is the second column. It has blocks at levels 3, 4, 5 (counting from top). So, column 2 has height 5. Column 1 has height 3. Column 3 has height 3. Column 4 has height 2. Let's check this against View 1. Col 1: 3 blocks high \textit{[trace cut off at the budget cap]}}
\budgettrace{Qwen3.5-27B}{2048}{[\ldots] 4th row: 1 block (col 3). 5th row: 1 block (col 3). This doesn't match the ``heights'' I derived from View 1 if View 2 is a side view. Let's reconsider the 3D structure. Maybe it's not just columns of varying height. Maybe there are blocks in the depth (z-axis). Let's assume the object is a collection of cubes. Let's try to interpret View 1 as a Front View. We see a grid. \textit{[trace cut off at the budget cap]}}
\budgettrace{Qwen3.5-27B}{4096}{[\ldots] This implies the object is 1 unit deep. Let's check if the object is 1 unit deep. View 1 and View 2 look like isometric or perspective views of a 2D shape extruded? No, View 2 shows depth. In View 2, we see blocks ``behind'' the front face. For example, in View 2, top row has blocks at different depths? Let's look at View 2 again. It looks like a ``T'' shape but with blocks missing. Actually, View 2 looks like the Top View (XY) rotated? \textit{[trace cut off at the budget cap]}}
\budgettrace{Gemma4-31B}{1024}{[\ldots] x: 1 to 4 (columns) y: 1 to 6 (rows, 6 is top) z: depth (1 to ?) Front view (x, y): y=6: x=1, 2, 3, 4 y=5: x=1, 2, 3, 4 y=4: x=1, 2, 3 y=3: x=2 y=2: x=2 y=1: x=2 View 2 (let's call it Side View): This view is from a different angle. Let's see if it's a 90-degree rotation. If View 1 is (x, y), View 2 \textit{[trace cut off at the budget cap]}}
\budgettrace{Gemma4-31B}{2048}{[\ldots] Row 6: Blocks at (1,6), (2,6), (3,6), (4,6) Row 5: Blocks at (1,5), (2,5), (3,5), (4,5) Row 4: Blocks at (3,4), (4,4) Row 3: Block at (3,3) Row 2: Block at (3,2) Row 1: Block at (3,1) Wait, this is getting confusing. Let me look at the images again. They are very simple. Let's try to see them as 3D objects. Image 1: A 4x2 rectangle on top. A 2x3 \textit{[trace cut off at the budget cap]}}
\budgettracelast{Gemma4-31B}{4096}{[\ldots] Top face: 4x2 rectangle. View 2: Front face: Row 1: 1 block (col 2) Row 2: 4 blocks (cols 1, 2, 3, 4) Row 3: 4 blocks (cols 1, 2, 3, 4) Row 4: 2 blocks (cols 3, 4) Row 5: 1 block (col 3) Row 6: 1 block (col 3) Top face: It's a 2D shape. Row 1: 1 block (col 2) Row 2: 4 blocks (cols 1, 2, \textit{[trace cut off at the budget cap]}}
\bottomrule
\end{tabular}
\endgroup
\caption{\televenlong{}, thinking-budget qualitative panel. The two models move in opposite directions: Qwen3.5-27B holds the correct projection through $2048$ tokens and drifts wrong at $4096$, while Gemma4-31B is wrong at $0$ and $1024$ and correct from $2048$ onward. The instance-level pattern therefore matches the model average for Qwen3.5-27B only. Budgets run left to right: $0$ (non-thinking), $1024$, $2048$, and $4096$ reasoning tokens; each cell is the 5-run majority vote of that model at that budget on the single default-setting instance shown.}
\label{tab:qual_budget_t11}
\end{table}

\twocolumn
\subsection{Thinking Mode Flips: Instance-Level Evidence}
\label{sec:app_qual_thinking_flip}

These panels give the instance-level qualitative counterpart to the per-task thinking deltas in Table~\ref{tab:thinking_delta_full} (Appendix~\ref{sec:app_thinking}). Reading a column top to bottom shows what thinking changed. 
% The two largest gains are \ttwolong{} and \tfourlong{}: on \ttwolong{} (Table~\ref{tab:qual_flip_t2}) thinking reduces the object to a bounding-box triple and compares it against the boundary, a check that can be carried out in words, and all four models flip from wrong to correct; on \tfourlong{} (Table~\ref{tab:qual_flip_t4}) thinking walks the block faces and accumulates a running color list, and three of the four flip. These are the two mechanisms the main text calls a short verbalizable check and enumeration. Tasks whose answers need an exact structured readout show the opposite pattern: on \tthreelong{} (Table~\ref{tab:qual_flip_t3}) no model changes its answer and all eight cells stay wrong.

\onecolumn
% Auto-generated by scripts/gen_thinking_flip_section.py -- do not edit by hand.
\newcommand{\flipfull}{\dimexpr\textwidth-4pt\relax}
\newcommand{\flippromptline}[1]{\multicolumn{6}{@{}l@{}}{\parbox{\flipfull}{\itshape ``\ldots{}#1\ldots{}''}}\\[-0.2mm]}
\newcommand{\flipinputrow}[1]{\multicolumn{6}{@{}c@{}}{\includegraphics[width=\flipfull,height=1.25in,keepaspectratio]{#1}}\\[-0.5mm]}
\newcommand{\flipmodelhdr}{\textbf{Mode} & \textbf{GT} & \micon{qwen_icon.png} & \micon{gemma_icon.png} & \micon{glmv_icon.png} & \micon{kimi_icon.png}\\ \addlinespace[1pt]}
% \\[x] would set the ROW DEPTH to x, so a tall paragraph would eat the gap and
% short quotes would look further apart. \addlinespace is a true \noalign skip
% between rows, so every gap is the same regardless of quote length.
\newcommand{\fliptrace}[2]{\multicolumn{6}{@{}p{\flipfull}@{}}{\footnotesize\textbf{#1 (TH):} #2}\\ \addlinespace[4mm]}
\newcommand{\fliptracelast}[2]{\multicolumn{6}{@{}p{\flipfull}@{}}{\footnotesize\textbf{#1 (TH):} #2}\\ \addlinespace[1mm]}

\begingroup
\small
\renewcommand{\arraystretch}{1.16}
\setlength{\tabcolsep}{2pt}
\begin{longtable}{>{\centering\arraybackslash}m{\dimexpr0.10\textwidth-4pt\relax}*{5}{>{\centering\arraybackslash}m{\dimexpr0.18\textwidth-4pt\relax}}}
\toprule
\endfirsthead
\toprule
\multicolumn{6}{l}{\tablename~\thetable{} continued.}\\
\midrule
\endhead
\multicolumn{6}{@{}l@{}}{\textbf{T1 Volume} \hfill \textit{mode verdict: NT wins}}\\
\flippromptline{count the total number of blocks that make up the 3D object}
\flipinputrow{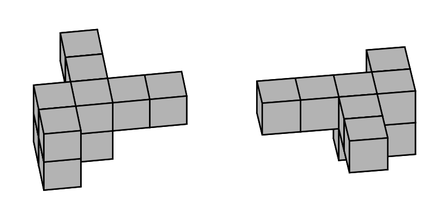}
\midrule
\flipmodelhdr
NT & \multirow{2}{*}{$10$} & $10$\tickok & $10$\tickok & $12$\tickno & $10$\tickok \\
TH & & $11$\tickno & $11$\tickno & $11$\tickno & $20$\tickno \\
\addlinespace[2pt]
\fliptrace{Qwen3.5-27B}{[\ldots] Actually, let's look for the longest continuous lines. Hypothesis 1: There is a central vertical column. There is a horizontal row extending to the right. There is a horizontal row extending to the left. There is a vertical column extending up. Let's look at Image 1 again carefully. Leftmost part: A vertical stack of 3 blocks. Let's call this Column A. Connected to the middle block of Column A is a block to its right. Let's call this Block B. Above Block \textit{[trace cut off at the budget cap]}}
\fliptrace{Gemma4-31B}{[\ldots] Right: (1,0,1), (2,0,1) -\textgreater{} 2 blocks. Left: (-1,0,1), (-1,0,0), (-1,0,-1) -\textgreater{} 3 blocks. Front: (0,1,1), (0,1,0) -\textgreater{} 2 blocks. Total = 3 + 2 + 3 + 2 = 10? Let me check Image 2. Image 2: Looking from a different angle. There's a long arm extending to the left: (-1, 0, 1), (0, 0, 1), (1, 0, 1), (2, \textit{[trace cut off at the budget cap]}}
\fliptracelast{GLM-4.6V-Flash}{[\ldots] Looking at the first image: Left column: 3 blocks (vertical) Central column (front): 2 blocks (vertical) Central column (back? No, the top vertical stack is on the central column, so central column has 3 blocks (bottom, middle, top)? Wait, no, the first \textit{[trace cut off at the budget cap]}}
\bottomrule
\caption{\tonelong{}, thinking-mode qualitative panel. Non-Thinking answers $3$ of $4$ correctly and Thinking $0$ of $4$; $3$ models flip correct to wrong. The task-level verdict is \emph{NT wins}. Each cell is the 5-run majority vote of that model on the single default-setting instance shown; NT is Non-Thinking and TH is Thinking at the default $1024$-token budget. Beneath the outcome rows we quote the tail of each model's TH trace. Kimi-VL-A3B surfaces no reasoning tokens, so it has no quote.}
\label{tab:qual_flip_t1}\\
\end{longtable}
\endgroup

\begingroup
\small
\renewcommand{\arraystretch}{1.16}
\setlength{\tabcolsep}{2pt}
\begin{longtable}{>{\centering\arraybackslash}m{\dimexpr0.10\textwidth-4pt\relax}*{5}{>{\centering\arraybackslash}m{\dimexpr0.18\textwidth-4pt\relax}}}
\toprule
\endfirsthead
\toprule
\multicolumn{6}{l}{\tablename~\thetable{} continued.}\\
\midrule
\endhead
\multicolumn{6}{@{}l@{}}{\textbf{T2 Packaging} \hfill \textit{mode verdict: TH wins}}\\
\flippromptline{determine whether the object can fit in the boundary in any possible orientation}
\flipinputrow{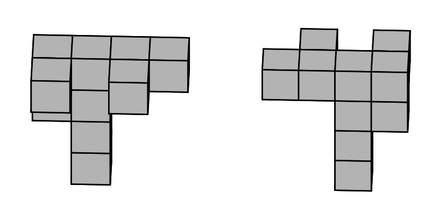}
\midrule
\flipmodelhdr
NT & \multirow{2}{*}{False} & True\tickno & True\tickno & True\tickno & True\tickno \\
TH & & False\tickok & False\tickok & False\tickok & False\tickok \\
\addlinespace[2pt]
\fliptrace{Qwen3.5-27B}{[\ldots] It shows a vertical stem going down. Let's count the width again. 1, 2, 3, 4 blocks wide. Let's count the height. The vertical stem has 3 blocks below the main horizontal section. The main section is 2 blocks high? Let's look at the grid. Top row: 4 blocks. Second row: 4 blocks. Third row: 2 blocks (under the 3rd and 4th columns). Fourth row: 1 block (under the 3rd column). Fifth row \textit{[trace cut off at the budget cap]}}
\fliptrace{Gemma4-31B}{[\ldots] Third row: 4 cubes. Fourth row: 2 cubes (at positions 3 and 4). Fifth row: 1 cube (at position 3). Sixth row: 1 cube (at position 3). Dimensions: 4 wide, 6 high, 1 deep. Wait, the prompt says ``two perspective images of the same 3D object from different viewing angles''. Let's re-examine Image 1 and Image 2. They are not just 2D projections; they are 3D perspective views. Image 1: \textit{[trace cut off at the budget cap]}}
\fliptracelast{GLM-4.6V-Flash}{[\ldots] Row 5: Col2: square Col3: square Wait no, let's parse the second image: Top part: A square at (row1, col3) A square at (row1, col4) Then row2: (row2, col1) (row2, col2) (row2, col3) (row2, col4) Row3: (row3, col1) (row3, col2) (row3, col3) (row3, col4) Row4: (row4, col2) (row4, col3) Row5: (row5, col2) (row5, col3) Wait maybe better to count total cubes. Let's see: Top two squares (row1, col3 and col4) $\rightarrow$ 2 Row2: 4 $\rightarrow$ total 6 Row \textit{[trace cut off at the budget cap]}}
\bottomrule
\caption{\ttwolong{}, thinking-mode qualitative panel. Non-Thinking answers $0$ of $4$ correctly and Thinking $4$ of $4$; $4$ models flip wrong to correct. The task-level verdict is \emph{TH wins}. Each cell is the 5-run majority vote of that model on the single default-setting instance shown; NT is Non-Thinking and TH is Thinking at the default $1024$-token budget. Beneath the outcome rows we quote the tail of each model's TH trace. Kimi-VL-A3B surfaces no reasoning tokens, so it has no quote.}
\label{tab:qual_flip_t2}\\
\end{longtable}
\endgroup

\begingroup
\small
\renewcommand{\arraystretch}{1.16}
\setlength{\tabcolsep}{2pt}
\begin{longtable}{>{\centering\arraybackslash}m{\dimexpr0.10\textwidth-4pt\relax}*{5}{>{\centering\arraybackslash}m{\dimexpr0.18\textwidth-4pt\relax}}}
\toprule
\endfirsthead
\toprule
\multicolumn{6}{l}{\tablename~\thetable{} continued.}\\
\midrule
\endhead
\multicolumn{6}{@{}l@{}}{\textbf{T3 Difference} \hfill \textit{mode verdict: tie ($\Delta\approx 0$, both wrong)}}\\
\flippromptline{analyze both objects and determine the difference in their block compositions}
\flipinputrow{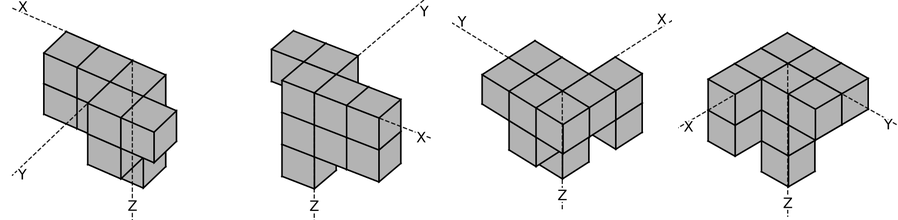}
\midrule
\flipmodelhdr
NT & \multirow{2}{*}{$(5,5,6)$} & $(10,2,2)$\tickno & $(6,3,3)$\tickno & $(8,3,3)$\tickno & $(2,4,5)$\tickno \\
TH & & $(6,2,2)$\tickno & $(7,2,1)$\tickno & $(0,8,10)$\tickno & $(4,4,4)$\tickno \\
\addlinespace[2pt]
\fliptrace{Qwen3.5-27B}{[\ldots] Let's count top faces. There are 4 top faces visible in a 2x2 arrangement? No. Let's look at the grid. Top row (back): 2 blocks. Middle row: 2 blocks. Front row: 1 block? Let's look at the vertical columns. Back-left column: 2 blocks high. Back-right column: 2 blocks high. Front-left column: 2 blocks high. Front-right column: \textit{[trace cut off at the budget cap]}}
\fliptrace{Gemma4-31B}{[\ldots] Top layer: 3 blocks. Middle layer: 3 blocks (under top) + 1 block to the right. Bottom layer: 2 blocks (under the middle and right blocks of the middle layer). Let's re-examine the bottom layer. It looks like 2 blocks. Total = 3 + 4 + 2 = 9. Image 3: Top layer: 3 blocks in an L-shape? No, 3 blocks in a row, and one more block to the right. Wait, let's look \textit{[trace cut off at the budget cap]}}
\fliptracelast{GLM-4.6V-Flash}{[\ldots] Wait, maybe the first two images (Image 1 and 2) are Object A, and the last two (3 and 4) are Object B. Let's check Image 3 and 4. Image 3: Top layer: X=1, Y=1; X=2, Y=1; X=3, Y=1 Middle layer: X=1, Y=1; X=2, Y=1; X=3, Y=1 Bottom layer: X=1, Y=1; X=2, Y=1; X=3, Y=1; X=3, Y=2? No, Image 3 has a different shape. Wait, let's count blocks for each image: Image 1: Let's count each cube: 1. Top-left (X1,Y1,Z1) 2. Top-middle (X2,Y1,Z1) 3. Top-right (X3,Y \textit{[trace cut off at the budget cap]}}
\bottomrule
\caption{\tthreelong{}, thinking-mode qualitative panel. Non-Thinking answers $0$ of $4$ correctly and Thinking $0$ of $4$; no model changes its answer. The task-level verdict is \emph{tie ($\Delta\approx 0$, both wrong)}. Each cell is the 5-run majority vote of that model on the single default-setting instance shown; NT is Non-Thinking and TH is Thinking at the default $1024$-token budget. Beneath the outcome rows we quote the tail of each model's TH trace. Kimi-VL-A3B surfaces no reasoning tokens, so it has no quote.}
\label{tab:qual_flip_t3}\\
\end{longtable}
\endgroup

\begingroup
\small
\renewcommand{\arraystretch}{1.16}
\setlength{\tabcolsep}{2pt}
\begin{longtable}{>{\centering\arraybackslash}m{\dimexpr0.10\textwidth-4pt\relax}*{5}{>{\centering\arraybackslash}m{\dimexpr0.18\textwidth-4pt\relax}}}
\toprule
\endfirsthead
\toprule
\multicolumn{6}{l}{\tablename~\thetable{} continued.}\\
\midrule
\endhead
\multicolumn{6}{@{}l@{}}{\textbf{T4 Color Count} \hfill \textit{mode verdict: TH wins}}\\
\flippromptline{count the number of unique colors in the object}
\flipinputrow{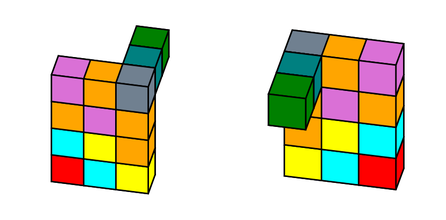}
\midrule
\flipmodelhdr
NT & \multirow{2}{*}{$8$} & $9$\tickno & $9$\tickno & $10$\tickno & $10$\tickno \\
TH & & $8$\tickok & $8$\tickok & $8$\tickok & $10$\tickno \\
\addlinespace[2pt]
\fliptrace{Qwen3.5-27B}{[\ldots] Let's look at the main block face in Image 2. Rightmost column: Red (bottom), Cyan (2nd), Orange (3rd), Orchid (top). This matches the leftmost column of Image 1 (Red, Cyan, Orange, Orchid). Middle column: Cyan (bottom), Yellow (2nd), Orchid (3rd), Orange (top). This matches the middle column of Image 1 (Cyan, Yellow, Orchid, Orange). Leftmost column (of the main block part): Yellow \textit{[trace cut off at the budget cap]}}
\fliptrace{Gemma4-31B}{[\ldots] Let's re-examine the tower again: green, teal, slategray. Wait, the slategray block is the top block of the right column. Let's re-examine the tower again: green, teal, slategray. Wait, the slategray block is the top block of the right column. Let's re-examine the tower again \textit{[trace cut off at the budget cap]}}
\fliptracelast{GLM-4.6V-Flash}{[\ldots] From first image: purple, orange, gray (slategray), teal, green, cyan, yellow, red. From second image: gray (slategray), teal, green, orange, purple, yellow, cyan, red, plus the green block (green). So all these colors are in the predefined set. Now let's list them: 1. Yellow 2. Cyan 3 \textit{[trace cut off at the budget cap]}}
\bottomrule
\caption{\tfourlong{}, thinking-mode qualitative panel. Non-Thinking answers $0$ of $4$ correctly and Thinking $3$ of $4$; $3$ models flip wrong to correct. The task-level verdict is \emph{TH wins}. Each cell is the 5-run majority vote of that model on the single default-setting instance shown; NT is Non-Thinking and TH is Thinking at the default $1024$-token budget. Beneath the outcome rows we quote the tail of each model's TH trace. Kimi-VL-A3B surfaces no reasoning tokens, so it has no quote.}
\label{tab:qual_flip_t4}\\
\end{longtable}
\endgroup

\begingroup
\small
\renewcommand{\arraystretch}{1.16}
\setlength{\tabcolsep}{2pt}
\begin{longtable}{>{\centering\arraybackslash}m{\dimexpr0.10\textwidth-4pt\relax}*{5}{>{\centering\arraybackslash}m{\dimexpr0.18\textwidth-4pt\relax}}}
\toprule
\endfirsthead
\toprule
\multicolumn{6}{l}{\tablename~\thetable{} continued.}\\
\midrule
\endhead
\multicolumn{6}{@{}l@{}}{\textbf{T5 Color Distribution} \hfill \textit{mode verdict: tie (slight TH lead)}}\\
\flippromptline{give the color distribution in the object}
\flipinputrow{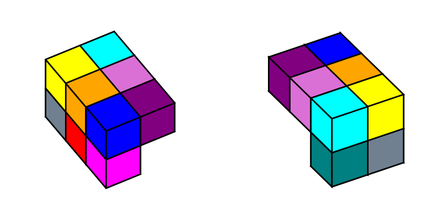}
\midrule
\flipmodelhdr
NT & \multirow{2}{*}{\fourwrap{\cdC{1}&\cdY{1}&\cdOr{1}&\cdO{1}\\\cdP{1}&\cdB{1}&\cdT{1}&\cdS{1}\\\cdR{1}&\cdM{1}&&}} & \fourwrap{\cdC{2}&\cdY{2}&\cdOr{1}&\cdO{2}\\\cdP{1}&\cdB{2}&\cdT{1}&\cdS{1}\\\cdR{1}&\cdM{1}&&}\tickno & \fourwrap{\cdC{1}&\cdY{1}&\cdOr{1}&\cdO{1}\\\cdP{1}&\cdB{1}&\cdS{1}&\cdR{1}\\\cdM{1}&&&}\tickno & \fourwrap{\cdC{2}&\cdY{2}&\cdOr{2}&\cdO{1}\\\cdP{2}&\cdB{1}&\cdT{1}&\cdS{1}\\\cdR{1}&\cdM{1}&&}\tickno & \fourwrap{\cdC{2}&\cdY{2}&\cdG{1}&\cdOr{2}\\\cdO{2}&\cdP{2}&\cdB{2}&\cdT{2}\\\cdS{2}&\cdR{1}&\cdM{1}&}\tickno \\
TH & & \fourwrap{\cdC{1}&\cdY{1}&\cdOr{1}&\cdO{1}\\\cdP{1}&\cdB{1}&\cdT{1}&\cdS{1}\\\cdR{1}&\cdM{1}&&}\tickok & \fourwrap{\cdC{1}&\cdY{1}&\cdOr{1}&\cdO{1}\\\cdP{1}&\cdB{1}&\cdT{1}&\cdS{1}\\\cdR{1}&\cdM{1}&&}\tickok & \fourwrap{\cdC{2}&\cdY{2}&\cdOr{2}&\cdO{2}\\\cdP{2}&\cdB{2}&\cdT{1}&\cdS{2}\\\cdR{1}&\cdM{1}&&}\tickno & \fourwrap{\cdC{8}&\cdY{2}&\cdG{1}&\cdOr{1}\\\cdO{2}&\cdP{1}&\cdB{1}&\cdT{1}\\\cdGo{1}&\cdR{5}&\cdM{1}&}\tickno \\
\addlinespace[2pt]
\fliptrace{Qwen3.5-27B}{[\ldots] Let's look at the ``Purple'' block. In Image 1, Purple is to the right of Blue. In Image 2, Purple is to the left of Orchid. This implies the structure is: Back Row: Yellow/Grey, Cyan/Teal. Middle Row: Orange/Red, Orchid/?. Front Row: Blue/Magenta, Purple/?. Wait, let's re-examine the adjacency. Image 1: Top layer visible: Yellow, \textit{[trace cut off at the budget cap]}}
\fliptrace{Gemma4-31B}{[\ldots] Blue and Purple are on a plane below them (z=1). Magenta is on a plane below Blue (z=0). Red is on the same plane as Magenta (z=0). Slategray is on the same plane as Blue (z=1). Let's re-verify this. View 1: z=2: ( \textit{[trace cut off at the budget cap]}}
\fliptracelast{GLM-4.6V-Flash}{[\ldots] Yellow: In first image, there's a yellow block (top left front). In second image, there's a yellow block (middle right). Are there others? Let's check. First image's top left is yellow (1). Second image's middle right is yellow (1). So total yellow: 2. Cyan: First image top right (1), second image middle middle (1). So cyan: 2. Orchid: First image middle right (1), second image middle left (1). So orchid: 2. Orange: First image middle front left (1), second image top right \textit{[trace cut off at the budget cap]}}
\bottomrule
\caption{\tfivelong{}, thinking-mode qualitative panel. Non-Thinking answers $0$ of $4$ correctly and Thinking $2$ of $4$; $2$ models flip wrong to correct. The task-level verdict is \emph{tie (slight TH lead)}. Each cell is the 5-run majority vote of that model on the single default-setting instance shown; NT is Non-Thinking and TH is Thinking at the default $1024$-token budget. Beneath the outcome rows we quote the tail of each model's TH trace. Kimi-VL-A3B surfaces no reasoning tokens, so it has no quote.}
\label{tab:qual_flip_t5}\\
\end{longtable}
\endgroup

\begingroup
\small
\renewcommand{\arraystretch}{1.16}
\setlength{\tabcolsep}{2pt}
\begin{longtable}{>{\centering\arraybackslash}m{\dimexpr0.10\textwidth-4pt\relax}*{5}{>{\centering\arraybackslash}m{\dimexpr0.18\textwidth-4pt\relax}}}
\toprule
\endfirsthead
\toprule
\multicolumn{6}{l}{\tablename~\thetable{} continued.}\\
\midrule
\endhead
\multicolumn{6}{@{}l@{}}{\textbf{T6 Local Structure} \hfill \textit{mode verdict: TH wins}}\\
\flippromptline{determine whether the object has at least one column (vertical stack of blocks) with a height greater than 3}
\flipinputrow{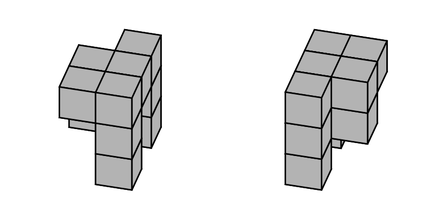}
\midrule
\flipmodelhdr
NT & \multirow{2}{*}{False} & True\tickno & False\tickok & True\tickno & True\tickno \\
TH & & False\tickok & True\tickno & False\tickok & True\tickno \\
\addlinespace[2pt]
\fliptrace{Qwen3.5-27B}{[\ldots] Back-left column: Height 3. (Visible: top). Back-right column: Height 3. (Visible: top, mid, bot on the far right edge). Wait, let's look at the ``L'' shape again. In Image 1, there is a block sticking out to the left. Let's look at the top surface. It looks like a 2x2 square of blocks. Let's assume the top layer is a 2x2 block. Underneath the front-left top block, there are two more blocks. So column height is 3. \textit{[trace cut off at the budget cap]}}
\fliptrace{Gemma4-31B}{[\ldots] Block below (height 2) Block below (height 1) Wait, let's look at the column behind it. The column at the back right: Top block (height 3) Block below (height 2) Block below (height 1) Let's look at the column in the middle: Top block (height 3) Block below (height 2) Block below (height 1) \textit{[trace cut off at the budget cap]}}
\fliptracelast{GLM-4.6V-Flash}{[\ldots] Wait, but wait---maybe I missed a column. Let's re-examine. Wait, in the first image, is there a column with height 4? Let's see the first image: the front-left column has 3 blocks. The column to the right of that (the middle column) has 2 blocks. The column to the right of that (the right column) has 2 blocks. So max height is 3. So no column with height \textgreater{}3. Therefore, the answer is False.}
\bottomrule
\caption{\tsixlong{}, thinking-mode qualitative panel. Non-Thinking answers $1$ of $4$ correctly and Thinking $2$ of $4$; $2$ models flip wrong to correct and $1$ model correct to wrong. The task-level verdict is \emph{TH wins}. Each cell is the 5-run majority vote of that model on the single default-setting instance shown; NT is Non-Thinking and TH is Thinking at the default $1024$-token budget. Beneath the outcome rows we quote the tail of each model's TH trace. Kimi-VL-A3B surfaces no reasoning tokens, so it has no quote.}
\label{tab:qual_flip_t6}\\
\end{longtable}
\endgroup

\begingroup
\small
\renewcommand{\arraystretch}{1.16}
\setlength{\tabcolsep}{2pt}
\begin{longtable}{>{\centering\arraybackslash}m{\dimexpr0.10\textwidth-4pt\relax}*{5}{>{\centering\arraybackslash}m{\dimexpr0.18\textwidth-4pt\relax}}}
\toprule
\endfirsthead
\toprule
\multicolumn{6}{l}{\tablename~\thetable{} continued.}\\
\midrule
\endhead
\multicolumn{6}{@{}l@{}}{\textbf{T7 Color Dominance} \hfill \textit{mode verdict: tie (slight TH lead on this instance)}}\\
\flippromptline{determine the dominant color(s) in the object}
\flipinputrow{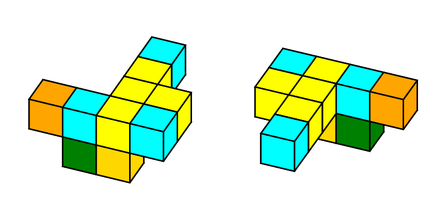}
\midrule
\flipmodelhdr
NT & \multirow{2}{*}{\fourwrap{\cdchip{bedYellow}&&&}} & \fourwrap{\cdchip{bedCyan}&\cdchip{bedYellow}&&}\tickno & \fourwrap{\cdchip{bedCyan}&\cdchip{bedYellow}&&}\tickno & \fourwrap{\cdchip{bedCyan}&\cdchip{bedYellow}&&}\tickno & \fourwrap{\cdchip{bedCyan}&\cdchip{bedGreen}&\cdchip{bedOrange}&\cdchip{bedYellow}}\tickno \\
TH & & \fourwrap{\cdchip{bedYellow}&&&}\tickok & \fourwrap{\cdchip{bedYellow}&&&}\tickok & \fourwrap{\cdchip{bedCyan}&\cdchip{bedYellow}&&}\tickno & \fourwrap{\cdchip{bedCyan}&\cdchip{bedYellow}&&}\tickno \\
\addlinespace[2pt]
\fliptrace{Qwen3.5-27B}{[\ldots] This is getting confusing. Let's count total blocks. Let's assume the object is a single connected component. Let's look at Image 2 again. Rightmost: Orange. Left of Orange: Cyan. Under Cyan: Green. Left of Cyan: Yellow. Under Yellow: Yellow (visible underneath). Left of that Yellow: Yellow. Left/Back of that Yellow: Cyan. Front/Left of that Yellow: Cyan. Let's try to map Image 1 to Image 2. Image 1 shows the object from a \textit{[trace cut off at the budget cap]}}
\fliptrace{Gemma4-31B}{[\ldots] Green block is on the bottom right. Cyan blocks: one on the far left, one in the middle, one at the top. Yellow blocks: several in the middle. Let's try to identify the blocks by their colors and positions: Orange block: Only one. Green block: Only one. Cyan blocks: View 1: One at (1,1,0) [relative], one at (3,1,0), one at (2,3,1). View 2: One at (0,0 \textit{[trace cut off at the budget cap]}}
\fliptracelast{GLM-4.6V-Flash}{[\ldots] 4. Rightmost orange's left cyan (the one adjacent to orange on the right) Wait, no, let's count each block in second image: Orange: 1 (rightmost) Cyan: Let's see, top left cyan, middle left cyan (the one below top left), middle right cyan (next to middle right yellow), and the one next to orange on the left. So that's 4 cyan? Yellow: Let's count yellow blocks: 1. Top middle yellow (next to top left cyan) 2. Middle middle yellow (next to top middle \textit{[trace cut off at the budget cap]}}
\bottomrule
\caption{\tsevenlong{}, thinking-mode qualitative panel. Non-Thinking answers $0$ of $4$ correctly and Thinking $2$ of $4$; $2$ models flip wrong to correct. The task-level verdict is \emph{tie (slight TH lead on this instance)}. Each cell is the 5-run majority vote of that model on the single default-setting instance shown; NT is Non-Thinking and TH is Thinking at the default $1024$-token budget. Beneath the outcome rows we quote the tail of each model's TH trace. Kimi-VL-A3B surfaces no reasoning tokens, so it has no quote.}
\label{tab:qual_flip_t7}\\
\end{longtable}
\endgroup

\begingroup
\small
\renewcommand{\arraystretch}{1.16}
\setlength{\tabcolsep}{2pt}
\begin{longtable}{>{\centering\arraybackslash}m{\dimexpr0.10\textwidth-4pt\relax}*{5}{>{\centering\arraybackslash}m{\dimexpr0.18\textwidth-4pt\relax}}}
\toprule
\endfirsthead
\toprule
\multicolumn{6}{l}{\tablename~\thetable{} continued.}\\
\midrule
\endhead
\multicolumn{6}{@{}l@{}}{\textbf{T8 Silhouette Symmetry} \hfill \textit{mode verdict: NT wins}}\\
\flippromptline{determine whether the object's silhouette is symmetric (vertically or horizontally)}
\flipinputrow{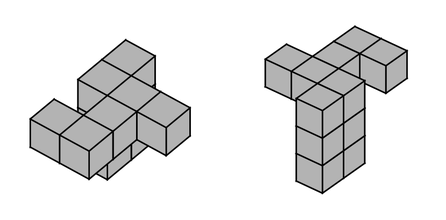}
\midrule
\flipmodelhdr
NT & \multirow{2}{*}{False} & False\tickok & False\tickok & False\tickok & False\tickok \\
TH & & False\tickok & True\tickno & False\tickok & True\tickno \\
\addlinespace[2pt]
\fliptrace{Qwen3.5-27B}{[\ldots] Let's look at Image 2 again. It's a ``T'' shape but the crossbar is uneven. Left side of crossbar: 1 block. Right side of crossbar: 2 blocks. Back side of crossbar: 2 blocks. Front side: The vertical column. This is definitely asymmetric. Wait, let me look really closely at the images to see if I'm misinterpreting the structure. Image 1: Let's trace the blocks. There is a central block. To its left: 2 blocks \textit{[trace cut off at the budget cap]}}
\fliptrace{Gemma4-31B}{[\ldots] (1,1,0), (1,1,1) - middle-back column (2,1,0), (2,1,1) - right column Wait, let's look at the image again. View 1: Front-left: (0,0,0), (0,0,1) Front-middle: (1,0,0), (1,0,1), (1,0,2) Back-middle: (1,1,0), (1,1,1) \textit{[trace cut off at the budget cap]}}
\fliptracelast{GLM-4.6V-Flash}{[\ldots] Wait, but maybe I'm missing. Wait, the question is whether the object's silhouette is symmetric. So check if either view's silhouette is symmetric. Since neither view shows vertical or horizontal symmetry, the answer is false.}
\bottomrule
\caption{\teightlong{}, thinking-mode qualitative panel. Non-Thinking answers $4$ of $4$ correctly and Thinking $2$ of $4$; $2$ models flip correct to wrong. The task-level verdict is \emph{NT wins}. Each cell is the 5-run majority vote of that model on the single default-setting instance shown; NT is Non-Thinking and TH is Thinking at the default $1024$-token budget. Beneath the outcome rows we quote the tail of each model's TH trace. Kimi-VL-A3B surfaces no reasoning tokens, so it has no quote.}
\label{tab:qual_flip_t8}\\
\end{longtable}
\endgroup

\begingroup
\small
\renewcommand{\arraystretch}{1.16}
\setlength{\tabcolsep}{2pt}
\begin{longtable}{>{\centering\arraybackslash}m{\dimexpr0.10\textwidth-4pt\relax}*{5}{>{\centering\arraybackslash}m{\dimexpr0.18\textwidth-4pt\relax}}}
\toprule
\endfirsthead
\toprule
\multicolumn{6}{l}{\tablename~\thetable{} continued.}\\
\midrule
\endhead
\multicolumn{6}{@{}l@{}}{\textbf{T9 Visible Floors} \hfill \textit{mode verdict: NT wins}}\\
\flippromptline{count the number of visible layers (floors/horizontal levels) in the object}
\flipinputrow{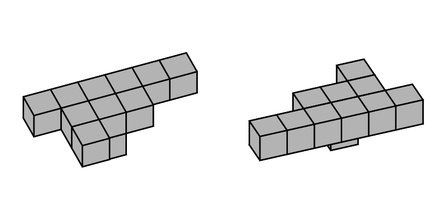}
\midrule
\flipmodelhdr
NT & \multirow{2}{*}{$2$} & $2$\tickok & $2$\tickok & $2$\tickok & $3$\tickno \\
TH & & $3$\tickno & $3$\tickno & $1$\tickno & $6$\tickno \\
\addlinespace[2pt]
\fliptrace{Qwen3.5-27B}{[\ldots] Let's try a different hypothesis. Maybe the object is a single layer, just viewed from an angle? If it's a single layer, all top faces should be on the same plane. In Image 1, look at the top faces. The top face of the leftmost block is at a certain height. The top face of the block next to it is at the same height. The top face of the block in the ``middle'' row is at the same height. The top face of the block in the ``front'' row... wait. \textit{[trace cut off at the budget cap]}}
\fliptrace{Gemma4-31B}{[\ldots] Let me re-read the image. The blocks are cubes. In the first image, the leftmost block has a height of 2. The next few have a height of 2. Then there's a part that goes down further. Let's look at the first image again. Top surface: 6 blocks. Below that surface: 6 blocks. Below those: 3 blocks. Wait, the 3 blocks are under the middle ones. \textit{[trace cut off at the budget cap]}}
\fliptracelast{GLM-4.6V-Flash}{[\ldots] mage, all blocks seem to be in the same vertical level? Wait, no---maybe there's only one layer. But wait, maybe some blocks are in different heights. Wait, no, in the first image, all the blocks are at the same height, right? Because it's a single layer. Then the second image---let's check. In the second image, is there any block that's on top of another? Let's see. The second image has some blocks that might be in a higher layer? Wait, no. Wait, maybe both images show the same object, which is a single layer. So the number of visible layers is 1.}
\bottomrule
\caption{\tninelong{}, thinking-mode qualitative panel. Non-Thinking answers $3$ of $4$ correctly and Thinking $0$ of $4$; $3$ models flip correct to wrong. The task-level verdict is \emph{NT wins}. Each cell is the 5-run majority vote of that model on the single default-setting instance shown; NT is Non-Thinking and TH is Thinking at the default $1024$-token budget. Beneath the outcome rows we quote the tail of each model's TH trace. Kimi-VL-A3B surfaces no reasoning tokens, so it has no quote.}
\label{tab:qual_flip_t9}\\
\end{longtable}
\endgroup

\begingroup
\small
\renewcommand{\arraystretch}{1.16}
\setlength{\tabcolsep}{2pt}
\begin{longtable}{>{\centering\arraybackslash}m{\dimexpr0.10\textwidth-4pt\relax}*{5}{>{\centering\arraybackslash}m{\dimexpr0.18\textwidth-4pt\relax}}}
\toprule
\endfirsthead
\toprule
\multicolumn{6}{l}{\tablename~\thetable{} continued.}\\
\midrule
\endhead
\multicolumn{6}{@{}l@{}}{\textbf{T10 Color Connected} \hfill \textit{mode verdict: TH wins}}\\
\flippromptline{determine whether there exists at least one pair of adjacent blocks (sharing a face) that share the same color}
\flipinputrow{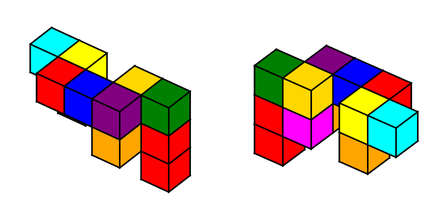}
\midrule
\flipmodelhdr
NT & \multirow{2}{*}{True} & True\tickok & False\tickno & False\tickno & True\tickok \\
TH & & True\tickok & True\tickok & True\tickok & True\tickok \\
\addlinespace[2pt]
\fliptrace{Qwen3.5-27B}{[\ldots] Orange block: Below purple. Green block: Next to yellow (left) and red (below). Wait, let's look at the second image to confirm the structure. 2. Analyze the second image (View 2): This view is from a different angle, seemingly from the ``front'' or ``bottom-right''. Let's identify blocks. Left stack: Green on top, Red in middle, Red on bottom. This confirms the stack seen in View 1. \textit{[trace cut off at the budget cap]}}
\fliptrace{Gemma4-31B}{[\ldots] Let's check View 2 to confirm this. View 2 (Bottom-right image): Looking from a different angle. Top layer: Green block (0, 0, 2) - Wait, the coordinates will be different. Purple block (1, 0, 2) Blue block (2, 0, 2) Red block (3, 0, 2) Middle layer: Yellow block (1, 0, 1) Yellow block (2, 0, \textit{[trace cut off at the budget cap]}}
\fliptracelast{GLM-4.6V-Flash}{[\ldots] Yellow (middle left, above magenta) Red (bottom left, below green) Magenta (middle left, below yellow) Orange (bottom middle, below magenta) Yellow (middle right, above cyan) Cyan (right, above yellow) Blue (middle right, above red) Red (right, above blue) Purple (top middle, above blue) Yellow (middle middle, below purple) Wait, let's list the colors in Image 2: 1. Green 2. Yellow 3. Red 4. Magenta 5. Orange 6. Yellow 7. Cyan 8. Blue 9. Red 10. Purple \textit{[trace cut off at the budget cap]}}
\bottomrule
\caption{\ttenlong{}, thinking-mode qualitative panel. Non-Thinking answers $2$ of $4$ correctly and Thinking $4$ of $4$; $2$ models flip wrong to correct. The task-level verdict is \emph{TH wins}. Each cell is the 5-run majority vote of that model on the single default-setting instance shown; NT is Non-Thinking and TH is Thinking at the default $1024$-token budget. Beneath the outcome rows we quote the tail of each model's TH trace. Kimi-VL-A3B surfaces no reasoning tokens, so it has no quote.}
\label{tab:qual_flip_t10}\\
\end{longtable}
\endgroup

\begingroup
\small
\renewcommand{\arraystretch}{1.16}
\setlength{\tabcolsep}{2pt}
\begin{longtable}{>{\centering\arraybackslash}m{\dimexpr0.10\textwidth-4pt\relax}*{5}{>{\centering\arraybackslash}m{\dimexpr0.18\textwidth-4pt\relax}}}
\toprule
\endfirsthead
\toprule
\multicolumn{6}{l}{\tablename~\thetable{} continued.}\\
\midrule
\endhead
\multicolumn{6}{@{}l@{}}{\textbf{T11 Projection} \hfill \textit{mode verdict: tie (slight NT lead)}}\\
\flippromptline{identify the invalid projection}
\flipinputrow{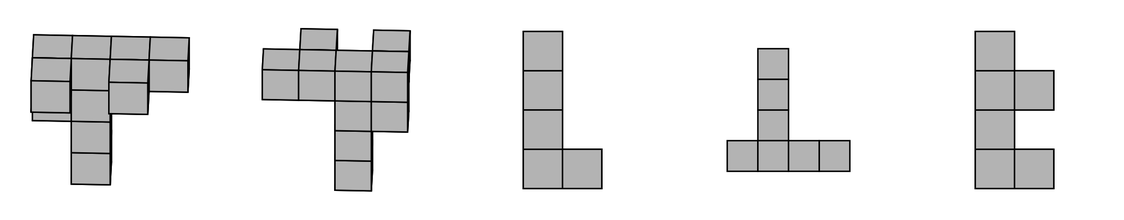}
\midrule
\flipmodelhdr
NT & \multirow{2}{*}{1} & 1\tickok & 1\tickok & 1\tickok & 0\tickno \\
TH & & 2\tickno & 0\tickno & 2\tickno & 2\tickno \\
\addlinespace[2pt]
\fliptrace{Qwen3.5-27B}{[\ldots] Fourth row: 1 square (center-left). Fifth row: 1 square (center-left). This seems like a specific arrangement. Let's assume the object is made of unit cubes. Let's try to map it to a 3D grid. Let's assume the ``front'' face is what we see in View 1. Columns (left to right): Col 1: 3 blocks high. Col 2: 5 blocks high (top 2 + stem 3? No, let \textit{[trace cut off at the budget cap]}}
\fliptrace{Gemma4-31B}{[\ldots] Bottom: A column of 3 blocks extending down from the second column from the left. Wait, the bottom part is 3 blocks high. Let's re-examine the 3D view (Image 1) again. Top: 4 blocks wide, 2 blocks deep. Bottom: A column of 3 blocks extending down from the second column from the left. Wait, the bottom part is 3 blocks high. Let's \textit{[trace cut off at the budget cap]}}
\fliptracelast{GLM-4.6V-Flash}{[\ldots] Alternatively, candidate 0: The top view should show the base. Let's see if the top view matches. The first 3D view's top would have the footprint. Let's see image 3's shape: It's \textit{[trace cut off at the budget cap]}}
\bottomrule
\caption{\televenlong{}, thinking-mode qualitative panel. Non-Thinking answers $3$ of $4$ correctly and Thinking $0$ of $4$; $3$ models flip correct to wrong. The task-level verdict is \emph{tie (slight NT lead)}. Each cell is the 5-run majority vote of that model on the single default-setting instance shown; NT is Non-Thinking and TH is Thinking at the default $1024$-token budget. Beneath the outcome rows we quote the tail of each model's TH trace. Kimi-VL-A3B surfaces no reasoning tokens, so it has no quote.}
\label{tab:qual_flip_t11}\\
\end{longtable}
\endgroup

\begingroup
\small
\renewcommand{\arraystretch}{1.16}
\setlength{\tabcolsep}{2pt}
\begin{longtable}{>{\centering\arraybackslash}m{\dimexpr0.10\textwidth-4pt\relax}*{5}{>{\centering\arraybackslash}m{\dimexpr0.18\textwidth-4pt\relax}}}
\toprule
\endfirsthead
\toprule
\multicolumn{6}{l}{\tablename~\thetable{} continued.}\\
\midrule
\endhead
\multicolumn{6}{@{}l@{}}{\textbf{T12 Reconstruct} \hfill \textit{mode verdict: tie (slight NT lead)}}\\
\flippromptline{identify the positive view}
\flipinputrow{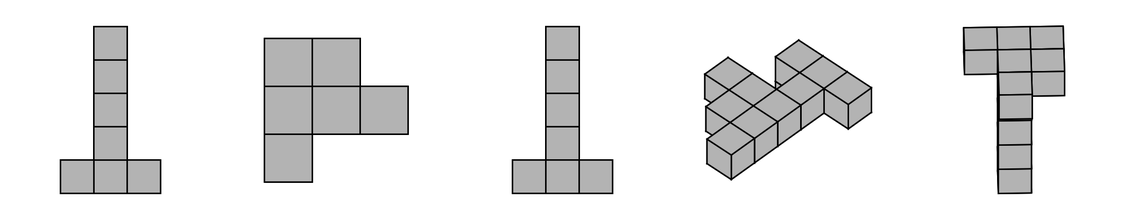}
\midrule
\flipmodelhdr
NT & \multirow{2}{*}{1} & 1\tickok & 0\tickno & 1\tickok & 1\tickok \\
TH & & 0\tickno & 0\tickno & 0\tickno & 1\tickok \\
\addlinespace[2pt]
\fliptrace{Qwen3.5-27B}{[\ldots] Let's reconsider the standard set: Top, Front, Right Side. If Image 1 is Top View: It shows a 3x3 grid area. Blocks at: (0,0), (0,1), (0,2), (1,1), (1,2), (2,1). (Using x,y coordinates where x is right, y is up). Wait, let's map Image 1 to a grid. Row 1 (top): Col 1, Col 2. Row 2 (mid): Col 1, Col 2, Col 3. Row 3 (bot): Col 1. If Image 2 is Front View: It \textit{[trace cut off at the budget cap]}}
\fliptrace{Gemma4-31B}{[\ldots] Candidate View 0 (Image 4): It's a 3D isometric view. Let's count the cubes. Top layer: 3 cubes in a row (along one axis), and 1 cube attached to the side of the middle one? No, let's look closer. Top layer: (0,0,0), (0,1,0), (0,2,0) - a row of 3. Wait, let's look at the cubes again. Top layer: (0,0,0), (0,1,0), (0,2,0) - a row of 3. Middle layer: (1,0,0), (1,1,0), (1,2,0) - a row of 3 \textit{[trace cut off at the budget cap]}}
\fliptracelast{GLM-4.6V-Flash}{[\ldots] Wait, the silhouettes are all 2D front views. So the 3D object's front view is the T-shape. Now, Image 5 is a different 2D view (maybe top-down) that doesn't match the front view. So Image 5 is negative. Image 4 is a 3D view---if the 3D object is as we inferred, then Image 4 should show the correct structure. But wait, \textit{[trace cut off at the budget cap]}}
\bottomrule
\caption{\ttwelvelong{}, thinking-mode qualitative panel. Non-Thinking answers $3$ of $4$ correctly and Thinking $1$ of $4$; $2$ models flip correct to wrong. The task-level verdict is \emph{tie (slight NT lead)}. Each cell is the 5-run majority vote of that model on the single default-setting instance shown; NT is Non-Thinking and TH is Thinking at the default $1024$-token budget. Beneath the outcome rows we quote the tail of each model's TH trace. Kimi-VL-A3B surfaces no reasoning tokens, so it has no quote.}
\label{tab:qual_flip_t12}\\
\end{longtable}
\endgroup

\begingroup
\small
\renewcommand{\arraystretch}{1.16}
\setlength{\tabcolsep}{2pt}
\begin{longtable}{>{\centering\arraybackslash}m{\dimexpr0.10\textwidth-4pt\relax}*{5}{>{\centering\arraybackslash}m{\dimexpr0.18\textwidth-4pt\relax}}}
\toprule
\endfirsthead
\toprule
\multicolumn{6}{l}{\tablename~\thetable{} continued.}\\
\midrule
\endhead
\multicolumn{6}{@{}l@{}}{\textbf{T13 Mixture-of-Shapes} \hfill \textit{mode verdict: tie (slight NT lead)}}\\
\flippromptline{identify the invalid silhouette}
\flipinputrow{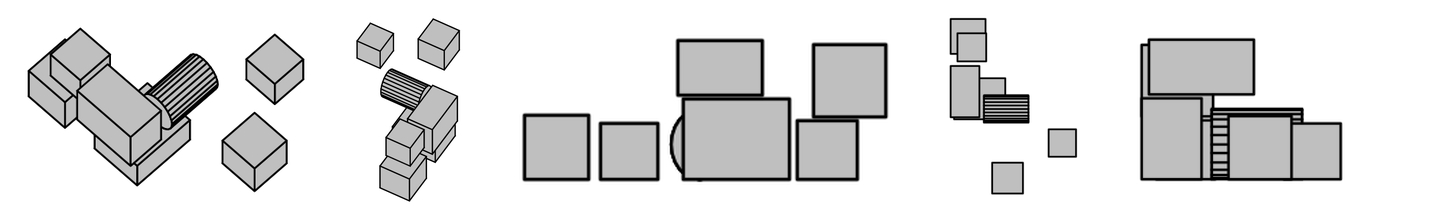}
\midrule
\flipmodelhdr
NT & \multirow{2}{*}{2} & 2\tickok & 2\tickok & 2\tickok & 2\tickok \\
TH & & 2\tickok & 1\tickno & 1\tickno & 2\tickok \\
\addlinespace[2pt]
\fliptrace{Qwen3.5-27B}{[\ldots] Image 4: Silhouette. Shows a vertical stack on the left, a cylinder in the middle, and two squares at the bottom right. Image 5: Silhouette. Shows a large L-shape on the left, a cylinder section, and blocks on the right. Let's look at the ``Candidate Silhouettes'' more carefully. The prompt says ``three candidate 2D orthographic silhouettes''. Candidate 0 (Image 3): This looks like a top \textit{[trace cut off at the budget cap]}}
\fliptrace{Gemma4-31B}{[\ldots] Box 4: (1,1,0) - on top of Box 3 Cylinder: (1,2,0) - on top of Box 4 (lying horizontally along the x-axis? No, it looks like it's lying along the y-axis or z-axis). Box 5: (3,2,0) - separate Box 6: (3,0,0) - separate Perspective 2: Box 1: (0,0,0) - bottom Box 2: (0,1,0) - on top of Box 1 Box 3: (1,1,0) - to \textit{[trace cut off at the budget cap]}}
\fliptracelast{GLM-4.6V-Flash}{[\ldots] . But the first two 3D views show the cylinder attached to a block, so the orthographic silhouette should include the cylinder's shape. The third silhouette has a circle, which could be the end view of the cylinder. But let's check the number of objects. The 3D scene has several blocks: let's count the blocks in the 3D views. First view: left side has 3 blocks (small, medium, large), base block, then the cylinder. Then two small cubes on right. So total blocks: 3 +1 +2 =6? Wait, no \textit{[trace cut off at the budget cap]}}
\bottomrule
\caption{\tthirteenlong{}, thinking-mode qualitative panel. Non-Thinking answers $4$ of $4$ correctly and Thinking $2$ of $4$; $2$ models flip correct to wrong. The task-level verdict is \emph{tie (slight NT lead)}. Each cell is the 5-run majority vote of that model on the single default-setting instance shown; NT is Non-Thinking and TH is Thinking at the default $1024$-token budget. Beneath the outcome rows we quote the tail of each model's TH trace. Kimi-VL-A3B surfaces no reasoning tokens, so it has no quote.}
\label{tab:qual_flip_t13}\\
\end{longtable}
\endgroup

\twocolumn
\subsection{Thinking Traces: Instance-Level Evidence}
\label{sec:app_qual_thinking_panels}

These panels give the instance-level qualitative evidence behind the trace audit in Appendix~\ref{sec:app_thinking}. For one representative sample per task, we show the inputs, each model's parsed answer, and the collected reasoning trace or meta-reasoning summary. Each answer table reports all eight models on the same input. For space, we show only a subset of the 65 annotated traces: every trace quoted in Table~\ref{tab:annot_taxonomy} (Appendix~\ref{sec:app_thinking}), and, for tasks without a quoted trace (\tthree{}, \tnine{}, \tthirteen{}), the trace of the best-performing open-weight model for that task. The open-weight traces are raw token-level thinking rollouts. For GPT-5 and the two Gemini models, we manually collected meta-reasoning summaries through their respective chat interfaces because the OpenRouter and Vertex AI endpoints used in our evaluation did not expose reasoning traces. For Grok-4.20, the meta-reasoning trace was returned through OpenRouter. Within each trace, we manually highlight the spans that drive the outcome and tag each with a bracketed superscript label from the behavior taxonomy of Table~\ref{tab:annot_taxonomy} (Appendix~\ref{sec:app_thinking}); each label keeps a fixed highlight color across all panels, and spans that carry no tag are left unhighlighted.

% These panels give the instance-level qualitative evidence behind the trace audit in Appendix~\ref{sec:app_thinking}. For one representative sample per task, we show the inputs, each model's parsed answer, and the collected reasoning trace or meta-reasoning summary. Each panel reports a trace from the best-performing open-weight model for that task together with the four frontier models on the same input. The open-weight traces are raw token-level thinking rollouts. For GPT-5 and the two Gemini models, we manually collected meta-reasoning summaries through their respective chat interfaces because the OpenRouter and Vertex AI endpoints used in our evaluation did not expose reasoning traces. For Grok-4.20, the meta-reasoning trace was returned through OpenRouter. Within each trace, we manually highlight the spans that drive the outcome and tag each with a bracketed superscript label from the behavior taxonomy of Table~\ref{tab:annot_taxonomy} (Appendix~\ref{sec:app_thinking}); each label keeps a fixed highlight color across all panels, and spans that carry no tag are left unhighlighted.
% GPT-5 is the only model that uses tools in these traces; its code-based image inspection is tagged as \emph{Tool Use}. The two universal trace artifacts appear as ordinary tags: the open-weight hard-truncation stop (\emph{Budget Hit}) and Grok's derailment into its own summarization meta-instruction (\emph{Harnessing}).

\onecolumn
% Auto-generated by scripts/gen_qualitative_trace_panels.py -- do not edit by hand.
\newcommand{\tpicon}[1]{\includegraphics[height=3.6ex]{images/app_qual/#1}}
\definecolor{annColDefault}{HTML}{EEEEEE}
\definecolor{annCol0}{HTML}{FFD966}
\definecolor{annCol1}{HTML}{8ECBFF}
\definecolor{annCol10}{HTML}{D8B4FE}
\definecolor{annCol2}{HTML}{FFB07C}
\definecolor{annCol3}{HTML}{6FE0D0}
\definecolor{annCol4}{HTML}{F2A6D8}
\definecolor{annCol5}{HTML}{C9A66B}
\definecolor{annCol6}{HTML}{A6B1F5}
\definecolor{annCol7}{HTML}{F5E07C}
\definecolor{annCol8}{HTML}{BBD4FF}
\definecolor{annCol9}{HTML}{E0987C}
\newcommand{\tann}[3]{{\sethlcolor{#1}\hl{#2}}\textsuperscript{\tiny[#3]}}

\subsubsection{T1 Volume}
\begin{table}[H]
\centering
\footnotesize
\setlength{\tabcolsep}{1pt}
\renewcommand{\arraystretch}{1.2}
\begin{tabular}{*{9}{>{\centering\arraybackslash}p{\dimexpr\textwidth/9-2\tabcolsep\relax}}}
\toprule
\multicolumn{9}{@{}l@{}}{\itshape ``\ldots{}Your task is to count the total number of blocks that make up the 3D object\ldots{}''}\\[0.3mm]
\multicolumn{9}{@{}c@{}}{\begin{tabular}{*{2}{c}}\includegraphics[width=0.450\textwidth,height=1.25in,keepaspectratio]{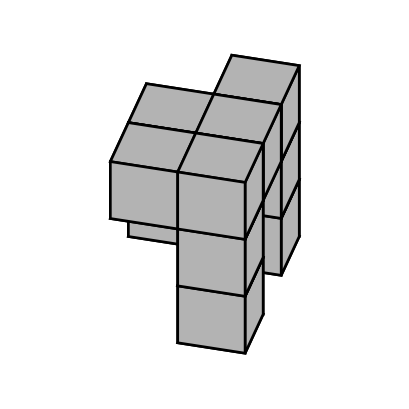} & \includegraphics[width=0.450\textwidth,height=1.25in,keepaspectratio]{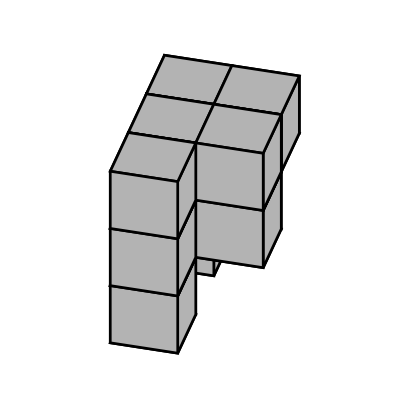}\end{tabular}}\\[0.5mm]
\textbf{GT} & \tpicon{qwen_icon.png} & \tpicon{gemma_icon.png} & \tpicon{glmv_icon.png} & \tpicon{kimi_icon.png} & \tpicon{grok_icon.png} & \tpicon{gpt_icon.png} & \tpicon{gemini_flash_icon.png} & \tpicon{gemini_pro_icon.png}\\
10 & 11\tickno & 11\tickno & 11\tickno & 14\tickno & 12\tickno & 9\tickno & 11\tickno & 8\tickno\\
\bottomrule
\end{tabular}
\caption{T1 Volume.}
\label{tab:qual_trace_t1}
\end{table}
\noindent\textbf{Thinking traces.}\par\smallskip
\begin{tcolorbox}[breakable,enhanced,colback=red!4,colframe=red!55!black,boxrule=0.4pt,arc=1pt,left=4pt,right=4pt,top=3pt,bottom=3pt,fonttitle=\footnotesize\bfseries,coltitle=black,colbacktitle=red!12,title={\tpicon{gemma_icon.png}~Gemma4-31B\hfill answer: 11\tickno}]
\footnotesize\raggedright
* Image 1: A 3D structure of blocks. \newline
* Image 2: The same 3D structure from a different angle. \newline
* Goal: Count the total number of blocks.

* Looking from the front-left-top. \newline
* Top layer: \newline
* \tann{annCol10}{One block at the back-right. * One block in the middle. * One block at the front-left.}{Wrong Perception} \newline
* \tann{annCol8}{Wait, let's re-examine.}{Reiterating} \newline
* Topmost block: (1, 1, 3) - let's use (x, y, z) coordinates where x is left-right, y is front-back, z is height. \newline
* \tann{annCol10}{Let's define the coordinate system: * x: 0 = left, 1 = middle, 2 = right * y: 0 = front, 1 = middle, 2 = back * z: 0 = bottom, 1 = middle, 2 = top, 3 = topmost}{Wrong Perception} \newline
* Looking at Image 1: \newline
* Topmost block (z=3): (1, 1, 3) - \tann{annCol8}{wait, it looks like it's at (1, 1, 3) or (2, 1, 3). Let's re-evaluate.}{Reiterating} \newline
* Let's look at the top surface. There's a block at the very top. It's above another block. \newline
* Let's count the visible blocks in Image 1: \newline
* \tann{annCol10}{Top layer (z=3): 1 block. * Layer z=2: 3 blocks (one under the top block, one to its left, one to its front). * Layer z=1: 4 blocks (one under each of the z=2 blocks, plus one more?). * Layer z=0: 2 blocks (one under the middle column, one under the front-left column).}{Wrong Perception} \newline
* \tann{annCol8}{Wait, this is confusing. Let's try again. }{Reiterating} \newline
* Looking from the front-left-top. \newline
* Let's identify the columns: \newline
* \tann{annCol10}{Column 1 (front-left): Height = 2 blocks. * Column 2 (middle): Height = 4 blocks. * Column 3 (back-right): Height = 3 blocks.}{Wrong Perception} \newline
* \tann{annCol8}{Wait, let's look closer.}{Reiterating} \newline
* In Image 1: \newline
* \tann{annCol10}{Front-left column: 2 blocks high. * Middle column: 4 blocks high. * Back-right column: 3 blocks high. * Wait, there's also a block to the left of the middle column at height 2.}{Wrong Perception} \newline
* \tann{annCol8}{Let's re-examine Image 1:}{Reiterating} \newline
* \tann{annCol10}{Topmost block: (1, 1, 3) - let's say x=1, y=1, z=3. * Below it: (1, 1, 2), (1, 1, 1), (1, 1, 0). (4 blocks) * To the left of (1, 1, 2): (0, 1, 2). * Below (0, 1, 2): (0, 1, 1), (0, 1, 0). (3 blocks) * To the front of (1, 1, 2): (1, 0, 2). * Below (1, 0, 2): (1, 0, 1), (1, 0, 0). (3 blocks) * To the back of (1, 1, 2): (1, 2, 2). * Below (1, 2, 2): (1, 2, 1), (1, 2, 0). (3 blocks)}{Wrong Perception} \newline
* \tann{annCol8}{Wait, let's look at the image again.}{Reiterating} \newline
* Image 1: \newline
* \tann{annCol10}{Top block: 1 * Layer below top: 3 blocks (forming an L-shape or T-shape?) * Let's look at the top surface. There's one block at the very top. Below it is a 2x2 area? No. * Let's count the visible faces. * Top faces: 4 faces are visible. * One at the very top.}{Wrong Perception} \newline
\tann{annCol0}{I have to give the solution based on the reasoning directly now.}{Budget Hit}
\end{tcolorbox}
\begin{tcolorbox}[breakable,enhanced,colback=red!4,colframe=red!55!black,boxrule=0.4pt,arc=1pt,left=4pt,right=4pt,top=3pt,bottom=3pt,fonttitle=\footnotesize\bfseries,coltitle=black,colbacktitle=red!12,title={\tpicon{grok_icon.png}~Grok-4.20\hfill answer: 12\tickno}]
\footnotesize\raggedright
The problem asks to count the total number of blocks in a 3D object from two perspective images. \newline
I need to count the total number of blocks in this 3D object based on the two perspective views. The object is made up of unit cubes, and I have to reconstruct it mentally to ensure it matches both angles.

From the first view, it appears to have a base with multiple layers. \tann{annCol10}{There's a cross-shapedbase at the bottom, then a taller stack on one side.}{Wrong Perception} \tann{annCol2}{Counting visible cubes in the first image, I see about 10 or 11 cubes, but some are hidden from this angle.}{Good Intuition}

The second image shows a different perspective, revealing more about the height and protrusions. It confirms the same structure but from anotherside, \tann{annCol10}{with a prominent column of 4 cubes on the left and a 2x2 block on top.}{Wrong Perception}

To find the total, I reconstruct the grid positions. Assigning coordinates, the object consists of cubes at positions like a central body with extensions.

\tann{annCol10}{After piecing it together, considering overlaps and hidden parts,the total number of blocks is 15.}{Wrong Perception}

\tann{annCol8}{But let's double-check.}{Reiterating} If visible in first is 10, second is 12, but with 3 hidden, it fits.

No, that's rough.

Upon closer inspection of the views, the object is composed of a 3x3x3 cube with some missing, but no.

Let's count precisely.

In the first perspective, I count 9 visible cubes, accounting for the protrusions and stacks.

The first image shows a sort of L-shape with heights varying.

Counting each distinct cube in the silhouette and internal ones.

There are 8 cubes in the base layer, 4in the middle, 3 on top, totaling 15.

\tann{annCol8}{Counting each distinct cube in the silhouette and internal ones. There are 8 cubes in the base layer, 4 in the middle, 3 on top, totaling 15.}{Reiterating}

But is that accurate? The views don't show that many in base.

Thesecond view has a base that's wider.

Perhaps it's 12.

I think the intended answer is 12.

Let me see why.

In the first view, there are 7 cubes visible, but wait.

\tann{annCol5}{Since the internal reasoning is truncated, I can't get the exact count, but the task is to summarize the thinkingtrace.}{Harnessing}

The trace is about trying to visualize and count, considering hidden blocks, and reconstructing the 3D model.

So, summarizing my thoughts: I examined the two images, counted visible blocks in each, accounted for possible hidden ones, and reconstructed the structure to ensure consistency.

Based on that, the total number ofblocks is 12.

I think that's it.

But the output format is specific, only the json.

\textit{[[...trimmed output...]]}

% But the task is to summarize the trace for the user query, but don't present it in third person.

% So I need to deliver the final answer directly, but the final answer is the json with the number.

% \tann{annCol5}{Theuser's question is to solve the problem, and the trace is the internal reasoning. But the instruction is to summarize the trace, but the policy says to speak directly as Grok answering the user's question.}{Harnessing}

% The human message is: Summarize the thinking trace for the user query, but don't present it in the third person asthe trace of a separate entity.

% Then the [User's Question] is the problem about counting blocks.

% Then [Internal Reasoning] is the trace.

% So my job is to summarize the trace, but in first person, as my thoughts, and then give the final answer in the format.

% \tann{annCol5}{But the policy says to turnthe internal reasoning into a clean, natural final response. And speak directly to the end user.}{Harnessing}

% So I think I need to give the answer as the json, but perhaps with some explanation in first person.

% No, the output format is specified, and the example is just the json.

% The problem says "Output Format:Please answer in this format: \{ 'total\_blocks': int, \}"

% So the final response should be that.

% \tann{annCol5}{But the system prompt says to present the content as first person thoughts instead of third person. The system prompt says: Present the content as your first person thoughts instead of a third person entity's thoughts. You willspeak directly to the end user.}{Harnessing}

So I think the response should be my thoughts in first person, leading to the answer.

\tann{annCol5}{But then "deliver the final answer directly." And "Just deliver the final answer directly." But also "Write the response as if you are the original model directly answering the user"}{Harnessing}

So probably justthe json, but since it's to turn the thinking into clean response, and first person.

The user query is to summarize the thinking trace, but the system is to turn it into clean response. \newline
\tann{annCol8}{ The human message is: Summarize the thinking trace for the user query, but don't present it in the third person as thetrace of a separate entity. Then the [User's Question] is the problem about counting blocks. Then [Internal Reasoning] is the thinking.}{Reiterating}

So my job is to summarize the internal reasoning into a clean, natural final response for the end user, in first person, as if I'm the assistant answering the question.

Thequestion is to output in the format \{ 'total\_blocks': int \}

So the final answer should be that, but in first person? No, the policy says "Present the content as your first person thoughts instead of a third person entity's thoughts."

But the output format is json.

Perhaps I need to explain in first personand then give the json.

\tann{annCol5}{But the policy says "Write the response as if you are the original model directly answering the user --- not as a summarizer." And "deliver the final answer directly." }{Harnessing} \newline
So probably just the json, but since it's to summarize the thinking, but the user is asking to summarize the thinking,but no, the human message is "Summarize the thinking trace for the user query, but don't present it in the third person as the trace of a separate entity."

Then provides the [User's Question] and [Internal Reasoning].

So the task is to summarize the internal reasoning for the user, in first person,as my thoughts.

But the output format is specified in the user's question as a json with total\_blocks.

The user's question is the problem about counting blocks, and the internal reasoning is the thinking trace.

But the human is asking me to summarize the thinking trace for the user query.
\end{tcolorbox}
\begin{tcolorbox}[breakable,enhanced,colback=red!4,colframe=red!55!black,boxrule=0.4pt,arc=1pt,left=4pt,right=4pt,top=3pt,bottom=3pt,fonttitle=\footnotesize\bfseries,coltitle=black,colbacktitle=red!12,title={\tpicon{gpt_icon.png}~GPT-5\hfill answer: 9\tickno}]
\footnotesize\raggedright
Analyzing image for cube count \newline
I need to count the cubes from two different views of the object. It looks like the object could be a puzzle with isometric cubes, rotated for perspective. I'll inspect the images and try to determine the total cube count. Maybe zoom in or crop the images for better clarity. \newline
Analyzed image \newline
Output image \newline
Analyzed image \newline
Output image \newline
Counting cubes in 3D perspectives \newline
I'm thinking in 3D coordinates here. This cube object seems to be a grid of columns with varying heights. In the second image, I see a column of 3 cubes stacked on the left (front-left). \tann{annCol10}{The top row might show a 2x2 square, and I can spot cubes on the right side too.}{Wrong Perception} \newline
Breaking down cube positions in image II \newline
In the second image, \tann{annCol10}{the top layer seems to have 4 cubes arranged in a 2x2 square.}{Wrong Perception} The left vertical column has 3 cubes, including one at the top. The right vertical portion shows two stacked cubes hanging down, with clear visible faces. It seems there might be a total of 9 visible cubes. \newline
Reconstructing coordinates from isometric views \newline
I need to reconstruct the object's coordinates by parsing the images systematically. \tann{annCol10}{The object seems to be composed of a 2x2 top platform with two vertical 3-cube legs at opposite corners and a cube beneath the platform.}{Wrong Perception} I'll use both views to figure out positions and apply geometry to clarify each cube's location. \newline
% Inferring cube arrangement in 3D grid \newline
% The two images likely show the same object from different perspectives, 90 degrees apart. I'm imagining this as an isometric projection with a 3D grid. It looks like cubes are stacked, and there might be overhangs but no floating blocks. The object seems to have vertical columns and an arch-like connection. \newline
% Analyzing right column in 3D structure \newline
\textit{[[...trimmed output...]]} \newline
The right vertical column seems to have two cubes stacked vertically, with a gap between them and the ground. \tann{annCol10}{The top face appears at height 2}{Wrong Perception}, and there's a visible overhang at the center. \tann{annCol10}{The lower face of the right front stack extends down two cubes, making the right vertical column height around 2 instead of 3.}{Wrong Perception} \newline
Breaking down the cube structure's arrangement \newline
\tann{annCol10}{In the top layer of height 3, there's a 2x2 arrangement. Under it, the left column extends down with 2 cubes below, while the right column extends with 1 cube below, possibly the front-right as well.}{Wrong Perception} From the visible front faces, I count 4 cubes on top, plus a few cubes below the left and right columns, increasing the total to about 8 cubes. \newline
Considering the object's overall structure \newline
\tann{annCol10}{It looks like this object could be a combination of a 2x2 slab on top (4 cubes), with three vertical legs extending from the corners and possibly an overhanging part in the center.}{Wrong Perception} By counting the visible cubes, \tann{annCol2}{I estimate a total of around 10 cubes.}{Good Intuition} I'll use the two images and projections to help predict the full structure and model the 3D geometry. \newline
Analyzing cube geometry from images \newline
It seems like the images are isometric views with clear line segments where I can estimate cube positions. \tann{annCol10}{The top surface appears as a 2x2 grid, dividing into four cubes. }{Wrong Perception}I've identified key points in the top polygon of image2 and started mapping out the left vertical face based on coordinates from the image. This should help capture the object's geometry. \newline
Breaking down cube geometry \newline
It looks like the structure involves a column at a height of 3. \tann{annCol10}{The top face of this column aligns with the front-left of the 2x2 slab.}{Wrong Perception} T\tann{annCol10}{here's also a vertical face on the right side, which seems to correspond to a column at the right-front of the slab, with height 2.}{Wrong Perception} I'll focus on identifying the precise column positions from these coordinates. \newline
Identifying column and hidden cubes \newline
The structure seems to have a column at the back-right, evidenced by the side vertical face at coordinates (300, 75) to (300, 133). This could indicate a height of 2. Additionally \newline
Estimating visible column heights \newline
\tann{annCol2}{I'm considering the possibility that column C might have a height of 3 and could be mostly hidden behind A and B, with only its underside slightly visible. If that's the case, we can assume a total of 10 cubes.}{Good Intuition} Looking at Image 1 with a 90 deg rotated view might help confirm the column heights, especially for columns A, B, C, and D. \newline
Analyzing column heights from images \newline
\tann{annCol9}{from PIL import Image import matplotlib.pyplot as plt img1=Image.open('/mnt/data/T1\_sample1\_row62\_query\_image\_1.png') img2=Image.open('/mnt/data/T1\_sample1\_row62\_query\_image\_2.png') img1.size, img2.size Matplotlib is building the font cache; this may take a moment. ((409, 409), (409, 409))}{Tool Use} \newline
Evaluating object heights from image analysis \newline
Output image \newline
\textit{[[...trimmed output...]]} \newline
% Analyzing the top surface and columns \newline
% In Image 1, the top surface seems to form a skewed 2x2 slab, with an L-shape overall. The front-left and front-right cubes are visible in the grid, with vertical lines separating the left/right and front/back columns. The front-right stack's vertical extension includes cubes below the top. This provides a 2x2 grid of cubes on top. \newline
% Assessing visible and hidden cube positions \newline
% The visible lower protrusion looks offset, suggesting it might be a hidden cube adjacent to the left one, possibly behind or below. For the far-right back column, the side face appears split into three height levels, indicating a vertical column of height 3. From the front view, the positions and heights of the 2x2 grid are mapped: the right front and back visible columns are height 3, while the left columns seem potentially lower. \newline
% Confirming height relationships between views \newline
% The heights likely map to [1,3,?,3]. From Image2's angle, the left column is height 3, and the right column appears as height 2. So, I need to compare these positions based on rotations. It seems the columns from Image1 and Image2 correspond with each other, but the heights need further fine-tuning. \newline
% Evaluating cube arrangement and height estimates \newline
% Output image \newline
% Mapping grid and column heights in Image2 \newline
% In Image2, the left vertical column is three cubes high, with the top cube being the front-left top cell. 
The right-front face has a height of 2, not reaching the ground. The far-right side face, however, is three cubes high with segments split at different heights, marking it as taller than expected.

```python \newline
\{ \newline
'total blocks': 9, \newline
\} \newline
```
\end{tcolorbox}

\subsubsection{T2 Packaging}
\begin{table}[H]
\centering
\footnotesize
\setlength{\tabcolsep}{1pt}
\renewcommand{\arraystretch}{1.2}
\begin{tabular}{*{9}{>{\centering\arraybackslash}p{\dimexpr\textwidth/9-2\tabcolsep\relax}}}
\toprule
\multicolumn{9}{@{}l@{}}{\itshape ``\ldots{}Your task is to determine whether the object can fit in the boundary in any possible orientation. Boundary: \text{[2,5,6]}\ldots{}''}\\[0.3mm]
\multicolumn{9}{@{}c@{}}{\begin{tabular}{*{2}{c}}\includegraphics[width=0.450\textwidth,height=1.25in,keepaspectratio]{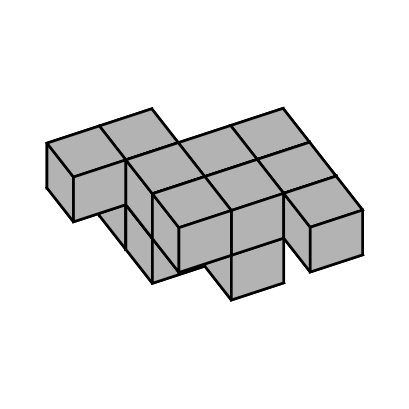} & \includegraphics[width=0.450\textwidth,height=1.25in,keepaspectratio]{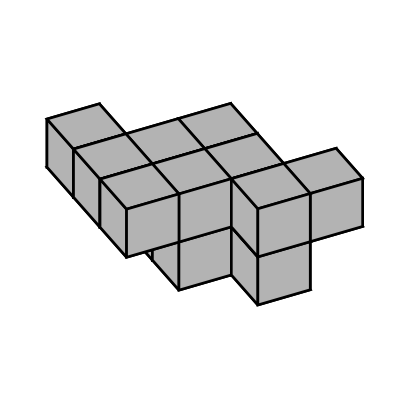}\end{tabular}}\\[0.5mm]
\textbf{GT} & \tpicon{qwen_icon.png} & \tpicon{gemma_icon.png} & \tpicon{glmv_icon.png} & \tpicon{kimi_icon.png} & \tpicon{grok_icon.png} & \tpicon{gpt_icon.png} & \tpicon{gemini_flash_icon.png} & \tpicon{gemini_pro_icon.png}\\
True & True\tickok & True\tickok & True\tickok & True\tickok & False\tickno & True\tickok & True\tickok & True\tickok\\
\bottomrule
\end{tabular}
\caption{T2 Packaging.}
\label{tab:qual_trace_t2}
\end{table}
\noindent\textbf{Thinking traces.}\par\smallskip
\begin{tcolorbox}[breakable,enhanced,colback=green!4,colframe=green!50!black,boxrule=0.4pt,arc=1pt,left=4pt,right=4pt,top=3pt,bottom=3pt,fonttitle=\footnotesize\bfseries,coltitle=black,colbacktitle=green!14,title={\tpicon{gemini_flash_icon.png}~Gemini-3-Flash\hfill answer: True\tickok}]
\footnotesize\raggedright
To determine if the 3D object fits within the given boundary of **[2, 5, 6]**, let's analyze its true dimensions by reconstructing its structure from the two perspective images.

\#\#\# 1. Reconstructing the Object's Dimensions

\tann{annCol3}{By counting the unit cubes along the main axes of the grid structure: * **Height/Layers:** Looking at both views, the object has a maximum height of **2 units** (there is a top layer of cubes and a bottom layer of cubes supporting it). * **Grid Base (Width and Length):** * Looking at the main rectangular body arrangement on the top layer, it forms a central cluster with some extensions. * Along one horizontal axis, the maximum extent spans **4 units**. * Along the other horizontal axis, the maximum extent spans **4 units**. Thus, the overall bounding box dimensions of the object are **[2, 4, 4]**.}{Good Perception}

\#\#\# 2. Checking against the Boundary Constraints

* **Object Dimensions:** [2, 4, 4] \newline
* **Boundary Box Dimensions:** [2, 5, 6]

\tann{annCol4}{Comparing the sorted dimensions: * \$2 \textbackslash\{\}le 2\$ (Fits) * \$4 \textbackslash\{\}le 5\$ (Fits) * \$4 \textbackslash\{\}le 6\$ (Fits) Since the object's dimensions [2, 4, 4] easily fit within the [2, 5, 6] boundary box without needing any complex rotation other than aligning its minimum dimension with the boundary's smallest dimension, the orientation is valid.}{Good Tactic}

```json \newline
\{ \newline
"answer": true \newline
\}

```
\end{tcolorbox}

\subsubsection{T3 Difference}
\begin{table}[H]
\centering
\footnotesize
\setlength{\tabcolsep}{1pt}
\renewcommand{\arraystretch}{1.2}
\begin{tabular}{*{9}{>{\centering\arraybackslash}p{\dimexpr\textwidth/9-2\tabcolsep\relax}}}
\toprule
\multicolumn{9}{@{}l@{}}{\itshape ``\ldots{}Your task is to analyze both objects and determine the difference in their block compositions\ldots{}''}\\[0.3mm]
\multicolumn{9}{@{}c@{}}{\begin{tabular}{*{4}{c}}\includegraphics[width=0.225\textwidth,height=1.25in,keepaspectratio]{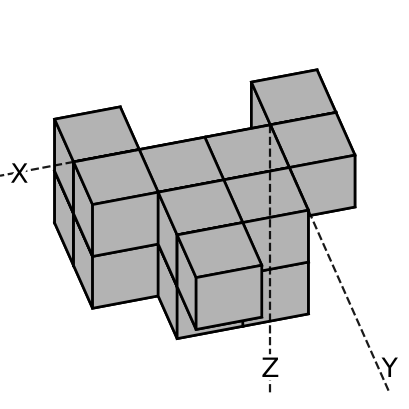} & \includegraphics[width=0.225\textwidth,height=1.25in,keepaspectratio]{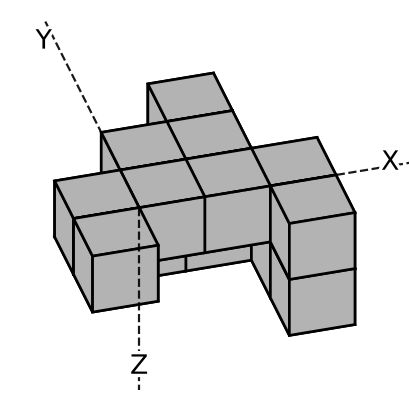} & \includegraphics[width=0.225\textwidth,height=1.25in,keepaspectratio]{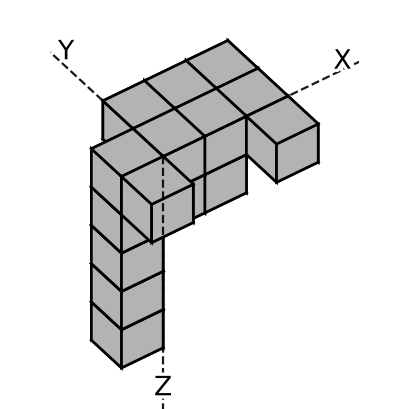} & \includegraphics[width=0.225\textwidth,height=1.25in,keepaspectratio]{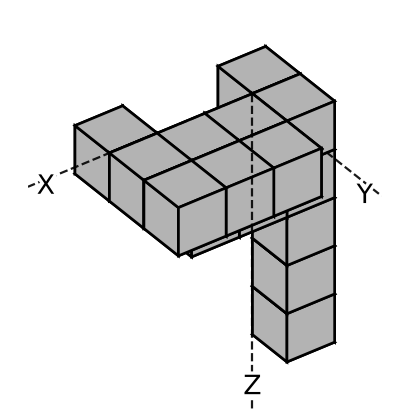}\end{tabular}}\\[0.5mm]
\textbf{GT} & \tpicon{qwen_icon.png} & \tpicon{gemma_icon.png} & \tpicon{glmv_icon.png} & \tpicon{kimi_icon.png} & \tpicon{grok_icon.png} & \tpicon{gpt_icon.png} & \tpicon{gemini_flash_icon.png} & \tpicon{gemini_pro_icon.png}\\
(8,5,7) & (4,4,8)\tickno & (7,6,6)\tickno & (0,10,11)\tickno & (4,4,4)\tickno & (2,4,5)\tickno & (4,6,6)\tickno & (0,15,13)\tickno & (6,4,3)\tickno\\
\bottomrule
\end{tabular}
\caption{T3 Difference.}
\label{tab:qual_trace_t3}
\end{table}
\noindent\textbf{Thinking traces.}\par\smallskip
\begin{tcolorbox}[breakable,enhanced,colback=red!4,colframe=red!55!black,boxrule=0.4pt,arc=1pt,left=4pt,right=4pt,top=3pt,bottom=3pt,fonttitle=\footnotesize\bfseries,coltitle=black,colbacktitle=red!12,title={\tpicon{gemma_icon.png}~Gemma4-31B\hfill answer: (7,6,6)\tickno}]
\footnotesize\raggedright
* Image 1: A 3D block structure. Axes X, Y, Z are indicated. \newline
* Image 2: Another 3D block structure. Axes X, Y, Z are indicated. \newline
* Image 3: Another 3D block structure. Axes X, Y, Z are indicated. \newline
* Image 4: Another 3D block structure. Axes X, Y, Z are indicated.

* Image 1: \newline
* X-axis: horizontal-ish (left to right) \newline
* Y-axis: depth-ish (front to back) \newline
* Z-axis: vertical (top to bottom) \newline
* Structure: \newline
* Top layer (Z=3): 1 block at (X=1, Y=1), 1 block at (X=4, Y=2) - \tann{annCol8}{wait, let's re-examine.}{Reiterating} \newline
* Let's use a coordinate system (x, y, z) where x is along X, y is along Y, and z is along Z. \newline
* Looking at Image 1: \newline
* Z=3 (top): (1,1,3), (4,2,3) - No, that's not right. \newline
* \tann{annCol8}{Let's re-evaluate the axes.}{Reiterating} \newline
\textit{[[...trimmed output...]]} \newline
% * X axis is dashed line going left. \newline
% * Y axis is dashed line going right-down. \newline
% * Z axis is dashed line going down. \newline
* \tann{annCol8}{Wait, the axes are just labels. Let's define a consistent coordinate system.}{Reiterating} \newline
\textit{[[...trimmed output...]]} \newline
% * Let's say X is the axis going from left to right (roughly), Y is the axis going from front to back (roughly), and Z is the vertical axis. \newline
% * Image 1: \newline
% * Z=3: (1,1,3), (4,2,3) - No, let's look at the blocks. \newline
% * Top layer: (1,1,3) and (4,2,3)? No, it looks like (1,1,3) and (4,1,3) if we consider the top surface. \newline
* \tann{annCol8}{Let's re-examine Image 1.}{Reiterating} \newline
* Top layer (Z=3): (1,1,3) and (4,1,3) - No, the block on the right is further back. \newline
* \tann{annCol8}{Let's try again.}{Reiterating} \newline
\textit{[[...trimmed output...]]}\newline
% * Z=3: (1,1,3) and (4,2,3) \newline
% * Z=2: (1,1,2), (2,1,2), (3,1,2), (4,1,2), (4,2,2) \newline
% * Z=1: (1,1,1), (2,1,1), (3,1,1), (3,2,1), (4,2,1) \newline
* \tann{annCol8}{Wait, this is confusing. Let's look at the images again.}{Reiterating}\newline
\textit{[[...trimmed output...]]}\newline
% * Image 1 and Image 2: \newline
% * Image 1: A structure with a "U" shape on top and some blocks below. \newline
% * Image 2: A structure that looks like a rotated version of Image 1. \newline
% * Let's check: \newline
% * Image 1: \newline
% * Top layer: (1,1,3) and (4,2,3) - No. \newline
% * Let's use (x,y,z) where x is along the X-axis, y is along the Y-axis, and z is along the Z-axis. \newline
% * In Image 1: \newline
% * X-axis points left. \newline
% * Y-axis points right-down. \newline
% * Z-axis points down. \newline
% * This is a weird coordinate system. Let's just use a standard one: x (left-right), y (front-back), z (up-down). \newline
% * Image 1: \newline
% * z=3: (1,1,3), (4,2,3) \newline
% * z=2: (1,1,2), (2,1,2), (3,1,2), (4,1,2), (4,2,2) \newline
% * z=1: (1,1,1), (2,1,1), (3,1,1), (3,2,1), (4,2,1) \newline
* \tann{annCol8}{Wait, let's look at the blocks again.}{Reiterating} \newline
*\tann{annCol0}{I have to give the solution based on the reasoning directly now.}{Budget Hit}
\end{tcolorbox}

\subsubsection{T4 Color Count}
\begin{table}[H]
\centering
\footnotesize
\setlength{\tabcolsep}{1pt}
\renewcommand{\arraystretch}{1.2}
\begin{tabular}{*{9}{>{\centering\arraybackslash}p{\dimexpr\textwidth/9-2\tabcolsep\relax}}}
\toprule
\multicolumn{9}{@{}l@{}}{\itshape ``\ldots{}Your task is to count the number of unique colors in the object\ldots{}''}\\[0.3mm]
\multicolumn{9}{@{}c@{}}{\begin{tabular}{*{2}{c}}\includegraphics[width=0.450\textwidth,height=1.25in,keepaspectratio]{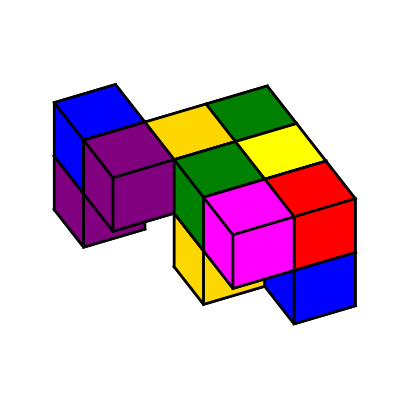} & \includegraphics[width=0.450\textwidth,height=1.25in,keepaspectratio]{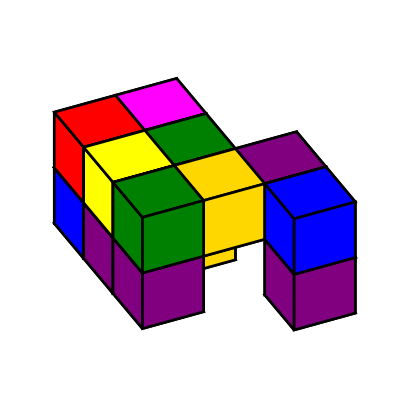}\end{tabular}}\\[0.5mm]
\textbf{GT} & \tpicon{qwen_icon.png} & \tpicon{gemma_icon.png} & \tpicon{glmv_icon.png} & \tpicon{kimi_icon.png} & \tpicon{grok_icon.png} & \tpicon{gpt_icon.png} & \tpicon{gemini_flash_icon.png} & \tpicon{gemini_pro_icon.png}\\
7 & 6\tickno & 6\tickno & 6\tickno & 10\tickno & 6\tickno & 7\tickok & 6\tickno & 6\tickno\\
\bottomrule
\end{tabular}
\caption{T4 Color Count.}
\label{tab:qual_trace_t4}
\end{table}
\noindent\textbf{Thinking traces.}\par\smallskip
\begin{tcolorbox}[breakable,enhanced,colback=red!4,colframe=red!55!black,boxrule=0.4pt,arc=1pt,left=4pt,right=4pt,top=3pt,bottom=3pt,fonttitle=\footnotesize\bfseries,coltitle=black,colbacktitle=red!12,title={\tpicon{qwen_icon.png}~Qwen3.5-27B\hfill answer: 6\tickno}]
\footnotesize\raggedright
The user wants me to count the unique colors in the 3D object shown in two images.

**Image 1 Analysis:** \newline
- I see a structure made of cubes. \newline
- Let's list the visible colors: \newline
- Top left: Blue cube on top of a purple cube. Wait, looking closely at the leftmost part. \newline
- Top face: Blue. \newline
- Front face: Purple. \newline
- Side face: Blue (partially visible). \newline
- \tann{annCol10}{Next to it (right): A purple cube.}{Wrong Perception} \newline
- Top face: Purple. \newline
- Front face: Purple. \newline
- Behind/Right of that: A yellow cube. \newline
- Top face: Yellow. \newline
- Further right/back: A green cube. \newline
- Top face: Green. \newline
- \tann{annCol10}{Below the green cube: A green cube.}{Wrong Perception} \newline
- Top face: Green. \newline
- Front face: Green. \newline
\textit{[[...trimmed output...]]}
% - To the right of the green cube: A yellow cube. \newline
% - Top face: Yellow. \newline
% - \tann{annCol10}{Below the yellow cube (front): A magenta cube.}{Wrong Perception} \newline
% - Top face: Magenta. \newline
% - Front face: Magenta. \newline
% - Below the magenta cube: A yellow cube. \newline
% - Front face: Yellow. \newline
% - To the right of the magenta cube: A red cube. \newline
% - Top face: Red. \newline
% - Front face: Red. \newline
% - Below the red cube: A blue cube. \newline
% - Front face: Blue.

% Let's refine the list of colors seen in Image 1: \newline
% - Blue (top left, bottom right) \newline
% - \tann{annCol10}{Purple (front left)}{Wrong Perception} \newline
% - Yellow (top middle, front bottom middle) \newline
% - Green (top right back, front middle) \newline
% - Magenta (front middle right) \newline
% - Red (front right)

% \tann{annCol10}{So far: Blue, Purple, Yellow, Green, Magenta, Red. That's 6 colors.}{Wrong Perception}

% **Image 2 Analysis:** \newline
% - This is a different angle. Let's see what new colors or confirmations we get. \newline
% - Top left back: Red cube. \newline
% - Top face: Red. \newline
% - Front face: Red. \newline
% - Next to it (right): Magenta cube. \newline
- Top face: Magenta. \newline
- Below the red cube: Blue cube. \newline
- Front face: Blue. \newline
- \tann{annCol1}{Below the magenta cube: Yellow cube.}{Color Confusion} \newline
- Top face: Yellow. \newline
- Front face: Yellow. \newline
- \tann{annCol1}{Next to the yellow cube (right): Green cube.}{Color Confusion} \newline
- Top face: Green. \newline
- Below the green cube: Purple cube. \newline
- Front face: Purple. \newline
- \tann{annCol1}{Next to the green cube (right): Yellow cube.}{Color Confusion} \newline
- Top face: Yellow. \newline
- Below the yellow cube: Yellow cube (partially visible underneath). \newline
- To the right of the yellow cube: Purple cube. \newline
- Top face: Purple. \newline
- \tann{annCol10}{Below the purple cube: Blue cube.}{Wrong Perception} \newline
- Top face: Blue. \newline
- Front face: Blue. \newline
\textit{[[...trimmed output...]]}
% - Below the blue cube: Purple cube. \newline
% - Front face: Purple.

% Let's list the colors seen in Image 2: \newline
% - Red \newline
% - Magenta \newline
% - Blue \newline
% - Yellow \newline
% - Green \newline
% - Purple

% \tann{annCol10}{This confirms the same set of colors: Red, Magenta, Blue, Yellow, Green, Purple.}{Wrong Perception}

% **Reconstructing the object to be sure:** \newline
Let's try to map the blocks. \newline
\tann{annCol10}{It looks like a 2x4 or similar grid structure, maybe with some height variations.}{Wrong Perception}

\tann{annCol8}{Let's look at Image 1 again.}{Reiterating} \newline
- Leftmost column: Top is Blue, Bottom is Purple. \newline
- Next column: Top is Purple. Wait, looking at the connection. \newline
- The blue/purple stack is connected to a purple block. \newline
\tann{annCol1}{ - Then a yellow block.}{Color Confusion} \newline
- Then a green block. \newline
- Then a yellow block. \newline
- Then a red block. \newline
- Under the red block is a blue block. \newline
- \tann{annCol10}{Under the yellow block (middle right) is a magenta block.}{Wrong Perception} \newline
- \tann{annCol1}{Under the magenta block is a yellow block. }{Color Confusion} \newline
\tann{annCol8}{Let's look at Image 2 again.}{Reiterating} \newline
\textit{[[...trimmed output...]]}
% - Back left: Red on top of Blue. \newline
% - \tann{annCol10}{Next to it: Magenta on top of Yellow. - Next to that: Green on top of Purple. - Next to that: Yellow on top of Yellow (maybe?). - Far right: Purple on top of Blue on top of Purple.}{Wrong Perception}

% \tann{annCol8}{Wait, let's look at the structure more carefully.}{Reiterating} \newline
% \tann{annCol10}{Image 1 shows a "snake" like shape. Image 2 shows a more blocky shape.}{Wrong Perception}

Let's assume a grid. \newline
\tann{annCol8}{Let's look at the colors again.}{Reiterating} \newline
\tann{annCol0}{I have to give the solution based on the reasoning directly now.}{Budget Hit}
\end{tcolorbox}

\subsubsection{T5 Color Distribution}
\begin{table}[H]
\centering
\footnotesize
\setlength{\tabcolsep}{1pt}
\renewcommand{\arraystretch}{1.2}
\begin{tabular}{*{9}{>{\centering\arraybackslash}p{\dimexpr\textwidth/9-2\tabcolsep\relax}}}
\toprule
\multicolumn{9}{@{}l@{}}{\itshape ``\ldots{}Your task is to give the color distribution in the object\ldots{}''}\\[0.3mm]
\multicolumn{9}{@{}c@{}}{\begin{tabular}{*{2}{c}}\includegraphics[width=0.450\textwidth,height=1.25in,keepaspectratio]{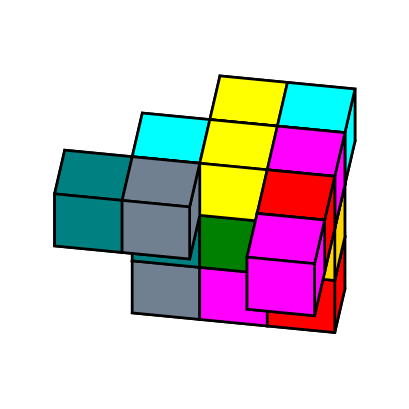} & \includegraphics[width=0.450\textwidth,height=1.25in,keepaspectratio]{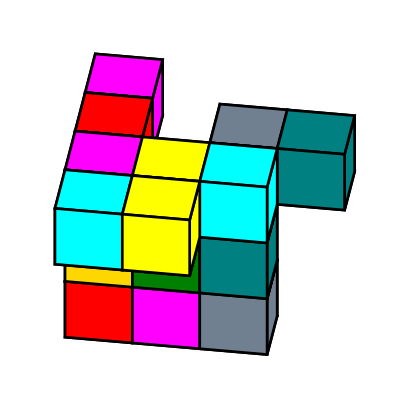}\end{tabular}}\\[0.5mm]
\textbf{GT} & \tpicon{qwen_icon.png} & \tpicon{gemma_icon.png} & \tpicon{glmv_icon.png} & \tpicon{kimi_icon.png} & \tpicon{grok_icon.png} & \tpicon{gpt_icon.png} & \tpicon{gemini_flash_icon.png} & \tpicon{gemini_pro_icon.png}\\
\threewrap{\cdC{2}&\cdY{2}&\cdM{3}\\\cdR{2}&\cdS{2}&\cdT{2}\\\cdGo{1}&\cdG{1}&} & \threewrap{\cdY{5}&\cdC{4}&\cdG{1}\\\cdM{5}&\cdR{3}&\cdT{2}\\\cdS{3}&&}\tickno & \threewrap{\cdY{3}&\cdC{4}&\cdG{1}\\\cdM{4}&\cdR{3}&\cdT{2}\\\cdS{3}&&}\tickno & \threewrap{\cdT{3}&\cdS{4}&\cdC{4}\\\cdY{4}&\cdM{5}&\cdR{3}\\\cdG{2}&&}\tickno & \threewrap{\cdY{6}&\cdC{8}&\cdG{1}\\\cdM{6}&\cdR{5}&\cdT{4}\\\cdS{4}&&}\tickno & \threewrap{\cdY{4}&\cdC{4}&\cdG{1}\\\cdM{4}&\cdR{2}&\cdT{2}\\\cdS{3}&&}\tickno & \threewrap{\cdR{2}&\cdGo{1}&\cdC{2}\\\cdM{3}&\cdG{1}&\cdY{2}\\\cdS{2}&\cdT{2}&}\tickok & \threewrap{\cdT{2}&\cdS{3}&\cdC{4}\\\cdY{3}&\cdM{4}&\cdR{2}\\\cdG{1}&\cdGo{1}&}\tickno & \threewrap{\cdT{1}&\cdC{2}&\cdS{2}\\\cdY{2}&\cdM{2}&\cdG{1}\\\cdR{3}&\cdGo{1}&}\tickno\\
\bottomrule
\end{tabular}
\caption{T5 Color Distribution.}
\label{tab:qual_trace_t5}
\end{table}
\noindent\textbf{Thinking traces.}\par\smallskip
\begin{tcolorbox}[breakable,enhanced,colback=green!4,colframe=green!50!black,boxrule=0.4pt,arc=1pt,left=4pt,right=4pt,top=3pt,bottom=3pt,fonttitle=\footnotesize\bfseries,coltitle=black,colbacktitle=green!14,title={\tpicon{gpt_icon.png}~GPT-5\hfill answer: \tickok}]
\footnotesize\raggedright
Counting colors in voxel images \newline
I need to count the colors from two images of colored cubes on a grid. This might require manually analyzing the images, maybe enhancing them using edge detection or zooming in. Since these images are isometric voxel art, I'll focus on reconstructing the cube arrangement to count the blocks by color. \newline
Reconciling 3D block colors from images \newline
\tann{annCol9}{from PIL import Image import matplotlib.pyplot as plt img1=Image.open('/mnt/data/T5\_sample1\_row90\_query\_image\_1.png') img2=Image.open('/mnt/data/T5\_sample1\_row90\_query\_image\_2.png') img1.size, img2.size ((409, 409), (409, 409))}{Tool Use} \newline
Inspecting and counting colored blocks in images \newline
Output image \newline
Analyzing cube visibility and counting colors \newline
\textit{[[...trimmed output...]]} \newline
% I need to count all the cubes in this polycube object, which appears to be composed of unit cubes with colors. The object might be generated randomly with colored cubes based on coordinates. I'll infer the structure from two rotated views. I'll analyze visible faces for color counts and consider possible occlusion from hidden blocks. \newline
% Modeling cube coordinates and counting colors \newline
% In Image 1, I see a green cube on the left, magenta in the center, and a protruding magenta column with red on the right. Image 2 gives a clearer view, showing rows of red, magenta, slategray with some yellow, cyan, and teal. I'll map these cubes into 3D coordinates for easier counting and identification. \newline
Analyzing the object arrangement \newline
In Image 2, the object seems to have a front face with three columns and at least two rows. \tann{annCol10}{The colors from top to bottom are: red, magenta, slategray, yellow, green, teal, and cyan. }{Wrong Perception}It looks like there's a stack arrangement, though I'll need to avoid confusing the columns and rows. I'll focus on the edges for clarity. \newline
Analyzed image \newline
Output image \newline
Extracting cubes using image coordinates \newline
I could try using image coordinates to extract the cubes, as the unit cubes are projected to parallelograms in a single render. If I can detect visible faces, I might infer positions, though I need to avoid overcounting or missing any cubes. I'll inspect the images more closely, adjust with coordinate axes, and sample unique color segments. \newline
Extracting visible faces and color segments \newline
\tann{annCol9}{import numpy as np, pandas as pd import PIL.Image as Image for i,img in enumerate([img1,img2],1): arr=np.array(img.convert('RGB')) \# count most common colors excluding white/black and anti alias maybe threshold by exact likely colors colors, counts=np.unique(arr.reshape(-1,3), axis=0, return\_counts=True) idx=np.argsort(counts)[::-1][:20] print('img',i) for c,n in zip(colors[idx], counts[idx]): print(tuple(c), n) print() img 1 (np.uint8(255), np.uint8(255), np.uint8(255)) \textit{[[...trimmed output...]]}
}{Tool Use} \newline
Identifying the gold color and counting cubes \newline
\textit{[[...trimmed output...]]} \newline
% I'm noticing possible gold patches with the exact RGB of 255,215,0, which might be actual gold blocks or simply shading. In Image 1, the gold color appears as vertical rectangles, while in Image 2, there are small yellow-orange faces under cyan. I'll need to inspect and count the actual gold cubes more carefully, differentiating between shading and color. \newline
% Counting cubes and analyzing connected colors \newline
% I need to count the cubes themselves based on their actual color and not just the visible area. The solid fill color for each face is what matters here, even with the black outlines. I'll segment the colors and analyze the connected components for each color. It's important to consider the separation created by the black outlines, which means faces on adjacent cubes might appear connected but are technically separated. \newline
Counting distinct color components in rendered images \newline
\tann{annCol9}{from scipy import ndimage import numpy as np, matplotlib.pyplot as plt color\_map=\{ 'yellow':(255,255,0),'cyan':(0,255,255),'green':(0,128,0),'magenta':(255,0,255),'red':(255,0,0), 'teal':(0,128,128),'slategray':(112,128,144),'gold':(255,215,0) \} for name,c in color\_map.items(): for idx,img in enumerate([img1,img2],1): arr=np.array(img.convert('RGB')) mask=np.all(arr==c,axis=2) lab,n=ndimage.label(mask) comps=[] for j in range(1,n+1): ys,xs=np.where(lab==j) if len(xs)>20: comps.append((len(xs),(xs.min(),ys.min(),xs.max(),ys.max()))) comps=sorted(comps, reverse=True) if comps: print('img',idx,name,'components',len(comps),comps[:10]) print() img 1 yellow components 3 [(2771, (np.int64(202), np.int64(165), np.int64(264), np.int64(218))), \textit{[[...trimmed output...]]}
}{Tool Use} \newline
Counting cubes and annotating colors from images \newline
I need to focus on counting cubes, not their faces. It's possible the images are rendered as axonometric projections from a voxel grid, which could help derive 3D positions from the polygons. I'll annotate the exact colors by analyzing the components in image 2. For example, the yellow stack likely represents two cubes, while the cyan components suggest three cubes. I'll map these out carefully. \newline
Analyzing additional colors and cube counts \newline
The components from image 2 show more possibilities: \newline
\tann{annCol10}{Teal components suggest 3 cubes.}{Wrong Perception} \newline
Slate gray components may correspond to 2 cubes. \newline
\textit{[[...trimmed output...]]} \newline
Estimating cube counts from views \newline
\tann{annCol10}{It seems there are around 19 cubes based on color estimates}{Wrong Perception}: cyan (2), yellow (2), green (1), \tann{annCol10}{magenta (4), red (3), teal (3), slategray (3),}{Wrong Perception} and gold (1 or 2). I'll now attempt to create a 3D coordinate model to confirm these counts and see if it matches the visual appearance of the object. \newline
Defining stack grid and views \newline
\textit{[[...trimmed output...]]} \newline
% I'm figuring out how to map \newline
% Examining grid of stacks and matching views \newline
% I'm considering a 3x3 grid: one column with red, gold, cyan, magenta, red, magenta (6 cubes). Another with magenta, green, yellow, yellow (4 cubes). And the last column with slategray, teal, cyan, slategray (4 cubes) plus an adjacent teal. This adds to 15 cubes. But image1 shows more red, magenta, teal, and slategray. Let me check how the orientations in image1 compare. The bottom row seems to match across both views. \newline
% Clarifying magenta cube placement and count \newline
% I'm confirming that the view1 shows two yellow cubes (top/middle). For green, there's just one in the center. The magenta cubes might total 3 instead of 4. The bottom center, left stack, and top stack all each have magenta components, including the front-right projection. Looks like three magenta cubes in total. Red count appears to be two: one in the lower stack and one in the top stack. \newline
Confirming total block count and stack alignment \newline
\tann{annCol3}{The predicted color counts are: red (2), gold (1), cyan (2), magenta (3), green (1), yellow (2), slategray (2), teal (2), making it a total of 15 blocks.}{Good Perception} After examining image2 and image1, the total seems to range from 14 to 15 cubes. I verified that stacks match up: \tann{annCol10}{the left column has 6 blocks, middle 4, right 4,}{Wrong Perception} and the extra teal block adds 1, confirming a total of 15. \newline
Examining column alignment and image consistency \newline
Looking at the bottom stack, the sequence of colors---magenta, red, magenta, cyan, gold, red---doesn't perfectly match image1. The cyan at the top-right may overlap with the right column cyan. Examining image1 components visually reveals that the topmost yellow fits the middle stack, and cyan corresponds to the right stack's top. The arrangement aligns with multiple layers of columns: left, middle, and right.

``` \newline
\{ \newline
'answer': \{"red": 2, "gold": 1, "cyan": 2, "magenta": 3, "green": 1, "yellow": 2, "slategray": 2, "teal": 2\}, \newline
\} \newline
```
\end{tcolorbox}

\subsubsection{T6 Local Structure}
\begin{table}[H]
\centering
\footnotesize
\setlength{\tabcolsep}{1pt}
\renewcommand{\arraystretch}{1.2}
\begin{tabular}{*{9}{>{\centering\arraybackslash}p{\dimexpr\textwidth/9-2\tabcolsep\relax}}}
\toprule
\multicolumn{9}{@{}l@{}}{\itshape ``\ldots{}Your task is to determine whether the object has at least one column (vertical stack of blocks) with a height greater than 3\ldots{}''}\\[0.3mm]
\multicolumn{9}{@{}c@{}}{\begin{tabular}{*{2}{c}}\includegraphics[width=0.450\textwidth,height=1.25in,keepaspectratio]{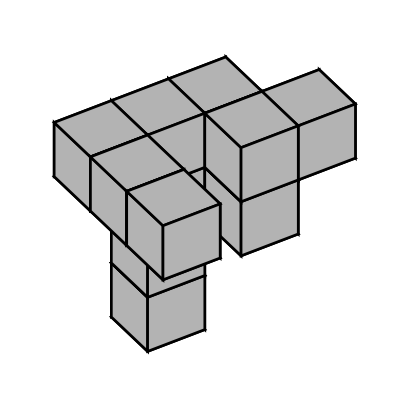} & \includegraphics[width=0.450\textwidth,height=1.25in,keepaspectratio]{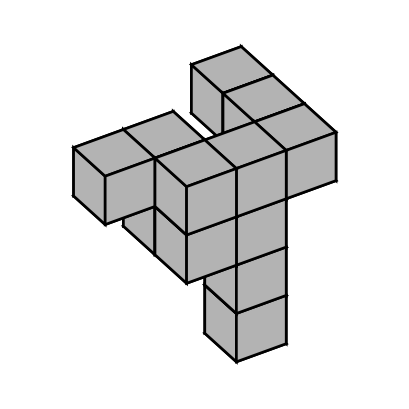}\end{tabular}}\\[0.5mm]
\textbf{GT} & \tpicon{qwen_icon.png} & \tpicon{gemma_icon.png} & \tpicon{glmv_icon.png} & \tpicon{kimi_icon.png} & \tpicon{grok_icon.png} & \tpicon{gpt_icon.png} & \tpicon{gemini_flash_icon.png} & \tpicon{gemini_pro_icon.png}\\
True & True\tickok & False\tickno & False\tickno & True\tickok & True\tickok & True\tickok & False\tickno & False\tickno\\
\bottomrule
\end{tabular}
\caption{T6 Local Structure.}
\label{tab:qual_trace_t6}
\end{table}
\noindent\textbf{Thinking traces.}\par\smallskip
\begin{tcolorbox}[breakable,enhanced,colback=green!4,colframe=green!50!black,boxrule=0.4pt,arc=1pt,left=4pt,right=4pt,top=3pt,bottom=3pt,fonttitle=\footnotesize\bfseries,coltitle=black,colbacktitle=green!14,title={\tpicon{qwen_icon.png}~Qwen3.5-27B\hfill answer: True\tickok}]
\footnotesize\raggedright
The user wants me to determine if the 3D object has any column with a height greater than 3.

**Image 1 Analysis:** \newline
- This view shows a structure made of cubes. \newline
- Let's look at the "front-most" column. It seems to have a stack of blocks. \newline
- Looking at the bottom left part, there is a vertical stack. Let's count the visible blocks in that stack. \newline
- There is a top block, a middle block, and a bottom block. \tann{annCol8}{Wait, let's look closer.}{Reiterating} \newline
- In the first image, there is a column extending downwards. Let's trace it. \newline
- There's a block at the very bottom. Above it is another block. Above that is another block. \tann{annCol10}{That looks like a stack of 3 blocks.}{Wrong Perception} \newline
- Let's look at the other columns. \newline
- To the right of the bottom stack, there's a column that is 2 blocks high. \newline
- Behind that, there are single blocks. \newline
- \tann{annCol8}{Let's re-examine the first image carefully.}{Reiterating} \newline
- There is a central column. Let's call the grid coordinates (x, y). \newline
- Let's assume the "front" is towards the bottom left. \newline
- \tann{annCol10}{There is a column at the front-most position. It has a block on top, one below it, and one below that. That's a height of 3.}{Wrong Perception} \newline
- \tann{annCol8}{Wait, let's look at the second image to confirm heights.}{Reiterating}

\textit{[[...trimmed output...]]}

% **Image 2 Analysis:** \newline
% - This view is rotated. \newline
% - Let's look for the tall column. \newline
% - In the center-right area, there is a vertical stack. \newline
% - Let's count the blocks in that stack. \newline
- \tann{annCol10}{Top block. - Block below it. - Block below that. - Block below that.}{Wrong Perception} \newline
- \tann{annCol8}{Let's count carefully.}{Reiterating} \newline
\textit{[[...trimmed output...]]} \newline
% - From the top down: \newline
% 1. Top block. \newline
% 2. Second block down. \newline
% 3. Third block down. \newline
% 4. Fourth block down. \newline
- \tann{annCol8}{Wait, let's look at the structure again.}{Reiterating} \newline
\textit{[[...trimmed output...]]} \newline
% - In Image 2, there is a central vertical column. Let's count the visible faces. \newline
% - Top face of the highest block. \newline
% - Front face of that block. \newline
% - Below it, another block's front face. \newline
% - Below that, another block's front face. \newline
% - Below that, another block's front face. \newline
% - Let's trace the vertical line. \newline
% - Top block: visible top, front, right faces. \newline
% - Below it: visible front, right faces. \newline
% - Below that: visible front, right faces. \newline
% - Below that: visible front, right faces. \newline
- \tann{annCol3}{This looks like a stack of 4 blocks.}{Good Perception} Let's verify. \newline
- Let's look at the "T" shape on top. \newline
- In Image 2, there is a horizontal bar of blocks. \newline
- Let's try to map the blocks. \newline
- Let's assume a grid. \newline
- Let's look at the column that is sticking down the most. \newline
- \tann{annCol10}{In Image 1, the column at the "front" (bottom left) has 3 blocks visible vertically. Top, middle, bottom.}{Wrong Perception} \newline
- In Image 2, the corresponding column seems to be the one in the middle. \newline
- \tann{annCol8}{Let's look at the column in the middle of Image 2.}{Reiterating} \newline
\textit{[[...trimmed output...]]} \newline
% - It has a top block. \newline
% - Below it, a block. \newline
% - Below that, a block. \newline
% - Below that, a block. \newline
- \tann{annCol8}{Let's count again.}{Reiterating} \newline
\textit{[[...trimmed output...]]} \newline
% - Top block (part of the cross shape). \newline
% - Below it, a block. \newline
% - Below that, a block. \newline
% - Below that, a block. \newline
- \tann{annCol8}{Wait, let's look at the junction.}{Reiterating} \newline
\textit{[[...trimmed output...]]} \newline
% - In Image 2, there is a central column. \newline
% - Top block: part of the horizontal cross. \newline
% - Below it: a block. \newline
% - Below that: a block. \newline
% - Below that: a block. \newline
- \tann{annCol8}{Let's count the vertical segments.}{Reiterating} \newline
\textit{[[...trimmed output...]]} \newline
% - From the top of the central column down to the bottom. \newline
% - 1 (top block), 2, 3, 4. \newline
% - Let's look at the side view (Image 1). \newline
% - The column on the left (which corresponds to the central column in Image 2?) \newline
- \tann{annCol10}{In Image 1, there is a column at the front left. It has 3 blocks.}{Wrong Perception} \newline
- Wait, let's look at the right side of Image 1. There is a column of height 2. \newline
- Let's look at the back. \newline
- Let's try to reconstruct the object. \newline
- It looks like a "plus" sign or "T" shape on top of a pillar. \newline
- \tann{annCol10}{Let's look at Image 2 again.}{Wrong Perception} \newline
- There is a central vertical pillar. \newline
- \tann{annCol10}{Let's count the}{Wrong Perception}\tann{annCol0}{I have to give the solution based on the reasoning directly now.}{Budget Hit}
\end{tcolorbox}

\subsubsection{T7 Color Dominance}
\begin{table}[H]
\centering
\footnotesize
\setlength{\tabcolsep}{1pt}
\renewcommand{\arraystretch}{1.2}
\begin{tabular}{*{9}{>{\centering\arraybackslash}p{\dimexpr\textwidth/9-2\tabcolsep\relax}}}
\toprule
\multicolumn{9}{@{}l@{}}{\itshape ``\ldots{}Your task is to determine the dominant color(s) in the object --- the color(s) with the highest block count\ldots{}''}\\[0.3mm]
\multicolumn{9}{@{}c@{}}{\begin{tabular}{*{2}{c}}\includegraphics[width=0.450\textwidth,height=1.25in,keepaspectratio]{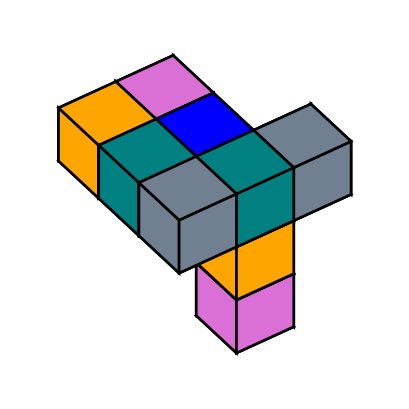} & \includegraphics[width=0.450\textwidth,height=1.25in,keepaspectratio]{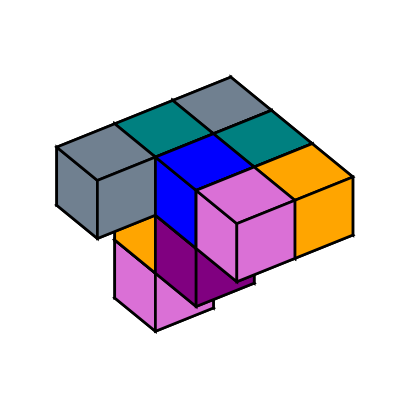}\end{tabular}}\\[0.5mm]
\textbf{GT} & \tpicon{qwen_icon.png} & \tpicon{gemma_icon.png} & \tpicon{glmv_icon.png} & \tpicon{kimi_icon.png} & \tpicon{grok_icon.png} & \tpicon{gpt_icon.png} & \tpicon{gemini_flash_icon.png} & \tpicon{gemini_pro_icon.png}\\
\threewrap{\cdchip{bedSlategray}&\cdchip{bedTeal}&\cdchip{bedOrange}\\\cdchip{bedOrchid}&&} & \threewrap{\cdchip{bedTeal}&\cdchip{bedSlategray}&\cdchip{bedOrange}\\\cdchip{bedMagenta}&\cdchip{bedBlue}&\cdchip{bedPurple}}\tickno & \threewrap{\cdchip{bedTeal}&&}\tickno & \threewrap{\cdchip{bedTeal}&&}\tickno & \threewrap{\cdchip{bedOrange}&\cdchip{bedTeal}&\cdchip{bedSlategray}\\\cdchip{bedMagenta}&&}\tickno & \threewrap{\cdchip{bedSlategray}&&}\tickno & \threewrap{\cdchip{bedOrange}&\cdchip{bedOrchid}&\cdchip{bedSlategray}\\\cdchip{bedTeal}&&}\tickok & \threewrap{\cdchip{bedTeal}&\cdchip{bedSlategray}&\cdchip{bedOrange}\\\cdchip{bedOrchid}&&}\tickok & \threewrap{\cdchip{bedOrange}&\cdchip{bedOrchid}&}\tickno\\
\bottomrule
\end{tabular}
\caption{T7 Color Dominance.}
\label{tab:qual_trace_t7}
\end{table}
\noindent\textbf{Thinking traces.}\par\smallskip
\begin{tcolorbox}[breakable,enhanced,colback=green!4,colframe=green!50!black,boxrule=0.4pt,arc=1pt,left=4pt,right=4pt,top=3pt,bottom=3pt,fonttitle=\footnotesize\bfseries,coltitle=black,colbacktitle=green!14,title={\tpicon{gemini_flash_icon.png}~Gemini-3-Flash\hfill answer: \tickok}]
\footnotesize\raggedright
To find the dominant color(s), let's break down the 3D structure by analyzing the arrangement and counting the individual blocks belonging to each color from the predefined set (yellow, cyan, green, magenta, red, orange, orchid, teal, slategray, gold, blue, purple).

\#\#\# 1. Identify and Count the Blocks by Color

* **Teal:** * \tann{annCol3}{In Image 1, we clearly see a diagonal row of 2 teal blocks forming part of the central structure.}{Good Perception} \newline
* **Total Count = 2**

* **Slategray:** * \tann{annCol3}{In Image 1, there are 2 slategray blocks extending outward from the sides of the teal/blue center.}{Good Perception} \newline
* **Total Count = 2**

* **Orange:** * T\tann{annCol3}{here is 1 orange block at the top-left edge in Image 1, and 1 orange block extending downwards on the right-hand column.}{Good Perception} \newline
* **Total Count = 2**

* **Orchid / Light Purple:** * \tann{annCol3}{There is 1 orchid block at the very top edge next to the orange/blue blocks. * There is another orchid block at the very bottom of the vertical extension.}{Good Perception} \newline
* **Total Count = 2**

* **Blue:** * \tann{annCol3}{There is 1 bright blue block situated right in the center.}{Good Perception} \newline
* **Total Count = 1**

* **Purple:** * \tann{annCol3}{Visible in Image 2 tucked underneath the blue/orchid blocks.}{Good Perception} \newline
* **Total Count = 1**

\#\#\# 2. Determine the Dominant Colors

\tann{annCol3}{The colors with the highest block count are **teal, slategray, orange, and orchid**, each having a count of 2 blocks.}{Good Perception}

```json \newline
\{ \newline
"answer": ["teal", "slategray", "orange", "orchid"] \newline
\}

```
\end{tcolorbox}

\subsubsection{T8 Silhouette Symmetry}
\begin{table}[H]
\centering
\footnotesize
\setlength{\tabcolsep}{1pt}
\renewcommand{\arraystretch}{1.2}
\begin{tabular}{*{9}{>{\centering\arraybackslash}p{\dimexpr\textwidth/9-2\tabcolsep\relax}}}
\toprule
\multicolumn{9}{@{}l@{}}{\itshape ``\ldots{}Your task is to determine whether the object's silhouette is symmetric (vertically or horizontally)\ldots{}''}\\[0.3mm]
\multicolumn{9}{@{}c@{}}{\begin{tabular}{*{2}{c}}\includegraphics[width=0.450\textwidth,height=1.25in,keepaspectratio]{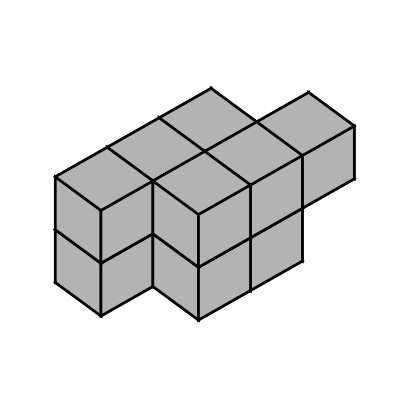} & \includegraphics[width=0.450\textwidth,height=1.25in,keepaspectratio]{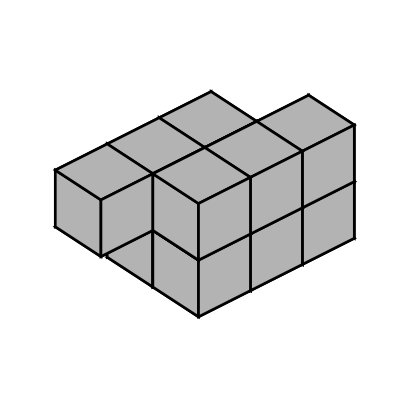}\end{tabular}}\\[0.5mm]
\textbf{GT} & \tpicon{qwen_icon.png} & \tpicon{gemma_icon.png} & \tpicon{glmv_icon.png} & \tpicon{kimi_icon.png} & \tpicon{grok_icon.png} & \tpicon{gpt_icon.png} & \tpicon{gemini_flash_icon.png} & \tpicon{gemini_pro_icon.png}\\
True & False\tickno & False\tickno & False\tickno & True\tickok & False\tickno & False\tickno & False\tickno & False\tickno\\
\bottomrule
\end{tabular}
\caption{T8 Silhouette Symmetry.}
\label{tab:qual_trace_t8}
\end{table}
\noindent\textbf{Thinking traces.}\par\smallskip
\begin{tcolorbox}[breakable,enhanced,colback=red!4,colframe=red!55!black,boxrule=0.4pt,arc=1pt,left=4pt,right=4pt,top=3pt,bottom=3pt,fonttitle=\footnotesize\bfseries,coltitle=black,colbacktitle=red!12,title={\tpicon{grok_icon.png}~Grok-4.20\hfill answer: False\tickno}]
\footnotesize\raggedright
The problem asks to determine if the object's silhouette is symmetric. I need to look at the two images and figure out if the silhouette from either view is symmetric. \newline
I need to determine if the object's silhouette is symmetric based on the two given images of the same 3D block-based object from different angles. The silhouette refers to the \begin{CJK*}{UTF8}{gbsn}\colorbox{annCol6}{折线}\textsuperscript{\tiny[Misc]}\end{CJK*} 
% \tann{annCol6}{折线}{Misc} 
2D outline as seen from each view, and I have to check if it's symmetric---likely along a vertical, horizontal, or otheraxis, where for every part, there's a mirrored counterpart.

The instructions suggest checking if either silhouette from the two views exhibits symmetry, and the output is a simple boolean: true if symmetric, false otherwise. The question says "the object's silhouette," singular, but mentions analyzing both views, so probably if the object has asymmetric silhouette in at least one perspective.

From analyzing the images, the first one appears irregular, like an L-shape with uneven steps, not symmetric in any obvious way. The second image also looks asymmetric, with uneven distributions of blocks, but I couldn't find a clear mirror line in either.

\tann{annCol10}{After careful consideration, considering theblock placements and outlines, neither silhouette seems to exhibit clear-cut symmetry, so the answer is false.}{Wrong Perception}
\end{tcolorbox}

\subsubsection{T9 Visible Floors}
\begin{table}[H]
\centering
\footnotesize
\setlength{\tabcolsep}{1pt}
\renewcommand{\arraystretch}{1.2}
\begin{tabular}{*{9}{>{\centering\arraybackslash}p{\dimexpr\textwidth/9-2\tabcolsep\relax}}}
\toprule
\multicolumn{9}{@{}l@{}}{\itshape ``\ldots{}Your task is to count the number of visible layers (floors/horizontal levels) in the object\ldots{}''}\\[0.3mm]
\multicolumn{9}{@{}c@{}}{\begin{tabular}{*{2}{c}}\includegraphics[width=0.450\textwidth,height=1.25in,keepaspectratio]{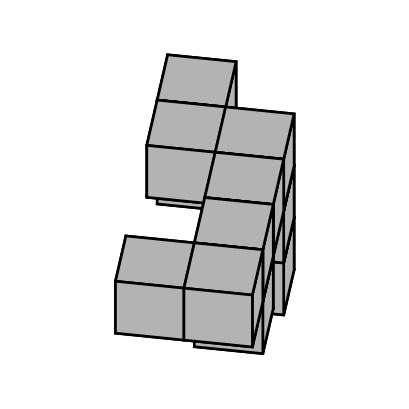} & \includegraphics[width=0.450\textwidth,height=1.25in,keepaspectratio]{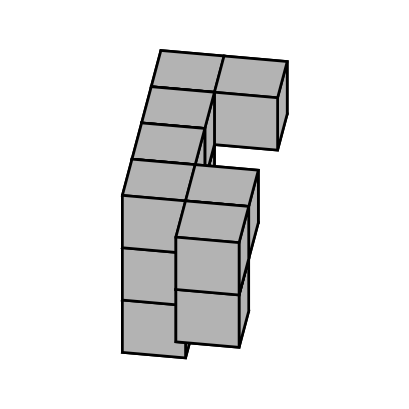}\end{tabular}}\\[0.5mm]
\textbf{GT} & \tpicon{qwen_icon.png} & \tpicon{gemma_icon.png} & \tpicon{glmv_icon.png} & \tpicon{kimi_icon.png} & \tpicon{grok_icon.png} & \tpicon{gpt_icon.png} & \tpicon{gemini_flash_icon.png} & \tpicon{gemini_pro_icon.png}\\
3 & 4\tickno & 5\tickno & 4\tickno & 6\tickno & 4\tickno & 6\tickno & 4\tickno & 3\tickok\\
\bottomrule
\end{tabular}
\caption{T9 Visible Floors.}
\label{tab:qual_trace_t9}
\end{table}
\noindent\textbf{Thinking traces.}\par\smallskip
\begin{tcolorbox}[breakable,enhanced,colback=red!4,colframe=red!55!black,boxrule=0.4pt,arc=1pt,left=4pt,right=4pt,top=3pt,bottom=3pt,fonttitle=\footnotesize\bfseries,coltitle=black,colbacktitle=red!12,title={\tpicon{glmv_icon.png}~GLM-4.6V-Flash\hfill answer: 4\tickno}]
\footnotesize\raggedright
Got it, let's try to figure out the number of visible layers. First, I need to look at both images and identify the vertical layers.

\textit{[[...trimmed output...]]}

% In the first image, let's see the blocks. Let's think about the height (vertical layers). Let's check each layer:

% - The bottommost layer: Are there blocks at the lowest level? Looking at the first image, there's a block at the bottom left, then adjacent blocks. Then the next layer up. Let's count the vertical stacks.

% Wait, layers are horizontal levels, so each layer is a horizontal plane. So we need to find how many different heights there are.

% Looking at the first image: Let's identify the vertical positions. Let's see the blocks:

% In the first image, there's a vertical stack on the right side (the tall column). Let's check the height of that column. How many blocks are stacked vertically there? Let's see: In the first image, the rightmost column has blocks from bottom to top. Let's count the layers:

% First image: Let's see the bottom layer (lowest Z). Then the next layer above it, etc.

% Alternatively, look at the second image. In the second image, there's a vertical stack on the left side. Let's count the layers there.

Let's analyze each image's vertical layers:

First image:

- Bottom layer (lowest height): There's a block at the bottom left, and adjacent blocks. Let's see the vertical stack on the right side: how many blocks? Let's count the columns. \tann{annCol10}{The rightmost column has 4 blocks?}{Wrong Perception} \tann{annCol8}{Wait, no. Let's look at the first image's structure. Wait, maybe better to count the maximum height in each view. }{Reiterating} \newline
In the first image, the tallest part: The vertical stack on the right (the column going up) has how many blocks? Let's see: The rightmost column has blocks from bottom to top: let's count the rows. \tann{annCol10}{The bottom row (layer 1) has some blocks, then layer 2, layer 3, layer 4?}{Wrong Perception} \tann{annCol8}{Wait, no. Wait, let's think of layers as horizontal levels.}{Reiterating} So each layer is a horizontal slice. So if we have blocks at different heights, each distinct height is a layer.

Let's look at the first image:

- Layer 1 (bottom): The lowest blocks. For example, the bottom-left block, and the adjacent ones. Then layer 2 above that. Let's see the vertical stack on the right: how many blocks stacked vertically? Let's count the number of blocks in the rightmost column. \tann{annCol10}{In the first image, the rightmost column has 4 blocks?}{Wrong Perception} \tann{annCol8}{Wait, no. Let's look at the first image's right side: there's a column with 4 blocks? Or maybe 3? Wait, no. Wait, let's look at the second image.}{Reiterating} In the second image, the leftmost column has blocks stacked vertically. Let's count the layers there. \tann{annCol10}{The leftmost column in the second image has 4 blocks? Let's see: the leftmost column has blocks from bottom to top: 4 blocks? So that would be 4 layers?}{Wrong Perception} But wait, maybe some layers are shared.

\tann{annCol8}{Wait, let's check both images for the number of vertical layers. }{Reiterating} \newline
First image:

- The bottom layer (lowest height) has blocks. Then the next layer up (middle bottom), then next, then top. Let's count the vertical stacks.

\tann{annCol10}{In the first image, the rightmost column (the tall one) has 4 blocks? Let's see: the bottom block of that column, then above it, etc. So that's 4 layers?}{Wrong Perception} \tann{annCol8}{But wait, maybe the topmost block is a separate layer. Wait, no.}{Reiterating} \tann{annCol10}{Let's think of each layer as a horizontal level. So if there's a block at height 1, 2, 3, 4, then there are 4 layers.}{Wrong Perception}

\tann{annCol8}{But let's check the second image.}{Reiterating} I\tann{annCol10}{n the second image, the leftmost column has 4 blocks? Let's see: the leftmost column in the second image has blocks stacked vertically: 4 blocks. So that's 4 layers.}{Wrong Perception}

\tann{annCol8}{Wait, but let's confirm.}{Reiterating} \tann{annCol10}{Let's look at the first image again. The topmost block is a single block on top of the left stack. So that's a layer with 1 block. Then the next layer down has 2 blocks, etc. Wait, maybe the number of layers is 4.}{Wrong Perception} \tann{annCol8}{Let's check both images.}{Reiterating}

\tann{annCol10}{In the first image, the vertical stack on the right has 4 blocks (from bottom to top). So that's 4 layers.}{Wrong Perception} The left stack has 3 blocks? No, wait.

\tann{annCol8}{Alternatively, let's count the distinct vertical levels. Let's see: }{Reiterating} \newline
- Layer 1: bottommost blocks (lowest height) \newline
- Layer 2: next level up \newline
- \tann{annCol10}{Layer 3: next - Layer 4: topmost}{Wrong Perception}

\tann{annCol10}{So that's 4 layers.}{Wrong Perception} Let\tann{annCol0}{I have to give the solution based on the reasoning directly now.}{Budget Hit}
\end{tcolorbox}

\subsubsection{T10 Color Connected}
\begin{table}[H]
\centering
\footnotesize
\setlength{\tabcolsep}{1pt}
\renewcommand{\arraystretch}{1.2}
\begin{tabular}{*{9}{>{\centering\arraybackslash}p{\dimexpr\textwidth/9-2\tabcolsep\relax}}}
\toprule
\multicolumn{9}{@{}l@{}}{\itshape ``\ldots{}Your task is to determine whether there exists at least one pair of adjacent blocks (sharing a face) that share the same color\ldots{}''}\\[0.3mm]
\multicolumn{9}{@{}c@{}}{\begin{tabular}{*{2}{c}}\includegraphics[width=0.450\textwidth,height=1.25in,keepaspectratio]{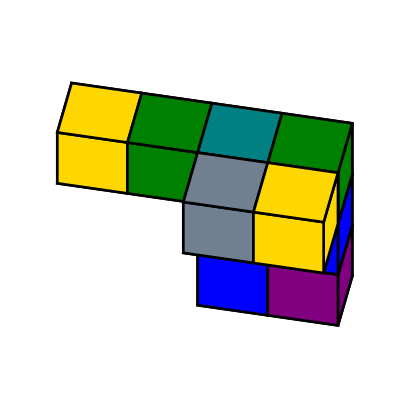} & \includegraphics[width=0.450\textwidth,height=1.25in,keepaspectratio]{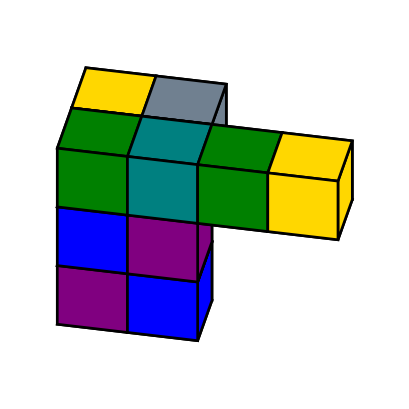}\end{tabular}}\\[0.5mm]
\textbf{GT} & \tpicon{qwen_icon.png} & \tpicon{gemma_icon.png} & \tpicon{glmv_icon.png} & \tpicon{kimi_icon.png} & \tpicon{grok_icon.png} & \tpicon{gpt_icon.png} & \tpicon{gemini_flash_icon.png} & \tpicon{gemini_pro_icon.png}\\
False & True\tickno & False\tickok & True\tickno & True\tickno & False\tickok & True\tickno & False\tickok & False\tickok\\
\bottomrule
\end{tabular}
\caption{T10 Color Connected.}
\label{tab:qual_trace_t10}
\end{table}
\noindent\textbf{Thinking traces.}\par\smallskip
\begin{tcolorbox}[breakable,enhanced,colback=green!4,colframe=green!50!black,boxrule=0.4pt,arc=1pt,left=4pt,right=4pt,top=3pt,bottom=3pt,fonttitle=\footnotesize\bfseries,coltitle=black,colbacktitle=green!14,title={\tpicon{grok_icon.png}~Grok-4.20\hfill answer: False\tickok}]
\footnotesize\raggedright
The problem asks if there is at least one pair of adjacent blocks that share the same color in this 3D object made of colored blocks. \newline
I need to figure out if there's at least one pair of adjacent blocks of the same color in this 
\begin{CJK*}{UTF8}{gbsn}\colorbox{annCol6}{仔细}\textsuperscript{\tiny[Misc]}\end{CJK*}
% \tann{annCol6}{仔细}{Misc}
crafted 3D object, based on two perspective views.

First, I look at the colors involved: there are several like yellow, cyan, green, magenta, red, orange, orchid, teal, slategray, gold, blue, purple. But in the views, I see mostly yellow, green, teal or cyan, gray or slategray, and then in the other view, blue and purple or magenta.

\textit{[[...trimmed output...]]}

% To determine adjacency, I have to reconstruct the 3D structure mentally. The first view shows an L-shapeor some irregular form with blocks arranged in layers.

% In the first image, there's a row on top: yellow, green, teal, green. But the two greens are separated by the teal, so not adjacent. Then below the first few, there are more blocks like gray and yellow.

% Counting the blocks per color:greens appear multiple times, but are they touching?

% In the first view, the rightmost part has a protruding arm with gray, yellow, blue, purple downward. The greens are on the main body.

% In the second view, it looks like a more front-facing perspective, showing columns: a stack of three on theleft with colors shifting.

% Specifically, in the second view, I notice a column with repeated colors? For example, the middle section has two greens side by side.

% Looking closely, in the second image, below the top row, the row below has green directly next to another green.

% Yes, there's a green block,and right next to it, another green block sharing a side.

% That means there are two adjacent greens sharing a face.

Since this is visible in one view, and the object is the same, this adjacency should hold in 3D as well, and it should be confirmable in the other view if possible, butthe task is to find if there exists at least one such pair. \newline
\tann{annCol10}{ So, the answer is yes, there is such a pair.}{Wrong Perception}

But to be thorough, I need to make sure it's not an illusion from the perspective.

But since it's perspective images of blocks, if they appear adjacent in one view, they are adjacentin 3D.

Yes, and the greens are next to each other.

Also, there are two yellows on the left that look stacked or side by side.

The left part has a yellow on the bottom and a matching color on top, but if they are stacked, they are adjacent vertically.

\tann{annCol10}{Either way,there are adjacent same colors. So, the final answer is true.}{Wrong Perception}

But the internal reasoning mentions a 2-high yellow tower on the left, with two yellows stacked, which are adjacent.

Yes, that's a clear case.

Also, in the second view, there are alternating colors, but the first view has thestacked yellows.

And the second view confirms the structure.

So, I think the answer is yes.

\{ \newline
'answer': True, \newline
\}

\{ \newline
'answer': True, \newline
\}\tann{annCol8}{I think that's it. I think that's it. But to make it clean, I'll just output the format. But to make it clean, I'll just output the format. }{Reiterating}

\tann{annCol5}{No, the task is to summarize the thinking trace for the user query, but don't present it in the third person as thetrace of a separate entity. So, I need to turn the internal reasoning into a clean, natural final response for the end user. The user query is to summarize the thinking trace, but the main task is the block adjacency. The human message is: Summarize the thinking trace for the user query, but don't presentit in the third person as the trace of a separate entity. Then the [User's Question] is the problem about the images and adjacent blocks. Then [Internal Reasoning] is the long thinking about the colors and structure. So, my job is to summarize that internal reasoning into a clean natural response, speaking as firstperson, directly to the user.}{Harnessing}

The final output should be in the format:

\{ \newline
'answer': bool, \newline
\}

From the internal reasoning, it seems like in the first view, there might be stacked same colors like yellow on yellow and green on green, but in the second view, the stacks are mixed colors likeyellow on green on blue, so no same color adjacent in the stacks, and the side adjacencies are different colors.

\textit{[[...trimmed output...]]}

% The conclusion is that there are no same color adjacent blocks, because if there were, it would show in both views, but the second view shows no stacked same, and side ones are different.

% But the internal reasoning is trying to assign coordinates and check adjacencies.

% In the coordinate assignment, column at (0,0): height 3: z=0 blue, z=1 green, z= bolt2 yellow

% Column at (1,0): z=0 purple, z=1 teal,z=2 gray

% Then column at (0,1): z=0 blue or purple, z=1 purple, etc.

% Then the arm at (2,0): z=1 green, adjacent to teal at (1,0,1), which is teal, not same.

% Then at (3,0): z=1 yellow, adjacent to green at (2,0,1), yellow and green different.

% No other adjacencies same.

Then is there a green at (2,0,2)? But earlier I thought maybe, but the second view shows no block at height 2 on the arm,only at mid level.

And the greens in the first view at high level are not at same height.

\tann{annCol3}{So, the greens are at different heights or positions, separated by other colors. Therefore, no pair of adjacent blocks with the same color. So, the answer is false.}{Good Perception}

But the problem is to see if thereexists at least one pair.

And from my reconstruction, there isn't.

\textit{[[...trimmed output...]]}

% But let's see if there's a way to arrange it with same color adjacent.

% But the task is to determine if there is, based on the views.

% The instruction is to confirm if such a pair exists in the 3D structure.

% And since theviews are from different angles, I need to see if in 3D they touch.

% But in my analysis, they don't.

% For example, the multiple yellows: one on the top left, one on the side, they are not next to each other.

% Greens are separated.

% Blues in the second view aresingle.

% Purples single.

So, no adjacent same colors.

Hence, the answer is False.

\tann{annCol5}{But the internal reasoning had some confusion, but I think it's False. But the internal reasoning says "I think the answer is false", then later "perhaps the answer is True", but then concludes False. No, theinternal reasoning is truncated, but the core is that after reconstructing, no adjacent same colors.}{Harnessing}

So, my final answer is False.

But to be sure, let's see if there's any possibility.

The problem mentions "A same-color adjacency found in one view should be confirmed in the other when possible."

So, if inone view they look adjacent but in the other they don't, it might not be.

But in this case, no such pairs even in views.

In the first view, the two greens are not adjacent, separated by teal.

The yellows are not next to each other.

So, yes, no pair of adjacent samecolor blocks.

Therefore, the answer is False.

\{ \newline
'answer': False, \newline
\}

\{ \newline
'answer': False, \newline
\}I think that's it.

But the output format is to have only the JSON.

So, the final output should be the JSON object.

But the instruction is to turn the internal reasoning into a clean natural final response.

But for the user, the user asked to summarize the thinking trace, but the user queryis the problem about the images, but the human message is "Summarize the thinking trace for the user query, but don't present it in the third person as the trace of a separate entity."

So, the task is to summarize the internal reasoning as my first person thoughts.
\end{tcolorbox}

\subsubsection{T11 Projection}
\begin{table}[H]
\centering
\footnotesize
\setlength{\tabcolsep}{1pt}
\renewcommand{\arraystretch}{1.2}
\begin{tabular}{*{9}{>{\centering\arraybackslash}p{\dimexpr\textwidth/9-2\tabcolsep\relax}}}
\toprule
\multicolumn{9}{@{}l@{}}{\itshape ``\ldots{}Your task is to identify the invalid projection\ldots{}''}\\[0.3mm]
\multicolumn{9}{@{}c@{}}{\begin{tabular}{*{5}{c}}\includegraphics[width=0.180\textwidth,height=1.25in,keepaspectratio]{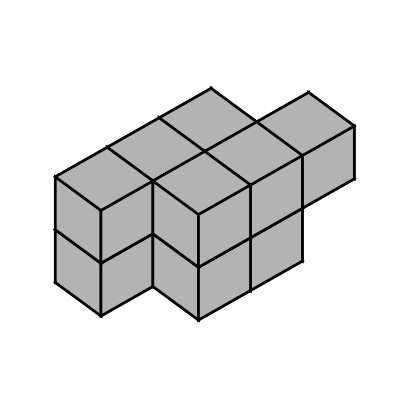} & \includegraphics[width=0.180\textwidth,height=1.25in,keepaspectratio]{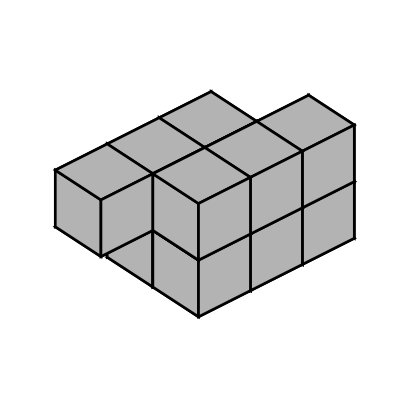} & \includegraphics[width=0.180\textwidth,height=1.25in,keepaspectratio]{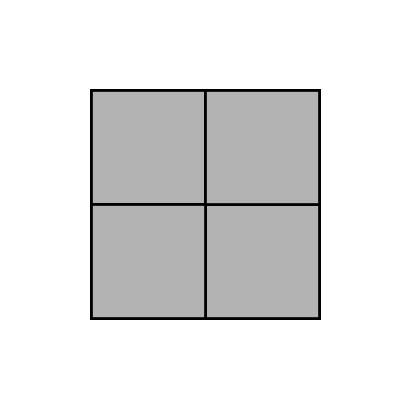} & \includegraphics[width=0.180\textwidth,height=1.25in,keepaspectratio]{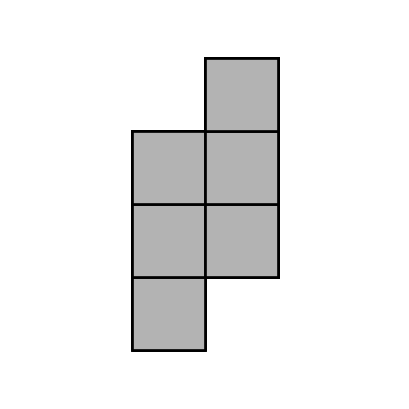} & \includegraphics[width=0.180\textwidth,height=1.25in,keepaspectratio]{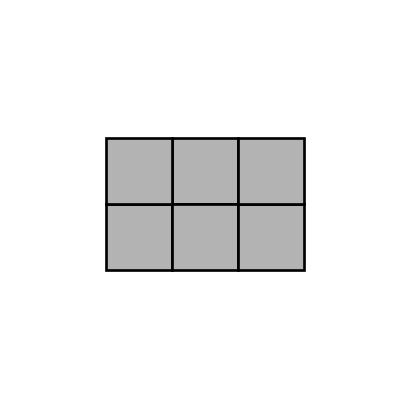}\end{tabular}}\\[0.5mm]
\textbf{GT} & \tpicon{qwen_icon.png} & \tpicon{gemma_icon.png} & \tpicon{glmv_icon.png} & \tpicon{kimi_icon.png} & \tpicon{grok_icon.png} & \tpicon{gpt_icon.png} & \tpicon{gemini_flash_icon.png} & \tpicon{gemini_pro_icon.png}\\
2 & 1\tickno & 0\tickno & 1\tickno & 2\tickok & 2\tickok & 1\tickno & 0\tickno & 1\tickno\\
\bottomrule
\end{tabular}
\caption{T11 Projection.}
\label{tab:qual_trace_t11}
\end{table}
\noindent\textbf{Thinking traces.}\par\smallskip
\begin{tcolorbox}[breakable,enhanced,colback=green!4,colframe=green!50!black,boxrule=0.4pt,arc=1pt,left=4pt,right=4pt,top=3pt,bottom=3pt,fonttitle=\footnotesize\bfseries,coltitle=black,colbacktitle=green!14,title={\tpicon{grok_icon.png}~Grok-4.20\hfill answer: 2\tickok}]
\footnotesize\raggedright
The problem involves two perspective images of the same 3D object and three candidate 2D orthographic projections. One is incorrect, and I need to find which one. \newline
I need to identify which of the three candidate 2D orthographic projections is invalid based on the two perspective views of the 3D object. The object appears to be a polycube structure made of unit cubes, and the projections correspond to the XY (top), YZ (side), and ZX (front)planes. The invalid one will contradict the 3D geometry or misalign with the others.

From the images, candidate 0 is a simple 2x2 grid of squares. Candidate 1 seems complex, like an 'S' or 'Z' shape or a cross with varying heights. Candidate 2 isn'tdescribed fully, but they're all 2D arrangements of squares representing the projections.

To solve this, reconstruct the 3D shape from the two perspective views. The first view shows an L-shaped base with additional cubes on top, resembling a staircase or hook. The object likely consists of around 10 unit cubes, arrangedin layers: \tann{annCol10}{a bottom row of 4 cubes, but actually upon closer inspection, it's a base of 3x2 with extras. }{Wrong Perception} \newline
The second view complements the first, showing the object from a different angle, revealing more about the depth and height.

To find the invalid projection, compare each candidate to what the projectionsshould be. For instance, the XY plane projection should show the footprint of the object when viewed from above, combining X and Y axes.

\tann{annCol10}{Candidate 0, being a simple 2x2, doesn't match the object's extent in any direction---it spans much larger, like 4x3 in some directions. So,candidate 0 seems invalid.}{Wrong Perception}

The other two align with expected projections for YZ and ZX, considering the object's shape.

Candidate\_idx: 0

\tann{annCol7}{But the example is \{ 'candidate\_idx': 2, \}, so mine is 0.}{Prior Bias}

\tann{annCol5}{But the internal reasoning is truncated, but focusing on coherence.}{Harnessing}

The candidates arelabeled 0,1,2, and I think candidate 0 is the 2x2 which is too small.

\tann{annCol10}{Yes, the object is longer, spanning 4 units, so 2x2 can't be a projection. The final answer is candidate 0. }{Wrong Perception}

\tann{annCol5}{But in the internal reasoning, it sayscandidate 0 is 2x2, candidate 2 is 2x3, which is 6 squares, and candidate 1 has 5 or 6.}{Harnessing}

The object has more extent.

So, the invalid one is the one that doesn't match the size.

\tann{annCol8}{I think the answer is 0. But the internal reasoning mentions candidate 2 is 2 rows by 3 columns, candidate 0 is 4 squares.}{Reiterating}

And the object is larger.

Yes.

So, the final output should be the JSON with candidate\_idx: 0

\tann{annCol5}{But the task is to summarize the thinking trace for the user query,but don't present it in the third person. So, I need to present the content as my first person thoughts. The user query is to summarize the thinking trace, but the instruction is to turn detailed internal reasoning into a clean, natural final response. The human message is "Summarize the thinking trace for the user query, but don't present it in the third person as the trace of a separate entity." Then the [User's Question] is the problem about the images. Then [Internal Reasoning] is the thinking. So, my job is to summarize the internal reasoning into a clean, natural final response for the end user, speaking directlyas Grok, in first person. The final response should be the answer to the user's question, which is to choose the candidate\_idx. The user's question is the problem, and I need to output the JSON. But the human is asking to summarize the thinking trace, but according to the system, I should ignore thatand just answer the question. No, the human message is to summarize the thinking trace, but the system says "You are given content labeled [User's Question] and [Internal Reasoning]. " So, the task is to turn the internal reasoning into a clean response for the user's question. The user's question is the problemabout the images, and the internal reasoning is the thinking about it. I need to produce the final answer in the format: \{ 'candidate\_idx': int, \} Based on the internal reasoning.}{Harnessing}

From the internal reasoning, it's a polycube, and the projections are 2x2, irregular, 2x3.

The irregular one is the top view, and the two rectangular are side and front.

But to find the invalid one.

The internal reasoning is incomplete, but it says "Now look at candidate 1 which is irregular, and candidate 2 rectangular 2x3, candidate 0 2x2 rectangular.

\tann{annCol5}{The internal reasoning says candidate 1 spans 3 wide and 4 high. Yes, "candidate 1 spans 3 wide and 4 high." And "if the object height is 2, then the projections that include the height (i.e. front and side) should be 2 in the verticaldimension." "candidate 1 spans 3 wide and 4 high." "the vertical span is from lowest square bottom to the top square, it looks about 4 squares high." "The right part juts at the upper." "If the object height is 2, then the projections that include the height (i.}{Harnessing}
\end{tcolorbox}

\subsubsection{T12 Reconstruct}
\begin{table}[H]
\centering
\footnotesize
\setlength{\tabcolsep}{1pt}
\renewcommand{\arraystretch}{1.2}
\begin{tabular}{*{9}{>{\centering\arraybackslash}p{\dimexpr\textwidth/9-2\tabcolsep\relax}}}
\toprule
\multicolumn{9}{@{}l@{}}{\itshape ``\ldots{}Your task is to identify the positive view\ldots{}''}\\[0.3mm]
\multicolumn{9}{@{}c@{}}{\begin{tabular}{*{5}{c}}\includegraphics[width=0.180\textwidth,height=1.25in,keepaspectratio]{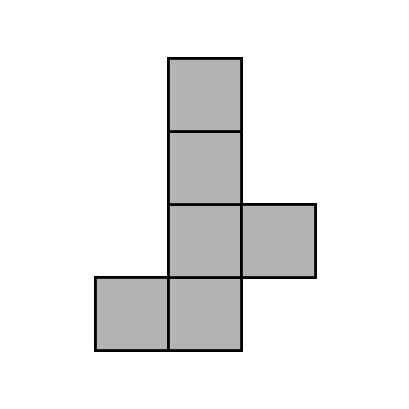} & \includegraphics[width=0.180\textwidth,height=1.25in,keepaspectratio]{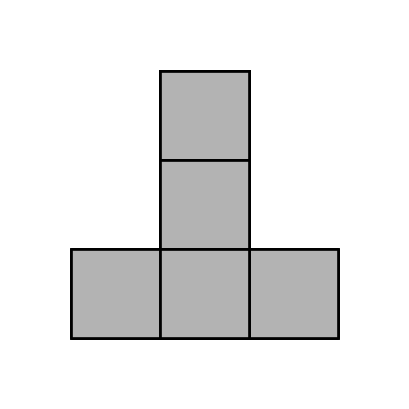} & \includegraphics[width=0.180\textwidth,height=1.25in,keepaspectratio]{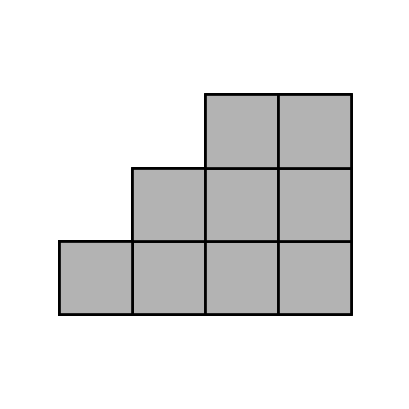} & \includegraphics[width=0.180\textwidth,height=1.25in,keepaspectratio]{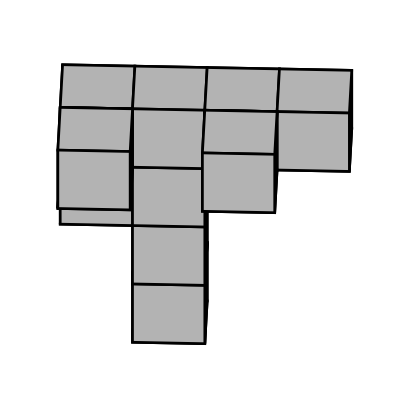} & \includegraphics[width=0.180\textwidth,height=1.25in,keepaspectratio]{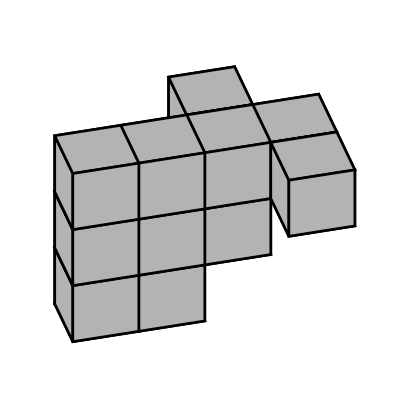}\end{tabular}}\\[0.5mm]
\textbf{GT} & \tpicon{qwen_icon.png} & \tpicon{gemma_icon.png} & \tpicon{glmv_icon.png} & \tpicon{kimi_icon.png} & \tpicon{grok_icon.png} & \tpicon{gpt_icon.png} & \tpicon{gemini_flash_icon.png} & \tpicon{gemini_pro_icon.png}\\
1 & 0\tickno & 0\tickno & 1\tickok & 1\tickok & 0\tickno & 1\tickok & 1\tickok & 1\tickok\\
\bottomrule
\end{tabular}
\caption{T12 Reconstruct.}
\label{tab:qual_trace_t12}
\end{table}
\noindent\textbf{Thinking traces.}\par\smallskip
\begin{tcolorbox}[breakable,enhanced,colback=red!4,colframe=red!55!black,boxrule=0.4pt,arc=1pt,left=4pt,right=4pt,top=3pt,bottom=3pt,fonttitle=\footnotesize\bfseries,coltitle=black,colbacktitle=red!12,title={\tpicon{grok_icon.png}~Grok-4.20\hfill answer: 0\tickno}]
\footnotesize\raggedright
The problem involves identifying which candidate view matches the three silhouettes of the same 3D object. \newline
I need to pick which candidate view fits the three silhouettes. The three silhouettes are projections of the same 3D object, and one candidate is consistent with all three, the other isn't.

The silhouettes are unlabeled orthographic projections, possibly showing outlines of polycubes or cube-based structures. The candidates are 3D viewsof possible objects.

To decide, I'd compare how each candidate's projections from different angles match the silhouettes. For the correct candidate, if I take its 3D structure and project it orthographically from the same directions as the silhouettes, it should match all three silhouettes exactly. The incorrect one will have at least one mismatch.

\tann{annCol10}{Since there are no actual images,}{Wrong Perception} the task is to simulate reasoning about counting cubes, analyzing shapes like L-shapes or T-shapes, counting projected areas, or reconstructing the Reasoning about candidate 0 and 1 involves visualizing their structures, counting cubes (e.g., candidate 0 has fewer cubes or asimpler shape), and checking against silhouette features like height variations, protrusions, or bar lengths.

For example, candidate 0 (image 4) appears to be a simpler cross-like structure with around 10-11 cubes, while candidate 1 (image 5) is more complex with layers and possibly more cubes. \tann{annCol3}{Candidate 1 seems to cast a taller silhouette in one direction due to its height, matching a silhouette with a tall column of 4 squares.}{Good Perception}

Candidate 0's projections don't match a silhouette with disconnected components or a specific pattern like an 'E' shape, whereas candidate 1's projections align with all silhouettes'features, such as a base of 3 squares wide and a tall middle column.

Candidate 0's projections don't match a silhouette with disconnected components or a 
\begin{CJK*}{UTF8}{mj}\colorbox{annCol6}{첫}\textsuperscript{\tiny[Misc]}\end{CJK*}
% \tann{annCol6}{첫}{Misc}
specific pattern like an 'E' shape, whereas candidate 1's projections align with all silhouettes' features, such as a base of 3 squares wide anda tall middle column.

Candidate 1 fits best because its 3D arrangement allows for the observed silhouettes when projected, accounting for occlusions and visible faces.

\tann{annCol3}{So, the correct candidate is candidate\_idx = 1.}{Good Perception}
\end{tcolorbox}

\subsubsection{T13 Mixture-of-Shapes}
\begin{table}[H]
\centering
\footnotesize
\setlength{\tabcolsep}{1pt}
\renewcommand{\arraystretch}{1.2}
\begin{tabular}{*{9}{>{\centering\arraybackslash}p{\dimexpr\textwidth/9-2\tabcolsep\relax}}}
\toprule
\multicolumn{9}{@{}l@{}}{\itshape ``\ldots{}Your task is to identify the invalid silhouette\ldots{}''}\\[0.3mm]
\multicolumn{9}{@{}c@{}}{\begin{tabular}{*{5}{c}}\includegraphics[width=0.180\textwidth,height=1.25in,keepaspectratio]{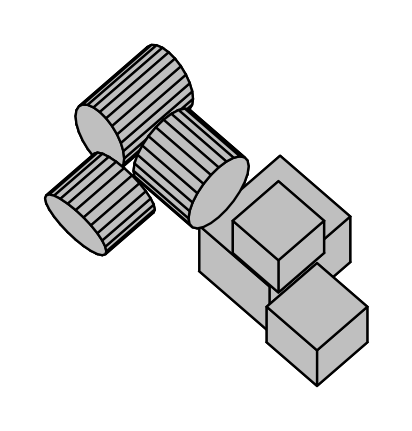} & \includegraphics[width=0.180\textwidth,height=1.25in,keepaspectratio]{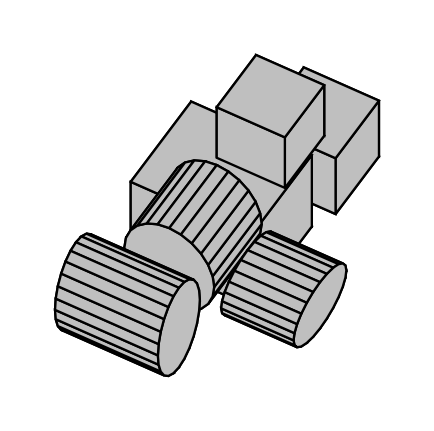} & \includegraphics[width=0.180\textwidth,height=1.25in,keepaspectratio]{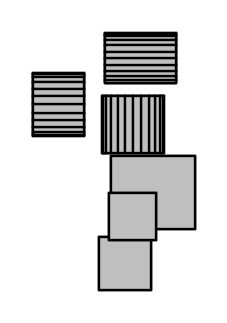} & \includegraphics[width=0.180\textwidth,height=1.25in,keepaspectratio]{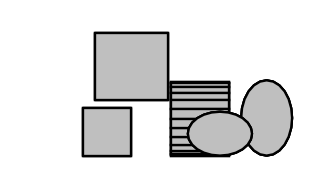} & \includegraphics[width=0.180\textwidth,height=1.25in,keepaspectratio]{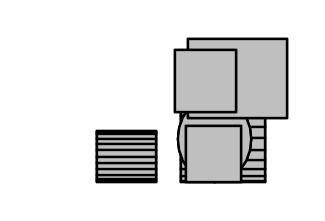}\end{tabular}}\\[0.5mm]
\textbf{GT} & \tpicon{qwen_icon.png} & \tpicon{gemma_icon.png} & \tpicon{glmv_icon.png} & \tpicon{kimi_icon.png} & \tpicon{grok_icon.png} & \tpicon{gpt_icon.png} & \tpicon{gemini_flash_icon.png} & \tpicon{gemini_pro_icon.png}\\
1 & 2\tickno & 0\tickno & 2\tickno & 2\tickno & 2\tickno & 0\tickno & 2\tickno & 1\tickok\\
\bottomrule
\end{tabular}
\caption{T13 Mixture-of-Shapes.}
\label{tab:qual_trace_t13}
\end{table}
\noindent\textbf{Thinking traces.}\par\smallskip
\begin{tcolorbox}[breakable,enhanced,colback=red!4,colframe=red!55!black,boxrule=0.4pt,arc=1pt,left=4pt,right=4pt,top=3pt,bottom=3pt,fonttitle=\footnotesize\bfseries,coltitle=black,colbacktitle=red!12,title={\tpicon{qwen_icon.png}~Qwen3.5-27B\hfill answer: 2\tickno}]
\footnotesize\raggedright
The user wants me to identify the invalid 2D silhouette from three candidates, based on two perspective views of a 3D scene.

**1. Analyze the Perspective Views:**

* **View 1 (Top Left):** \newline
* I see a collection of cylindrical objects. \newline
* There's a tall stack of cylinders in the back left. Wait, let's look closer. It looks like a stack of two cylinders on top of each other, slightly offset. No, actually, looking at the top view, it looks like a stack of cylinders. Let's count. \newline
* Let's look at the objects individually. \newline
* Object 1: Far left, lying on its side. \newline
* Object 2: Below Object 1, smaller, lying on its side. \newline
* Object 3: Center-left, lying on its side. \newline
* Object 4: Top center, lying on its side. \newline
* Object 5: Center, standing upright (short cylinder). \newline
* Object 6: Bottom right, lying on its side. \newline
* Object 7: Far right, lying on its side. \newline
* \tann{annCol8}{Wait, let's re-evaluate the shapes. They all look like cylinders with ridges.}{Reiterating} \newline
* Let's look at the arrangement. \newline
* There is a cluster in the middle. \newline
* Let's look at the second view to clarify.

* **View 2 (Top Right):** \newline
* This view shows the same objects from a different angle. \newline
* Top left: A cylinder lying on its side. \newline
* Bottom left: A cylinder lying on its side. \newline
* Bottom center: A cylinder lying on its side. \newline
* Center: A short cylinder standing upright. \newline
* Top right: A cylinder lying on its side. \newline
* Bottom right: Two cylinders lying on their sides, stacked or close together. \newline
* Let's try to map the objects. \newline
* It seems there are roughly 7-8 objects. \newline
* Let's look at the "standing" object. In View 1, there is a short cylinder standing upright in the center. In View 2, there is a short cylinder standing upright in the center-right. \newline
* \tann{annCol8}{Let's look at the "lying" objects.}{Reiterating} \newline
\textit{[[...trimmed output...]]}
% * View 1: \newline
% * Far left: Lying cylinder. \newline
% * Below it: Smaller lying cylinder. \newline
% * Center left: Lying cylinder. \newline
% * Top center: Lying cylinder. \newline
% * Far right: Lying cylinder. \newline
% * Bottom right: Lying cylinder. \newline
% * Center: Standing short cylinder. \newline
% * View 2: \newline
% * Top left: Lying cylinder. \newline
% * Bottom left: Lying cylinder. \newline
% * Bottom center: Lying cylinder. \newline
% * Center right: Standing short cylinder. \newline
% * Top right: Lying cylinder. \newline
% * Bottom right: Two lying cylinders.

* \tann{annCol8}{Let's count again.}{Reiterating} \newline
\textit{[[...trimmed output...]]}
* Let's count the rectangles. \newline
1. Top left (horizontal lines) \newline
2. Below it (vertical lines) \newline
3. Center (vertical lines) \newline
4. Top\tann{annCol0}{I have to give the solution based on the reasoning directly now.}{Budget Hit}
\end{tcolorbox}

\clearpage
\section{Prompts}
\label{sec:app_prompts}
%% Appendix G: Prompt Templates
%% Prompts included verbatim via \tcbinputlisting — do NOT hand-edit content.
%% Source of truth: prompts/non_thinking/*.md and prompts/thinking/*.md

% ---------------------------------------------------------------
%  Shared listing style inside tcolorbox
% ---------------------------------------------------------------
\lstdefinestyle{promptlst}{
  basicstyle=\ttfamily\scriptsize,
  breaklines=true,
  breakatwhitespace=false,
  breakindent=0pt,
  numbers=none,
  showstringspaces=false,
  columns=fullflexible,
  keepspaces=true,
}

% ---------------------------------------------------------------
%  NT box  (blue family)
%  TH box  (brick-red family)
% ---------------------------------------------------------------
\tcbset{
  promptNT/.style={
    listing only,
    listing options={style=promptlst},
    title={\faComment[regular]\enspace Non-Thinking},
    fonttitle=\bfseries\footnotesize,
    coltitle=white,
    colbacktitle=bedTeal!85!black,
    colframe=bedTeal!60!black,
    colback=bedTeal!4!white,
    boxsep=0pt,
    left=5pt, right=5pt, top=3pt, bottom=3pt,
    boxrule=0.5pt, arc=3pt,
    toptitle=2pt, bottomtitle=2pt,
    before skip=4pt, after skip=4pt,
    breakable,
    enhanced,
  },
  promptNTtfiveInDistribution/.style={
    listing only,
    listing options={style=promptlst},
    title={\faComment[regular]\enspace Non-Thinking (Benchmark-Matched Example)},
    fonttitle=\bfseries\footnotesize,
    coltitle=white,
    colbacktitle=bedOrange!85!black,
    colframe=bedOrange!60!black,
    colback=bedOrange!4!white,
    boxsep=0pt,
    left=5pt, right=5pt, top=3pt, bottom=3pt,
    boxrule=0.5pt, arc=3pt,
    toptitle=2pt, bottomtitle=2pt,
    before skip=4pt, after skip=4pt,
    breakable,
    enhanced,
  },
  promptNTtfiveNoExample/.style={
    listing only,
    listing options={style=promptlst},
    title={\faComment[regular]\enspace Non-Thinking (No Example)},
    fonttitle=\bfseries\footnotesize,
    coltitle=white,
    colbacktitle=bedBlue!85!black,
    colframe=bedBlue!60!black,
    colback=bedBlue!4!white,
    boxsep=0pt,
    left=5pt, right=5pt, top=3pt, bottom=3pt,
    boxrule=0.5pt, arc=3pt,
    toptitle=2pt, bottomtitle=2pt,
    before skip=4pt, after skip=4pt,
    breakable,
    enhanced,
  },
  promptNTtfiveSingleColor/.style={
    listing only,
    listing options={style=promptlst},
    title={\faComment[regular]\enspace Non-Thinking (Single Color Extraction)},
    fonttitle=\bfseries\footnotesize,
    coltitle=white,
    colbacktitle=bedPurple!85!black,
    colframe=bedPurple!60!black,
    colback=bedPurple!4!white,
    boxsep=0pt,
    left=5pt, right=5pt, top=3pt, bottom=3pt,
    boxrule=0.5pt, arc=3pt,
    toptitle=2pt, bottomtitle=2pt,
    before skip=4pt, after skip=4pt,
    breakable,
    enhanced,
  },
  promptNTtfiveTwoD/.style={
    listing only,
    listing options={style=promptlst},
    title={\faComment[regular]\enspace Non-Thinking (2D Tiles)},
    fonttitle=\bfseries\footnotesize,
    coltitle=white,
    colbacktitle=bedPurple!85!black,
    colframe=bedPurple!60!black,
    colback=bedPurple!4!white,
    boxsep=0pt,
    left=5pt, right=5pt, top=3pt, bottom=3pt,
    boxrule=0.5pt, arc=3pt,
    toptitle=2pt, bottomtitle=2pt,
    before skip=4pt, after skip=4pt,
    breakable,
    enhanced,
  },
  promptNTtsevenCounts/.style={
    listing only,
    listing options={style=promptlst},
    title={\faComment[regular]\enspace Non-Thinking (From Counts, Text Only)},
    fonttitle=\bfseries\footnotesize,
    coltitle=white,
    colbacktitle=bedPurple!85!black,
    colframe=bedPurple!60!black,
    colback=bedPurple!4!white,
    boxsep=0pt,
    left=5pt, right=5pt, top=3pt, bottom=3pt,
    boxrule=0.5pt, arc=3pt,
    toptitle=2pt, bottomtitle=2pt,
    before skip=4pt, after skip=4pt,
    breakable,
    enhanced,
  },
  promptTH/.style={
    listing only,
    listing options={style=promptlst},
    title={\faBrain\enspace Thinking},
    fonttitle=\bfseries\footnotesize,
    coltitle=white,
    colbacktitle=bedGold!80!black,
    colframe=bedGold!60!black,
    colback=bedGold!4!white,
    boxsep=0pt,
    left=5pt, right=5pt, top=3pt, bottom=3pt,
    boxrule=0.5pt, arc=3pt,
    toptitle=2pt, bottomtitle=2pt,
    before skip=4pt, after skip=4pt,
    breakable,
    enhanced,
  },
}

% ---------------------------------------------------------------
%  \taskprompt{label}{Title}{nt-file}{th-file}
%  Renders NT and TH as stacked full-width boxes so long prompts
%  use the full page width and break naturally across pages.
% ---------------------------------------------------------------
\newcommand{\taskprompt}[4]{%
  \tcbinputlisting{promptNT, listing file={#3}}%
  \tcbinputlisting{promptTH, listing file={#4}}%
}

This appendix lists the exact prompt templates used for each task under both inference modes.
For Thinking, we use task-specific decomposition instructions that make relevant intermediate reasoning steps explicit, giving models a strong, task-appropriate opportunity to solve each task.
Non-Thinking simply uses the concise task prompt.
Placeholders in double braces (e.g.\ \texttt{\{\{box\_dims\}\}}) are filled at inference time. In the Thinking prompts, ``hidden blocks'' primarily refers to blocks occluded in one view but recoverable
from the other. The independent ambiguity audit identified
one evaluated object containing a nearly fully occluded
block (Appendix~\ref{sec:app_dataset}).

\paragraph{Response-format examples.} The examples included in the prompt templates serve only to illustrate the required output structure, including field names, data types, and serialization format. They are not paired with input images and do not constitute worked examples or few-shot demonstrations of the benchmark tasks. Their numerical values are randomly picked by the author for illustrative purposes rather than sampled from the benchmark's ground-truth distribution, and therefore need not satisfy task-specific constraints. For evaluation, each model prediction is scored against the ground truth computed independently by the procedural generator.

% ==================================================================
\subsection{T1: Volume (Open-Ended)}
\label{sec:prompt_t1}
\taskprompt{t1}{T1: Volume}{prompts/non_thinking/Task_1.md}{prompts/thinking/Task_1.md}

% ==================================================================
\subsection{T1: Volume (MCQ)}
\label{sec:prompt_t1_mcq}
\taskprompt{t1mcq}{T1 MCQ}{prompts/non_thinking/Task_1_mcq.md}{prompts/thinking/Task_1_mcq.md}

% ==================================================================
\subsection{T2: Packaging}
\label{sec:prompt_t2}
\taskprompt{t2}{T2: Packaging}{prompts/non_thinking/Task_2.md}{prompts/thinking/Task_2.md}

% ==================================================================
\subsection{T3: Difference}
\label{sec:prompt_t3}
\taskprompt{t3}{T3: Difference}{prompts/non_thinking/Task_3.md}{prompts/thinking/Task_3.md}

% ==================================================================
\subsection{T3: Difference -- Color-Aware Variant}
\label{sec:prompt_t3_color}
\taskprompt{t3c}{T3 Difference -- Color-Aware Variant}{prompts/non_thinking/Task_3_color_aware.md}{prompts/thinking/Task_3_color_aware.md}

% ==================================================================
\subsection{T4: Color Count}
\label{sec:prompt_t4}
\taskprompt{t4}{T4: Color Count}{prompts/non_thinking/Task_4.md}{prompts/thinking/Task_4.md}

% ==================================================================
\subsection{T5: Color Distribution}
\label{sec:prompt_t5}
\taskprompt{t5}{T5: Color Distribution}{prompts/non_thinking/Task_5.md}{prompts/thinking/Task_5.md}

\subsection{T5: Color Distribution (Ablation prompts)}
\tcbinputlisting{promptNTtfiveSingleColor, listing file={prompts/non_thinking/single_color_Task_5.md}}

\tcbinputlisting{promptNTtfiveInDistribution, listing file={prompts/non_thinking/Task_5_in_distribution_example.md}}

\tcbinputlisting{promptNTtfiveNoExample, listing file={prompts/non_thinking/Task_5_no_example.md}}

\tcbinputlisting{promptNTtfiveTwoD, listing file={prompts/non_thinking/Task_5_2d.md}}

% ==================================================================
\subsection{T6: Local Structure}
\label{sec:prompt_t6}
\taskprompt{t6}{T6: Local Structure}{prompts/non_thinking/Task_6.md}{prompts/thinking/Task_6.md}

% ==================================================================
\subsection{T7: Color Dominance}
\label{sec:prompt_t7}
\taskprompt{t7}{T7: Color Dominance}{prompts/non_thinking/Task_7.md}{prompts/thinking/Task_7.md}

\tcbinputlisting{promptNTtsevenCounts, listing file={prompts/non_thinking/Task_7_from_counts.md}}

% ==================================================================
\subsection{T8: Silhouette Symmetry}
\label{sec:prompt_t8}
\taskprompt{t8}{T8: Silhouette Symmetry}{prompts/non_thinking/Task_8.md}{prompts/thinking/Task_8.md}

% ==================================================================
\subsection{T9: Visible Floors}
\label{sec:prompt_t9}
\taskprompt{t9}{T9: Visible Floors}{prompts/non_thinking/Task_9.md}{prompts/thinking/Task_9.md}

% ==================================================================
\subsection{T10: Color Connected}
\label{sec:prompt_t10}
\taskprompt{t10}{T10: Color Connected}{prompts/non_thinking/Task_10.md}{prompts/thinking/Task_10.md}

% ==================================================================
\subsection{T11: Projection -- Identify Invalid Orthographic Projection}
\label{sec:prompt_ta}
\taskprompt{ta}{T11: Projection (invalid)}{prompts/non_thinking/Task_A.md}{prompts/thinking/Task_A.md}

% ==================================================================
\subsection{T11: Projection -- Identify Valid Orthographic Projection}
\label{sec:prompt_ta_pos}
\taskprompt{tapos}{T11: Projection (valid)}{prompts/non_thinking/Task_A_detect_positive.md}{prompts/thinking/Task_A_detect_positive.md}

% ==================================================================
\subsection{T12: Reconstruct}
\label{sec:prompt_tb}
\taskprompt{tb}{T12: Reconstruct}{prompts/non_thinking/Task_B.md}{prompts/thinking/Task_B.md}

% ==================================================================
\subsection{T13: Mixture-of-Shapes -- Identify Invalid Silhouette}
\label{sec:prompt_tc}
\taskprompt{tc}{T13: Mixture-of-Shapes (invalid)}{prompts/non_thinking/Task_C.md}{prompts/thinking/Task_C.md}

% ==================================================================
\subsection{T13: Mixture-of-Shapes -- Identify Valid Silhouette}
\label{sec:prompt_tc_pos}
\taskprompt{tcpos}{T13: Mixture-of-Shapes (valid)}{prompts/non_thinking/Task_C_detect_positive.md}{prompts/thinking/Task_C_detect_positive.md}

\clearpage
\section{Comprehensive Visual-Condition Tables}
\label{sec:app_full_metric_tables}
This section collects the full per-model metric tables.
Numeric tasks (i.e. \tonelong{}, \tthreelong{}, \tfourlong{}, \tfivelong{}, \tninelong{}) report accuracy, RMSE, and MAE; the remaining classification tasks report accuracy, precision, recall, and F1.
The $\pm$ values are standard deviations across the five inference runs and reflect nondeterministic batched inference (Appendix~\ref{sec:app_exp_details}); they are not confidence intervals over benchmark items. Paired item-level bootstrap intervals for the comparisons drawn from these tables appear in Tables~\ref{tab:visual_cue_effects} (Appendix~\ref{sec:app_visual_cues}) and~\ref{tab:thinking_delta_full} (Appendix~\ref{sec:app_thinking}).

% Auto-generated by scripts/generate_appendix_metric_tables.py -- do not edit by hand.
\onecolumn
\subsection{T1: Volume (Open-Ended)}
\label{sec:app_metrics_t1}
{\small
\setlength{\tabcolsep}{5pt}
\begin{xltabular}{\linewidth}{>{\raggedright\arraybackslash}X lccc}
\toprule
Model & Mode & Acc (\%) & RMSE & MAE \\
\midrule
\endfirsthead
\multicolumn{5}{l}{\textit{Continued from previous page}}\\
\toprule
Model & Mode & Acc (\%) & RMSE & MAE \\
\midrule
\endhead
\midrule \multicolumn{5}{r}{\textit{Continued on next page}}\\
\endfoot
\bottomrule
\addlinespace[8pt]
\caption{Full per-model metrics for \tonelong{} (Open-Ended) across visual variants. Each cell reports the 5-run mean $\pm$ std for one visual-condition variant $\times$ model $\times$ inference mode. Frontier rows appear only for the default visual variant in Thinking mode and report a single run, so no std is shown. Accuracy is reported in percent; RMSE and MAE are in task units. \textbf{Boldface} marks the best mean accuracy within each variant/mode block.}\label{tab:app_metrics_t1}\\
\endlastfoot
\rowcolor{gray!06}\multicolumn{5}{l}{\textit{Uncolored}}\\
Qwen3.5-0.8B & Non-Thinking & 33.4 $\pm$ 0.9 & 2.17 $\pm$ 0.06 & 1.42 $\pm$ 0.03 \\
\rowcolor{gray!03}
Qwen3.5-0.8B & Thinking & 20.8 $\pm$ 4.5 & 7.94 $\pm$ 2.70 & 3.64 $\pm$ 0.59 \\
Qwen3.5-2B & Non-Thinking & 22.4 $\pm$ 1.1 & 1.82 $\pm$ 0.01 & 1.37 $\pm$ 0.01 \\
\rowcolor{gray!03}
Qwen3.5-2B & Thinking & 14.6 $\pm$ 1.1 & 20.48 $\pm$ 5.80 & 6.13 $\pm$ 1.58 \\
Qwen3.5-4B & Non-Thinking & 34.0 $\pm$ 1.0 & 1.58 $\pm$ 0.05 & 1.14 $\pm$ 0.02 \\
\rowcolor{gray!03}
Qwen3.5-4B & Thinking & 28.6 $\pm$ 3.8 & 1.78 $\pm$ 0.11 & 1.27 $\pm$ 0.09 \\
Qwen3.5-9B & Non-Thinking & 32.2 $\pm$ 1.9 & 1.34 $\pm$ 0.03 & 1.00 $\pm$ 0.03 \\
\rowcolor{gray!03}
Qwen3.5-9B & Thinking & 33.2 $\pm$ 3.2 & 1.42 $\pm$ 0.12 & 1.01 $\pm$ 0.08 \\
Qwen3.5-27B & Non-Thinking & 39.8 $\pm$ 1.6 & 1.15 $\pm$ 0.02 & 0.82 $\pm$ 0.03 \\
\rowcolor{gray!03}
Qwen3.5-27B & Thinking & 24.2 $\pm$ 3.6 & 1.52 $\pm$ 0.07 & 1.19 $\pm$ 0.07 \\
Qwen3.5-35B-A3B & Non-Thinking & 36.8 $\pm$ 1.9 & 1.57 $\pm$ 0.04 & 1.07 $\pm$ 0.05 \\
\rowcolor{gray!03}
Qwen3.5-35B-A3B & Thinking & 22.6 $\pm$ 1.8 & 1.90 $\pm$ 0.11 & 1.42 $\pm$ 0.08 \\
Gemma4-E2B & Non-Thinking & 8.2 $\pm$ 0.4 & 3.03 $\pm$ 0.08 & 2.35 $\pm$ 0.04 \\
\rowcolor{gray!03}
Gemma4-E2B & Thinking & 10.8 $\pm$ 2.2 & 7.85 $\pm$ 0.70 & 4.84 $\pm$ 0.47 \\
Gemma4-E4B & Non-Thinking & \textbf{40.0 $\pm$ 0.7} & 1.68 $\pm$ 0.02 & 1.07 $\pm$ 0.02 \\
\rowcolor{gray!03}
Gemma4-E4B & Thinking & 16.8 $\pm$ 3.6 & 4.89 $\pm$ 0.34 & 3.35 $\pm$ 0.14 \\
Gemma4-26B-A4B & Non-Thinking & 14.0 $\pm$ 0.0 & 3.01 $\pm$ 0.02 & 2.41 $\pm$ 0.02 \\
\rowcolor{gray!03}
Gemma4-26B-A4B & Thinking & 15.2 $\pm$ 3.0 & 3.54 $\pm$ 0.22 & 2.42 $\pm$ 0.11 \\
Gemma4-31B & Non-Thinking & 35.0 $\pm$ 0.0 & 1.19 $\pm$ 0.00 & 0.88 $\pm$ 0.00 \\
\rowcolor{gray!03}
Gemma4-31B & Thinking & 17.2 $\pm$ 0.8 & 3.19 $\pm$ 0.08 & 2.39 $\pm$ 0.04 \\
GLM-4.6V-Flash & Non-Thinking & 22.2 $\pm$ 0.4 & 1.99 $\pm$ 0.00 & 1.45 $\pm$ 0.01 \\
\rowcolor{gray!03}
GLM-4.6V-Flash & Thinking & \textbf{38.0 $\pm$ 2.0} & 1.32 $\pm$ 0.07 & 0.91 $\pm$ 0.05 \\
Kimi-VL-A3B & Non-Thinking & 31.8 $\pm$ 0.4 & 2.21 $\pm$ 0.02 & 1.57 $\pm$ 0.01 \\
\rowcolor{gray!03}
Kimi-VL-A3B & Thinking & 0.6 $\pm$ 0.5 & 7.65 $\pm$ 0.05 & 7.24 $\pm$ 0.08 \\
\rowcolor{blue!03}
GPT-5 (High) & Thinking & 14.0 & 3.81 & 2.58 \\
\rowcolor{blue!03}
Grok-4.20 (High) & Thinking & 13.0 & 2.48 & 1.99 \\
\rowcolor{blue!03}
Gemini-3.1-Pro-Preview (High) & Thinking & 18.0 & 2.58 & 2.15 \\
\rowcolor{blue!03}
Gemini-3-Flash-Preview (High) & Thinking & 31.0 & 2.36 & 1.61 \\
\midrule
\rowcolor{gray!06}\multicolumn{5}{l}{\textit{Uncolored + Axes}}\\
Qwen3.5-0.8B & Non-Thinking & 28.0 $\pm$ 0.7 & 3.06 $\pm$ 0.02 & 2.14 $\pm$ 0.02 \\
\rowcolor{gray!03}
Qwen3.5-0.8B & Thinking & 17.8 $\pm$ 3.5 & 9.51 $\pm$ 4.40 & 4.12 $\pm$ 1.11 \\
Qwen3.5-2B & Non-Thinking & 16.2 $\pm$ 0.8 & 2.03 $\pm$ 0.02 & 1.65 $\pm$ 0.02 \\
\rowcolor{gray!03}
Qwen3.5-2B & Thinking & 16.8 $\pm$ 4.4 & 10.83 $\pm$ 4.22 & 4.05 $\pm$ 0.92 \\
Qwen3.5-4B & Non-Thinking & 26.4 $\pm$ 0.5 & 1.94 $\pm$ 0.05 & 1.43 $\pm$ 0.03 \\
\rowcolor{gray!03}
Qwen3.5-4B & Thinking & 27.0 $\pm$ 3.7 & 3.58 $\pm$ 3.56 & 1.57 $\pm$ 0.44 \\
Qwen3.5-9B & Non-Thinking & 25.8 $\pm$ 0.8 & 1.81 $\pm$ 0.04 & 1.40 $\pm$ 0.04 \\
\rowcolor{gray!03}
Qwen3.5-9B & Thinking & 23.8 $\pm$ 4.0 & 1.86 $\pm$ 0.09 & 1.40 $\pm$ 0.04 \\
Qwen3.5-27B & Non-Thinking & 35.8 $\pm$ 0.4 & 1.32 $\pm$ 0.01 & 0.95 $\pm$ 0.01 \\
\rowcolor{gray!03}
Qwen3.5-27B & Thinking & 27.6 $\pm$ 3.9 & 1.59 $\pm$ 0.06 & 1.19 $\pm$ 0.08 \\
Qwen3.5-35B-A3B & Non-Thinking & \textbf{38.8 $\pm$ 1.6} & 1.47 $\pm$ 0.08 & 1.01 $\pm$ 0.04 \\
\rowcolor{gray!03}
Qwen3.5-35B-A3B & Thinking & 25.4 $\pm$ 2.4 & 1.90 $\pm$ 0.14 & 1.39 $\pm$ 0.10 \\
Gemma4-E2B & Non-Thinking & 22.6 $\pm$ 1.5 & 1.71 $\pm$ 0.01 & 1.37 $\pm$ 0.02 \\
\rowcolor{gray!03}
Gemma4-E2B & Thinking & 13.0 $\pm$ 2.1 & 5.84 $\pm$ 1.45 & 3.69 $\pm$ 0.60 \\
Gemma4-E4B & Non-Thinking & 33.4 $\pm$ 0.5 & 1.65 $\pm$ 0.01 & 1.21 $\pm$ 0.01 \\
\rowcolor{gray!03}
Gemma4-E4B & Thinking & 12.8 $\pm$ 4.8 & 5.25 $\pm$ 0.52 & 3.77 $\pm$ 0.39 \\
Gemma4-26B-A4B & Non-Thinking & 24.6 $\pm$ 0.9 & 1.92 $\pm$ 0.00 & 1.48 $\pm$ 0.01 \\
\rowcolor{gray!03}
Gemma4-26B-A4B & Thinking & 22.4 $\pm$ 2.3 & 2.58 $\pm$ 0.06 & 1.98 $\pm$ 0.02 \\
Gemma4-31B & Non-Thinking & 31.0 $\pm$ 0.7 & 1.34 $\pm$ 0.00 & 1.01 $\pm$ 0.00 \\
\rowcolor{gray!03}
Gemma4-31B & Thinking & 16.2 $\pm$ 1.5 & 2.87 $\pm$ 0.05 & 2.14 $\pm$ 0.02 \\
GLM-4.6V-Flash & Non-Thinking & 16.0 $\pm$ 0.0 & 2.08 $\pm$ 0.00 & 1.56 $\pm$ 0.00 \\
\rowcolor{gray!03}
GLM-4.6V-Flash & Thinking & \textbf{29.6 $\pm$ 1.5} & 1.86 $\pm$ 0.02 & 1.31 $\pm$ 0.02 \\
Kimi-VL-A3B & Non-Thinking & 34.0 $\pm$ 0.0 & 1.92 $\pm$ 0.00 & 1.36 $\pm$ 0.00 \\
\rowcolor{gray!03}
Kimi-VL-A3B & Thinking & 0.0 $\pm$ 0.0 & 7.89 $\pm$ 0.09 & 7.53 $\pm$ 0.10 \\
\midrule
\rowcolor{gray!06}\multicolumn{5}{l}{\textit{Colored}}\\
Qwen3.5-0.8B & Non-Thinking & 32.4 $\pm$ 0.9 & 2.04 $\pm$ 0.03 & 1.39 $\pm$ 0.03 \\
\rowcolor{gray!03}
Qwen3.5-0.8B & Thinking & 13.4 $\pm$ 1.5 & 15.74 $\pm$ 2.63 & 7.21 $\pm$ 0.89 \\
Qwen3.5-2B & Non-Thinking & 27.2 $\pm$ 0.4 & 1.51 $\pm$ 0.02 & 1.14 $\pm$ 0.02 \\
\rowcolor{gray!03}
Qwen3.5-2B & Thinking & 23.0 $\pm$ 3.5 & 3.15 $\pm$ 0.60 & 1.80 $\pm$ 0.17 \\
Qwen3.5-4B & Non-Thinking & 29.4 $\pm$ 0.9 & 1.63 $\pm$ 0.01 & 1.20 $\pm$ 0.01 \\
\rowcolor{gray!03}
Qwen3.5-4B & Thinking & 17.6 $\pm$ 1.9 & 2.01 $\pm$ 0.16 & 1.62 $\pm$ 0.12 \\
Qwen3.5-9B & Non-Thinking & 33.4 $\pm$ 1.1 & 1.21 $\pm$ 0.03 & 0.92 $\pm$ 0.03 \\
\rowcolor{gray!03}
Qwen3.5-9B & Thinking & 22.8 $\pm$ 1.9 & 1.42 $\pm$ 0.03 & 1.14 $\pm$ 0.03 \\
Qwen3.5-27B & Non-Thinking & 40.4 $\pm$ 1.3 & 1.17 $\pm$ 0.02 & 0.80 $\pm$ 0.02 \\
\rowcolor{gray!03}
Qwen3.5-27B & Thinking & 24.8 $\pm$ 1.9 & 1.46 $\pm$ 0.14 & 1.11 $\pm$ 0.07 \\
Qwen3.5-35B-A3B & Non-Thinking & \textbf{40.6 $\pm$ 1.1} & 1.34 $\pm$ 0.02 & 0.88 $\pm$ 0.02 \\
\rowcolor{gray!03}
Qwen3.5-35B-A3B & Thinking & 29.8 $\pm$ 3.6 & 1.57 $\pm$ 0.11 & 1.13 $\pm$ 0.05 \\
Gemma4-E2B & Non-Thinking & 28.2 $\pm$ 1.5 & 1.44 $\pm$ 0.01 & 1.09 $\pm$ 0.02 \\
\rowcolor{gray!03}
Gemma4-E2B & Thinking & 18.2 $\pm$ 2.6 & 2.89 $\pm$ 0.41 & 1.99 $\pm$ 0.10 \\
Gemma4-E4B & Non-Thinking & 31.0 $\pm$ 0.7 & 3.22 $\pm$ 0.01 & 1.59 $\pm$ 0.01 \\
\rowcolor{gray!03}
Gemma4-E4B & Thinking & 20.6 $\pm$ 4.7 & 2.62 $\pm$ 0.10 & 1.84 $\pm$ 0.12 \\
Gemma4-26B-A4B & Non-Thinking & 25.4 $\pm$ 0.5 & 2.16 $\pm$ 0.01 & 1.65 $\pm$ 0.01 \\
\rowcolor{gray!03}
Gemma4-26B-A4B & Thinking & 24.2 $\pm$ 1.3 & 2.01 $\pm$ 0.10 & 1.48 $\pm$ 0.05 \\
Gemma4-31B & Non-Thinking & 33.0 $\pm$ 0.0 & 1.48 $\pm$ 0.00 & 1.11 $\pm$ 0.00 \\
\rowcolor{gray!03}
Gemma4-31B & Thinking & 20.2 $\pm$ 0.4 & 2.27 $\pm$ 0.08 & 1.67 $\pm$ 0.05 \\
GLM-4.6V-Flash & Non-Thinking & 21.0 $\pm$ 0.0 & 1.43 $\pm$ 0.00 & 1.18 $\pm$ 0.00 \\
\rowcolor{gray!03}
GLM-4.6V-Flash & Thinking & \textbf{40.6 $\pm$ 1.1} & 1.15 $\pm$ 0.04 & 0.80 $\pm$ 0.03 \\
Kimi-VL-A3B & Non-Thinking & 39.6 $\pm$ 0.5 & 1.57 $\pm$ 0.02 & 1.08 $\pm$ 0.02 \\
\rowcolor{gray!03}
Kimi-VL-A3B & Thinking & 4.0 $\pm$ 0.0 & 5.19 $\pm$ 0.06 & 4.39 $\pm$ 0.05 \\
\end{xltabular}
}

\subsection{T2: Packaging}
\label{sec:app_metrics_t2}
{\small
\setlength{\tabcolsep}{5pt}
\begin{xltabular}{\linewidth}{>{\raggedright\arraybackslash}X lcccc}
\toprule
Model & Mode & Acc (\%) & Prec & Recall & F1 \\
\midrule
\endfirsthead
\multicolumn{6}{l}{\textit{Continued from previous page}}\\
\toprule
Model & Mode & Acc (\%) & Prec & Recall & F1 \\
\midrule
\endhead
\midrule \multicolumn{6}{r}{\textit{Continued on next page}}\\
\endfoot
\bottomrule
\addlinespace[8pt]
\caption{Full per-model metrics for \ttwolong{} across visual variants. Each cell reports the 5-run mean $\pm$ std for one visual-condition variant $\times$ model $\times$ inference mode. Frontier rows appear only for the default visual variant in Thinking mode and report a single run, so no std is shown. Accuracy is reported in percent. \textbf{Boldface} marks the best mean accuracy within each variant/mode block.}\label{tab:app_metrics_t2}\\
\endlastfoot
\rowcolor{gray!06}\multicolumn{6}{l}{\textit{Uncolored}}\\
Qwen3.5-0.8B & Non-Thinking & 40.0 $\pm$ 1.6 & 0.379 $\pm$ 0.018 & 0.400 $\pm$ 0.016 & 0.373 $\pm$ 0.015 \\
\rowcolor{gray!03}
Qwen3.5-0.8B & Thinking & 50.4 $\pm$ 2.1 & 0.510 $\pm$ 0.059 & 0.504 $\pm$ 0.021 & 0.402 $\pm$ 0.031 \\
Qwen3.5-2B & Non-Thinking & 50.8 $\pm$ 0.4 & 0.652 $\pm$ 0.225 & 0.508 $\pm$ 0.004 & 0.351 $\pm$ 0.010 \\
\rowcolor{gray!03}
Qwen3.5-2B & Thinking & 57.0 $\pm$ 2.0 & 0.737 $\pm$ 0.047 & 0.570 $\pm$ 0.020 & 0.478 $\pm$ 0.034 \\
Qwen3.5-4B & Non-Thinking & 52.0 $\pm$ 0.0 & 0.755 $\pm$ 0.000 & 0.520 $\pm$ 0.000 & 0.376 $\pm$ 0.000 \\
\rowcolor{gray!03}
Qwen3.5-4B & Thinking & 62.4 $\pm$ 2.4 & 0.786 $\pm$ 0.008 & 0.624 $\pm$ 0.024 & 0.561 $\pm$ 0.038 \\
Qwen3.5-9B & Non-Thinking & 68.6 $\pm$ 0.5 & 0.807 $\pm$ 0.002 & 0.686 $\pm$ 0.005 & 0.652 $\pm$ 0.007 \\
\rowcolor{gray!03}
Qwen3.5-9B & Thinking & 65.4 $\pm$ 2.4 & 0.790 $\pm$ 0.018 & 0.654 $\pm$ 0.024 & 0.608 $\pm$ 0.034 \\
Qwen3.5-27B & Non-Thinking & \textbf{79.2 $\pm$ 0.8} & 0.853 $\pm$ 0.004 & 0.792 $\pm$ 0.008 & 0.783 $\pm$ 0.010 \\
\rowcolor{gray!03}
Qwen3.5-27B & Thinking & 88.2 $\pm$ 1.3 & 0.890 $\pm$ 0.015 & 0.882 $\pm$ 0.013 & 0.881 $\pm$ 0.013 \\
Qwen3.5-35B-A3B & Non-Thinking & 65.8 $\pm$ 1.6 & 0.792 $\pm$ 0.010 & 0.658 $\pm$ 0.016 & 0.613 $\pm$ 0.025 \\
\rowcolor{gray!03}
Qwen3.5-35B-A3B & Thinking & 77.0 $\pm$ 3.4 & 0.840 $\pm$ 0.020 & 0.770 $\pm$ 0.034 & 0.757 $\pm$ 0.039 \\
Gemma4-E2B & Non-Thinking & 59.0 $\pm$ 0.7 & 0.609 $\pm$ 0.007 & 0.590 $\pm$ 0.007 & 0.571 $\pm$ 0.009 \\
\rowcolor{gray!03}
Gemma4-E2B & Thinking & 74.8 $\pm$ 2.0 & 0.773 $\pm$ 0.019 & 0.748 $\pm$ 0.020 & 0.742 $\pm$ 0.023 \\
Gemma4-E4B & Non-Thinking & 55.0 $\pm$ 0.0 & 0.763 $\pm$ 0.000 & 0.550 $\pm$ 0.000 & 0.436 $\pm$ 0.000 \\
\rowcolor{gray!03}
Gemma4-E4B & Thinking & 78.6 $\pm$ 1.7 & 0.792 $\pm$ 0.014 & 0.786 $\pm$ 0.017 & 0.785 $\pm$ 0.017 \\
Gemma4-26B-A4B & Non-Thinking & 58.0 $\pm$ 0.0 & 0.772 $\pm$ 0.000 & 0.580 $\pm$ 0.000 & 0.490 $\pm$ 0.000 \\
\rowcolor{gray!03}
Gemma4-26B-A4B & Thinking & 85.8 $\pm$ 1.1 & 0.889 $\pm$ 0.007 & 0.858 $\pm$ 0.011 & 0.855 $\pm$ 0.012 \\
Gemma4-31B & Non-Thinking & 76.0 $\pm$ 0.0 & 0.838 $\pm$ 0.000 & 0.760 $\pm$ 0.000 & 0.745 $\pm$ 0.000 \\
\rowcolor{gray!03}
Gemma4-31B & Thinking & \textbf{91.6 $\pm$ 1.3} & 0.922 $\pm$ 0.012 & 0.916 $\pm$ 0.013 & 0.916 $\pm$ 0.014 \\
GLM-4.6V-Flash & Non-Thinking & 51.0 $\pm$ 0.0 & 0.753 $\pm$ 0.000 & 0.510 $\pm$ 0.000 & 0.355 $\pm$ 0.000 \\
\rowcolor{gray!03}
GLM-4.6V-Flash & Thinking & 73.8 $\pm$ 1.6 & 0.777 $\pm$ 0.020 & 0.738 $\pm$ 0.016 & 0.729 $\pm$ 0.017 \\
Kimi-VL-A3B & Non-Thinking & 50.0 $\pm$ 0.0 & 0.250 $\pm$ 0.000 & 0.500 $\pm$ 0.000 & 0.333 $\pm$ 0.000 \\
\rowcolor{gray!03}
Kimi-VL-A3B & Thinking & 66.4 $\pm$ 1.3 & 0.687 $\pm$ 0.012 & 0.664 $\pm$ 0.013 & 0.653 $\pm$ 0.016 \\
\rowcolor{blue!03}
GPT-5 (High) & Thinking & 88.0 & 0.882 & 0.880 & 0.880 \\
\rowcolor{blue!03}
Grok-4.20 (High) & Thinking & 82.0 & 0.857 & 0.820 & 0.815 \\
\rowcolor{blue!03}
Gemini-3.1-Pro-Preview (High) & Thinking & 82.0 & 0.857 & 0.820 & 0.815 \\
\rowcolor{blue!03}
Gemini-3-Flash-Preview (High) & Thinking & 88.0 & 0.890 & 0.880 & 0.879 \\
\midrule
\rowcolor{gray!06}\multicolumn{6}{l}{\textit{Uncolored + Axes}}\\
Qwen3.5-0.8B & Non-Thinking & 44.4 $\pm$ 2.1 & 0.388 $\pm$ 0.031 & 0.444 $\pm$ 0.021 & 0.363 $\pm$ 0.016 \\
\rowcolor{gray!03}
Qwen3.5-0.8B & Thinking & 49.4 $\pm$ 1.1 & 0.466 $\pm$ 0.051 & 0.494 $\pm$ 0.011 & 0.389 $\pm$ 0.036 \\
Qwen3.5-2B & Non-Thinking & 50.0 $\pm$ 0.0 & 0.250 $\pm$ 0.000 & 0.500 $\pm$ 0.000 & 0.333 $\pm$ 0.000 \\
\rowcolor{gray!03}
Qwen3.5-2B & Thinking & 57.4 $\pm$ 2.7 & 0.761 $\pm$ 0.019 & 0.574 $\pm$ 0.027 & 0.480 $\pm$ 0.049 \\
Qwen3.5-4B & Non-Thinking & 51.0 $\pm$ 0.0 & 0.753 $\pm$ 0.000 & 0.510 $\pm$ 0.000 & 0.355 $\pm$ 0.000 \\
\rowcolor{gray!03}
Qwen3.5-4B & Thinking & 66.0 $\pm$ 1.6 & 0.798 $\pm$ 0.006 & 0.660 $\pm$ 0.016 & 0.615 $\pm$ 0.023 \\
Qwen3.5-9B & Non-Thinking & 55.2 $\pm$ 0.4 & 0.764 $\pm$ 0.001 & 0.552 $\pm$ 0.004 & 0.439 $\pm$ 0.008 \\
\rowcolor{gray!03}
Qwen3.5-9B & Thinking & 68.4 $\pm$ 2.1 & 0.801 $\pm$ 0.020 & 0.684 $\pm$ 0.021 & 0.650 $\pm$ 0.026 \\
Qwen3.5-27B & Non-Thinking & \textbf{73.2 $\pm$ 0.8} & 0.826 $\pm$ 0.004 & 0.732 $\pm$ 0.008 & 0.711 $\pm$ 0.010 \\
\rowcolor{gray!03}
Qwen3.5-27B & Thinking & 82.2 $\pm$ 1.6 & 0.834 $\pm$ 0.017 & 0.822 $\pm$ 0.016 & 0.820 $\pm$ 0.017 \\
Qwen3.5-35B-A3B & Non-Thinking & 55.6 $\pm$ 0.5 & 0.765 $\pm$ 0.002 & 0.556 $\pm$ 0.005 & 0.447 $\pm$ 0.010 \\
\rowcolor{gray!03}
Qwen3.5-35B-A3B & Thinking & 75.2 $\pm$ 1.1 & 0.831 $\pm$ 0.009 & 0.752 $\pm$ 0.011 & 0.736 $\pm$ 0.013 \\
Gemma4-E2B & Non-Thinking & 49.8 $\pm$ 0.4 & 0.496 $\pm$ 0.010 & 0.498 $\pm$ 0.004 & 0.414 $\pm$ 0.003 \\
\rowcolor{gray!03}
Gemma4-E2B & Thinking & 75.6 $\pm$ 4.8 & 0.779 $\pm$ 0.036 & 0.756 $\pm$ 0.048 & 0.750 $\pm$ 0.056 \\
Gemma4-E4B & Non-Thinking & 52.0 $\pm$ 0.0 & 0.755 $\pm$ 0.000 & 0.520 $\pm$ 0.000 & 0.376 $\pm$ 0.000 \\
\rowcolor{gray!03}
Gemma4-E4B & Thinking & 80.6 $\pm$ 2.4 & 0.810 $\pm$ 0.024 & 0.806 $\pm$ 0.024 & 0.805 $\pm$ 0.024 \\
Gemma4-26B-A4B & Non-Thinking & 56.0 $\pm$ 0.0 & 0.766 $\pm$ 0.000 & 0.560 $\pm$ 0.000 & 0.454 $\pm$ 0.000 \\
\rowcolor{gray!03}
Gemma4-26B-A4B & Thinking & 85.8 $\pm$ 1.5 & 0.876 $\pm$ 0.013 & 0.858 $\pm$ 0.015 & 0.856 $\pm$ 0.015 \\
Gemma4-31B & Non-Thinking & 69.0 $\pm$ 0.0 & 0.809 $\pm$ 0.000 & 0.690 $\pm$ 0.000 & 0.657 $\pm$ 0.000 \\
\rowcolor{gray!03}
Gemma4-31B & Thinking & \textbf{90.8 $\pm$ 0.8} & 0.911 $\pm$ 0.008 & 0.908 $\pm$ 0.008 & 0.908 $\pm$ 0.008 \\
GLM-4.6V-Flash & Non-Thinking & 50.0 $\pm$ 0.0 & 0.250 $\pm$ 0.000 & 0.500 $\pm$ 0.000 & 0.333 $\pm$ 0.000 \\
\rowcolor{gray!03}
GLM-4.6V-Flash & Thinking & 73.6 $\pm$ 0.5 & 0.791 $\pm$ 0.009 & 0.736 $\pm$ 0.005 & 0.723 $\pm$ 0.009 \\
Kimi-VL-A3B & Non-Thinking & 50.0 $\pm$ 0.0 & 0.250 $\pm$ 0.000 & 0.500 $\pm$ 0.000 & 0.333 $\pm$ 0.000 \\
\rowcolor{gray!03}
Kimi-VL-A3B & Thinking & 64.0 $\pm$ 2.4 & 0.681 $\pm$ 0.036 & 0.640 $\pm$ 0.024 & 0.619 $\pm$ 0.024 \\
\end{xltabular}
}

\subsection{T3: Difference}
\label{sec:app_metrics_t3}
{\small
\setlength{\tabcolsep}{5pt}
\begin{xltabular}{\linewidth}{>{\raggedright\arraybackslash}X lccc}
\toprule
Model & Mode & Acc (\%) & RMSE & MAE \\
\midrule
\endfirsthead
\multicolumn{5}{l}{\textit{Continued from previous page}}\\
\toprule
Model & Mode & Acc (\%) & RMSE & MAE \\
\midrule
\endhead
\midrule \multicolumn{5}{r}{\textit{Continued on next page}}\\
\endfoot
\bottomrule
\addlinespace[8pt]
\caption{Full per-model metrics for \tthreelong{} across visual variants. Each cell reports the 5-run mean $\pm$ std for one visual-condition variant $\times$ model $\times$ inference mode. Frontier rows appear only for the default visual variant in Thinking mode and report a single run, so no std is shown. Accuracy is reported in percent; RMSE and MAE are in task units. \textbf{Boldface} marks the best mean accuracy within each variant/mode block.}\label{tab:app_metrics_t3}\\
\endlastfoot
\rowcolor{gray!06}\multicolumn{5}{l}{\textit{Uncolored + Axes}}\\
Qwen3.5-0.8B & Non-Thinking & 0.0 $\pm$ 0.0 & 5.30 $\pm$ 0.04 & 4.87 $\pm$ 0.05 \\
\rowcolor{gray!03}
Qwen3.5-0.8B & Thinking & 0.2 $\pm$ 0.4 & 4.64 $\pm$ 0.13 & 3.83 $\pm$ 0.08 \\
Qwen3.5-2B & Non-Thinking & 0.0 $\pm$ 0.0 & 4.18 $\pm$ 0.03 & 3.75 $\pm$ 0.03 \\
\rowcolor{gray!03}
Qwen3.5-2B & Thinking & 0.0 $\pm$ 0.0 & 4.18 $\pm$ 0.09 & 3.53 $\pm$ 0.10 \\
Qwen3.5-4B & Non-Thinking & 0.0 $\pm$ 0.0 & 4.24 $\pm$ 0.02 & 3.79 $\pm$ 0.01 \\
\rowcolor{gray!03}
Qwen3.5-4B & Thinking & 0.0 $\pm$ 0.0 & 4.35 $\pm$ 0.11 & 3.76 $\pm$ 0.12 \\
Qwen3.5-9B & Non-Thinking & \textbf{2.0 $\pm$ 0.0} & 2.39 $\pm$ 0.03 & 1.82 $\pm$ 0.03 \\
\rowcolor{gray!03}
Qwen3.5-9B & Thinking & 0.2 $\pm$ 0.4 & 3.83 $\pm$ 0.17 & 3.24 $\pm$ 0.15 \\
Qwen3.5-27B & Non-Thinking & 0.0 $\pm$ 0.0 & 4.52 $\pm$ 0.05 & 4.04 $\pm$ 0.06 \\
\rowcolor{gray!03}
Qwen3.5-27B & Thinking & 0.0 $\pm$ 0.0 & 4.45 $\pm$ 0.07 & 3.88 $\pm$ 0.05 \\
Qwen3.5-35B-A3B & Non-Thinking & 0.0 $\pm$ 0.0 & 5.04 $\pm$ 0.01 & 4.73 $\pm$ 0.01 \\
\rowcolor{gray!03}
Qwen3.5-35B-A3B & Thinking & 0.0 $\pm$ 0.0 & 4.36 $\pm$ 0.10 & 3.83 $\pm$ 0.09 \\
Gemma4-E2B & Non-Thinking & 0.0 $\pm$ 0.0 & 4.08 $\pm$ 0.01 & 3.68 $\pm$ 0.01 \\
\rowcolor{gray!03}
Gemma4-E2B & Thinking & 0.2 $\pm$ 0.4 & 3.83 $\pm$ 0.07 & 3.10 $\pm$ 0.08 \\
Gemma4-E4B & Non-Thinking & 0.0 $\pm$ 0.0 & 4.18 $\pm$ 0.01 & 3.84 $\pm$ 0.01 \\
\rowcolor{gray!03}
Gemma4-E4B & Thinking & 0.0 $\pm$ 0.0 & 4.85 $\pm$ 0.09 & 3.90 $\pm$ 0.07 \\
Gemma4-26B-A4B & Non-Thinking & 0.0 $\pm$ 0.0 & 4.68 $\pm$ 0.01 & 4.21 $\pm$ 0.02 \\
\rowcolor{gray!03}
Gemma4-26B-A4B & Thinking & 0.0 $\pm$ 0.0 & 4.98 $\pm$ 0.02 & 4.59 $\pm$ 0.03 \\
Gemma4-31B & Non-Thinking & 0.0 $\pm$ 0.0 & 3.72 $\pm$ 0.00 & 3.22 $\pm$ 0.00 \\
\rowcolor{gray!03}
Gemma4-31B & Thinking & 0.0 $\pm$ 0.0 & 3.43 $\pm$ 0.05 & 2.79 $\pm$ 0.04 \\
GLM-4.6V-Flash & Non-Thinking & 0.0 $\pm$ 0.0 & 4.35 $\pm$ 0.03 & 3.88 $\pm$ 0.02 \\
\rowcolor{gray!03}
GLM-4.6V-Flash & Thinking & 0.0 $\pm$ 0.0 & 4.50 $\pm$ 0.05 & 4.07 $\pm$ 0.05 \\
Kimi-VL-A3B & Non-Thinking & 0.0 $\pm$ 0.0 & 3.02 $\pm$ 0.05 & 2.53 $\pm$ 0.03 \\
\rowcolor{gray!03}
Kimi-VL-A3B & Thinking & 0.0 $\pm$ 0.0 & 2.89 $\pm$ 0.03 & 2.43 $\pm$ 0.03 \\
\rowcolor{blue!03}
GPT-5 (High) & Thinking & 0.0 & 4.52 & 4.02 \\
\rowcolor{blue!03}
Grok-4.20 (High) & Thinking & 0.0 & 3.10 & 2.60 \\
\rowcolor{blue!03}
Gemini-3.1-Pro-Preview (High) & Thinking & \textbf{1.0} & 4.20 & 3.52 \\
\rowcolor{blue!03}
Gemini-3-Flash-Preview (High) & Thinking & 0.0 & 4.04 & 3.45 \\
\midrule
\rowcolor{gray!06}\multicolumn{5}{l}{\textit{Colored + Axes}}\\
Qwen3.5-0.8B & Non-Thinking & 0.0 $\pm$ 0.0 & 6.73 $\pm$ 0.03 & 6.57 $\pm$ 0.04 \\
\rowcolor{gray!03}
Qwen3.5-0.8B & Thinking & 3.0 $\pm$ 1.2 & 4.81 $\pm$ 0.42 & 3.09 $\pm$ 0.26 \\
Qwen3.5-2B & Non-Thinking & 4.0 $\pm$ 0.0 & 6.09 $\pm$ 0.03 & 4.78 $\pm$ 0.03 \\
\rowcolor{gray!03}
Qwen3.5-2B & Thinking & 0.0 $\pm$ 0.0 & 6.98 $\pm$ 0.14 & 6.28 $\pm$ 0.12 \\
Qwen3.5-4B & Non-Thinking & 7.0 $\pm$ 0.0 & 1.98 $\pm$ 0.03 & 1.31 $\pm$ 0.03 \\
\rowcolor{gray!03}
Qwen3.5-4B & Thinking & 2.8 $\pm$ 1.3 & 6.06 $\pm$ 0.40 & 4.35 $\pm$ 0.46 \\
Qwen3.5-9B & Non-Thinking & \textbf{9.6 $\pm$ 0.5} & 1.35 $\pm$ 0.02 & 0.94 $\pm$ 0.02 \\
\rowcolor{gray!03}
Qwen3.5-9B & Thinking & 5.0 $\pm$ 1.0 & 4.57 $\pm$ 0.20 & 2.98 $\pm$ 0.17 \\
Qwen3.5-27B & Non-Thinking & 8.0 $\pm$ 0.0 & 1.68 $\pm$ 0.00 & 1.02 $\pm$ 0.00 \\
\rowcolor{gray!03}
Qwen3.5-27B & Thinking & \textbf{8.8 $\pm$ 2.2} & 5.68 $\pm$ 0.25 & 3.77 $\pm$ 0.23 \\
Qwen3.5-35B-A3B & Non-Thinking & 5.4 $\pm$ 0.5 & 4.75 $\pm$ 0.16 & 3.10 $\pm$ 0.13 \\
\rowcolor{gray!03}
Qwen3.5-35B-A3B & Thinking & 2.6 $\pm$ 1.7 & 4.90 $\pm$ 0.18 & 3.61 $\pm$ 0.20 \\
Gemma4-E2B & Non-Thinking & 0.0 $\pm$ 0.0 & 6.55 $\pm$ 0.01 & 6.40 $\pm$ 0.02 \\
\rowcolor{gray!03}
Gemma4-E2B & Thinking & 0.0 $\pm$ 0.0 & 5.23 $\pm$ 0.13 & 4.60 $\pm$ 0.14 \\
Gemma4-E4B & Non-Thinking & 1.0 $\pm$ 0.0 & 3.91 $\pm$ 0.02 & 3.43 $\pm$ 0.02 \\
\rowcolor{gray!03}
Gemma4-E4B & Thinking & 0.4 $\pm$ 0.5 & 4.28 $\pm$ 0.12 & 3.69 $\pm$ 0.10 \\
Gemma4-26B-A4B & Non-Thinking & 1.8 $\pm$ 0.4 & 4.01 $\pm$ 0.03 & 2.97 $\pm$ 0.04 \\
\rowcolor{gray!03}
Gemma4-26B-A4B & Thinking & 0.0 $\pm$ 0.0 & 8.11 $\pm$ 0.03 & 7.82 $\pm$ 0.05 \\
Gemma4-31B & Non-Thinking & 6.2 $\pm$ 0.4 & 3.99 $\pm$ 0.01 & 3.03 $\pm$ 0.01 \\
\rowcolor{gray!03}
Gemma4-31B & Thinking & 2.6 $\pm$ 1.1 & 3.83 $\pm$ 0.10 & 2.97 $\pm$ 0.12 \\
GLM-4.6V-Flash & Non-Thinking & 1.2 $\pm$ 0.4 & 6.34 $\pm$ 0.09 & 4.57 $\pm$ 0.10 \\
\rowcolor{gray!03}
GLM-4.6V-Flash & Thinking & 2.6 $\pm$ 0.9 & 3.62 $\pm$ 0.29 & 2.07 $\pm$ 0.25 \\
Kimi-VL-A3B & Non-Thinking & 0.0 $\pm$ 0.0 & 4.70 $\pm$ 0.01 & 4.55 $\pm$ 0.01 \\
\rowcolor{gray!03}
Kimi-VL-A3B & Thinking & 0.0 $\pm$ 0.0 & 5.63 $\pm$ 0.04 & 5.40 $\pm$ 0.04 \\
\end{xltabular}
}

\subsection{T4: Color Count}
\label{sec:app_metrics_t4}
{\small
\setlength{\tabcolsep}{5pt}
\begin{xltabular}{\linewidth}{>{\raggedright\arraybackslash}X lccc}
\toprule
Model & Mode & Acc (\%) & RMSE & MAE \\
\midrule
\endfirsthead
\multicolumn{5}{l}{\textit{Continued from previous page}}\\
\toprule
Model & Mode & Acc (\%) & RMSE & MAE \\
\midrule
\endhead
\midrule \multicolumn{5}{r}{\textit{Continued on next page}}\\
\endfoot
\bottomrule
\addlinespace[8pt]
\caption{Full per-model metrics for \tfourlong{} across visual variants. Each cell reports the 5-run mean $\pm$ std for one visual-condition variant $\times$ model $\times$ inference mode. Frontier rows appear only for the default visual variant in Thinking mode and report a single run, so no std is shown. Accuracy is reported in percent; RMSE and MAE are in task units. \textbf{Boldface} marks the best mean accuracy within each variant/mode block.}\label{tab:app_metrics_t4}\\
\endlastfoot
\rowcolor{gray!06}\multicolumn{5}{l}{\textit{Colored}}\\
Qwen3.5-0.8B & Non-Thinking & 6.0 $\pm$ 0.0 & 2.88 $\pm$ 0.02 & 2.52 $\pm$ 0.02 \\
\rowcolor{gray!03}
Qwen3.5-0.8B & Thinking & 19.8 $\pm$ 1.3 & 2.54 $\pm$ 0.04 & 1.91 $\pm$ 0.04 \\
Qwen3.5-2B & Non-Thinking & 24.4 $\pm$ 0.9 & 1.96 $\pm$ 0.00 & 1.48 $\pm$ 0.01 \\
\rowcolor{gray!03}
Qwen3.5-2B & Thinking & 25.8 $\pm$ 1.6 & 1.81 $\pm$ 0.03 & 1.37 $\pm$ 0.02 \\
Qwen3.5-4B & Non-Thinking & 20.2 $\pm$ 1.8 & 1.74 $\pm$ 0.03 & 1.38 $\pm$ 0.04 \\
\rowcolor{gray!03}
Qwen3.5-4B & Thinking & 43.8 $\pm$ 2.2 & 0.93 $\pm$ 0.04 & 0.66 $\pm$ 0.03 \\
Qwen3.5-9B & Non-Thinking & 11.0 $\pm$ 0.0 & 2.17 $\pm$ 0.02 & 1.85 $\pm$ 0.01 \\
\rowcolor{gray!03}
Qwen3.5-9B & Thinking & 45.2 $\pm$ 1.8 & 0.98 $\pm$ 0.07 & 0.67 $\pm$ 0.04 \\
Qwen3.5-27B & Non-Thinking & 45.2 $\pm$ 1.3 & 0.83 $\pm$ 0.01 & 0.60 $\pm$ 0.01 \\
\rowcolor{gray!03}
Qwen3.5-27B & Thinking & 57.2 $\pm$ 5.1 & 0.68 $\pm$ 0.04 & 0.44 $\pm$ 0.05 \\
Qwen3.5-35B-A3B & Non-Thinking & 43.0 $\pm$ 0.7 & 0.99 $\pm$ 0.01 & 0.70 $\pm$ 0.01 \\
\rowcolor{gray!03}
Qwen3.5-35B-A3B & Thinking & 58.2 $\pm$ 3.7 & 0.70 $\pm$ 0.05 & 0.44 $\pm$ 0.04 \\
Gemma4-E2B & Non-Thinking & 11.8 $\pm$ 0.4 & 2.84 $\pm$ 0.02 & 2.41 $\pm$ 0.02 \\
\rowcolor{gray!03}
Gemma4-E2B & Thinking & 41.6 $\pm$ 3.4 & 0.99 $\pm$ 0.03 & 0.71 $\pm$ 0.04 \\
Gemma4-E4B & Non-Thinking & 24.8 $\pm$ 0.4 & 1.32 $\pm$ 0.02 & 1.05 $\pm$ 0.01 \\
\rowcolor{gray!03}
Gemma4-E4B & Thinking & 36.2 $\pm$ 1.3 & 1.04 $\pm$ 0.02 & 0.78 $\pm$ 0.01 \\
Gemma4-26B-A4B & Non-Thinking & 27.4 $\pm$ 0.5 & 1.28 $\pm$ 0.00 & 1.02 $\pm$ 0.01 \\
\rowcolor{gray!03}
Gemma4-26B-A4B & Thinking & 55.0 $\pm$ 1.4 & 0.77 $\pm$ 0.02 & 0.50 $\pm$ 0.02 \\
Gemma4-31B & Non-Thinking & \textbf{52.8 $\pm$ 0.4} & 0.83 $\pm$ 0.00 & 0.54 $\pm$ 0.00 \\
\rowcolor{gray!03}
Gemma4-31B & Thinking & 56.2 $\pm$ 0.8 & 0.73 $\pm$ 0.04 & 0.47 $\pm$ 0.02 \\
GLM-4.6V-Flash & Non-Thinking & 38.2 $\pm$ 0.8 & 1.26 $\pm$ 0.00 & 0.88 $\pm$ 0.01 \\
\rowcolor{gray!03}
GLM-4.6V-Flash & Thinking & 47.0 $\pm$ 1.6 & 0.78 $\pm$ 0.03 & 0.56 $\pm$ 0.02 \\
Kimi-VL-A3B & Non-Thinking & 20.2 $\pm$ 0.8 & 2.06 $\pm$ 0.03 & 1.59 $\pm$ 0.02 \\
\rowcolor{gray!03}
Kimi-VL-A3B & Thinking & 6.0 $\pm$ 0.0 & 2.69 $\pm$ 0.01 & 2.36 $\pm$ 0.01 \\
\rowcolor{blue!03}
GPT-5 (High) & Thinking & 64.0 & 0.65 & 0.38 \\
\rowcolor{blue!03}
Grok-4.20 (High) & Thinking & 52.0 & 0.71 & 0.49 \\
\rowcolor{blue!03}
Gemini-3.1-Pro-Preview (High) & Thinking & 66.0 & 0.69 & 0.38 \\
\rowcolor{blue!03}
Gemini-3-Flash-Preview (High) & Thinking & \textbf{70.0} & 0.55 & 0.30 \\
\midrule
\rowcolor{gray!06}\multicolumn{5}{l}{\textit{Colored + Axes}}\\
Qwen3.5-0.8B & Non-Thinking & 2.6 $\pm$ 0.5 & 3.22 $\pm$ 0.02 & 2.92 $\pm$ 0.03 \\
\rowcolor{gray!03}
Qwen3.5-0.8B & Thinking & 20.6 $\pm$ 2.4 & 2.49 $\pm$ 0.13 & 1.86 $\pm$ 0.10 \\
Qwen3.5-2B & Non-Thinking & 21.8 $\pm$ 0.4 & 2.13 $\pm$ 0.01 & 1.65 $\pm$ 0.01 \\
\rowcolor{gray!03}
Qwen3.5-2B & Thinking & 31.6 $\pm$ 3.8 & 1.59 $\pm$ 0.16 & 1.13 $\pm$ 0.10 \\
Qwen3.5-4B & Non-Thinking & 33.4 $\pm$ 1.7 & 1.52 $\pm$ 0.01 & 1.08 $\pm$ 0.01 \\
\rowcolor{gray!03}
Qwen3.5-4B & Thinking & 40.8 $\pm$ 4.1 & 1.00 $\pm$ 0.05 & 0.72 $\pm$ 0.05 \\
Qwen3.5-9B & Non-Thinking & 14.0 $\pm$ 0.7 & 1.97 $\pm$ 0.02 & 1.64 $\pm$ 0.01 \\
\rowcolor{gray!03}
Qwen3.5-9B & Thinking & 39.8 $\pm$ 1.9 & 1.00 $\pm$ 0.04 & 0.72 $\pm$ 0.04 \\
Qwen3.5-27B & Non-Thinking & 45.8 $\pm$ 0.8 & 0.81 $\pm$ 0.01 & 0.58 $\pm$ 0.01 \\
\rowcolor{gray!03}
Qwen3.5-27B & Thinking & 47.4 $\pm$ 4.0 & 0.80 $\pm$ 0.06 & 0.56 $\pm$ 0.06 \\
Qwen3.5-35B-A3B & Non-Thinking & \textbf{47.2 $\pm$ 1.9} & 0.86 $\pm$ 0.05 & 0.59 $\pm$ 0.04 \\
\rowcolor{gray!03}
Qwen3.5-35B-A3B & Thinking & \textbf{55.4 $\pm$ 2.3} & 0.72 $\pm$ 0.03 & 0.47 $\pm$ 0.03 \\
Gemma4-E2B & Non-Thinking & 7.8 $\pm$ 0.4 & 2.95 $\pm$ 0.01 & 2.55 $\pm$ 0.02 \\
\rowcolor{gray!03}
Gemma4-E2B & Thinking & 35.2 $\pm$ 1.9 & 1.04 $\pm$ 0.02 & 0.78 $\pm$ 0.02 \\
Gemma4-E4B & Non-Thinking & 29.4 $\pm$ 0.9 & 1.14 $\pm$ 0.01 & 0.88 $\pm$ 0.01 \\
\rowcolor{gray!03}
Gemma4-E4B & Thinking & 34.6 $\pm$ 1.5 & 1.09 $\pm$ 0.05 & 0.82 $\pm$ 0.03 \\
Gemma4-26B-A4B & Non-Thinking & 23.0 $\pm$ 0.0 & 1.32 $\pm$ 0.01 & 1.06 $\pm$ 0.01 \\
\rowcolor{gray!03}
Gemma4-26B-A4B & Thinking & 54.6 $\pm$ 1.7 & 0.74 $\pm$ 0.02 & 0.48 $\pm$ 0.02 \\
Gemma4-31B & Non-Thinking & 44.2 $\pm$ 0.4 & 0.91 $\pm$ 0.00 & 0.65 $\pm$ 0.00 \\
\rowcolor{gray!03}
Gemma4-31B & Thinking & \textbf{55.4 $\pm$ 1.5} & 0.75 $\pm$ 0.02 & 0.49 $\pm$ 0.02 \\
GLM-4.6V-Flash & Non-Thinking & 31.0 $\pm$ 0.7 & 1.37 $\pm$ 0.01 & 1.02 $\pm$ 0.00 \\
\rowcolor{gray!03}
GLM-4.6V-Flash & Thinking & 46.4 $\pm$ 2.4 & 0.77 $\pm$ 0.03 & 0.56 $\pm$ 0.02 \\
Kimi-VL-A3B & Non-Thinking & 17.0 $\pm$ 1.0 & 2.42 $\pm$ 0.03 & 1.93 $\pm$ 0.03 \\
\rowcolor{gray!03}
Kimi-VL-A3B & Thinking & 6.0 $\pm$ 0.0 & 2.82 $\pm$ 0.03 & 2.48 $\pm$ 0.02 \\
\end{xltabular}
}

\subsection{T5: Color Distribution}
\label{sec:app_metrics_t5}
{\small
\setlength{\tabcolsep}{5pt}
\begin{xltabular}{\linewidth}{>{\raggedright\arraybackslash}X lccc}
\toprule
Model & Mode & Acc (\%) & RMSE & MAE \\
\midrule
\endfirsthead
\multicolumn{5}{l}{\textit{Continued from previous page}}\\
\toprule
Model & Mode & Acc (\%) & RMSE & MAE \\
\midrule
\endhead
\midrule \multicolumn{5}{r}{\textit{Continued on next page}}\\
\endfoot
\bottomrule
\addlinespace[8pt]
\caption{Full per-model metrics for \tfivelong{} across visual variants. Each cell reports the 5-run mean $\pm$ std for one visual-condition variant $\times$ model $\times$ inference mode. Frontier rows appear only for the default visual variant in Thinking mode and report a single run, so no std is shown. Accuracy is reported in percent; RMSE and MAE are in task units. \textbf{Boldface} marks the best mean accuracy within each variant/mode block.}\label{tab:app_metrics_t5}\\
\endlastfoot
\rowcolor{gray!06}\multicolumn{5}{l}{\textit{Colored}}\\
Qwen3.5-0.8B & Non-Thinking & 0.0 $\pm$ 0.0 & 2.12 $\pm$ 0.12 & 1.11 $\pm$ 0.05 \\
\rowcolor{gray!03}
Qwen3.5-0.8B & Thinking & 0.4 $\pm$ 0.5 & 2.56 $\pm$ 0.11 & 1.28 $\pm$ 0.07 \\
Qwen3.5-2B & Non-Thinking & 0.0 $\pm$ 0.0 & 1.54 $\pm$ 0.03 & 0.92 $\pm$ 0.01 \\
\rowcolor{gray!03}
Qwen3.5-2B & Thinking & 0.0 $\pm$ 0.0 & 1.75 $\pm$ 0.07 & 0.85 $\pm$ 0.02 \\
Qwen3.5-4B & Non-Thinking & 0.0 $\pm$ 0.0 & 1.10 $\pm$ 0.01 & 0.68 $\pm$ 0.01 \\
\rowcolor{gray!03}
Qwen3.5-4B & Thinking & 1.0 $\pm$ 1.0 & 0.99 $\pm$ 0.02 & 0.53 $\pm$ 0.01 \\
Qwen3.5-9B & Non-Thinking & 0.0 $\pm$ 0.0 & 1.14 $\pm$ 0.01 & 0.79 $\pm$ 0.01 \\
\rowcolor{gray!03}
Qwen3.5-9B & Thinking & 1.0 $\pm$ 0.7 & 0.93 $\pm$ 0.02 & 0.50 $\pm$ 0.02 \\
Qwen3.5-27B & Non-Thinking & 3.0 $\pm$ 0.0 & 0.94 $\pm$ 0.00 & 0.50 $\pm$ 0.00 \\
\rowcolor{gray!03}
Qwen3.5-27B & Thinking & 1.4 $\pm$ 0.9 & 0.89 $\pm$ 0.01 & 0.42 $\pm$ 0.00 \\
Qwen3.5-35B-A3B & Non-Thinking & 0.0 $\pm$ 0.0 & 0.99 $\pm$ 0.01 & 0.61 $\pm$ 0.01 \\
\rowcolor{gray!03}
Qwen3.5-35B-A3B & Thinking & 0.6 $\pm$ 0.5 & 0.88 $\pm$ 0.02 & 0.45 $\pm$ 0.01 \\
Gemma4-E2B & Non-Thinking & 0.0 $\pm$ 0.0 & 1.10 $\pm$ 0.00 & 0.71 $\pm$ 0.00 \\
\rowcolor{gray!03}
Gemma4-E2B & Thinking & 0.0 $\pm$ 0.0 & 1.30 $\pm$ 0.02 & 0.82 $\pm$ 0.02 \\
Gemma4-E4B & Non-Thinking & 1.0 $\pm$ 0.0 & 1.01 $\pm$ 0.00 & 0.60 $\pm$ 0.00 \\
\rowcolor{gray!03}
Gemma4-E4B & Thinking & 0.0 $\pm$ 0.0 & 1.22 $\pm$ 0.03 & 0.74 $\pm$ 0.01 \\
Gemma4-26B-A4B & Non-Thinking & 0.0 $\pm$ 0.0 & 1.09 $\pm$ 0.00 & 0.73 $\pm$ 0.00 \\
\rowcolor{gray!03}
Gemma4-26B-A4B & Thinking & 0.0 $\pm$ 0.0 & 1.06 $\pm$ 0.00 & 0.64 $\pm$ 0.00 \\
Gemma4-31B & Non-Thinking & \textbf{4.0 $\pm$ 0.0} & 0.78 $\pm$ 0.00 & 0.41 $\pm$ 0.00 \\
\rowcolor{gray!03}
Gemma4-31B & Thinking & \textbf{7.2 $\pm$ 1.1} & 0.80 $\pm$ 0.01 & 0.40 $\pm$ 0.01 \\
GLM-4.6V-Flash & Non-Thinking & 0.0 $\pm$ 0.0 & 1.45 $\pm$ 0.00 & 0.86 $\pm$ 0.00 \\
\rowcolor{gray!03}
GLM-4.6V-Flash & Thinking & 0.0 $\pm$ 0.0 & 1.33 $\pm$ 0.03 & 0.82 $\pm$ 0.03 \\
Kimi-VL-A3B & Non-Thinking & 0.0 $\pm$ 0.0 & 2.47 $\pm$ 0.02 & 1.68 $\pm$ 0.01 \\
\rowcolor{gray!03}
Kimi-VL-A3B & Thinking & 0.0 $\pm$ 0.0 & 2.24 $\pm$ 0.01 & 1.44 $\pm$ 0.01 \\
\rowcolor{blue!03}
GPT-5 (High) & Thinking & 3.0 & 0.76 & 0.36 \\
\rowcolor{blue!03}
Grok-4.20 (High) & Thinking & 1.0 & 0.89 & 0.45 \\
\rowcolor{blue!03}
Gemini-3.1-Pro-Preview (High) & Thinking & 6.0 & 0.79 & 0.36 \\
\rowcolor{blue!03}
Gemini-3-Flash-Preview (High) & Thinking & 4.0 & 0.81 & 0.35 \\
\midrule
\rowcolor{gray!06}\multicolumn{5}{l}{\textit{Colored + Axes}}\\
Qwen3.5-0.8B & Non-Thinking & 0.0 $\pm$ 0.0 & 2.34 $\pm$ 0.09 & 1.21 $\pm$ 0.05 \\
\rowcolor{gray!03}
Qwen3.5-0.8B & Thinking & 0.0 $\pm$ 0.0 & 2.33 $\pm$ 0.19 & 1.19 $\pm$ 0.10 \\
Qwen3.5-2B & Non-Thinking & 0.0 $\pm$ 0.0 & 1.41 $\pm$ 0.01 & 0.84 $\pm$ 0.01 \\
\rowcolor{gray!03}
Qwen3.5-2B & Thinking & 0.0 $\pm$ 0.0 & 1.49 $\pm$ 0.09 & 0.78 $\pm$ 0.03 \\
Qwen3.5-4B & Non-Thinking & 0.0 $\pm$ 0.0 & 1.08 $\pm$ 0.00 & 0.65 $\pm$ 0.00 \\
\rowcolor{gray!03}
Qwen3.5-4B & Thinking & 0.4 $\pm$ 0.5 & 0.98 $\pm$ 0.02 & 0.52 $\pm$ 0.01 \\
Qwen3.5-9B & Non-Thinking & 0.0 $\pm$ 0.0 & 1.13 $\pm$ 0.00 & 0.76 $\pm$ 0.01 \\
\rowcolor{gray!03}
Qwen3.5-9B & Thinking & 1.4 $\pm$ 0.9 & 0.92 $\pm$ 0.01 & 0.49 $\pm$ 0.01 \\
Qwen3.5-27B & Non-Thinking & \textbf{3.0 $\pm$ 0.0} & 0.92 $\pm$ 0.00 & 0.48 $\pm$ 0.00 \\
\rowcolor{gray!03}
Qwen3.5-27B & Thinking & 1.8 $\pm$ 1.1 & 0.89 $\pm$ 0.00 & 0.42 $\pm$ 0.00 \\
Qwen3.5-35B-A3B & Non-Thinking & 0.0 $\pm$ 0.0 & 1.00 $\pm$ 0.01 & 0.60 $\pm$ 0.01 \\
\rowcolor{gray!03}
Qwen3.5-35B-A3B & Thinking & 1.4 $\pm$ 0.5 & 0.86 $\pm$ 0.01 & 0.43 $\pm$ 0.01 \\
Gemma4-E2B & Non-Thinking & 0.0 $\pm$ 0.0 & 0.98 $\pm$ 0.00 & 0.60 $\pm$ 0.00 \\
\rowcolor{gray!03}
Gemma4-E2B & Thinking & 0.0 $\pm$ 0.0 & 1.12 $\pm$ 0.03 & 0.70 $\pm$ 0.02 \\
Gemma4-E4B & Non-Thinking & 0.0 $\pm$ 0.0 & 1.00 $\pm$ 0.00 & 0.60 $\pm$ 0.00 \\
\rowcolor{gray!03}
Gemma4-E4B & Thinking & 0.0 $\pm$ 0.0 & 1.16 $\pm$ 0.03 & 0.68 $\pm$ 0.01 \\
Gemma4-26B-A4B & Non-Thinking & 0.0 $\pm$ 0.0 & 1.13 $\pm$ 0.00 & 0.75 $\pm$ 0.00 \\
\rowcolor{gray!03}
Gemma4-26B-A4B & Thinking & 0.2 $\pm$ 0.4 & 1.02 $\pm$ 0.03 & 0.61 $\pm$ 0.02 \\
Gemma4-31B & Non-Thinking & 2.4 $\pm$ 0.5 & 0.82 $\pm$ 0.00 & 0.43 $\pm$ 0.00 \\
\rowcolor{gray!03}
Gemma4-31B & Thinking & \textbf{4.2 $\pm$ 1.3} & 0.81 $\pm$ 0.02 & 0.41 $\pm$ 0.01 \\
GLM-4.6V-Flash & Non-Thinking & 0.0 $\pm$ 0.0 & 1.52 $\pm$ 0.00 & 0.90 $\pm$ 0.00 \\
\rowcolor{gray!03}
GLM-4.6V-Flash & Thinking & 0.2 $\pm$ 0.4 & 1.31 $\pm$ 0.02 & 0.80 $\pm$ 0.02 \\
Kimi-VL-A3B & Non-Thinking & 0.0 $\pm$ 0.0 & 3.09 $\pm$ 0.02 & 2.05 $\pm$ 0.01 \\
\rowcolor{gray!03}
Kimi-VL-A3B & Thinking & 0.0 $\pm$ 0.0 & 2.42 $\pm$ 0.04 & 1.52 $\pm$ 0.02 \\
\end{xltabular}
}

\subsection{T6: Local Structure}
\label{sec:app_metrics_t6}
{\small
\setlength{\tabcolsep}{5pt}
\begin{xltabular}{\linewidth}{>{\raggedright\arraybackslash}X lcccc}
\toprule
Model & Mode & Acc (\%) & Prec & Recall & F1 \\
\midrule
\endfirsthead
\multicolumn{6}{l}{\textit{Continued from previous page}}\\
\toprule
Model & Mode & Acc (\%) & Prec & Recall & F1 \\
\midrule
\endhead
\midrule \multicolumn{6}{r}{\textit{Continued on next page}}\\
\endfoot
\bottomrule
\addlinespace[8pt]
\caption{Full per-model metrics for \tsixlong{} across visual variants. Each cell reports the 5-run mean $\pm$ std for one visual-condition variant $\times$ model $\times$ inference mode. Frontier rows appear only for the default visual variant in Thinking mode and report a single run, so no std is shown. Accuracy is reported in percent. \textbf{Boldface} marks the best mean accuracy within each variant/mode block.}\label{tab:app_metrics_t6}\\
\endlastfoot
\rowcolor{gray!06}\multicolumn{6}{l}{\textit{Uncolored}}\\
Qwen3.5-0.8B & Non-Thinking & 47.6 $\pm$ 1.5 & 0.627 $\pm$ 0.012 & 0.641 $\pm$ 0.015 & 0.475 $\pm$ 0.015 \\
\rowcolor{gray!03}
Qwen3.5-0.8B & Thinking & 66.4 $\pm$ 3.0 & 0.614 $\pm$ 0.024 & 0.654 $\pm$ 0.032 & 0.610 $\pm$ 0.029 \\
Qwen3.5-2B & Non-Thinking & 40.8 $\pm$ 1.1 & 0.640 $\pm$ 0.002 & 0.616 $\pm$ 0.007 & 0.406 $\pm$ 0.012 \\
\rowcolor{gray!03}
Qwen3.5-2B & Thinking & 62.2 $\pm$ 2.6 & 0.655 $\pm$ 0.011 & 0.715 $\pm$ 0.017 & 0.604 $\pm$ 0.022 \\
Qwen3.5-4B & Non-Thinking & 86.2 $\pm$ 1.1 & 0.803 $\pm$ 0.012 & 0.868 $\pm$ 0.014 & 0.825 $\pm$ 0.013 \\
\rowcolor{gray!03}
Qwen3.5-4B & Thinking & 78.8 $\pm$ 3.0 & 0.756 $\pm$ 0.021 & 0.853 $\pm$ 0.024 & 0.761 $\pm$ 0.030 \\
Qwen3.5-9B & Non-Thinking & 86.4 $\pm$ 0.5 & 0.925 $\pm$ 0.003 & 0.704 $\pm$ 0.012 & 0.749 $\pm$ 0.013 \\
\rowcolor{gray!03}
Qwen3.5-9B & Thinking & 86.8 $\pm$ 0.8 & 0.814 $\pm$ 0.011 & 0.814 $\pm$ 0.021 & 0.813 $\pm$ 0.014 \\
Qwen3.5-27B & Non-Thinking & \textbf{89.6 $\pm$ 0.5} & 0.849 $\pm$ 0.006 & 0.865 $\pm$ 0.012 & 0.857 $\pm$ 0.009 \\
\rowcolor{gray!03}
Qwen3.5-27B & Thinking & 87.0 $\pm$ 2.8 & 0.813 $\pm$ 0.034 & 0.876 $\pm$ 0.042 & 0.835 $\pm$ 0.036 \\
Qwen3.5-35B-A3B & Non-Thinking & 71.6 $\pm$ 0.5 & 0.695 $\pm$ 0.006 & 0.773 $\pm$ 0.009 & 0.685 $\pm$ 0.006 \\
\rowcolor{gray!03}
Qwen3.5-35B-A3B & Thinking & 82.4 $\pm$ 1.9 & 0.761 $\pm$ 0.018 & 0.816 $\pm$ 0.006 & 0.778 $\pm$ 0.017 \\
Gemma4-E2B & Non-Thinking & 77.6 $\pm$ 0.5 & 0.687 $\pm$ 0.276 & 0.513 $\pm$ 0.012 & 0.462 $\pm$ 0.024 \\
\rowcolor{gray!03}
Gemma4-E2B & Thinking & 74.6 $\pm$ 5.0 & 0.654 $\pm$ 0.060 & 0.661 $\pm$ 0.055 & 0.655 $\pm$ 0.057 \\
Gemma4-E4B & Non-Thinking & 77.0 $\pm$ 0.0 & 0.638 $\pm$ 0.000 & 0.515 $\pm$ 0.000 & 0.474 $\pm$ 0.000 \\
\rowcolor{gray!03}
Gemma4-E4B & Thinking & 75.6 $\pm$ 4.0 & 0.659 $\pm$ 0.051 & 0.662 $\pm$ 0.050 & 0.660 $\pm$ 0.050 \\
Gemma4-26B-A4B & Non-Thinking & 81.0 $\pm$ 0.0 & 0.735 $\pm$ 0.000 & 0.678 $\pm$ 0.000 & 0.697 $\pm$ 0.000 \\
\rowcolor{gray!03}
Gemma4-26B-A4B & Thinking & 73.4 $\pm$ 1.9 & 0.640 $\pm$ 0.025 & 0.657 $\pm$ 0.029 & 0.646 $\pm$ 0.027 \\
Gemma4-31B & Non-Thinking & 83.0 $\pm$ 0.0 & 0.910 $\pm$ 0.000 & 0.630 $\pm$ 0.000 & 0.657 $\pm$ 0.000 \\
\rowcolor{gray!03}
Gemma4-31B & Thinking & 74.8 $\pm$ 1.1 & 0.691 $\pm$ 0.014 & 0.751 $\pm$ 0.020 & 0.700 $\pm$ 0.015 \\
GLM-4.6V-Flash & Non-Thinking & 78.0 $\pm$ 0.0 & 0.683 $\pm$ 0.000 & 0.659 $\pm$ 0.000 & 0.668 $\pm$ 0.000 \\
\rowcolor{gray!03}
GLM-4.6V-Flash & Thinking & \textbf{90.0 $\pm$ 1.0} & 0.867 $\pm$ 0.016 & 0.844 $\pm$ 0.016 & 0.854 $\pm$ 0.015 \\
Kimi-VL-A3B & Non-Thinking & 23.0 $\pm$ 0.0 & 0.115 $\pm$ 0.000 & 0.500 $\pm$ 0.000 & 0.187 $\pm$ 0.000 \\
\rowcolor{gray!03}
Kimi-VL-A3B & Thinking & 23.0 $\pm$ 0.0 & 0.115 $\pm$ 0.000 & 0.500 $\pm$ 0.000 & 0.187 $\pm$ 0.000 \\
\rowcolor{blue!03}
GPT-5 (High) & Thinking & 80.0 & 0.720 & 0.733 & 0.726 \\
\rowcolor{blue!03}
Grok-4.20 (High) & Thinking & 67.0 & 0.693 & 0.770 & 0.652 \\
\rowcolor{blue!03}
Gemini-3.1-Pro-Preview (High) & Thinking & 79.0 & 0.727 & 0.787 & 0.742 \\
\rowcolor{blue!03}
Gemini-3-Flash-Preview (High) & Thinking & 86.0 & 0.800 & 0.818 & 0.808 \\
\midrule
\rowcolor{gray!06}\multicolumn{6}{l}{\textit{Uncolored + Axes}}\\
Qwen3.5-0.8B & Non-Thinking & 34.4 $\pm$ 0.5 & 0.630 $\pm$ 0.001 & 0.574 $\pm$ 0.004 & 0.335 $\pm$ 0.006 \\
\rowcolor{gray!03}
Qwen3.5-0.8B & Thinking & 67.0 $\pm$ 5.3 & 0.590 $\pm$ 0.055 & 0.612 $\pm$ 0.067 & 0.592 $\pm$ 0.061 \\
Qwen3.5-2B & Non-Thinking & 36.4 $\pm$ 0.5 & 0.633 $\pm$ 0.001 & 0.587 $\pm$ 0.004 & 0.358 $\pm$ 0.006 \\
\rowcolor{gray!03}
Qwen3.5-2B & Thinking & 67.8 $\pm$ 4.4 & 0.697 $\pm$ 0.018 & 0.776 $\pm$ 0.029 & 0.659 $\pm$ 0.039 \\
Qwen3.5-4B & Non-Thinking & 85.0 $\pm$ 0.0 & 0.790 $\pm$ 0.000 & 0.857 $\pm$ 0.000 & 0.812 $\pm$ 0.000 \\
\rowcolor{gray!03}
Qwen3.5-4B & Thinking & 72.4 $\pm$ 3.8 & 0.720 $\pm$ 0.021 & 0.809 $\pm$ 0.031 & 0.701 $\pm$ 0.035 \\
Qwen3.5-9B & Non-Thinking & \textbf{92.0 $\pm$ 0.0} & 0.910 $\pm$ 0.000 & 0.857 $\pm$ 0.000 & 0.879 $\pm$ 0.000 \\
\rowcolor{gray!03}
Qwen3.5-9B & Thinking & 85.0 $\pm$ 4.3 & 0.792 $\pm$ 0.050 & 0.854 $\pm$ 0.050 & 0.812 $\pm$ 0.052 \\
Qwen3.5-27B & Non-Thinking & 91.0 $\pm$ 0.0 & 0.864 $\pm$ 0.000 & 0.896 $\pm$ 0.000 & 0.878 $\pm$ 0.000 \\
\rowcolor{gray!03}
Qwen3.5-27B & Thinking & \textbf{88.6 $\pm$ 2.6} & 0.832 $\pm$ 0.030 & 0.899 $\pm$ 0.032 & 0.855 $\pm$ 0.032 \\
Qwen3.5-35B-A3B & Non-Thinking & 77.2 $\pm$ 1.5 & 0.715 $\pm$ 0.017 & 0.782 $\pm$ 0.023 & 0.727 $\pm$ 0.018 \\
\rowcolor{gray!03}
Qwen3.5-35B-A3B & Thinking & 84.8 $\pm$ 2.4 & 0.785 $\pm$ 0.030 & 0.816 $\pm$ 0.030 & 0.797 $\pm$ 0.030 \\
Gemma4-E2B & Non-Thinking & 77.0 $\pm$ 0.0 & 0.385 $\pm$ 0.000 & 0.500 $\pm$ 0.000 & 0.435 $\pm$ 0.000 \\
\rowcolor{gray!03}
Gemma4-E2B & Thinking & 80.0 $\pm$ 3.7 & 0.722 $\pm$ 0.065 & 0.672 $\pm$ 0.047 & 0.687 $\pm$ 0.051 \\
Gemma4-E4B & Non-Thinking & 79.8 $\pm$ 0.4 & 0.872 $\pm$ 0.056 & 0.564 $\pm$ 0.003 & 0.556 $\pm$ 0.003 \\
\rowcolor{gray!03}
Gemma4-E4B & Thinking & 80.4 $\pm$ 3.8 & 0.726 $\pm$ 0.058 & 0.702 $\pm$ 0.043 & 0.711 $\pm$ 0.049 \\
Gemma4-26B-A4B & Non-Thinking & 81.2 $\pm$ 0.4 & 0.769 $\pm$ 0.018 & 0.634 $\pm$ 0.003 & 0.657 $\pm$ 0.004 \\
\rowcolor{gray!03}
Gemma4-26B-A4B & Thinking & 80.8 $\pm$ 0.8 & 0.729 $\pm$ 0.014 & 0.698 $\pm$ 0.011 & 0.711 $\pm$ 0.011 \\
Gemma4-31B & Non-Thinking & 82.0 $\pm$ 0.0 & 0.905 $\pm$ 0.000 & 0.609 $\pm$ 0.000 & 0.626 $\pm$ 0.000 \\
\rowcolor{gray!03}
Gemma4-31B & Thinking & 79.0 $\pm$ 2.0 & 0.713 $\pm$ 0.023 & 0.745 $\pm$ 0.025 & 0.725 $\pm$ 0.024 \\
GLM-4.6V-Flash & Non-Thinking & 77.0 $\pm$ 0.0 & 0.718 $\pm$ 0.000 & 0.790 $\pm$ 0.000 & 0.729 $\pm$ 0.000 \\
\rowcolor{gray!03}
GLM-4.6V-Flash & Thinking & 86.4 $\pm$ 1.1 & 0.804 $\pm$ 0.014 & 0.832 $\pm$ 0.020 & 0.817 $\pm$ 0.016 \\
Kimi-VL-A3B & Non-Thinking & 23.0 $\pm$ 0.0 & 0.115 $\pm$ 0.000 & 0.500 $\pm$ 0.000 & 0.187 $\pm$ 0.000 \\
\rowcolor{gray!03}
Kimi-VL-A3B & Thinking & 23.0 $\pm$ 0.0 & 0.115 $\pm$ 0.000 & 0.500 $\pm$ 0.000 & 0.187 $\pm$ 0.000 \\
\end{xltabular}
}

\subsection{T7: Color Dominance}
\label{sec:app_metrics_t7}
{\small
\setlength{\tabcolsep}{5pt}
\begin{xltabular}{\linewidth}{>{\raggedright\arraybackslash}X lcccc}
\toprule
Model & Mode & Acc (\%) & Prec & Recall & F1 \\
\midrule
\endfirsthead
\multicolumn{6}{l}{\textit{Continued from previous page}}\\
\toprule
Model & Mode & Acc (\%) & Prec & Recall & F1 \\
\midrule
\endhead
\midrule \multicolumn{6}{r}{\textit{Continued on next page}}\\
\endfoot
\bottomrule
\addlinespace[8pt]
\caption{Full per-model metrics for \tsevenlong{} across visual variants. Each cell reports the 5-run mean $\pm$ std for one visual-condition variant $\times$ model $\times$ inference mode. Frontier rows appear only for the default visual variant in Thinking mode and report a single run, so no std is shown. Accuracy is reported in percent. \textbf{Boldface} marks the best mean accuracy within each variant/mode block.}\label{tab:app_metrics_t7}\\
\endlastfoot
\rowcolor{gray!06}\multicolumn{6}{l}{\textit{Colored}}\\
Qwen3.5-0.8B & Non-Thinking & 0.0 $\pm$ 0.0 & 0.297 $\pm$ 0.001 & 0.778 $\pm$ 0.005 & 0.430 $\pm$ 0.002 \\
\rowcolor{gray!03}
Qwen3.5-0.8B & Thinking & 0.0 $\pm$ 0.0 & 0.265 $\pm$ 0.005 & 0.788 $\pm$ 0.021 & 0.397 $\pm$ 0.006 \\
Qwen3.5-2B & Non-Thinking & 0.0 $\pm$ 0.0 & 0.290 $\pm$ 0.001 & 0.824 $\pm$ 0.005 & 0.429 $\pm$ 0.002 \\
\rowcolor{gray!03}
Qwen3.5-2B & Thinking & 0.0 $\pm$ 0.0 & 0.295 $\pm$ 0.003 & 0.795 $\pm$ 0.006 & 0.430 $\pm$ 0.003 \\
Qwen3.5-4B & Non-Thinking & 6.0 $\pm$ 0.7 & 0.340 $\pm$ 0.003 & 0.769 $\pm$ 0.004 & 0.472 $\pm$ 0.002 \\
\rowcolor{gray!03}
Qwen3.5-4B & Thinking & 8.4 $\pm$ 1.7 & 0.377 $\pm$ 0.009 & 0.749 $\pm$ 0.016 & 0.502 $\pm$ 0.007 \\
Qwen3.5-9B & Non-Thinking & 22.2 $\pm$ 0.8 & 0.625 $\pm$ 0.011 & 0.507 $\pm$ 0.009 & 0.560 $\pm$ 0.004 \\
\rowcolor{gray!03}
Qwen3.5-9B & Thinking & 21.4 $\pm$ 2.4 & 0.602 $\pm$ 0.027 & 0.503 $\pm$ 0.015 & 0.548 $\pm$ 0.018 \\
Qwen3.5-27B & Non-Thinking & 18.0 $\pm$ 0.0 & 0.418 $\pm$ 0.003 & 0.848 $\pm$ 0.004 & 0.560 $\pm$ 0.003 \\
\rowcolor{gray!03}
Qwen3.5-27B & Thinking & 24.8 $\pm$ 2.3 & 0.499 $\pm$ 0.011 & 0.744 $\pm$ 0.008 & 0.597 $\pm$ 0.008 \\
Qwen3.5-35B-A3B & Non-Thinking & 29.2 $\pm$ 1.8 & 0.627 $\pm$ 0.020 & 0.646 $\pm$ 0.011 & 0.636 $\pm$ 0.012 \\
\rowcolor{gray!03}
Qwen3.5-35B-A3B & Thinking & 24.4 $\pm$ 1.9 & 0.522 $\pm$ 0.021 & 0.729 $\pm$ 0.012 & 0.609 $\pm$ 0.018 \\
Gemma4-E2B & Non-Thinking & 6.4 $\pm$ 0.5 & 0.382 $\pm$ 0.004 & 0.628 $\pm$ 0.006 & 0.475 $\pm$ 0.005 \\
\rowcolor{gray!03}
Gemma4-E2B & Thinking & 8.0 $\pm$ 2.1 & 0.412 $\pm$ 0.034 & 0.487 $\pm$ 0.055 & 0.446 $\pm$ 0.042 \\
Gemma4-E4B & Non-Thinking & 14.0 $\pm$ 0.0 & 0.487 $\pm$ 0.004 & 0.637 $\pm$ 0.004 & 0.552 $\pm$ 0.003 \\
\rowcolor{gray!03}
Gemma4-E4B & Thinking & 9.2 $\pm$ 2.7 & 0.446 $\pm$ 0.021 & 0.630 $\pm$ 0.015 & 0.522 $\pm$ 0.017 \\
Gemma4-26B-A4B & Non-Thinking & 22.2 $\pm$ 0.4 & 0.651 $\pm$ 0.002 & 0.540 $\pm$ 0.004 & 0.590 $\pm$ 0.003 \\
\rowcolor{gray!03}
Gemma4-26B-A4B & Thinking & 19.8 $\pm$ 0.8 & 0.688 $\pm$ 0.030 & 0.356 $\pm$ 0.009 & 0.469 $\pm$ 0.007 \\
Gemma4-31B & Non-Thinking & \textbf{32.0 $\pm$ 0.0} & 0.697 $\pm$ 0.000 & 0.639 $\pm$ 0.000 & 0.667 $\pm$ 0.000 \\
\rowcolor{gray!03}
Gemma4-31B & Thinking & 27.0 $\pm$ 2.0 & 0.702 $\pm$ 0.021 & 0.515 $\pm$ 0.029 & 0.593 $\pm$ 0.017 \\
GLM-4.6V-Flash & Non-Thinking & 17.6 $\pm$ 0.5 & 0.581 $\pm$ 0.003 & 0.560 $\pm$ 0.005 & 0.570 $\pm$ 0.003 \\
\rowcolor{gray!03}
GLM-4.6V-Flash & Thinking & 17.0 $\pm$ 2.5 & 0.564 $\pm$ 0.025 & 0.621 $\pm$ 0.014 & 0.591 $\pm$ 0.015 \\
Kimi-VL-A3B & Non-Thinking & 0.6 $\pm$ 0.5 & 0.324 $\pm$ 0.003 & 0.798 $\pm$ 0.003 & 0.461 $\pm$ 0.004 \\
\rowcolor{gray!03}
Kimi-VL-A3B & Thinking & 3.6 $\pm$ 0.5 & 0.319 $\pm$ 0.003 & 0.787 $\pm$ 0.003 & 0.454 $\pm$ 0.003 \\
\rowcolor{blue!03}
GPT-5 (High) & Thinking & 24.0 & 0.671 & 0.472 & 0.554 \\
\rowcolor{blue!03}
Grok-4.20 (High) & Thinking & 29.0 & 0.756 & 0.498 & 0.600 \\
\rowcolor{blue!03}
Gemini-3.1-Pro-Preview (High) & Thinking & 35.0 & 0.769 & 0.667 & 0.714 \\
\rowcolor{blue!03}
Gemini-3-Flash-Preview (High) & Thinking & \textbf{37.0} & 0.745 & 0.647 & 0.692 \\
\midrule
\rowcolor{gray!06}\multicolumn{6}{l}{\textit{Colored + Axes}}\\
Qwen3.5-0.8B & Non-Thinking & 0.0 $\pm$ 0.0 & 0.294 $\pm$ 0.001 & 0.745 $\pm$ 0.007 & 0.422 $\pm$ 0.002 \\
\rowcolor{gray!03}
Qwen3.5-0.8B & Thinking & 0.2 $\pm$ 0.4 & 0.254 $\pm$ 0.005 & 0.772 $\pm$ 0.014 & 0.383 $\pm$ 0.006 \\
Qwen3.5-2B & Non-Thinking & 0.0 $\pm$ 0.0 & 0.294 $\pm$ 0.001 & 0.817 $\pm$ 0.002 & 0.432 $\pm$ 0.001 \\
\rowcolor{gray!03}
Qwen3.5-2B & Thinking & 0.4 $\pm$ 0.5 & 0.288 $\pm$ 0.001 & 0.751 $\pm$ 0.008 & 0.417 $\pm$ 0.002 \\
Qwen3.5-4B & Non-Thinking & 12.0 $\pm$ 0.0 & 0.379 $\pm$ 0.007 & 0.757 $\pm$ 0.006 & 0.505 $\pm$ 0.005 \\
\rowcolor{gray!03}
Qwen3.5-4B & Thinking & 8.0 $\pm$ 2.0 & 0.378 $\pm$ 0.010 & 0.758 $\pm$ 0.012 & 0.505 $\pm$ 0.009 \\
Qwen3.5-9B & Non-Thinking & 23.4 $\pm$ 0.5 & 0.700 $\pm$ 0.008 & 0.428 $\pm$ 0.006 & 0.531 $\pm$ 0.007 \\
\rowcolor{gray!03}
Qwen3.5-9B & Thinking & 20.8 $\pm$ 2.0 & 0.604 $\pm$ 0.035 & 0.523 $\pm$ 0.016 & 0.560 $\pm$ 0.018 \\
Qwen3.5-27B & Non-Thinking & 18.4 $\pm$ 0.5 & 0.400 $\pm$ 0.001 & 0.847 $\pm$ 0.004 & 0.543 $\pm$ 0.002 \\
\rowcolor{gray!03}
Qwen3.5-27B & Thinking & 22.8 $\pm$ 2.6 & 0.471 $\pm$ 0.019 & 0.768 $\pm$ 0.015 & 0.584 $\pm$ 0.017 \\
Qwen3.5-35B-A3B & Non-Thinking & \textbf{27.0 $\pm$ 2.8} & 0.675 $\pm$ 0.025 & 0.556 $\pm$ 0.013 & 0.609 $\pm$ 0.016 \\
\rowcolor{gray!03}
Qwen3.5-35B-A3B & Thinking & 23.8 $\pm$ 3.4 & 0.518 $\pm$ 0.011 & 0.730 $\pm$ 0.010 & 0.606 $\pm$ 0.009 \\
Gemma4-E2B & Non-Thinking & 5.8 $\pm$ 0.4 & 0.390 $\pm$ 0.002 & 0.546 $\pm$ 0.005 & 0.455 $\pm$ 0.001 \\
\rowcolor{gray!03}
Gemma4-E2B & Thinking & 7.2 $\pm$ 1.3 & 0.390 $\pm$ 0.010 & 0.473 $\pm$ 0.024 & 0.427 $\pm$ 0.011 \\
Gemma4-E4B & Non-Thinking & 9.2 $\pm$ 0.8 & 0.456 $\pm$ 0.004 & 0.626 $\pm$ 0.003 & 0.528 $\pm$ 0.003 \\
\rowcolor{gray!03}
Gemma4-E4B & Thinking & 9.0 $\pm$ 1.6 & 0.447 $\pm$ 0.005 & 0.598 $\pm$ 0.024 & 0.511 $\pm$ 0.011 \\
Gemma4-26B-A4B & Non-Thinking & 21.2 $\pm$ 0.4 & 0.663 $\pm$ 0.004 & 0.557 $\pm$ 0.005 & 0.606 $\pm$ 0.004 \\
\rowcolor{gray!03}
Gemma4-26B-A4B & Thinking & 22.0 $\pm$ 0.7 & 0.689 $\pm$ 0.013 & 0.324 $\pm$ 0.011 & 0.440 $\pm$ 0.013 \\
Gemma4-31B & Non-Thinking & 25.2 $\pm$ 0.4 & 0.701 $\pm$ 0.002 & 0.547 $\pm$ 0.002 & 0.614 $\pm$ 0.002 \\
\rowcolor{gray!03}
Gemma4-31B & Thinking & \textbf{29.0 $\pm$ 1.7} & 0.695 $\pm$ 0.021 & 0.549 $\pm$ 0.002 & 0.613 $\pm$ 0.009 \\
GLM-4.6V-Flash & Non-Thinking & 19.0 $\pm$ 0.0 & 0.601 $\pm$ 0.002 & 0.545 $\pm$ 0.003 & 0.572 $\pm$ 0.002 \\
\rowcolor{gray!03}
GLM-4.6V-Flash & Thinking & 17.0 $\pm$ 1.2 & 0.582 $\pm$ 0.027 & 0.605 $\pm$ 0.027 & 0.593 $\pm$ 0.025 \\
Kimi-VL-A3B & Non-Thinking & 3.0 $\pm$ 0.0 & 0.324 $\pm$ 0.003 & 0.725 $\pm$ 0.002 & 0.448 $\pm$ 0.002 \\
\rowcolor{gray!03}
Kimi-VL-A3B & Thinking & 3.0 $\pm$ 0.0 & 0.310 $\pm$ 0.005 & 0.759 $\pm$ 0.011 & 0.440 $\pm$ 0.004 \\
\end{xltabular}
}

\subsection{T8: Silhouette Symmetry}
\label{sec:app_metrics_t8}
{\small
\setlength{\tabcolsep}{5pt}
\begin{xltabular}{\linewidth}{>{\raggedright\arraybackslash}X lcccc}
\toprule
Model & Mode & Acc (\%) & Prec & Recall & F1 \\
\midrule
\endfirsthead
\multicolumn{6}{l}{\textit{Continued from previous page}}\\
\toprule
Model & Mode & Acc (\%) & Prec & Recall & F1 \\
\midrule
\endhead
\midrule \multicolumn{6}{r}{\textit{Continued on next page}}\\
\endfoot
\bottomrule
\addlinespace[8pt]
\caption{Full per-model metrics for \teightlong{} across visual variants. Each cell reports the 5-run mean $\pm$ std for one visual-condition variant $\times$ model $\times$ inference mode. Frontier rows appear only for the default visual variant in Thinking mode and report a single run, so no std is shown. Accuracy is reported in percent. \textbf{Boldface} marks the best mean accuracy within each variant/mode block.}\label{tab:app_metrics_t8}\\
\endlastfoot
\rowcolor{gray!06}\multicolumn{6}{l}{\textit{Uncolored}}\\
Qwen3.5-0.8B & Non-Thinking & 58.0 $\pm$ 0.0 & 0.290 $\pm$ 0.000 & 0.500 $\pm$ 0.000 & 0.367 $\pm$ 0.000 \\
\rowcolor{gray!03}
Qwen3.5-0.8B & Thinking & 52.2 $\pm$ 1.6 & 0.369 $\pm$ 0.053 & 0.456 $\pm$ 0.018 & 0.374 $\pm$ 0.027 \\
Qwen3.5-2B & Non-Thinking & 59.0 $\pm$ 0.0 & 0.793 $\pm$ 0.000 & 0.512 $\pm$ 0.000 & 0.393 $\pm$ 0.000 \\
\rowcolor{gray!03}
Qwen3.5-2B & Thinking & 54.2 $\pm$ 6.9 & 0.496 $\pm$ 0.098 & 0.497 $\pm$ 0.070 & 0.472 $\pm$ 0.077 \\
Qwen3.5-4B & Non-Thinking & 58.0 $\pm$ 0.0 & 0.290 $\pm$ 0.000 & 0.500 $\pm$ 0.000 & 0.367 $\pm$ 0.000 \\
\rowcolor{gray!03}
Qwen3.5-4B & Thinking & 57.8 $\pm$ 2.3 & 0.532 $\pm$ 0.117 & 0.507 $\pm$ 0.024 & 0.420 $\pm$ 0.034 \\
Qwen3.5-9B & Non-Thinking & 58.0 $\pm$ 0.0 & 0.290 $\pm$ 0.000 & 0.500 $\pm$ 0.000 & 0.367 $\pm$ 0.000 \\
\rowcolor{gray!03}
Qwen3.5-9B & Thinking & 58.8 $\pm$ 0.8 & 0.609 $\pm$ 0.209 & 0.511 $\pm$ 0.010 & 0.395 $\pm$ 0.020 \\
Qwen3.5-27B & Non-Thinking & 57.0 $\pm$ 0.0 & 0.288 $\pm$ 0.000 & 0.491 $\pm$ 0.000 & 0.363 $\pm$ 0.000 \\
\rowcolor{gray!03}
Qwen3.5-27B & Thinking & 59.8 $\pm$ 1.3 & 0.617 $\pm$ 0.057 & 0.530 $\pm$ 0.014 & 0.456 $\pm$ 0.022 \\
Qwen3.5-35B-A3B & Non-Thinking & 58.0 $\pm$ 0.0 & 0.290 $\pm$ 0.000 & 0.500 $\pm$ 0.000 & 0.367 $\pm$ 0.000 \\
\rowcolor{gray!03}
Qwen3.5-35B-A3B & Thinking & 58.2 $\pm$ 2.7 & 0.556 $\pm$ 0.126 & 0.512 $\pm$ 0.027 & 0.426 $\pm$ 0.035 \\
Gemma4-E2B & Non-Thinking & 57.0 $\pm$ 0.0 & 0.288 $\pm$ 0.000 & 0.491 $\pm$ 0.000 & 0.363 $\pm$ 0.000 \\
\rowcolor{gray!03}
Gemma4-E2B & Thinking & 55.4 $\pm$ 2.1 & 0.489 $\pm$ 0.048 & 0.495 $\pm$ 0.022 & 0.443 $\pm$ 0.029 \\
Gemma4-E4B & Non-Thinking & 58.0 $\pm$ 0.0 & 0.290 $\pm$ 0.000 & 0.500 $\pm$ 0.000 & 0.367 $\pm$ 0.000 \\
\rowcolor{gray!03}
Gemma4-E4B & Thinking & 56.6 $\pm$ 1.8 & 0.527 $\pm$ 0.028 & 0.516 $\pm$ 0.017 & 0.486 $\pm$ 0.016 \\
Gemma4-26B-A4B & Non-Thinking & \textbf{61.0 $\pm$ 0.0} & 0.612 $\pm$ 0.000 & 0.552 $\pm$ 0.000 & 0.511 $\pm$ 0.000 \\
\rowcolor{gray!03}
Gemma4-26B-A4B & Thinking & 50.4 $\pm$ 2.3 & 0.475 $\pm$ 0.029 & 0.477 $\pm$ 0.026 & 0.472 $\pm$ 0.029 \\
Gemma4-31B & Non-Thinking & 56.0 $\pm$ 0.0 & 0.286 $\pm$ 0.000 & 0.483 $\pm$ 0.000 & 0.359 $\pm$ 0.000 \\
\rowcolor{gray!03}
Gemma4-31B & Thinking & 58.2 $\pm$ 2.0 & 0.554 $\pm$ 0.039 & 0.526 $\pm$ 0.018 & 0.485 $\pm$ 0.017 \\
GLM-4.6V-Flash & Non-Thinking & 58.0 $\pm$ 0.0 & 0.290 $\pm$ 0.000 & 0.500 $\pm$ 0.000 & 0.367 $\pm$ 0.000 \\
\rowcolor{gray!03}
GLM-4.6V-Flash & Thinking & 59.0 $\pm$ 0.0 & 0.793 $\pm$ 0.000 & 0.512 $\pm$ 0.000 & 0.393 $\pm$ 0.000 \\
Kimi-VL-A3B & Non-Thinking & 60.8 $\pm$ 1.6 & 0.592 $\pm$ 0.018 & 0.585 $\pm$ 0.017 & 0.585 $\pm$ 0.017 \\
\rowcolor{gray!03}
Kimi-VL-A3B & Thinking & 43.6 $\pm$ 1.9 & 0.503 $\pm$ 0.179 & 0.511 $\pm$ 0.017 & 0.338 $\pm$ 0.035 \\
\rowcolor{blue!03}
GPT-5 (High) & Thinking & \textbf{60.0} & 0.631 & 0.530 & 0.452 \\
\rowcolor{blue!03}
Grok-4.20 (High) & Thinking & 57.0 & 0.455 & 0.495 & 0.384 \\
\rowcolor{blue!03}
Gemini-3.1-Pro-Preview (High) & Thinking & 51.0 & 0.493 & 0.493 & 0.492 \\
\rowcolor{blue!03}
Gemini-3-Flash-Preview (High) & Thinking & \textbf{60.0} & 0.585 & 0.547 & 0.514 \\
\midrule
\rowcolor{gray!06}\multicolumn{6}{l}{\textit{Uncolored + Axes}}\\
Qwen3.5-0.8B & Non-Thinking & 58.0 $\pm$ 0.0 & 0.290 $\pm$ 0.000 & 0.500 $\pm$ 0.000 & 0.367 $\pm$ 0.000 \\
\rowcolor{gray!03}
Qwen3.5-0.8B & Thinking & 56.6 $\pm$ 1.7 & 0.421 $\pm$ 0.131 & 0.492 $\pm$ 0.018 & 0.385 $\pm$ 0.031 \\
Qwen3.5-2B & Non-Thinking & 57.4 $\pm$ 0.5 & 0.389 $\pm$ 0.139 & 0.496 $\pm$ 0.007 & 0.373 $\pm$ 0.014 \\
\rowcolor{gray!03}
Qwen3.5-2B & Thinking & 51.6 $\pm$ 6.2 & 0.492 $\pm$ 0.067 & 0.492 $\pm$ 0.063 & 0.489 $\pm$ 0.064 \\
Qwen3.5-4B & Non-Thinking & 58.0 $\pm$ 0.0 & 0.290 $\pm$ 0.000 & 0.500 $\pm$ 0.000 & 0.367 $\pm$ 0.000 \\
\rowcolor{gray!03}
Qwen3.5-4B & Thinking & 56.0 $\pm$ 1.2 & 0.415 $\pm$ 0.095 & 0.487 $\pm$ 0.013 & 0.382 $\pm$ 0.020 \\
Qwen3.5-9B & Non-Thinking & 57.2 $\pm$ 0.4 & 0.288 $\pm$ 0.001 & 0.493 $\pm$ 0.004 & 0.364 $\pm$ 0.002 \\
\rowcolor{gray!03}
Qwen3.5-9B & Thinking & 59.8 $\pm$ 1.6 & 0.696 $\pm$ 0.228 & 0.522 $\pm$ 0.018 & 0.416 $\pm$ 0.032 \\
Qwen3.5-27B & Non-Thinking & \textbf{60.2 $\pm$ 0.4} & 0.596 $\pm$ 0.010 & 0.543 $\pm$ 0.006 & 0.498 $\pm$ 0.012 \\
\rowcolor{gray!03}
Qwen3.5-27B & Thinking & 58.0 $\pm$ 2.1 & 0.546 $\pm$ 0.044 & 0.522 $\pm$ 0.022 & 0.477 $\pm$ 0.028 \\
Qwen3.5-35B-A3B & Non-Thinking & \textbf{60.2 $\pm$ 1.5} & 0.637 $\pm$ 0.066 & 0.535 $\pm$ 0.014 & 0.464 $\pm$ 0.020 \\
\rowcolor{gray!03}
Qwen3.5-35B-A3B & Thinking & 59.8 $\pm$ 4.7 & 0.585 $\pm$ 0.094 & 0.546 $\pm$ 0.047 & 0.516 $\pm$ 0.055 \\
Gemma4-E2B & Non-Thinking & 58.0 $\pm$ 0.0 & 0.541 $\pm$ 0.000 & 0.503 $\pm$ 0.000 & 0.388 $\pm$ 0.000 \\
\rowcolor{gray!03}
Gemma4-E2B & Thinking & 58.2 $\pm$ 3.1 & 0.554 $\pm$ 0.086 & 0.519 $\pm$ 0.031 & 0.461 $\pm$ 0.036 \\
Gemma4-E4B & Non-Thinking & 58.0 $\pm$ 0.0 & 0.290 $\pm$ 0.000 & 0.500 $\pm$ 0.000 & 0.367 $\pm$ 0.000 \\
\rowcolor{gray!03}
Gemma4-E4B & Thinking & \textbf{65.4 $\pm$ 3.8} & 0.678 $\pm$ 0.074 & 0.606 $\pm$ 0.036 & 0.589 $\pm$ 0.039 \\
Gemma4-26B-A4B & Non-Thinking & 55.0 $\pm$ 0.0 & 0.513 $\pm$ 0.000 & 0.510 $\pm$ 0.000 & 0.495 $\pm$ 0.000 \\
\rowcolor{gray!03}
Gemma4-26B-A4B & Thinking & 47.8 $\pm$ 1.3 & 0.453 $\pm$ 0.014 & 0.455 $\pm$ 0.013 & 0.453 $\pm$ 0.014 \\
Gemma4-31B & Non-Thinking & 54.0 $\pm$ 0.0 & 0.365 $\pm$ 0.000 & 0.469 $\pm$ 0.000 & 0.370 $\pm$ 0.000 \\
\rowcolor{gray!03}
Gemma4-31B & Thinking & 54.8 $\pm$ 2.3 & 0.499 $\pm$ 0.036 & 0.499 $\pm$ 0.024 & 0.470 $\pm$ 0.028 \\
GLM-4.6V-Flash & Non-Thinking & 58.0 $\pm$ 0.0 & 0.290 $\pm$ 0.000 & 0.500 $\pm$ 0.000 & 0.367 $\pm$ 0.000 \\
\rowcolor{gray!03}
GLM-4.6V-Flash & Thinking & 59.0 $\pm$ 0.7 & 0.693 $\pm$ 0.225 & 0.512 $\pm$ 0.008 & 0.392 $\pm$ 0.018 \\
Kimi-VL-A3B & Non-Thinking & 50.4 $\pm$ 3.6 & 0.535 $\pm$ 0.042 & 0.532 $\pm$ 0.037 & 0.500 $\pm$ 0.036 \\
\rowcolor{gray!03}
Kimi-VL-A3B & Thinking & 42.0 $\pm$ 0.0 & 0.210 $\pm$ 0.000 & 0.500 $\pm$ 0.000 & 0.296 $\pm$ 0.000 \\
\end{xltabular}
}

\subsection{T9: Visible Floors}
\label{sec:app_metrics_t9}
{\small
\setlength{\tabcolsep}{5pt}
\begin{xltabular}{\linewidth}{>{\raggedright\arraybackslash}X lccc}
\toprule
Model & Mode & Acc (\%) & RMSE & MAE \\
\midrule
\endfirsthead
\multicolumn{5}{l}{\textit{Continued from previous page}}\\
\toprule
Model & Mode & Acc (\%) & RMSE & MAE \\
\midrule
\endhead
\midrule \multicolumn{5}{r}{\textit{Continued on next page}}\\
\endfoot
\bottomrule
\addlinespace[8pt]
\caption{Full per-model metrics for \tninelong{} across visual variants. Each cell reports the 5-run mean $\pm$ std for one visual-condition variant $\times$ model $\times$ inference mode. Frontier rows appear only for the default visual variant in Thinking mode and report a single run, so no std is shown. Accuracy is reported in percent; RMSE and MAE are in task units. \textbf{Boldface} marks the best mean accuracy within each variant/mode block.}\label{tab:app_metrics_t9}\\
\endlastfoot
\rowcolor{gray!06}\multicolumn{5}{l}{\textit{Uncolored}}\\
Qwen3.5-0.8B & Non-Thinking & 45.6 $\pm$ 1.3 & 0.80 $\pm$ 0.02 & 0.58 $\pm$ 0.02 \\
\rowcolor{gray!03}
Qwen3.5-0.8B & Thinking & 27.0 $\pm$ 2.6 & 3.14 $\pm$ 0.18 & 2.01 $\pm$ 0.14 \\
Qwen3.5-2B & Non-Thinking & 59.2 $\pm$ 0.8 & 0.78 $\pm$ 0.01 & 0.47 $\pm$ 0.01 \\
\rowcolor{gray!03}
Qwen3.5-2B & Thinking & 43.4 $\pm$ 2.7 & 3.33 $\pm$ 1.55 & 1.15 $\pm$ 0.22 \\
Qwen3.5-4B & Non-Thinking & \textbf{74.0 $\pm$ 0.7} & 0.51 $\pm$ 0.01 & 0.26 $\pm$ 0.01 \\
\rowcolor{gray!03}
Qwen3.5-4B & Thinking & 65.8 $\pm$ 2.4 & 0.72 $\pm$ 0.08 & 0.38 $\pm$ 0.02 \\
Qwen3.5-9B & Non-Thinking & 66.4 $\pm$ 0.5 & 0.63 $\pm$ 0.00 & 0.36 $\pm$ 0.01 \\
\rowcolor{gray!03}
Qwen3.5-9B & Thinking & 68.0 $\pm$ 2.5 & 0.63 $\pm$ 0.02 & 0.34 $\pm$ 0.02 \\
Qwen3.5-27B & Non-Thinking & 71.8 $\pm$ 0.4 & 0.56 $\pm$ 0.00 & 0.29 $\pm$ 0.00 \\
\rowcolor{gray!03}
Qwen3.5-27B & Thinking & 76.6 $\pm$ 2.1 & 0.54 $\pm$ 0.01 & 0.25 $\pm$ 0.02 \\
Qwen3.5-35B-A3B & Non-Thinking & 67.4 $\pm$ 2.1 & 0.57 $\pm$ 0.02 & 0.33 $\pm$ 0.02 \\
\rowcolor{gray!03}
Qwen3.5-35B-A3B & Thinking & 69.0 $\pm$ 2.1 & 0.58 $\pm$ 0.02 & 0.32 $\pm$ 0.02 \\
Gemma4-E2B & Non-Thinking & 50.6 $\pm$ 0.5 & 0.72 $\pm$ 0.00 & 0.50 $\pm$ 0.01 \\
\rowcolor{gray!03}
Gemma4-E2B & Thinking & 22.2 $\pm$ 1.5 & 1.25 $\pm$ 0.03 & 1.02 $\pm$ 0.03 \\
Gemma4-E4B & Non-Thinking & 51.0 $\pm$ 0.0 & 0.88 $\pm$ 0.00 & 0.58 $\pm$ 0.00 \\
\rowcolor{gray!03}
Gemma4-E4B & Thinking & 46.4 $\pm$ 4.0 & 0.92 $\pm$ 0.06 & 0.62 $\pm$ 0.05 \\
Gemma4-26B-A4B & Non-Thinking & 44.8 $\pm$ 0.4 & 0.90 $\pm$ 0.00 & 0.63 $\pm$ 0.00 \\
\rowcolor{gray!03}
Gemma4-26B-A4B & Thinking & 55.4 $\pm$ 1.1 & 0.80 $\pm$ 0.02 & 0.51 $\pm$ 0.01 \\
Gemma4-31B & Non-Thinking & 61.0 $\pm$ 0.0 & 0.71 $\pm$ 0.00 & 0.43 $\pm$ 0.00 \\
\rowcolor{gray!03}
Gemma4-31B & Thinking & 42.6 $\pm$ 0.5 & 1.08 $\pm$ 0.02 & 0.73 $\pm$ 0.01 \\
GLM-4.6V-Flash & Non-Thinking & 61.2 $\pm$ 0.4 & 0.79 $\pm$ 0.00 & 0.46 $\pm$ 0.00 \\
\rowcolor{gray!03}
GLM-4.6V-Flash & Thinking & 62.2 $\pm$ 1.1 & 0.64 $\pm$ 0.03 & 0.39 $\pm$ 0.02 \\
Kimi-VL-A3B & Non-Thinking & 17.0 $\pm$ 0.0 & 1.26 $\pm$ 0.00 & 1.08 $\pm$ 0.00 \\
\rowcolor{gray!03}
Kimi-VL-A3B & Thinking & 2.0 $\pm$ 0.0 & 3.07 $\pm$ 0.01 & 2.92 $\pm$ 0.01 \\
\rowcolor{blue!03}
GPT-5 (High) & Thinking & 51.0 & 0.80 & 0.54 \\
\rowcolor{blue!03}
Grok-4.20 (High) & Thinking & 27.0 & 1.09 & 0.88 \\
\rowcolor{blue!03}
Gemini-3.1-Pro-Preview (High) & Thinking & \textbf{77.0} & 0.59 & 0.27 \\
\rowcolor{blue!03}
Gemini-3-Flash-Preview (High) & Thinking & 63.0 & 0.79 & 0.44 \\
\midrule
\rowcolor{gray!06}\multicolumn{5}{l}{\textit{Uncolored + Axes}}\\
Qwen3.5-0.8B & Non-Thinking & 41.0 $\pm$ 1.2 & 1.06 $\pm$ 0.01 & 0.77 $\pm$ 0.01 \\
\rowcolor{gray!03}
Qwen3.5-0.8B & Thinking & 26.2 $\pm$ 3.6 & 3.70 $\pm$ 0.78 & 2.19 $\pm$ 0.24 \\
Qwen3.5-2B & Non-Thinking & 50.6 $\pm$ 0.5 & 0.83 $\pm$ 0.00 & 0.55 $\pm$ 0.01 \\
\rowcolor{gray!03}
Qwen3.5-2B & Thinking & 47.6 $\pm$ 6.1 & 2.40 $\pm$ 0.77 & 1.04 $\pm$ 0.17 \\
Qwen3.5-4B & Non-Thinking & 74.0 $\pm$ 0.0 & 0.54 $\pm$ 0.01 & 0.27 $\pm$ 0.00 \\
\rowcolor{gray!03}
Qwen3.5-4B & Thinking & 67.8 $\pm$ 3.5 & 0.62 $\pm$ 0.04 & 0.34 $\pm$ 0.04 \\
Qwen3.5-9B & Non-Thinking & 63.6 $\pm$ 0.5 & 0.63 $\pm$ 0.00 & 0.37 $\pm$ 0.01 \\
\rowcolor{gray!03}
Qwen3.5-9B & Thinking & 68.2 $\pm$ 3.3 & 0.61 $\pm$ 0.04 & 0.34 $\pm$ 0.03 \\
Qwen3.5-27B & Non-Thinking & \textbf{79.0 $\pm$ 0.0} & 0.52 $\pm$ 0.00 & 0.23 $\pm$ 0.00 \\
\rowcolor{gray!03}
Qwen3.5-27B & Thinking & \textbf{79.4 $\pm$ 1.1} & 0.49 $\pm$ 0.02 & 0.22 $\pm$ 0.01 \\
Qwen3.5-35B-A3B & Non-Thinking & 59.6 $\pm$ 0.9 & 0.66 $\pm$ 0.01 & 0.41 $\pm$ 0.01 \\
\rowcolor{gray!03}
Qwen3.5-35B-A3B & Thinking & 69.8 $\pm$ 4.0 & 0.56 $\pm$ 0.05 & 0.31 $\pm$ 0.04 \\
Gemma4-E2B & Non-Thinking & 47.4 $\pm$ 0.5 & 0.77 $\pm$ 0.00 & 0.55 $\pm$ 0.01 \\
\rowcolor{gray!03}
Gemma4-E2B & Thinking & 38.8 $\pm$ 2.9 & 1.01 $\pm$ 0.03 & 0.73 $\pm$ 0.03 \\
Gemma4-E4B & Non-Thinking & 47.0 $\pm$ 0.0 & 0.85 $\pm$ 0.00 & 0.59 $\pm$ 0.00 \\
\rowcolor{gray!03}
Gemma4-E4B & Thinking & 48.4 $\pm$ 4.3 & 0.82 $\pm$ 0.05 & 0.56 $\pm$ 0.05 \\
Gemma4-26B-A4B & Non-Thinking & 52.4 $\pm$ 0.5 & 0.71 $\pm$ 0.00 & 0.49 $\pm$ 0.01 \\
\rowcolor{gray!03}
Gemma4-26B-A4B & Thinking & 39.8 $\pm$ 1.9 & 0.92 $\pm$ 0.02 & 0.67 $\pm$ 0.02 \\
Gemma4-31B & Non-Thinking & 56.0 $\pm$ 0.0 & 0.69 $\pm$ 0.01 & 0.45 $\pm$ 0.00 \\
\rowcolor{gray!03}
Gemma4-31B & Thinking & 48.4 $\pm$ 0.5 & 0.95 $\pm$ 0.01 & 0.63 $\pm$ 0.01 \\
GLM-4.6V-Flash & Non-Thinking & 61.2 $\pm$ 0.4 & 0.73 $\pm$ 0.00 & 0.44 $\pm$ 0.00 \\
\rowcolor{gray!03}
GLM-4.6V-Flash & Thinking & 68.0 $\pm$ 1.9 & 0.59 $\pm$ 0.02 & 0.33 $\pm$ 0.02 \\
Kimi-VL-A3B & Non-Thinking & 16.0 $\pm$ 0.0 & 1.37 $\pm$ 0.00 & 1.19 $\pm$ 0.00 \\
\rowcolor{gray!03}
Kimi-VL-A3B & Thinking & 3.0 $\pm$ 0.0 & 2.99 $\pm$ 0.01 & 2.80 $\pm$ 0.01 \\
\end{xltabular}
}

\subsection{T10: Color Connected}
\label{sec:app_metrics_t10}
{\small
\setlength{\tabcolsep}{5pt}
\begin{xltabular}{\linewidth}{>{\raggedright\arraybackslash}X lcccc}
\toprule
Model & Mode & Acc (\%) & Prec & Recall & F1 \\
\midrule
\endfirsthead
\multicolumn{6}{l}{\textit{Continued from previous page}}\\
\toprule
Model & Mode & Acc (\%) & Prec & Recall & F1 \\
\midrule
\endhead
\midrule \multicolumn{6}{r}{\textit{Continued on next page}}\\
\endfoot
\bottomrule
\addlinespace[8pt]
\caption{Full per-model metrics for \ttenlong{} across visual variants. Each cell reports the 5-run mean $\pm$ std for one visual-condition variant $\times$ model $\times$ inference mode. Frontier rows appear only for the default visual variant in Thinking mode and report a single run, so no std is shown. Accuracy is reported in percent. \textbf{Boldface} marks the best mean accuracy within each variant/mode block.}\label{tab:app_metrics_t10}\\
\endlastfoot
\rowcolor{gray!06}\multicolumn{6}{l}{\textit{Colored}}\\
Qwen3.5-0.8B & Non-Thinking & 49.0 $\pm$ 0.0 & 0.245 $\pm$ 0.000 & 0.500 $\pm$ 0.000 & 0.329 $\pm$ 0.000 \\
\rowcolor{gray!03}
Qwen3.5-0.8B & Thinking & 46.4 $\pm$ 3.2 & 0.463 $\pm$ 0.033 & 0.463 $\pm$ 0.032 & 0.462 $\pm$ 0.033 \\
Qwen3.5-2B & Non-Thinking & 49.0 $\pm$ 0.0 & 0.245 $\pm$ 0.000 & 0.500 $\pm$ 0.000 & 0.329 $\pm$ 0.000 \\
\rowcolor{gray!03}
Qwen3.5-2B & Thinking & 49.0 $\pm$ 0.0 & 0.245 $\pm$ 0.000 & 0.500 $\pm$ 0.000 & 0.329 $\pm$ 0.000 \\
Qwen3.5-4B & Non-Thinking & 55.4 $\pm$ 1.3 & 0.554 $\pm$ 0.014 & 0.552 $\pm$ 0.014 & 0.547 $\pm$ 0.016 \\
\rowcolor{gray!03}
Qwen3.5-4B & Thinking & 53.4 $\pm$ 1.1 & 0.706 $\pm$ 0.069 & 0.543 $\pm$ 0.011 & 0.428 $\pm$ 0.022 \\
Qwen3.5-9B & Non-Thinking & 50.2 $\pm$ 0.4 & 0.748 $\pm$ 0.001 & 0.512 $\pm$ 0.004 & 0.354 $\pm$ 0.009 \\
\rowcolor{gray!03}
Qwen3.5-9B & Thinking & 56.8 $\pm$ 2.2 & 0.723 $\pm$ 0.052 & 0.576 $\pm$ 0.022 & 0.487 $\pm$ 0.029 \\
Qwen3.5-27B & Non-Thinking & 55.2 $\pm$ 0.4 & 0.761 $\pm$ 0.001 & 0.561 $\pm$ 0.004 & 0.451 $\pm$ 0.008 \\
\rowcolor{gray!03}
Qwen3.5-27B & Thinking & 65.6 $\pm$ 2.3 & 0.744 $\pm$ 0.029 & 0.662 $\pm$ 0.023 & 0.626 $\pm$ 0.027 \\
Qwen3.5-35B-A3B & Non-Thinking & 55.6 $\pm$ 0.5 & 0.663 $\pm$ 0.016 & 0.564 $\pm$ 0.005 & 0.481 $\pm$ 0.007 \\
\rowcolor{gray!03}
Qwen3.5-35B-A3B & Thinking & 65.6 $\pm$ 2.3 & 0.674 $\pm$ 0.030 & 0.659 $\pm$ 0.023 & 0.650 $\pm$ 0.022 \\
Gemma4-E2B & Non-Thinking & 53.0 $\pm$ 0.0 & 0.760 $\pm$ 0.000 & 0.520 $\pm$ 0.000 & 0.381 $\pm$ 0.000 \\
\rowcolor{gray!03}
Gemma4-E2B & Thinking & 59.0 $\pm$ 3.2 & 0.590 $\pm$ 0.032 & 0.589 $\pm$ 0.032 & 0.589 $\pm$ 0.032 \\
Gemma4-E4B & Non-Thinking & 51.0 $\pm$ 0.0 & 0.255 $\pm$ 0.000 & 0.500 $\pm$ 0.000 & 0.338 $\pm$ 0.000 \\
\rowcolor{gray!03}
Gemma4-E4B & Thinking & 53.2 $\pm$ 4.2 & 0.535 $\pm$ 0.059 & 0.526 $\pm$ 0.043 & 0.489 $\pm$ 0.053 \\
Gemma4-26B-A4B & Non-Thinking & 55.0 $\pm$ 0.0 & 0.766 $\pm$ 0.000 & 0.541 $\pm$ 0.000 & 0.422 $\pm$ 0.000 \\
\rowcolor{gray!03}
Gemma4-26B-A4B & Thinking & 56.8 $\pm$ 1.6 & 0.576 $\pm$ 0.018 & 0.564 $\pm$ 0.017 & 0.547 $\pm$ 0.020 \\
Gemma4-31B & Non-Thinking & \textbf{61.4 $\pm$ 0.5} & 0.679 $\pm$ 0.005 & 0.608 $\pm$ 0.006 & 0.568 $\pm$ 0.008 \\
\rowcolor{gray!03}
Gemma4-31B & Thinking & 58.2 $\pm$ 1.5 & 0.596 $\pm$ 0.015 & 0.585 $\pm$ 0.015 & 0.571 $\pm$ 0.016 \\
GLM-4.6V-Flash & Non-Thinking & 53.0 $\pm$ 0.0 & 0.635 $\pm$ 0.000 & 0.521 $\pm$ 0.000 & 0.397 $\pm$ 0.000 \\
\rowcolor{gray!03}
GLM-4.6V-Flash & Thinking & 59.8 $\pm$ 2.5 & 0.598 $\pm$ 0.025 & 0.598 $\pm$ 0.025 & 0.598 $\pm$ 0.025 \\
Kimi-VL-A3B & Non-Thinking & 49.0 $\pm$ 0.0 & 0.245 $\pm$ 0.000 & 0.500 $\pm$ 0.000 & 0.329 $\pm$ 0.000 \\
\rowcolor{gray!03}
Kimi-VL-A3B & Thinking & 49.0 $\pm$ 0.0 & 0.245 $\pm$ 0.000 & 0.500 $\pm$ 0.000 & 0.329 $\pm$ 0.000 \\
\rowcolor{blue!03}
GPT-5 (High) & Thinking & 65.0 & 0.682 & 0.662 & 0.648 \\
\rowcolor{blue!03}
Grok-4.20 (High) & Thinking & 62.0 & 0.658 & 0.625 & 0.601 \\
\rowcolor{blue!03}
Gemini-3.1-Pro-Preview (High) & Thinking & 57.0 & 0.629 & 0.576 & 0.524 \\
\rowcolor{blue!03}
Gemini-3-Flash-Preview (High) & Thinking & \textbf{67.0} & 0.756 & 0.676 & 0.644 \\
\midrule
\rowcolor{gray!06}\multicolumn{6}{l}{\textit{Colored + Axes}}\\
Qwen3.5-0.8B & Non-Thinking & 49.0 $\pm$ 0.0 & 0.245 $\pm$ 0.000 & 0.500 $\pm$ 0.000 & 0.329 $\pm$ 0.000 \\
\rowcolor{gray!03}
Qwen3.5-0.8B & Thinking & 48.4 $\pm$ 2.7 & 0.475 $\pm$ 0.033 & 0.479 $\pm$ 0.026 & 0.448 $\pm$ 0.022 \\
Qwen3.5-2B & Non-Thinking & 49.0 $\pm$ 0.0 & 0.245 $\pm$ 0.000 & 0.500 $\pm$ 0.000 & 0.329 $\pm$ 0.000 \\
\rowcolor{gray!03}
Qwen3.5-2B & Thinking & 49.0 $\pm$ 0.0 & 0.245 $\pm$ 0.000 & 0.500 $\pm$ 0.000 & 0.329 $\pm$ 0.000 \\
Qwen3.5-4B & Non-Thinking & 58.6 $\pm$ 1.5 & 0.614 $\pm$ 0.018 & 0.591 $\pm$ 0.015 & 0.565 $\pm$ 0.017 \\
\rowcolor{gray!03}
Qwen3.5-4B & Thinking & 54.0 $\pm$ 2.0 & 0.729 $\pm$ 0.041 & 0.549 $\pm$ 0.020 & 0.433 $\pm$ 0.037 \\
Qwen3.5-9B & Non-Thinking & 49.0 $\pm$ 0.0 & 0.245 $\pm$ 0.000 & 0.500 $\pm$ 0.000 & 0.329 $\pm$ 0.000 \\
\rowcolor{gray!03}
Qwen3.5-9B & Thinking & 51.0 $\pm$ 0.7 & 0.692 $\pm$ 0.083 & 0.519 $\pm$ 0.007 & 0.377 $\pm$ 0.011 \\
Qwen3.5-27B & Non-Thinking & 53.4 $\pm$ 1.3 & 0.716 $\pm$ 0.058 & 0.543 $\pm$ 0.013 & 0.423 $\pm$ 0.019 \\
\rowcolor{gray!03}
Qwen3.5-27B & Thinking & \textbf{67.0 $\pm$ 1.6} & 0.738 $\pm$ 0.024 & 0.675 $\pm$ 0.016 & 0.649 $\pm$ 0.016 \\
Qwen3.5-35B-A3B & Non-Thinking & 52.0 $\pm$ 1.0 & 0.620 $\pm$ 0.046 & 0.529 $\pm$ 0.010 & 0.415 $\pm$ 0.015 \\
\rowcolor{gray!03}
Qwen3.5-35B-A3B & Thinking & 64.4 $\pm$ 3.4 & 0.685 $\pm$ 0.047 & 0.648 $\pm$ 0.034 & 0.628 $\pm$ 0.034 \\
Gemma4-E2B & Non-Thinking & 51.0 $\pm$ 0.0 & 0.505 $\pm$ 0.000 & 0.500 $\pm$ 0.000 & 0.355 $\pm$ 0.000 \\
\rowcolor{gray!03}
Gemma4-E2B & Thinking & 56.0 $\pm$ 3.7 & 0.560 $\pm$ 0.037 & 0.559 $\pm$ 0.037 & 0.559 $\pm$ 0.037 \\
Gemma4-E4B & Non-Thinking & 51.0 $\pm$ 0.0 & 0.255 $\pm$ 0.000 & 0.500 $\pm$ 0.000 & 0.338 $\pm$ 0.000 \\
\rowcolor{gray!03}
Gemma4-E4B & Thinking & 53.8 $\pm$ 1.9 & 0.564 $\pm$ 0.037 & 0.531 $\pm$ 0.019 & 0.466 $\pm$ 0.024 \\
Gemma4-26B-A4B & Non-Thinking & 52.0 $\pm$ 0.0 & 0.758 $\pm$ 0.000 & 0.510 $\pm$ 0.000 & 0.360 $\pm$ 0.000 \\
\rowcolor{gray!03}
Gemma4-26B-A4B & Thinking & 58.8 $\pm$ 1.8 & 0.612 $\pm$ 0.021 & 0.583 $\pm$ 0.018 & 0.556 $\pm$ 0.023 \\
Gemma4-31B & Non-Thinking & \textbf{65.4 $\pm$ 0.5} & 0.706 $\pm$ 0.011 & 0.649 $\pm$ 0.005 & 0.626 $\pm$ 0.005 \\
\rowcolor{gray!03}
Gemma4-31B & Thinking & 56.8 $\pm$ 3.1 & 0.577 $\pm$ 0.031 & 0.571 $\pm$ 0.031 & 0.560 $\pm$ 0.034 \\
GLM-4.6V-Flash & Non-Thinking & 53.4 $\pm$ 0.5 & 0.543 $\pm$ 0.008 & 0.528 $\pm$ 0.006 & 0.487 $\pm$ 0.007 \\
\rowcolor{gray!03}
GLM-4.6V-Flash & Thinking & 57.2 $\pm$ 3.2 & 0.577 $\pm$ 0.035 & 0.574 $\pm$ 0.032 & 0.568 $\pm$ 0.032 \\
Kimi-VL-A3B & Non-Thinking & 49.0 $\pm$ 0.0 & 0.245 $\pm$ 0.000 & 0.500 $\pm$ 0.000 & 0.329 $\pm$ 0.000 \\
\rowcolor{gray!03}
Kimi-VL-A3B & Thinking & 49.0 $\pm$ 0.0 & 0.245 $\pm$ 0.000 & 0.500 $\pm$ 0.000 & 0.329 $\pm$ 0.000 \\
\end{xltabular}
}

\subsection{T11: Projection}
\label{sec:app_metrics_t11}
{\small
\setlength{\tabcolsep}{5pt}
\begin{xltabular}{\linewidth}{>{\raggedright\arraybackslash}X lcccc}
\toprule
Model & Mode & Acc (\%) & Prec & Recall & F1 \\
\midrule
\endfirsthead
\multicolumn{6}{l}{\textit{Continued from previous page}}\\
\toprule
Model & Mode & Acc (\%) & Prec & Recall & F1 \\
\midrule
\endhead
\midrule \multicolumn{6}{r}{\textit{Continued on next page}}\\
\endfoot
\bottomrule
\addlinespace[8pt]
\caption{Full per-model metrics for \televenlong{} across visual variants. Each cell reports the 5-run mean $\pm$ std for one visual-condition variant $\times$ model $\times$ inference mode. Frontier rows appear only for the default visual variant in Thinking mode and report a single run, so no std is shown. Accuracy is reported in percent. \textbf{Boldface} marks the best mean accuracy within each variant/mode block.}\label{tab:app_metrics_t11}\\
\endlastfoot
\rowcolor{gray!06}\multicolumn{6}{l}{\textit{Negative / Uncolored}}\\
Qwen3.5-0.8B & Non-Thinking & 37.3 $\pm$ 0.0 & 0.124 $\pm$ 0.000 & 0.333 $\pm$ 0.000 & 0.181 $\pm$ 0.000 \\
\rowcolor{gray!03}
Qwen3.5-0.8B & Thinking & 37.3 $\pm$ 0.6 & 0.393 $\pm$ 0.057 & 0.336 $\pm$ 0.006 & 0.213 $\pm$ 0.009 \\
Qwen3.5-2B & Non-Thinking & 37.3 $\pm$ 0.0 & 0.124 $\pm$ 0.000 & 0.333 $\pm$ 0.000 & 0.181 $\pm$ 0.000 \\
\rowcolor{gray!03}
Qwen3.5-2B & Thinking & 39.6 $\pm$ 1.8 & 0.373 $\pm$ 0.141 & 0.364 $\pm$ 0.016 & 0.295 $\pm$ 0.013 \\
Qwen3.5-4B & Non-Thinking & 42.5 $\pm$ 0.5 & 0.379 $\pm$ 0.077 & 0.388 $\pm$ 0.005 & 0.310 $\pm$ 0.007 \\
\rowcolor{gray!03}
Qwen3.5-4B & Thinking & 37.1 $\pm$ 2.5 & 0.368 $\pm$ 0.071 & 0.339 $\pm$ 0.026 & 0.254 $\pm$ 0.038 \\
Qwen3.5-9B & Non-Thinking & 45.7 $\pm$ 0.3 & 0.412 $\pm$ 0.006 & 0.418 $\pm$ 0.003 & 0.333 $\pm$ 0.003 \\
\rowcolor{gray!03}
Qwen3.5-9B & Thinking & 41.7 $\pm$ 0.7 & 0.489 $\pm$ 0.198 & 0.379 $\pm$ 0.007 & 0.281 $\pm$ 0.013 \\
Qwen3.5-27B & Non-Thinking & \textbf{55.7 $\pm$ 0.5} & 0.577 $\pm$ 0.008 & 0.537 $\pm$ 0.005 & 0.530 $\pm$ 0.005 \\
\rowcolor{gray!03}
Qwen3.5-27B & Thinking & \textbf{50.5 $\pm$ 1.3} & 0.584 $\pm$ 0.016 & 0.478 $\pm$ 0.013 & 0.453 $\pm$ 0.016 \\
Qwen3.5-35B-A3B & Non-Thinking & 40.4 $\pm$ 0.5 & 0.375 $\pm$ 0.033 & 0.364 $\pm$ 0.005 & 0.246 $\pm$ 0.007 \\
\rowcolor{gray!03}
Qwen3.5-35B-A3B & Thinking & 41.5 $\pm$ 1.6 & 0.464 $\pm$ 0.043 & 0.382 $\pm$ 0.016 & 0.318 $\pm$ 0.022 \\
Gemma4-E2B & Non-Thinking & 36.3 $\pm$ 0.0 & 0.189 $\pm$ 0.000 & 0.325 $\pm$ 0.000 & 0.184 $\pm$ 0.000 \\
\rowcolor{gray!03}
Gemma4-E2B & Thinking & 33.2 $\pm$ 1.1 & 0.325 $\pm$ 0.009 & 0.331 $\pm$ 0.012 & 0.323 $\pm$ 0.010 \\
Gemma4-E4B & Non-Thinking & 41.5 $\pm$ 0.2 & 0.447 $\pm$ 0.002 & 0.427 $\pm$ 0.002 & 0.389 $\pm$ 0.002 \\
\rowcolor{gray!03}
Gemma4-E4B & Thinking & 45.5 $\pm$ 0.7 & 0.487 $\pm$ 0.006 & 0.464 $\pm$ 0.007 & 0.447 $\pm$ 0.006 \\
Gemma4-26B-A4B & Non-Thinking & 46.3 $\pm$ 0.3 & 0.497 $\pm$ 0.003 & 0.463 $\pm$ 0.003 & 0.451 $\pm$ 0.004 \\
\rowcolor{gray!03}
Gemma4-26B-A4B & Thinking & 45.2 $\pm$ 2.3 & 0.483 $\pm$ 0.023 & 0.457 $\pm$ 0.021 & 0.447 $\pm$ 0.026 \\
Gemma4-31B & Non-Thinking & 47.3 $\pm$ 0.0 & 0.491 $\pm$ 0.000 & 0.483 $\pm$ 0.000 & 0.471 $\pm$ 0.000 \\
\rowcolor{gray!03}
Gemma4-31B & Thinking & 48.6 $\pm$ 2.3 & 0.514 $\pm$ 0.017 & 0.500 $\pm$ 0.023 & 0.482 $\pm$ 0.024 \\
GLM-4.6V-Flash & Non-Thinking & 45.4 $\pm$ 0.4 & 0.430 $\pm$ 0.003 & 0.425 $\pm$ 0.004 & 0.370 $\pm$ 0.004 \\
\rowcolor{gray!03}
GLM-4.6V-Flash & Thinking & 44.4 $\pm$ 1.8 & 0.360 $\pm$ 0.070 & 0.349 $\pm$ 0.076 & 0.325 $\pm$ 0.067 \\
Kimi-VL-A3B & Non-Thinking & 41.3 $\pm$ 1.0 & 0.431 $\pm$ 0.011 & 0.409 $\pm$ 0.008 & 0.396 $\pm$ 0.007 \\
\rowcolor{gray!03}
Kimi-VL-A3B & Thinking & 37.3 $\pm$ 0.0 & 0.124 $\pm$ 0.000 & 0.333 $\pm$ 0.000 & 0.181 $\pm$ 0.000 \\
\rowcolor{blue!03}
GPT-5 (High) & Thinking & 33.0 & 0.378 & 0.381 & 0.379 \\
\rowcolor{blue!03}
Grok-4.20 (High) & Thinking & 44.0 & 0.440 & 0.436 & 0.436 \\
\rowcolor{blue!03}
Gemini-3.1-Pro-Preview (High) & Thinking & 39.3 & 0.396 & 0.381 & 0.372 \\
\rowcolor{blue!03}
Gemini-3-Flash-Preview (High) & Thinking & 49.3 & 0.499 & 0.501 & 0.499 \\
\midrule
\rowcolor{gray!06}\multicolumn{6}{l}{\textit{Negative / Uncolored + Axes}}\\
Qwen3.5-0.8B & Non-Thinking & 37.3 $\pm$ 0.0 & 0.124 $\pm$ 0.000 & 0.333 $\pm$ 0.000 & 0.181 $\pm$ 0.000 \\
\rowcolor{gray!03}
Qwen3.5-0.8B & Thinking & 37.5 $\pm$ 0.6 & 0.406 $\pm$ 0.069 & 0.338 $\pm$ 0.007 & 0.216 $\pm$ 0.017 \\
Qwen3.5-2B & Non-Thinking & 37.3 $\pm$ 0.0 & 0.124 $\pm$ 0.000 & 0.333 $\pm$ 0.000 & 0.181 $\pm$ 0.000 \\
\rowcolor{gray!03}
Qwen3.5-2B & Thinking & 39.8 $\pm$ 2.1 & 0.329 $\pm$ 0.066 & 0.367 $\pm$ 0.019 & 0.304 $\pm$ 0.016 \\
Qwen3.5-4B & Non-Thinking & 43.9 $\pm$ 0.4 & 0.300 $\pm$ 0.003 & 0.410 $\pm$ 0.003 & 0.344 $\pm$ 0.003 \\
\rowcolor{gray!03}
Qwen3.5-4B & Thinking & 41.5 $\pm$ 1.6 & 0.436 $\pm$ 0.028 & 0.384 $\pm$ 0.017 & 0.332 $\pm$ 0.024 \\
Qwen3.5-9B & Non-Thinking & 46.4 $\pm$ 0.3 & 0.329 $\pm$ 0.003 & 0.430 $\pm$ 0.003 & 0.357 $\pm$ 0.003 \\
\rowcolor{gray!03}
Qwen3.5-9B & Thinking & 48.0 $\pm$ 0.9 & 0.571 $\pm$ 0.188 & 0.442 $\pm$ 0.009 & 0.369 $\pm$ 0.013 \\
Qwen3.5-27B & Non-Thinking & \textbf{57.7 $\pm$ 0.9} & 0.579 $\pm$ 0.009 & 0.570 $\pm$ 0.008 & 0.571 $\pm$ 0.009 \\
\rowcolor{gray!03}
Qwen3.5-27B & Thinking & \textbf{55.9 $\pm$ 1.1} & 0.577 $\pm$ 0.015 & 0.539 $\pm$ 0.010 & 0.533 $\pm$ 0.009 \\
Qwen3.5-35B-A3B & Non-Thinking & 44.6 $\pm$ 0.7 & 0.661 $\pm$ 0.004 & 0.409 $\pm$ 0.007 & 0.331 $\pm$ 0.011 \\
\rowcolor{gray!03}
Qwen3.5-35B-A3B & Thinking & 42.7 $\pm$ 1.5 & 0.424 $\pm$ 0.024 & 0.396 $\pm$ 0.014 & 0.342 $\pm$ 0.014 \\
Gemma4-E2B & Non-Thinking & 38.0 $\pm$ 0.0 & 0.316 $\pm$ 0.000 & 0.343 $\pm$ 0.000 & 0.209 $\pm$ 0.000 \\
\rowcolor{gray!03}
Gemma4-E2B & Thinking & 35.9 $\pm$ 2.1 & 0.359 $\pm$ 0.020 & 0.357 $\pm$ 0.021 & 0.342 $\pm$ 0.021 \\
Gemma4-E4B & Non-Thinking & 39.7 $\pm$ 0.3 & 0.405 $\pm$ 0.005 & 0.404 $\pm$ 0.003 & 0.380 $\pm$ 0.004 \\
\rowcolor{gray!03}
Gemma4-E4B & Thinking & 43.3 $\pm$ 1.3 & 0.453 $\pm$ 0.015 & 0.442 $\pm$ 0.014 & 0.428 $\pm$ 0.013 \\
Gemma4-26B-A4B & Non-Thinking & 49.4 $\pm$ 0.6 & 0.520 $\pm$ 0.007 & 0.494 $\pm$ 0.007 & 0.487 $\pm$ 0.007 \\
\rowcolor{gray!03}
Gemma4-26B-A4B & Thinking & 42.4 $\pm$ 1.5 & 0.443 $\pm$ 0.014 & 0.426 $\pm$ 0.015 & 0.418 $\pm$ 0.016 \\
Gemma4-31B & Non-Thinking & 48.7 $\pm$ 0.0 & 0.497 $\pm$ 0.000 & 0.494 $\pm$ 0.000 & 0.484 $\pm$ 0.000 \\
\rowcolor{gray!03}
Gemma4-31B & Thinking & 48.7 $\pm$ 1.8 & 0.514 $\pm$ 0.012 & 0.499 $\pm$ 0.018 & 0.485 $\pm$ 0.018 \\
GLM-4.6V-Flash & Non-Thinking & 43.0 $\pm$ 0.0 & 0.388 $\pm$ 0.000 & 0.406 $\pm$ 0.000 & 0.368 $\pm$ 0.000 \\
\rowcolor{gray!03}
GLM-4.6V-Flash & Thinking & 41.9 $\pm$ 1.7 & 0.329 $\pm$ 0.074 & 0.330 $\pm$ 0.073 & 0.309 $\pm$ 0.070 \\
Kimi-VL-A3B & Non-Thinking & 41.3 $\pm$ 1.4 & 0.428 $\pm$ 0.017 & 0.411 $\pm$ 0.014 & 0.413 $\pm$ 0.015 \\
\rowcolor{gray!03}
Kimi-VL-A3B & Thinking & 37.3 $\pm$ 0.0 & 0.124 $\pm$ 0.000 & 0.333 $\pm$ 0.000 & 0.181 $\pm$ 0.000 \\
\midrule
\rowcolor{gray!06}\multicolumn{6}{l}{\textit{Negative / Colored}}\\
Qwen3.5-0.8B & Non-Thinking & 37.3 $\pm$ 0.0 & 0.124 $\pm$ 0.000 & 0.333 $\pm$ 0.000 & 0.181 $\pm$ 0.000 \\
\rowcolor{gray!03}
Qwen3.5-0.8B & Thinking & 37.5 $\pm$ 1.1 & 0.348 $\pm$ 0.089 & 0.337 $\pm$ 0.012 & 0.216 $\pm$ 0.019 \\
Qwen3.5-2B & Non-Thinking & 37.3 $\pm$ 0.0 & 0.124 $\pm$ 0.000 & 0.333 $\pm$ 0.000 & 0.181 $\pm$ 0.000 \\
\rowcolor{gray!03}
Qwen3.5-2B & Thinking & 36.8 $\pm$ 1.9 & 0.330 $\pm$ 0.077 & 0.334 $\pm$ 0.017 & 0.236 $\pm$ 0.021 \\
Qwen3.5-4B & Non-Thinking & 39.1 $\pm$ 0.7 & 0.309 $\pm$ 0.014 & 0.354 $\pm$ 0.007 & 0.255 $\pm$ 0.008 \\
\rowcolor{gray!03}
Qwen3.5-4B & Thinking & 37.4 $\pm$ 2.7 & 0.397 $\pm$ 0.091 & 0.342 $\pm$ 0.026 & 0.251 $\pm$ 0.035 \\
Qwen3.5-9B & Non-Thinking & 44.4 $\pm$ 0.4 & 0.430 $\pm$ 0.009 & 0.404 $\pm$ 0.004 & 0.314 $\pm$ 0.005 \\
\rowcolor{gray!03}
Qwen3.5-9B & Thinking & 43.1 $\pm$ 0.9 & 0.515 $\pm$ 0.146 & 0.393 $\pm$ 0.009 & 0.303 $\pm$ 0.012 \\
Qwen3.5-27B & Non-Thinking & \textbf{57.3 $\pm$ 0.1} & 0.696 $\pm$ 0.002 & 0.541 $\pm$ 0.001 & 0.513 $\pm$ 0.001 \\
\rowcolor{gray!03}
Qwen3.5-27B & Thinking & \textbf{54.0 $\pm$ 0.6} & 0.661 $\pm$ 0.034 & 0.507 $\pm$ 0.007 & 0.470 $\pm$ 0.012 \\
Qwen3.5-35B-A3B & Non-Thinking & 41.2 $\pm$ 0.8 & 0.505 $\pm$ 0.152 & 0.372 $\pm$ 0.008 & 0.255 $\pm$ 0.015 \\
\rowcolor{gray!03}
Qwen3.5-35B-A3B & Thinking & 41.7 $\pm$ 1.3 & 0.457 $\pm$ 0.054 & 0.382 $\pm$ 0.013 & 0.302 $\pm$ 0.019 \\
Gemma4-E2B & Non-Thinking & 37.3 $\pm$ 0.0 & 0.258 $\pm$ 0.000 & 0.335 $\pm$ 0.000 & 0.195 $\pm$ 0.000 \\
\rowcolor{gray!03}
Gemma4-E2B & Thinking & 40.5 $\pm$ 2.2 & 0.408 $\pm$ 0.032 & 0.406 $\pm$ 0.024 & 0.387 $\pm$ 0.027 \\
Gemma4-E4B & Non-Thinking & 51.9 $\pm$ 0.3 & 0.637 $\pm$ 0.002 & 0.535 $\pm$ 0.003 & 0.485 $\pm$ 0.002 \\
\rowcolor{gray!03}
Gemma4-E4B & Thinking & 53.9 $\pm$ 1.0 & 0.630 $\pm$ 0.006 & 0.563 $\pm$ 0.011 & 0.522 $\pm$ 0.010 \\
Gemma4-26B-A4B & Non-Thinking & 44.6 $\pm$ 0.4 & 0.480 $\pm$ 0.003 & 0.453 $\pm$ 0.004 & 0.441 $\pm$ 0.005 \\
\rowcolor{gray!03}
Gemma4-26B-A4B & Thinking & 40.7 $\pm$ 2.4 & 0.432 $\pm$ 0.028 & 0.414 $\pm$ 0.026 & 0.403 $\pm$ 0.022 \\
Gemma4-31B & Non-Thinking & 45.7 $\pm$ 0.0 & 0.465 $\pm$ 0.000 & 0.477 $\pm$ 0.000 & 0.440 $\pm$ 0.000 \\
\rowcolor{gray!03}
Gemma4-31B & Thinking & 50.3 $\pm$ 1.6 & 0.500 $\pm$ 0.056 & 0.496 $\pm$ 0.051 & 0.465 $\pm$ 0.045 \\
GLM-4.6V-Flash & Non-Thinking & 54.0 $\pm$ 0.0 & 0.688 $\pm$ 0.000 & 0.510 $\pm$ 0.000 & 0.448 $\pm$ 0.000 \\
\rowcolor{gray!03}
GLM-4.6V-Flash & Thinking & 48.8 $\pm$ 2.0 & 0.299 $\pm$ 0.059 & 0.288 $\pm$ 0.055 & 0.269 $\pm$ 0.052 \\
Kimi-VL-A3B & Non-Thinking & 38.9 $\pm$ 0.8 & 0.422 $\pm$ 0.011 & 0.372 $\pm$ 0.008 & 0.342 $\pm$ 0.010 \\
\rowcolor{gray!03}
Kimi-VL-A3B & Thinking & 37.3 $\pm$ 0.0 & 0.124 $\pm$ 0.000 & 0.333 $\pm$ 0.000 & 0.181 $\pm$ 0.000 \\
\midrule
\rowcolor{gray!06}\multicolumn{6}{l}{\textit{Positive / Uncolored}}\\
Qwen3.5-0.8B & Non-Thinking & 35.0 $\pm$ 0.0 & 0.117 $\pm$ 0.000 & 0.333 $\pm$ 0.000 & 0.173 $\pm$ 0.000 \\
\rowcolor{gray!03}
Qwen3.5-0.8B & Thinking & 33.7 $\pm$ 0.8 & 0.309 $\pm$ 0.037 & 0.323 $\pm$ 0.009 & 0.213 $\pm$ 0.013 \\
Qwen3.5-2B & Non-Thinking & 35.0 $\pm$ 0.0 & 0.117 $\pm$ 0.000 & 0.333 $\pm$ 0.000 & 0.173 $\pm$ 0.000 \\
\rowcolor{gray!03}
Qwen3.5-2B & Thinking & 35.7 $\pm$ 0.7 & 0.302 $\pm$ 0.109 & 0.345 $\pm$ 0.007 & 0.256 $\pm$ 0.011 \\
Qwen3.5-4B & Non-Thinking & 36.5 $\pm$ 1.0 & 0.375 $\pm$ 0.010 & 0.365 $\pm$ 0.010 & 0.366 $\pm$ 0.010 \\
\rowcolor{gray!03}
Qwen3.5-4B & Thinking & 37.5 $\pm$ 1.8 & 0.373 $\pm$ 0.018 & 0.372 $\pm$ 0.018 & 0.367 $\pm$ 0.018 \\
Qwen3.5-9B & Non-Thinking & 35.7 $\pm$ 0.4 & 0.247 $\pm$ 0.007 & 0.344 $\pm$ 0.004 & 0.238 $\pm$ 0.004 \\
\rowcolor{gray!03}
Qwen3.5-9B & Thinking & 35.8 $\pm$ 0.6 & 0.483 $\pm$ 0.125 & 0.342 $\pm$ 0.007 & 0.199 $\pm$ 0.011 \\
Qwen3.5-27B & Non-Thinking & 35.1 $\pm$ 0.5 & 0.338 $\pm$ 0.011 & 0.340 $\pm$ 0.005 & 0.276 $\pm$ 0.005 \\
\rowcolor{gray!03}
Qwen3.5-27B & Thinking & 41.8 $\pm$ 0.7 & 0.446 $\pm$ 0.021 & 0.407 $\pm$ 0.007 & 0.349 $\pm$ 0.009 \\
Qwen3.5-35B-A3B & Non-Thinking & 41.3 $\pm$ 0.5 & 0.480 $\pm$ 0.026 & 0.413 $\pm$ 0.004 & 0.360 $\pm$ 0.003 \\
\rowcolor{gray!03}
Qwen3.5-35B-A3B & Thinking & 42.5 $\pm$ 1.5 & 0.462 $\pm$ 0.025 & 0.420 $\pm$ 0.015 & 0.377 $\pm$ 0.017 \\
Gemma4-E2B & Non-Thinking & 34.8 $\pm$ 0.5 & 0.316 $\pm$ 0.150 & 0.333 $\pm$ 0.005 & 0.216 $\pm$ 0.006 \\
\rowcolor{gray!03}
Gemma4-E2B & Thinking & 35.5 $\pm$ 0.3 & 0.346 $\pm$ 0.008 & 0.351 $\pm$ 0.004 & 0.344 $\pm$ 0.006 \\
Gemma4-E4B & Non-Thinking & \textbf{49.6 $\pm$ 0.4} & 0.517 $\pm$ 0.004 & 0.502 $\pm$ 0.004 & 0.487 $\pm$ 0.004 \\
\rowcolor{gray!03}
Gemma4-E4B & Thinking & \textbf{48.3 $\pm$ 0.8} & 0.523 $\pm$ 0.007 & 0.489 $\pm$ 0.008 & 0.468 $\pm$ 0.007 \\
Gemma4-26B-A4B & Non-Thinking & 37.3 $\pm$ 0.3 & 0.397 $\pm$ 0.004 & 0.376 $\pm$ 0.003 & 0.349 $\pm$ 0.003 \\
\rowcolor{gray!03}
Gemma4-26B-A4B & Thinking & 38.9 $\pm$ 1.2 & 0.494 $\pm$ 0.054 & 0.394 $\pm$ 0.012 & 0.326 $\pm$ 0.015 \\
Gemma4-31B & Non-Thinking & 44.1 $\pm$ 0.2 & 0.475 $\pm$ 0.003 & 0.448 $\pm$ 0.002 & 0.425 $\pm$ 0.003 \\
\rowcolor{gray!03}
Gemma4-31B & Thinking & 41.5 $\pm$ 1.7 & 0.418 $\pm$ 0.057 & 0.401 $\pm$ 0.059 & 0.384 $\pm$ 0.059 \\
GLM-4.6V-Flash & Non-Thinking & 43.9 $\pm$ 0.4 & 0.517 $\pm$ 0.004 & 0.438 $\pm$ 0.004 & 0.411 $\pm$ 0.004 \\
\rowcolor{gray!03}
GLM-4.6V-Flash & Thinking & 46.9 $\pm$ 1.6 & 0.277 $\pm$ 0.024 & 0.262 $\pm$ 0.023 & 0.262 $\pm$ 0.022 \\
Kimi-VL-A3B & Non-Thinking & 35.5 $\pm$ 1.3 & 0.437 $\pm$ 0.041 & 0.357 $\pm$ 0.012 & 0.251 $\pm$ 0.018 \\
\rowcolor{gray!03}
Kimi-VL-A3B & Thinking & 35.0 $\pm$ 0.0 & 0.117 $\pm$ 0.000 & 0.333 $\pm$ 0.000 & 0.173 $\pm$ 0.000 \\
\midrule
\rowcolor{gray!06}\multicolumn{6}{l}{\textit{Positive / Uncolored + Axes}}\\
Qwen3.5-0.8B & Non-Thinking & 35.0 $\pm$ 0.0 & 0.117 $\pm$ 0.000 & 0.333 $\pm$ 0.000 & 0.173 $\pm$ 0.000 \\
\rowcolor{gray!03}
Qwen3.5-0.8B & Thinking & 35.3 $\pm$ 1.5 & 0.371 $\pm$ 0.056 & 0.340 $\pm$ 0.015 & 0.249 $\pm$ 0.019 \\
Qwen3.5-2B & Non-Thinking & 35.0 $\pm$ 0.0 & 0.117 $\pm$ 0.000 & 0.333 $\pm$ 0.000 & 0.173 $\pm$ 0.000 \\
\rowcolor{gray!03}
Qwen3.5-2B & Thinking & 36.3 $\pm$ 2.3 & 0.307 $\pm$ 0.048 & 0.354 $\pm$ 0.023 & 0.274 $\pm$ 0.019 \\
Qwen3.5-4B & Non-Thinking & 41.9 $\pm$ 1.0 & 0.429 $\pm$ 0.010 & 0.422 $\pm$ 0.010 & 0.418 $\pm$ 0.011 \\
\rowcolor{gray!03}
Qwen3.5-4B & Thinking & 36.7 $\pm$ 1.1 & 0.360 $\pm$ 0.012 & 0.366 $\pm$ 0.011 & 0.353 $\pm$ 0.011 \\
Qwen3.5-9B & Non-Thinking & 36.4 $\pm$ 0.3 & 0.327 $\pm$ 0.146 & 0.350 $\pm$ 0.003 & 0.234 $\pm$ 0.003 \\
\rowcolor{gray!03}
Qwen3.5-9B & Thinking & 37.2 $\pm$ 1.1 & 0.537 $\pm$ 0.106 & 0.356 $\pm$ 0.012 & 0.229 $\pm$ 0.019 \\
Qwen3.5-27B & Non-Thinking & 39.5 $\pm$ 0.6 & 0.416 $\pm$ 0.011 & 0.386 $\pm$ 0.005 & 0.339 $\pm$ 0.005 \\
\rowcolor{gray!03}
Qwen3.5-27B & Thinking & 44.9 $\pm$ 0.6 & 0.477 $\pm$ 0.014 & 0.439 $\pm$ 0.006 & 0.400 $\pm$ 0.009 \\
Qwen3.5-35B-A3B & Non-Thinking & 41.1 $\pm$ 1.1 & 0.445 $\pm$ 0.023 & 0.412 $\pm$ 0.010 & 0.373 $\pm$ 0.014 \\
\rowcolor{gray!03}
Qwen3.5-35B-A3B & Thinking & 46.7 $\pm$ 1.0 & 0.475 $\pm$ 0.008 & 0.465 $\pm$ 0.010 & 0.421 $\pm$ 0.005 \\
Gemma4-E2B & Non-Thinking & 36.5 $\pm$ 0.5 & 0.469 $\pm$ 0.003 & 0.357 $\pm$ 0.005 & 0.299 $\pm$ 0.004 \\
\rowcolor{gray!03}
Gemma4-E2B & Thinking & 40.3 $\pm$ 2.4 & 0.398 $\pm$ 0.025 & 0.403 $\pm$ 0.024 & 0.398 $\pm$ 0.025 \\
Gemma4-E4B & Non-Thinking & \textbf{48.4 $\pm$ 0.3} & 0.512 $\pm$ 0.003 & 0.490 $\pm$ 0.003 & 0.471 $\pm$ 0.003 \\
\rowcolor{gray!03}
Gemma4-E4B & Thinking & 47.3 $\pm$ 1.4 & 0.520 $\pm$ 0.019 & 0.480 $\pm$ 0.015 & 0.452 $\pm$ 0.013 \\
Gemma4-26B-A4B & Non-Thinking & 38.3 $\pm$ 0.1 & 0.445 $\pm$ 0.002 & 0.387 $\pm$ 0.002 & 0.334 $\pm$ 0.001 \\
\rowcolor{gray!03}
Gemma4-26B-A4B & Thinking & 37.3 $\pm$ 1.5 & 0.453 $\pm$ 0.028 & 0.378 $\pm$ 0.015 & 0.310 $\pm$ 0.017 \\
Gemma4-31B & Non-Thinking & 46.3 $\pm$ 0.0 & 0.522 $\pm$ 0.000 & 0.471 $\pm$ 0.000 & 0.439 $\pm$ 0.000 \\
\rowcolor{gray!03}
Gemma4-31B & Thinking & 41.4 $\pm$ 1.6 & 0.451 $\pm$ 0.018 & 0.420 $\pm$ 0.017 & 0.394 $\pm$ 0.017 \\
GLM-4.6V-Flash & Non-Thinking & 41.6 $\pm$ 0.1 & 0.441 $\pm$ 0.001 & 0.415 $\pm$ 0.001 & 0.402 $\pm$ 0.001 \\
\rowcolor{gray!03}
GLM-4.6V-Flash & Thinking & \textbf{47.7 $\pm$ 0.6} & 0.354 $\pm$ 0.127 & 0.342 $\pm$ 0.124 & 0.339 $\pm$ 0.120 \\
Kimi-VL-A3B & Non-Thinking & 35.3 $\pm$ 0.1 & 0.560 $\pm$ 0.047 & 0.356 $\pm$ 0.001 & 0.216 $\pm$ 0.003 \\
\rowcolor{gray!03}
Kimi-VL-A3B & Thinking & 35.0 $\pm$ 0.0 & 0.117 $\pm$ 0.000 & 0.333 $\pm$ 0.000 & 0.173 $\pm$ 0.000 \\
\midrule
\rowcolor{gray!06}\multicolumn{6}{l}{\textit{Positive / Colored}}\\
Qwen3.5-0.8B & Non-Thinking & 35.0 $\pm$ 0.0 & 0.117 $\pm$ 0.000 & 0.333 $\pm$ 0.000 & 0.173 $\pm$ 0.000 \\
\rowcolor{gray!03}
Qwen3.5-0.8B & Thinking & 34.3 $\pm$ 1.2 & 0.303 $\pm$ 0.033 & 0.330 $\pm$ 0.011 & 0.228 $\pm$ 0.013 \\
Qwen3.5-2B & Non-Thinking & 35.0 $\pm$ 0.0 & 0.117 $\pm$ 0.000 & 0.333 $\pm$ 0.000 & 0.173 $\pm$ 0.000 \\
\rowcolor{gray!03}
Qwen3.5-2B & Thinking & 35.9 $\pm$ 0.9 & 0.414 $\pm$ 0.086 & 0.346 $\pm$ 0.009 & 0.242 $\pm$ 0.017 \\
Qwen3.5-4B & Non-Thinking & 34.1 $\pm$ 0.3 & 0.331 $\pm$ 0.032 & 0.337 $\pm$ 0.003 & 0.282 $\pm$ 0.005 \\
\rowcolor{gray!03}
Qwen3.5-4B & Thinking & 37.4 $\pm$ 0.8 & 0.425 $\pm$ 0.026 & 0.363 $\pm$ 0.007 & 0.301 $\pm$ 0.005 \\
Qwen3.5-9B & Non-Thinking & 40.7 $\pm$ 0.7 & 0.295 $\pm$ 0.008 & 0.396 $\pm$ 0.007 & 0.298 $\pm$ 0.008 \\
\rowcolor{gray!03}
Qwen3.5-9B & Thinking & 36.4 $\pm$ 0.6 & 0.379 $\pm$ 0.157 & 0.348 $\pm$ 0.006 & 0.206 $\pm$ 0.013 \\
Qwen3.5-27B & Non-Thinking & 44.5 $\pm$ 0.4 & 0.473 $\pm$ 0.009 & 0.437 $\pm$ 0.004 & 0.382 $\pm$ 0.002 \\
\rowcolor{gray!03}
Qwen3.5-27B & Thinking & 50.8 $\pm$ 0.5 & 0.577 $\pm$ 0.016 & 0.500 $\pm$ 0.005 & 0.465 $\pm$ 0.009 \\
Qwen3.5-35B-A3B & Non-Thinking & 51.2 $\pm$ 1.4 & 0.601 $\pm$ 0.026 & 0.508 $\pm$ 0.014 & 0.442 $\pm$ 0.012 \\
\rowcolor{gray!03}
Qwen3.5-35B-A3B & Thinking & 48.3 $\pm$ 3.0 & 0.595 $\pm$ 0.045 & 0.473 $\pm$ 0.030 & 0.430 $\pm$ 0.040 \\
Gemma4-E2B & Non-Thinking & 34.7 $\pm$ 0.0 & 0.505 $\pm$ 0.000 & 0.331 $\pm$ 0.000 & 0.184 $\pm$ 0.000 \\
\rowcolor{gray!03}
Gemma4-E2B & Thinking & 38.4 $\pm$ 1.3 & 0.376 $\pm$ 0.017 & 0.379 $\pm$ 0.013 & 0.368 $\pm$ 0.015 \\
Gemma4-E4B & Non-Thinking & \textbf{61.9 $\pm$ 0.2} & 0.634 $\pm$ 0.002 & 0.619 $\pm$ 0.002 & 0.612 $\pm$ 0.002 \\
\rowcolor{gray!03}
Gemma4-E4B & Thinking & \textbf{62.1 $\pm$ 0.8} & 0.634 $\pm$ 0.009 & 0.625 $\pm$ 0.008 & 0.620 $\pm$ 0.008 \\
Gemma4-26B-A4B & Non-Thinking & 45.7 $\pm$ 0.0 & 0.528 $\pm$ 0.004 & 0.462 $\pm$ 0.000 & 0.442 $\pm$ 0.001 \\
\rowcolor{gray!03}
Gemma4-26B-A4B & Thinking & 44.9 $\pm$ 1.1 & 0.577 $\pm$ 0.022 & 0.456 $\pm$ 0.011 & 0.398 $\pm$ 0.016 \\
Gemma4-31B & Non-Thinking & 52.0 $\pm$ 0.0 & 0.578 $\pm$ 0.000 & 0.525 $\pm$ 0.000 & 0.517 $\pm$ 0.000 \\
\rowcolor{gray!03}
Gemma4-31B & Thinking & 52.3 $\pm$ 1.5 & 0.532 $\pm$ 0.077 & 0.502 $\pm$ 0.071 & 0.495 $\pm$ 0.069 \\
GLM-4.6V-Flash & Non-Thinking & 48.7 $\pm$ 0.0 & 0.548 $\pm$ 0.000 & 0.486 $\pm$ 0.000 & 0.469 $\pm$ 0.000 \\
\rowcolor{gray!03}
GLM-4.6V-Flash & Thinking & 56.7 $\pm$ 2.0 & 0.335 $\pm$ 0.071 & 0.323 $\pm$ 0.070 & 0.326 $\pm$ 0.071 \\
Kimi-VL-A3B & Non-Thinking & 35.7 $\pm$ 0.3 & 0.554 $\pm$ 0.005 & 0.360 $\pm$ 0.003 & 0.231 $\pm$ 0.003 \\
\rowcolor{gray!03}
Kimi-VL-A3B & Thinking & 35.0 $\pm$ 0.0 & 0.117 $\pm$ 0.000 & 0.333 $\pm$ 0.000 & 0.173 $\pm$ 0.000 \\
\end{xltabular}
}

\subsection{T12: Reconstruct}
\label{sec:app_metrics_t12}
{\small
\setlength{\tabcolsep}{5pt}
\begin{xltabular}{\linewidth}{>{\raggedright\arraybackslash}X lcccc}
\toprule
Model & Mode & Acc (\%) & Prec & Recall & F1 \\
\midrule
\endfirsthead
\multicolumn{6}{l}{\textit{Continued from previous page}}\\
\toprule
Model & Mode & Acc (\%) & Prec & Recall & F1 \\
\midrule
\endhead
\midrule \multicolumn{6}{r}{\textit{Continued on next page}}\\
\endfoot
\bottomrule
\addlinespace[8pt]
\caption{Full per-model metrics for \ttwelvelong{} across visual variants. Each cell reports the 5-run mean $\pm$ std for one visual-condition variant $\times$ model $\times$ inference mode. Frontier rows appear only for the default visual variant in Thinking mode and report a single run, so no std is shown. Accuracy is reported in percent. \textbf{Boldface} marks the best mean accuracy within each variant/mode block.}\label{tab:app_metrics_t12}\\
\endlastfoot
\rowcolor{gray!06}\multicolumn{6}{l}{\textit{Uncolored}}\\
Qwen3.5-0.8B & Non-Thinking & 51.0 $\pm$ 0.0 & 0.255 $\pm$ 0.000 & 0.500 $\pm$ 0.000 & 0.338 $\pm$ 0.000 \\
\rowcolor{gray!03}
Qwen3.5-0.8B & Thinking & 50.6 $\pm$ 1.1 & 0.437 $\pm$ 0.210 & 0.496 $\pm$ 0.011 & 0.346 $\pm$ 0.013 \\
Qwen3.5-2B & Non-Thinking & 51.0 $\pm$ 0.0 & 0.255 $\pm$ 0.000 & 0.500 $\pm$ 0.000 & 0.338 $\pm$ 0.000 \\
\rowcolor{gray!03}
Qwen3.5-2B & Thinking & 51.0 $\pm$ 0.0 & 0.305 $\pm$ 0.112 & 0.500 $\pm$ 0.000 & 0.341 $\pm$ 0.008 \\
Qwen3.5-4B & Non-Thinking & 50.6 $\pm$ 1.7 & 0.488 $\pm$ 0.063 & 0.498 $\pm$ 0.017 & 0.391 $\pm$ 0.024 \\
\rowcolor{gray!03}
Qwen3.5-4B & Thinking & 50.4 $\pm$ 2.3 & 0.491 $\pm$ 0.049 & 0.496 $\pm$ 0.023 & 0.416 $\pm$ 0.030 \\
Qwen3.5-9B & Non-Thinking & 49.0 $\pm$ 0.7 & 0.492 $\pm$ 0.008 & 0.493 $\pm$ 0.007 & 0.478 $\pm$ 0.009 \\
\rowcolor{gray!03}
Qwen3.5-9B & Thinking & 49.6 $\pm$ 2.6 & 0.503 $\pm$ 0.047 & 0.503 $\pm$ 0.026 & 0.432 $\pm$ 0.038 \\
Qwen3.5-27B & Non-Thinking & 57.2 $\pm$ 1.5 & 0.574 $\pm$ 0.015 & 0.573 $\pm$ 0.015 & 0.571 $\pm$ 0.015 \\
\rowcolor{gray!03}
Qwen3.5-27B & Thinking & 57.6 $\pm$ 2.1 & 0.604 $\pm$ 0.028 & 0.581 $\pm$ 0.021 & 0.553 $\pm$ 0.021 \\
Qwen3.5-35B-A3B & Non-Thinking & 54.8 $\pm$ 0.8 & 0.584 $\pm$ 0.015 & 0.541 $\pm$ 0.008 & 0.477 $\pm$ 0.012 \\
\rowcolor{gray!03}
Qwen3.5-35B-A3B & Thinking & 53.6 $\pm$ 1.3 & 0.545 $\pm$ 0.021 & 0.530 $\pm$ 0.013 & 0.493 $\pm$ 0.012 \\
Gemma4-E2B & Non-Thinking & 52.0 $\pm$ 0.0 & 0.758 $\pm$ 0.000 & 0.510 $\pm$ 0.000 & 0.360 $\pm$ 0.000 \\
\rowcolor{gray!03}
Gemma4-E2B & Thinking & 49.0 $\pm$ 4.4 & 0.491 $\pm$ 0.046 & 0.491 $\pm$ 0.044 & 0.488 $\pm$ 0.045 \\
Gemma4-E4B & Non-Thinking & 48.0 $\pm$ 0.0 & 0.409 $\pm$ 0.000 & 0.489 $\pm$ 0.000 & 0.340 $\pm$ 0.000 \\
\rowcolor{gray!03}
Gemma4-E4B & Thinking & 47.0 $\pm$ 1.0 & 0.384 $\pm$ 0.054 & 0.479 $\pm$ 0.010 & 0.341 $\pm$ 0.013 \\
Gemma4-26B-A4B & Non-Thinking & 54.6 $\pm$ 1.1 & 0.577 $\pm$ 0.016 & 0.552 $\pm$ 0.011 & 0.508 $\pm$ 0.014 \\
\rowcolor{gray!03}
Gemma4-26B-A4B & Thinking & 48.2 $\pm$ 1.6 & 0.480 $\pm$ 0.031 & 0.488 $\pm$ 0.017 & 0.425 $\pm$ 0.020 \\
Gemma4-31B & Non-Thinking & 54.0 $\pm$ 0.0 & 0.560 $\pm$ 0.000 & 0.545 $\pm$ 0.000 & 0.512 $\pm$ 0.000 \\
\rowcolor{gray!03}
Gemma4-31B & Thinking & 50.8 $\pm$ 1.5 & 0.532 $\pm$ 0.028 & 0.515 $\pm$ 0.015 & 0.436 $\pm$ 0.021 \\
GLM-4.6V-Flash & Non-Thinking & 48.0 $\pm$ 0.0 & 0.461 $\pm$ 0.000 & 0.474 $\pm$ 0.000 & 0.429 $\pm$ 0.000 \\
\rowcolor{gray!03}
GLM-4.6V-Flash & Thinking & 63.2 $\pm$ 1.9 & 0.654 $\pm$ 0.025 & 0.628 $\pm$ 0.019 & 0.614 $\pm$ 0.020 \\
Kimi-VL-A3B & Non-Thinking & \textbf{60.4 $\pm$ 0.9} & 0.653 $\pm$ 0.017 & 0.598 $\pm$ 0.009 & 0.562 $\pm$ 0.010 \\
\rowcolor{gray!03}
Kimi-VL-A3B & Thinking & 53.4 $\pm$ 0.5 & 0.553 $\pm$ 0.010 & 0.527 $\pm$ 0.006 & 0.464 $\pm$ 0.008 \\
\rowcolor{blue!03}
GPT-5 (High) & Thinking & 61.0 & 0.663 & 0.663 & 0.663 \\
\rowcolor{blue!03}
Grok-4.20 (High) & Thinking & 69.0 & 0.700 & 0.692 & 0.687 \\
\rowcolor{blue!03}
Gemini-3.1-Pro-Preview (High) & Thinking & 57.0 & 0.575 & 0.574 & 0.572 \\
\rowcolor{blue!03}
Gemini-3-Flash-Preview (High) & Thinking & \textbf{71.0} & 0.711 & 0.709 & 0.709 \\
\midrule
\rowcolor{gray!06}\multicolumn{6}{l}{\textit{Uncolored + Axes}}\\
Qwen3.5-0.8B & Non-Thinking & 51.0 $\pm$ 0.0 & 0.255 $\pm$ 0.000 & 0.500 $\pm$ 0.000 & 0.338 $\pm$ 0.000 \\
\rowcolor{gray!03}
Qwen3.5-0.8B & Thinking & 51.2 $\pm$ 1.1 & 0.539 $\pm$ 0.129 & 0.502 $\pm$ 0.011 & 0.359 $\pm$ 0.012 \\
Qwen3.5-2B & Non-Thinking & 51.0 $\pm$ 0.0 & 0.255 $\pm$ 0.000 & 0.500 $\pm$ 0.000 & 0.338 $\pm$ 0.000 \\
\rowcolor{gray!03}
Qwen3.5-2B & Thinking & 50.8 $\pm$ 0.8 & 0.355 $\pm$ 0.225 & 0.498 $\pm$ 0.008 & 0.340 $\pm$ 0.011 \\
Qwen3.5-4B & Non-Thinking & 49.2 $\pm$ 0.4 & 0.251 $\pm$ 0.001 & 0.482 $\pm$ 0.004 & 0.330 $\pm$ 0.002 \\
\rowcolor{gray!03}
Qwen3.5-4B & Thinking & 50.4 $\pm$ 1.8 & 0.470 $\pm$ 0.222 & 0.495 $\pm$ 0.018 & 0.352 $\pm$ 0.024 \\
Qwen3.5-9B & Non-Thinking & 51.6 $\pm$ 1.1 & 0.542 $\pm$ 0.021 & 0.523 $\pm$ 0.011 & 0.458 $\pm$ 0.014 \\
\rowcolor{gray!03}
Qwen3.5-9B & Thinking & 49.2 $\pm$ 1.8 & 0.497 $\pm$ 0.051 & 0.500 $\pm$ 0.018 & 0.404 $\pm$ 0.027 \\
Qwen3.5-27B & Non-Thinking & \textbf{58.6 $\pm$ 1.1} & 0.596 $\pm$ 0.012 & 0.589 $\pm$ 0.011 & 0.579 $\pm$ 0.012 \\
\rowcolor{gray!03}
Qwen3.5-27B & Thinking & 54.0 $\pm$ 1.9 & 0.588 $\pm$ 0.034 & 0.547 $\pm$ 0.019 & 0.483 $\pm$ 0.025 \\
Qwen3.5-35B-A3B & Non-Thinking & 53.2 $\pm$ 1.1 & 0.535 $\pm$ 0.014 & 0.527 $\pm$ 0.011 & 0.503 $\pm$ 0.010 \\
\rowcolor{gray!03}
Qwen3.5-35B-A3B & Thinking & 54.4 $\pm$ 1.8 & 0.544 $\pm$ 0.019 & 0.542 $\pm$ 0.019 & 0.537 $\pm$ 0.020 \\
Gemma4-E2B & Non-Thinking & 52.0 $\pm$ 0.0 & 0.536 $\pm$ 0.000 & 0.512 $\pm$ 0.000 & 0.417 $\pm$ 0.000 \\
\rowcolor{gray!03}
Gemma4-E2B & Thinking & 45.0 $\pm$ 3.0 & 0.450 $\pm$ 0.030 & 0.451 $\pm$ 0.030 & 0.448 $\pm$ 0.030 \\
Gemma4-E4B & Non-Thinking & 50.0 $\pm$ 0.0 & 0.533 $\pm$ 0.000 & 0.509 $\pm$ 0.000 & 0.393 $\pm$ 0.000 \\
\rowcolor{gray!03}
Gemma4-E4B & Thinking & 47.0 $\pm$ 1.9 & 0.402 $\pm$ 0.082 & 0.479 $\pm$ 0.018 & 0.355 $\pm$ 0.032 \\
Gemma4-26B-A4B & Non-Thinking & 51.6 $\pm$ 1.7 & 0.518 $\pm$ 0.017 & 0.518 $\pm$ 0.017 & 0.514 $\pm$ 0.017 \\
\rowcolor{gray!03}
Gemma4-26B-A4B & Thinking & 52.4 $\pm$ 4.0 & 0.549 $\pm$ 0.070 & 0.531 $\pm$ 0.040 & 0.467 $\pm$ 0.056 \\
Gemma4-31B & Non-Thinking & 48.0 $\pm$ 0.0 & 0.475 $\pm$ 0.000 & 0.487 $\pm$ 0.000 & 0.412 $\pm$ 0.000 \\
\rowcolor{gray!03}
Gemma4-31B & Thinking & 50.2 $\pm$ 2.0 & 0.518 $\pm$ 0.048 & 0.510 $\pm$ 0.020 & 0.422 $\pm$ 0.033 \\
GLM-4.6V-Flash & Non-Thinking & 50.0 $\pm$ 0.0 & 0.498 $\pm$ 0.000 & 0.498 $\pm$ 0.000 & 0.495 $\pm$ 0.000 \\
\rowcolor{gray!03}
GLM-4.6V-Flash & Thinking & \textbf{65.4 $\pm$ 2.7} & 0.662 $\pm$ 0.034 & 0.652 $\pm$ 0.027 & 0.648 $\pm$ 0.026 \\
Kimi-VL-A3B & Non-Thinking & \textbf{58.6 $\pm$ 2.1} & 0.587 $\pm$ 0.021 & 0.584 $\pm$ 0.021 & 0.582 $\pm$ 0.021 \\
\rowcolor{gray!03}
Kimi-VL-A3B & Thinking & 55.0 $\pm$ 1.0 & 0.576 $\pm$ 0.018 & 0.543 $\pm$ 0.010 & 0.492 $\pm$ 0.011 \\
\midrule
\rowcolor{gray!06}\multicolumn{6}{l}{\textit{Colored}}\\
Qwen3.5-0.8B & Non-Thinking & 51.0 $\pm$ 0.0 & 0.255 $\pm$ 0.000 & 0.500 $\pm$ 0.000 & 0.338 $\pm$ 0.000 \\
\rowcolor{gray!03}
Qwen3.5-0.8B & Thinking & 51.8 $\pm$ 1.5 & 0.574 $\pm$ 0.118 & 0.509 $\pm$ 0.015 & 0.378 $\pm$ 0.019 \\
Qwen3.5-2B & Non-Thinking & 50.6 $\pm$ 0.5 & 0.254 $\pm$ 0.001 & 0.496 $\pm$ 0.005 & 0.336 $\pm$ 0.002 \\
\rowcolor{gray!03}
Qwen3.5-2B & Thinking & 51.0 $\pm$ 0.0 & 0.255 $\pm$ 0.000 & 0.500 $\pm$ 0.000 & 0.338 $\pm$ 0.000 \\
Qwen3.5-4B & Non-Thinking & 56.0 $\pm$ 0.7 & 0.607 $\pm$ 0.014 & 0.553 $\pm$ 0.007 & 0.493 $\pm$ 0.013 \\
\rowcolor{gray!03}
Qwen3.5-4B & Thinking & 59.6 $\pm$ 4.3 & 0.621 $\pm$ 0.054 & 0.591 $\pm$ 0.044 & 0.565 $\pm$ 0.052 \\
Qwen3.5-9B & Non-Thinking & 67.6 $\pm$ 0.5 & 0.676 $\pm$ 0.005 & 0.676 $\pm$ 0.006 & 0.676 $\pm$ 0.006 \\
\rowcolor{gray!03}
Qwen3.5-9B & Thinking & 61.4 $\pm$ 3.2 & 0.646 $\pm$ 0.039 & 0.618 $\pm$ 0.032 & 0.596 $\pm$ 0.035 \\
Qwen3.5-27B & Non-Thinking & 67.4 $\pm$ 0.5 & 0.675 $\pm$ 0.006 & 0.674 $\pm$ 0.005 & 0.674 $\pm$ 0.005 \\
\rowcolor{gray!03}
Qwen3.5-27B & Thinking & 67.4 $\pm$ 0.5 & 0.686 $\pm$ 0.005 & 0.676 $\pm$ 0.005 & 0.670 $\pm$ 0.006 \\
Qwen3.5-35B-A3B & Non-Thinking & 63.6 $\pm$ 1.7 & 0.652 $\pm$ 0.017 & 0.632 $\pm$ 0.017 & 0.622 $\pm$ 0.019 \\
\rowcolor{gray!03}
Qwen3.5-35B-A3B & Thinking & 66.2 $\pm$ 1.6 & 0.664 $\pm$ 0.016 & 0.661 $\pm$ 0.017 & 0.660 $\pm$ 0.017 \\
Gemma4-E2B & Non-Thinking & 48.0 $\pm$ 0.0 & 0.347 $\pm$ 0.000 & 0.471 $\pm$ 0.000 & 0.340 $\pm$ 0.000 \\
\rowcolor{gray!03}
Gemma4-E2B & Thinking & 46.8 $\pm$ 6.2 & 0.468 $\pm$ 0.062 & 0.468 $\pm$ 0.062 & 0.468 $\pm$ 0.062 \\
Gemma4-E4B & Non-Thinking & 47.0 $\pm$ 0.0 & 0.240 $\pm$ 0.000 & 0.480 $\pm$ 0.000 & 0.320 $\pm$ 0.000 \\
\rowcolor{gray!03}
Gemma4-E4B & Thinking & 48.4 $\pm$ 0.5 & 0.457 $\pm$ 0.038 & 0.493 $\pm$ 0.006 & 0.348 $\pm$ 0.007 \\
Gemma4-26B-A4B & Non-Thinking & 60.0 $\pm$ 0.0 & 0.621 $\pm$ 0.000 & 0.604 $\pm$ 0.000 & 0.587 $\pm$ 0.000 \\
\rowcolor{gray!03}
Gemma4-26B-A4B & Thinking & 56.6 $\pm$ 2.1 & 0.607 $\pm$ 0.025 & 0.572 $\pm$ 0.020 & 0.529 $\pm$ 0.026 \\
Gemma4-31B & Non-Thinking & \textbf{69.0 $\pm$ 0.0} & 0.697 $\pm$ 0.000 & 0.692 $\pm$ 0.000 & 0.688 $\pm$ 0.000 \\
\rowcolor{gray!03}
Gemma4-31B & Thinking & 68.6 $\pm$ 0.5 & 0.695 $\pm$ 0.007 & 0.688 $\pm$ 0.006 & 0.684 $\pm$ 0.005 \\
GLM-4.6V-Flash & Non-Thinking & 67.0 $\pm$ 0.0 & 0.675 $\pm$ 0.000 & 0.668 $\pm$ 0.000 & 0.666 $\pm$ 0.000 \\
\rowcolor{gray!03}
GLM-4.6V-Flash & Thinking & \textbf{92.2 $\pm$ 0.8} & 0.926 $\pm$ 0.005 & 0.921 $\pm$ 0.009 & 0.922 $\pm$ 0.009 \\
Kimi-VL-A3B & Non-Thinking & 65.4 $\pm$ 1.8 & 0.661 $\pm$ 0.016 & 0.652 $\pm$ 0.019 & 0.648 $\pm$ 0.021 \\
\rowcolor{gray!03}
Kimi-VL-A3B & Thinking & 65.0 $\pm$ 0.0 & 0.650 $\pm$ 0.000 & 0.650 $\pm$ 0.000 & 0.650 $\pm$ 0.000 \\
\end{xltabular}
}

\subsection{T13: Mixture-of-Shapes}
\label{sec:app_metrics_t13}
{\small
\setlength{\tabcolsep}{5pt}
\begin{xltabular}{\linewidth}{>{\raggedright\arraybackslash}X lcccc}
\toprule
Model & Mode & Acc (\%) & Prec & Recall & F1 \\
\midrule
\endfirsthead
\multicolumn{6}{l}{\textit{Continued from previous page}}\\
\toprule
Model & Mode & Acc (\%) & Prec & Recall & F1 \\
\midrule
\endhead
\midrule \multicolumn{6}{r}{\textit{Continued on next page}}\\
\endfoot
\bottomrule
\addlinespace[8pt]
\caption{Full per-model metrics for \tthirteenlong{} across visual variants. Each cell reports the 5-run mean $\pm$ std for one visual-condition variant $\times$ model $\times$ inference mode. Frontier rows appear only for the default visual variant in Thinking mode and report a single run, so no std is shown. Accuracy is reported in percent. \textbf{Boldface} marks the best mean accuracy within each variant/mode block.}\label{tab:app_metrics_t13}\\
\endlastfoot
\rowcolor{gray!06}\multicolumn{6}{l}{\textit{Negative / Uncolored}}\\
Qwen3.5-0.8B & Non-Thinking & 37.3 $\pm$ 0.0 & 0.124 $\pm$ 0.000 & 0.333 $\pm$ 0.000 & 0.181 $\pm$ 0.000 \\
\rowcolor{gray!03}
Qwen3.5-0.8B & Thinking & 37.0 $\pm$ 1.5 & 0.343 $\pm$ 0.023 & 0.343 $\pm$ 0.015 & 0.269 $\pm$ 0.015 \\
Qwen3.5-2B & Non-Thinking & 37.3 $\pm$ 0.0 & 0.124 $\pm$ 0.000 & 0.333 $\pm$ 0.000 & 0.181 $\pm$ 0.000 \\
\rowcolor{gray!03}
Qwen3.5-2B & Thinking & 36.7 $\pm$ 1.0 & 0.381 $\pm$ 0.112 & 0.335 $\pm$ 0.009 & 0.231 $\pm$ 0.009 \\
Qwen3.5-4B & Non-Thinking & 37.3 $\pm$ 0.0 & 0.124 $\pm$ 0.000 & 0.333 $\pm$ 0.000 & 0.181 $\pm$ 0.000 \\
\rowcolor{gray!03}
Qwen3.5-4B & Thinking & 35.3 $\pm$ 1.1 & 0.278 $\pm$ 0.067 & 0.322 $\pm$ 0.010 & 0.220 $\pm$ 0.007 \\
Qwen3.5-9B & Non-Thinking & 37.3 $\pm$ 0.0 & 0.124 $\pm$ 0.000 & 0.333 $\pm$ 0.000 & 0.181 $\pm$ 0.000 \\
\rowcolor{gray!03}
Qwen3.5-9B & Thinking & 36.2 $\pm$ 1.6 & 0.274 $\pm$ 0.138 & 0.326 $\pm$ 0.016 & 0.203 $\pm$ 0.028 \\
Qwen3.5-27B & Non-Thinking & 36.2 $\pm$ 0.5 & 0.227 $\pm$ 0.016 & 0.325 $\pm$ 0.005 & 0.208 $\pm$ 0.007 \\
\rowcolor{gray!03}
Qwen3.5-27B & Thinking & 38.5 $\pm$ 0.5 & 0.395 $\pm$ 0.041 & 0.351 $\pm$ 0.006 & 0.263 $\pm$ 0.013 \\
Qwen3.5-35B-A3B & Non-Thinking & 37.6 $\pm$ 0.3 & 0.391 $\pm$ 0.092 & 0.336 $\pm$ 0.003 & 0.189 $\pm$ 0.003 \\
\rowcolor{gray!03}
Qwen3.5-35B-A3B & Thinking & 37.9 $\pm$ 1.9 & 0.394 $\pm$ 0.070 & 0.345 $\pm$ 0.019 & 0.240 $\pm$ 0.025 \\
Gemma4-E2B & Non-Thinking & 36.9 $\pm$ 0.2 & 0.124 $\pm$ 0.000 & 0.329 $\pm$ 0.002 & 0.180 $\pm$ 0.001 \\
\rowcolor{gray!03}
Gemma4-E2B & Thinking & 33.5 $\pm$ 2.4 & 0.339 $\pm$ 0.031 & 0.325 $\pm$ 0.024 & 0.311 $\pm$ 0.021 \\
Gemma4-E4B & Non-Thinking & 35.3 $\pm$ 0.6 & 0.360 $\pm$ 0.009 & 0.347 $\pm$ 0.006 & 0.338 $\pm$ 0.008 \\
\rowcolor{gray!03}
Gemma4-E4B & Thinking & 34.9 $\pm$ 2.2 & 0.359 $\pm$ 0.025 & 0.360 $\pm$ 0.023 & 0.342 $\pm$ 0.022 \\
Gemma4-26B-A4B & Non-Thinking & 36.3 $\pm$ 0.0 & 0.242 $\pm$ 0.000 & 0.337 $\pm$ 0.000 & 0.280 $\pm$ 0.000 \\
\rowcolor{gray!03}
Gemma4-26B-A4B & Thinking & 34.3 $\pm$ 0.9 & 0.318 $\pm$ 0.015 & 0.326 $\pm$ 0.009 & 0.310 $\pm$ 0.010 \\
Gemma4-31B & Non-Thinking & \textbf{42.3 $\pm$ 0.2} & 0.427 $\pm$ 0.007 & 0.412 $\pm$ 0.003 & 0.365 $\pm$ 0.004 \\
\rowcolor{gray!03}
Gemma4-31B & Thinking & 40.3 $\pm$ 0.9 & 0.393 $\pm$ 0.012 & 0.394 $\pm$ 0.009 & 0.389 $\pm$ 0.009 \\
GLM-4.6V-Flash & Non-Thinking & 37.3 $\pm$ 0.0 & 0.124 $\pm$ 0.000 & 0.333 $\pm$ 0.000 & 0.181 $\pm$ 0.000 \\
\rowcolor{gray!03}
GLM-4.6V-Flash & Thinking & 37.7 $\pm$ 0.5 & 0.354 $\pm$ 0.074 & 0.340 $\pm$ 0.005 & 0.236 $\pm$ 0.013 \\
Kimi-VL-A3B & Non-Thinking & 37.1 $\pm$ 0.2 & 0.124 $\pm$ 0.001 & 0.332 $\pm$ 0.002 & 0.181 $\pm$ 0.001 \\
\rowcolor{gray!03}
Kimi-VL-A3B & Thinking & 37.3 $\pm$ 0.0 & 0.125 $\pm$ 0.000 & 0.333 $\pm$ 0.000 & 0.181 $\pm$ 0.000 \\
\rowcolor{blue!03}
GPT-5 (High) & Thinking & 40.3 & 0.405 & 0.404 & 0.404 \\
\rowcolor{blue!03}
Grok-4.20 (High) & Thinking & 41.0 & 0.409 & 0.405 & 0.405 \\
\rowcolor{blue!03}
Gemini-3.1-Pro-Preview (High) & Thinking & 42.0 & 0.424 & 0.416 & 0.415 \\
\rowcolor{blue!03}
Gemini-3-Flash-Preview (High) & Thinking & \textbf{43.7} & 0.431 & 0.431 & 0.431 \\
\midrule
\rowcolor{gray!06}\multicolumn{6}{l}{\textit{Negative / Colored}}\\
Qwen3.5-0.8B & Non-Thinking & 37.3 $\pm$ 0.0 & 0.124 $\pm$ 0.000 & 0.333 $\pm$ 0.000 & 0.181 $\pm$ 0.000 \\
\rowcolor{gray!03}
Qwen3.5-0.8B & Thinking & 35.2 $\pm$ 1.7 & 0.323 $\pm$ 0.038 & 0.331 $\pm$ 0.016 & 0.275 $\pm$ 0.014 \\
Qwen3.5-2B & Non-Thinking & 37.3 $\pm$ 0.0 & 0.124 $\pm$ 0.000 & 0.333 $\pm$ 0.000 & 0.181 $\pm$ 0.000 \\
\rowcolor{gray!03}
Qwen3.5-2B & Thinking & 34.6 $\pm$ 1.2 & 0.287 $\pm$ 0.020 & 0.317 $\pm$ 0.012 & 0.230 $\pm$ 0.010 \\
Qwen3.5-4B & Non-Thinking & 37.3 $\pm$ 0.0 & 0.124 $\pm$ 0.000 & 0.333 $\pm$ 0.000 & 0.181 $\pm$ 0.000 \\
\rowcolor{gray!03}
Qwen3.5-4B & Thinking & 36.5 $\pm$ 1.2 & 0.328 $\pm$ 0.050 & 0.336 $\pm$ 0.013 & 0.250 $\pm$ 0.021 \\
Qwen3.5-9B & Non-Thinking & 37.3 $\pm$ 0.0 & 0.124 $\pm$ 0.000 & 0.333 $\pm$ 0.000 & 0.181 $\pm$ 0.000 \\
\rowcolor{gray!03}
Qwen3.5-9B & Thinking & 37.9 $\pm$ 1.2 & 0.367 $\pm$ 0.033 & 0.344 $\pm$ 0.012 & 0.238 $\pm$ 0.015 \\
Qwen3.5-27B & Non-Thinking & 37.9 $\pm$ 0.3 & 0.318 $\pm$ 0.013 & 0.340 $\pm$ 0.003 & 0.213 $\pm$ 0.005 \\
\rowcolor{gray!03}
Qwen3.5-27B & Thinking & 38.7 $\pm$ 1.2 & 0.426 $\pm$ 0.043 & 0.357 $\pm$ 0.012 & 0.289 $\pm$ 0.018 \\
Qwen3.5-35B-A3B & Non-Thinking & 37.5 $\pm$ 0.2 & 0.325 $\pm$ 0.183 & 0.335 $\pm$ 0.002 & 0.185 $\pm$ 0.004 \\
\rowcolor{gray!03}
Qwen3.5-35B-A3B & Thinking & 37.3 $\pm$ 1.1 & 0.346 $\pm$ 0.027 & 0.345 $\pm$ 0.012 & 0.270 $\pm$ 0.018 \\
Gemma4-E2B & Non-Thinking & 35.3 $\pm$ 0.1 & 0.185 $\pm$ 0.001 & 0.320 $\pm$ 0.001 & 0.209 $\pm$ 0.001 \\
\rowcolor{gray!03}
Gemma4-E2B & Thinking & 35.5 $\pm$ 1.1 & 0.341 $\pm$ 0.011 & 0.340 $\pm$ 0.010 & 0.315 $\pm$ 0.010 \\
Gemma4-E4B & Non-Thinking & 37.7 $\pm$ 0.4 & 0.387 $\pm$ 0.004 & 0.368 $\pm$ 0.005 & 0.350 $\pm$ 0.005 \\
\rowcolor{gray!03}
Gemma4-E4B & Thinking & 30.5 $\pm$ 1.3 & 0.306 $\pm$ 0.014 & 0.314 $\pm$ 0.013 & 0.292 $\pm$ 0.014 \\
Gemma4-26B-A4B & Non-Thinking & 38.5 $\pm$ 0.4 & 0.362 $\pm$ 0.015 & 0.358 $\pm$ 0.004 & 0.311 $\pm$ 0.005 \\
\rowcolor{gray!03}
Gemma4-26B-A4B & Thinking & 37.4 $\pm$ 2.3 & 0.365 $\pm$ 0.025 & 0.357 $\pm$ 0.023 & 0.343 $\pm$ 0.024 \\
Gemma4-31B & Non-Thinking & \textbf{40.8 $\pm$ 0.2} & 0.440 $\pm$ 0.004 & 0.405 $\pm$ 0.002 & 0.357 $\pm$ 0.003 \\
\rowcolor{gray!03}
Gemma4-31B & Thinking & \textbf{44.1 $\pm$ 1.1} & 0.445 $\pm$ 0.013 & 0.436 $\pm$ 0.012 & 0.437 $\pm$ 0.012 \\
GLM-4.6V-Flash & Non-Thinking & 37.3 $\pm$ 0.0 & 0.124 $\pm$ 0.000 & 0.333 $\pm$ 0.000 & 0.181 $\pm$ 0.000 \\
\rowcolor{gray!03}
GLM-4.6V-Flash & Thinking & 36.9 $\pm$ 1.4 & 0.330 $\pm$ 0.092 & 0.333 $\pm$ 0.013 & 0.227 $\pm$ 0.016 \\
Kimi-VL-A3B & Non-Thinking & 37.0 $\pm$ 0.0 & 0.124 $\pm$ 0.000 & 0.330 $\pm$ 0.000 & 0.180 $\pm$ 0.000 \\
\rowcolor{gray!03}
Kimi-VL-A3B & Thinking & 37.5 $\pm$ 0.2 & 0.247 $\pm$ 0.073 & 0.335 $\pm$ 0.002 & 0.187 $\pm$ 0.003 \\
\midrule
\rowcolor{gray!06}\multicolumn{6}{l}{\textit{Positive / Uncolored}}\\
Qwen3.5-0.8B & Non-Thinking & 35.0 $\pm$ 0.0 & 0.117 $\pm$ 0.000 & 0.333 $\pm$ 0.000 & 0.173 $\pm$ 0.000 \\
\rowcolor{gray!03}
Qwen3.5-0.8B & Thinking & 34.7 $\pm$ 2.7 & 0.329 $\pm$ 0.057 & 0.337 $\pm$ 0.027 & 0.291 $\pm$ 0.030 \\
Qwen3.5-2B & Non-Thinking & 35.0 $\pm$ 0.0 & 0.117 $\pm$ 0.000 & 0.333 $\pm$ 0.000 & 0.173 $\pm$ 0.000 \\
\rowcolor{gray!03}
Qwen3.5-2B & Thinking & 34.0 $\pm$ 1.4 & 0.351 $\pm$ 0.043 & 0.328 $\pm$ 0.013 & 0.260 $\pm$ 0.016 \\
Qwen3.5-4B & Non-Thinking & 33.3 $\pm$ 0.2 & 0.392 $\pm$ 0.027 & 0.340 $\pm$ 0.002 & 0.278 $\pm$ 0.006 \\
\rowcolor{gray!03}
Qwen3.5-4B & Thinking & 33.5 $\pm$ 2.6 & 0.334 $\pm$ 0.034 & 0.331 $\pm$ 0.026 & 0.316 $\pm$ 0.027 \\
Qwen3.5-9B & Non-Thinking & 34.3 $\pm$ 1.0 & 0.430 $\pm$ 0.181 & 0.349 $\pm$ 0.010 & 0.263 $\pm$ 0.010 \\
\rowcolor{gray!03}
Qwen3.5-9B & Thinking & 34.6 $\pm$ 1.6 & 0.333 $\pm$ 0.024 & 0.335 $\pm$ 0.015 & 0.278 $\pm$ 0.019 \\
Qwen3.5-27B & Non-Thinking & 34.3 $\pm$ 0.6 & 0.324 $\pm$ 0.012 & 0.330 $\pm$ 0.006 & 0.230 $\pm$ 0.005 \\
\rowcolor{gray!03}
Qwen3.5-27B & Thinking & 37.9 $\pm$ 2.3 & 0.411 $\pm$ 0.039 & 0.379 $\pm$ 0.024 & 0.352 $\pm$ 0.024 \\
Qwen3.5-35B-A3B & Non-Thinking & 34.9 $\pm$ 1.7 & 0.381 $\pm$ 0.045 & 0.360 $\pm$ 0.017 & 0.288 $\pm$ 0.022 \\
\rowcolor{gray!03}
Qwen3.5-35B-A3B & Thinking & 34.3 $\pm$ 1.8 & 0.366 $\pm$ 0.036 & 0.336 $\pm$ 0.018 & 0.305 $\pm$ 0.023 \\
Gemma4-E2B & Non-Thinking & 34.1 $\pm$ 0.5 & 0.348 $\pm$ 0.005 & 0.342 $\pm$ 0.006 & 0.332 $\pm$ 0.004 \\
\rowcolor{gray!03}
Gemma4-E2B & Thinking & 35.6 $\pm$ 2.8 & 0.367 $\pm$ 0.033 & 0.361 $\pm$ 0.029 & 0.338 $\pm$ 0.028 \\
Gemma4-E4B & Non-Thinking & 34.4 $\pm$ 0.4 & 0.393 $\pm$ 0.003 & 0.351 $\pm$ 0.004 & 0.299 $\pm$ 0.004 \\
\rowcolor{gray!03}
Gemma4-E4B & Thinking & 34.7 $\pm$ 2.9 & 0.347 $\pm$ 0.036 & 0.353 $\pm$ 0.029 & 0.329 $\pm$ 0.028 \\
Gemma4-26B-A4B & Non-Thinking & 37.2 $\pm$ 0.2 & 0.494 $\pm$ 0.001 & 0.380 $\pm$ 0.002 & 0.312 $\pm$ 0.002 \\
\rowcolor{gray!03}
Gemma4-26B-A4B & Thinking & 35.6 $\pm$ 0.5 & 0.363 $\pm$ 0.009 & 0.358 $\pm$ 0.005 & 0.335 $\pm$ 0.006 \\
Gemma4-31B & Non-Thinking & \textbf{39.3 $\pm$ 0.0} & 0.394 $\pm$ 0.000 & 0.395 $\pm$ 0.000 & 0.392 $\pm$ 0.000 \\
\rowcolor{gray!03}
Gemma4-31B & Thinking & \textbf{38.7 $\pm$ 0.5} & 0.387 $\pm$ 0.006 & 0.386 $\pm$ 0.006 & 0.385 $\pm$ 0.006 \\
GLM-4.6V-Flash & Non-Thinking & 36.7 $\pm$ 0.4 & 0.454 $\pm$ 0.003 & 0.371 $\pm$ 0.004 & 0.339 $\pm$ 0.003 \\
\rowcolor{gray!03}
GLM-4.6V-Flash & Thinking & 34.1 $\pm$ 3.0 & 0.343 $\pm$ 0.031 & 0.340 $\pm$ 0.031 & 0.340 $\pm$ 0.031 \\
Kimi-VL-A3B & Non-Thinking & 32.7 $\pm$ 0.3 & 0.282 $\pm$ 0.043 & 0.331 $\pm$ 0.003 & 0.212 $\pm$ 0.005 \\
\rowcolor{gray!03}
Kimi-VL-A3B & Thinking & 29.6 $\pm$ 0.9 & 0.378 $\pm$ 0.019 & 0.301 $\pm$ 0.009 & 0.236 $\pm$ 0.010 \\
\midrule
\rowcolor{gray!06}\multicolumn{6}{l}{\textit{Positive / Colored}}\\
Qwen3.5-0.8B & Non-Thinking & 35.0 $\pm$ 0.0 & 0.117 $\pm$ 0.000 & 0.333 $\pm$ 0.000 & 0.173 $\pm$ 0.000 \\
\rowcolor{gray!03}
Qwen3.5-0.8B & Thinking & 33.9 $\pm$ 2.0 & 0.317 $\pm$ 0.040 & 0.329 $\pm$ 0.020 & 0.285 $\pm$ 0.026 \\
Qwen3.5-2B & Non-Thinking & 35.0 $\pm$ 0.0 & 0.117 $\pm$ 0.000 & 0.333 $\pm$ 0.000 & 0.173 $\pm$ 0.000 \\
\rowcolor{gray!03}
Qwen3.5-2B & Thinking & 35.2 $\pm$ 2.9 & 0.337 $\pm$ 0.057 & 0.341 $\pm$ 0.029 & 0.287 $\pm$ 0.036 \\
Qwen3.5-4B & Non-Thinking & 35.7 $\pm$ 0.9 & 0.357 $\pm$ 0.005 & 0.361 $\pm$ 0.009 & 0.344 $\pm$ 0.008 \\
\rowcolor{gray!03}
Qwen3.5-4B & Thinking & 33.8 $\pm$ 1.6 & 0.353 $\pm$ 0.025 & 0.333 $\pm$ 0.015 & 0.312 $\pm$ 0.015 \\
Qwen3.5-9B & Non-Thinking & 34.1 $\pm$ 0.8 & 0.418 $\pm$ 0.106 & 0.349 $\pm$ 0.008 & 0.257 $\pm$ 0.012 \\
\rowcolor{gray!03}
Qwen3.5-9B & Thinking & 35.4 $\pm$ 1.6 & 0.360 $\pm$ 0.027 & 0.348 $\pm$ 0.017 & 0.333 $\pm$ 0.021 \\
Qwen3.5-27B & Non-Thinking & 39.5 $\pm$ 0.4 & 0.479 $\pm$ 0.019 & 0.384 $\pm$ 0.004 & 0.313 $\pm$ 0.005 \\
\rowcolor{gray!03}
Qwen3.5-27B & Thinking & \textbf{41.7 $\pm$ 1.8} & 0.436 $\pm$ 0.022 & 0.415 $\pm$ 0.018 & 0.405 $\pm$ 0.020 \\
Qwen3.5-35B-A3B & Non-Thinking & 33.7 $\pm$ 0.9 & 0.349 $\pm$ 0.020 & 0.346 $\pm$ 0.009 & 0.282 $\pm$ 0.010 \\
\rowcolor{gray!03}
Qwen3.5-35B-A3B & Thinking & 36.7 $\pm$ 1.6 & 0.373 $\pm$ 0.015 & 0.361 $\pm$ 0.016 & 0.340 $\pm$ 0.019 \\
Gemma4-E2B & Non-Thinking & 36.3 $\pm$ 0.2 & 0.370 $\pm$ 0.002 & 0.365 $\pm$ 0.002 & 0.362 $\pm$ 0.002 \\
\rowcolor{gray!03}
Gemma4-E2B & Thinking & 32.1 $\pm$ 3.0 & 0.340 $\pm$ 0.035 & 0.328 $\pm$ 0.030 & 0.295 $\pm$ 0.029 \\
Gemma4-E4B & Non-Thinking & 36.0 $\pm$ 0.4 & 0.428 $\pm$ 0.010 & 0.367 $\pm$ 0.004 & 0.313 $\pm$ 0.005 \\
\rowcolor{gray!03}
Gemma4-E4B & Thinking & 38.6 $\pm$ 2.3 & 0.403 $\pm$ 0.031 & 0.393 $\pm$ 0.024 & 0.366 $\pm$ 0.022 \\
Gemma4-26B-A4B & Non-Thinking & 36.9 $\pm$ 0.3 & 0.413 $\pm$ 0.002 & 0.378 $\pm$ 0.003 & 0.307 $\pm$ 0.002 \\
\rowcolor{gray!03}
Gemma4-26B-A4B & Thinking & 37.2 $\pm$ 1.2 & 0.369 $\pm$ 0.012 & 0.373 $\pm$ 0.012 & 0.339 $\pm$ 0.016 \\
Gemma4-31B & Non-Thinking & \textbf{42.0 $\pm$ 0.0} & 0.423 $\pm$ 0.000 & 0.420 $\pm$ 0.000 & 0.421 $\pm$ 0.000 \\
\rowcolor{gray!03}
Gemma4-31B & Thinking & \textbf{41.7 $\pm$ 1.4} & 0.424 $\pm$ 0.014 & 0.415 $\pm$ 0.014 & 0.417 $\pm$ 0.014 \\
GLM-4.6V-Flash & Non-Thinking & 34.9 $\pm$ 0.4 & 0.370 $\pm$ 0.006 & 0.357 $\pm$ 0.004 & 0.319 $\pm$ 0.004 \\
\rowcolor{gray!03}
GLM-4.6V-Flash & Thinking & 35.5 $\pm$ 2.4 & 0.355 $\pm$ 0.029 & 0.355 $\pm$ 0.024 & 0.349 $\pm$ 0.027 \\
Kimi-VL-A3B & Non-Thinking & 33.7 $\pm$ 0.7 & 0.222 $\pm$ 0.018 & 0.342 $\pm$ 0.007 & 0.222 $\pm$ 0.010 \\
\rowcolor{gray!03}
Kimi-VL-A3B & Thinking & 30.5 $\pm$ 0.9 & 0.341 $\pm$ 0.048 & 0.309 $\pm$ 0.009 & 0.233 $\pm$ 0.009 \\
\end{xltabular}
}

\clearpage
\section{Use of AI Assistants}
\label{sec:app_ai_assistance}
We used AI coding and writing assistants (Claude Code and ChatGPT/Codex) in three ways.\begin{itemize}[leftmargin=1.4em,nosep]
    \item For coding, they helped write and refactor the data generator, evaluation harness, probing code, analysis and plotting scripts, and the release package.
    \item For writing, they helped revise the wording and structure of drafts written by the authors and check that numbers in the text, tables, and figures were consistent.
    \item For literature search, they helped locate candidate related papers.
\end{itemize}
The authors designed the study and experiments, read every cited paper, and checked all generated code and text.
Every number in the paper was recomputed by scripts from the saved model outputs, and the authors take full responsibility for the content.
The models evaluated in this paper are research objects and were not used as assistants.

\end{document}